\documentclass{article} 
\usepackage{iclr2027_conference,times}

\usepackage[utf8]{inputenc}
\usepackage[T1]{fontenc}
\usepackage{hyperref}
\usepackage{url}
\usepackage{booktabs}
\usepackage{longtable}
\usepackage{array}
\usepackage{amsfonts}
\usepackage{amsmath}
\usepackage{amssymb}
\usepackage{amsthm}
\usepackage{mathrsfs}
\newtheorem{sdtheorem}{Theorem}[section]
\newtheorem{sdproposition}[sdtheorem]{Proposition}

\theoremstyle{definition}
\newtheorem{sddefinition}[sdtheorem]{Definition}
\theoremstyle{plain}
\usepackage{nicefrac}
\usepackage{microtype}
\usepackage{xcolor}
\usepackage{colortbl} 
\usepackage{graphicx}
\usepackage{wrapfig}
\usepackage{pifont}
\graphicspath{{figures/}}
\usepackage[section]{placeins} 
\usepackage{tikz}
\usetikzlibrary{positioning, fit, calc, arrows.meta, shapes.geometric, backgrounds, decorations.markings}
\usepackage{listings}
\usepackage{enumitem}
\usepackage{inconsolata}
\usepackage[most]{tcolorbox}
\tcbuselibrary{skins,breakable,listings}


\lstdefinestyle{prompt}{
  basicstyle=\ttfamily\footnotesize,
  breaklines=true,
  breakindent=0pt,
  breakatwhitespace=true,
  columns=fullflexible,
  keepspaces=true,
  showstringspaces=false,
  frame=none,
  aboveskip=0pt,
  belowskip=0pt,
}
\tcbset{
  promptboxbase/.style={
    enhanced,
    colback=gray!5,
    colframe=black,
    coltitle=white,
    fonttitle=\bfseries\small,
    boxrule=0.6pt,
    left=4pt,
    right=4pt,
    top=4pt,
    bottom=4pt,
    before skip=6pt,
    after skip=6pt,
  },
  promptbox/.style={
    promptboxbase,
    unbreakable,
  },
  longpromptbox/.style={
    promptboxbase,
    breakable,
  },
  toolbox/.style={
    unbreakable,
    enhanced,
    colback=green!3,
    colframe=green!40!black,
    coltitle=white,
    fonttitle=\bfseries\small,
    boxrule=0.6pt,
    left=4pt,
    right=4pt,
    top=4pt,
    bottom=4pt,
    before skip=6pt,
    after skip=6pt,
  },
  vllmbox/.style={
    unbreakable,
    enhanced,
    colback=teal!4,
    colframe=teal!55!black,
    coltitle=white,
    fonttitle=\bfseries\small,
    boxrule=0.6pt,
    left=4pt,
    right=4pt,
    top=4pt,
    bottom=4pt,
    before skip=6pt,
    after skip=6pt,
  },
  professionalbox/.style={
    boxrule=0.5pt,
    arc=0.3mm,
    left=2mm, right=2mm, top=3mm, bottom=2mm,
    width=\linewidth,
    boxsep=0pt,
    enhanced,
    unbreakable,
    shadow={1mm}{-1mm}{0.5mm}{black!30},
    attach boxed title to top left={yshift=-1.5mm, xshift=1mm},
    fonttitle=\scriptsize\sffamily\fontseries{b}\selectfont,
    coltitle=white,
    boxed title style={
      colback=#1!70!black,
      colframe=#1!60!black,
      boxrule=0.3pt,
      arc=0.5mm,
      left=1.5mm, right=1.5mm, top=0.4mm, bottom=0.4mm,
    },
    colback=#1!4,
    colframe=#1!50!black,
  },
}
\definecolor{toolgreen}{RGB}{34, 139, 34}
\definecolor{tooldescblue}{RGB}{70, 130, 180}
\definecolor{toolparamsorange}{RGB}{210, 105, 30}
\definecolor{reviewblue}{RGB}{0, 92, 170}
\definecolor{revieworange}{RGB}{180, 95, 0}
\definecolor{reviewpurple}{RGB}{110, 45, 150}
\definecolor{abstractgray}{RGB}{245, 246, 248}

\definecolor{pathblue}{RGB}{35,55,145}
\definecolor{pathred}{RGB}{215,30,30}
\definecolor{pathorange}{RGB}{235,140,20}
\DeclareRobustCommand{\pblue}[1]{\textcolor{pathblue}{#1}}
\DeclareRobustCommand{\pred}[1]{\textcolor{pathred}{#1}}
\DeclareRobustCommand{\porange}[1]{\textcolor{pathorange}{#1}}
\definecolor{markrep}{RGB}{34,120,60}
\definecolor{markexc}{RGB}{0,84,40}
\definecolor{markabv}{RGB}{196,108,0}
\DeclareRobustCommand{\rep}{\textcolor{markrep}{\ding{51}}}
\DeclareRobustCommand{\exc}{\textcolor{markexc}{\ding{51}\kern-0.28em\ding{51}}}
\DeclareRobustCommand{\abv}{\textcolor{markabv}{\ding{115}}}
\DeclareRobustCommand{\nopub}{\textcolor{black!35}{---}}

\newcommand{\famrow}[2]{\multicolumn{#1}{@{}l}{\textsc{\textcolor{black!60}{#2}}}}
\newcommand{\modelname}[1]{\hspace{0.9em}#1}
\newcommand{\marklegend}{{\footnotesize \rep\ reproduced \qquad \exc\ exceeded \qquad \abv\ not reached \qquad \nopub\ nothing published}}

\newtcolorbox{toolsection}[1]{%
  professionalbox=toolgreen,
  title={\fontseries{b}\selectfont TOOL: #1},
}
\newtcolorbox{toolconfigbox}[1]{%
  colback=toolgreen!4,
  colframe=toolgreen!60!black,
  boxrule=0.8pt,
  arc=0.5mm,
  left=2mm, right=2mm, top=4mm, bottom=2mm,
  enhanced,
  unbreakable,
  shadow={1mm}{-1mm}{0.5mm}{black!25},
  title={\normalfont\bfseries\sffamily TOOL: #1},
  coltitle=white,
  fonttitle=\scriptsize\sffamily\bfseries,
  attach boxed title to top left={yshift=-2mm, xshift=2mm},
  boxed title style={
    colback=toolgreen!80!black,
    colframe=toolgreen!90!black,
    boxrule=0.5pt,
    arc=0.3mm,
    left=1.5mm, right=1.5mm, top=0.4mm, bottom=0.4mm,
  },
}
\newtcolorbox{tooldescriptionsection}{%
  professionalbox=tooldescblue,
  title={\fontseries{b}\selectfont DESCRIPTION},
  left=1.5mm, right=1.5mm, top=3mm, bottom=1mm,
}
\newtcolorbox{toolparamssection}{%
  professionalbox=toolparamsorange,
  title={\fontseries{b}\selectfont PARAMETERS},
  left=1.5mm, right=1.5mm, top=3mm, bottom=1mm,
  fontupper=\ttfamily\footnotesize,
}
\definecolor{intentblue}{RGB}{70, 110, 175}
\definecolor{execplum}{RGB}{150, 70, 95}
\newtcolorbox{turnbox}[1]{%
  colback=black!2,
  colframe=black!45,
  boxrule=0.8pt,
  arc=0.5mm,
  boxsep=1pt,
  left=2mm, right=2mm, top=2.6mm, bottom=1.2mm,
  enhanced,
  unbreakable,
  shadow={0.7mm}{-0.7mm}{0.4mm}{black!18},
  title={\normalfont\bfseries\sffamily #1},
  coltitle=white,
  fonttitle=\scriptsize\sffamily\bfseries,
  attach boxed title to top left={yshift=-1.8mm, xshift=2mm},
  boxed title style={
    colback=black!62,
    colframe=black!75,
    boxrule=0.5pt,
    arc=0.3mm,
    left=1.5mm, right=1.5mm, top=0.4mm, bottom=0.4mm,
  },
  before skip=3pt, after skip=4pt,
}
\newtcolorbox{intentsection}{%
  professionalbox=intentblue,
  title={\fontseries{b}\selectfont INTENT \textnormal{(model's reasoning)}},
  boxsep=1pt,
  left=1.5mm, right=1.5mm, top=2.3mm, bottom=0.8mm,
  before skip=2pt, after skip=3pt,
}
\newtcolorbox{executionsection}{%
  professionalbox=execplum,
  title={\fontseries{b}\selectfont EXECUTION \textnormal{(harness closed-loop)}},
  boxsep=1pt,
  left=1.5mm, right=1.5mm, top=2.3mm, bottom=0.8mm,
  before skip=2pt, after skip=2pt,
}

\title{Dissecting model behavior through agent trajectories}

\author{%
  Gaurav Gupta, Vatshank Chaturvedi, Sudipta Sengupta, Jun Huan \& Anoop Deoras \\
  AWS AI Labs \\
  \texttt{\{gauravaz, vatshc, sudipta, lukehuan, adeoras\}@amazon.com}
}

\iclrfinalcopy 

\fancypagestyle{titlepage}{%
  \fancyhf{}%
  \renewcommand{\headrulewidth}{0.4pt}%
  \lhead{\includegraphics[height=0.7cm]{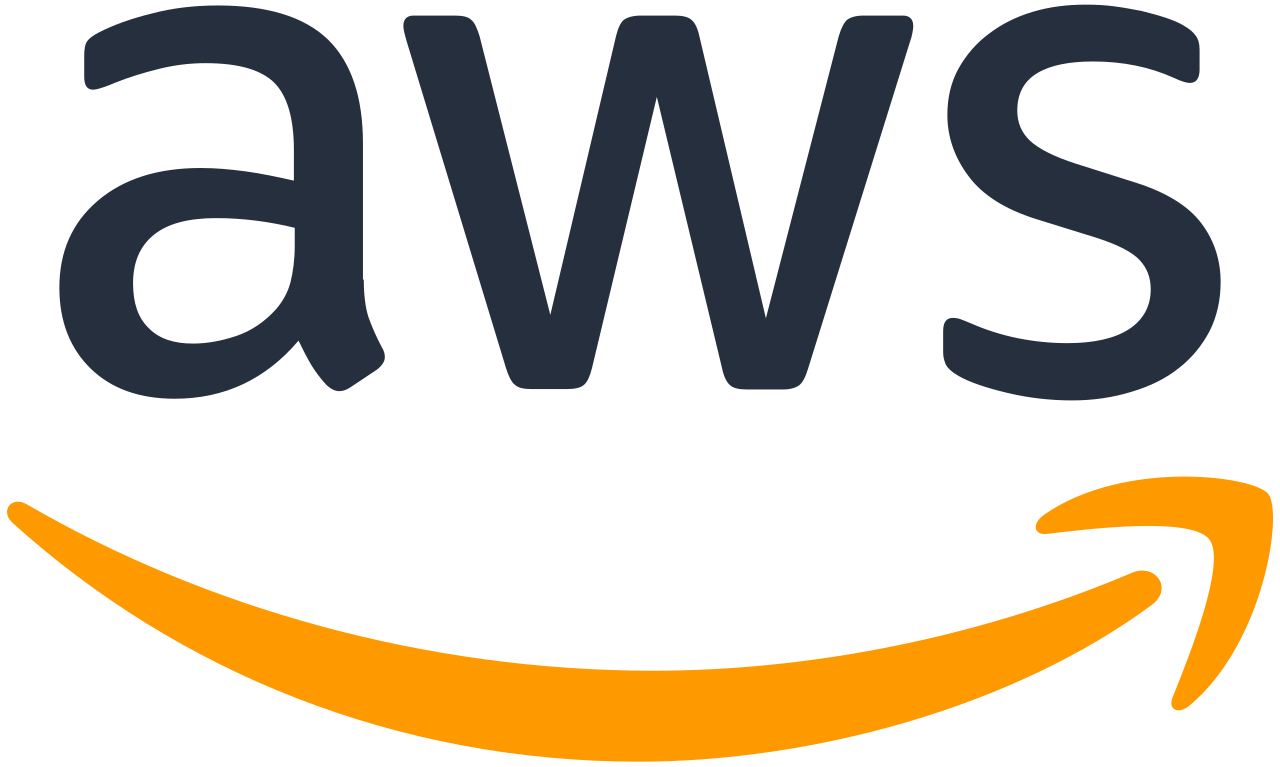}}%
  \cfoot{\thepage}%
}
\begin{document}

\raggedbottom

\maketitle
\thispagestyle{titlepage}

\begin{abstract}
AI agent performance is not just a modeling problem, it is fundamentally a
systems problem. The advanced capabilities of models are realized through agent harnesses. 
Therefore, a gap between model assumptions and harness behavior can easily prevent the model's full capabilities from translating into agent performance.
We formalize this as the `intent-execution' gap: the mismatch between
what the model intends and what the harness executes, and vice versa. 
We argue that minimizing this intent-execution gap is as important as other aspects of harness design such as tools and execution loops.
To illustrate the impact of this harness-model alignment, we develop a simple and customizable harness called `Simple Strands Agent' (SSA). SSA aims to find the bulk of common patterns which generalize across different model families (such as Claude, Gemini, GPT, Grok, Qwen), as well as a small number of model-specific preferences.
We make two contributions: (i) we \textbf{reproduce or improve on the pass@1} performance reported by diverse model-provider families on popular agentic benchmarks (SWE-Pro, SWE-Verified, Terminal-Bench-2, and Deep-SWE), and (ii) building on an \textbf{analysis of 154\texttt{k} trajectories generated by SSA}, we look beyond the \texttt{pass@1} numbers which tend to be relatively even across frontier models. 
By representing agent trajectories in code state-spaces, we compute solution distance and backtracking to observe model-level differences in problem-solving behavior.
Finer-grained metrics such as edit frequency, testing activity, and phase-transitions reveal how individual models allocate effort across different stages of problem solving.

\vspace{4pt}
\noindent\textbf{Code:} \url{https://github.com/strands-labs/benchmark-harnesses}
\end{abstract}

\section{Introduction}
\label{sec:introduction}

A modern coding agent couples an LLM with a
\emph{harness}, the software layer that mediates every tool call, feeds execution outcomes back into the context window, and manages the thought-action-observation
loop~\citep{wang2024agentsurvey,xi2023agentsurvey,anthropic2024agents,yang2024sweagent}.

As base models improve, the alignment of the model with the harness becomes critical to realizing  the model’s full capabilities. Informally, given two models $\mathcal{M}_A > \mathcal{M}_B$ in raw capability, a harness $H$ without such alignment can yield $\mathrm{acc}(H, \mathcal{M}_A) \approx \mathrm{acc}(H, \mathcal{M}_B)$ where $acc$ is accuracy.

We formalize this misalignment between harness and model as the \textbf{intent-execution gap}: a bidirectional mismatch between what the model intends and what the harness
executes, and between what happened in the environment and what is revealed to the model 
. The gap is bidirectional because failures compound in both
directions as a misinterpreted edit silently corrupts state, and an
under-specified execution result leads the model to reason from false
premises. For instance, a model may intend to replace a single occurrence of a
function call, while the harness silently patches every match; or a long shell
output may be truncated in a way that hides the test failure the model needs
to see. 

Problem-specific tunings such as bespoke
prompts~\citep{brown2020gpt3,kojima2022zerocot}, specialized tool
schemas~\citep{schick2023toolformer,qin2024toolllm}, or hand-crafted execution
graphs~\citep{yao2023react,shinn2023reflexion,wu2023autogen} can reduce this
gap for a given model on a given benchmark. But such gains are brittle. As
models improve~\citep{wei2022emergent,guo2025deepseekr1}, behaviors shift
and optimizations that overfit one model's quirks can degrade on the
next~\citep{anthropic2024agents,wang2024agentsurvey}. This motivates a different strategy: rather than only chasing model-specific tricks, identify
\emph{invariant} interface principles that remain effective
across model families, model generations, and benchmarks, and adapt only where
families genuinely differ. This perspective raises two fundamental questions at the model-harness boundary:
\begin{enumerate}
    \item Does the harness faithfully execute what the model intended?
    \item Is the model given an accurate account of what the harness did?
\end{enumerate}
We show that minimizing this bidirectional gap without task-specific
tuning is sufficient to match or exceed official benchmark numbers for
models we evaluate across seven families (Claude, GPT, Gemini,
Grok, Kimi, Muse Spark, Qwen) on SWE-Bench-Verified, SWE-Bench-Pro, Terminal-Bench-2 and Deep-SWE.

We also analyze SSA trajectories to answer a question that
pass@1 cannot: \emph{how} do different model families solve the same
problems? By projecting agent state into a text-level code space and measuring
distance to verified solutions at each cycle, we surface stable behavioral
signatures (Section \ref{sec:distance}) in the form of tool-call intent, phase (exploration vs implementation), backtracking rates that differentiate families even when their pass@1 scores are
indistinguishable.

\paragraph{Main Contributions.}
\begin{enumerate}[leftmargin=*]
    \item We introduce Simple Strands Agent (SSA), an open-source harness
    designed around the intent-execution gap, and show that it reproduces or
    exceeds the published benchmark numbers for 61 of 66 model--benchmark pairs across seven model families on SWE-Bench-Verified,
    SWE-Bench-Pro, Terminal-Bench-2 and Deep-SWE.

    \item We define a solution-distance metric over a text-level code space that
    tracks, at each agent cycle, how far the repository state is from a
    verified-correct fix, and decompose it exactly into best-so-far
    progress and a backtracking term, giving backtracking a precise and time-dependent
    definition.

    \item We present a trajectory-level analysis across all 38 evaluated models revealing family-specific behavioral
    signatures such as edit/test ratios, phase schedules, and backtracking patterns, that are stable
    across benchmarks and invisible to aggregate metrics.

    \item We identify and quantify a git-history leakage channel in the public
    SWE-Bench-Pro containers that inflates measured pass@1 by up to $6.9\%$
     for the affected models (Appendix~\ref{app:eval-integrity}).
\end{enumerate}

Section~\ref{sec:methods} explains harness improvements that make SSA effective,
Section~\ref{sec:distance} introduces a framework for tracking agent code-states as they evolve over the course of a trajectory, and
Section~\ref{sec:results} reports SSA's performance on agentic benchmarks and the analysis of model trajectories.
Appendix~\ref{app:architecture} outlines the SSA architecture.

\section{SSA Refinements}
\label{sec:methods}

We now describe methods that make SSA effective, namely, feedback-rich tool boundaries in Section\,\ref{sec:methods:harness-improvements}, our approach to prompting agents for a better balance between reasoning and tool-use in Section\,\ref{ssec:reasoning_nudge}, as well as leaning on the model's preferences for tool interfaces in Section\,\ref{ssec:model_tool_preference}.

\subsection{Feedback-rich tool boundaries}
\label{sec:methods:harness-improvements}

The key design principle of the SSA harness is to make the information exchanged at the model-tools boundary transparent and explicit. We do so through the following mechanisms:

\textbf{Structured execution feedback.}
All shell calls return the following information: command, status, exit code, output,
and error text. When interrupted by a timeout, partial output is returned when available. Long outputs are clipped in the middle but preserve the start and end of the outputs so that failure summaries and final test results are not lost.

\textbf{Safer editing semantics.}
All edits are anchored to neighboring lines. 
Partial-line matches and duplicate matches are rejected with feedback that helps the model disambiguate and issue a more precise edit in the next turn. Successful edits return a unified diff reducing the need for the model to verify if the intended change was actually applied.

\subsection{Balancing reasoning with tool-calling}
\label{ssec:reasoning_nudge}

Chain-of-thought reasoning~\citep{wei2022chainofthought,kojima2022zerocot} is clearly valuable. It allows the
model to decompose a problem, plan next-steps, and decide which tool to invoke. 
Without sufficient reasoning, tool-usage becomes reactive, leading to shallow exploration, redundant calls, or poor sequencing of actions. 
However, excessive thinking introduces its own failure mode. When the model spends too long reasoning internally, it begins to form assumptions about the environment rather than verifying them.
As a result, the agent may issue poorly grounded tool calls or skip necessary validation steps altogether, creating a fundamental tension between the model's internal state and the external environment.

Effective agents must continuously reconcile these two demands, and we
refer to this balance as \textit{tool calling with a reasoning nudge}.
The idea is to encourage the model to perform just enough reasoning to decide the
next action and then prioritize evidence-gathering interactions with the
environment over further reasoning. Rather than extending internal chains-of-thought, the agent is nudged toward validating its hypotheses through tool outputs~\citep{yao2023react,shinn2023reflexion,schick2023toolformer}.

In practice, we did not find a single ``golden prompt'' that reliably balances reasoning and tool interaction across all model families. 
For the Claude variants, we found that a quantitative target, e.g., ``make 50+ tool calls'' or ``ideal tool call count is 100''\footnote{This style of quantitative tool-call prompt first appeared in the Claude Sonnet 4.5 release: \url{https://www.anthropic.com/news/claude-sonnet-4-5}.} helps shift the balance from reasoning toward interaction with the environment.
While the exact number of target tool calls is not important, it serves as a useful north star that biases the model toward
action.
Under the same prompt, Gemini and Grok runs frequently emitted empty tool calls up to the stated mark.
Here, we found that using a qualitative direction such as ``You should use tools as much as possible'' achieves the same balance.

\begin{figure}[h]
    \centering
    \includegraphics[width=0.55\linewidth]{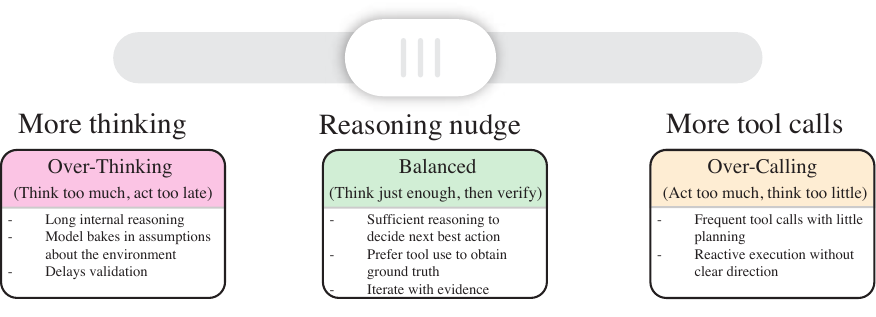}
    \caption{\textbf{Reasoning nudges in SSA}: Each family is nudged toward tool use in the form, quantitative or qualitative, it responds to.}
    \label{fig:reasoning-nudge}
\end{figure}

\subsection{Aligning harness with the model's tool use preferences}
\label{ssec:model_tool_preference}

Across agents, while tools function in exactly the same way, models tend to
exhibit distinct preferences in how they invoke them. For example, GPT
models prefer to update code by using an \texttt{apply\_patch} command to splice in text from a separate 
file, formatted in a particular way. 
Performance depends on honoring these formatting preferences~\citep{yang2024sweagent,kim2026codestruct,wang2024codeact,xia2024agentless}.

Similarly, for Grok-4.20, a single monolithic tool for editing and viewing performs poorly, whereas splitting functionality into atomic operations yields better results even though the functionality remains unchanged.
Additionally, viewing line numbers in a file helps most models, but Grok has trouble separating prefixes from line numbers. Here, disabling line-numbers helps the model use the view-tool effectively.
These preferences are a by-product of training, and an effective harness should leverage them rather than fight them~\citep{anthropic2024agents,wang2024agentsurvey}.

This reinforces a broader design principle that agent performance is a function of not only what tools are available but also how naturally those tools align with the model's learned behaviors. 
A well-designed harness should meet the model where it is, adapting interfaces and feedback to its strengths while still enforcing the invariants needed for reliable execution.

\section{Solution distance: measuring trajectory quality}
\label{sec:distance}

\paragraph{Motivation.}
Standard agentic-coding metrics such as pass@1, tool-calls distribution, number of output-tokens summarize \emph{outcome} and \emph{cost} but not the \emph{trajectory quality}. Two runs that achieve similar pass@1 on the same instance can still differ markedly. One may proceed through a clean explore$\to$localize$\to$implement$\to$verify arc with a single decisive edit while another may oscillate between candidate fixes, undo and reapply edits, or commit a correct patch and subsequently retract it. To study and isolate this behavior, we introduce metrics in this subsubsection that answer a single, mechanistic question at every point in a trajectory: \emph{how close is the agent's current repository state to a verified-correct solution} for this instance, and \emph{does it move monotonically towards one or does it backtrack?}

\paragraph{Text-level code space.}
Let $\mathcal{P}$ be the universe of absolute file paths and $\Sigma^\ast$ the set of exact file-content strings. Let $x$ be a map from a file-path to its contents such that $x:\mathcal{P}\to\Sigma^\ast\cup\{\bot\}$. The ambient text-level code space is
\begin{equation}
  \mathcal{X}^{\mathrm{text}}
  =
  \{\,x
  \;|\;
  |\{f:x(f)\neq\bot\}|<\infty\,\},
  \label{eq:soldist:text-code-space}
\end{equation}
where $x(f)$ is file $f$'s exact text and $\bot$ denotes absence of content. Therefore, creation and deletion are represented by transitions from or to $\bot$.
For brevity, we drop the superscript in $\mathcal{X}^{\mathrm{text}}$ and write this text-level space as $\mathcal{X}$.

For a problem instance $i$ (such as a repository-level coding task), let $\mathcal{R}_i$ be the given workspace root, $\mathcal{P}_{\mathcal{R}_i}\subset\mathcal{P}$ the paths under that root, $x_i^0$ the given code, and $\mathcal{F}_i\subset\mathcal{P}_{\mathcal{R}_i}$ the retained source-path set. The instance-level text subspace is defined as follows:
\begin{equation}
  \mathcal{X}_i
  =
  \{\,x\in\mathcal{X}
  : x(f)=x_i^0(f)\ \forall\ f\in\mathcal{P}\setminus\mathcal{P}_{\mathcal{R}_i}\,\}.
  \label{eq:soldist:instance-text-code-space}
\end{equation}
Thus states in $\mathcal{X}_i$ may vary only on the retained paths under the workspace root. A trajectory state $x_i(t)$ is the accepted textual mutations applied to $x_i^0$ and projected to this subspace. We write the retained-path projection as $\pi_i(x)=x|_{\mathcal{P}_{\mathcal{R}_i}}$.

\textbf{Prior Work} Generate-and-validate repair systems define \emph{patch} or \emph{search} spaces over syntactic program modifications, AST edits, edit locations, mutation operators, and repair ingredients~\citep{legoues2019automatedrepair,legoues2012systematic,long2016analysis,barr2014plastic}; program-representation work maps code into token, AST-path, data-flow, or graph-structured representations~\citep{alon2018general,guo2021graphcodebert}; semantic genetic programming defines distances in a semantic output space rather than directly over program syntax~\citep{moraglio2012geometric}; and recent work contrasts discrete symbolic source-token spaces with continuous numerical program spaces~\citep{silva2025gradientrepair}.

Since most coding agents (including SSA) operate on the observable source text and apply edits as text file mutations and evaluation also uses text patches, using $\mathcal{X}$ as a text-level code state space is the most useful. Note that $\mathcal{X}_i$ is an instance-level subspace and not a semantic program space.

Let $O_i:\mathcal{X}_i\to\{0,1\}$ be the evaluation oracle, with $O_i(x)=1$ iff materializing state $x$ in the testing environment passes all of the instance's tests. The true solution set in this space is
\begin{equation}
  \mathcal{S}_i
  =
  \{\,x\in\mathcal{X}_i:O_i(x)=1\,\}.
  \label{eq:soldist:true-solution-set}
\end{equation}
Next, to be able to express similarity in the observable space, we define the text-diff projection as
\begin{equation}
  \phi_i(x)
  =
  \tau\!\left(\mathrm{diff}(\pi_i(x_i^0),\pi_i(x))\right),
  \label{eq:soldist:text-diff-projection}
\end{equation}

\begin{figure}[t]
    \centering
    \includegraphics[width=0.9\linewidth]{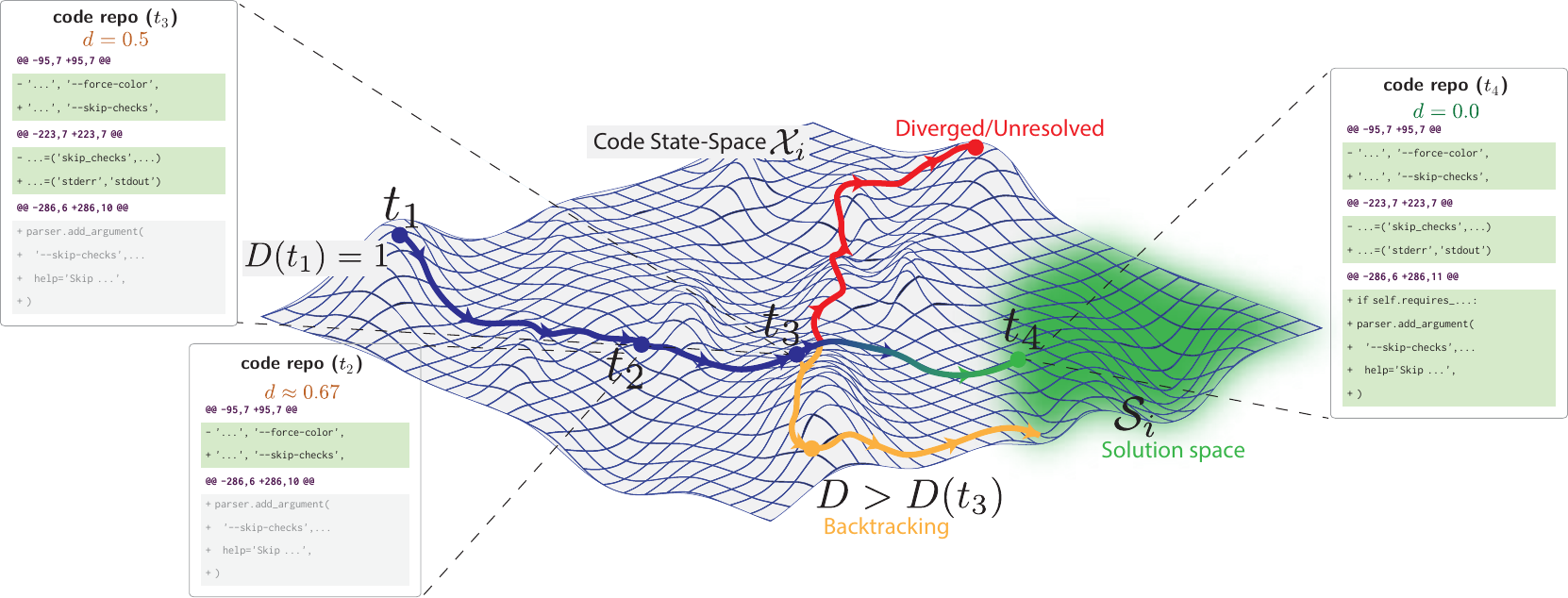}
    \caption{\textbf{Agent trajectory in code state-space}. Solution divergence $D(t)$ evolution as agent traverses code state-space $\mathcal{X}_{i}$ from the given initial state at $t_1$
    till it reaches solution space $\mathcal{S}_i$ at step $t_4$ along the \pblue{blue} path. Each inset compares the reconstructed live
    diff with the closest element in $\mathcal{S}_i$. Green
    lines are reproduced patch features and gray lines are still missing. For the successful work, the
    trace follows a staircase $D=1.0 \to D\approx0.67 \to D=0.5 \to D=0$ as the
    agent accumulates the repair. The \pred{red} path is an unresolved run, which terminates outside $\mathcal{S}_i$. The \porange{orange} path backtracks: it transiently moves away from the solution space, $D > D(t_3)$, before returning and resolving. Appendix~\ref{app:metrics:solution-distance}
    describes the empirical reference construction and replay procedure.}
    \label{fig:distance-demo}
\end{figure}

where \texttt{diff} is the unified diff against the projected base state and
$\tau$ extracts the patch features defined in
Appendix~\ref{app:metrics:solution-distance}.

\begin{sddefinition}[Recall divergence]
\label{def:soldist:recall-divergence}
The recall divergence from the current state $x_i(t)$ to the nearest
oracle-correct state is
\begin{equation}
  D_i(t) \;=\; \min_{x^\star \in \mathcal{S}_i}
              \left(
              1 - \frac{|\,\phi_i(x_i(t)) \cap \phi_i(x^\star)\,|}
                       {|\,\phi_i(x^\star)\,|}
              \right),
  \label{eq:distance-main}
\end{equation}
where the minimization is over oracle-correct states with non-empty projected
source diffs. The recall denominator is deliberate as it measures recovery of a
known-correct fix, while the solution oracle remains responsible for judging
whether additional edits are harmless.
\end{sddefinition}

In practice, $\mathcal{S}_i$ is not available or enumerable. Our experiments
therefore approximate Equation~\ref{eq:distance-main} with a finite empirical
reference set of verified patches and faithful replay of the live repository
state (see Appendix~\ref{app:metrics:solution-distance}). The resulting time-series is used for mean-distance curves, per-edit
$\Delta D$ backtracking, area under the distance curve, and time-to-threshold
summaries such as $t$ at $D \le 0.1$.

{\textbf{Note}}: $D$ starts at $1$ (no edits, no overlap) and falls to $0$ once the state $x(t)$
fully reproduces a correct repair under the projection. Since $D$ could
rise or fall, the per-edit change
$\Delta D > 0$ becomes a clean signal of backtracking as we formalize in Section\,\ref{sec:soldist:cohort-refinement}.
\subsection{Level sets and backtracking}
\label{sec:soldist:cohort-refinement}

Consider $N$ distance (Eq.(\ref{eq:distance-main}) trajectories
$D_r:[0,1]\to[0,1]$, each a right-continuous step function on normalized
time $t$ with $t=0/1$ at beginning/end of the agent run.
The mean, $\mathbb{E}\,D(t)$ (mean over all instances) can descend gradually even
when individual runs make a single sharp drop at different times (see Figure\,\ref{fig:solution-distance-illustration} in Appendix\,\ref{app:metrics:soldist:examples}).

\begin{sddefinition}[Level-set processes]
\label{def:soldist:level-set-processes}
For each run $r$ and level $\theta\in[0,1]$, define the first entry into the
distance sublevel set and the best-so-far distance by
\begin{equation}
  T_r(\theta)=\inf\{t\in[0,1]:D_r(t)\le\theta\},
  \qquad M_r(t)=\min_{0\le s\le t}D_r(s),
  \label{eq:soldist:cohort-refinement:crossing-time}
\end{equation}
Finite step trajectories ensure that every finite crossing is attained.
Let $R\sim\operatorname{Unif}\{1,\ldots,N\}$ denote a uniformly selected run.
The corresponding \emph{level-survival curve} is
\begin{equation}
  S_\theta(t)
  :=\Pr\{T_R(\theta)>t\}
   =\Pr\{M_R(t)>\theta\}
   =1/N\sum_{r=1}^N\mathbf1\{T_r(\theta)>t\}.
  \label{eq:soldist:cohort-refinement:level-curve}
\end{equation}
It is the fraction of runs that have not yet reached level $\theta$, and is
non-increasing in both $t$ and $\theta$.
\end{sddefinition}

Next, we write the cohort averages as
\[
  \textstyle
  \bar D(t)=1/N\sum_{r=1}^N D_r(t),\quad
  \bar M(t)=1/N\sum_{r=1}^N M_r(t).
\]

\begin{sdtheorem}[Level-set decomposition]
\label{thm:soldist:cohort-refinement:decomposition}
For every $t\in[0,1]$, these averages satisfy
\begin{equation}
  \bar D(t)=\bar M(t)+\bar B(t),
  \label{eq:soldist:cohort-refinement:level-decomposition}
\end{equation}
where $\bar M(t)=\int_0^1S_\theta(t)\,d\theta$ and
$\bar B(t):=1/N\textstyle\sum_{r=1}^N[D_r(t)-M_r(t)]\ge0$.
\end{sdtheorem}

\begin{sdproposition}[Zero backtracking]
\label{prop:soldist:cohort-refinement:zero-backtracking}
For any fixed $t\in[0,1]$, $\bar B(t)=0$ iff $D_r(t)=M_r(t)$ for every
run $r$. Consequently, $\bar B\equiv0$ iff every run is non-increasing.
\end{sdproposition}
Proofs of Theorem~\ref{thm:soldist:cohort-refinement:decomposition} and
Proposition~\ref{prop:soldist:cohort-refinement:zero-backtracking} are given in
Appendix~\ref{app:metrics:soldist:proofs}.

Thus, backtracking $\bar B(t)$ is the mean distance above each run's best previous state:
coverage previously attained but currently lost. Interestingly, this signal can remain positive while a run recovers from an earlier reversal.


\section{Experimental Results}
\label{sec:results}

We evaluate SSA on four agentic benchmarks, namely, SWE-Bench-Verified~\citep{jimenez2024swebench}, 
SWE-Bench-Pro~\citep{deng2025swebenchpro},
Terminal-Bench-2~\citep{merrill2026terminalbench} and Deep-SWE~\citep{datacurve2026deepswe} and seven model families: Claude, GPT, Gemini, Grok, Kimi, Muse Spark and Qwen. 
Under one harness, SSA reproduces or exceeds the published pass@1 for 61 of the 66 model--benchmark pairs 
with a published figure, and is ahead of widely-used open-source harnesses. Next, in this Section, we analyze trajectories ($154$k in total) from all these runs to better understand how each model-family approaches problem solving, information that is not expressed just looking at pass@1. 

\subsection{Setup and protocol}
\label{sec:results:setup}

Experimental-setup details (compute cluster, model serving, per-model inference settings, and the evaluation-hygiene protocol) are given in Appendix~\ref{sec:appendix-configs}, and the prompts in Appendix~\ref{app:prompts}. 
All four benchmark ship with pre-existing tests. The SWE benchmarks serialize the agent's workspace changes
into a patch and replay it in a separate evaluator container. Terminal-Bench-2 evaluates in-place but imposes explicit per-task compute and wall-time limits that the SWE benchmarks do not. 
Throughout, we report pass@1 following the convention of \citet{chen2021codex}. We follow a strict routine of counting every failure (agent or infra) into pass@1 and not alter denominator which otherwise can inflate the scores. Each model is run five times on the full benchmark.
In total we evaluate 38 models (see Appendix~\ref{app:passrates}). The main sweep of 21 models over three benchmarks yields $(500+731+89)\times5\times21=138{,}600$ trajectories; the second Terminal-Bench-2 setting for the 13 closed models adds $89\times5\times13=5{,}785$, and the 17 models on Deep-SWE add $113\times5\times17=9{,}605$ yielding $153{,}990$ trajectories in total, $\approx154$k, which we analyze further for model behavior.


\subsection{Published pass rates are recovered under one harness}
\label{sec:results:headline}

\textbf{SSA reproduces or exceeds official model-card numbers under a single harness}.
Every model is run under the same execution loop, feedback semantics and
evaluation protocol (Section~\ref{sec:results:setup}); only the narrow
interface adapters of Appendix~\ref{sec:appendix-configs} differ.
Tables~\ref{tab:results-reproduction} and~\ref{tab:results-reproduction-deep} record, for each model and benchmark,
whether the figure the provider publishes falls inside SSA's $95\%$ confidence
interval, below it, or above it. Appendix~\ref{app:passrates} reports the mean
per-run resolution rate with $95\%$ confidence intervals over all per-instance
trials ($2{,}500$ for SWE-Bench-Verified, $3{,}655$ for SWE-Bench-Pro, $445$ for
Terminal-Bench-2, and $565$ for Deep-SWE). Of the 66 model--benchmark pairs for which a
published figure exists (43 in Table~\ref{tab:results-reproduction}, 17 in Table~\ref{tab:results-reproduction-deep}, and 6, all reproduced, from models added after the main sweep that appear only in Appendix~\ref{app:passrates}), 61 fall within or below SSA's interval (44
reproduced and 17 exceeded) and 5 lie above it, four of them on SWE-Bench-Pro, whose public containers leak future git history (Appendix~\ref{app:eval-integrity}; per-model figures in Appendix~\ref{app:passrates}). These results characterize model–harness configurations under the stated evaluation conditions and they do not establish a general ordering of provider capability.

For Terminal-Bench-2 we report two settings (motivated in
Appendix~\ref{app:tb2-infra}): \emph{Constrained}, which honors the benchmark's
memory and time budgets, and \emph{Unconstrained}, which removes them and acts
as an upper bound on performance. Providers report their figures under the
Constrained setting, except for the Qwen models, whose published figures are
Unconstrained; Table~\ref{tab:results-reproduction} compares each published
figure against the matching setting. Open-source models were run only
Unconstrained. Overall, we found that removing the memory and agent-time limits lifts pass@1 by
roughly $5$-$10\%$ across models (see Appendix~\ref{app:tb2-infra}).

\providecommand{\famname}[1]{\textsc{\textcolor{black!60}{#1}}}
\begin{table}[h]
\centering
\caption{\textbf{SSA reproduces or exceeds published pass@1.} Each provider's model-card figure is compared with SSA's five-run pass@1 and its $95\%$ confidence interval. Terminal-Bench-2 uses the setting the provider reports (Constrained for closed models, Unconstrained for Qwen). Rows are grouped by model family, with families in alphabetical order and models by version within a family. The order carries no ranking. Full numbers, and six further pairs from models added after the main sweep, are in Appendix~\ref{app:passrates}.}
\label{tab:results-reproduction}
\footnotesize
\setlength{\tabcolsep}{7pt}
\renewcommand{\arraystretch}{1.0}
\begin{tabular}{@{}llccc@{}}
\toprule
Family & Model & SWE-Bench-Verified & SWE-Bench-Pro & Terminal-Bench-2 \\
\midrule
\famname{Claude} & Opus 4.6 & \rep & \exc & \exc \\
 & Opus 4.5 & \rep & \exc & \exc \\
 & Sonnet 4.6 & \rep & \nopub & \exc \\
 & Sonnet 4.5 & \rep & \nopub & \rep \\
 & Sonnet 4.0 & \rep & \nopub & \nopub \\
 & Haiku 4.5 & \rep & \nopub & \rep \\
\addlinespace[3pt]
\famname{Gemini} & Gemini 3.1 Pro & \rep & \rep & \rep \\
 & Gemini 3 Flash & \rep & \nopub & \exc \\
\addlinespace[3pt]
\famname{GPT} & GPT-5.4 & \nopub & \exc & \rep \\
 & GPT-5.3 & \nopub & \exc & \rep \\
 & GPT-5.2 & \rep & \rep & \rep \\
 & GPT-5.2-codex & \nopub & \abv & \rep \\
 & GPT-OSS 120B & \exc & \nopub & \nopub \\
 & GPT-OSS 20B & \rep & \nopub & \nopub \\
\addlinespace[3pt]
\famname{Grok} & Grok 4.2 & \nopub & \nopub & \nopub \\
\addlinespace[3pt]
\famname{Qwen} & Qwen3-Coder Next & \rep & \exc & \rep \\
 & Qwen3-Coder 480B & \abv & \nopub & \nopub \\
 & Qwen3-Coder 30B & \rep & \nopub & \nopub \\
 & Qwen3.6 35B-A3B & \rep & \abv & \rep \\
 & Qwen3.6 27B & \rep & \abv & \rep \\
 & Qwen3.5 27B & \exc & \abv & \exc \\
\midrule
\multicolumn{2}{@{}l}{\emph{Totals}} & 14\,\rep\;\; 2\,\exc\;\; 1\,\abv & 2\,\rep\;\; 5\,\exc\;\; 4\,\abv & 10\,\rep\;\; 5\,\exc\;\; 0\,\abv \\
\bottomrule
\end{tabular}
\\[4pt]
\marklegend
\end{table}
\FloatBarrier

\begin{wraptable}{r}{0.415\textwidth}
\vspace{-\baselineskip}
\centering
\footnotesize
\setlength{\tabcolsep}{2.5pt}
\renewcommand{\arraystretch}{1.0}
\begin{tabular}{@{}llc@{}}
\toprule
Family & Model & Deep-SWE \\
\midrule
\famname{Claude} & Opus 5.5 & \rep \\
 & Opus 5.0 & \exc \\
 & Opus 4.8 & \rep \\
 & Sonnet 5.0 & \rep \\
 & Fable 5 & \rep \\
\addlinespace[3pt]
\famname{Gemini} & Gemini 3.8 Flash & \rep \\
\addlinespace[3pt]
\famname{GPT} & \shortstack[l]{GPT-6\\{\scriptsize (Astra/Sol/Luna)}} & \exc/\exc/\rep \\
 & \shortstack[l]{GPT-5.6\\{\scriptsize (Sol/Terra/Luna)}} & \rep/\rep/\rep \\
 & GPT-5.5 & \exc \\
\addlinespace[3pt]
\famname{Grok} & Grok 4.7 & \rep \\
\addlinespace[3pt]
\famname{Kimi} & Kimi-K3 & \exc \\
\addlinespace[3pt]
\famname{Muse Spark} & Muse Spark 1.3 & \rep \\
\addlinespace[3pt]
\famname{Qwen} & Qwen3.8 27B & \rep \\
\bottomrule
\end{tabular}
\caption{\textbf{Deep-SWE.} Marks as in Table~\ref{tab:results-reproduction}, 
against the provider's system or model card where published, otherwise 
the DataCurve v1.1 artifact.}
\label{tab:results-reproduction-deep}
\vspace*{-40pt}
\end{wraptable}
On Deep-SWE, whose tasks require substantially larger repairs than either SWE benchmark, the reference numbers come from vendor model cards where one exists
and otherwise from the DataCurve v1.1 artifact, against SSA's five attempts at a single fixed
configuration (Appendix~\ref{app:passrates}). The token-cost view of the SWE
results, a Pareto plot of total output tokens against pass@1, is in
Appendix~\ref{app:metrics:sbv} (Figure~\ref{fig:metrics-sbv-pass-vs-tokens}).

Taken together, the benchmark results show that a harness
built to minimize the intent-execution gap, with only narrow family-level
interface adapters, recovers published performance, within confidence intervals, for 61 of the 66 model-benchmark pairs, and out-performs
the default open-source harnesses practitioners typically use.

\subsection{Trajectory-level analysis}
\label{sec:results:trajectory-metrics}

Performance via pass@1 reports \emph{whether} a run was successfully resolved but says nothing about \emph{how}. Using an (i) LLM-judge (see Appendix\,\ref{app:metrics:judge} for details), we label each trajectory's tool-calls across R1-R8 rubrics as detailed in Table\,\ref{tab:judge-buckets}.
We partition tool calls into source-edit, scratch-testing (tests written by the agent itself), and suite-testing labels (pre-existing tests in the repository). We calculate the per-cycle R6 phase composition (\texttt{explore}/\texttt{localize}/\texttt{implement}/\texttt{verify}). Next, (ii) with code state-space, the solution-distance curve $D(t)$ and backtracking signal $B(t)$ (see Appendix\,\ref{app:metrics:soldist:backtrack}) are captured.
Together, these metrics give us a better understanding of how the models work in the setting of an agentic problem solving.

The full per-model data on these metrics for both SWE benchmarks can be found in Appendix~\ref{app:metrics:solution-distance}
(Figures~\ref{fig:metrics-sbv-editread}--\ref{fig:metrics-sbp-soldist-b}), and we summarize three findings below.

\begin{figure}[h]
\centering
\includegraphics[width=0.44\linewidth]{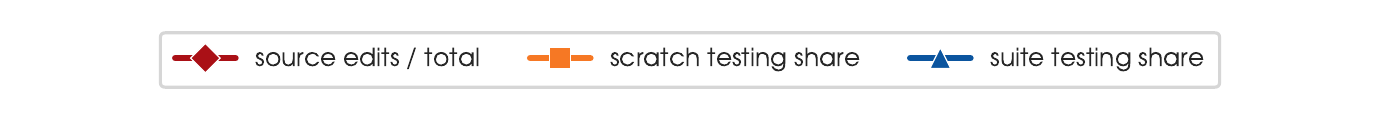}\\[2pt]
\setlength{\tabcolsep}{2pt}
\renewcommand{\arraystretch}{0.6}
\begin{tabular}{cccc}
\includegraphics[width=0.21\linewidth]{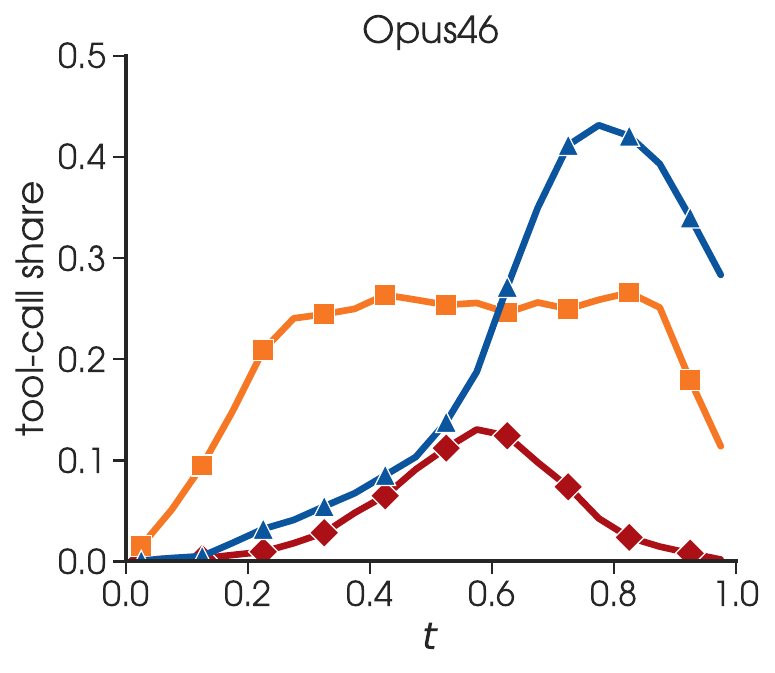} &
\includegraphics[width=0.21\linewidth]{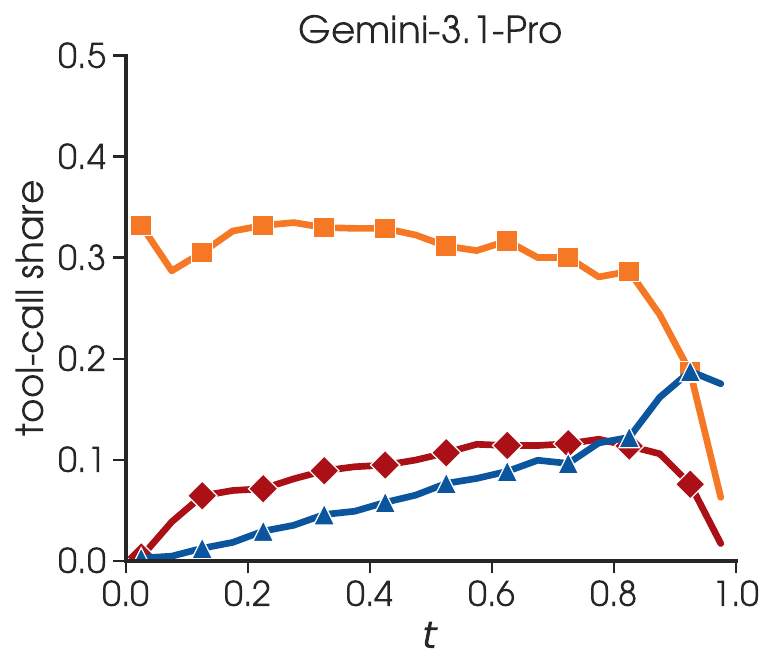} &
\includegraphics[width=0.21\linewidth]{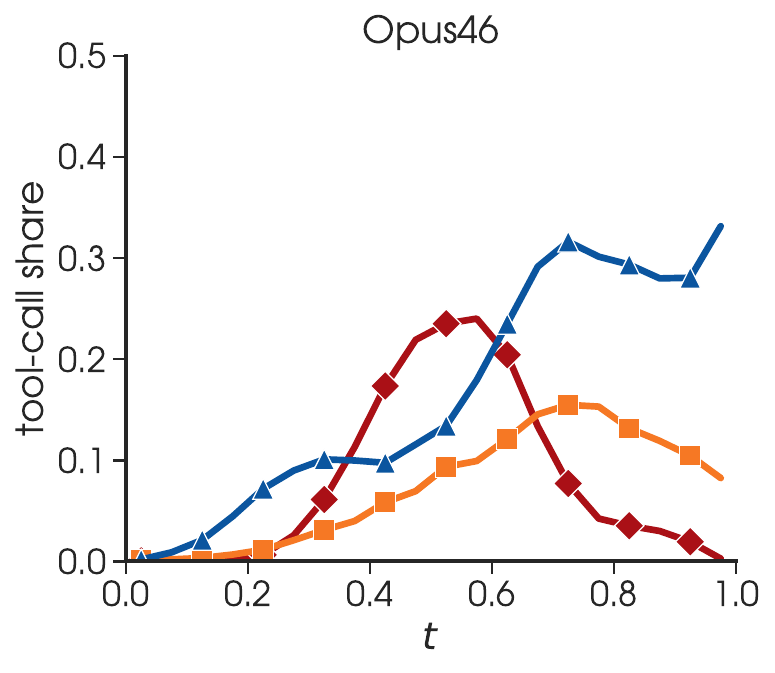} &
\includegraphics[width=0.21\linewidth]{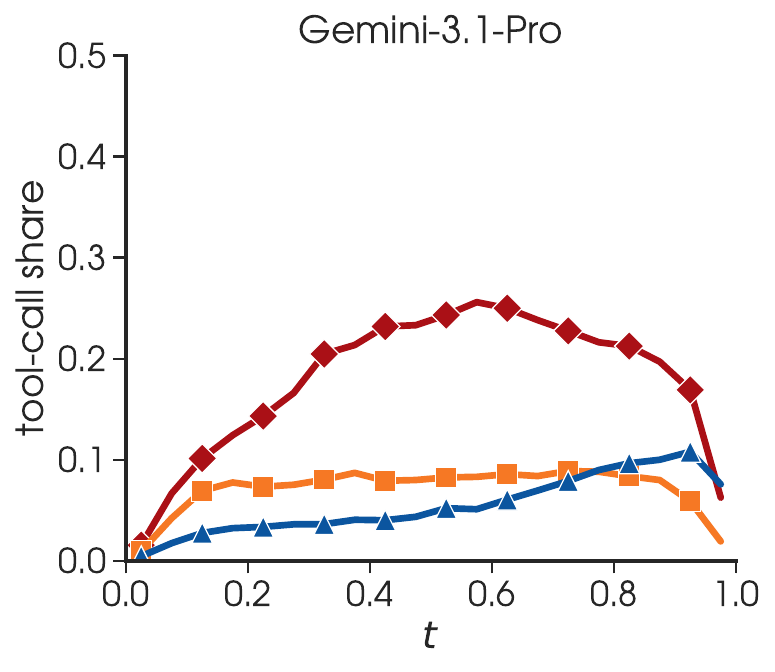} \\
\multicolumn{2}{c}{\textbf{SWE-Bench-Verified}} & \multicolumn{2}{c}{\textbf{SWE-Bench-Pro}}\\
\end{tabular}
\caption{\textbf{Composition of activities is a model signature}. Share of three activities (source editing, scratch testing, suite testing) vs normalized cycle position of agent trajectories between $0$ and $1$, averaged across full benchmark instances (5 runs per instance). For SWE-Bench-Verified (left), Opus~4.6 scratch-tests from early on, edits peak mid-trajectory, and suite testing dominates towards the end; Gemini~3.1~Pro instead scratch-tests at a steady, high rate for almost the whole trajectory, with a low edit share and suite testing only in the final stretch. On SWE-Bench-Pro (right), where tasks need more lines of code, both models shift towards editing and the scratch-testing share falls, Gemini's most sharply. Remaining $19$ models in
Appendix~\ref{app:metrics:solution-distance}
(Figures~\ref{fig:metrics-sbv-editread}--\ref{fig:metrics-sbv-editread-b},
\ref{fig:metrics-sbp-editread}--\ref{fig:metrics-sbp-editread-b}).
}
\label{fig:results-editread-slice}
\end{figure}

\begin{figure}[h]
\centering
\setlength{\tabcolsep}{2pt}
\renewcommand{\arraystretch}{0.6}
\begin{tabular}{@{}m{1.2em}@{\hspace{3pt}}m{0.40\linewidth}@{\hspace{3pt}}m{0.40\linewidth}@{}}
\rotatebox{90}{\small\textbf{SWE-Bench-Verified}} &
\includegraphics[width=0.9\linewidth]{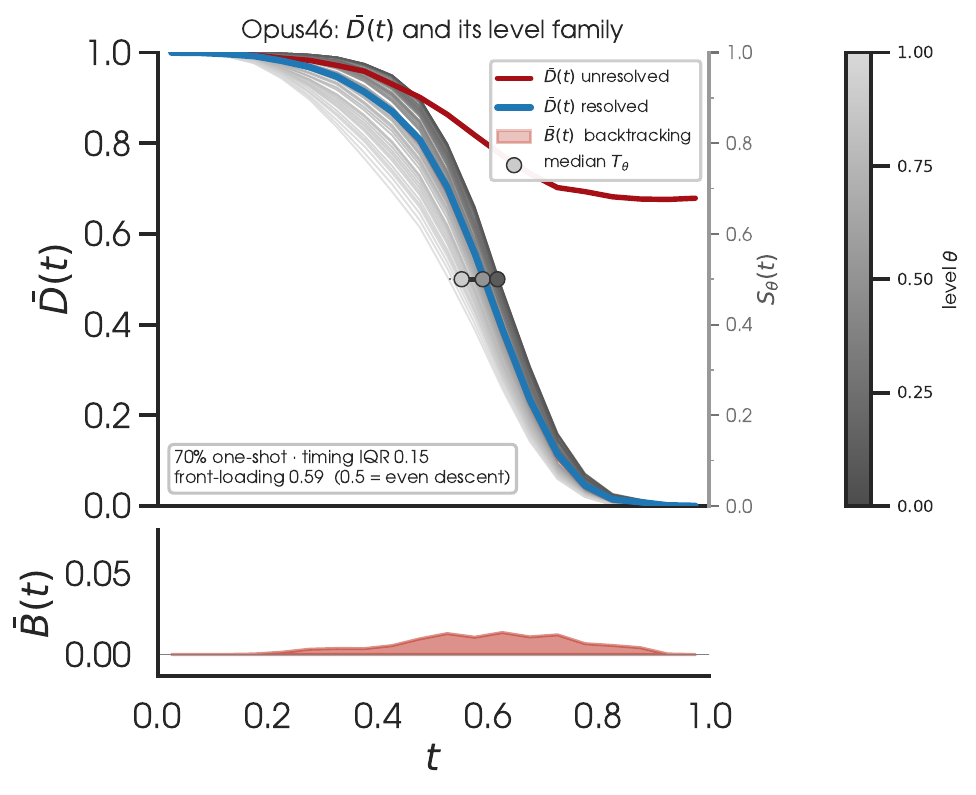} &
\includegraphics[width=0.9\linewidth]{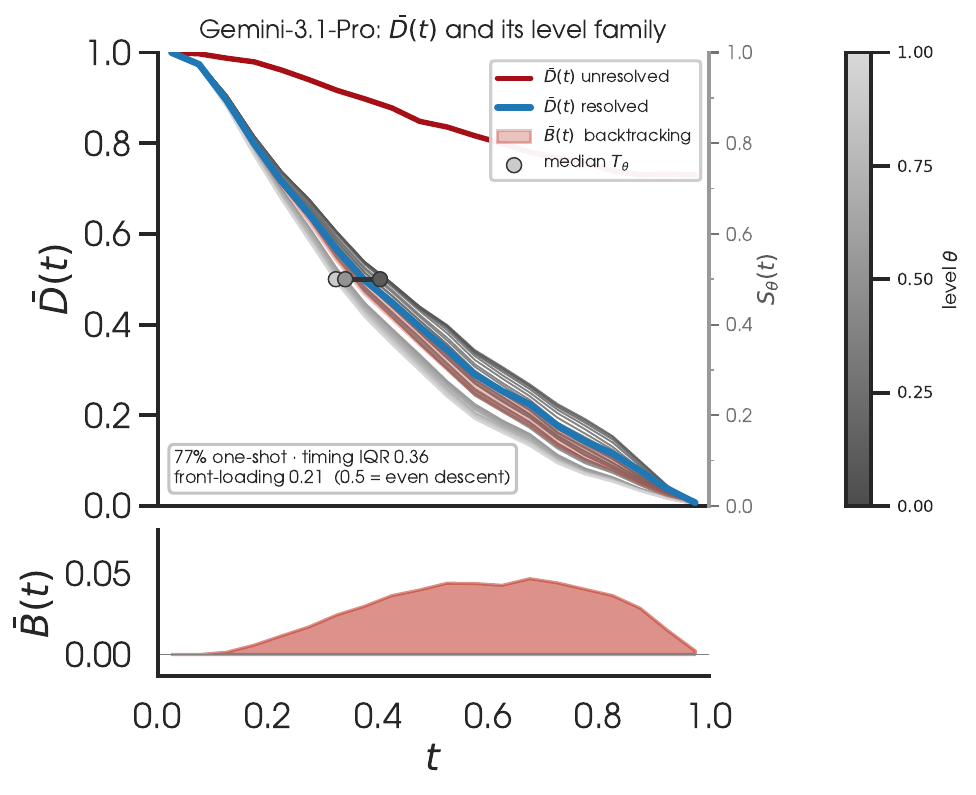} \\
\rotatebox{90}{\small\textbf{Deep-SWE}} &
\includegraphics[width=0.9\linewidth]{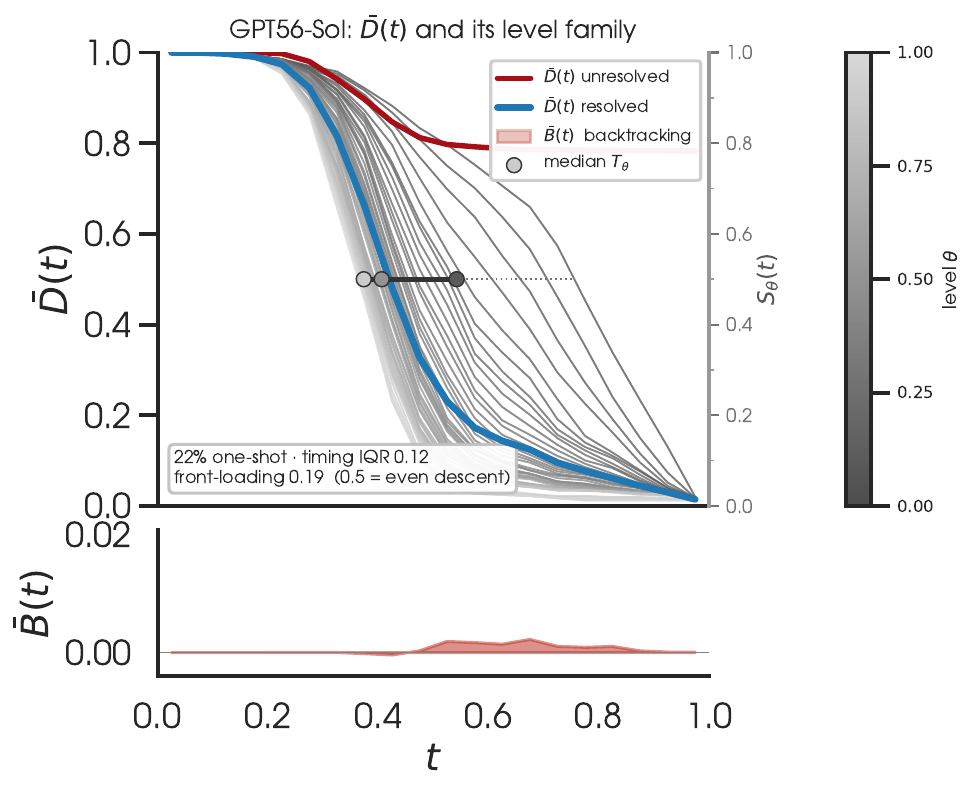} &
\includegraphics[width=0.9\linewidth]{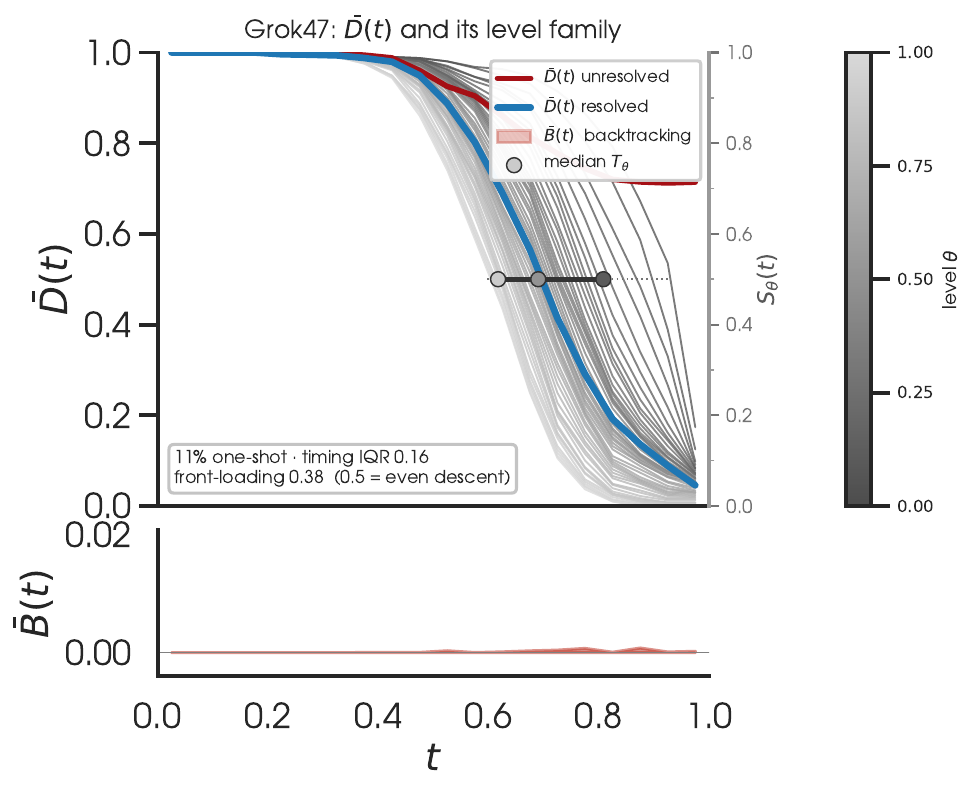} \\
\end{tabular}
\caption{\textbf{Level families separate shape, timing and backtracking.} The level families of Section~\ref{sec:distance} for two models on
SWE-Bench-Verified (top) and two on Deep-SWE (bottom); blue is the mean $\bar D(t)$ of the resolved runs, dark red that of the unresolved runs. On SWE-Bench-Verified
both Opus~4.6 and Gemini~3.1~Pro write the patch in a single edit on most runs (70\% and 77\% one-shot, respectively), so their mean curves
differ not in shape but in timing and backtracking: Gemini~3.1~Pro's solve
times are more than twice as dispersed (timing IQR, the interquartile range of the runs' half-way times $T_r(0.5)$: $0.36$ vs.\ $0.15$) and its
$\bar B(t)$ peaks three times higher ($0.047$ vs.\ $0.014$), which is what makes its
mean descent smooth and near-linear. On Deep-SWE, GPT-5.6-Sol and Grok~4.7 are with backtracking near-zero, and what separates them is \emph{when} progress happens. GPT-5.6-Sol closes half the distance by $T_{0.5}=0.41$; Grok~4.7 stays near $D=1$ until $t\approx0.6$ and then falls steeply, reaching $T_{0.5}=0.69$. The front-loading index in each subplot is $(T_{0.5}-T_{0.9})/(T_{0.1}-T_{0.9})$, 
with $T_\theta$ the median over runs of the first-entry time $T_r(\theta)$ of Section~\ref{sec:distance} (where $S_\theta$ crosses $1/2$): $0.5$ is an even descent, lower means most of the distance falls early. Level families for all models are in
Appendix~\ref{app:metrics:solution-distance}.}
\label{fig:results-soldist-slice}
\end{figure}

\textbf{Trajectories taken by different models can display very diverse behavior even when they have similar resolve rates}. 
On both SWE-Bench-Verified and SWE-Bench-Pro, the trajectory metrics (edits/testing share across agent steps) 
for Opus~4.6 and Gemini~3.1~Pro look very different (Figure~\ref{fig:results-editread-slice}) even though 
their pass@1 on SWE-Bench-Verified is identical and close on SWE-Bench-Pro. 
The two treat testing differently: Opus~4.6 ramps up suite testing in the second half and devotes most of its late-trajectory calls to it, whereas Gemini~3.1~Pro relies on scratch tests throughout and turns to the suite only at the end. For both models, we see a higher share of edits in SWE-Bench-Pro compared to SWE-Bench-Verified owing to the higher difficulty of tasks in the former.


\textbf{Backtracking is decoupled from solve rate.} Figure~\ref{fig:results-soldist-slice}, on SWE-Bench-Verified (top row), Opus~4.6 and Gemini~3.1~Pro post the same pass@1 but their behavior in code state-space is significantly diverse. Gemini~3.1~Pro has $3\times$ higher backtracking than Opus~4.6; Appendix Figure~\ref{fig:solution-distance-pingpong-astropy} traces one such Gemini~3.1~Pro backtrack, a deliberate revert that uncovers a second bug before the run resolves. The distance curves $\bar D(t)$ follows different shapes to reach the solution space with Opus~4.6 more flat and then steep descent while Gemini~3.1~Pro has more steady decay over the normalized time. The level curves $S_\theta(t)$ are tighter in spread due to dataset requiring atomic edits and mostly single-file. Next, on the more complicated Deep-SWE dataset, GPT-5.6-Sol and Grok~4.7 also post the same pass@1 and both have backtracking terms near zero, yet their level families occupy different parts of the trajectory. GPT-5.6-Sol closes most of the distance in a short burst before mid-run, while Grok~4.7 makes almost no progress until the final third and then closes it in a short, steep drop. It is also coherent from edit/testing share plots in Appendix~\ref{app:metrics:deep} that edits land later for Grok~4.7, hence distance drops later in time. Overall, we find that neither backtracking nor timing is recoverable from pass@1; both reflect model-internal behavior on the way to the same success. Per-model backtracking rates for every model and benchmark are given in Appendix~\ref{app:metrics:sbv}.

\textbf{Different model-families execute same tasks with different phase schedules.} 
Appendix Figures~\ref{fig:metrics-sbv-phase}--\ref{fig:metrics-sbv-phase-b} split each trajectory into explore/localize/implement/verify phases for every model. The implementation phase peaks at different positions and widths,
and the verify phase carries different long-tails. Note that the sharp drop in `explore' in Opus~4.6, GPT-5.4, and Qwen3 coder 480B is almost always due to doing an initial workspace-tree survey such as with \texttt{ls -l} tool-calls.

\textbf{Terminal-Bench-2 scores are sensitive to evaluation infrastructure.}
Beyond model quality, Terminal-Bench-2 results also reflect the stability of the
serving and execution stack: inference-backend throughput, node packing,
command batching, and the benchmark's own memory and time budgets each move
pass@1 by $5$--$10$ points. We analyse these effects in Appendix~\ref{app:tb2-infra}.

\section{Conclusion}
\label{sec:conclusion}

As frontier models grow more capable, the harness is critical in determining whether that capability is translated into agent 
performance or lost in the interface between model and environment. A good harness (i) applies the principles that help every model family, such
as line-anchored edits, explicit post-edit diffs, and faithful tool-output
serialization, and (ii) adapts to the model's preferences where model-families genuinely differ. Our experiments show that minimizing the 
bidirectional intent--execution gap is enough to bring the models we test to or above their reported model card numbers, within confidence intervals, in 61 of 66 cases on SWE-Bench-Pro, 
SWE-Bench-Verified, Terminal-Bench-2 and Deep-SWE with no task-specific tuning; 4 of the remaining 5 are on SWE-Bench-Pro, whose public containers carry a leakage artifact (Section~\ref{sec:results:headline}). The gap between documented and reproduced agent performance 
is therefore in large part harness friction rather than any fundamental limitation of a simple ReACT~\citep{yao2023react} based single-agent design.
And because no tool surface is uniformly best, the goal is to have a harness that holds execution invariants fixed while meeting each model-family
where its training left it.

We see that as models converge near the
top of these benchmarks, pass@1 stops separating them but their trajectories
do not. Our code state-space analysis (solution-distance, backtracking) and phase decomposition show that models with the same solve rate can differ in the path taken by their corresponding agents in code-state space.
Going beyond accuracy numbers and understanding trajectory behavior is what makes model-harness co-design possible. 

\subsubsection*{Acknowledgements}
We thank Murali Krishna Ramanathan, Narayanan Sadagopan, and Varun Kumar for providing
their feedback on this work, and Gouri Pandeshwar for resolving an
infrastructure issue.

We acknowledge the SWE-agent project for agent-computer interface designs.
Their SWE-Bench prompt templates served as starting points for our SWE
prompts~\citep{yang2024sweagent}. We also acknowledge the OpenAI Cookbook for
public \texttt{apply\_patch} descriptions and examples. These informed our
GPT-family patching interface~\citep{openai2025cookbookapplypatch}.


\subsection*{AI use statement}

In this work, we used generative AI tools as writing and editing assistants. Specifically,
(a) to polish the wording of the introduction, (b) developing figures and graphics in tables which
the authors then reviewed and edited. AI was also used to help collect, format and check
bibliographic references. We have not used generative AI
tools to generate research ideas, to design experiments, or to produce or
alter experimental results. All four benchmark runs, trajectory measurements and solution distance analyses
reported here were carried out with the authors' harness and evaluation
pipeline.
We have reviewed all AI-assisted work and was checked against the
underlying data and edited by the authors. We take responsibility for the final content of this work,
including text, claims or artifacts produced with the aid of generative AI.

\subsection*{Reproducibility statement}

We have aimed to make every number in this paper traceable to a released artifact or a 
documented procedure. SSA source code, together with the exact configuration files, prompts 
and tool specifications used for every run, will be released publicly. Appendix~\ref{sec:appendix-configs} 
lists, per benchmark, the compute-setup, inference and reasoning settings, prompts and agent tools 
for every model. Appendix~\ref{app:prompts} specifies all system and user prompts. Appendix~\ref{sec:appendix-tools} 
gives the complete tool-schemas. Appendix~\ref{app:vllm} lists the exact vLLM serve command for 
every open-weights model. Finally, Appendix~\ref{app:gptoss} describes the one harness-side parsing 
fix we applied.

Appendix~\ref{app:passrates} reports every model's pass@1 with its $95\%$ confidence interval, 
pass@2, and the published figure.

Tool-call labels come from a Claude Opus~4.6 judge whose windowed, 
 forced-tool-output classification pipeline is specified in Appendix~\ref{app:metrics:judge} 
 and whose fixed rubric, with every label value defined, is given in 
 Table~\ref{tab:judge-buckets} (Appendix~\ref{app:metrics:rubric}).

\nocite{yang2024sweagent,wang2024openhands,guo2025deepseekr1,anthropic2025claude47,deepmind2025gemini3,xai2025grok42,strandsagents,kim2026codestruct,liu2024deepseekv3,hui2024qwen25coder,minimax2025m2,amazon2024nova,brown2020gpt3,wei2022emergent,kojima2022zerocot,shinn2023reflexion,wang2024agentsurvey,xi2023agentsurvey,qin2024toolllm,wu2023autogen,xia2024agentless,jain2024livecodebench,sainz2023contamination,magar2022contamination,zheng2023judging,anthropic2024agents,liu2024agentbench,zhou2023webarena,xie2024osworld,wang2024codeact,zhang2024autocoderover,liu2024repobench,yang2024swebenchmultimodal,rashid2025swepolybench}

\clearpage
\bibliographystyle{iclr2027_conference}
\bibliography{references}

\appendix
\clearpage
\section*{Appendix Contents}
\addcontentsline{toc}{section}{Appendix Contents}

The appendix is organized into the nine parts below. All entries are
hyperlinked; the number on the right is the page on which the
(sub)section begins.

\newcommand{\appidxA}[2]{\item \hyperref[#1]{\textbf{\ref*{#1}\quad #2}}\dotfill \textbf{\pageref{#1}}}
\newcommand{\appidxB}[2]{\item \hyperref[#1]{\ref*{#1}\quad #2}\dotfill \pageref{#1}}

\begingroup
\begin{itemize}[label={},leftmargin=1.4em,itemsep=2pt,topsep=2pt,parsep=0pt]
  \appidxA{app:architecture}{SSA Architecture}
  \appidxA{app:metrics}{Methodology and Metrics}
  \begin{itemize}[label={},leftmargin=2.0em,itemsep=1pt,topsep=1pt,parsep=0pt]
    \appidxB{app:metrics:methodology}{Methodology}
    \begin{itemize}[label={},leftmargin=2.4em,itemsep=0pt,topsep=0pt,parsep=0pt]
      \appidxB{app:metrics:judge}{LLM judge for tool-call classification}
      \appidxB{app:metrics:rubric}{Classification rubric}
      \appidxB{app:metrics:solution-distance}{Solution-distance methodology details}
      \appidxB{app:metrics:soldist:examples}{Illustrative examples}
      \appidxB{app:metrics:soldist:replay}{Faithful state replay}
      \appidxB{app:metrics:soldist:zero}{Resolved endpoints: self-anchor}
      \appidxB{app:metrics:soldist:backtrack}{Backtracking quantification}
      \appidxB{app:metrics:soldist:limits}{Limitations of the solution-distance metric}
    \end{itemize}
    \appidxB{app:metrics:metrics}{Metrics}
    \appidxB{app:metrics:sbv}{Metrics: SWE-Bench-Verified}
    \appidxB{app:metrics:sbp}{Metrics: SWE-Bench-Pro}
    \appidxB{app:metrics:tb2}{Metrics: Terminal-Bench-2}
    \appidxB{app:metrics:deep}{Metrics: Deep-SWE}
  \end{itemize}
  \appidxA{app:passrates}{Per-Model Pass Rates by Family}
  \appidxA{app:tb2-infra}{Terminal-Bench-2: Infrastructure Sensitivity}
  \appidxA{app:eval-integrity}{Evaluation Integrity}
  \appidxA{app:swe-pro-git-leakage}{SWE-Bench-Pro Git Leakage}
  \begin{itemize}[label={},leftmargin=2.0em,itemsep=1pt,topsep=1pt,parsep=0pt]
    \appidxB{app:swe-pro-git-leakage-protocol}{Sanitized evaluation}
    \appidxB{app:swe-pro-git-leakage-proxy}{Direct leakage metric}
    \appidxB{app:swe-pro-git-leakage-impact}{Empirical impact}
    \appidxB{app:swe-pro-git-leakage-opensource}{Open-source models and explicit git instructions}
  \end{itemize}
  \appidxA{sec:appendix-configs}{Configurations}
  \begin{itemize}[label={},leftmargin=2.0em,itemsep=1pt,topsep=1pt,parsep=0pt]
    \appidxB{app:configs:sbv}{SWE-Bench-Verified}
    \appidxB{app:configs:sbp}{SWE-Bench-Pro}
    \appidxB{app:configs:tb2}{Terminal-Bench-2}
    \appidxB{app:configs:deepswe}{Deep-SWE}
  \end{itemize}
  \appidxA{app:prompts}{Prompts}
  \begin{itemize}[label={},leftmargin=2.0em,itemsep=1pt,topsep=1pt,parsep=0pt]
    \appidxB{app:prompts:swe-generic}{Generic SWE prompts}
    \appidxB{app:prompts:gpt}{Git PR prompt with GPT-specific instructions}
    \appidxB{app:prompts:anthropic}{Claude specific prompt with reasoning/tool-call nudge}
    \appidxB{app:prompts:tb2}{Terminal-Bench-2 prompts}
    \appidxB{app:prompts:deepswe}{Deep-SWE prompts}
  \end{itemize}
  \appidxA{sec:appendix-tools}{Tool Definitions}
  \begin{itemize}[label={},leftmargin=2.0em,itemsep=1pt,topsep=1pt,parsep=0pt]
    \appidxB{app:tools:generic}{Generic shell and editor tools}
    \appidxB{app:tools:openai}{OpenAI-specific tools}
    \appidxB{app:tools:xai}{xAI-specific tools}
  \end{itemize}
  \appidxA{app:vllm}{vLLM Serving Configurations}
  \begin{itemize}[label={},leftmargin=2.0em,itemsep=1pt,topsep=1pt,parsep=0pt]
    \appidxB{app:vllm:serve}{Per-model serve commands}
    \appidxB{app:vllm:choices}{A note on vLLM flags}
  \end{itemize}
  \appidxA{app:gptoss}{Harmony-format parsing failures in the harness}
  \begin{itemize}[label={},leftmargin=2.0em,itemsep=1pt,topsep=1pt,parsep=0pt]
    \appidxB{app:gptoss:symptom}{The symptom: empty and malformed tool calls}
    \appidxB{app:gptoss:boundary}{The serving boundary: stateless re-tokenization}
    \appidxB{app:gptoss:observed}{Token-level evidence}
    \appidxB{app:gptoss:patches}{The fix: surface a single retryable error}
  \end{itemize}
\end{itemize}
\endgroup

\clearpage
\section{SSA Architecture}
\label{app:architecture}

We use Simple Strands Agent (SSA) as the harness for all our experiments (see Figure~\ref{fig:ssa-architecture} for architecture). 
It has the following components: tasks, prompts, models, tools, environments, and hooks.
The input task is converted into prompts, with the agent alternating between model calls and tool execution.
Tools act on an isolated environment (we use Docker for all our experiments) and hooks enforce control policies such as context management and content filtering.

\begin{figure}[!h]
\centering
\resizebox{\textwidth}{!}{%
\begin{tikzpicture}[
  every node/.style={inner sep=3.5pt, outer sep=0pt},
  comp/.style={draw=black!55, line width=0.5pt, rounded corners=2.5pt, align=center,
               font=\footnotesize, minimum height=1.05cm, text width=2.6cm, fill=white},
  cfgbox/.style={comp, fill=orange!8, draw=orange!55!black, text width=2.55cm},
  llmbox/.style={comp, fill=blue!9, draw=blue!55!black, text width=2.65cm},
  toolc/.style={comp, fill=green!9, draw=green!50!black, text width=2.9cm},
  cmgrbox/.style={comp, fill=violet!8, draw=violet!55!black, text width=2.9cm},
  envbox/.style={comp, fill=gray!8, draw=black!55, text width=9.1cm, minimum height=1.05cm},
  hookbox/.style={comp, fill=red!6, draw=red!55!black, text width=3.0cm},
  termbox/.style={comp, fill=yellow!12, draw=yellow!55!black, text width=2.9cm},
  corebox/.style={draw=blue!50!black, dashed, line width=0.7pt, rounded corners=4pt, fill=blue!2},
  flow/.style={-{Stealth[length=2.2mm]}, line width=0.5pt, black!70, shorten <=1pt, shorten >=1pt},
  loopflow/.style={-{Stealth[length=2.4mm]}, line width=0.8pt, blue!55!black, shorten <=1pt, shorten >=1pt},
  hookflow/.style={-{Stealth[length=2.0mm]}, line width=0.5pt, dashed, red!55!black, shorten <=1pt},
  flab/.style={font=\scriptsize\sffamily, text=black!70, fill=white, inner sep=1.4pt, align=center},
]

\node[cfgbox, sharp corners, fill=yellow!12, draw=yellow!55!black] (task) at (1.5, 8.25) {\textbf{Input/User task}};
\node[cfgbox] (pg)  at (1.5, 6.2) {\textbf{PromptGenerator}\\[1pt]{\scriptsize Jinja2 templates}};
\node[cfgbox] (mdl) at (1.5, 4.0) {\textbf{Model adapters}\\[1pt]{\scriptsize Bedrock $\cdot$ Anthropic\\OpenAI $\cdot$ Gemini $\cdot$ xAI\\LiteLLM / vLLM}};

\draw[flow] (task) -- (pg);

\node[llmbox]  (llm)   at (6.9, 7.7) {\textbf{Model call}\\[1pt]{\scriptsize stream assistant\\+ \texttt{tool\_use}}};
\node[toolc]   (tools) at (10.9, 7.7) {\textbf{Tool dispatch}\\[1pt]{\scriptsize \texttt{bash} $\cdot$ \texttt{editor}\\\texttt{view} $\cdot$ \texttt{search}}};
\node[cmgrbox] (cmgr)  at (6.9, 5.0) {\textbf{Conversation mgr}\\[1pt]{\scriptsize adaptive sliding\\window + truncation}};

\begin{scope}[on background layer]
  \node[corebox, fit=(llm)(tools)(cmgr), inner sep=16pt] (core) {};
\end{scope}
\node[anchor=north west, font=\scriptsize\sffamily\bfseries, text=blue!45!black]
  at ([xshift=2pt,yshift=-2pt]core.north west) {SimpleStrandAgent $\cdot$ event loop};

\node[envbox] (env) at (8.7, 1.8)
  {\textbf{Execution environment}\\[1pt]{\scriptsize Docker sandbox \emph{or} local\\streaming \texttt{bash} exec $\cdot$ file I/O $\cdot$ internet gate}};

\node[hookbox] (hooks) at (15.2, 7.8)
  {\textbf{Hooks}\\[1pt]{\scriptsize Content $\cdot$ Trajectory\\PromptCache $\cdot$ ContextWindow\\EventLoopLimiter\\DropUnknownTool}};
\node[termbox, sharp corners] (term) at (15.2, 4.2)
  {\textbf{Output/Terminate}\\[1pt]{\scriptsize environment final state\\diff patch}};

\draw[flow] (pg.east) -- ++(1.0,0) |- (llm.west);
\node[flab, anchor=south west] at ([xshift=0.18cm,yshift=0.6cm]pg.east) {system + user\\prompt};
\draw[flow] (mdl.north east) to[out=32,in=220]
  node[flab, below, pos=0.46]{model backend} (llm.south west);

\draw[loopflow] (llm.east) -- node[flab, above, yshift=0.08cm]{\texttt{tool\_use}} (tools.west);
\draw[loopflow] (tools.south) -- node[flab, anchor=west, xshift=0.18cm, pos=0.58]{execute} (tools.south |- env.north);
\draw[loopflow] (env.north -| cmgr.south) -- node[flab, anchor=east, xshift=-0.18cm, pos=0.45]{tool result} (cmgr.south);
\draw[loopflow] (cmgr.north) -- node[flab, anchor=west, xshift=0.18cm, pos=0.5]{next cycle} (llm.south);

\draw[hookflow] (hooks.west) -- (core.east);

\draw[flow] (env.east) -| (term.south);

\end{tikzpicture}%
}
\caption{\textbf{SSA architecture}. Tasks and model adapters feed a cyclic
\texttt{SimpleStrandAgent} loop. The loop streams model output, dispatches tools,
executes them in the environment, and appends tool results to conversation
state. Hooks validate bounds and record the loop; termination extracts
the final environment state and diff patch.}
\label{fig:ssa-architecture}
\end{figure}

\textbf{Task and prompts.} The task to be solved is converted to a user prompt and paired with system prompts that may include model-specific instructions (see Appendix\,\ref{app:prompts} for complete set of SSA prompts). Model specific changes may include reasoning nudge (see Section\,\ref{ssec:reasoning_nudge}) and tool-specific instructions such as \texttt{apply\_patch} for GPT family.

\textbf{Model interface.} Adapters map model-provider APIs into a common event interface: assistant text,
reasoning content, tool-use requests, stop reasons, usage, cache metadata, and
recoverable errors. The event loop is the same across all models.

\textbf{Agent core and conversation.} The core is a thin control loop: call the model, execute requested tools, append tool results, and repeat. The conversation manager preserves valid
tool-use/tool-result structure while compressing large outputs or trimming old
turns when context gets too long.

\textbf{Tools and environment.} SSA's tool surface is deliberately small as it mainly relies on bash and file viewing/editing tools. Tools delegate command execution to the environment. For a complete list of tools, see Appendix\,\ref{sec:appendix-tools}.

\textbf{Hooks and instrumentation.} Hooks validate model output, drop malformed tool calls, manage context-windows, enforce loop bounds, add cache markers, and record trajectories and metrics for analysis.

\clearpage
\section{Methodology and Metrics}
\label{app:metrics}

All quantitative results in Section~\ref{sec:results} are computed from two artifacts produced by every SSA evaluation. (1) A \texttt{metrics.json} with the per-cycle token, latency, cache and tool-success counters emitted by the harness, and (2) \texttt{trajectory.json} containing every user / assistant message (with reasoning blocks, tool calls and tool results) verbatim. The trajectory is the authoritative input to all behavioral analyses, whenever counters in \texttt{metrics.json} disagree with the trajectory (e.g.\ retry due to exception handling) the trajectory wins.

\subsection{Methodology}
\label{app:metrics:methodology}

All behavioral analyses are driven by a per-call LLM judge~\citep{zheng2023judging} that classifies every tool invocation against a fixed rubric. This subsection describes the judge pipeline, the rubric it operates on, and the system prompt under which it runs.

\subsubsection{LLM judge for tool-call classification}
\label{app:metrics:judge}

Every individual tool call in the trajectory is classified against a fixed rubric by a Claude Opus 4.6 judge \citep{anthropic2025claude47}. The pipeline works as follows:

\begin{enumerate}[leftmargin=*, itemsep=2pt, topsep=2pt]
  \item \textbf{Windowed classification.} Each trajectory is split into sliding windows of LLM calls. The judge sees the window rendered as a transcript (preceding events tagged \texttt{[context]}, and the callable event tagged with \texttt{<{<}CALL\ n{>}>}) and LLM labels the provided event in the window in a single response. Local context lets the judge disambiguate repeated commands by surrounding state.
  \item \textbf{Forced tool-use output.} The judge cannot reply with free-form text. Its only allowed output is a single call to the \texttt{submit\_classifications} tool whose schema only accepts an \texttt{enum}. Outputs are validated client-side before acceptance.
\end{enumerate}

\subsubsection{Classification rubric}
\label{app:metrics:rubric}

Table~\ref{tab:judge-buckets} lists every field the judge assigns to a call. R1--R5 carry the bulk of the behavioural signal, R6 supplies a coarse-phase label used in Gantt-style trajectory plots, R7 is a reliability self-report used to flag low-confidence buckets in downstream analyses, and R8 disambiguates the raw \texttt{error} status, because a non-zero \texttt{bash} exit (\texttt{grep} returning ``no match'', \texttt{pytest} reporting test failures, etc.)\ is often not a tool malfunction.

\definecolor{rubricheader}{RGB}{200, 220, 240}
\begin{longtable}{@{}>{\raggedright\arraybackslash}p{3.4cm}>{\raggedright\arraybackslash}p{10.1cm}@{}}
\caption{Rubrics the LLM judge assigns to every tool call, with per-value definitions. R5 (\texttt{target\_paths}) is the only open-valued field; the rest are picked from the listed enums. Coloured rows are field separators.}\label{tab:judge-buckets}\\
\toprule
Value & Description \\
\midrule
\endfirsthead
\multicolumn{2}{c}{\tablename\ \thetable{} -- continued}\\
\toprule
Value & Description \\
\midrule
\endhead
\midrule\multicolumn{2}{r}{continued on next page}\\
\endfoot
\bottomrule
\endlastfoot
\rowcolor{rubricheader}[0pt][0pt]\multicolumn{2}{@{}l@{}}{\textbf{R1 category}} \\
\texttt{explore\_navigate} & Orient in the repo: list / inspect structure, history, or state (\texttt{ls}, \texttt{tree}, \texttt{view} a directory, \texttt{git log/status}, \texttt{find -type d}). \\
\texttt{search\_locate} & Hunt for a symbol, string, or file across the tree (\texttt{grep -rn}, \texttt{rg}, \texttt{find -name}). \\
\texttt{read\_code} & Read the contents of a specific source / test / config file (\texttt{cat}, \texttt{head}, \texttt{view <file>}, \texttt{sed -n '1,80p'}). \\
\texttt{edit\_source} & Modify non-test project source code (\texttt{str\_replace} on \texttt{<file>.py}, \texttt{sed -i} on a source file, \texttt{apply\_patch} updating a source file). \\
\texttt{edit\_test} & Modify the project's own test files (under \texttt{tests/}, files matching \texttt{test\_*.py}). \\
\texttt{write\_scratch} & Create or modify an agent-authored throwaway repro / debug script (e.g.\ \texttt{cat > /testbed/reproduce.py}, \texttt{create /testbed/check.py}). \\
\texttt{run\_suite} & Run the project's real test suite (\texttt{pytest <project tests>}, \texttt{python -m unittest}, \texttt{./runtests.py}, \texttt{tox}). \\
\texttt{run\_scratch} & Execute a throwaway repro or probe (\texttt{python reproduce.py}, \texttt{python -c "<repro snippet>"}). \\
\texttt{inspect\_runtime} & Quick non-repro runtime probe (\texttt{python -c "import x; print(x.\_\_version\_\_)"}, \texttt{pip show}, \texttt{--version}). \\
\texttt{vcs} & Version-control mutation or inspection beyond log/status (\texttt{git diff}, \texttt{git apply}, \texttt{git stash}, \texttt{git checkout -- f}, \texttt{git add}). \\
\texttt{env\_setup} & Install / build / configure the environment (\texttt{pip install -e .}, \texttt{conda ...}, \texttt{python setup.py build\_ext}, \texttt{export}). \\
\texttt{other} & None of the above or undeterminable (miscellaneous shell plumbing). \\
\midrule
\rowcolor{rubricheader}[0pt][0pt]\multicolumn{2}{@{}l@{}}{\textbf{R2 target\_kind}} \\
\texttt{project\_source} & Real, non-test source code shipped by the project (under the package tree, not under \texttt{tests/}). \\
\texttt{project\_test} & The project's own test files (e.g.\ \texttt{astropy/table/tests/test\_*.py}). \\
\texttt{scratch} & Agent-authored throwaway file --- repro / verify / debug --- including \texttt{/tmp/*} and files at \texttt{/testbed/} root that don't belong to the package. \\
\texttt{build\_config} & Build, packaging, or CI config: \texttt{setup.py}, \texttt{pyproject.toml}, \texttt{conftest.py}, CI YAML. \\
\texttt{vcs\_internal} & VCS metadata only (the \texttt{.git} directory). \\
\texttt{none} & No file target (e.g.\ \texttt{python --version}, \texttt{cd}, \texttt{echo}). \\
\midrule
\rowcolor{rubricheader}[0pt][0pt]\multicolumn{2}{@{}l@{}}{\textbf{R3 mutation}} \\
\texttt{read\_only} & Pure read or inspection; no filesystem change. \\
\texttt{mutates\_source} & Writes / changes project source or test files. \\
\texttt{mutates\_scratch} & Writes / changes only scratch / throwaway files. \\
\texttt{destructive} & Broad delete / reset / overwrite: \texttt{rm -rf}, \texttt{git reset --hard}, \texttt{git checkout .}, \texttt{apply\_patch *** Delete File}. \\
\midrule
\rowcolor{rubricheader}[0pt][0pt]\multicolumn{2}{@{}l@{}}{\textbf{R4 test\_activity}} \\
\texttt{suite} & Runs or edits the project's real test suite. \\
\texttt{scratch} & Runs or writes a throwaway repro / verification script. \\
\texttt{none} & Not a testing action. \\
\midrule
\rowcolor{rubricheader}[0pt][0pt]\multicolumn{2}{@{}l@{}}{\textbf{R5 target\_paths}} \\
\textit{list} & Concrete file path(s) the call reads or writes. Resolves a leading \texttt{cd /testbed \&\&} into absolute paths where unambiguous; otherwise the path as written. \texttt{[]} when no file target. Deduplicated within a single call. \\
\midrule
\rowcolor{rubricheader}[0pt][0pt]\multicolumn{2}{@{}l@{}}{\textbf{R6 phase}} \\
\texttt{explore} & Understanding the repo or reproducing the bug. \\
\texttt{localize} & Narrowing in on the offending code. \\
\texttt{implement} & Writing the fix. \\
\texttt{verify} & Running tests or otherwise confirming the fix. \\
\texttt{other} & Setup, cleanup, or otherwise undeterminable. \\
\midrule
\rowcolor{rubricheader}[0pt][0pt]\multicolumn{2}{@{}l@{}}{\textbf{R7 confidence}} \\
\texttt{high} & Judge is highly confident in R1 and R2 for this call. \\
\texttt{medium} & Judge has moderate confidence; downstream code may down-weight or audit these. \\
\texttt{low} & Judge is unsure; flagged for review and excluded from precision-sensitive aggregates. \\
\midrule
\rowcolor{rubricheader}[0pt][0pt]\multicolumn{2}{@{}l@{}}{\textbf{R8 error\_kind}} \\
\texttt{none} & Call did not error (result status was success or unknown). Forced to \texttt{none} when result status is not \texttt{error}. \\
\texttt{tool\_invocation} & The tool (or the CLI it dispatched) could not process the call --- the failure happens \emph{before} the requested operation runs: \texttt{command not found}, shell syntax error, \texttt{No such file or directory}, permission denied, a subprocess rejecting its CLI arguments (\texttt{unrecognized argument}, argparse errors), an inline \texttt{python -c} \texttt{SyntaxError}, or a \texttt{str\_replace\_editor} / \texttt{apply\_patch} failure (target/\texttt{old\_str} not found, patch context mismatch). \\
\texttt{subprocess\_runtime} & The tool accepted and dispatched the call and the subprocess then failed mid-execution --- it parsed fine but the program inside threw: a Python \texttt{Traceback} from a repro script, a runtime \texttt{ImportError} / \texttt{AttributeError}, a \texttt{conftest} / fixture crash while loading tests, or a \texttt{pip install} whose build step failed. \\
\texttt{benign} & Expected non-zero exit that is not a tool malfunction: \texttt{grep} / \texttt{rg} no-match (exit 1), \texttt{pytest} reporting test failures or "no tests ran" (the run executed fine), \texttt{[ -f x ]} test conditional, \texttt{git diff --exit-code} / \texttt{--quiet} used as a check, a pre-fix repro deliberately demonstrated to fail. \\
\texttt{uncertain} & Status is \texttt{error} but the output is empty or truncated and the judge cannot tell which bucket. \\
\end{longtable}

The LLM judge rubrics obtained from Table\,\ref{tab:judge-buckets} are used to inform numerous metrics from the agent trajectories. The tool-call distribution (Figures\,\ref{fig:metrics-sbv-tooldist}-\ref{fig:metrics-sbv-tooldist-b}, \ref{fig:metrics-sbp-tooldist}-\ref{fig:metrics-sbp-tooldist-b}, \ref{fig:metrics-tb2-tooldist}-\ref{fig:metrics-tb2-tooldist-b}) and bash sub-command taxonomy come from R1, the edit-share (Figures\,\ref{fig:metrics-sbv-editread}-\ref{fig:metrics-sbv-editread-b}, \ref{fig:metrics-sbp-editread}-\ref{fig:metrics-sbp-editread-b}, \ref{fig:metrics-tb2-editread}-\ref{fig:metrics-tb2-editread-b}) from R3, the scratch-vs-suite testing share (Figures\,\ref{fig:metrics-sbv-editread}-\ref{fig:metrics-sbv-editread-b}, \ref{fig:metrics-sbp-editread}-\ref{fig:metrics-sbp-editread-b}, \ref{fig:metrics-tb2-editread}-\ref{fig:metrics-tb2-editread-b}), the explore$\to$localize$\to$implement$\to$verify Gantt plots (Figures\,\ref{fig:metrics-sbv-phase}-\ref{fig:metrics-sbv-phase-b}, \ref{fig:metrics-sbp-phase}-\ref{fig:metrics-sbp-phase-b}, \ref{fig:metrics-tb2-phase}-\ref{fig:metrics-tb2-phase-b}) from R6, and the tool-error rate (Figures\,\ref{fig:metrics-sbv-toolerr}-\ref{fig:metrics-sbv-toolerr-b}, \ref{fig:metrics-sbp-toolerr}-\ref{fig:metrics-sbp-toolerr-b}, \ref{fig:metrics-tb2-toolerr}-\ref{fig:metrics-tb2-toolerr-b}) from R8. The union of the two real-failure buckets (\texttt{tool\_invocation} and \texttt{subprocess\_runtime}) is substantially lower than the raw \texttt{tool\_use.error\_count} reported by the harness. Wherever the figures below refer to ``\texttt{R8 = genuine}'' they mean this union, benign and uncertain calls are excluded.

\subsubsection{Solution-distance methodology details}
\label{app:metrics:solution-distance}

Section~\ref{sec:distance} defines the text-level state space and the ideal
solution distance over the oracle-defined solution set $\mathcal{S}_i$. This
section provides the empirical approximation used in the experiments:
patch-feature extraction, reference-set construction, replay
indexing, and endpoint corrections.

\paragraph{Proofs of the level-set results.}
\phantomsection
\label{app:metrics:soldist:proofs}

\begin{proof}[Proof of Theorem~\ref{thm:soldist:cohort-refinement:decomposition}]
As defined in Section\,\ref{sec:soldist:cohort-refinement}, a run has not reached $\theta$ by $t$ exactly when $M_r(t)>\theta$.
Applying this straightforward identity
\begin{equation*}
z=\int_0^1\mathbf1\{z>\theta\}\,d\theta,
\end{equation*}
to each $M_r(t)\in[0,1]$, we get the following
\begin{eqnarray*}
 \int_0^1S_\theta(t)\,d\theta &=&\int_0^1 \frac{1}{N}\sum\nolimits_{r}\mathbf{1}\{M_r(t)>\theta\}d\theta \\
 &=&\frac{1}{N}\sum\nolimits_{r}\int_0^1\mathbf{1}\{M_r(t)>\theta\}d\theta \\
 &=&\frac{1}{N}\sum\nolimits_{r}M_r(t) = \bar M(t).
\end{eqnarray*}
By the definitions of $\bar D$ and $\bar M$,
$\bar D(t)-\bar M(t)=N^{-1}\sum_r[D_r(t)-M_r(t)]=\bar B(t)$.
Moreover, $M_r(t)\le D_r(t)$ for every run, hence $\bar B(t)\ge0$.
\end{proof}

\begin{proof}[Proof of Proposition~\ref{prop:soldist:cohort-refinement:zero-backtracking}]
Each difference $D_r(t)-M_r(t)$ is nonnegative, so their mean is zero
exactly when every difference is zero. If $D_r=M_r$ at all times $t$, then
$D_r$ is non-increasing because $M_r$ is non-increasing. Conversely, a non-increasing
trajectory always equals its running minimum, making backtracking to be zero.
\end{proof}

\paragraph{The empirical solution subset.}
The oracle-correct set $\mathcal{S}_i$ is unobservable, so we approximate it
by a deduplicated empirical subset
$\tilde{\mathcal{S}}_i \subseteq \mathcal{S}_i$ assembled from the runs we
actually have. Each SWE instance (in both SWE-Bench-Verified and SWE-Bench-Pro)
was attempted exactly $105$ times in our sweep ($21$ models $\times$ $5$ runs),
and every resolved run contributes one verified-correct final code state. This provides a 
rich set of data to construct/approximate $\mathcal{S}$. For data instances where no executed run
produced any resolved status, we at least have the provided gold patch. Near-identical states are merged at Jaccard
similarity $\rho=0.9$. Most resolved patches stay distinct, so
$|\tilde{\mathcal{S}}_i|$ should not be read as a count of semantic solution
classes. In our collective runs, on SWE-Bench-Verified, resolved-run references already cover $474$ of $500$ instances
with a mean of $\sim 82$ resolved runs per covered instance, and the remaining $26$
fall back to the gold patch. On SWE-Bench-Pro, the corresponding numbers are $562$ of $731$
instances covered, with a mean of $\sim 60$ resolved runs per covered instance, and
the remaining $169$ instances fall back to the gold patch.

Next, we represent each patch as a set of \emph{patch features}
$\tau(P)=\{(f,s,\ell): f\in\text{files}(P),\,s\in\{+,-\},\,\ell\in
\text{lines}(P)\}$, where $\ell$ is whitespace-stripped line
content, and headers and blank lines are dropped. Features are sets, so repeated
identical changed lines in the same file collapse intentionally. Patches with empty 
feature sets after filtering are excluded. We have used this simple representation in this work, and 
additional approaches such as AST-based~\citep{alon2018general,guo2021graphcodebert}
or semantic similarity~\citep{moraglio2012geometric,silva2025gradientrepair} could be explored further.

Before de-duplication, we filter out agent's scratch work. Roughly $31\%$ of raw mode
features come from files that are neither present at the base commit nor in the
gold patch, such as reproduction scripts, generated migrations, and build
artifacts. Such features make otherwise correct modes impossible to match. We
keep a feature only if its file is present at the base commit, appears in the
gold patch, or is created at least $30\%$ of the time across all runs. On SWE-Bench-Verified, the
consensus clause is empty in practice, so the filter removes scratch features
without dropping observed source fixes.

\paragraph{Divergence choices.}
Section~\ref{sec:distance} already gives the canonical solution distance as a
recall-style divergence (Eq.~\ref{eq:distance-main}); the implementation here
just substitutes the empirical $\tilde{\mathcal{S}}_i$ for the unavailable
$\mathcal{S}_i$. Writing $d_i(t)=\phi_i(x_i(t))$ for the source-projected
live-diff feature set after replay step $t$ (or $d(t)$ when the instance is
clear), the per-reference recall fraction is
\begin{equation}
  \mathrm{recall}_P(t) \;=\; \frac{|\, d(t) \cap \tau(P)\,|}{|\,\tau(P)\,|},
  \label{eq:soldist:progress}
\end{equation}
and Eq.~\ref{eq:distance-main} is evaluated empirically as
$D(t) = 1-\max_{P\in\tilde{\mathcal{S}}_i}\mathrm{recall}_P(t)$.

A more symmetric alternative is a Jaccard distance to the nearest reference,
\begin{equation}
  D^{\mathrm{Jac}}_P(t)
  \;=\;
  1 - \frac{|\,d(t)\cap\tau(P)\,|}{|\,d(t)\cup\tau(P)\,|},
  \label{eq:soldist:jaccard}
\end{equation}
which penalises every live feature not in the reference as well as every missing
one. We do not use it for live scoring because $D^{\mathrm{Jac}}$ penalizes harmless additions. If an agent reproduces all features of a verified repair
but also adds $m$ harmless features (comments, nearby cleanup, reproduction
helpers, or unrelated files), recall reports $0$ while Jaccard reports
$m/(|\tau(P)|+m)$. From a practical viewpoint, such incidental features are common in the agent
traces and difficult to filter without dropping real fix features. This is our primary motivation
to keep recall for live-state scoring and sacrificing the theoretical \textit{distance} property of Eq.\ref{eq:soldist:jaccard}.
As a reminder, we use Jaccard only when de-duplicating
verified reference patches, where both operands are themselves candidate
solutions. Finally, beyond these two options, one could plug in precision-style penalties on
extra live features, edit-distance over patch text, or semantic divergences
(AST alignment, embedding similarity, LLM-judged ``fraction of fix achieved'') and
we leave those to future work.

\textbf{Implementation specifics}: Two details are worth mentioning.
(1) $d(t)$ must be extracted with the \emph{same diff algorithm} that produced the solution set elements, a unified diff (for example, \texttt{difflib.unified\_diff(base, current)}), not a whole-file set difference. In an internal experiment, using set-diff inflated the residual end-distance by about $0.2$, because a changed line whose text recurs elsewhere in the file makes set-diff miss the corresponding `$-$' patch-feature while the git-diff mode has it.
(2) Edit-step (via edit tool, shell command calls) paths must be normalized relative to repo workspace before feature extraction. Without this, models that issue absolute paths (e.g. \texttt{/testbed/django/\ldots}) produce patch-features with file prefixes that never match the empirical solution relative paths, so $D$ stays at $1$.

\subsubsection{Illustrative examples}
\label{app:metrics:soldist:examples}

We elaborate the recall-divergence (Eq. (\ref{eq:distance-main})) on five sample trajectories that span the qualitative shapes
it should distinguish. (1) One decisive edit, (2) a gradual staircase with respect to (w.r.t.) a single element in solution set, (3) the same
gradual descent w.r.t. complete solution set, (4) found-it-and-lost-it behavior, and (5) revert as A/B testing. Throughout, green tiles mark features the live diff has reproduced
and gray tiles mark features still missing. Because $D=1$ leaves the best-mode
selection undefined (zero overlap with every reference in $\mathcal{S}_i$), we fix the displayed
reference for any $D=1$ panel to the benchmark gold patch as a visual
tie-breaker. Note that, the agent may later reach a different oracle solution.

\paragraph{Example 1. Decisive single-edit fix.}
Figure~\ref{fig:solution-distance-illustration} walks through the GPT-5.4
trajectory for \texttt{astropy\_\_astropy-12907}, whose gold patch is a
one-line replacement (two patch features: one deletion and one addition). A
single edit at cycle position $t \approx 0.53$ reproduces both features
simultaneously, so two snapshots fully characterise the run: the initial
state ($D{=}1$) and the post-edit state ($D{=}0$). This is the simplest
trajectory shape --- one edit collapses $D$ from $1$ to $0$ in a single
step.

\begin{figure}[!htbp]
\centering
\begin{tikzpicture}[
  every node/.style={inner sep=2pt, outer sep=0pt},
  panelbox/.style={draw=black!45, line width=0.4pt, rounded corners=2pt, fill=white},
  stagelabel/.style={font=\sffamily\bfseries\footnotesize, anchor=north, align=center},
  modelabel/.style={font=\sffamily\scriptsize\itshape, anchor=north, text=black!60, align=center},
  dvalbad/.style={font=\sffamily\bfseries\small, anchor=north, text=red!65!black},
  dvalmid/.style={font=\sffamily\bfseries\small, anchor=north, text=orange!75!black},
  dvalok/.style={font=\sffamily\bfseries\small, anchor=north, text=green!45!black},
  diffline/.style={font=\ttfamily\tiny, anchor=west, text width=3.25cm, minimum height=3.5mm, inner xsep=2pt, inner ysep=0pt, align=left},
  matched/.style={fill=green!18, draw=none, text=black},
  unmatched/.style={fill=gray!8, draw=none, text=black!55},
  hunkhdr/.style={font=\ttfamily\tiny\bfseries, anchor=west, text width=3.25cm, minimum height=3.5mm, inner xsep=2pt, inner ysep=0pt, align=left, fill=none, draw=none, text=violet!60!black},
  noteline/.style={font=\sffamily\scriptsize, anchor=west, text width=3.25cm, minimum height=3.5mm, inner xsep=2pt, inner ysep=0pt, align=left, text=black!55, draw=none, fill=none},
  tklabel/.style={font=\sffamily\bfseries\tiny, text=black!80},
  midarrow/.style={postaction={decorate, decoration={markings, mark=at position 0.5 with {\arrow{Stealth[length=4pt,width=4pt]}}}}},
]

\begin{scope}[shift={(2.7, 6.7)}]
  \draw[gray!55, line width=0.5pt] (0, 0) -- (8.2, 0);
  \draw[gray!55, line width=0.5pt] (0, 0) -- (0, 2.3);
  \foreach \x/\lab in {0/0, 4/0.5, 8/1} {
    \draw[gray!55] (\x, 0) -- (\x, -0.07);
    \node[below=1pt, font=\tiny, text=black!70] at (\x, 0) {\lab};
  }
  \foreach \y/\lab in {0/0, 1.0/0.5, 2.0/1} {
    \draw[gray!55] (0, \y) -- (-0.07, \y);
    \node[left=1pt, font=\tiny, text=black!70] at (0, \y) {\lab};
  }
  \node[below, font=\scriptsize, text=black!75] at (4, -0.45) {$t$};
  \node[rotate=90, above, font=\scriptsize, text=black!75] at (-0.7, 1.0) {$D(t)$};
  \draw[gray!20, dashed, very thin] (0, 2.0) -- (8, 2.0);
  \draw[gray!20, dashed, very thin] (0, 1.0) -- (8, 1.0);
  \draw[blue!60!black, line width=0.9pt, midarrow] (0, 2.0) -- (4.222, 2.0);
  \draw[blue!60!black, line width=0.9pt] (4.222, 2.0) -- (4.222, 0.0);
  \draw[blue!60!black, line width=0.9pt, midarrow] (4.222, 0.0) -- (8, 0.0);
  \filldraw[blue!60!black] (0, 2.0) circle (1.6pt);
  \filldraw[blue!60!black] (8, 0.0) circle (1.6pt);
  \node[tklabel, above=2pt] at (0, 2.0) {$t_1$};
  \node[tklabel, above right=1pt] at (8, 0.0) {$t_2$};
\end{scope}

\begin{scope}[shift={(1.5,0)}]
  \node[stagelabel] (t1) at (1.75, 5.35) {$x(t_1)$};
  \node[dvalbad]    (d1) at (1.75, 4.55) {$D = 1.0$};
  \node[modelabel]  (m1) at (1.75, 4.05) {gold patch: 2 features\\($D=1$ tie)};
  \node[noteline]   (n1) at (0.10, 3.25) {\textit{no edits applied}};
  \node[hunkhdr]    (h1) at (0.10, 2.55) {@@ separable.py @@};
  \node[diffline, unmatched] (r1a) at (0.10, 2.20) {- cright[...,:] = 1};
  \node[diffline, unmatched] (r1b) at (0.10, 1.85) {+ cright[...,:] = right};
  \node[noteline]   (p1) at (0.10, 1.20) {$0/2$ features reproduced};
  \begin{scope}[on background layer]
    \node[panelbox, fit=(t1)(d1)(m1)(n1)(h1)(r1b)(p1), inner sep=5pt] {};
  \end{scope}
\end{scope}

\begin{scope}[shift={(8.0,0)}]
  \node[stagelabel] (t2) at (1.75, 5.35) {$x(t_2)$};
  \node[dvalok]     (d2) at (1.75, 4.55) {$D = 0.0$};
  \node[modelabel]  (m2) at (1.75, 4.05) {gold patch: 2 features};
  \node[noteline]   (n2) at (0.10, 3.25) {\textit{correct fix landed}};
  \node[hunkhdr]    (h2) at (0.10, 2.55) {@@ separable.py @@};
  \node[diffline, matched] (r2a) at (0.10, 2.20) {- cright[...,:] = 1};
  \node[diffline, matched] (r2b) at (0.10, 1.85) {+ cright[...,:] = right};
  \node[noteline]   (p2) at (0.10, 1.20) {$2/2$ features reproduced};
  \begin{scope}[on background layer]
    \node[panelbox, fit=(t2)(d2)(m2)(n2)(h2)(r2b)(p2), inner sep=5pt] {};
  \end{scope}
\end{scope}

\end{tikzpicture}
\caption{\textbf{Single-edit collapse.} A single correct edit can move an
agent directly from the initial state to the solution set. Faithful replay of
the GPT-5.4 trajectory on \texttt{astropy\_\_astropy-12907} (SWE-Bench
Verified) has only two relevant states: the initial repository state $x(t_1)$
before any edit ($D{=}1$) and the post-edit state $x(t_2)$ after the decisive
edit at cycle position
$t \approx 0.53$ ($D{=}0$). Green tiles mark reference patch features already
present in the live diff; gray tiles mark features still absent. At $t_1$ every
reference in $\tilde{\mathcal{S}}_i$ has zero overlap, so the displayed
reference is fixed to the gold patch only as a visual tie-breaker. The curve
therefore shows the cleanest possible trajectory shape: one edit moves the
agent directly into the solution set.}
\label{fig:solution-distance-illustration}
\end{figure}

\paragraph{Example 2. Gradual staircase against a single reference.}
Most runs descend through several transient non-zero $D$ values before
reaching the solution set. Figure~\ref{fig:solution-distance-staircase} shows one
such descent on the Opus 4.6 trajectory for \texttt{django\_\_django-11292} (SWE-Bench-Verified),
whose reference fix touches three hunks. The divergence is scored against a single 8-feature
reference solution in $\mathcal{S}_i$, and $D$ records four non-zero values across the first three edits
at normalized cycle positions $t=0.27$, $0.28$, $0.29$, then holds at $0$ through three
refinements, each hunk landing in turn corresponds to one stair step.

\begin{figure}[!htbp]
\centering
\begin{tikzpicture}[
  every node/.style={inner sep=2pt, outer sep=0pt},
  panelbox/.style={draw=black!45, line width=0.4pt, rounded corners=2pt, fill=white},
  stagelabel/.style={font=\sffamily\bfseries\footnotesize, anchor=north, align=center},
  dvalbad/.style={font=\sffamily\bfseries\small, anchor=north, text=red!65!black},
  dvalmid/.style={font=\sffamily\bfseries\small, anchor=north, text=orange!75!black},
  dvalok/.style={font=\sffamily\bfseries\small, anchor=north, text=green!45!black},
  diffline/.style={font=\ttfamily\tiny, anchor=west, text width=3.05cm, minimum height=3.5mm, inner xsep=2pt, inner ysep=0pt, align=left},
  matched/.style={fill=green!18, draw=none, text=black},
  unmatched/.style={fill=gray!8, draw=none, text=black!55},
  hunkhdr/.style={font=\ttfamily\tiny\bfseries, anchor=west, text width=3.05cm, minimum height=3.5mm, inner xsep=2pt, inner ysep=0pt, align=left, fill=none, draw=none, text=violet!60!black},
  tklabel/.style={font=\sffamily\bfseries\tiny, text=black!80},
  midarrow/.style={postaction={decorate, decoration={markings, mark=at position 0.5 with {\arrow{Stealth[length=4pt,width=4pt]}}}}},
]

\begin{scope}[shift={(2.8, 8.6)}]
  \draw[gray!55, line width=0.5pt] (0, 0) -- (8.2, 0);
  \draw[gray!55, line width=0.5pt] (0, 0) -- (0, 2.3);
  \foreach \x/\lab in {0/0, 4/0.5, 8/1} {
    \draw[gray!55] (\x, 0) -- (\x, -0.07);
    \node[below=1pt, font=\tiny, text=black!70] at (\x, 0) {\lab};
  }
  \foreach \y/\lab in {0/0, 1.0/0.5, 2.0/1} {
    \draw[gray!55] (0, \y) -- (-0.07, \y);
    \node[left=1pt, font=\tiny, text=black!70] at (0, \y) {\lab};
  }
  \node[below, font=\scriptsize, text=black!75] at (4, -0.45) {$t$};
  \node[rotate=90, above, font=\scriptsize, text=black!75] at (-0.7, 1.0) {$D(t)$};
  \draw[gray!20, dashed, very thin] (0, 2.0) -- (8, 2.0);
  \draw[gray!20, dashed, very thin] (0, 1.0) -- (8, 1.0);
  \draw[blue!60!black, line width=0.9pt, midarrow] (0, 2.0) -- (2.185, 2.0);
  \draw[blue!60!black, line width=0.9pt]
    (2.185, 2.0) -- (2.185, 1.5)
              -- (2.259, 1.5) -- (2.259, 1.0)
              -- (2.333, 1.0) -- (2.333, 0.0);
  \draw[blue!60!black, line width=0.9pt, midarrow] (2.333, 0.0) -- (8, 0.0);
  \filldraw[blue!60!black] (0, 2.0) circle (1.6pt);
  \filldraw[blue!60!black] (2.185, 1.5) circle (1.6pt);
  \filldraw[blue!60!black] (2.259, 1.0) circle (1.6pt);
  \filldraw[blue!60!black] (8, 0.0)     circle (1.6pt);
  \node[tklabel, above=2pt] at (0, 2.0) {$t_1$};
  \draw[black!50, line width=0.3pt] (2.185, 1.5) -- (1.45, 2.0);
  \node[tklabel] at (1.30, 2.1) {$t_2$};
  \draw[black!50, line width=0.3pt] (2.259, 1.0) -- (3.15, 1.6);
  \node[tklabel] at (3.30, 1.7) {$t_3$};
  \node[tklabel, above=2pt] at (8, 0.0) {$t_4$};
\end{scope}

\begin{scope}[shift={(0, 0)}]
  \node[stagelabel] (t1) at (1.65, 7.05) {$x(t_1)$};
  \node[dvalbad]    (d1) at (1.65, 6.10) {$D = 1.0$};
  \node[hunkhdr]            (h1a) at (0.10, 5.30) {@@ -95,7 +95,7 @@};
  \node[diffline, unmatched](n1a) at (0.10, 4.95) {- '...', '-{}-force-color',};
  \node[diffline, unmatched](n1b) at (0.10, 4.60) {+ '...', '-{}-skip-checks',};
  \node[hunkhdr]            (h1b) at (0.10, 4.20) {@@ -223,7 +223,7 @@};
  \node[diffline, unmatched](n1c) at (0.10, 3.85) {- ...=('skip\_checks',...)};
  \node[diffline, unmatched](n1d) at (0.10, 3.50) {+ ...=('stderr','stdout')};
  \node[hunkhdr]            (h1c) at (0.10, 3.10) {@@ -286,6 +286,10 @@};
  \node[diffline, unmatched](n1f) at (0.10, 2.75) {+ parser.add\_argument(};
  \node[diffline, unmatched](n1g) at (0.10, 2.40) {+\ \ \ '-{}-skip-checks',...};
  \node[diffline, unmatched](n1h) at (0.10, 2.05) {+\ \ \ help='Skip ...',};
  \node[diffline, unmatched](n1i) at (0.10, 1.70) {+ )};
  \begin{scope}[on background layer]
    \node[panelbox, fit=(t1)(d1)(h1a)(n1i), inner sep=5pt] {};
  \end{scope}
\end{scope}

\begin{scope}[shift={(3.5, 0)}]
  \node[stagelabel] (t2) at (1.65, 7.05) {$x(t_2)$};
  \node[dvalmid]    (d2) at (1.65, 6.10) {$D = 0.75$};
  \node[hunkhdr]            (h2a) at (0.10, 5.30) {@@ -95,7 +95,7 @@};
  \node[diffline, matched]  (n2a) at (0.10, 4.95) {- '...', '-{}-force-color',};
  \node[diffline, matched]  (n2b) at (0.10, 4.60) {+ '...', '-{}-skip-checks',};
  \node[hunkhdr]            (h2b) at (0.10, 4.20) {@@ -223,7 +223,7 @@};
  \node[diffline, unmatched](n2c) at (0.10, 3.85) {- ...=('skip\_checks',...)};
  \node[diffline, unmatched](n2d) at (0.10, 3.50) {+ ...=('stderr','stdout')};
  \node[hunkhdr]            (h2c) at (0.10, 3.10) {@@ -286,6 +286,10 @@};
  \node[diffline, unmatched](n2f) at (0.10, 2.75) {+ parser.add\_argument(};
  \node[diffline, unmatched](n2g) at (0.10, 2.40) {+\ \ \ '-{}-skip-checks',...};
  \node[diffline, unmatched](n2h) at (0.10, 2.05) {+\ \ \ help='Skip ...',};
  \node[diffline, unmatched](n2i) at (0.10, 1.70) {+ )};
  \begin{scope}[on background layer]
    \node[panelbox, fit=(t2)(d2)(h2a)(n2i), inner sep=5pt] {};
  \end{scope}
\end{scope}

\begin{scope}[shift={(7.0, 0)}]
  \node[stagelabel] (t3) at (1.65, 7.05) {$x(t_3)$};
  \node[dvalmid]    (d3) at (1.65, 6.10) {$D = 0.5$};
  \node[hunkhdr]            (h3a) at (0.10, 5.30) {@@ -95,7 +95,7 @@};
  \node[diffline, matched]  (n3a) at (0.10, 4.95) {- '...', '-{}-force-color',};
  \node[diffline, matched]  (n3b) at (0.10, 4.60) {+ '...', '-{}-skip-checks',};
  \node[hunkhdr]            (h3b) at (0.10, 4.20) {@@ -223,7 +223,7 @@};
  \node[diffline, matched]  (n3c) at (0.10, 3.85) {- ...=('skip\_checks',...)};
  \node[diffline, matched]  (n3d) at (0.10, 3.50) {+ ...=('stderr','stdout')};
  \node[hunkhdr]            (h3c) at (0.10, 3.10) {@@ -286,6 +286,10 @@};
  \node[diffline, unmatched](n3f) at (0.10, 2.75) {+ parser.add\_argument(};
  \node[diffline, unmatched](n3g) at (0.10, 2.40) {+\ \ \ '-{}-skip-checks',...};
  \node[diffline, unmatched](n3h) at (0.10, 2.05) {+\ \ \ help='Skip ...',};
  \node[diffline, unmatched](n3i) at (0.10, 1.70) {+ )};
  \begin{scope}[on background layer]
    \node[panelbox, fit=(t3)(d3)(h3a)(n3i), inner sep=5pt] {};
  \end{scope}
\end{scope}

\begin{scope}[shift={(10.5, 0)}]
  \node[stagelabel] (t4) at (1.65, 7.05) {$x(t_4)$};
  \node[dvalok]     (d4) at (1.65, 6.10) {$D = 0.0$};
  \node[hunkhdr]            (h4a) at (0.10, 5.30) {@@ -95,7 +95,7 @@};
  \node[diffline, matched]  (n4a) at (0.10, 4.95) {- '...', '-{}-force-color',};
  \node[diffline, matched]  (n4b) at (0.10, 4.60) {+ '...', '-{}-skip-checks',};
  \node[hunkhdr]            (h4b) at (0.10, 4.20) {@@ -223,7 +223,7 @@};
  \node[diffline, matched]  (n4c) at (0.10, 3.85) {- ...=('skip\_checks',...)};
  \node[diffline, matched]  (n4d) at (0.10, 3.50) {+ ...=('stderr','stdout')};
  \node[hunkhdr]            (h4c) at (0.10, 3.10) {@@ -286,6 +286,10 @@};
  \node[diffline, matched]  (n4f) at (0.10, 2.75) {+ parser.add\_argument(};
  \node[diffline, matched]  (n4g) at (0.10, 2.40) {+\ \ \ '-{}-skip-checks',...};
  \node[diffline, matched]  (n4h) at (0.10, 2.05) {+\ \ \ help='Skip ...',};
  \node[diffline, matched]  (n4i) at (0.10, 1.70) {+ )};
  \begin{scope}[on background layer]
    \node[panelbox, fit=(t4)(d4)(h4a)(n4i), inner sep=5pt] {};
  \end{scope}
\end{scope}

\end{tikzpicture}
\caption{\textbf{Fixed-reference staircase.} Against a fixed reference,
solution distance behaves like recall over patch features. The Opus 4.6
trajectory for \texttt{django\_\_django-11292} is scored against one 8-feature
reference held constant across the four panels. Green tiles are reference
features reproduced by the live repository state, and gray tiles are missing
features. As the agent
lands successive hunks, the matched count increases from $0/8$ to $2/8$,
$4/8$, and finally $8/8$, producing the staircase
$D{=}1 \to 0.75 \to 0.5 \to 0$. This fixed-reference view makes the
recall-style construction concrete: distance decreases in proportion to the
fraction of the target patch already committed to disk. The $t_2$ value is
higher than the recorded trajectory value ($\approx 0.67$) because the actual
metric maximises over all empirical reference modes, as shown in
Figure~\ref{fig:solution-distance-staircase-bestmode}.}
\label{fig:solution-distance-staircase}
\end{figure}

\paragraph{Example 3. Same trajectory under the \texttt{max}-over-references selection.}
Figure~\ref{fig:solution-distance-staircase-bestmode} shows the same Opus 4.6
run (as in Example 2), but the displayed reference is now selected per step whenever the
live state has nonzero overlap with an element in $\mathcal{S}_i$. Panel 1 is the $D=1$ tie-breaker (gold patch), 
panel 2 scores against a popular 6-feature mode (22 of 104 resolved runs) which becomes closest in $\tilde{\mathcal{S}}_i$. Panels 3 and 4 score against the
8-feature three-hunk solution state shared with the Opus 4.5 runs as well (11 runs), which
the live diff matches at $4/8$ after the second edit and $8/8$ after the
third. Reported values $1.0 \to 0.667 \to 0.5 \to 0.0$ show the agent journey in the state space $\mathcal{T}_i$ towards $\mathcal{S}_i$ (or empirically approximated $\tilde{\mathcal{S}_i}$).

\begin{figure}[!htbp]
\centering
\begin{tikzpicture}[
  every node/.style={inner sep=2pt, outer sep=0pt},
  panelbox/.style={draw=black!45, line width=0.4pt, rounded corners=2pt, fill=white},
  stagelabel/.style={font=\sffamily\bfseries\footnotesize, anchor=north, align=center},
  modelabel/.style={font=\sffamily\scriptsize\itshape, anchor=north, text=black!60, align=center},
  dvalbad/.style={font=\sffamily\bfseries\small, anchor=north, text=red!65!black},
  dvalmid/.style={font=\sffamily\bfseries\small, anchor=north, text=orange!75!black},
  dvalok/.style={font=\sffamily\bfseries\small, anchor=north, text=green!45!black},
  diffline/.style={font=\ttfamily\tiny, anchor=west, text width=3.05cm, minimum height=3.5mm, inner xsep=2pt, inner ysep=0pt, align=left},
  matched/.style={fill=green!18, draw=none, text=black},
  unmatched/.style={fill=gray!8, draw=none, text=black!55},
  hunkhdr/.style={font=\ttfamily\tiny\bfseries, anchor=west, text width=3.05cm, minimum height=3.5mm, inner xsep=2pt, inner ysep=0pt, align=left, fill=none, draw=none, text=violet!60!black},
  tklabel/.style={font=\sffamily\bfseries\tiny, text=black!80},
  midarrow/.style={postaction={decorate, decoration={markings, mark=at position 0.5 with {\arrow{Stealth[length=4pt,width=4pt]}}}}},
]

\begin{scope}[shift={(2.8, 8.6)}]
  \draw[gray!55, line width=0.5pt] (0, 0) -- (8.2, 0);
  \draw[gray!55, line width=0.5pt] (0, 0) -- (0, 2.3);
  \foreach \x/\lab in {0/0, 4/0.5, 8/1} {
    \draw[gray!55] (\x, 0) -- (\x, -0.07);
    \node[below=1pt, font=\tiny, text=black!70] at (\x, 0) {\lab};
  }
  \foreach \y/\lab in {0/0, 1.0/0.5, 2.0/1} {
    \draw[gray!55] (0, \y) -- (-0.07, \y);
    \node[left=1pt, font=\tiny, text=black!70] at (0, \y) {\lab};
  }
  \node[below, font=\scriptsize, text=black!75] at (4, -0.45) {$t$};
  \node[rotate=90, above, font=\scriptsize, text=black!75] at (-0.7, 1.0) {$D(t)$};
  \draw[gray!20, dashed, very thin] (0, 2.0) -- (8, 2.0);
  \draw[gray!20, dashed, very thin] (0, 1.0) -- (8, 1.0);
  \draw[blue!60!black, line width=0.9pt, midarrow] (0, 2.0) -- (2.185, 2.0);
  \draw[blue!60!black, line width=0.9pt]
    (2.185, 2.0) -- (2.185, 1.333)
              -- (2.259, 1.333) -- (2.259, 1.0)
              -- (2.333, 1.0)   -- (2.333, 0.0);
  \draw[blue!60!black, line width=0.9pt, midarrow] (2.333, 0.0) -- (8, 0.0);
  \filldraw[blue!60!black] (0, 2.0) circle (1.6pt);
  \filldraw[blue!60!black] (2.185, 1.333) circle (1.6pt);
  \filldraw[blue!60!black] (2.259, 1.0)   circle (1.6pt);
  \filldraw[blue!60!black] (8, 0.0)       circle (1.6pt);
  \node[tklabel, above=2pt] at (0, 2.0) {$t_1$};
  \draw[black!50, line width=0.3pt] (2.185, 1.333) -- (1.45, 1.85);
  \node[tklabel] at (1.30, 1.95) {$t_2$};
  \draw[black!50, line width=0.3pt] (2.259, 1.0)   -- (3.15, 1.6);
  \node[tklabel] at (3.30, 1.7) {$t_3$};
  \node[tklabel, above=2pt] at (8, 0.0) {$t_4$};
\end{scope}

\begin{scope}[shift={(0, 0)}]
  \node[stagelabel] (T1) at (1.65, 7.30) {$x(t_1)$};
  \node[dvalbad]    (D1) at (1.65, 6.35) {$D = 1.0$};
  \node[modelabel]  (M1) at (1.65, 5.85) {shown: gold patch\\($D=1$ tie)};
  \node[hunkhdr]            (H1a) at (0.10, 5.10) {@@ -95,7 +95,7 @@};
  \node[diffline, unmatched](N1a) at (0.10, 4.75) {- '...', '-{}-force-color',};
  \node[diffline, unmatched](N1b) at (0.10, 4.40) {+ ..., '-{}-skip-checks',};
  \node[hunkhdr]            (H1b) at (0.10, 4.02) {@@ -223,7 +223,7 @@};
  \node[diffline, unmatched](N1c) at (0.10, 3.67) {- ...=('skip\_checks',...)};
  \node[diffline, unmatched](N1d) at (0.10, 3.32) {+ ...=('stderr','stdout')};
  \node[hunkhdr]            (H1c) at (0.10, 2.94) {@@ -286,6 +286,11 @@};
  \node[diffline, unmatched](N1e) at (0.10, 2.59) {+ if self.requires\_...:};
  \node[diffline, unmatched](N1f) at (0.10, 2.24) {+ parser.add\_argument(};
  \node[diffline, unmatched](N1g) at (0.10, 1.89) {+\ \ \ '-{}-skip-checks',...};
  \node[diffline, unmatched](N1h) at (0.10, 1.54) {+\ \ \ help='Skip ...',};
  \node[diffline, unmatched](N1i) at (0.10, 1.19) {+ )};
  \node[hunkhdr]            (H1d) at (0.10, 0.81) {@@ -357,7 +362,7 @@};
  \node[diffline, unmatched](N1j) at (0.10, 0.46) {- if ... not options.get(...):};
  \node[diffline, unmatched](N1k) at (0.10, 0.11) {+ if ... not skip\_checks:};
  \begin{scope}[on background layer]
    \node[panelbox, fit=(T1)(D1)(M1)(H1a)(N1k), inner sep=5pt] {};
  \end{scope}
\end{scope}

\begin{scope}[shift={(3.5, 0)}]
  \node[stagelabel] (T2) at (1.65, 7.30) {$x(t_2)$};
  \node[dvalmid]    (D2) at (1.65, 6.35) {$D \approx 0.67$};
  \node[modelabel]  (M2) at (1.65, 5.85) {best mode: 6 features\\(popular, 22 runs)};
  \node[hunkhdr]            (H2a) at (0.10, 4.85) {@@ -95,7 +95,7 @@};
  \node[diffline, matched]  (N2a) at (0.10, 4.50) {- '...', '-{}-force-color',};
  \node[diffline, matched]  (N2b) at (0.10, 4.15) {+ '...', '-{}-skip-checks',};
  \node[hunkhdr]            (H2c) at (0.10, 3.75) {@@ -286,6 +286,10 @@};
  \node[diffline, unmatched](N2f) at (0.10, 3.40) {+ parser.add\_argument(};
  \node[diffline, unmatched](N2g) at (0.10, 3.05) {+\ \ \ '-{}-skip-checks',...};
  \node[diffline, unmatched](N2h) at (0.10, 2.70) {+\ \ \ help='Skip ...',};
  \node[diffline, unmatched](N2i) at (0.10, 2.35) {+ )};
  \begin{scope}[on background layer]
    \node[panelbox, fit=(T2)(D2)(M2)(H2a)(N2i), inner sep=5pt] {};
  \end{scope}
\end{scope}

\begin{scope}[shift={(7.0, 0)}]
  \node[stagelabel] (T3) at (1.65, 7.30) {$x(t_3)$};
  \node[dvalmid]    (D3) at (1.65, 6.35) {$D = 0.5$};
  \node[modelabel]  (M3) at (1.65, 5.85) {best mode: 8 features\\(Opus 4.5 family, 11 runs)};
  \node[hunkhdr]            (H3a) at (0.10, 4.85) {@@ -95,7 +95,7 @@};
  \node[diffline, matched]  (N3a) at (0.10, 4.50) {- '...', '-{}-force-color',};
  \node[diffline, matched]  (N3b) at (0.10, 4.15) {+ '...', '-{}-skip-checks',};
  \node[hunkhdr]            (H3b) at (0.10, 3.75) {@@ -223,7 +223,7 @@};
  \node[diffline, matched]  (N3c) at (0.10, 3.40) {- ...=('skip\_checks',...)};
  \node[diffline, matched]  (N3d) at (0.10, 3.05) {+ ...=('stderr','stdout')};
  \node[hunkhdr]            (H3c) at (0.10, 2.65) {@@ -286,6 +286,10 @@};
  \node[diffline, unmatched](N3f) at (0.10, 2.30) {+ parser.add\_argument(};
  \node[diffline, unmatched](N3g) at (0.10, 1.95) {+\ \ \ '-{}-skip-checks',...};
  \node[diffline, unmatched](N3h) at (0.10, 1.60) {+\ \ \ help='Skip ...',};
  \node[diffline, unmatched](N3i) at (0.10, 1.25) {+ )};
  \begin{scope}[on background layer]
    \node[panelbox, fit=(T3)(D3)(M3)(H3a)(N3i), inner sep=5pt] {};
  \end{scope}
\end{scope}

\begin{scope}[shift={(10.5, 0)}]
  \node[stagelabel] (T4) at (1.65, 7.30) {$x(t_4)$};
  \node[dvalok]     (D4) at (1.65, 6.35) {$D = 0.0$};
  \node[modelabel]  (M4) at (1.65, 5.85) {best mode: 8 features\\(Opus final)};
  \node[hunkhdr]            (H4a) at (0.10, 4.60) {@@ -95,7 +95,7 @@};
  \node[diffline, matched]  (N4a) at (0.10, 4.25) {- '...', '-{}-force-color',};
  \node[diffline, matched]  (N4b) at (0.10, 3.90) {+ '...', '-{}-skip-checks',};
  \node[hunkhdr]            (H4b) at (0.10, 3.50) {@@ -223,7 +223,7 @@};
  \node[diffline, matched]  (N4c) at (0.10, 3.15) {- ...=('skip\_checks',...)};
  \node[diffline, matched]  (N4d) at (0.10, 2.80) {+ ...=('stderr','stdout')};
  \node[hunkhdr]            (H4c) at (0.10, 2.40) {@@ -286,6 +286,10 @@};
  \node[diffline, matched]  (N4e) at (0.10, 2.05) {+ parser.add\_argument(};
  \node[diffline, matched]  (N4f) at (0.10, 1.70) {+\ \ \ '-{}-skip-checks',...};
  \node[diffline, matched]  (N4g) at (0.10, 1.35) {+\ \ \ help='Skip ...',};
  \node[diffline, matched]  (N4h) at (0.10, 1.00) {+ )};
  \begin{scope}[on background layer]
    \node[panelbox, fit=(T4)(D4)(M4)(H4a)(N4h), inner sep=5pt] {};
  \end{scope}
\end{scope}

\end{tikzpicture}
\caption{\textbf{Max-over-modes view.} The full metric tracks the nearest
empirical solution mode, not a single fixed patch. The same Opus 4.6 run on
\texttt{django\_\_django-11292} is re-rendered with the displayed reference
chosen independently at each snapshot by the maximisation in
Eq.~\ref{eq:distance-main}. The trajectory descends
$1 \to 0.667 \to 0.5 \to 0$ between cycles $0.273$ and $0.292$, then remains
at zero through three later refinements. The best reference changes along the
way: panel 1 uses the gold patch only as a zero-overlap tie-breaker, panel 2
selects a smaller 6-feature mode shared by 22 resolved runs, and panels 3--4
select an 8-feature mode shared with the Opus 4.5 cluster. This view explains
why the empirical $D(t)$ values differ from the fixed-reference staircase in
Figure~\ref{fig:solution-distance-staircase}: progress is measured against the
closest element of $\tilde{\mathcal{S}}_i$ at each state.}
\label{fig:solution-distance-staircase-bestmode}
\end{figure}

\paragraph{Example 4. Found-it-then-lost-it.}
We observed that not every trajectory that reaches $\tilde{\mathcal{S}}_i$ stays there till the end.
Figure~\ref{fig:solution-distance-thrash-gemflash} shows a Gemini~3~Flash run on
\texttt{django\_\_django-14140} (SWE-Bench-Verified), where the target is \texttt{Q.deconstruct()}
in \texttt{django/db/models/query\_utils.py}. Edit \#1 (normalized cycle $t=0.15$) writes
exactly the popular $12$-feature minimal rewrite shared by $22$ resolved runs
and the agent reaches $\tilde{\mathcal{S}}_i$ ($D{=}0$). Edit \#2 (cycle $0.46$) reverts
that rewrite and replaces it with a $7$-line multi-condition guard that
shares only $1/12$ features with that reference element, so $D$ jumps to $0.917$. Three
further refinements (cycles $t=0.63$, $0.76$, $0.80$) trim the guard but never
restore the simple rewrite, and the run terminates at $D{=}0.9$, unresolved. Therefore,
we should be aware of events where agent successfully reached the desired state (resolved in this case), but can 
backtrack and reaches a wrong conclusion. Such dissection of model behavior is helpful to further improve the model behavior.

\begin{figure}[p]
\centering
\resizebox{\textwidth}{!}{%
\begin{tikzpicture}[
  every node/.style={inner sep=2pt, outer sep=0pt},
  panelbox/.style={draw=black!45, line width=0.4pt, rounded corners=2pt, fill=white},
  stagelabel/.style={font=\sffamily\bfseries\footnotesize, anchor=north, align=center},
  modelabel/.style={font=\sffamily\scriptsize\itshape, anchor=north, text=black!60, align=center},
  dvalbad/.style={font=\sffamily\bfseries\small, anchor=north, text=red!65!black},
  dvalmid/.style={font=\sffamily\bfseries\small, anchor=north, text=orange!75!black},
  dvalok/.style={font=\sffamily\bfseries\small, anchor=north, text=green!45!black},
  diffline/.style={font=\ttfamily\tiny, anchor=west, text width=3.05cm, minimum height=3.2mm, inner xsep=2pt, inner ysep=0pt, align=left},
  matched/.style={fill=green!18, draw=none, text=black},
  unmatched/.style={fill=gray!8, draw=none, text=black!55},
  hunkhdr/.style={font=\ttfamily\tiny\bfseries, anchor=west, text width=3.05cm, minimum height=3.2mm, inner xsep=2pt, inner ysep=0pt, align=left, fill=none, draw=none, text=violet!60!black},
  noteline/.style={font=\sffamily\scriptsize, anchor=west, text width=3.05cm, minimum height=3.2mm, inner xsep=2pt, inner ysep=0pt, align=left, text=black!55, draw=none, fill=none},
  tklabel/.style={font=\sffamily\bfseries\tiny, text=black!80},
  midarrow/.style={postaction={decorate, decoration={markings, mark=at position 0.5 with {\arrow{Stealth[length=4pt,width=4pt]}}}}},
]
\begin{scope}[shift={(2.8, 8.6)}]
  \draw[gray!55, line width=0.5pt] (0, 0) -- (8.2, 0);
  \draw[gray!55, line width=0.5pt] (0, 0) -- (0, 2.3);
  \foreach \x/\lab in {0/0, 4/0.5, 8/1} {
    \draw[gray!55] (\x, 0) -- (\x, -0.07);
    \node[below=1pt, font=\tiny, text=black!70] at (\x, 0) {\lab};
  }
  \foreach \y/\lab in {0/0, 1.0/0.5, 2.0/1} {
    \draw[gray!55] (0, \y) -- (-0.07, \y);
    \node[left=1pt, font=\tiny, text=black!70] at (0, \y) {\lab};
  }
  \node[below, font=\scriptsize, text=black!75] at (4, -0.45) {$t$};
  \node[rotate=90, above, font=\scriptsize, text=black!75] at (-0.7, 1.0) {$D(t)$};
  \draw[gray!20, dashed, very thin] (0, 2.0) -- (8, 2.0);
  \draw[gray!20, dashed, very thin] (0, 1.0) -- (8, 1.0);
  \draw[blue!60!black, line width=0.9pt, midarrow] (0, 2.0) -- (1.217, 2.0);
  \draw[blue!60!black, line width=0.9pt] (1.217, 2.0) -- (1.217, 0.0);
  \draw[blue!60!black, line width=0.9pt, midarrow] (1.217, 0.0) -- (3.652, 0.0);
  \draw[red!75!black, line width=1.1pt]
    (3.652, 0.0) -- (3.652, 1.8);
  \draw[blue!60!black, line width=0.9pt, midarrow] (3.652, 1.8) -- (8, 1.8);
  \filldraw[blue!60!black] (0, 2.0) circle (1.6pt);
  \filldraw[blue!60!black] (1.217, 0.0) circle (1.6pt);
  \filldraw[red!75!black]  (3.652, 1.8) circle (1.8pt);
  \filldraw[blue!60!black] (8, 1.8) circle (1.6pt);
  \node[tklabel, above=2pt] at (0, 2.0) {$t_1$};
  \node[tklabel, below right=1pt] at (1.217, 0.0) {$t_2$};
  \node[tklabel, above right=1pt, text=red!75!black] at (3.652, 1.8) {$t_3$};
  \node[tklabel, above=2pt] at (8, 1.8) {$t_4$};
\end{scope}
\begin{scope}[shift={(0, 0)}]
  \node[stagelabel] (T1) at (1.65, 7.60) {$x(t_1)$};
  \node[dvalbad]    (D1) at (1.65, 6.70) {$D = 1.0$};
  \node[modelabel]  (M1) at (1.65, 6.20) {shown: gold patch\\($D=1$ tie)};
  \node[noteline]   (NN1a) at (0.10, 5.30) {\textit{no edits applied}};
  \node[noteline]   (NN1b) at (0.10, 4.95) {$0/12$ features reproduced};
  \node[hunkhdr]            (H1a) at (0.10, 4.20) {@@ -85,11 +85,7 @@};
  \node[diffline, unmatched](N1a) at (0.10, 3.85) {- args, kwargs = (), \{\}};
  \node[diffline, unmatched](N1b) at (0.10, 3.50) {- if len(c)==1 and};
  \node[diffline, unmatched](N1c) at (0.10, 3.15) {-\ \ \ not isinst(c[0],Q):};
  \node[diffline, unmatched](N1d) at (0.10, 2.80) {-\ \ \ child = c[0]};
  \node[diffline, unmatched](N1e) at (0.10, 2.45) {-\ \ \ kwargs=\{c0:c1\}};
  \node[diffline, unmatched](N1f) at (0.10, 2.10) {- else:};
  \node[diffline, unmatched](N1g) at (0.10, 1.75) {-\ \ \ args=tuple(c)};
  \node[diffline, unmatched](N1h) at (0.10, 1.40) {-\ \ \ kwargs=\{'\_conn':...\}};
  \node[diffline, unmatched](N1i) at (0.10, 1.05) {+ args = tuple(c)};
  \node[diffline, unmatched](N1j) at (0.10, 0.70) {+ kwargs = \{\}};
  \node[diffline, unmatched](N1k) at (0.10, 0.35) {+ kwargs['\_conn']=conn};
  \begin{scope}[on background layer]
    \node[panelbox, fit=(T1)(D1)(M1)(H1a)(N1k), inner sep=5pt] {};
  \end{scope}
\end{scope}
\begin{scope}[shift={(3.5, 0)}]
  \node[stagelabel] (T2) at (1.65, 7.60) {$x(t_2)$};
  \node[dvalok]     (D2) at (1.65, 6.70) {$D = 0.0$};
  \node[modelabel]  (M2) at (1.65, 6.20) {best mode: 12 features\\(popular, 22 runs)};
  \node[noteline]   (NN2a) at (0.10, 5.30) {\textit{edit \#1: drop unpacking}};
  \node[noteline]   (NN2b) at (0.10, 4.95) {$12/12$ features reproduced};
  \node[hunkhdr]            (H2a) at (0.10, 4.20) {@@ -85,11 +85,7 @@};
  \node[diffline, matched]  (N2a) at (0.10, 3.85) {- args, kwargs = (), \{\}};
  \node[diffline, matched]  (N2b) at (0.10, 3.50) {- if len(c)==1 and};
  \node[diffline, matched]  (N2c) at (0.10, 3.15) {-\ \ \ not isinst(c[0],Q):};
  \node[diffline, matched]  (N2d) at (0.10, 2.80) {-\ \ \ child = c[0]};
  \node[diffline, matched]  (N2e) at (0.10, 2.45) {-\ \ \ kwargs=\{c0:c1\}};
  \node[diffline, matched]  (N2f) at (0.10, 2.10) {- else:};
  \node[diffline, matched]  (N2g) at (0.10, 1.75) {-\ \ \ args=tuple(c)};
  \node[diffline, matched]  (N2h) at (0.10, 1.40) {-\ \ \ kwargs=\{'\_conn':...\}};
  \node[diffline, matched]  (N2i) at (0.10, 1.05) {+ args = tuple(c)};
  \node[diffline, matched]  (N2j) at (0.10, 0.70) {+ kwargs = \{\}};
  \node[diffline, matched]  (N2k) at (0.10, 0.35) {+ kwargs['\_conn']=conn};
  \begin{scope}[on background layer]
    \node[panelbox, fit=(T2)(D2)(M2)(H2a)(N2k), inner sep=5pt] {};
  \end{scope}
\end{scope}
\begin{scope}[shift={(7.0, 0)}]
  \node[stagelabel] (T3) at (1.65, 7.60) {$x(t_3)$};
  \node[dvalbad]    (D3) at (1.65, 6.70) {$D = 0.9$};
  \node[modelabel]  (M3) at (1.65, 6.20) {best mode: 12 features\\(popular, 22 runs)};
  \node[noteline]   (NN3a) at (0.10, 5.30) {\textit{edit \#2: revert + 7-line guard}};
  \node[noteline]   (NN3b) at (0.10, 4.95) {$1/12$ features reproduced};
  \node[hunkhdr]            (H3a) at (0.10, 4.20) {@@ -85,1 +85,7 @@};
  \node[diffline, matched]  (N3a) at (0.10, 3.85) {- if len(c)==1 and};
  \node[diffline, matched]  (N3b) at (0.10, 3.50) {-\ \ \ not isinst(c[0],Q):};
  \node[diffline, unmatched](N3c) at (0.10, 3.15) {+ if (};
  \node[diffline, unmatched](N3d) at (0.10, 2.80) {+\ \ \ len(c)==1 and};
  \node[diffline, unmatched](N3e) at (0.10, 2.45) {+\ \ \ not isinst(c[0],Q) and};
  \node[diffline, unmatched](N3f) at (0.10, 2.10) {+\ \ \ hasattr(c[0],'\_\_gi\_\_')};
  \node[diffline, unmatched](N3g) at (0.10, 1.75) {+\ \ \ \ \ \ and};
  \node[diffline, unmatched](N3h) at (0.10, 1.40) {+\ \ \ not isinst(c[0],str) and};
  \node[diffline, unmatched](N3i) at (0.10, 1.05) {+\ \ \ len(c[0]) == 2};
  \node[diffline, unmatched](N3j) at (0.10, 0.70) {+ ):};
  \node[noteline]   (NN3c) at (0.10, 0.30) {+7 extra non-reference features};
  \begin{scope}[on background layer]
    \node[panelbox, fit=(T3)(D3)(M3)(H3a)(NN3c), inner sep=5pt] {};
  \end{scope}
\end{scope}
\begin{scope}[shift={(10.5, 0)}]
  \node[stagelabel] (T4) at (1.65, 7.60) {$x(t_4)$};
  \node[dvalbad]    (D4) at (1.65, 6.70) {$D = 0.9$};
  \node[modelabel]  (M4) at (1.65, 6.20) {best mode: 12 features\\(popular, 22 runs)};
  \node[noteline]   (NN4a) at (0.10, 5.30) {\textit{edits \#3--\#5: refine guard}};
  \node[noteline]   (NN4b) at (0.10, 4.95) {$1/12$ features reproduced};
  \node[hunkhdr]            (H4a) at (0.10, 4.20) {@@ -85,1 +85,6 @@};
  \node[diffline, matched]  (N4a) at (0.10, 3.85) {- if len(c)==1 and};
  \node[diffline, matched]  (N4b) at (0.10, 3.50) {-\ \ \ not isinst(c[0],Q):};
  \node[diffline, unmatched](N4c) at (0.10, 3.15) {+ if (};
  \node[diffline, unmatched](N4d) at (0.10, 2.80) {+\ \ \ len(c)==1 and};
  \node[diffline, unmatched](N4e) at (0.10, 2.45) {+\ \ \ not isinst(c[0],Q) and};
  \node[diffline, unmatched](N4f) at (0.10, 2.10) {+\ \ \ isinst(c[0],tuple) and};
  \node[diffline, unmatched](N4g) at (0.10, 1.75) {+\ \ \ len(c[0]) == 2};
  \node[diffline, unmatched](N4h) at (0.10, 1.40) {+ ):};
  \node[noteline]   (NN4c) at (0.10, 0.95) {final $x$: unresolved};
  \node[noteline]   (NN4d) at (0.10, 0.60) {`found-it-then-lost-it'};
  \begin{scope}[on background layer]
    \node[panelbox, fit=(T4)(D4)(M4)(H4a)(NN4d), inner sep=5pt] {};
  \end{scope}
\end{scope}
\end{tikzpicture}%
}
\caption{\textbf{Found-it-then-lost-it.} In this Gemini~3~Flash trajectory
on \texttt{django\_\_django-14140}, the first edit exactly reproduces a popular
12-feature rewrite in $\tilde{\mathcal{S}}_i$ (22 of $\sim$\,$30$ resolved
runs, also the dataset gold), so $D$ drops from $1$ to $0$ at cycle $0.15$.
The next edit reverts that rewrite and substitutes a multi-condition guard
that overlaps only one reference feature, so $D$ jumps back to $0.917$
(rounded to $0.9$); the upward segment in the curve is drawn red. Later edits
refine the guard but never recover the matching rewrite, and the run finishes
unresolved at $D{=}0.9$. The lower panels show the corresponding repository
states before editing, after the correct edit, after the destructive edit, and
at termination.}
\label{fig:solution-distance-thrash-gemflash}
\end{figure}

\paragraph{Example 5. Revert as discovery: a deliberate A/B test that uncovers a second bug.}
Not every backtrack is a mistake, and not every revert is a recovery
from one. Figure~\ref{fig:solution-distance-pingpong-astropy} shows a
Gemini~3.1~Pro trajectory whose middle event is a deliberate
\texttt{git checkout}, a revert the agent uses as a controlled
experiment. The revert surfaces a
\emph{second}, unrelated bug that the agent's first fix would otherwise
have shipped past. The instance is \texttt{astropy-12907} (same as discussed in Example\,1 for GPT\,5.4), whose gold
patch contains only one change in
\texttt{astropy/modeling/separable.py} (see Figure\,\ref{fig:solution-distance-pingpong-astropy} leftmost panel). The fix is enough to pass the
hidden test suite. This Gemini~3.1~Pro trajectory, also identifies a second,
unrelated bug in the same file: \texttt{\_coord\_matrix} calls
\texttt{np.roll(mat, k)} without \texttt{axis{=}0}, which flattens
before rolling and breaks compound models whose right child has
multiple outputs.

Edit~\#1 (normalized cycle $t=0.13$) is the obvious one: the agent replaces
\texttt{cright[\,..\,] = 1} with \texttt{cright[\,..\,] = right} and
its reproduction script passes. The benchmark gold solution is matched exactly, so
$D$ drops from $1$ to $0$. At this point the run is already
oracle-correct and a submission here would pass the hidden tests.

Rather than submit, the agent constructs a sharper compound case, and notices that \texttt{separability\_matrix}
still returns a result that does not look block-separable. Edit~\#2
(cycle $t=0.27$) is a \texttt{git checkout} on \texttt{separable.py}, and $D$ jumps back to $1$ (upward
segment drawn red). The agent then walks into \texttt{\_coord\_matrix}, isolates the rolled
coordinate matrix with a four-line probe, and observes that
\texttt{np.roll(mat, k)} is flattening the 2-D array before rolling.
Edit~\#3 (cycle $0.61$) adds \texttt{,\,axis{=}0} to that call. The
parseable repository state now contains the two \texttt{np.roll}
features but is still missing the reverted \texttt{\_cstack} features, which is later added at cycle $t_5$.

This is in contrast with Example\,4 (Gemini~3~Flash on
\texttt{django-14140}) which is a destructive backtrack, or the agent finds the
solution, edits away from it, and never recovers. The present run is
the opposite pattern, the revert is deliberate, and the post-revert
trajectory introduces a new feature (\texttt{np.roll axis{=}0}) that
the pre-revert state never contained. The resolved final state is
strictly richer in patch features than the state at $t_2$ even though
both achieve $D{=}0$. However, since the intended task is not regarding fixing the \texttt{np.roll(mat, k)}, this additional work 
has caused delayed distance reduction to reach the solution set $\mathcal{S}_i$. It is apparant from Figure~\ref{fig:metrics-sbv-soldist} that $D(t)$ decays much slower for Gemini-3.1 Pro compared to counterparts (Opus 4.6, GPT 5.4).

\begin{figure}[p]
\centering
\resizebox{\textwidth}{!}{%
\begin{tikzpicture}[
  every node/.style={inner sep=2pt, outer sep=0pt},
  panelbox/.style={draw=black!45, line width=0.4pt, rounded corners=2pt, fill=white},
  stagelabel/.style={font=\sffamily\bfseries\footnotesize, anchor=north, align=center},
  modelabel/.style={font=\sffamily\scriptsize\itshape, anchor=north, text=black!60, align=center},
  dvalbad/.style={font=\sffamily\bfseries\small, anchor=north, text=red!65!black},
  dvalmid/.style={font=\sffamily\bfseries\small, anchor=north, text=orange!75!black},
  dvalok/.style={font=\sffamily\bfseries\small, anchor=north, text=green!45!black},
  diffline/.style={font=\ttfamily\tiny, anchor=west, text width=3.05cm, minimum height=3.2mm, inner xsep=2pt, inner ysep=0pt, align=left},
  matched/.style={fill=green!18, draw=none, text=black},
  unmatched/.style={fill=gray!8, draw=none, text=black!55},
  hunkhdr/.style={font=\ttfamily\tiny\bfseries, anchor=west, text width=3.05cm, minimum height=3.2mm, inner xsep=2pt, inner ysep=0pt, align=left, fill=none, draw=none, text=violet!60!black},
  noteline/.style={font=\sffamily\scriptsize, anchor=west, text width=3.05cm, minimum height=3.2mm, inner xsep=2pt, inner ysep=0pt, align=left, text=black!55, draw=none, fill=none},
  tklabel/.style={font=\sffamily\bfseries\tiny, text=black!80},
  midarrow/.style={postaction={decorate, decoration={markings, mark=at position 0.5 with {\arrow{Stealth[length=4pt,width=4pt]}}}}},
]
\begin{scope}[shift={(1.5, 8.6)}]
  \draw[gray!55, line width=0.5pt] (0, 0) -- (14.2, 0);
  \draw[gray!55, line width=0.5pt] (0, 0) -- (0, 2.3);
  \foreach \x/\lab in {0/0, 7/0.5, 14/1} {
    \draw[gray!55] (\x, 0) -- (\x, -0.07);
    \node[below=1pt, font=\tiny, text=black!70] at (\x, 0) {\lab};
  }
  \foreach \y/\lab in {0/0, 1.0/0.5, 2.0/1} {
    \draw[gray!55] (0, \y) -- (-0.07, \y);
    \node[left=1pt, font=\tiny, text=black!70] at (0, \y) {\lab};
  }
  \node[below, font=\scriptsize, text=black!75] at (7, -0.45) {$t$};
  \node[rotate=90, above, font=\scriptsize, text=black!75] at (-0.7, 1.0) {$D(t)$};
  \draw[gray!20, dashed, very thin] (0, 2.0) -- (14, 2.0);
  \draw[gray!20, dashed, very thin] (0, 1.0) -- (14, 1.0);
  \draw[blue!60!black, line width=0.9pt, midarrow] (0, 2.0) -- (1.83, 2.0);
  \draw[blue!60!black, line width=0.9pt] (1.83, 2.0) -- (1.83, 0.0);
  \draw[blue!60!black, line width=0.9pt, midarrow] (1.83, 0.0) -- (3.84, 0.0);
  \draw[red!75!black, line width=1.1pt] (3.84, 0.0) -- (3.84, 2.0);
  \draw[blue!60!black, line width=0.9pt, midarrow] (3.84, 2.0) -- (8.50, 2.0);
  \draw[blue!60!black, line width=0.9pt] (8.50, 2.0) -- (8.50, 1.0);
  \draw[blue!60!black, line width=0.9pt, midarrow] (8.50, 1.0) -- (13.16, 1.0);
  \draw[blue!60!black, line width=0.9pt] (13.16, 1.0) -- (13.16, 0.0);
  \draw[blue!60!black, line width=0.9pt, midarrow] (13.16, 0.0) -- (14, 0.0);
  \filldraw[blue!60!black] (0, 2.0) circle (1.6pt);
  \filldraw[blue!60!black] (1.83, 0.0) circle (1.6pt);
  \filldraw[red!75!black]  (3.84, 2.0) circle (1.8pt);
  \filldraw[blue!60!black] (8.50, 1.0) circle (1.6pt);
  \filldraw[blue!60!black] (13.16, 0.0) circle (1.6pt);
  \node[tklabel, above=2pt] at (0, 2.0) {$t_1$};
  \node[tklabel, above right=1pt] at (1.83, 0.0) {$t_2$};
  \node[tklabel, above right=1pt, text=red!75!black] at (3.84, 2.0) {$t_3$};
  \node[tklabel, above right=1pt] at (8.50, 1.0) {$t_4$};
  \node[tklabel, above right=1pt] at (13.16, 0.0) {$t_5$};
\end{scope}
\begin{scope}[shift={(0, 0)}]
  \node[stagelabel] (T1) at (1.65, 7.60) {$x(t_1)$};
  \node[dvalbad]    (D1) at (1.65, 6.70) {$D = 1.0$};
  \node[modelabel]  (M1) at (1.65, 6.20) {best mode: gold\\(tie-break, $D{=}1$)};
  \node[noteline]   (NN1a) at (0.10, 5.30) {\textit{no edits applied}};
  \node[noteline]   (NN1b) at (0.10, 4.95) {$0/2$ gold features};
  \node[hunkhdr]            (H1a) at (0.10, 4.20) {astropy/modeling/separable.py};
  \node[hunkhdr]            (H1b) at (0.10, 3.85) {@@ \_cstack @@};
  \node[diffline, unmatched](N1a) at (0.10, 3.50) {-\ cright[..] = 1};
  \node[diffline, unmatched](N1b) at (0.10, 3.15) {+\ cright[..] = right};
  \node[noteline]   (NN1c) at (0.10, 2.50) {gold $\tau$: 2 \_cstack features};
  \node[noteline]   (NN1d) at (0.10, 2.15) {\_coord\_matrix bug exists in};
  \node[noteline]   (NN1e) at (0.10, 1.80) {code but not in best mode};
  \begin{scope}[on background layer]
    \node[panelbox, fit=(T1)(D1)(M1)(H1a)(NN1e), inner sep=5pt] {};
  \end{scope}
\end{scope}
\begin{scope}[shift={(3.5, 0)}]
  \node[stagelabel] (T2) at (1.65, 7.60) {$x(t_2)$};
  \node[dvalok]     (D2) at (1.65, 6.70) {$D = 0.0$};
  \node[modelabel]  (M2) at (1.65, 6.20) {best mode: gold\\(recall $2/2$)};
  \node[noteline]   (NN2a) at (0.10, 5.30) {\textit{edit \#1: \_cstack fix}};
  \node[noteline]   (NN2b) at (0.10, 4.95) {$2/2$ gold features};
  \node[hunkhdr]            (H2a) at (0.10, 4.20) {astropy/modeling/separable.py};
  \node[hunkhdr]            (H2b) at (0.10, 3.85) {@@ \_cstack @@};
  \node[diffline, matched]  (N2a) at (0.10, 3.50) {-\ cright[..] = 1};
  \node[diffline, matched]  (N2b) at (0.10, 3.15) {+\ cright[..] = right};
  \node[noteline]   (NN2c) at (0.10, 2.50) {headline case fixed;};
  \node[noteline]   (NN2d) at (0.10, 2.15) {oracle-correct here ---};
  \node[noteline]   (NN2e) at (0.10, 1.80) {agent probes further};
  \begin{scope}[on background layer]
    \node[panelbox, fit=(T2)(D2)(M2)(H2a)(NN2e), inner sep=5pt] {};
  \end{scope}
\end{scope}
\begin{scope}[shift={(7.0, 0)}]
  \node[stagelabel] (T3) at (1.65, 7.60) {$x(t_3)$};
  \node[dvalbad]    (D3) at (1.65, 6.70) {$D = 1.0$};
  \node[modelabel]  (M3) at (1.65, 6.20) {best mode: gold\\(tie-break, $D{=}1$)};
  \node[noteline]   (NN3a) at (0.10, 5.30) {\textit{edit \#2: git checkout}};
  \node[noteline]   (NN3b) at (0.10, 4.95) {$0/2$ gold features};
  \node[hunkhdr]            (H3a) at (0.10, 4.20) {astropy/modeling/separable.py};
  \node[hunkhdr]            (H3b) at (0.10, 3.85) {@@ \_cstack @@};
  \node[diffline, unmatched](N3a) at (0.10, 3.50) {-\ cright[..] = 1\ \ \ (back)};
  \node[diffline, unmatched](N3b) at (0.10, 3.15) {+\ cright[..] = right};
  \node[noteline]   (NN3c) at (0.10, 2.50) {revert reveals \texttt{cm2 \& cm}};
  \node[noteline]   (NN3d) at (0.10, 2.15) {still fails $\Rightarrow$ separate};
  \node[noteline]   (NN3e) at (0.10, 1.80) {bug in \_coord\_matrix};
  \begin{scope}[on background layer]
    \node[panelbox, fit=(T3)(D3)(M3)(H3a)(NN3e), inner sep=5pt] {};
  \end{scope}
\end{scope}
\begin{scope}[shift={(10.5, 0)}]
  \node[stagelabel] (T4) at (1.65, 7.60) {$x(t_4)$};
  \node[dvalmid]    (D4) at (1.65, 6.70) {$D = 0.5$};
  \node[modelabel]  (M4) at (1.65, 6.20) {best mode: 4-feat minority\\(recall $2/4$)};
  \node[noteline]   (NN4a) at (0.10, 5.30) {\textit{edit \#3: np.roll axis=0}};
  \node[noteline]   (NN4b) at (0.10, 4.95) {$2/4$ minority features};
  \node[hunkhdr]            (H4a) at (0.10, 4.20) {astropy/modeling/separable.py};
  \node[hunkhdr]            (H4b) at (0.10, 3.85) {@@ \_cstack @@};
  \node[diffline, unmatched](N4a) at (0.10, 3.50) {-\ cright[..] = 1\ \ \ (back)};
  \node[diffline, unmatched](N4b) at (0.10, 3.15) {+\ cright[..] = right};
  \node[hunkhdr]            (H4c) at (0.10, 2.65) {@@ \_coord\_matrix @@};
  \node[diffline, matched]  (N4c) at (0.10, 2.30) {-\ np.roll(mat,\,k)};
  \node[diffline, matched]  (N4d) at (0.10, 1.95) {+\ np.roll(..,\,axis=0)};
  \node[noteline]   (NN4c) at (0.10, 1.30) {2nd bug fix lands;};
  \node[noteline]   (NN4d) at (0.10, 0.95) {cstack re-apply via heredoc (unparsed)};
  \begin{scope}[on background layer]
    \node[panelbox, fit=(T4)(D4)(M4)(H4a)(NN4d), inner sep=5pt] {};
  \end{scope}
\end{scope}
\begin{scope}[shift={(14.0, 0)}]
  \node[stagelabel] (T5) at (1.65, 7.60) {$x(t_5)$};
  \node[dvalok]     (D5) at (1.65, 6.70) {$D = 0.0$};
  \node[modelabel]  (M5) at (1.65, 6.20) {best mode: gold\\(gap-attribution closes)};
  \node[noteline]   (NN5a) at (0.10, 5.30) {\textit{final \texttt{sra\_patch.patch}}};
  \node[noteline]   (NN5b) at (0.10, 4.95) {$2/2$ gold $+$ \texttt{np.roll}};
  \node[hunkhdr]            (H5a) at (0.10, 4.20) {astropy/modeling/separable.py};
  \node[hunkhdr]            (H5b) at (0.10, 3.85) {@@ \_cstack @@};
  \node[diffline, matched]  (N5a) at (0.10, 3.50) {-\ cright[..] = 1};
  \node[diffline, matched]  (N5b) at (0.10, 3.15) {+\ cright[..] = right};
  \node[hunkhdr]            (H5c) at (0.10, 2.65) {@@ \_coord\_matrix @@};
  \node[diffline, matched]  (N5c) at (0.10, 2.30) {-\ np.roll(mat,\,k)};
  \node[diffline, matched]  (N5d) at (0.10, 1.95) {+\ np.roll(..,\,axis=0)};
  \node[noteline]   (NN5c) at (0.10, 1.30) {final $x$: resolved;};
  \node[noteline]   (NN5d) at (0.10, 0.95) {strictly richer than $x(t_2)$};
  \begin{scope}[on background layer]
    \node[panelbox, fit=(T5)(D5)(M5)(H5a)(NN5d), inner sep=5pt] {};
  \end{scope}
\end{scope}
\end{tikzpicture}%
}
\caption{\textbf{Revert as discovery.} Gemini~3.1~Pro run on
\texttt{astropy-12907} (SWE-Bench-Verified), the agent first reaches the gold-mode fix, deliberately
reverts it to test the unfixed baseline, and uncovers an pre-existing unrelated bug. Faithful replay records the productive backtrack
($D:1\!\to\!0\!\to\!1\!\to\!0.5\!\to\!0$): the final resolved patch combines the
original \texttt{\_cstack} fix with a new \texttt{np.roll(..., axis=0)} fix.}
\label{fig:solution-distance-pingpong-astropy}
\end{figure}

\FloatBarrier

\subsubsection{Faithful state replay}
\label{app:metrics:soldist:replay}

The empirical approximation of Eq.~\ref{eq:distance-main} requires the agent's actual repository state after each edit event. A simple heuristic of using a monotonic surrogate (e.g., credit each final-patch feature at the cycle position it first appears in any tool call's text) is computationally cheap but cannot increase and decrease in value. It cannot capture an agent that lands the fix and subsequently retracts it. Therefore, we reconstruct the state from scratch by parsing every edit call, applying it to in-memory file contents, and recomputing $d(t)$ at each step.

\paragraph{Cross-tool edit identification.}
In SSA, models edit the code through different tool surfaces. We map all of them to a unified operation list $[(f, \text{op}, \text{payload}), \ldots]$. Table~\ref{tab:soldist:appliers} summarises the supported tool families and the corresponding parsers.

\begin{table}[h]
\centering
\small
\begin{tabular}{@{}>{\raggedright\arraybackslash}p{3.5cm}>{\raggedright\arraybackslash}p{2.5cm}>{\raggedright\arraybackslash}p{6.8cm}@{}}
\toprule
Tool / family & Models & Handled edit forms \\
\midrule
\texttt{str\_replace\_editor} & Claude, Qwen, Gemini & \texttt{str\_replace}, \texttt{create}, \texttt{insert}, \texttt{undo\_edit} \\
\texttt{file\_edit} & Grok & same shape as \texttt{str\_replace\_editor} \\
\texttt{file\_write} & Grok & whole-file write \\
\texttt{apply\_patch} & GPT-OSS & \texttt{Update File}, \texttt{Add File}, \texttt{Delete File} \\
\texttt{shell}, \texttt{bash} & GPT family and shell-tool users & \texttt{apply\_patch} heredocs, \texttt{cat} heredoc overwrites. Git reverts handled below \\
\bottomrule
\end{tabular}
\caption{Edit-tool interfaces in SSA and the corresponding faithful-replay parsers.}
\label{tab:soldist:appliers}
\end{table}

\paragraph{Bash-based edit extraction.}
For shell-style tools, the replay parser scans the command text. For example, if a command contains one or more
\texttt{apply\_patch} heredocs, each heredoc body is parsed with
the same \texttt{Update/Add/Delete File} parser used for the structured
\texttt{apply\_patch} tool. If the command contains
\texttt{cat \textgreater{} file \textless{}\textless{}TAG ... TAG}, replay treats it as a whole-file overwrite
of \texttt{file}. The bash edits identification is deliberately narrower than the behavioral edit classifier. The judge
and path-auditor recognize broader bash mutations such as \texttt{sed -i},
\texttt{perl -pi}, and wrapper scripts that open a source file for writing.
Faithful replay does not execute arbitrary shell or Python, so those commands
are not applied as exact intermediate states. For resolved runs, any own-patch
features still missing from the reconstructed state are treated as gap features.
Replay assigns each missing features to the last parsed edit on its file, or, if no
such edit exists, to the last bash command that textually mentioned that file.
Those features are then added for distance evaluation after that cycle position.
Thus, an un-parsed script edit can still reach $D(t_{\text{end}})=0$ through the
self-anchor, but by attribution rather than by exact replay.

\paragraph{Fuzzy matching.}
The standalone editing tools in SSA, for example, \texttt{str\_replace\_editor} and \texttt{apply\_patch}, tolerate whitespace and indent drift between the model's quoted context and the on-disk content. An exact-string applier in our faithful replay, therefore, under-applies edits that the real tool accepted. We first try an exact replacement. If that fails, \texttt{fuzzy\_replace} normalizes each non-blank line by collapsing whitespace, then searches for a contiguous file-line window whose normalized non-blank lines equal the quoted old block. The same fallback is used inside \texttt{apply\_patch} update hunks when the hunk's old block is not found exactly.

\paragraph{Revert handling.}
Agents undo edits through multiple channels and we track three revert categories: (1) \texttt{undo\_edit} from the text-editor family (see Table\,\ref{tab:soldist:appliers}), we maintain a per-file backup of pre-edit snapshots and pop on \texttt{undo\_edit} call, (2) \textit{file-scoped git reverts} parsed from shell/bash, such as \texttt{git checkout <file>} and \texttt{git restore <file>}, which reset the named file to its base contents, and (3) \textit{tree-scoped git reverts}, \texttt{git reset --hard} and \texttt{git checkout .}, which resets every edited file to base state. Note, one pattern we deliberately exclude is \emph{paired} \texttt{git stash} $\to$ \texttt{git stash pop}. This is a transient shelve-to-test maneuver (i.e., run tests on a clean tree and then restore), and hence we ignore counting its momentary dip as backtracking. A simple audit of the trajectories shows it is essentially always paired, so the final state is unaffected and the dip is purely an artefact of the intermediate state.

\subsubsection{Resolved endpoints: self-anchor}
\label{app:metrics:soldist:zero}

A resolved run for an instance $r$ is in the solution space by construction as its
final submitted patch passed the oracle. Therefore, the correct endpoint
for the resolved trajectory is $D(t_{\text{end}})=0$. We tested strict
leave-one-run-out (LOO) scoring as an intuitive diagnostic: score a run only
against solutions independently observed in other runs. This did not work for
our reported trajectory curves. The empirical solution set
$\tilde{\mathcal{S}}_r$ is a finite sample, not a dense cover of the true
solution space. Nearby correct solutions can differ in variable names, hunk
boundaries, guard placement, helper structure, or surrounding context, and the
nearest textual variant may simply be absent from the other observed runs. In
practice, LOO left some resolved trajectories with $D(t_{\text{end}})>0$, which
disturbed the average resolved $D(t)$ curve even though those runs solved their
instances.

For the trajectory curves we therefore use a self-anchor, or any resolved run $r$ has its corresponding generated patch alwsys in the empirical solution set $\tilde{\mathcal{S}_r}$. 
Unresolved runs receive no
self-anchor, since their final patches are not oracle-verified solutions.

Note that this is not an extra credit for solved runs, it is the endpoint-consistency
condition implied by the definition of the solution space. Hence, the nearest-reference recall
at the endpoint is $1$, and $D_r(t_\text{end})=0$ by construction. Without
self-anchor, a positive endpoint distance can reflect sparsity in the empirical
reference set $\tilde{\mathcal{S}_r}$ rather than a real failure to reach the solution set $\mathcal{S}_i$.

\subsubsection{Backtracking quantification}
\label{app:metrics:soldist:backtrack}

The main-text quantity $\bar B(t)$ is an instantaneous \emph{backtracking
depth}.  Writing the corresponding per-run depth as
$B_r(t):=D_r(t)-M_r(t)$ gives $\bar B(t)=N^{-1}\sum_r B_r(t)$.
Interestingly, faithful replay also supports a complementary local diagnostic.  For
successive successful edit events at $t_k$ and $t_{k+1}$, define
\begin{equation}
  \Delta D_{r,k}:=D_r(t_{k+1})-D_r(t_k),\qquad k=1,2,\ldots .
  \label{eq:soldist:local-backtrack}
\end{equation}
The two notions of Eq.(\ref{eq:soldist:cohort-refinement:crossing-time}) and Eq.(\ref{eq:soldist:local-backtrack}) are linked exactly by
\begin{equation}
  B_r(t_{k+1})
  =\bigl[B_r(t_k)+\Delta D_{r,k}\bigr]_+,
  \qquad\text{where}, [z]_+:=\max\{z,0\}.
  \label{eq:soldist:backtrack-recurrence}
\end{equation}
Thus, $\Delta D_{r,k}>0$ is a \emph{local backtracking event} that increases
the depth, while a negative change reduces the depth and may leave it
positive while the run recovers.  Accordingly, $\bar B(t)$ measures current
unrecovered depth, while the following summaries measure the frequency and
magnitude of local uphill edits.
\begin{description}
  \item[Backtrack fraction] the share of trajectories for which some
    $\Delta D_{r,k}>0$.
  \item[Backtrack-edit fraction]
    $\#\{(r,k):\Delta D_{r,k}>0\}/
     \#\{(r,k):|\Delta D_{r,k}|>0\}$.
  \item[Uphill mass]
    $\sum_{r,k}[\Delta D_{r,k}]_+/
     \sum_{r,k}|\Delta D_{r,k}|$.
\end{description}
The first counts trajectories that contain at least one local backtracking
event, while the second counts such events among distance-moving edits.  The
third weights them by size, so one large reversal contributes as much as many
small ones.  Per-model values are reported with the $\Delta D$ histograms in
Figures~\ref{fig:metrics-sbv-backtracking} and~\ref{fig:metrics-sbv-backtracking-b}
(green = progress edits, red = local backtracking events, annotated percentage
= backtrack-edit fraction).

\subsubsection{Limitations of the solution-distance metric}
\label{app:metrics:soldist:limits}

The recall-divergence metric (Eq.\,\ref{eq:distance-main}) we have described is deliberately structural rather than semantic, and inherits several limitations that we now list below.

\begin{enumerate}[leftmargin=*, itemsep=2pt, topsep=2pt]
  \item \textbf{Textual overlap, not semantics.} The recall fraction in Eq.~\ref{eq:soldist:progress} counts matching changed lines. A correct-but-textually-different fix, i.e., a different identifier choice, a refactored expression, a guard placed elsewhere scores $< 1$ against reference modes it does not textually match. The empirical subsets and the self-anchor for resolved endpoints mitigate this substantially, but the underlying limitation remains. For unresolved runs and intermediate states, a semantic distance (AST alignment, embedding similarity, or an LLM-judge ``fraction-of-fix-achieved'') would score the agent's intent rather than its output text.
  \item \textbf{Empirical subset $\neq$ solution space.} The space is defined by the test oracle and $\tilde{\mathcal{S}}_i$ approximates it with the patches we happened to observe (using 21 models run 5 times). A unique correct solution found by nobody else, a real possibility on novel instances, can be far from the observed empirical modes, which is why the self-anchor is reserved for resolved runs whose own patches are oracle-verified.
  \item \textbf{Reference-set sparsity scales with difficulty.} $\tilde{\mathcal{S}}_i$ is dense on easy instances (for example, in SWE-Bench-Verified $80+$ reference modes) and sparse on hard ones. Instances that no model in our sweep solved have only the gold patch as a reference (or no reference if gold is also missing). $D$ on these instances reads against a thin reference set and should be interpreted with care.
  \item \textbf{Replay fidelity.} $d(t)$ is reconstructed from the edit-tool calls we parse (Table~\ref{tab:soldist:appliers}). Therefore, exotic shell rewrites (e.g.\ Python scripts that open a source file in write mode) are not fully parsed. Self-anchor fixes the resolved endpoint, but it cannot recover the exact intermediate state of an unparsed edit, so a small number of trajectories' mid-run shapes remain approximate.
\end{enumerate}

\subsection{Metrics}
\label{app:metrics:metrics}

This subsection catalogs every per-trajectory and per-cycle metric either reported directly by SSA or extracted via the LLM judge or simple computations. Table~\ref{tab:metrics-inventory} gives a symbol or short name used elsewhere in this report, a one-line definition, and the figure(s) in which it is plotted.

\begin{longtable}{@{}>{\raggedright\arraybackslash}p{3.0cm}>{\raggedright\arraybackslash}p{7.4cm}>{\raggedright\arraybackslash}p{2.6cm}@{}}
\caption{Metrics glossary. Every per-trajectory / per-cycle quantity reported by SSA and the figure(s) in which it appears. ``Judge R\#'' refers to the rubric field defined in Table~\ref{tab:judge-buckets}.}\label{tab:metrics-inventory}\\
\toprule
Metric & Definition & Figures \\
\midrule
\endfirsthead
\multicolumn{3}{c}{\tablename\ \thetable{} -- continued}\\
\toprule
Metric & Definition & Figures \\
\midrule
\endhead
\midrule\multicolumn{3}{r}{continued on next page}\\
\endfoot
\bottomrule
\endlastfoot
\textbf{pass@1} & Mean resolution rate across the $n$ runs $\times$ $k$ instances per (model, benchmark); the fraction whose final state passes the benchmark's tests. & Table~\ref{tab:results-reproduction}; Appendix~\ref{app:passrates} \\
\textbf{Total output tokens} & Sum of \texttt{output\_tokens} across all $(run, instance)$ pairs for a given (model, benchmark) & Figures~\ref{fig:metrics-sbv-pass-vs-tokens}, \ref{fig:metrics-sbp-pass-vs-tokens}, \ref{fig:metrics-tb2-pass-vs-tokens} \\
\textbf{$D(t)$, solution distance} & Main-text recall divergence (Eq.~\ref{eq:distance-main}) evaluated on the empirical solution subset $\tilde{\mathcal{S}}_i$ using Eq.~\ref{eq:soldist:progress}. Distance is from the live state to the nearest verified-correct reference mode, with self-anchor for resolved endpoints. & Figures~ \ref{fig:metrics-sbv-soldist}, \ref{fig:metrics-sbv-soldist-b}, \ref{fig:metrics-sbp-soldist}, \ref{fig:metrics-sbp-soldist-b} \\
\textbf{$\Delta D_{r,k}$, local change} & $D_r(t_{k+1})-D_r(t_k)$ for successive successful edit events. A negative change is progress; a positive change is a local backtracking event. & Figures~\ref{fig:metrics-sbv-backtracking}, \ref{fig:metrics-sbv-backtracking-b}, \ref{fig:metrics-sbp-backtracking}, \ref{fig:metrics-sbp-backtracking-b} \\
\textbf{Backtrack-edit fraction} & Share of distance-moving edits with $\Delta D_{r,k}>10^{-9}$; changes with $|\Delta D_{r,k}|\le10^{-9}$ are excluded. & Figures~\ref{fig:metrics-sbv-backtracking}, \ref{fig:metrics-sbv-backtracking-b}, \ref{fig:metrics-sbp-backtracking}, \ref{fig:metrics-sbp-backtracking-b} (annotated) \\
\textbf{Backtrack fraction} & Share of trajectories with any $\Delta D_{r,k}>10^{-9}$, i.e., how often trajectories contain at least one local backtracking event. & cohort summary (text, \S\ref{app:metrics:sbv}) \\
\textbf{Uphill mass} & $\sum_{r,k}[\Delta D_{r,k}]_+/\sum_{r,k}|\Delta D_{r,k}|$, local backtracking weighted by step size. & cohort summary (\S\ref{app:metrics:sbv}) \\
\textbf{Edit/test ratio} & Per-cycle ratio of edit-oriented calls (\texttt{edit\_source}, \texttt{edit\_test}, \texttt{write\_scratch}) to read-oriented calls (\texttt{read\_code}, \texttt{search\_locate}, \texttt{explore\_navigate}); aggregated as a distribution per model. & Figures~\ref{fig:metrics-sbv-editread}, \ref{fig:metrics-sbv-editread-b}, \ref{fig:metrics-sbp-editread}, \ref{fig:metrics-sbp-editread-b}, \ref{fig:metrics-tb2-editread}, \ref{fig:metrics-tb2-editread-b} \\
\textbf{Genuine tool-error rate} & Per-cycle share of calls with R8 $=$ \texttt{genuine}, broken out by tool family (\texttt{bash} vs.\ \texttt{str\_replace\_editor}); excludes benign non-zero exits (\texttt{grep} no-match, \texttt{pytest} reporting test failures, etc.). & Figures~\ref{fig:metrics-sbv-toolerr}, \ref{fig:metrics-sbv-toolerr-b}, \ref{fig:metrics-sbp-toolerr}, \ref{fig:metrics-sbp-toolerr-b}, \ref{fig:metrics-tb2-toolerr}, \ref{fig:metrics-tb2-toolerr-b} \\
\textbf{Phase composition} & Share of cycles in each R6 phase (\texttt{explore}, \texttt{localize}, \texttt{implement}, \texttt{verify}, \texttt{other}); stacked-area over normalised cycle position. & Figures~\ref{fig:metrics-sbv-phase}, \ref{fig:metrics-sbv-phase-b}, \ref{fig:metrics-sbp-phase}, \ref{fig:metrics-sbp-phase-b}, \ref{fig:metrics-tb2-phase}, \ref{fig:metrics-tb2-phase-b} \\
\textbf{Token economics} & Per-cycle \texttt{input\_tokens}, \texttt{output\_tokens}, \texttt{reasoning\_tokens}, \texttt{cache\_read\_input\_tokens}, \texttt{cache\_write\_input\_tokens}, plus \texttt{cache\_hit\_rate}. & cost / cache analyses \\
\textbf{Latency} & \texttt{ttft\_ms} (time-to-first-token) and \texttt{cycle\_durations\_sec}; aggregated into \texttt{execution\_time\_seconds}. & cost analyses \\
\textbf{Per-tool counters} & \texttt{call\_count}, \texttt{success\_count}, \texttt{error\_count}, \texttt{total\_time}, \texttt{average\_time}, derived \texttt{success\_rate}; measured per tool name. & Figures~\ref{fig:metrics-sbv-tooldist}, \ref{fig:metrics-sbv-tooldist-b}, \ref{fig:metrics-sbp-tooldist}, \ref{fig:metrics-sbp-tooldist-b}, \ref{fig:metrics-tb2-tooldist}, \ref{fig:metrics-tb2-tooldist-b} \\
\end{longtable}

The metrics fall into three families. 
\textbf{(i) Outcome and cost} (pass@1, total output tokens) describe what the run achieved and what it cost. They come directly from the harness and the benchmark-specific evaluator. 
\textbf{(ii) Trajectory geometry} ($D(t)$ and the $\Delta D$-derived scalars) describe how the live repository state evolved over time relative to the empirical solution subset. They require the faithful-replay pipeline of Sections~\ref{app:metrics:soldist:replay}--\ref{app:metrics:soldist:zero}. \textbf{(iii) Behavioral composition} (edit ratio, genuine tool-error rate, phase composition) describe \emph{what} the agent was doing inside each cycle. They require the per-call LLM judge of Section~\ref{app:metrics:judge}.

\paragraph{Definition of pass@k.}
We follow the unbiased combinatorial estimator introduced by \citet{chen2021codex}. Let $n$ be the number of independent samples generated per instance and let $c \leq n$ be the number of those samples that pass the oracle. Then
\begin{equation}
  \mathrm{pass@k} \;=\; \mathbb{E}_{\text{instances}}\!\left[\, 1 \;-\; \dfrac{\binom{n-c}{k}}{\binom{n}{k}} \,\right],
  \label{eq:passatk}
\end{equation}
where the expectation is the mean over instances in the benchmark and $\binom{n-c}{k} = 0$ whenever $n-c < k$. The estimator is unbiased and is numerically stable to compute as $1 - \prod_{i=0}^{k-1} (n-c-i)/(n-i)$ when $n-c \geq k$. We report $\mathrm{pass@1}$ in Table~\ref{tab:results-reproduction} and Appendix~\ref{app:passrates}, with $95\%$ confidence intervals computed across the $5 \times |\text{instances}|$ trials per (model, benchmark). For $\mathrm{pass@2}$ we use $n=5$ runs per instance, so an instance contributes $1 - \binom{5-c}{2}/\binom{5}{2}$ for $c \in \{0, 1, \ldots, 5\}$ correct samples, then average over instances.

Sections~\ref{app:metrics:sbv}--\ref{app:metrics:tb2} below break the metrics down per benchmark, each in its own subsection. Plots are computed by aggregating the per-cycle counters above across the five runs we collect per (model, benchmark), then averaging at the model level (token totals are summed across all $(run, instance)$ pairs.

\subsection{Metrics: SWE-Bench-Verified}
\label{app:metrics:sbv}

\begin{figure}[h]
    \centering
    \includegraphics[width=\linewidth]{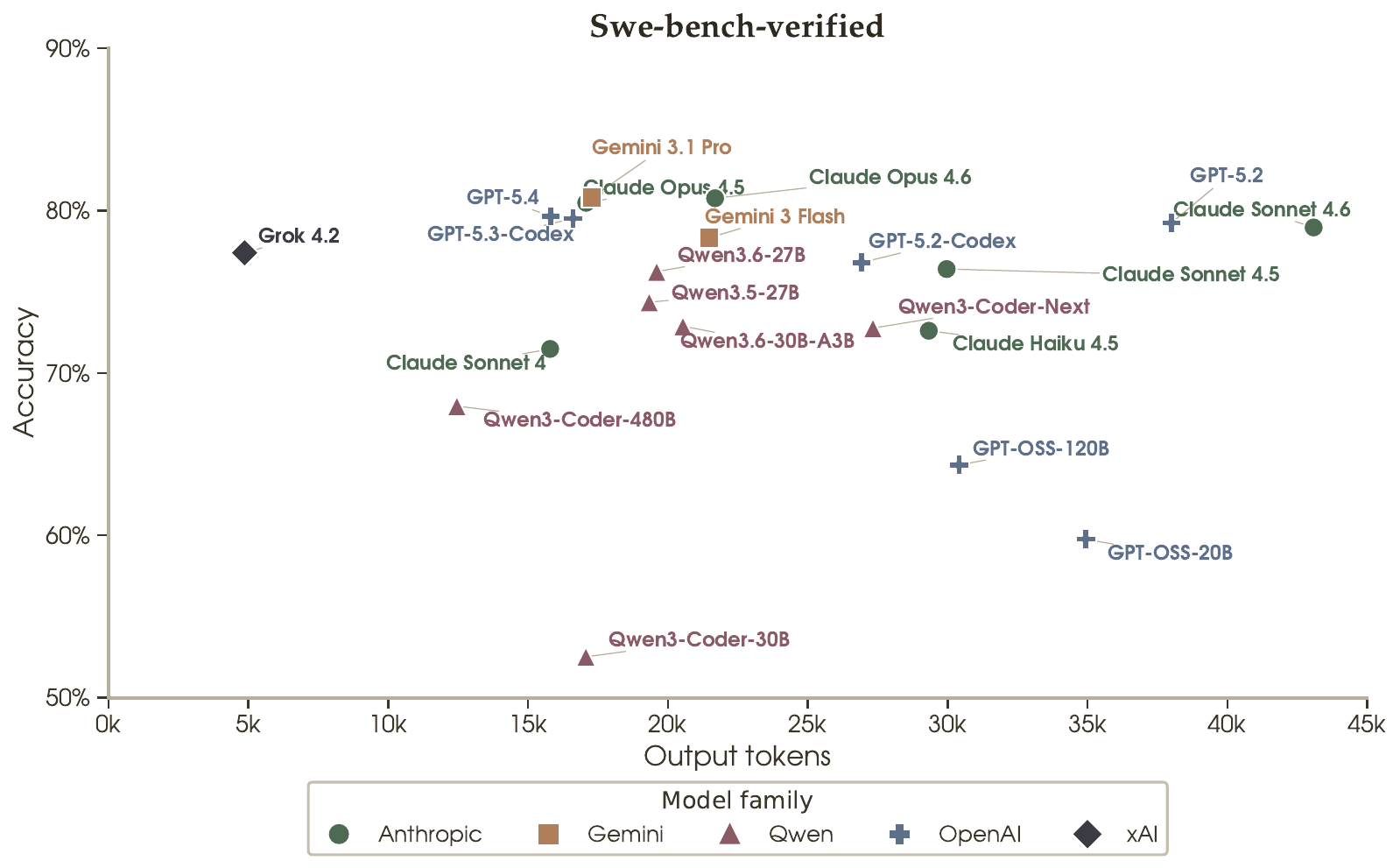}
    \caption{SWE-Bench-Verified: total output tokens vs.\ pass@1, one point per model. Output tokens are for each problem instance (summed over all LLM calls in an agent run), averaged across 500 instances and 5 runs per instance . Pass@1 is the mean of the five per-run resolution rates. Colors group models by family. Top-left corresponds to high accuracy with the fewest output tokens spent reasoning and acting.}
    \label{fig:metrics-sbv-pass-vs-tokens}
\end{figure}

Figure~\ref{fig:metrics-sbv-pass-vs-tokens} relates total output-token spend to pass@1 across every model we evaluate, giving a rough Pareto view of token cost vs.\ resolution rate. Top-left is the desirable region (high accuracy with the fewest output tokens spent reasoning and acting).

The next four figures show per-model behavioral metrics across all 21 models we evaluated on SWE-Bench-Verified, ordered by family so adjacent cells share lineage. Anthropic Claude (Opus 4.6, Opus 4.5, Sonnet 4.6, Sonnet 4.5, Sonnet 4.0, Haiku 4.5), OpenAI GPTs (GPT-5.4, GPT-5.3 Codex, GPT-5.2 Codex, GPT-5.2), OpenAI open-weights (GPT-OSS 120B, GPT-OSS 20B), Google Gemini (3.1 Pro, 3 Flash), xAI (Grok 4.20-reasoning), then Qwen3-Coder variants (Next, 480B, 30B) and Qwen3.5/3.6 think variants (35B-A3B, 27B, Qwen3.5-27B).

Table~\ref{tab:metrics-sbv-distance-backtrack-summary} gives a compact numeric summary of the trajectory-geometry metrics (solution distance and local backtracking) over the resolved trajectories of each model, alongside the outcome metric pass@1. Three trajectory scalars: \textbf{local backtrack \%} (share of distance-moving edits with $\Delta D > 0$ --- low value means a monotone descent), \textbf{Arrival-time} $\boldsymbol{t @ D \le 0.1}$ (the normalized cycle position at which the mean resolved $D(t)$ curve first reaches within $0.1$ of a solution --- high value means the model lands its solution-relevant edits late in the run), and \textbf{AUC} (area under the resolved $D(t)$ curve, lower means the live diff was, on average, closer to a reference solution throughout the run). The full outcome table including pass@2 with $95\%$ confidence intervals is in Appendix~\ref{app:passrates}.

\begin{table}[h]
\centering
\small
\caption{Distance, local backtracking and outcome summary over resolved SWE-Bench-Verified trajectories per model. Local backtrack \%, Arrival-time $t$ @ $D\!\le\!0.1$ and area-under-curve (AUC) are computed from the faithful state-replay; pass@1 is from Appendix~\ref{app:passrates}.}
\label{tab:metrics-sbv-distance-backtrack-summary}
\begin{tabular}{@{}lrrrr@{}}
\toprule
Model & Local backtrack \% & $t$ @ $D\!\le\!0.1$ & AUC & pass@1 \% \\
\midrule
Opus 5.0           &  9.0 & 0.37 & 0.25 & 96.60 \\
Opus 4.8           &  9.3 & 0.47 & 0.31 & 88.80 \\
Opus 4.6           &  9.4 & 0.79 & 0.58 & 80.80 \\
Opus 4.5           &  7.5 & 0.58 & 0.41 & 80.48 \\
Sonnet 5.0         &  9.9 & 0.47 & 0.27 & 84.04 \\
Sonnet 4.6         &  5.4 & 0.74 & 0.50 & 79.00 \\
Sonnet 4.5         &  6.1 & 0.47 & 0.28 & 76.40 \\
Sonnet 4.0         &  6.2 & 0.68 & 0.41 & 72.00 \\
Haiku 4.5          &  4.2 & 0.58 & 0.35 & 72.60 \\
\midrule
GPT-5.4            &  1.3 & 0.84 & 0.63 & 80.16 \\
GPT-5.3            &  1.6 & 0.79 & 0.56 & 79.52 \\
GPT-5.2            &  1.2 & 0.79 & 0.54 & 79.28 \\
GPT-5.2-codex      &  0.9 & 0.89 & 0.68 & 76.80 \\
GPT-OSS 120B       &  2.8 & 0.84 & 0.42 & 64.64 \\
GPT-OSS 20B        &  1.2 & 0.89 & 0.54 & 59.84 \\
\midrule
Gemini 3.1 Pro     & 14.2 & 0.89 & 0.43 & 80.80 \\
Gemini 3 Flash     & 16.9 & 0.79 & 0.42 & 78.32 \\
\midrule
Grok 4.2           &  9.6 & 0.95 & 0.57 & 77.56 \\
\midrule
Qwen3-Coder Next   &  5.6 & 0.68 & 0.44 & 72.44 \\
Qwen3-Coder 480B   &  6.5 & 0.68 & 0.44 & 67.56 \\
Qwen3-Coder 30B    &  4.2 & 0.68 & 0.45 & 51.36 \\
Qwen3.6 35B-A3B    &  5.0 & 0.63 & 0.40 & 72.92 \\
Qwen3.6 27B        &  5.0 & 0.58 & 0.37 & 76.24 \\
Qwen3.5 27B        &  6.4 & 0.63 & 0.38 & 74.36 \\
\bottomrule
\end{tabular}
\end{table}

Three salient observations from Table~\ref{tab:metrics-sbv-distance-backtrack-summary} carry through the per-model grids below. The \textbf{Backtracking-edit} (the share of distance-moving edits with $\Delta D > 0$) spans an order of magnitude across the 24 models. The GPT-5 family and GPT-OSS sit under $3\%$ (GPT-5.2-codex lowest at $0.9\%$), the Claude models at $4$--$10\%$, Grok 4.2 at $9.6\%$, Gemini 3.1 Pro at $14.2\%$ and Gemini 3 Flash at $16.9\%$. The backtrack fraction (the share of trajectories that backtrack at least once) tracks this ranking closely (Gemini 3 Flash at $36\%$; GPT-5.2-codex at $2\%$). Crucially, backtracking is roughly \emph{independent} of solve rate: Gemini 3.1 Pro backtracks more than any model except Gemini 3 Flash ($14.2\%$) yet ties Opus 4.6 ($9.4\%$) at $80.8\%$ pass@1 on SWE-Bench-Verified, while GPT-5.4 is among the least-backtracked ($1.3\%$) and sits within a point of both ($80.2\%$). This decoupling reveals need for measuring trajectory \emph{shape} alongside the outcome, or, \textit{two models with the same pass@1 can have radically different confidence profiles}. For \textbf{Arrival-time} ($t$ @ $D\!\le\!0.1$), Grok 4.2 ($0.95$) and GPT-5.2-codex / GPT-OSS 20B ($0.89$) push their solution-relevant edits the latest, while Opus 5.0 ($0.37$) and then Opus 4.8, Sonnet 5.0 and Sonnet 4.5 ($0.47$) converge the earliest. \textbf{AUC} (average distance) is lowest for Opus 5.0 ($0.25$), Sonnet 5.0 ($0.27$) and Sonnet 4.5 ($0.28$) and highest for GPT-5.2-codex ($0.68$), GPT-5.4 ($0.63$) and Opus 4.6 ($0.58$). This is consistent with the arrival-time signal, a model that converges late spends most of the cycle far from any solution, regardless of whether its final state is correct.

\paragraph{Independent corroboration via the judge.}
The metrics above are computed entirely from the reconstructed file state. We cross-check it against an orthogonal signal derived from the tool-call judge, the deterministic \emph{revert label} \texttt{is\_revert}. This label tags each tool-call as \texttt{undo\_edit} / \texttt{git\_revert\_file} / \texttt{git\_revert\_all} / \texttt{none}. Paired \texttt{git stash} $\to$ \texttt{git stash pop} calls are excluded as transient. The judge-derived revert/edit rate (calls labeled revert divided by \texttt{edit\_source} $+$ \texttt{edit\_test}) tracks the faithful-replay backtracking ranking almost exactly. Gemini-3.1-Pro scores $14.0\%$, Opus 4.6 scores $8.7\%$, Sonnet 4.6 scores $6.1\%$ and GPT/Codex sit near $0$. Overall, two independent methods, (i) full state reconstruction via $\Delta D > 0$ and (ii) syntactic call labeling, agree both on which models backtrack and on how frequently.

\begin{figure}[p]
\centering
\setlength{\tabcolsep}{1pt}
\renewcommand{\arraystretch}{0.5}
\begin{tabular}{ccc}
\includegraphics[width=0.31\linewidth]{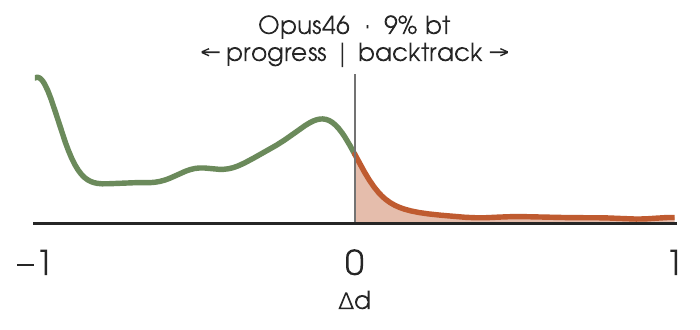} &
\includegraphics[width=0.31\linewidth]{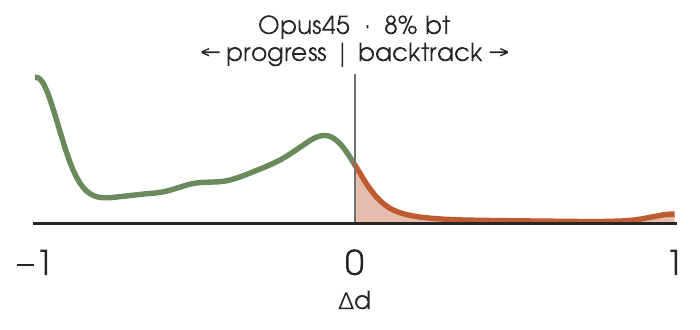} &
\includegraphics[width=0.31\linewidth]{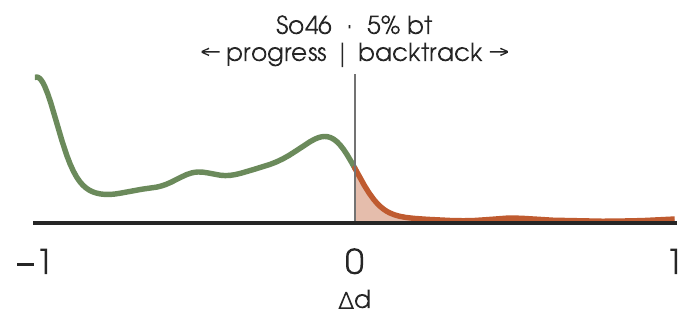} \\
\includegraphics[width=0.31\linewidth]{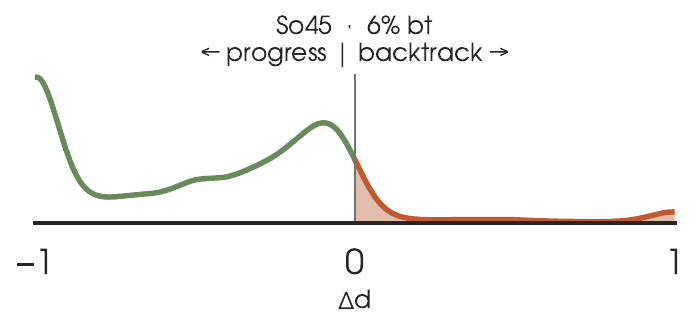} &
\includegraphics[width=0.31\linewidth]{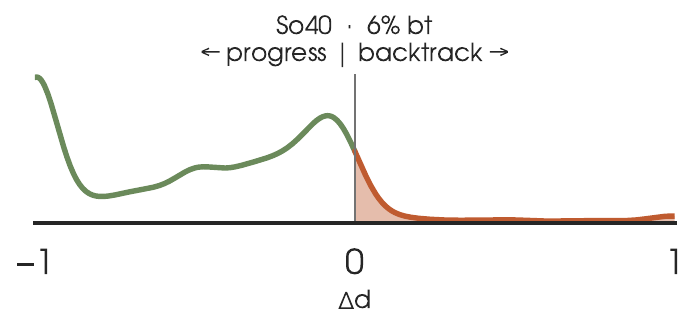} &
\includegraphics[width=0.31\linewidth]{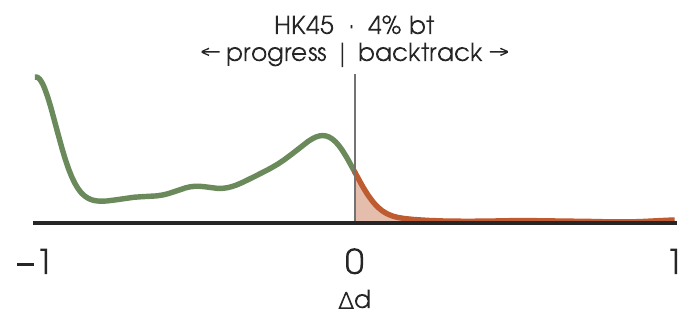} \\
\includegraphics[width=0.31\linewidth]{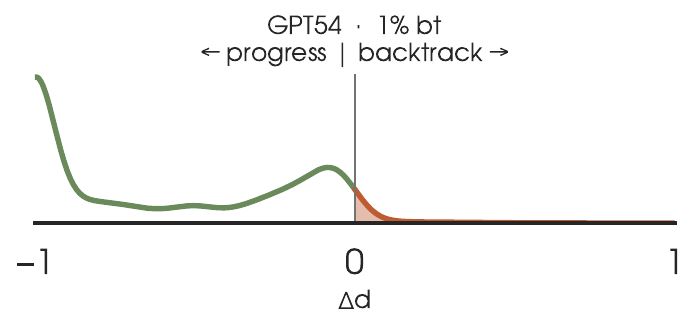} &
\includegraphics[width=0.31\linewidth]{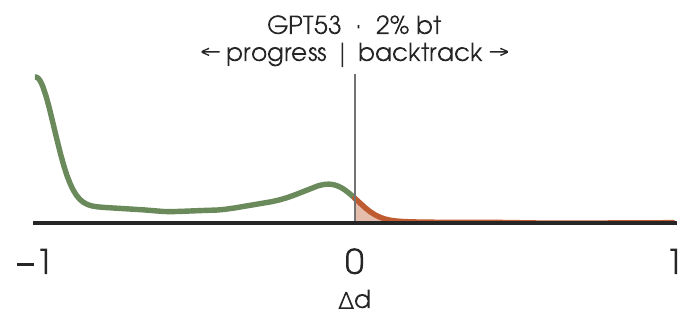} &
\includegraphics[width=0.31\linewidth]{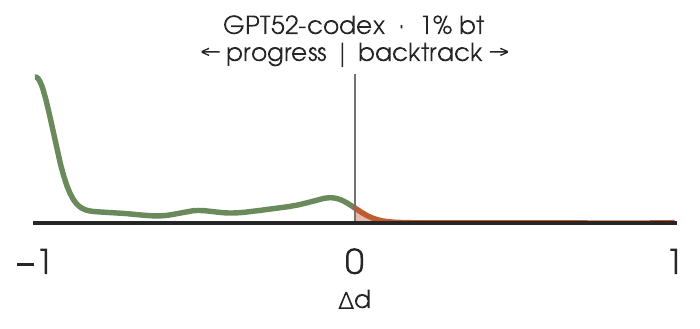} \\
\includegraphics[width=0.31\linewidth]{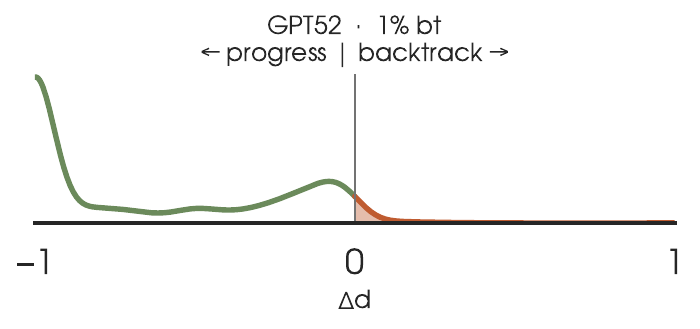} &
\includegraphics[width=0.31\linewidth]{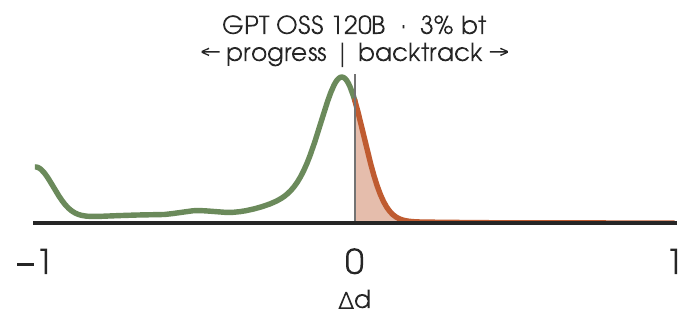} &
\includegraphics[width=0.31\linewidth]{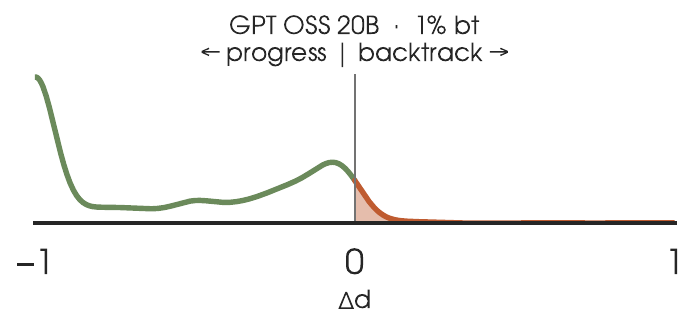} \\
\end{tabular}
\caption{\textbf{Local backtracking.} Histograms of $\Delta D$ on SWE-Bench-Verified (part 1 of 2): Anthropic Claude and OpenAI families. Each cell is one model, green = progress edits, red = local backtracking edits. Annotation shows the per-model local-backtrack share.}
\label{fig:metrics-sbv-backtracking}
\end{figure}

Figure~\ref{fig:metrics-sbv-backtracking} shows the per-edit $\Delta D$ histogram for each model: blue bars are progress edits ($\Delta D < 0$, agent moved towards a reference mode), red bars are backtracking edits ($\Delta D > 0$, agent moved away). The annotated percentage is the backtrack-edit share. Decisive solvers (GPT-5 family) have near-empty red mass, Qwen3-coder and Gemini-3.1-Pro carry the heaviest tails.

\begin{figure}[p]
\centering
\setlength{\tabcolsep}{1pt}
\renewcommand{\arraystretch}{0.5}
\begin{tabular}{ccc}
\includegraphics[width=0.31\linewidth]{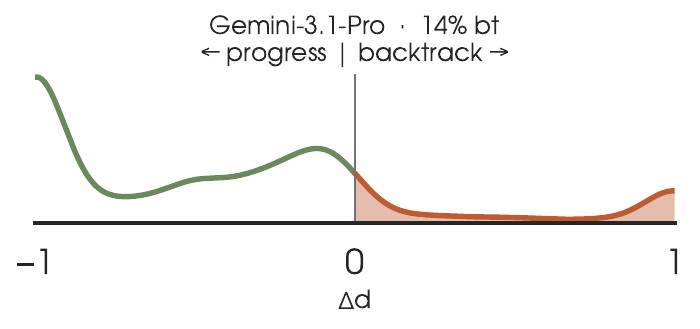} &
\includegraphics[width=0.31\linewidth]{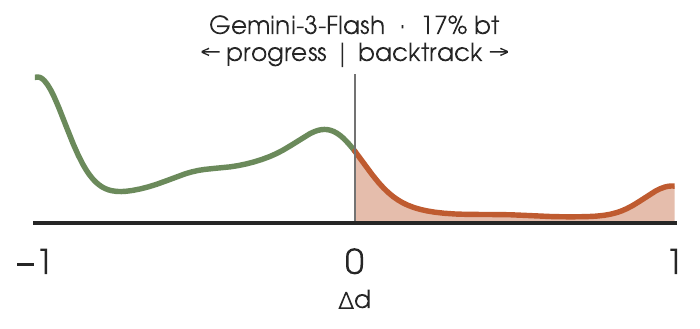} & \\
\includegraphics[width=0.31\linewidth]{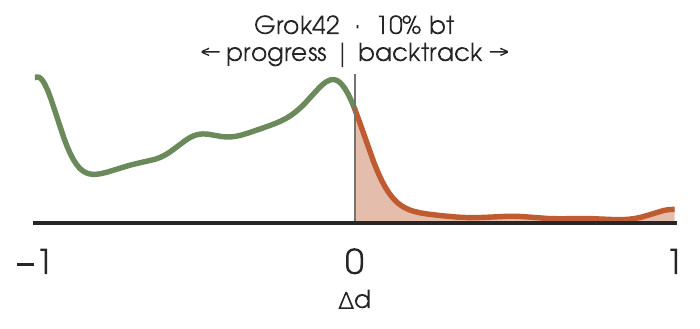} & & \\
\includegraphics[width=0.31\linewidth]{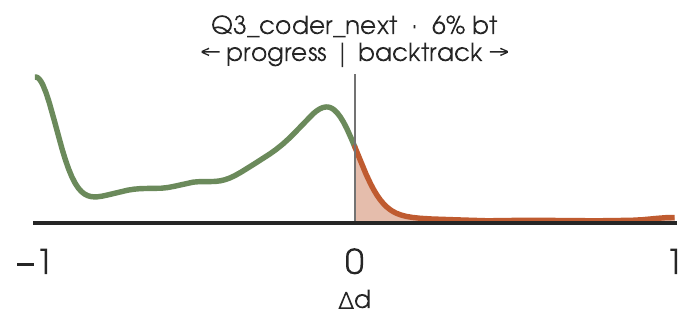} &
\includegraphics[width=0.31\linewidth]{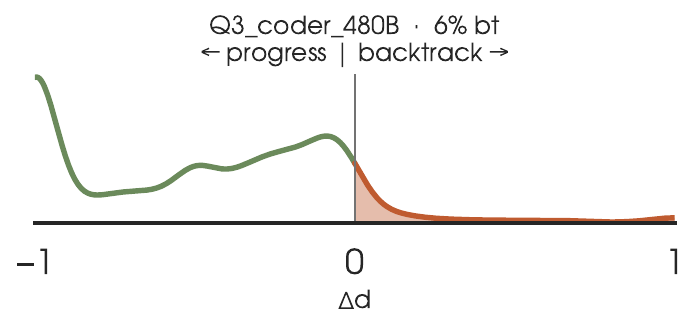} &
\includegraphics[width=0.31\linewidth]{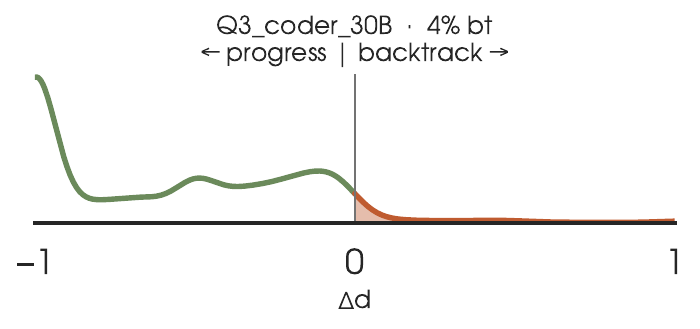} \\
\includegraphics[width=0.31\linewidth]{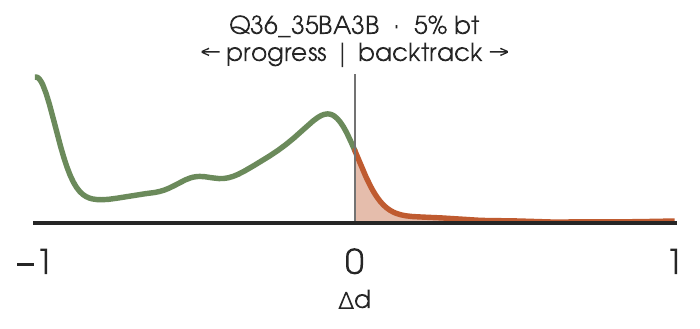} &
\includegraphics[width=0.31\linewidth]{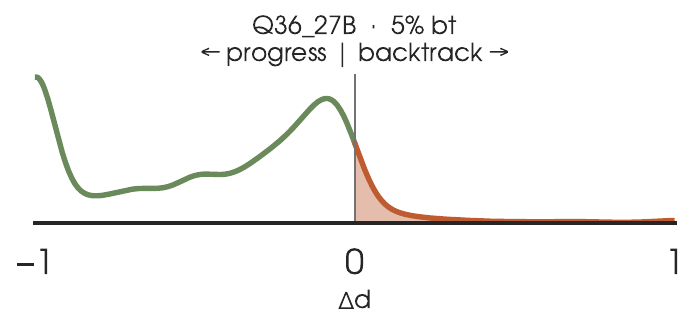} &
\includegraphics[width=0.31\linewidth]{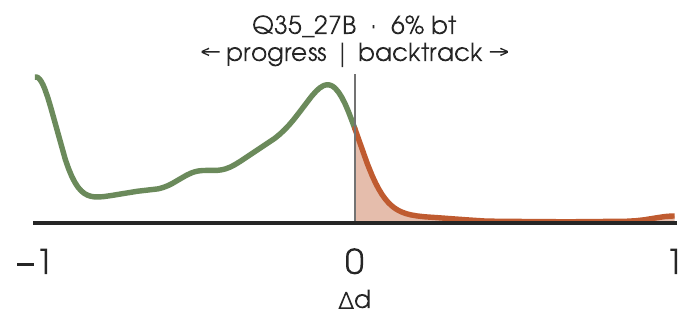} \\
\end{tabular}
\caption{\textbf{Local backtracking.} Histograms of $\Delta D$ on SWE-Bench-Verified (part 2 of 2): Gemini, Grok and Qwen families. Continued from Figure~\ref{fig:metrics-sbv-backtracking}.}
\label{fig:metrics-sbv-backtracking-b}
\end{figure}

\begin{figure}[p]
\centering
\includegraphics[width=0.75\linewidth]{swe_verif/edit_read_ratio_legend.pdf}\\[2pt]
\setlength{\tabcolsep}{1pt}
\renewcommand{\arraystretch}{0.5}
\begin{tabular}{ccc}
\includegraphics[width=0.31\linewidth]{swe_verif/edit_read_ratio/opus46.pdf} &
\includegraphics[width=0.31\linewidth]{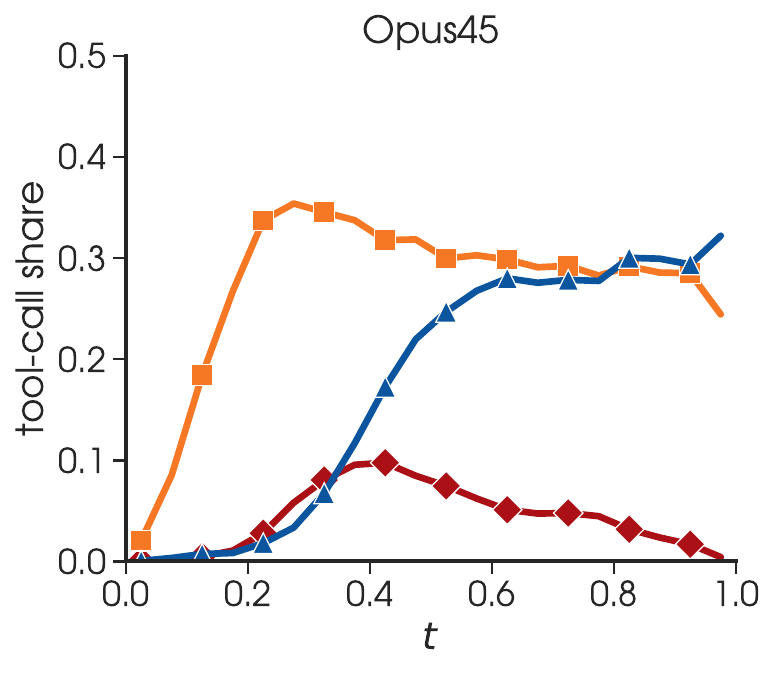} &
\includegraphics[width=0.31\linewidth]{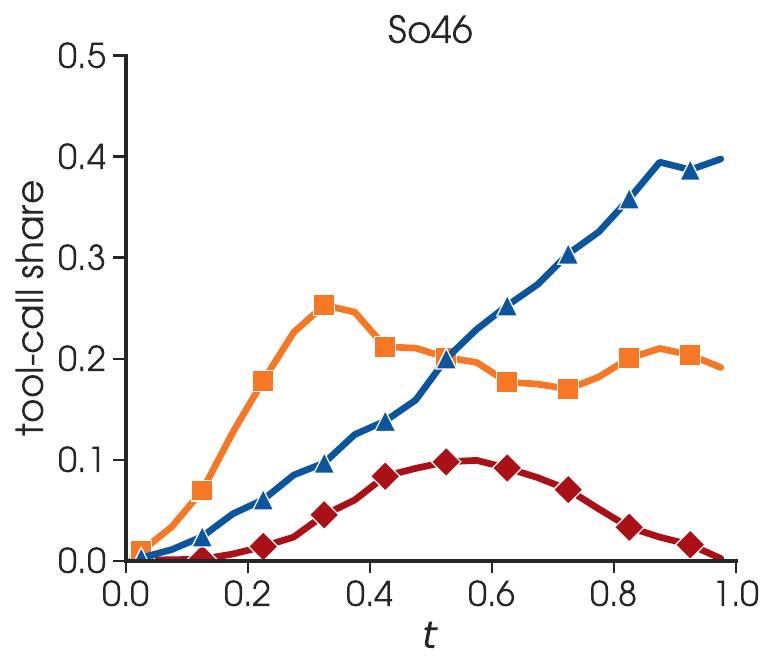} \\
\includegraphics[width=0.31\linewidth]{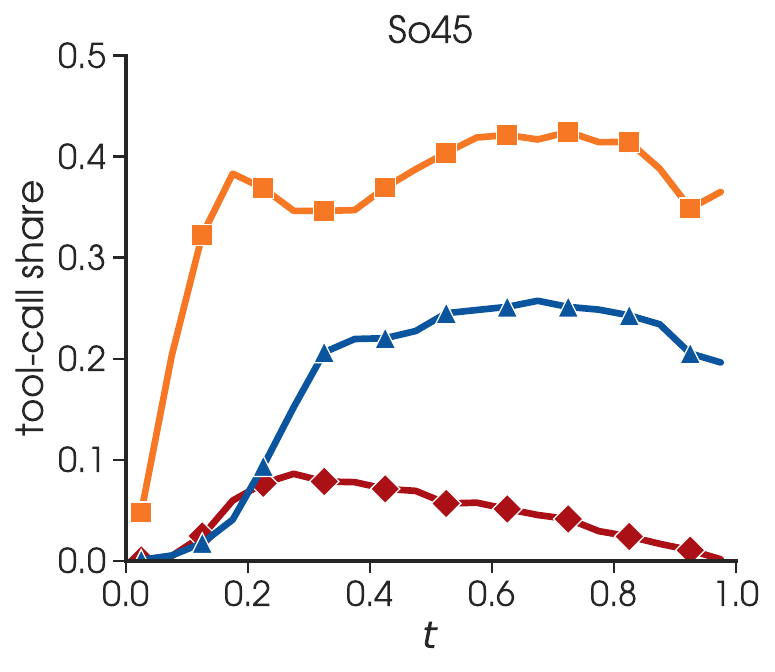} &
\includegraphics[width=0.31\linewidth]{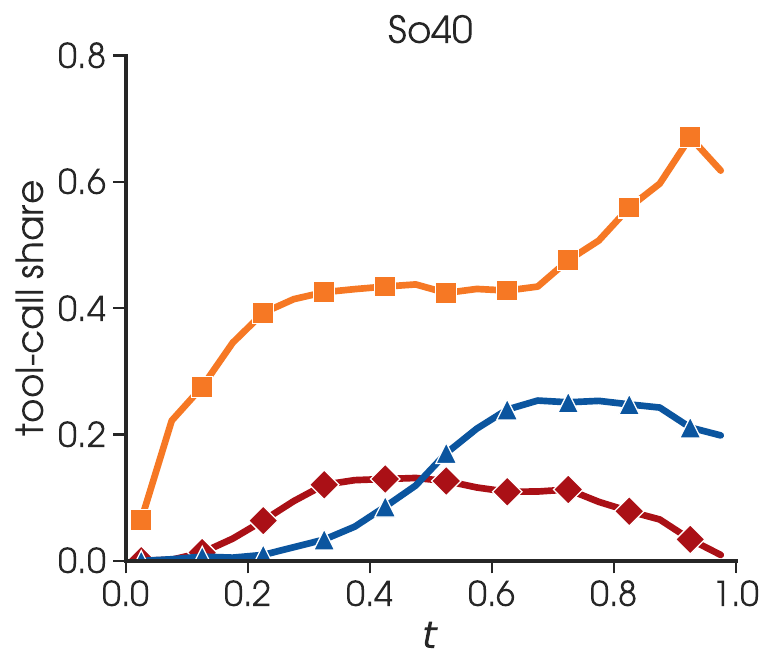} &
\includegraphics[width=0.31\linewidth]{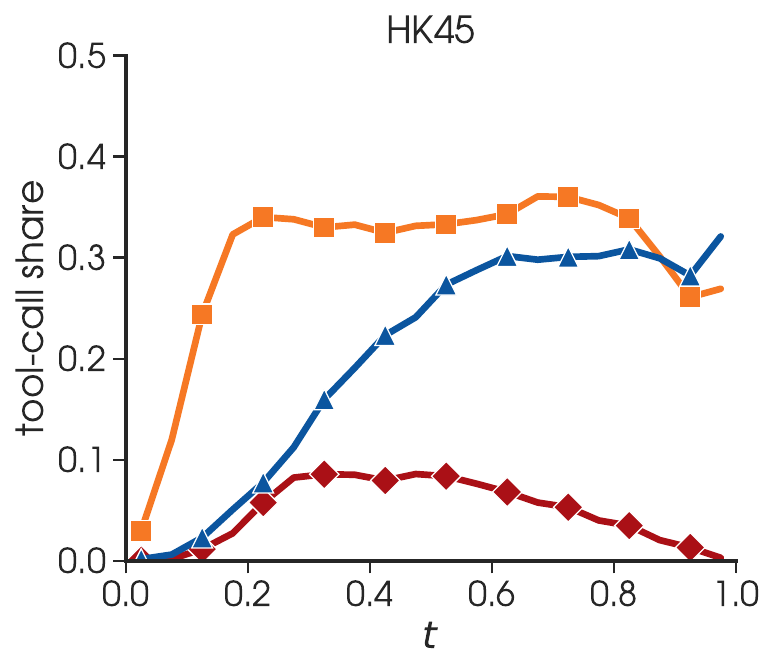} \\
\includegraphics[width=0.31\linewidth]{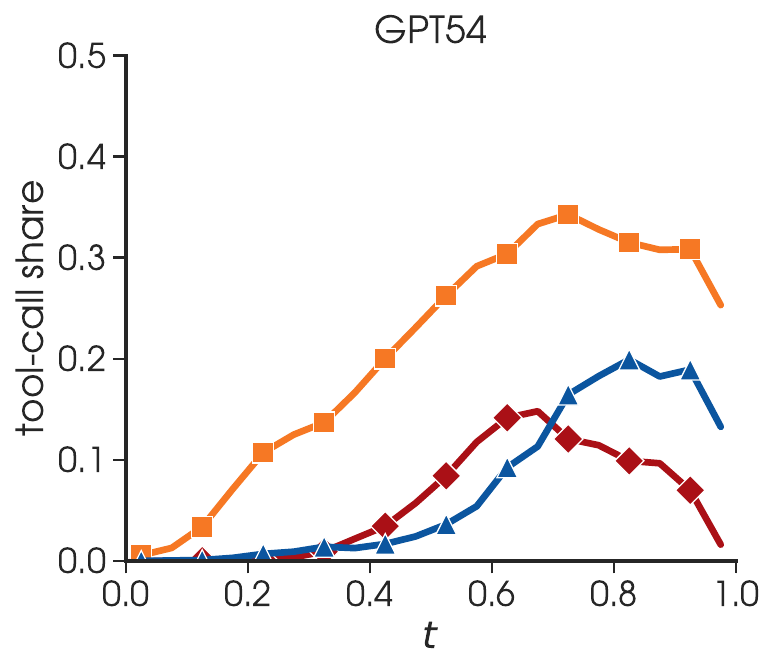} &
\includegraphics[width=0.31\linewidth]{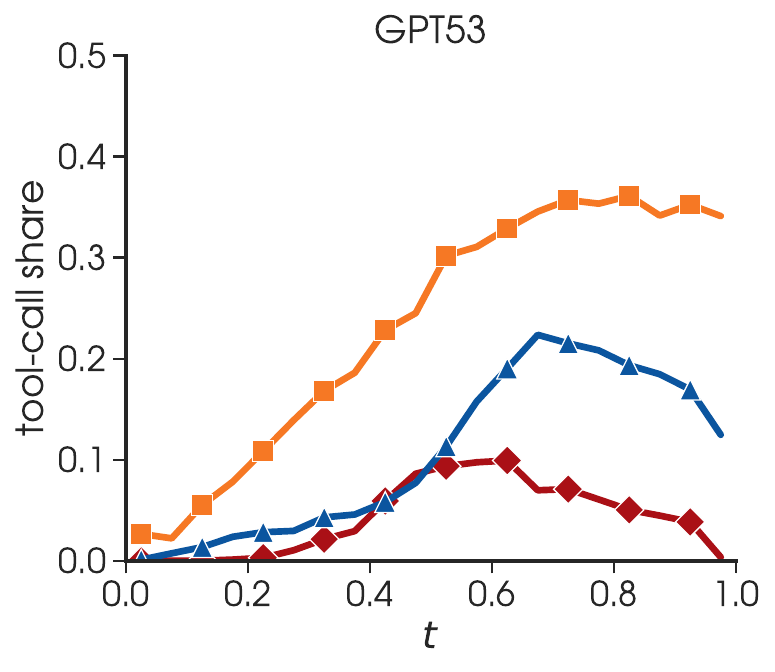} &
\includegraphics[width=0.31\linewidth]{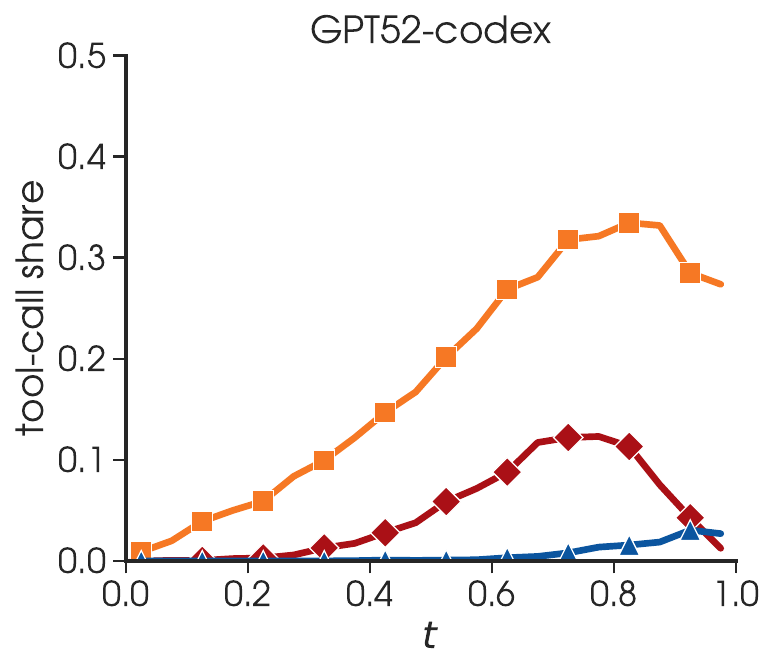} \\
\includegraphics[width=0.31\linewidth]{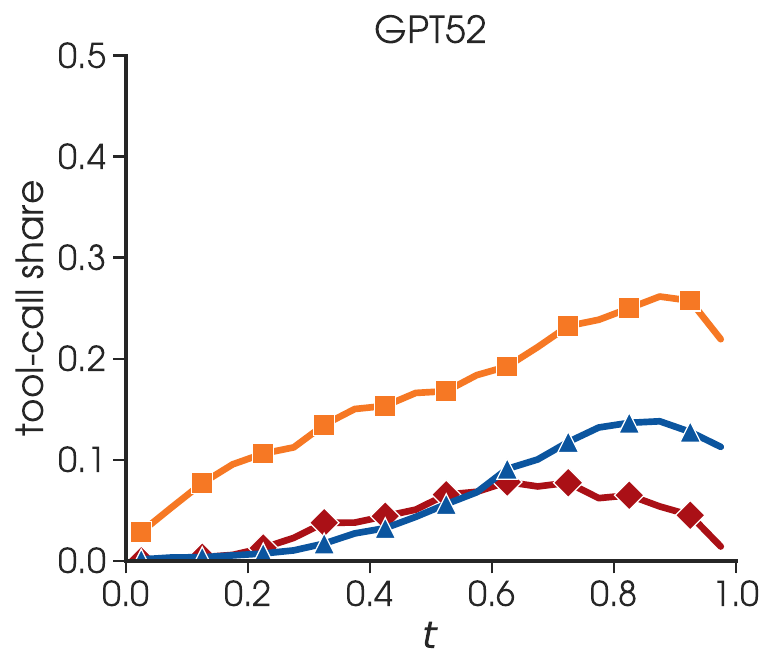} &
\includegraphics[width=0.31\linewidth]{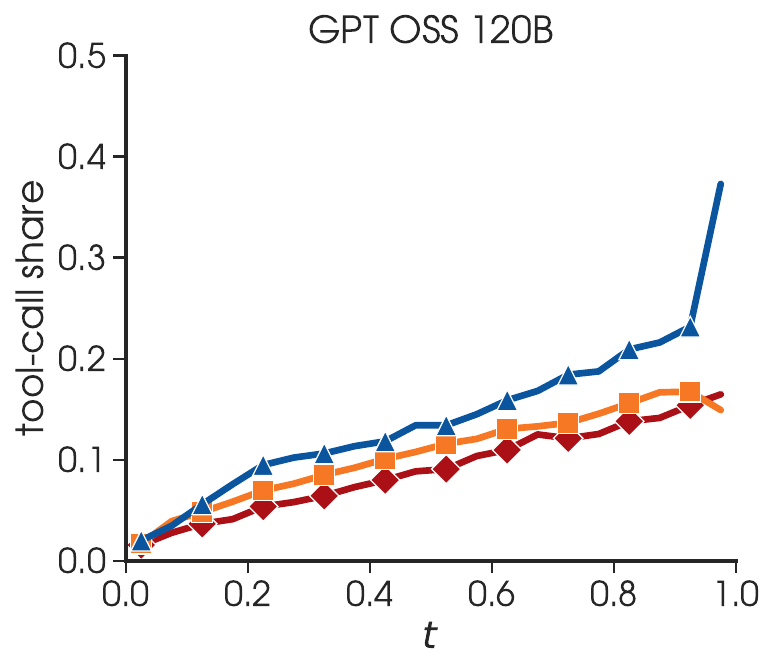} &
\includegraphics[width=0.31\linewidth]{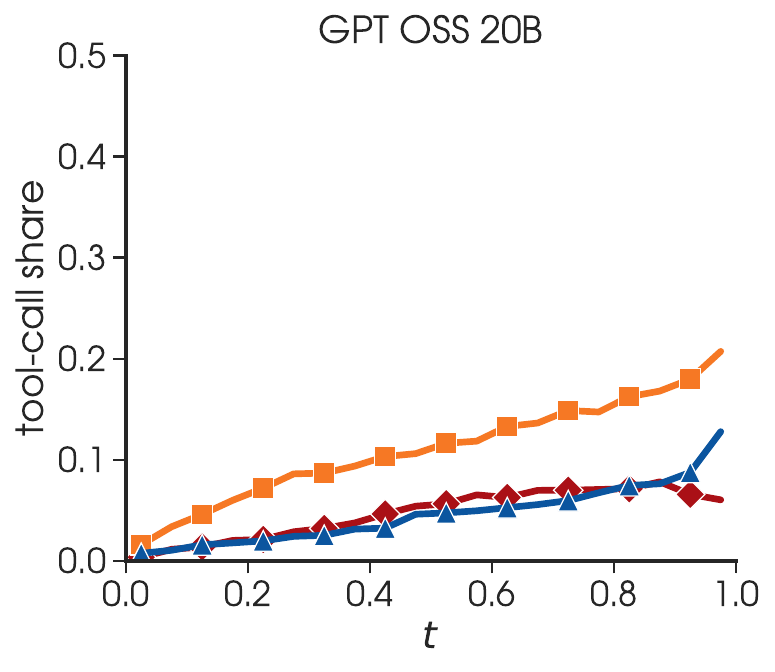} \\
\end{tabular}
\caption{Edit/test ratio per cycle (part 1 of 2): Anthropic Claude and OpenAI families. Each cell shows the model's distribution of ratios.}
\label{fig:metrics-sbv-editread}
\end{figure}

\begin{figure}[p]
\centering
\includegraphics[width=0.75\linewidth]{swe_verif/edit_read_ratio_legend.pdf}\\[2pt]
\setlength{\tabcolsep}{1pt}
\renewcommand{\arraystretch}{0.5}
\begin{tabular}{ccc}
\includegraphics[width=0.31\linewidth]{swe_verif/edit_read_ratio/gemini_3.1_pro.pdf} &
\includegraphics[width=0.31\linewidth]{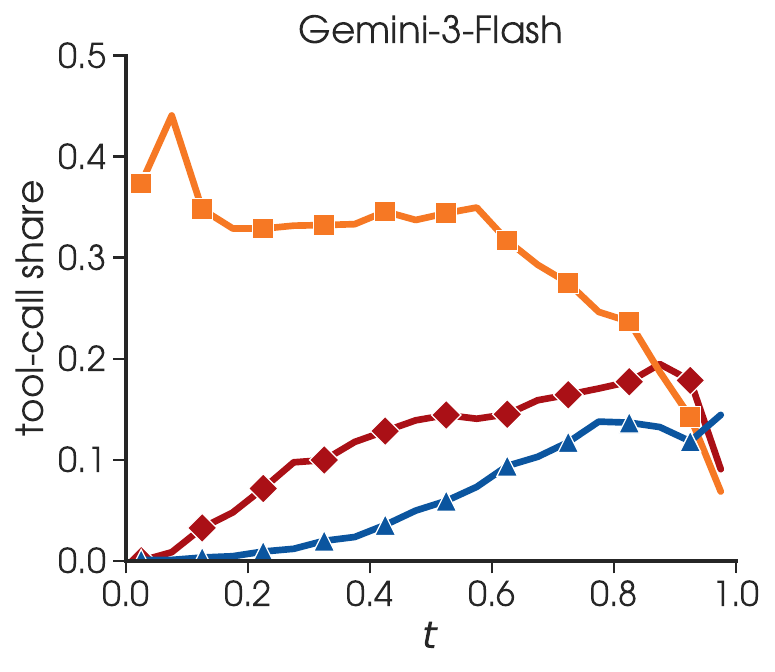} & \\
\includegraphics[width=0.31\linewidth]{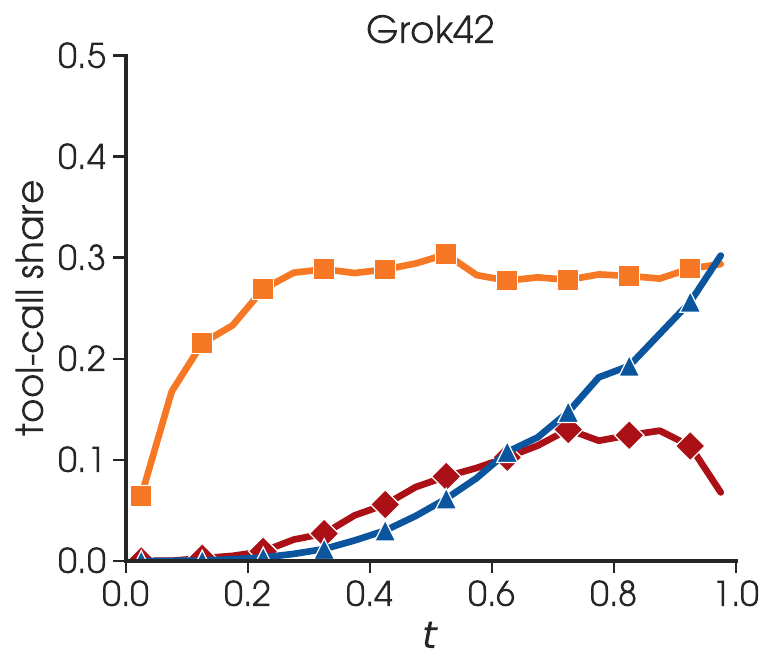} & & \\
\includegraphics[width=0.31\linewidth]{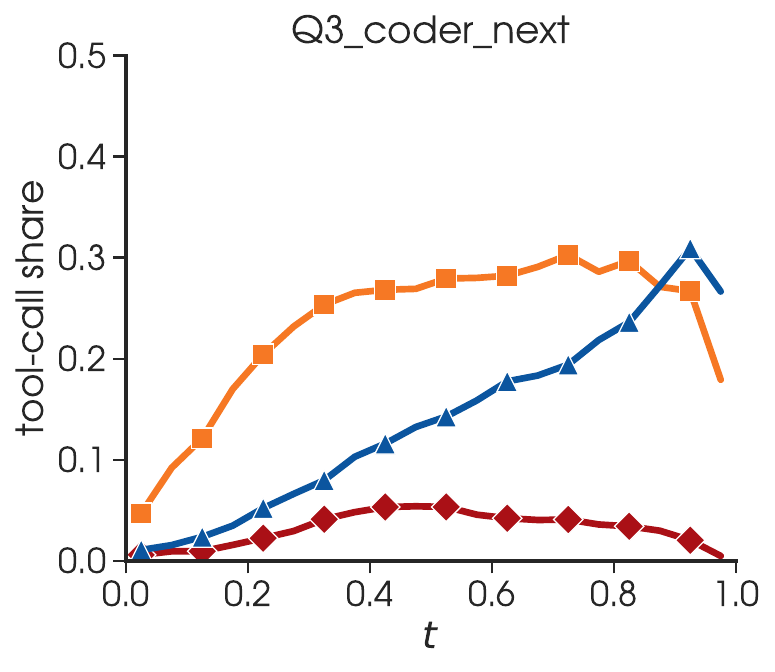} &
\includegraphics[width=0.31\linewidth]{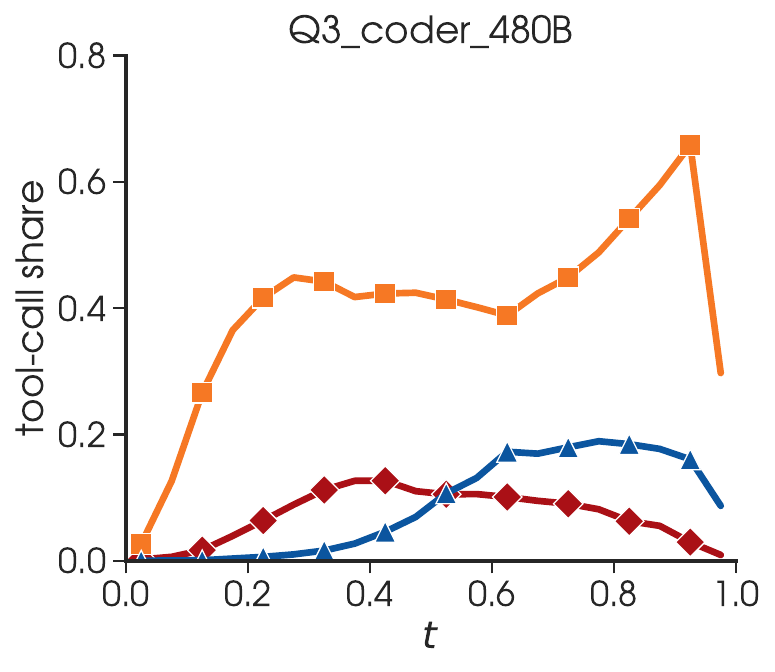} &
\includegraphics[width=0.31\linewidth]{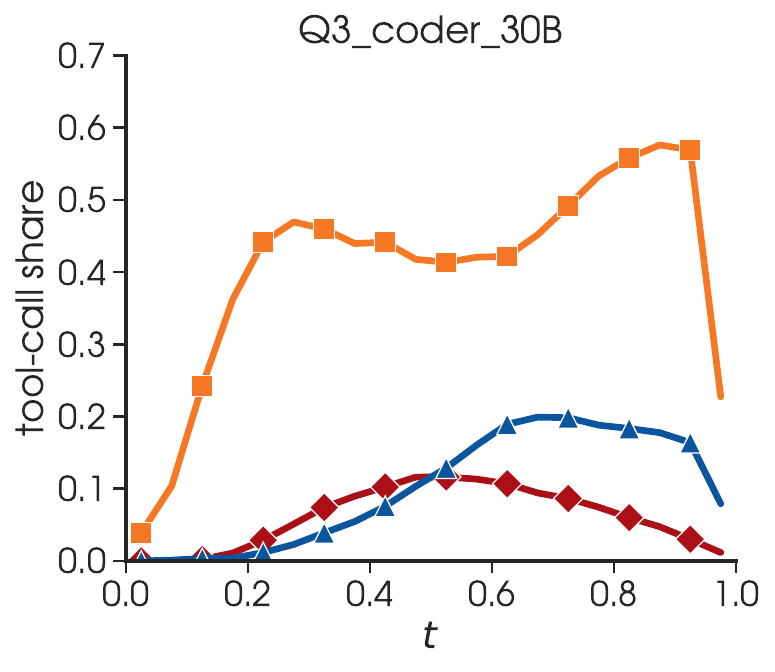} \\
\includegraphics[width=0.31\linewidth]{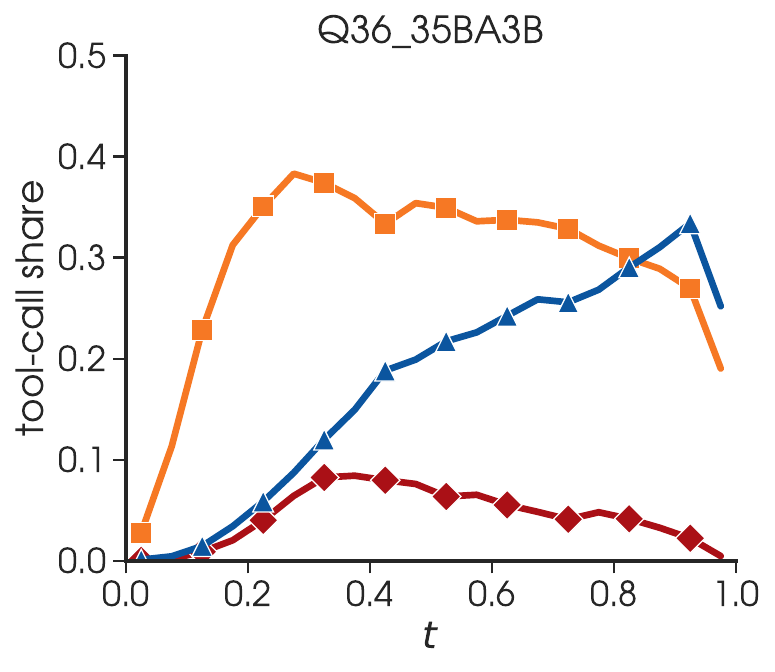} &
\includegraphics[width=0.31\linewidth]{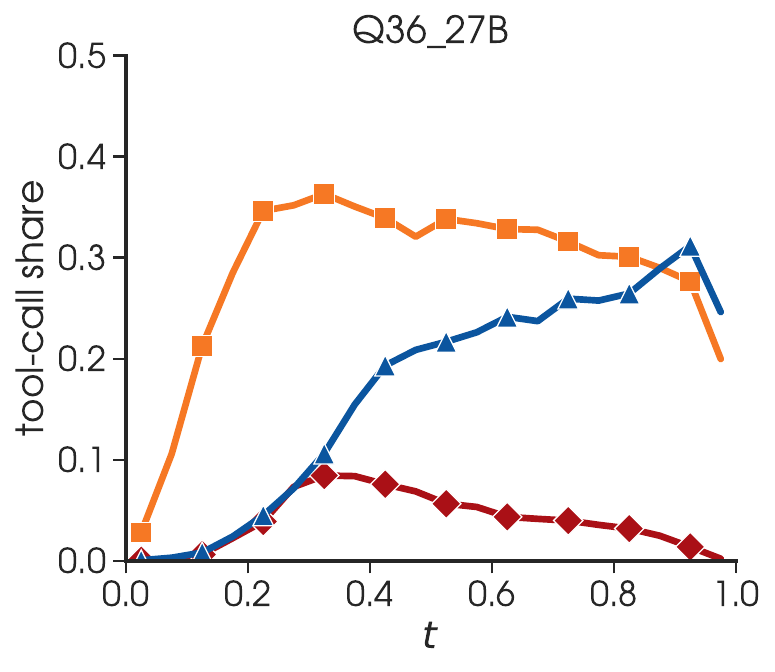} &
\includegraphics[width=0.31\linewidth]{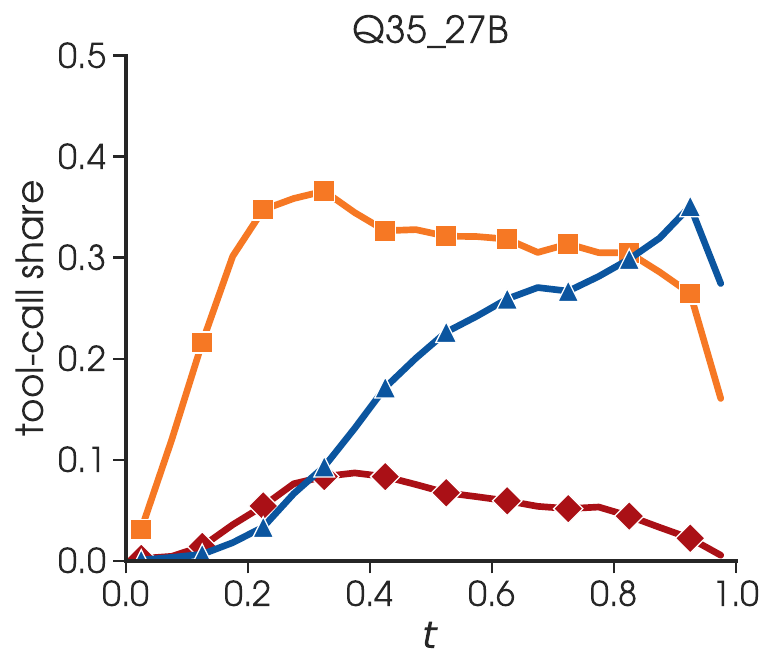} \\
\end{tabular}
\caption{Edit/test ratio per cycle (part 2 of 2). Gemini, Grok and Qwen families. Continued from Figure~\ref{fig:metrics-sbv-editread}.}
\label{fig:metrics-sbv-editread-b}
\end{figure}

\paragraph{Per-model edit/test ratio.}
Figure~\ref{fig:metrics-sbv-editread} shows the per-cycle ratio of edit-oriented tool calls to total ones across resolved + unresolved trajectories. A high ratio means the agent mostly edits and a low ratio means it mostly explores/test.

\begin{figure}[p]
\centering
\setlength{\tabcolsep}{1pt}
\renewcommand{\arraystretch}{0.5}
\begin{tabular}{ccc}
\includegraphics[width=0.31\linewidth]{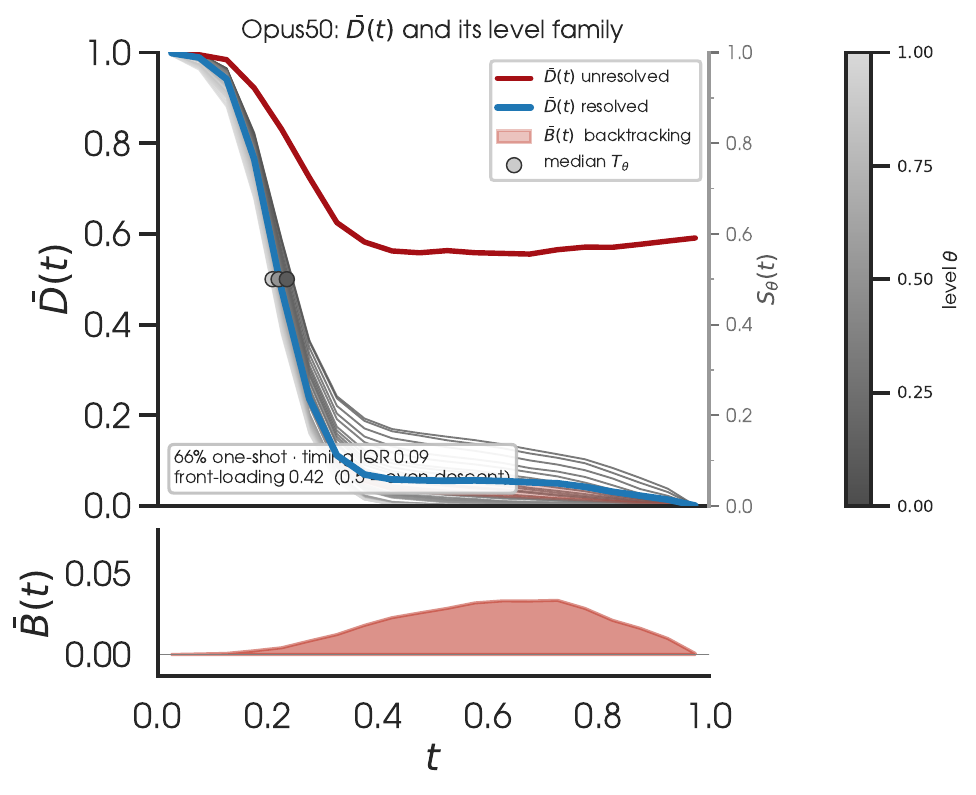} &
\includegraphics[width=0.31\linewidth]{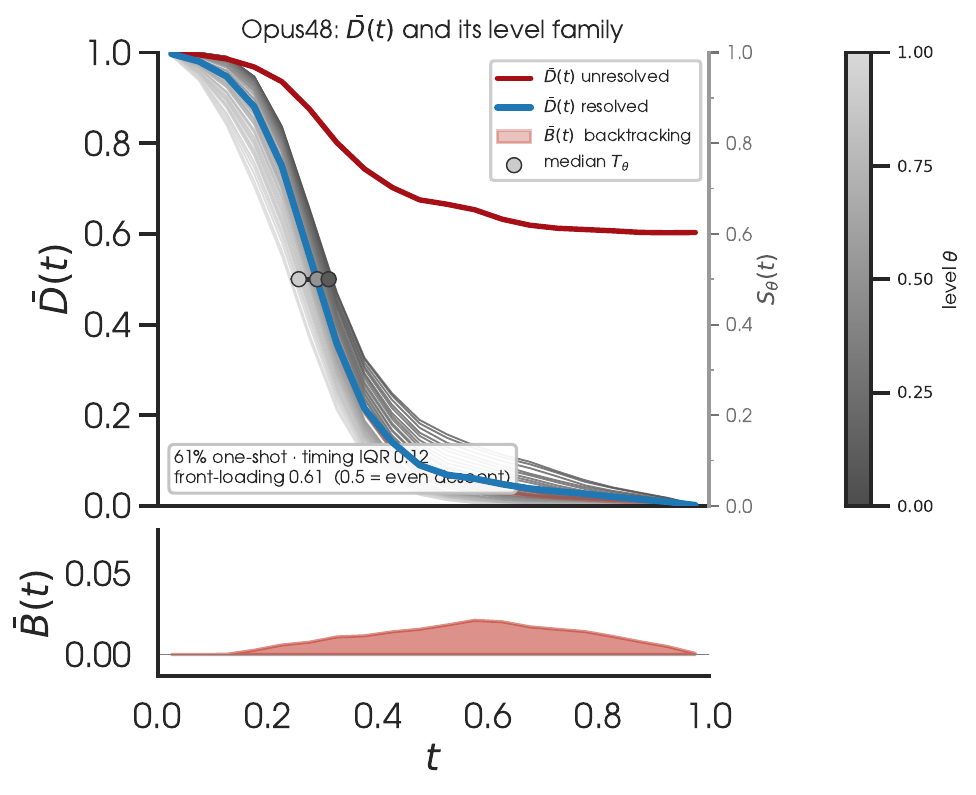} &
\includegraphics[width=0.31\linewidth]{swe_verif/solution_distance_fan/opus46.pdf} \\
\includegraphics[width=0.31\linewidth]{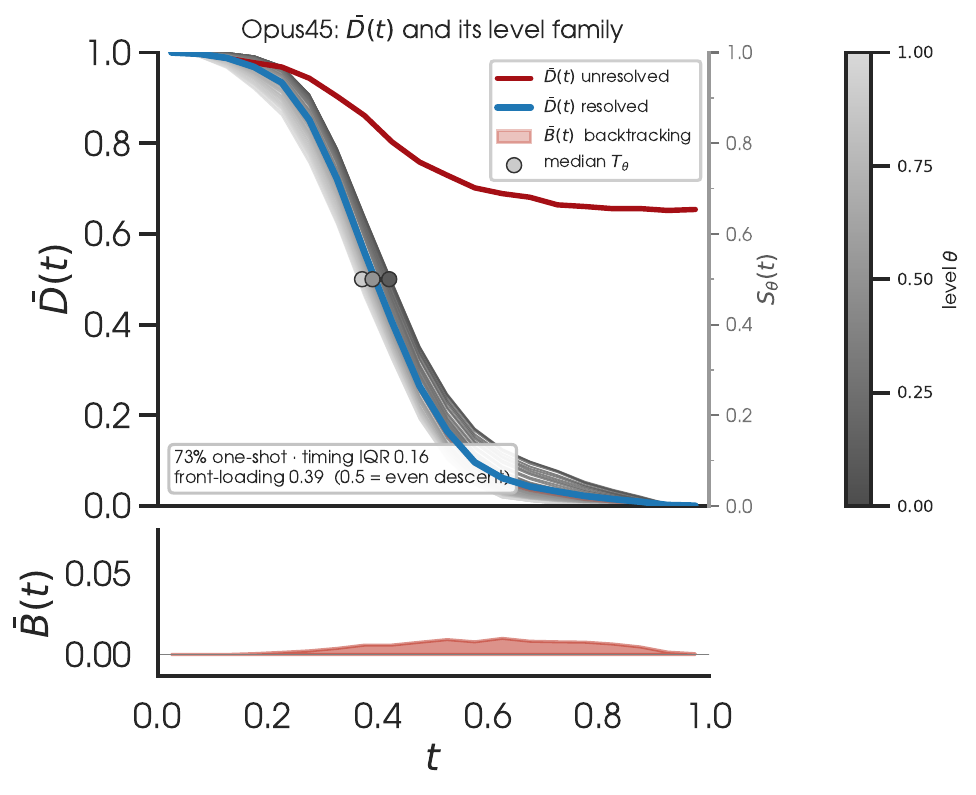} &
\includegraphics[width=0.31\linewidth]{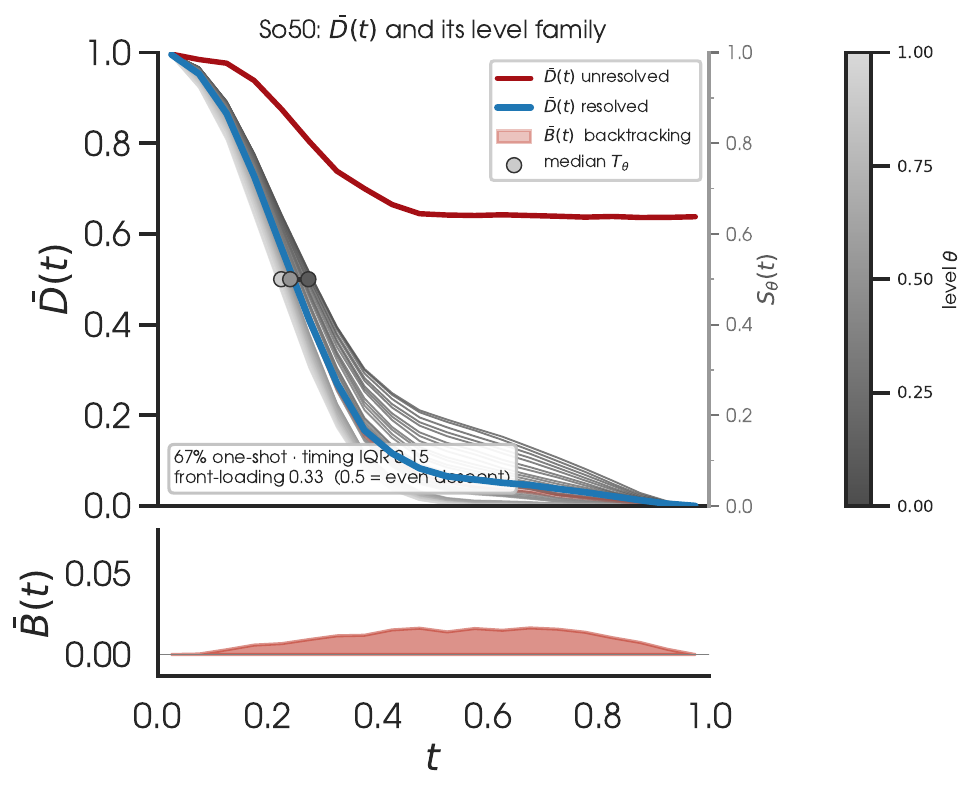} &
\includegraphics[width=0.31\linewidth]{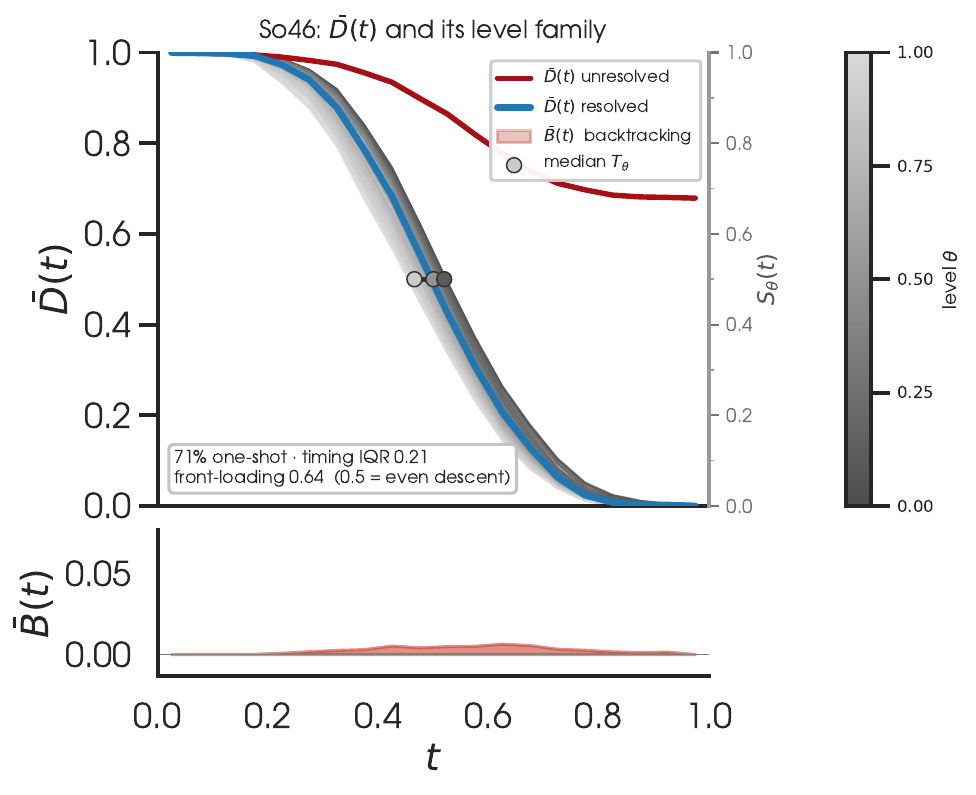} \\
\includegraphics[width=0.31\linewidth]{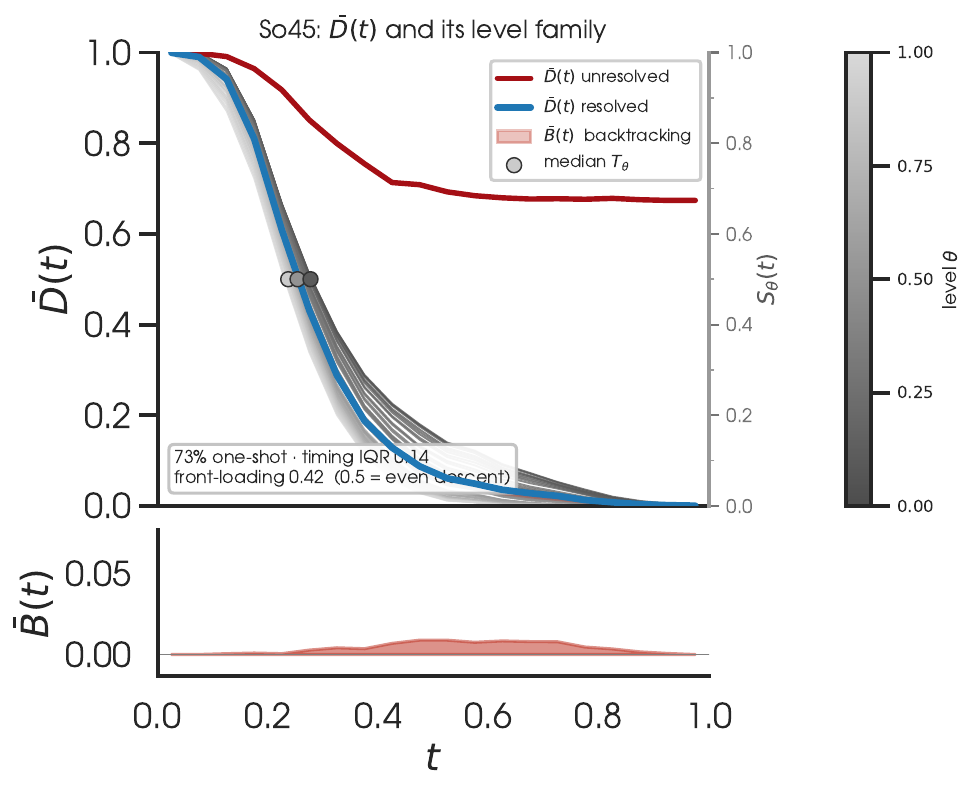} &
\includegraphics[width=0.31\linewidth]{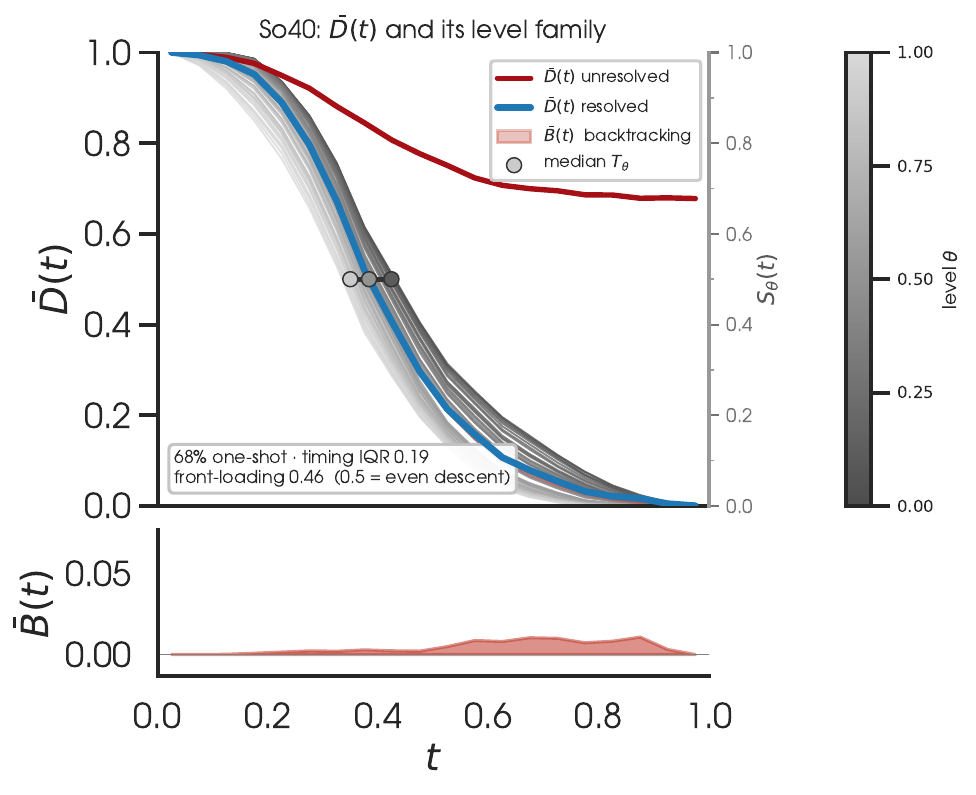} &
\includegraphics[width=0.31\linewidth]{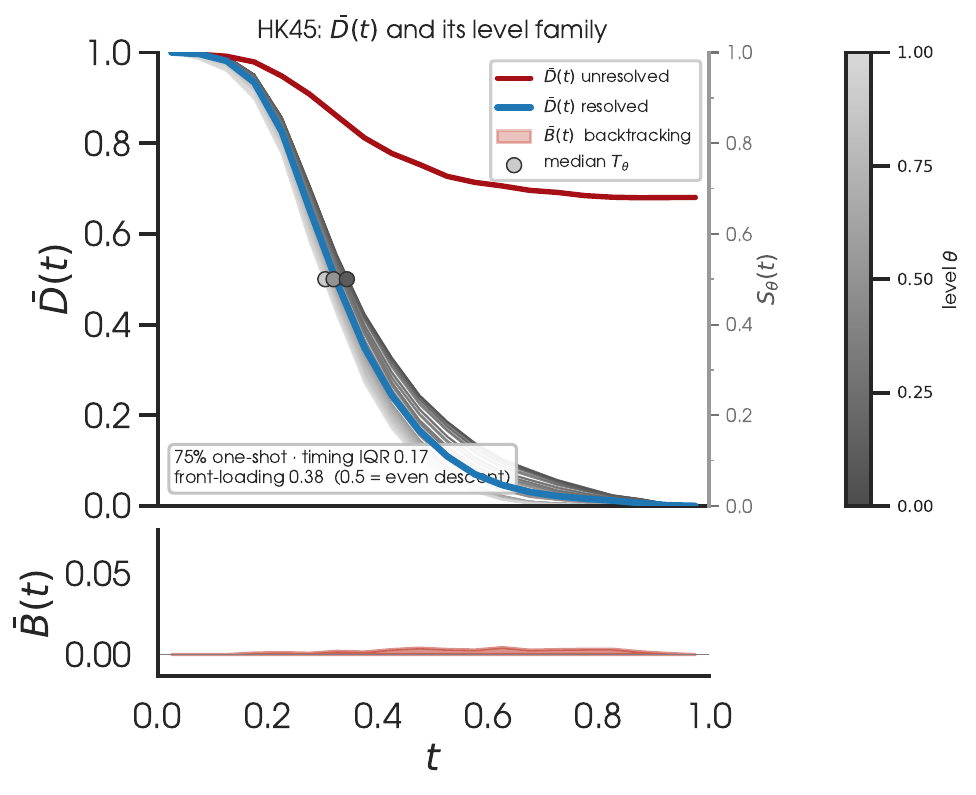} \\
\includegraphics[width=0.31\linewidth]{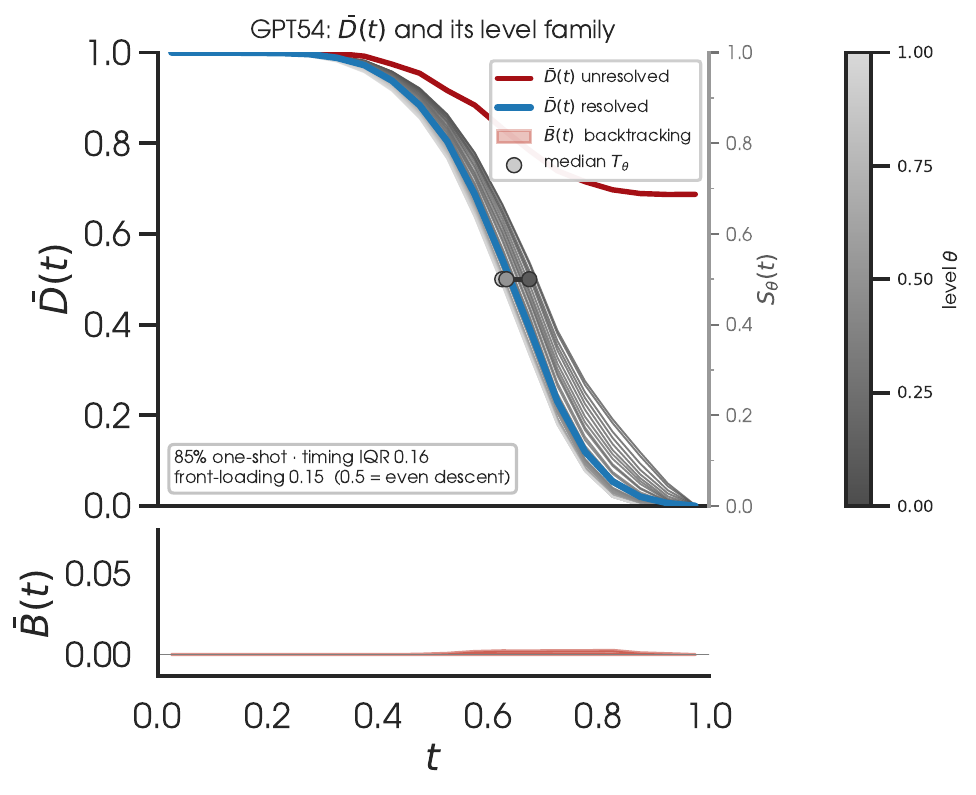} &
\includegraphics[width=0.31\linewidth]{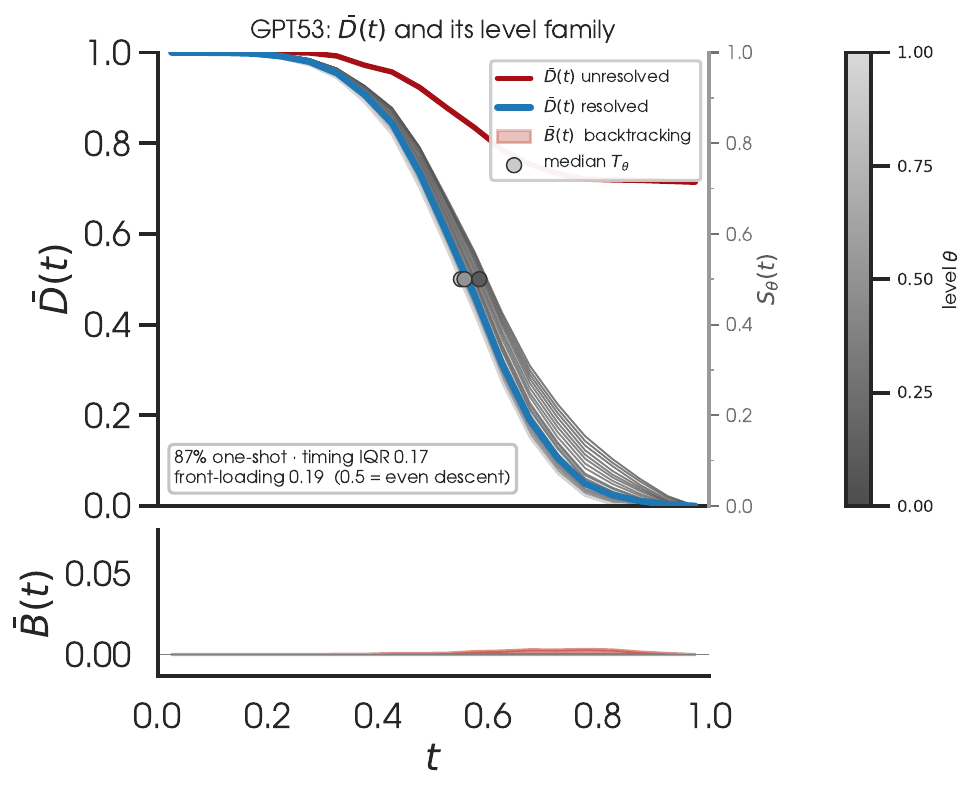} &
\includegraphics[width=0.31\linewidth]{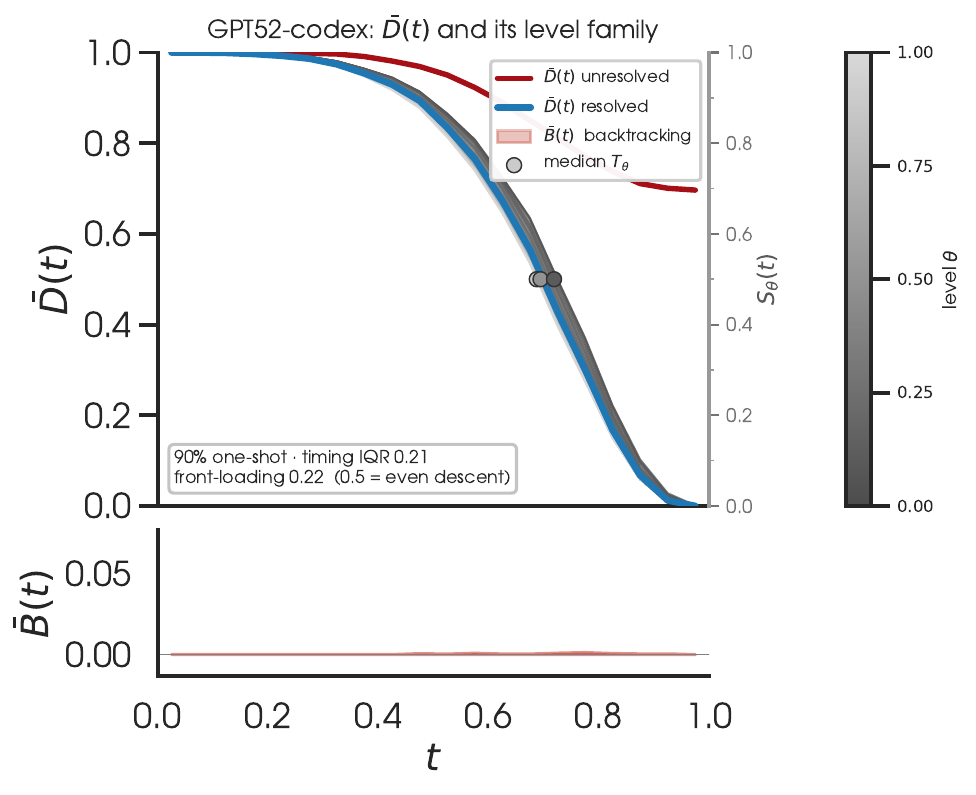} \\
\includegraphics[width=0.31\linewidth]{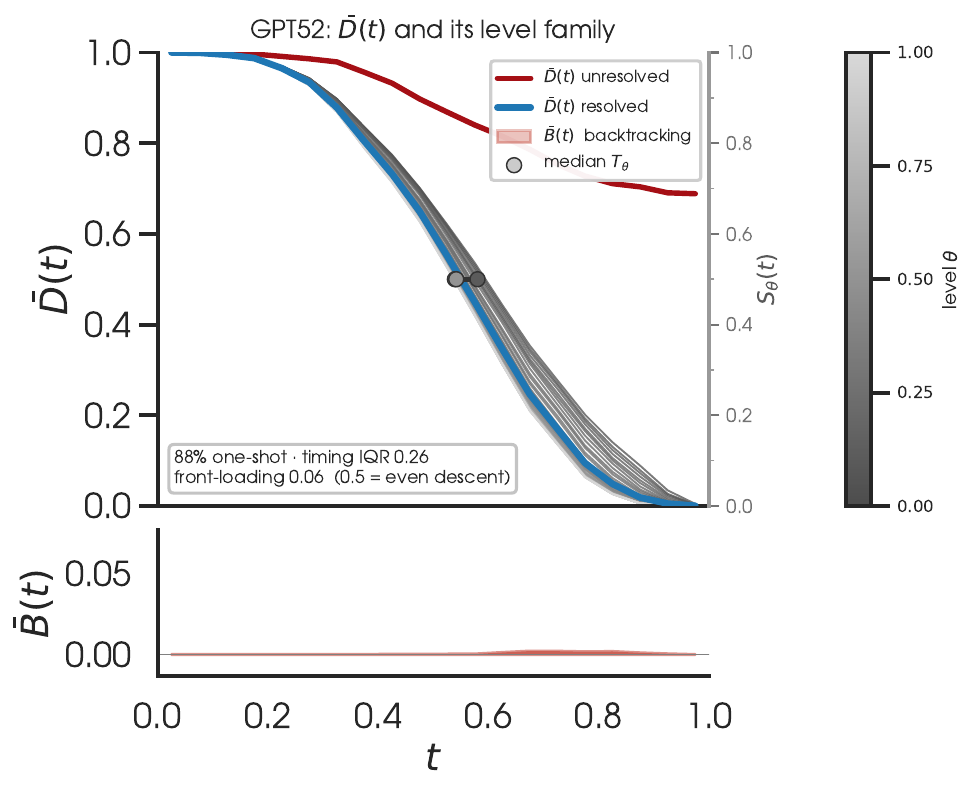} &
\includegraphics[width=0.31\linewidth]{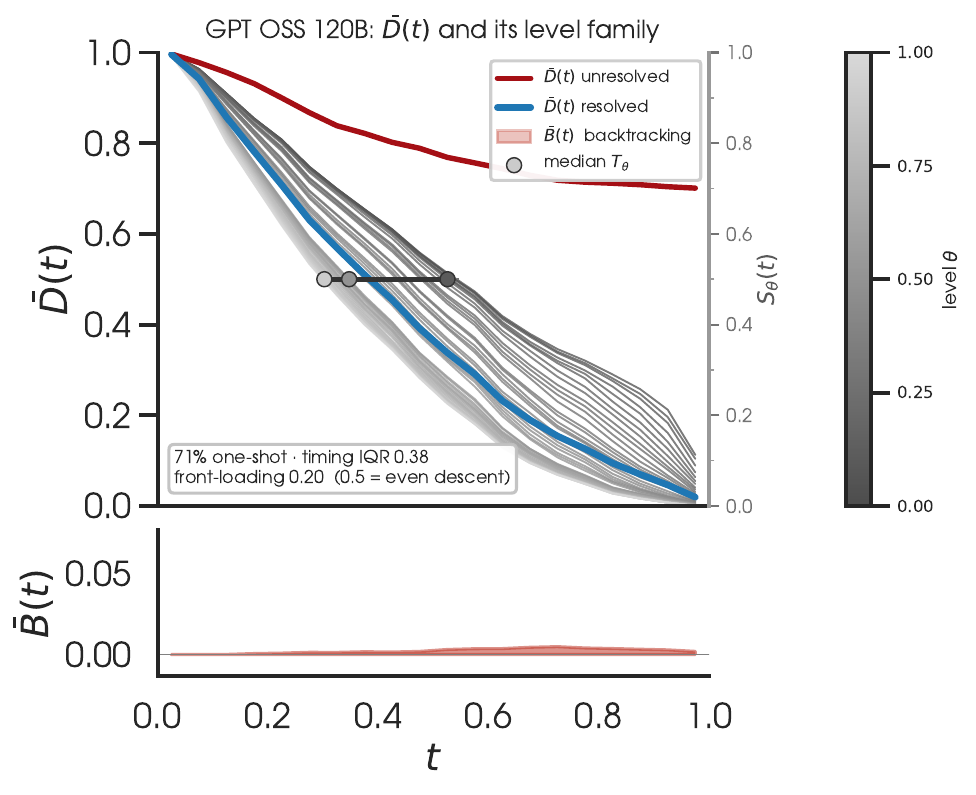} &
\includegraphics[width=0.31\linewidth]{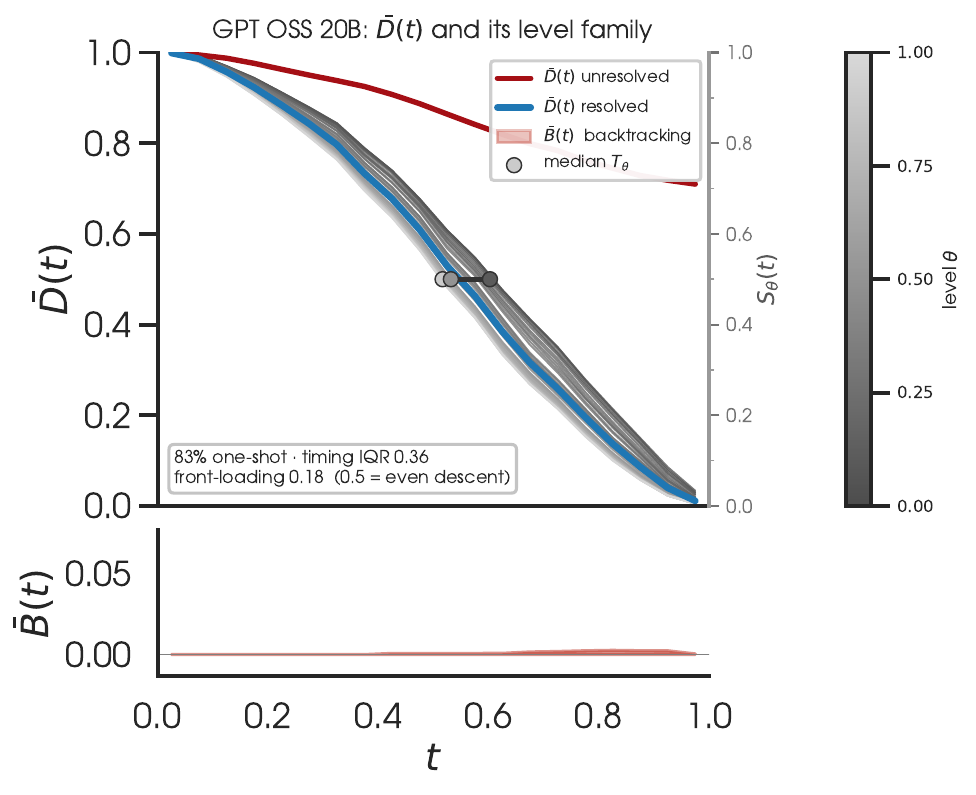} \\
\end{tabular}
\caption{Level families on SWE-Bench-Verified (part 1 of 2): Claude and GPT families. Grey curves are the level sets $S_\theta(t)$ of the resolved trajectories (Section~\ref{sec:distance}), the blue curve is their mean $\bar D(t)$, the dark-red curve is the mean $\bar D(t)$ of the unresolved runs (which plateaus where they stop), the markers give the median crossing time $T_\theta$ (the median over runs of the first-entry time $T_r(\theta)$, i.e.\ where $S_\theta$ crosses $1/2$) at $\theta=0.5$ with its interquartile range, and the strip beneath each panel is the backtracking term $\bar B(t)$. The inset reports the one-shot share (runs that land the patch in a single edit); the timing IQR, the interquartile range of the runs' half-way times $T_r(0.5)$, i.e.\ how dispersed solve timing is across runs; and the front-loading index $(T_{0.5}-T_{0.9})/(T_{0.1}-T_{0.9})$, the position of the median half-way time between the median crossing times of levels $0.9$ and $0.1$ ($0.5$ = even descent; lower = most of the distance falls early).}
\label{fig:metrics-sbv-soldist}
\end{figure}

\begin{figure}[p]
\centering
\setlength{\tabcolsep}{1pt}
\renewcommand{\arraystretch}{0.5}
\begin{tabular}{ccc}
\includegraphics[width=0.31\linewidth]{swe_verif/solution_distance_fan/gemini_3.1_pro.pdf} &
\includegraphics[width=0.31\linewidth]{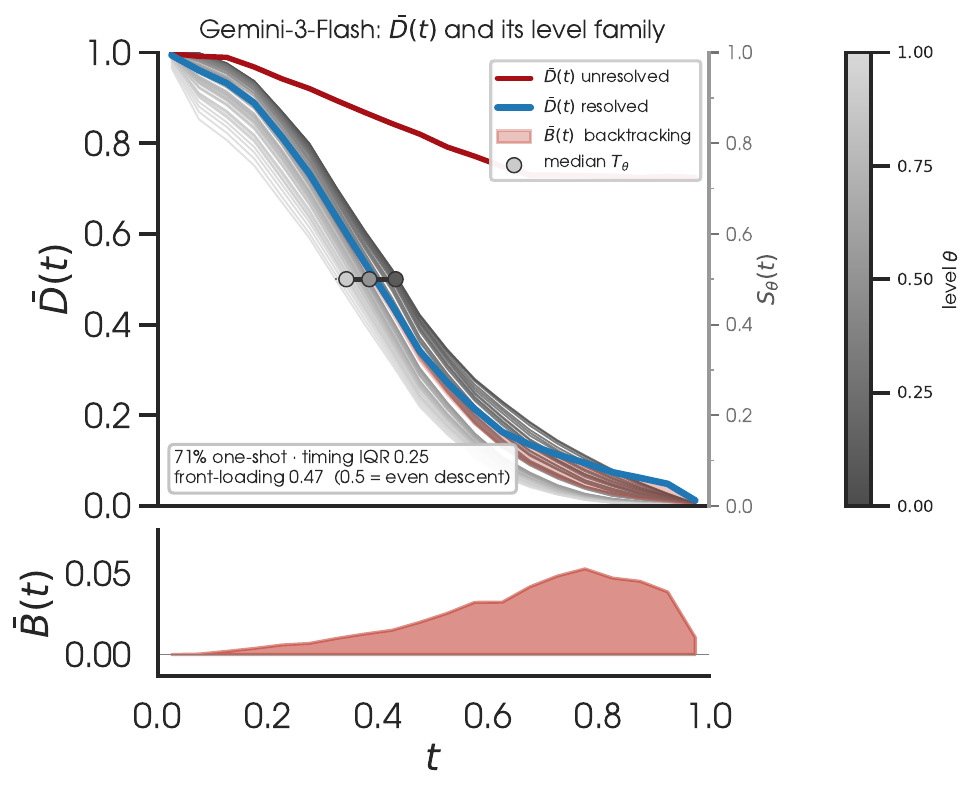} &
\includegraphics[width=0.31\linewidth]{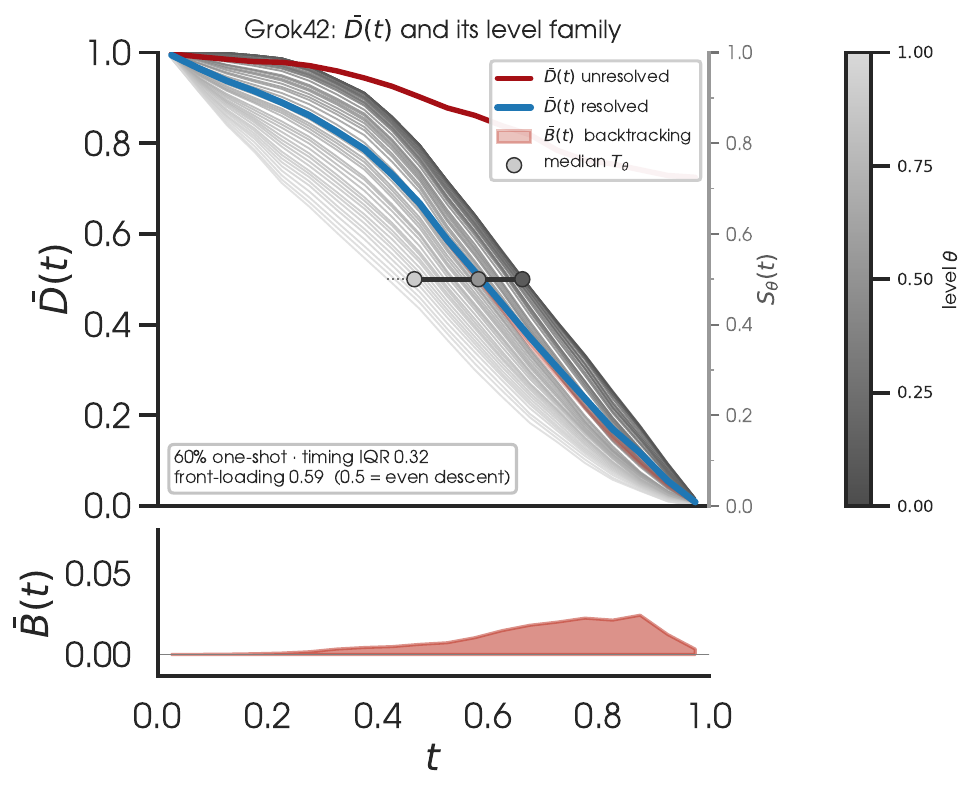} \\
\includegraphics[width=0.31\linewidth]{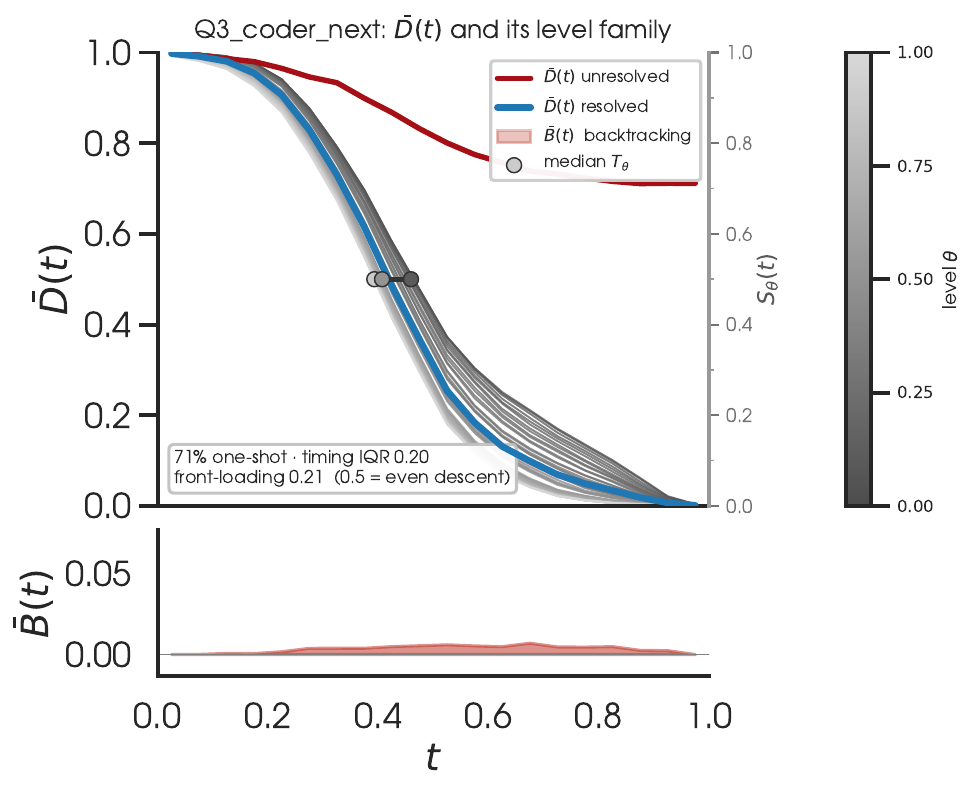} &
\includegraphics[width=0.31\linewidth]{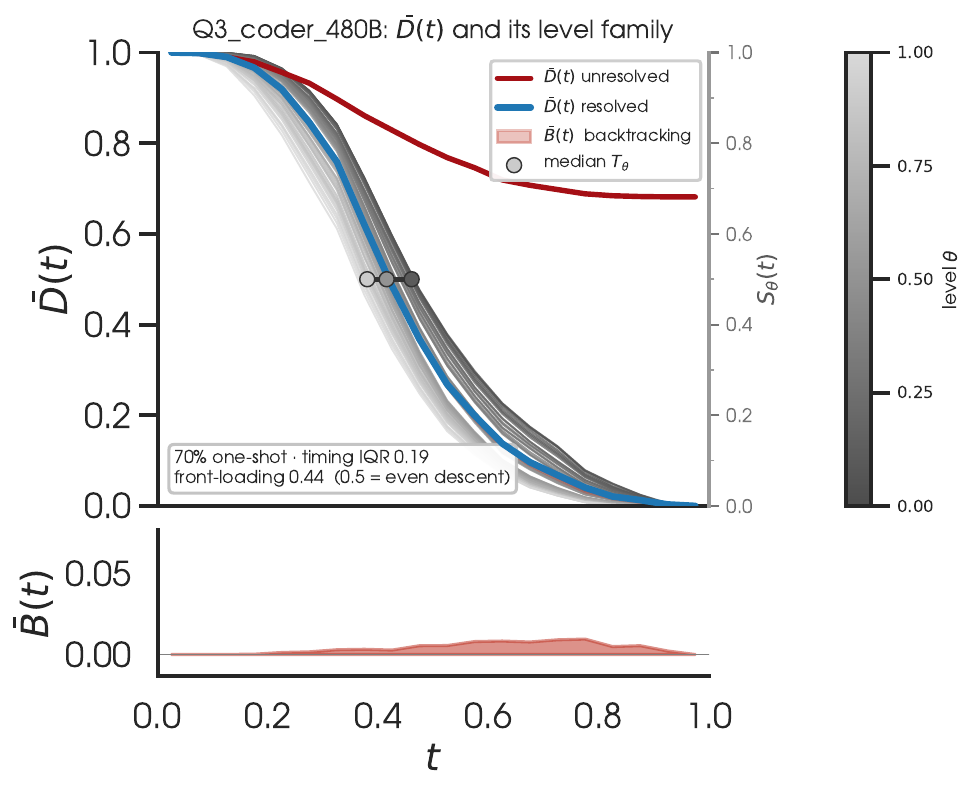} &
\includegraphics[width=0.31\linewidth]{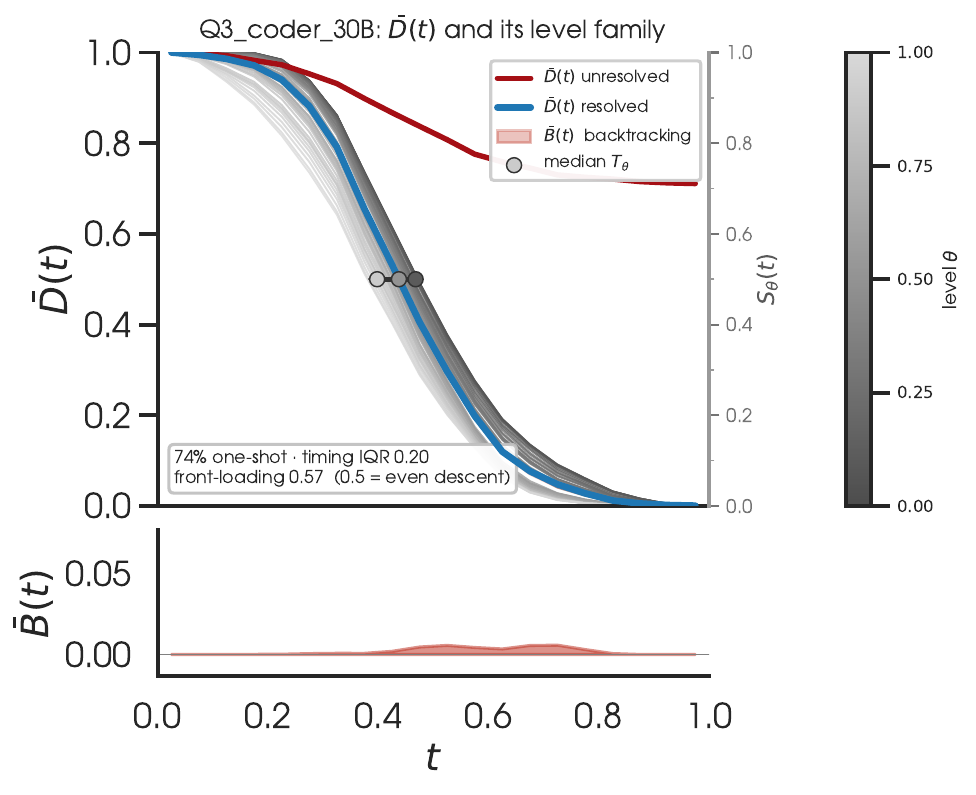} \\
\includegraphics[width=0.31\linewidth]{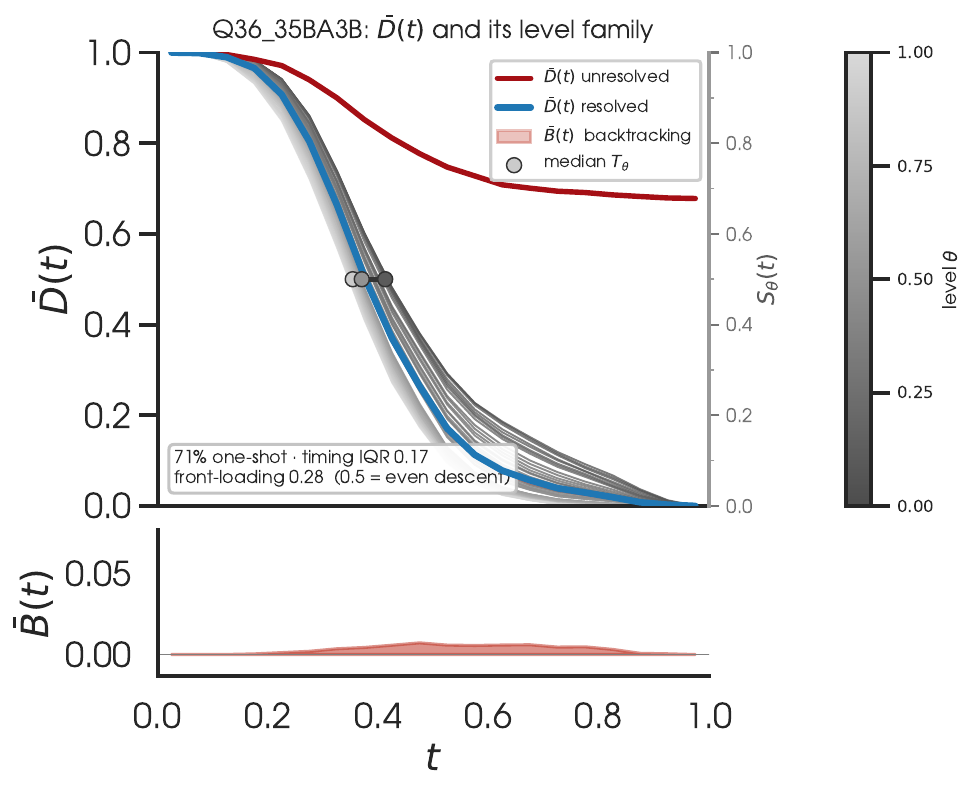} &
\includegraphics[width=0.31\linewidth]{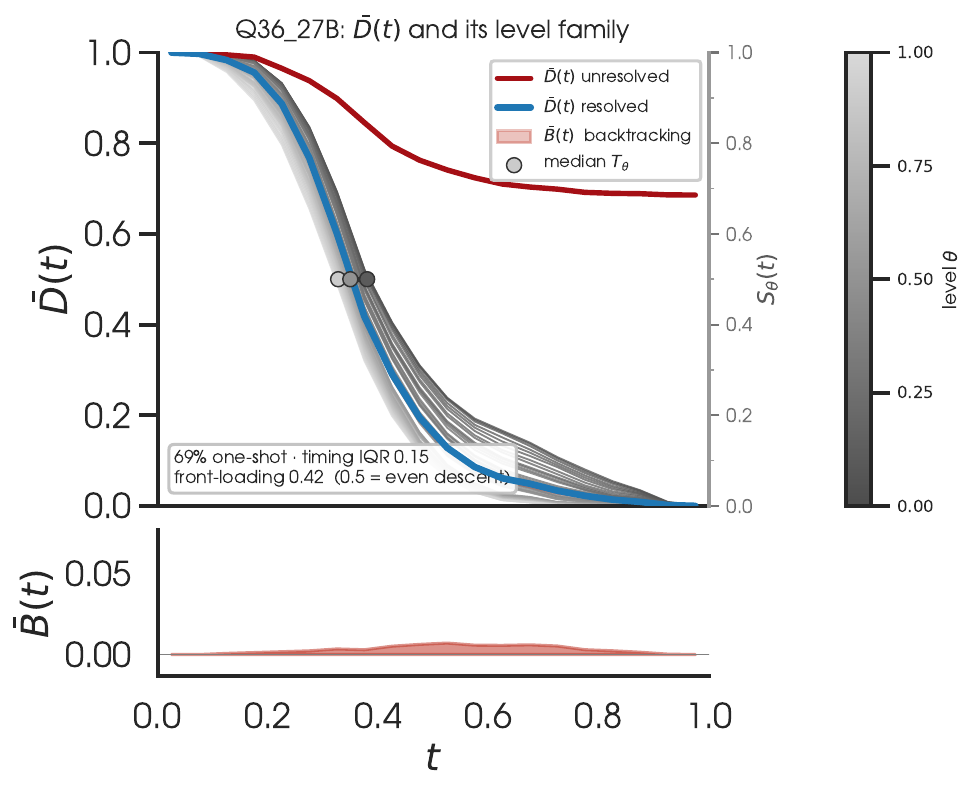} &
\includegraphics[width=0.31\linewidth]{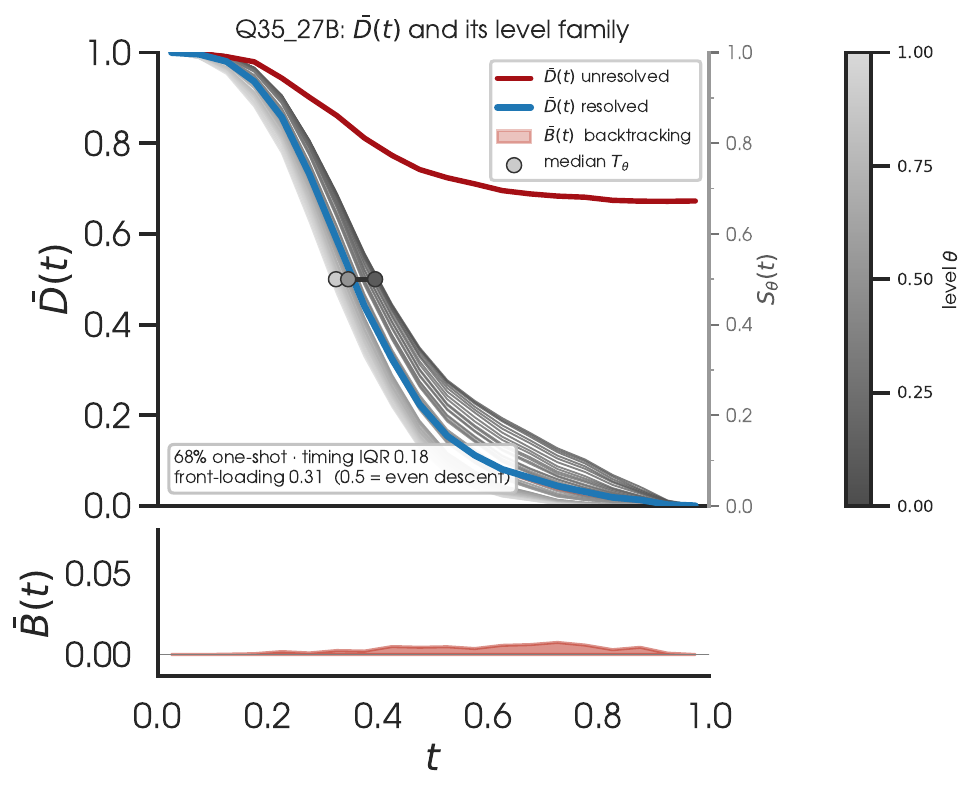} \\
\end{tabular}
\caption{Level families on SWE-Bench-Verified (part 2 of 2): Gemini, Grok and Qwen families. Continued from Figure~\ref{fig:metrics-sbv-soldist}.}
\label{fig:metrics-sbv-soldist-b}
\end{figure}

\paragraph{Per-model level families.}
Figures~\ref{fig:metrics-sbv-soldist}--\ref{fig:metrics-sbv-soldist-b} show, for each model, the level family of its resolved trajectories together with the mean $\bar D(t)$ and the backtracking term $\bar B(t)$ of Section~\ref{sec:distance}. Resolved curves are self-anchored to terminate at $0$ (Section~\ref{app:metrics:soldist:zero}); the dark-red curve is the unresolved mean.

\begin{figure}[p]
\centering
\setlength{\tabcolsep}{1pt}
\renewcommand{\arraystretch}{0.5}
\begin{tabular}{ccc}
\includegraphics[width=0.31\linewidth]{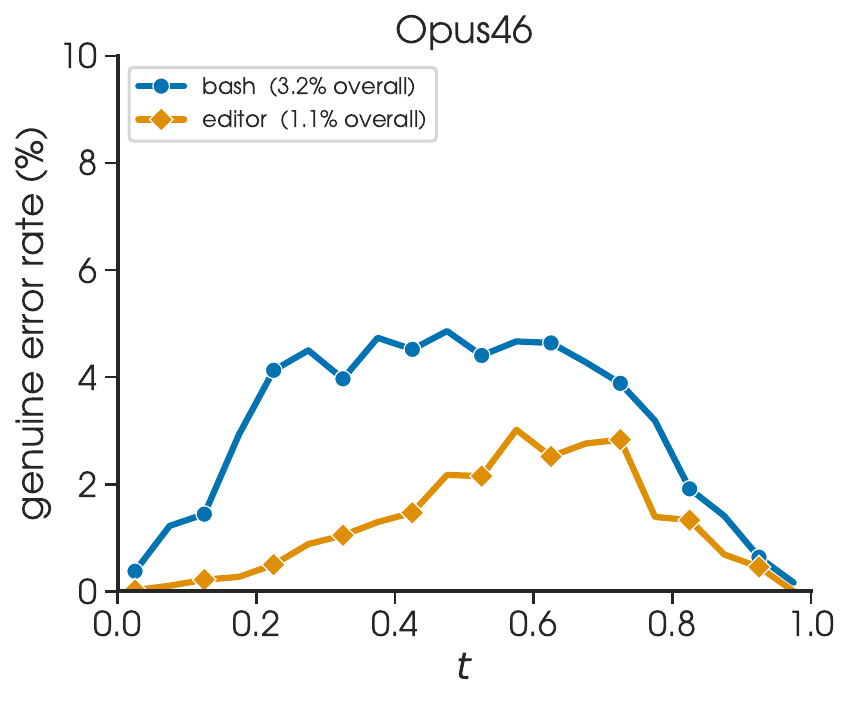} &
\includegraphics[width=0.31\linewidth]{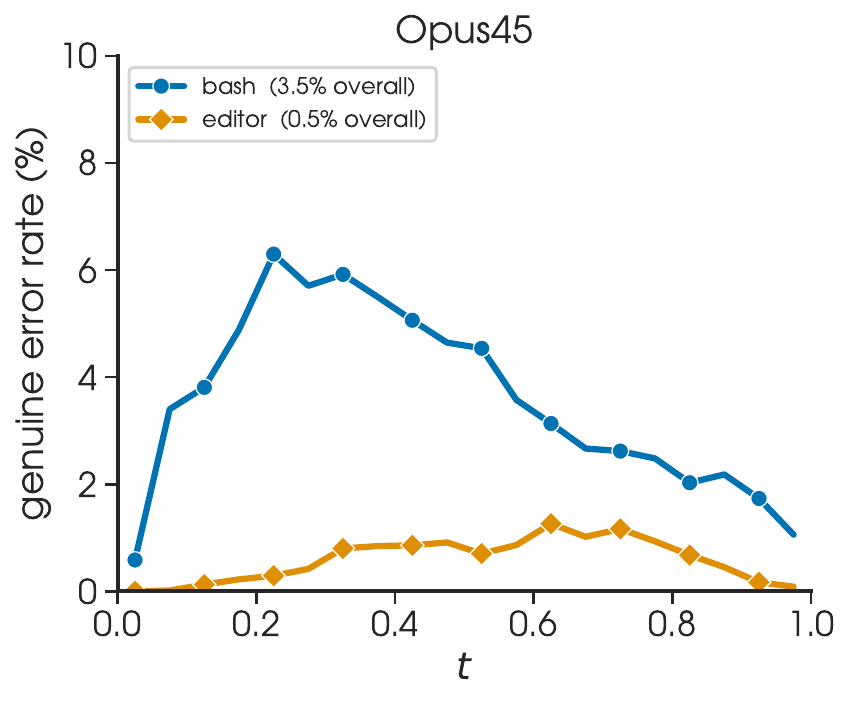} &
\includegraphics[width=0.31\linewidth]{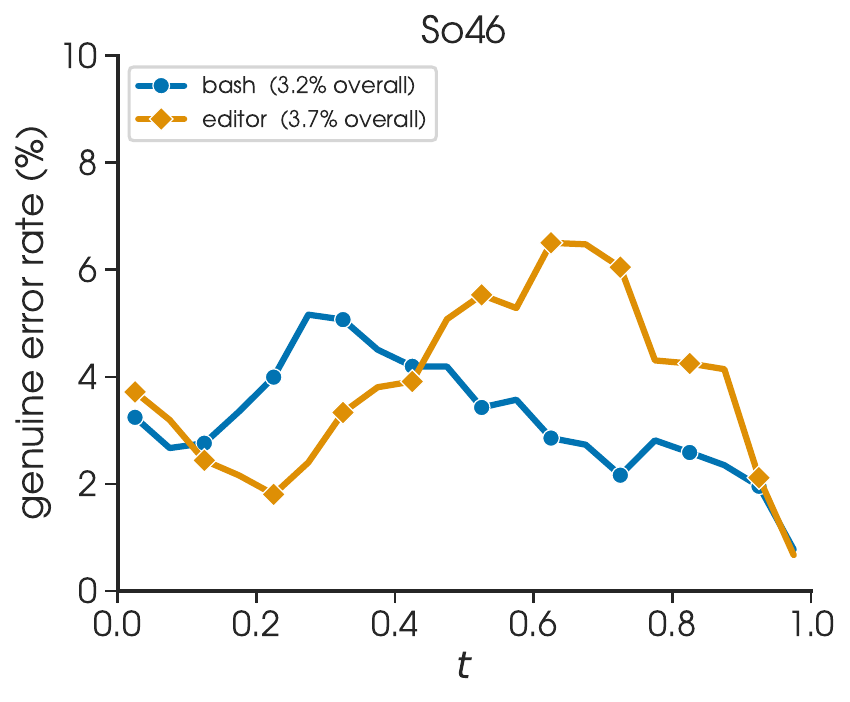} \\
\includegraphics[width=0.31\linewidth]{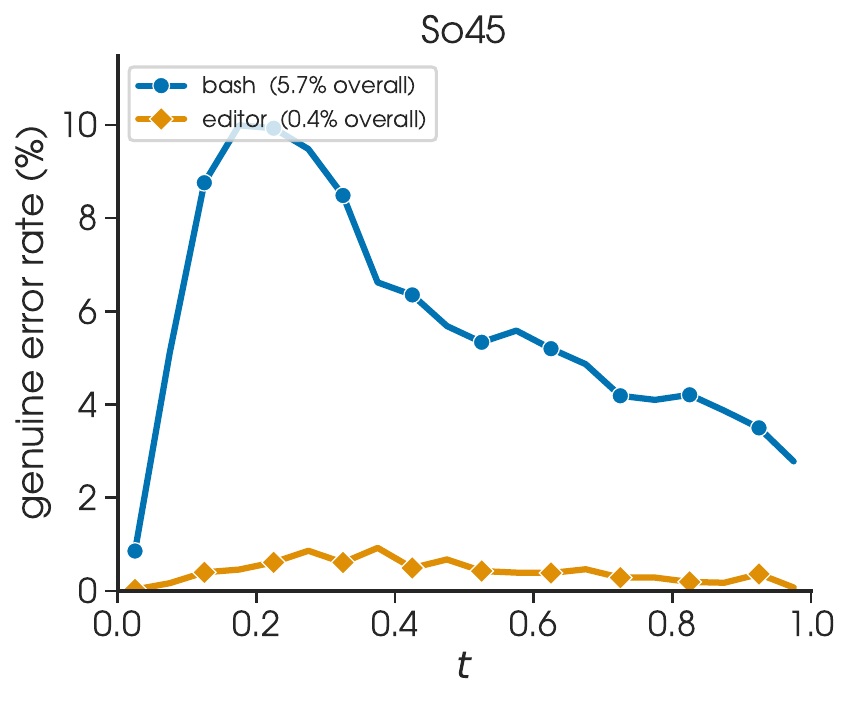} &
\includegraphics[width=0.31\linewidth]{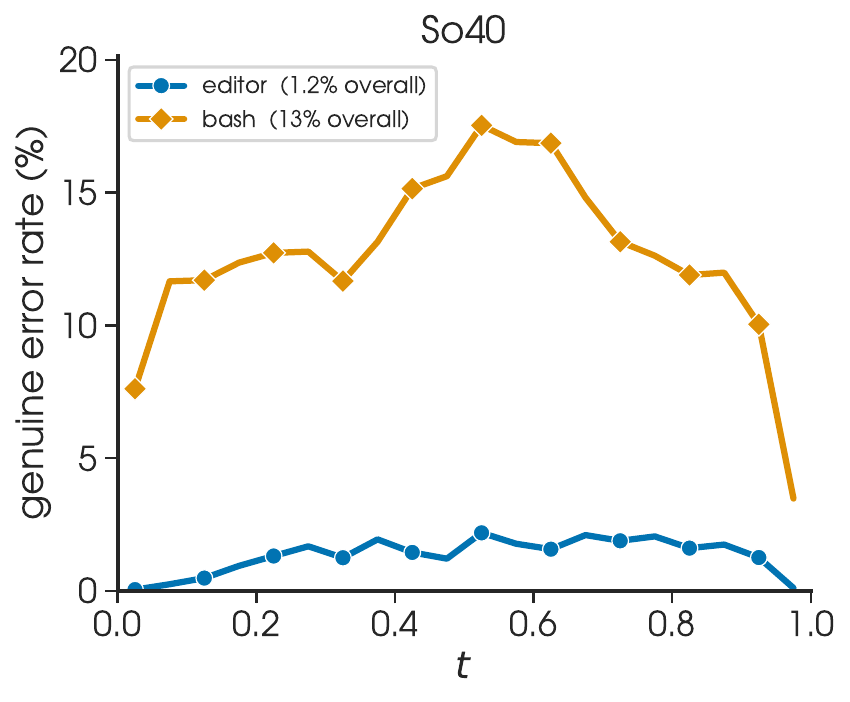} &
\includegraphics[width=0.31\linewidth]{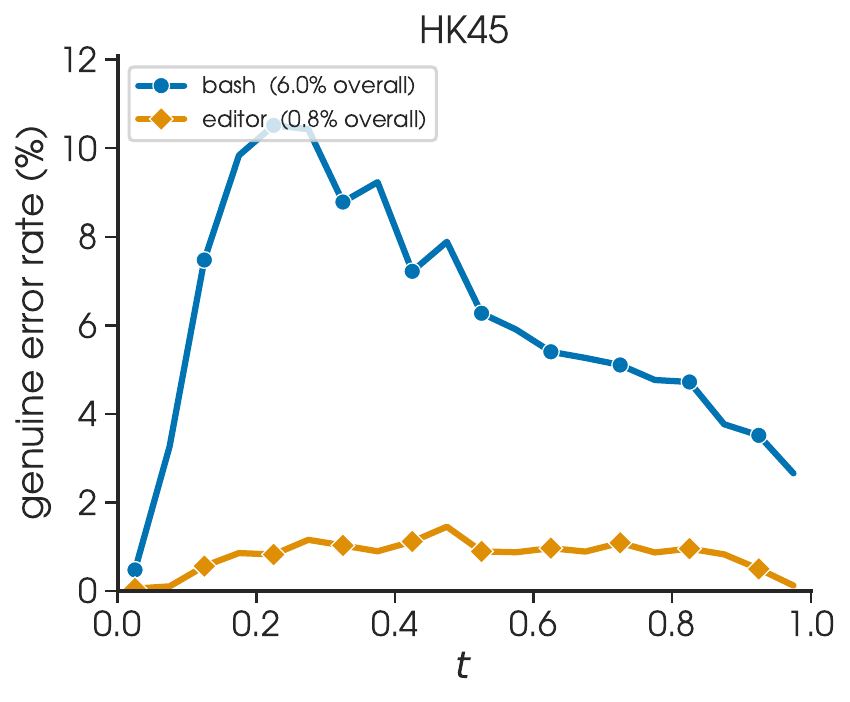} \\
\includegraphics[width=0.31\linewidth]{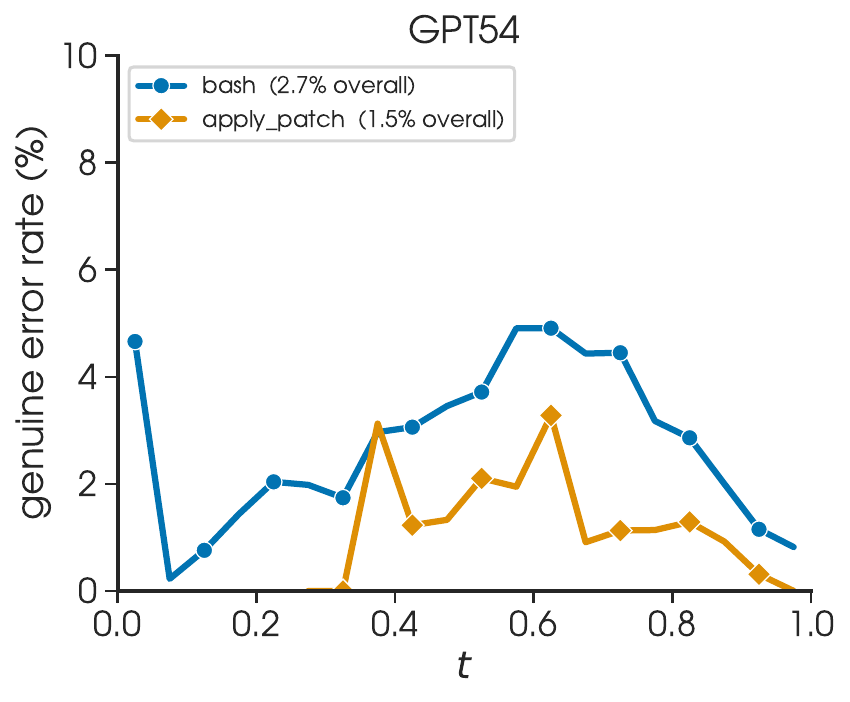} &
\includegraphics[width=0.31\linewidth]{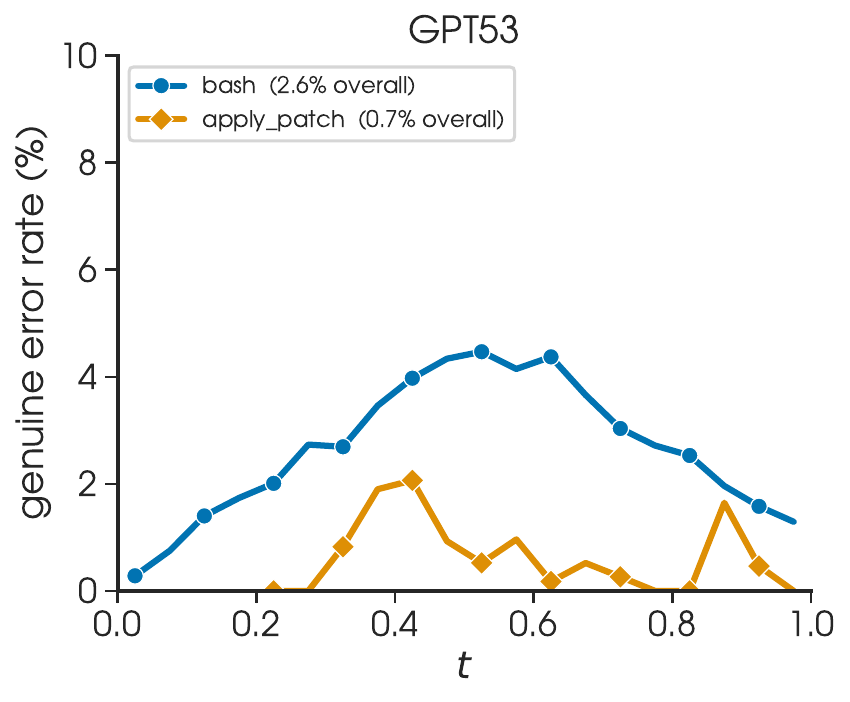} &
\includegraphics[width=0.31\linewidth]{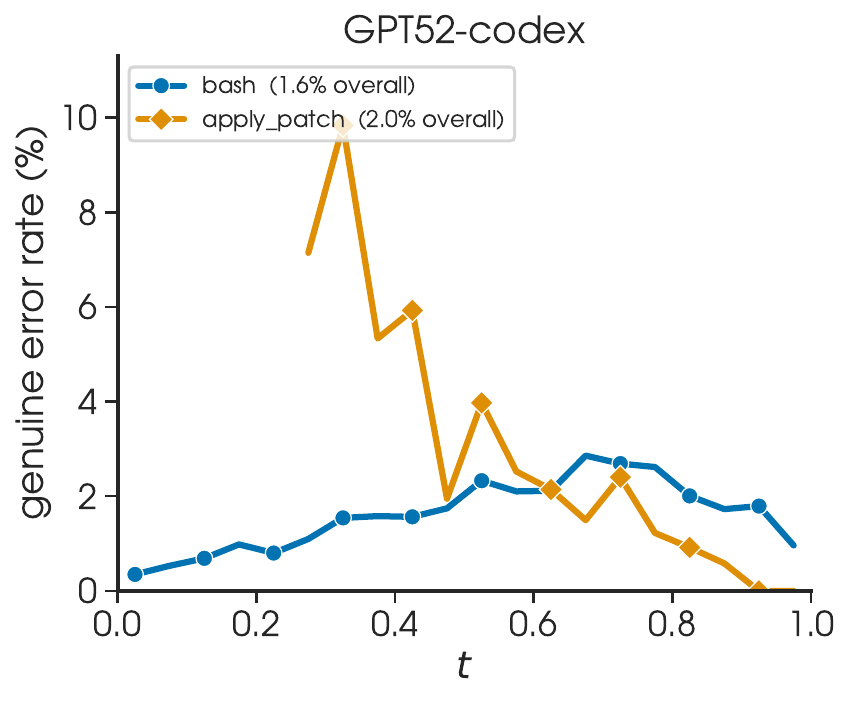} \\
\includegraphics[width=0.31\linewidth]{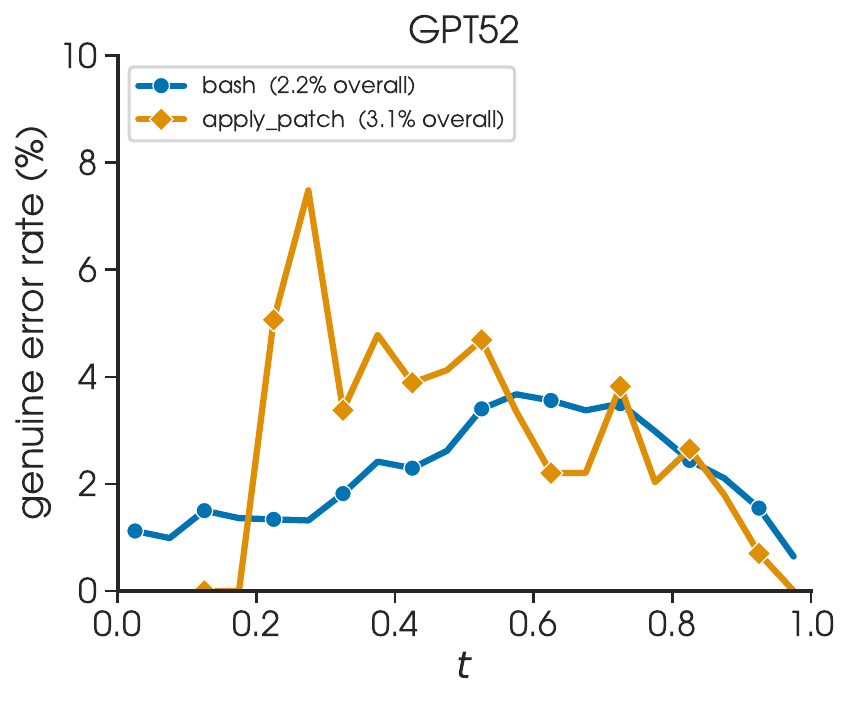} &
\includegraphics[width=0.31\linewidth]{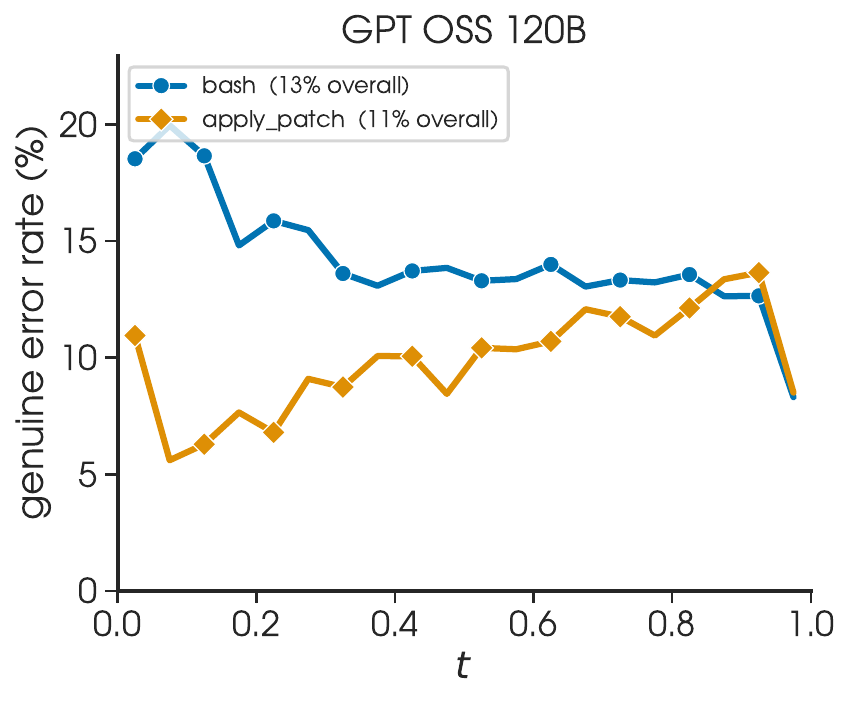} &
\includegraphics[width=0.31\linewidth]{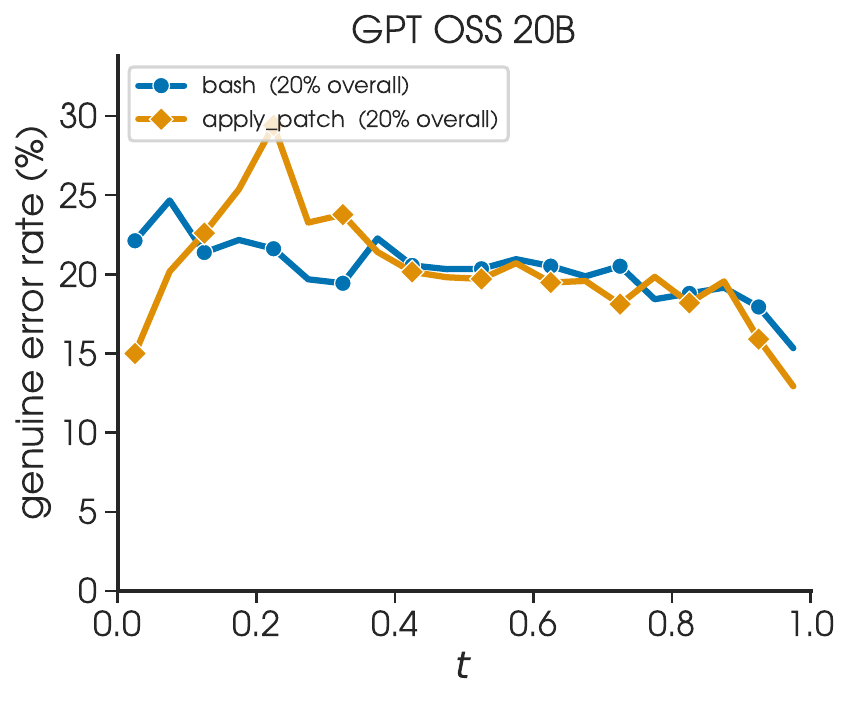} \\
\end{tabular}
\caption{Genuine tool-error rate per cycle (part 1 of 2): Anthropic Claude and OpenAI families. Benign exits (\texttt{grep} no-match, \texttt{pytest} test failures, etc.)\ are excluded. Tool families broken out within each sub-plot.}
\label{fig:metrics-sbv-toolerr}
\end{figure}

\begin{figure}[p]
\centering
\setlength{\tabcolsep}{1pt}
\renewcommand{\arraystretch}{0.5}
\begin{tabular}{ccc}
\includegraphics[width=0.31\linewidth]{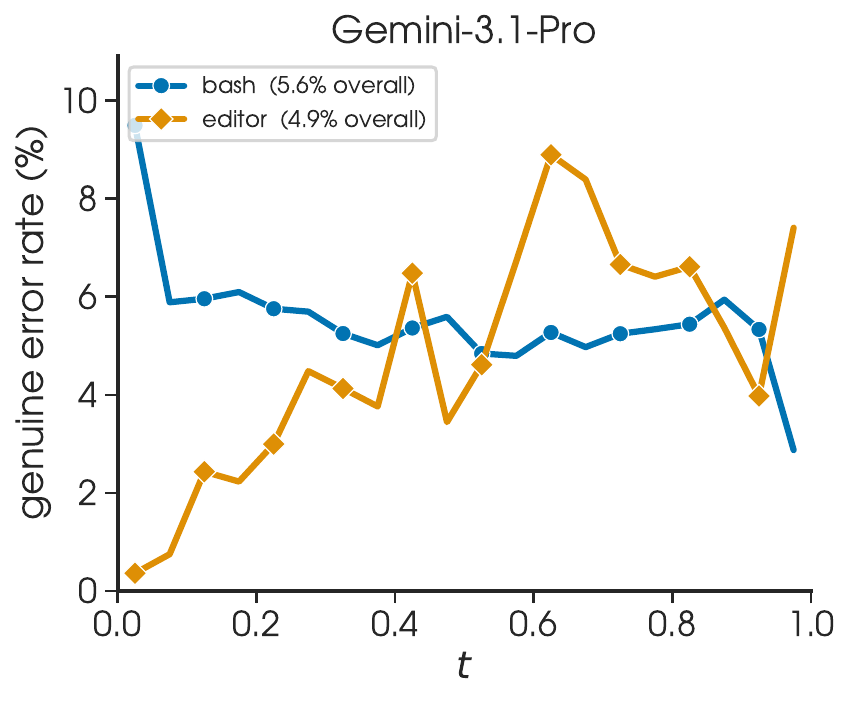} &
\includegraphics[width=0.31\linewidth]{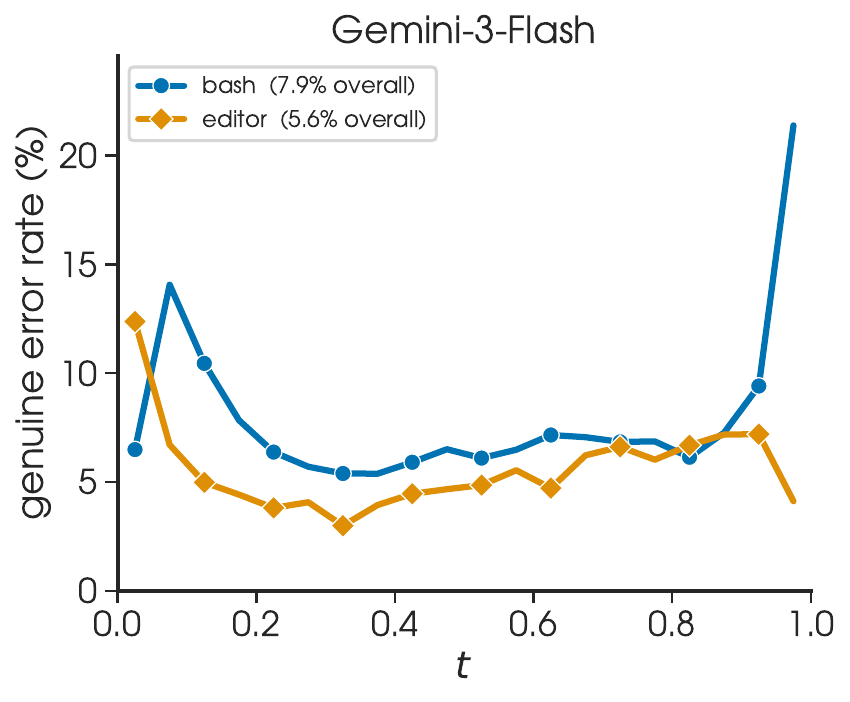} & \\
\includegraphics[width=0.31\linewidth]{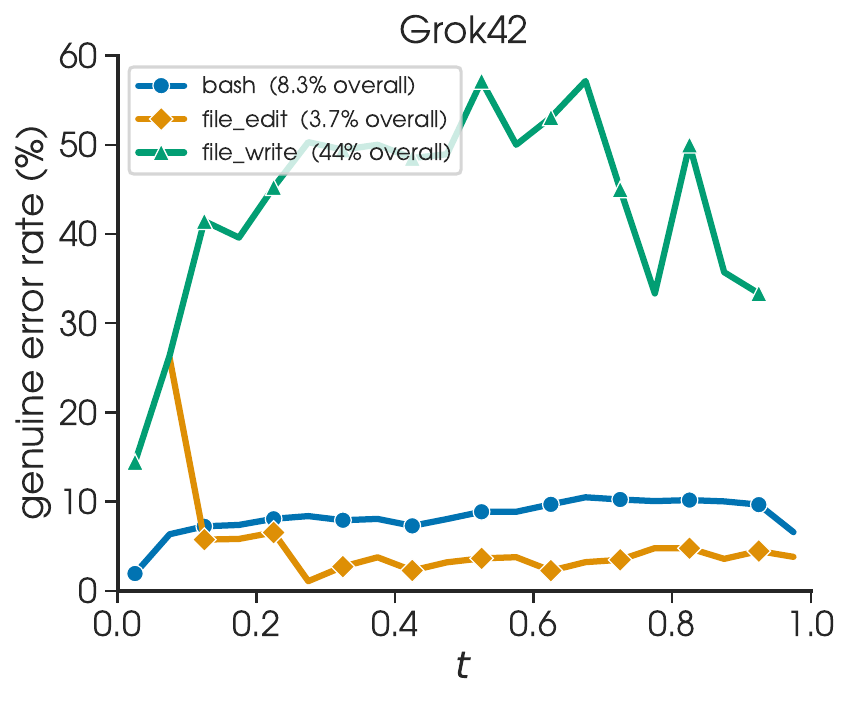} & & \\
\includegraphics[width=0.31\linewidth]{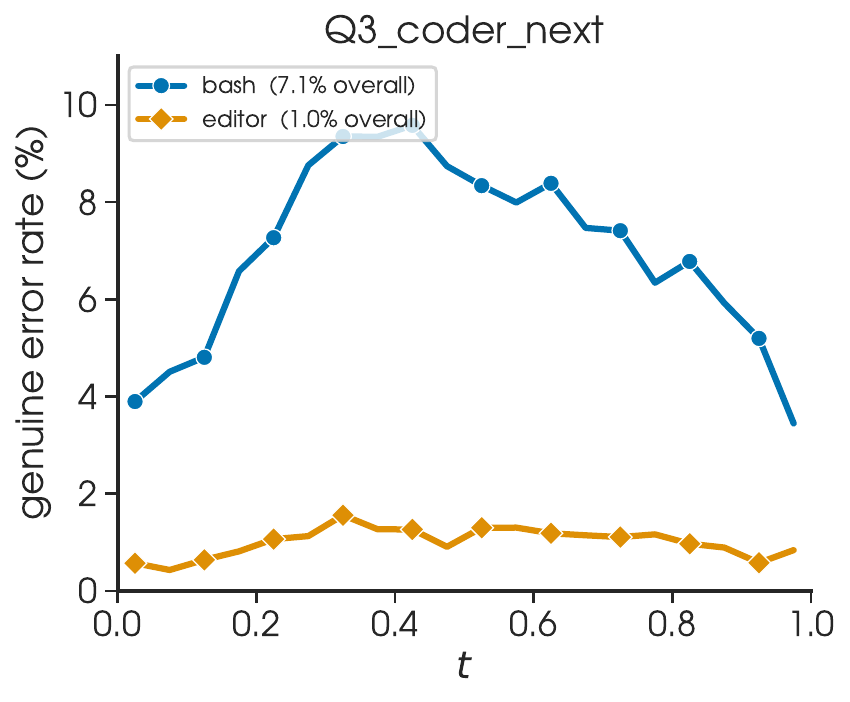} &
\includegraphics[width=0.31\linewidth]{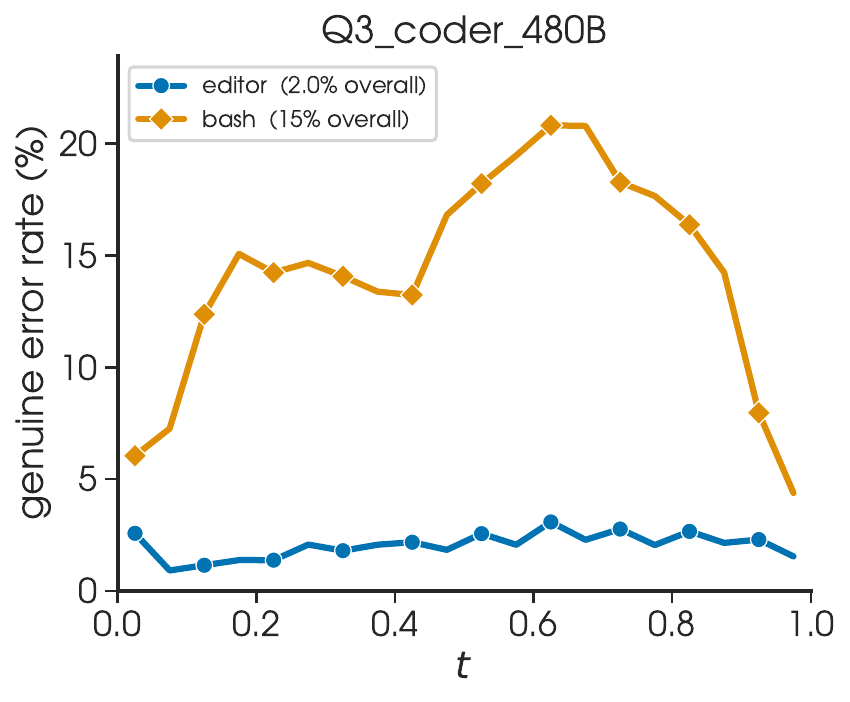} &
\includegraphics[width=0.31\linewidth]{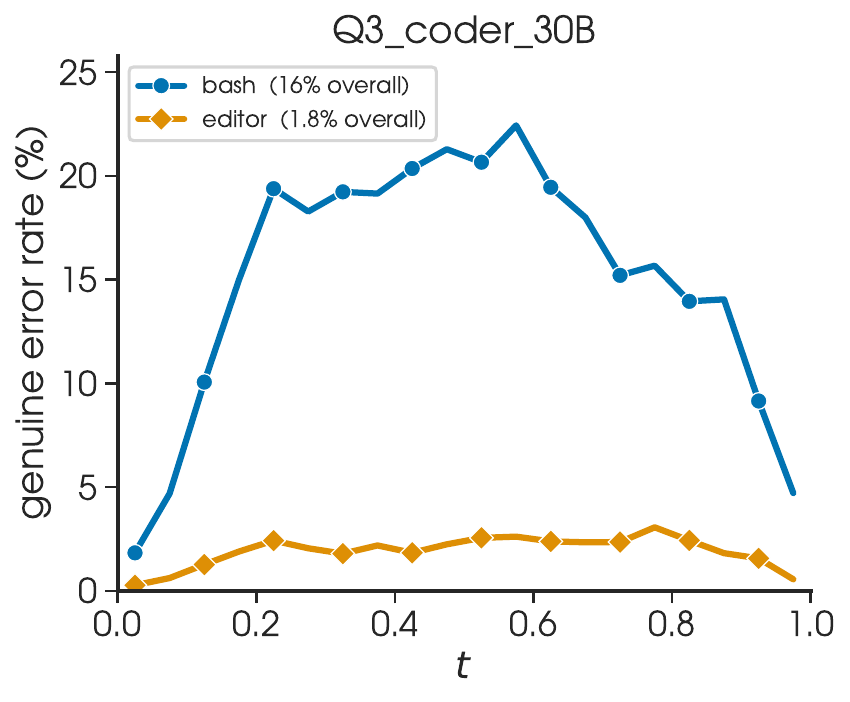} \\
\includegraphics[width=0.31\linewidth]{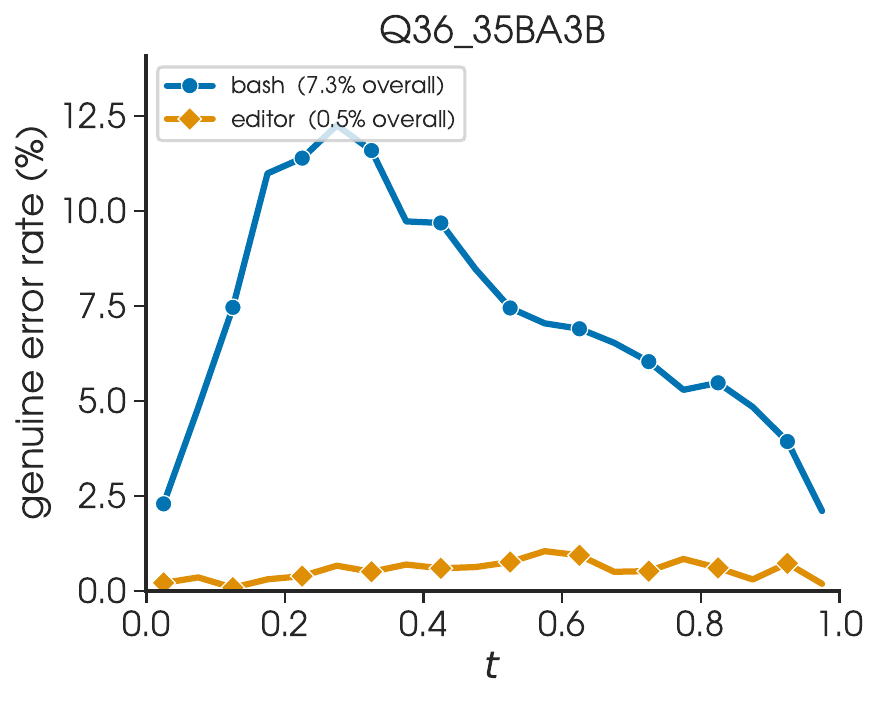} &
\includegraphics[width=0.31\linewidth]{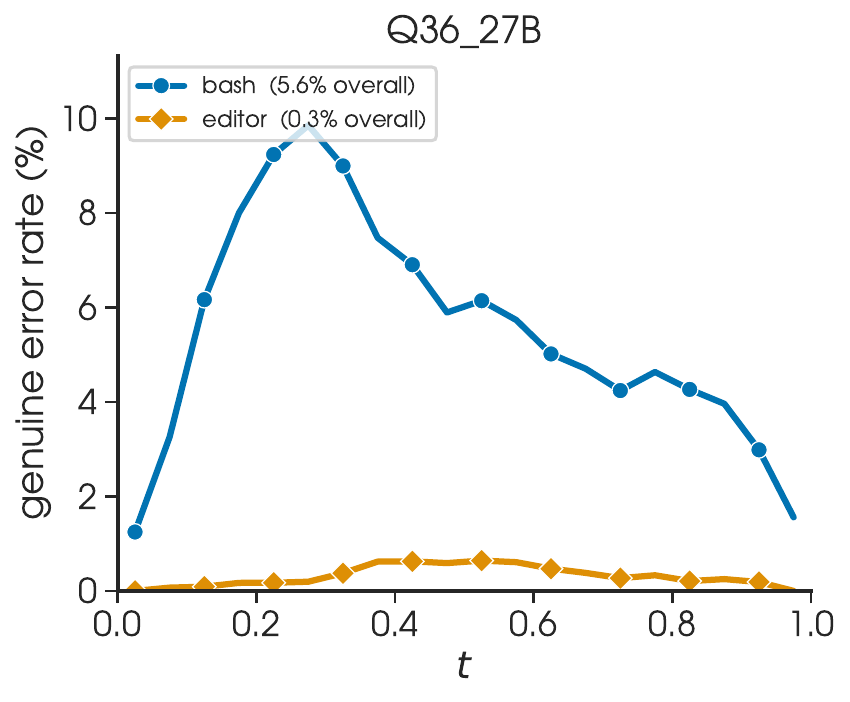} &
\includegraphics[width=0.31\linewidth]{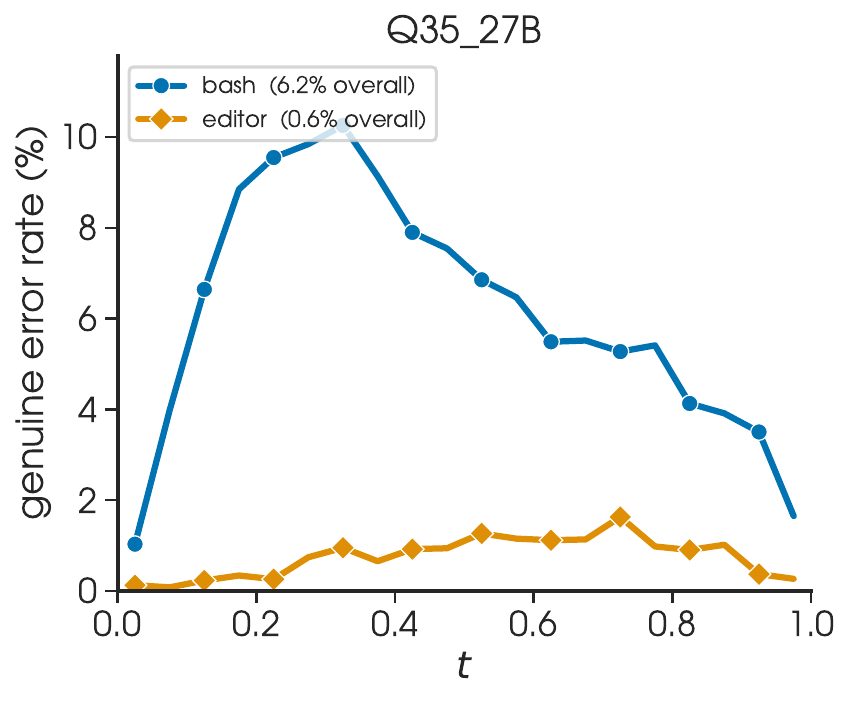} \\
\end{tabular}
\caption{Genuine tool-error rate per cycle (part 2 of 2): Gemini, Grok and Qwen families. Continued from Figure~\ref{fig:metrics-sbv-toolerr}.}
\label{fig:metrics-sbv-toolerr-b}
\end{figure}

\paragraph{Per-model genuine tool-error rate over cycles.}
Figure~\ref{fig:metrics-sbv-toolerr} plots the rate of \emph{genuine} tool errors per cycle (R8 = \texttt{genuine}, see Section~\ref{app:metrics:rubric}), broken out by tool family (\texttt{bash} vs. \texttt{editor}). Benign non-zero exits (\texttt{grep} no-match, \texttt{pytest} reporting test failures, etc.)\ are excluded so spikes in this metric are real harness friction and not test outcomes.
The standout is Grok~4.2, whose \texttt{file\_write} errors run near $44\%$: a tool-contract mismatch rather than a reasoning failure, since \texttt{file\_write} is create-only and confined to the workspace. Scratch scripts written to \texttt{/tmp/} ($74\%$) and re-writes of existing files instead of \texttt{file\_edit} ($26\%$) account for all of it, and both are recoverable on retry.

\begin{figure}[p]
\centering
\includegraphics[width=0.65\linewidth]{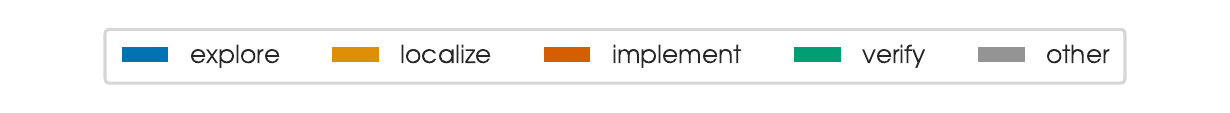}\\[2pt]
\setlength{\tabcolsep}{1pt}
\renewcommand{\arraystretch}{0.5}
\begin{tabular}{ccc}
\includegraphics[width=0.31\linewidth]{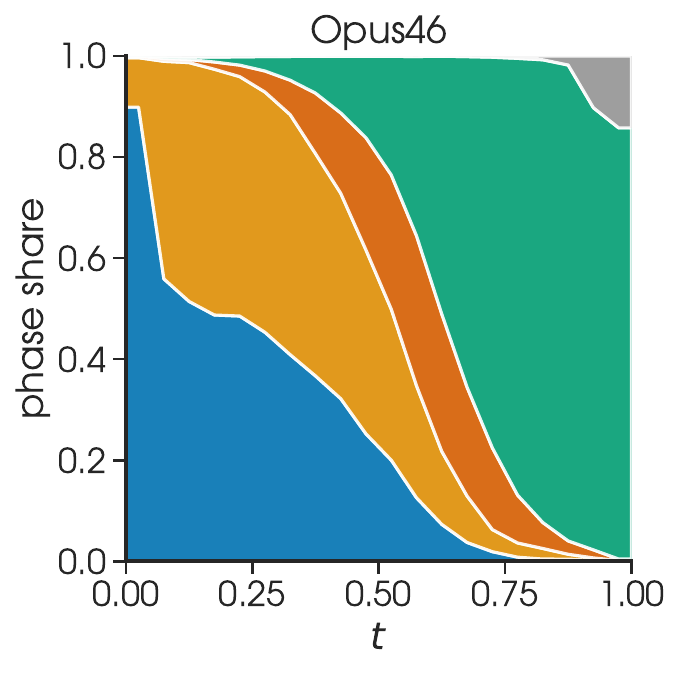} &
\includegraphics[width=0.31\linewidth]{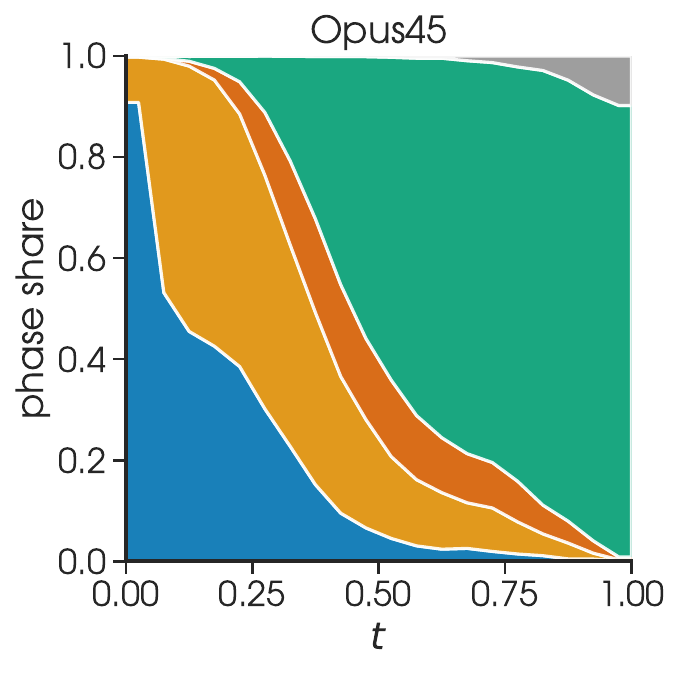} &
\includegraphics[width=0.31\linewidth]{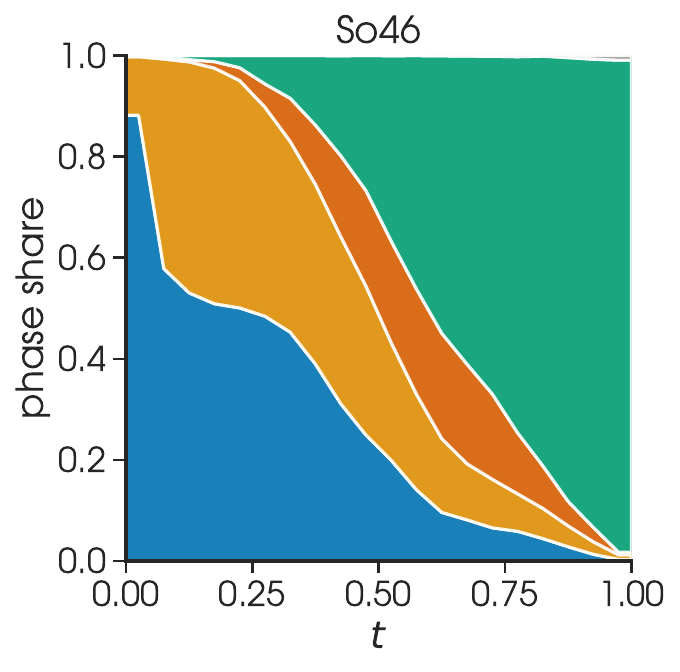} \\
\includegraphics[width=0.31\linewidth]{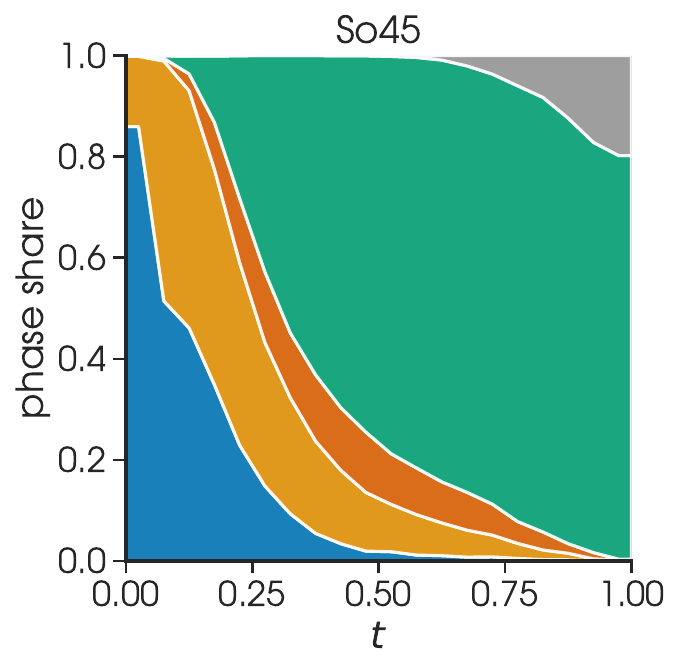} &
\includegraphics[width=0.31\linewidth]{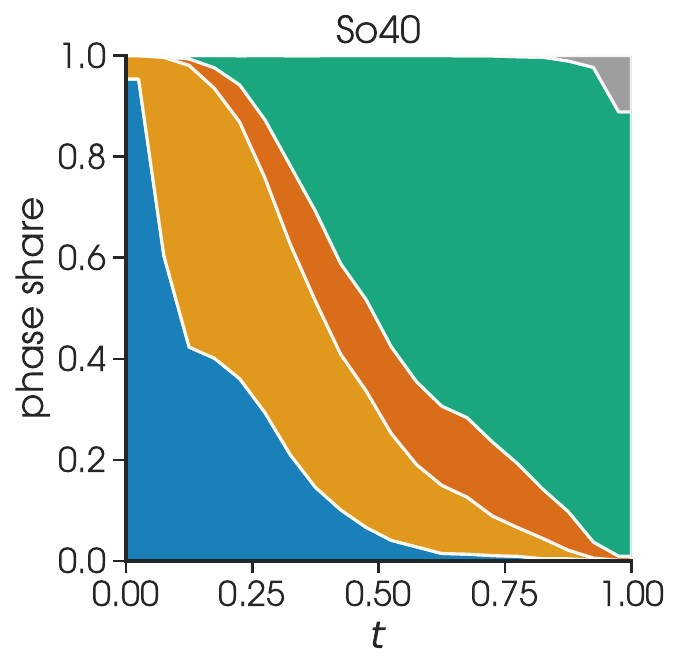} &
\includegraphics[width=0.31\linewidth]{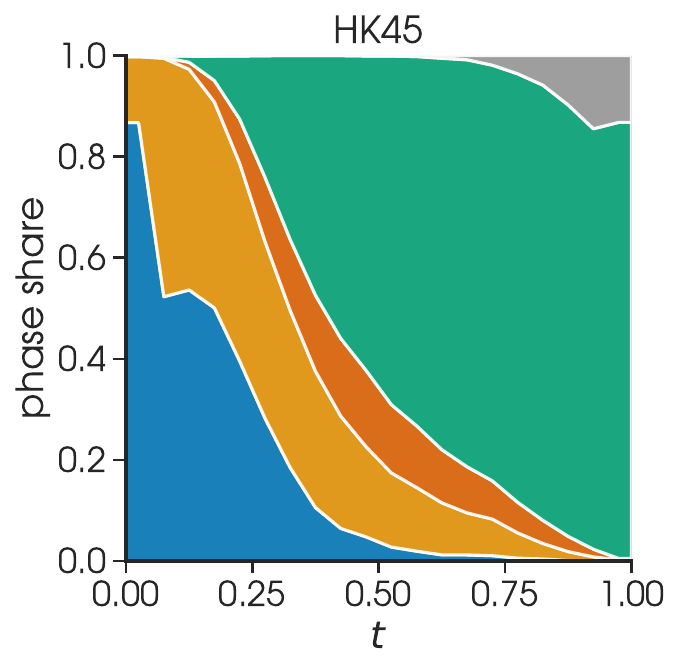} \\
\includegraphics[width=0.31\linewidth]{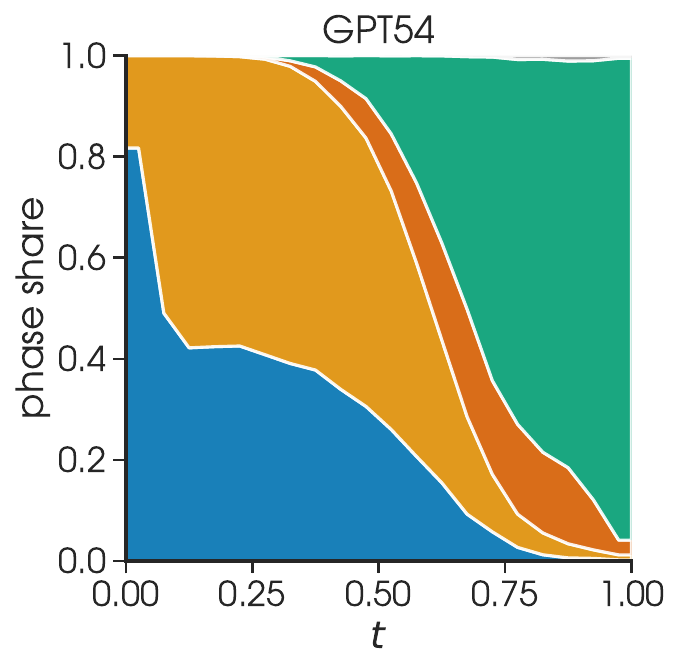} &
\includegraphics[width=0.31\linewidth]{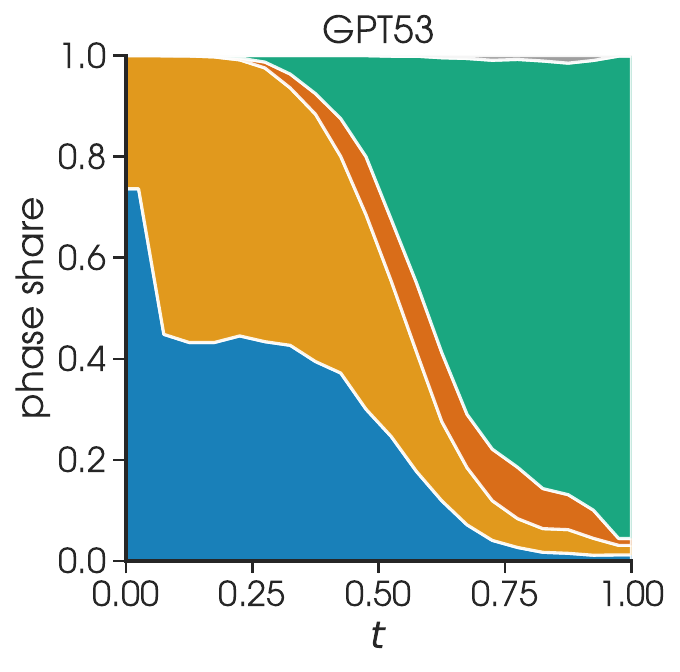} &
\includegraphics[width=0.31\linewidth]{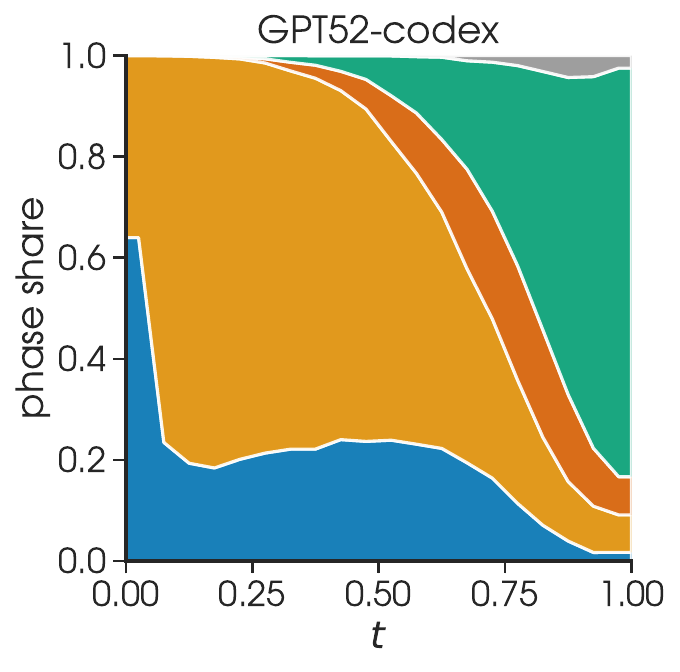} \\
\includegraphics[width=0.31\linewidth]{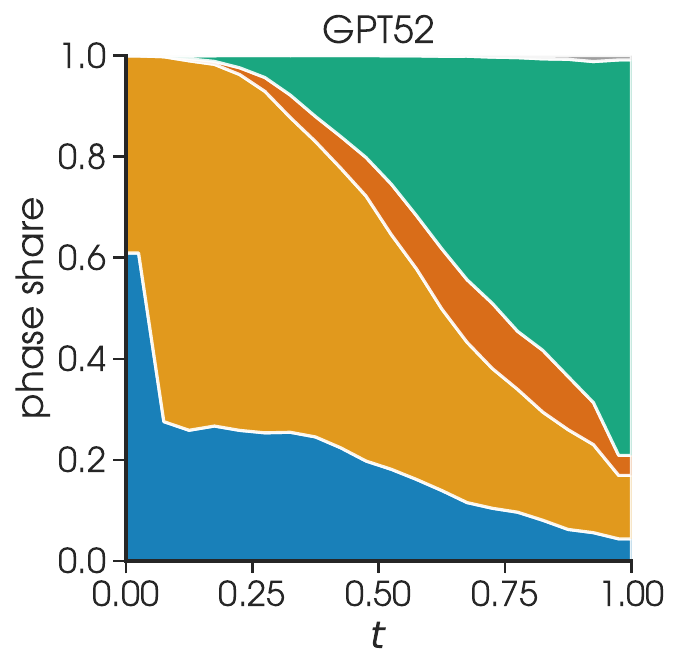} &
\includegraphics[width=0.31\linewidth]{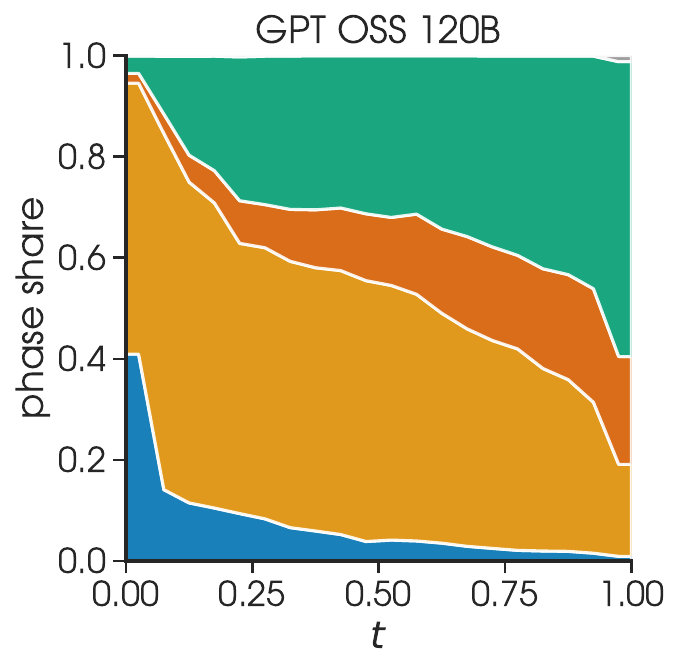} &
\includegraphics[width=0.31\linewidth]{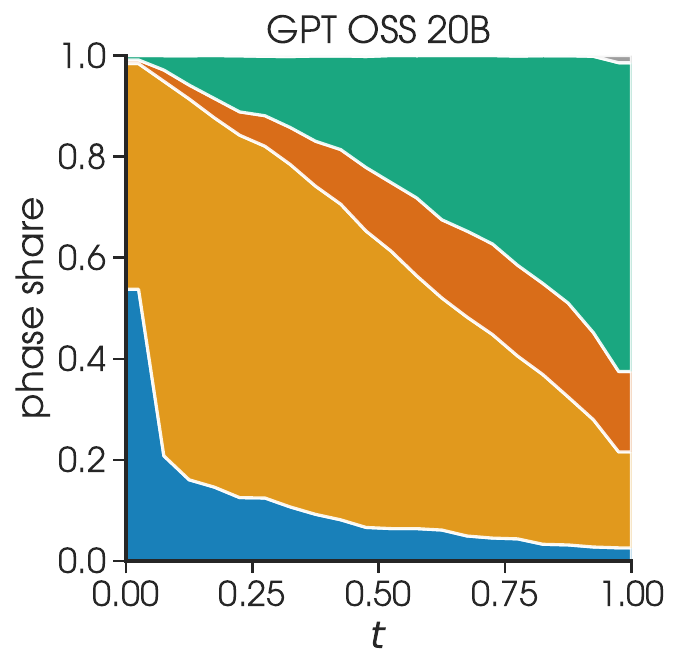} \\
\end{tabular}
\caption{\textbf{Phase composition surfaces model-specific schedules.} Per-cycle share of the four R6 (see LLM judge rubrics in Table\,\ref{tab:judge-buckets}) phases (\texttt{explore},
\texttt{localize}, \texttt{implement}, \texttt{verify}) vs.\ normalised cycle
position $t$ on SWE-Bench-Verified, for Claude and GPT families (part 1 of 2). Given the
same tasks, models schedule their cycles differently. The implementation peak shifts in position and width, and the 
verify phase carries different long-tail mass.}
\label{fig:metrics-sbv-phase}
\end{figure}

\begin{figure}[p]
\centering
\includegraphics[width=0.65\linewidth]{swe_verif/phase_legend.pdf}\\[2pt]
\setlength{\tabcolsep}{1pt}
\renewcommand{\arraystretch}{0.5}
\begin{tabular}{ccc}
\includegraphics[width=0.31\linewidth]{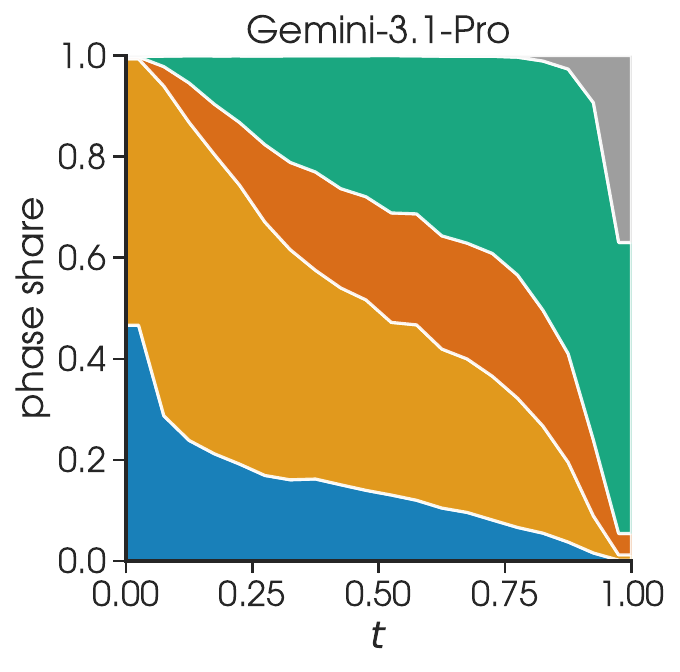} &
\includegraphics[width=0.31\linewidth]{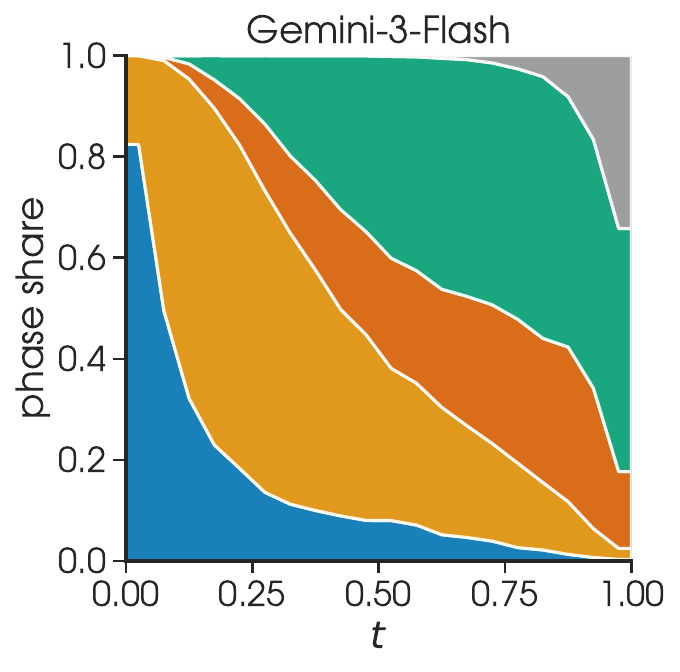} & \\
\includegraphics[width=0.31\linewidth]{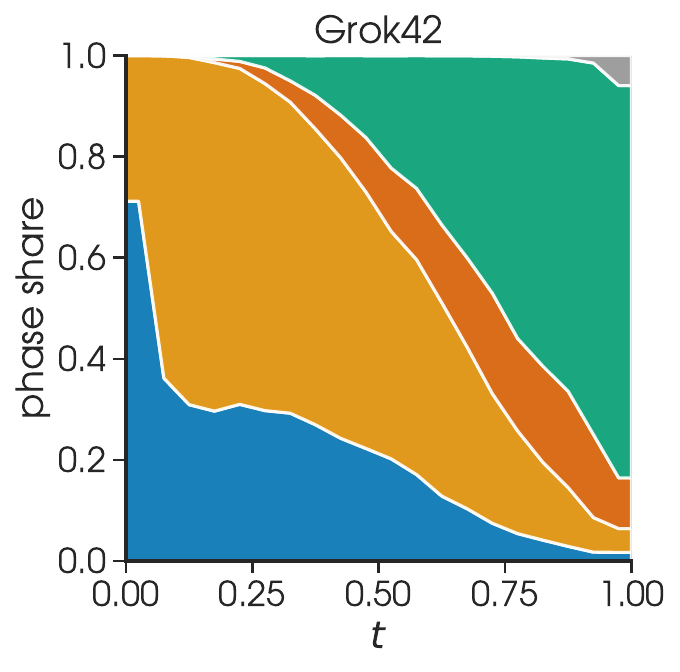} & & \\
\includegraphics[width=0.31\linewidth]{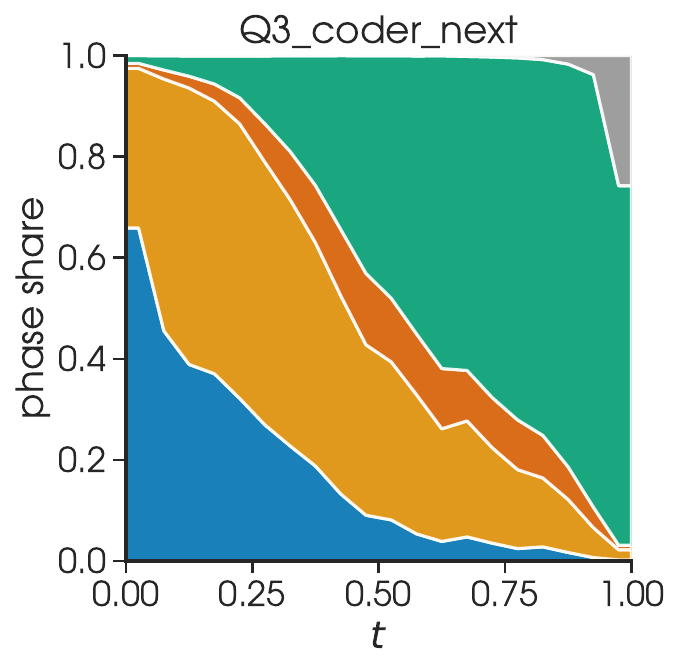} &
\includegraphics[width=0.31\linewidth]{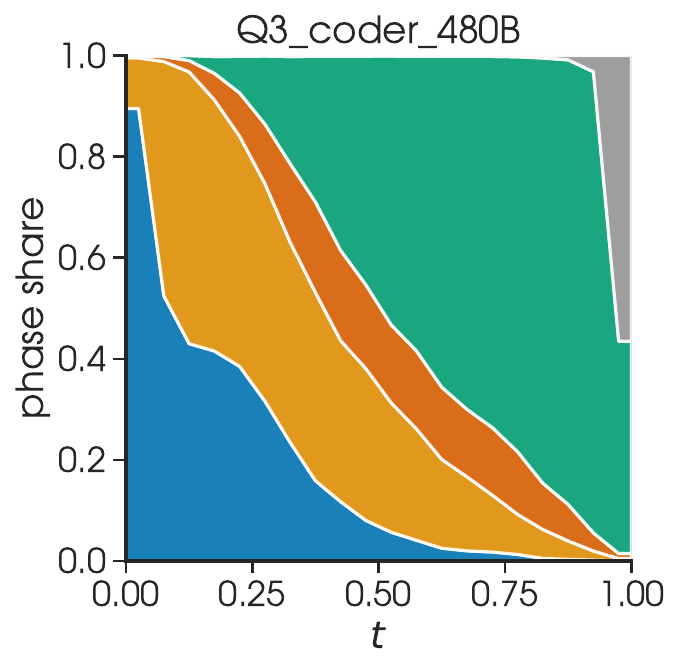} &
\includegraphics[width=0.31\linewidth]{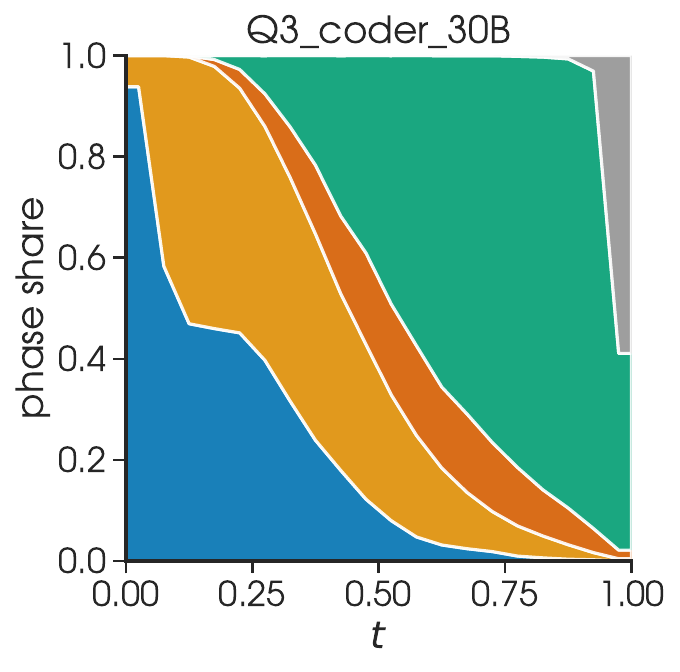} \\
\includegraphics[width=0.31\linewidth]{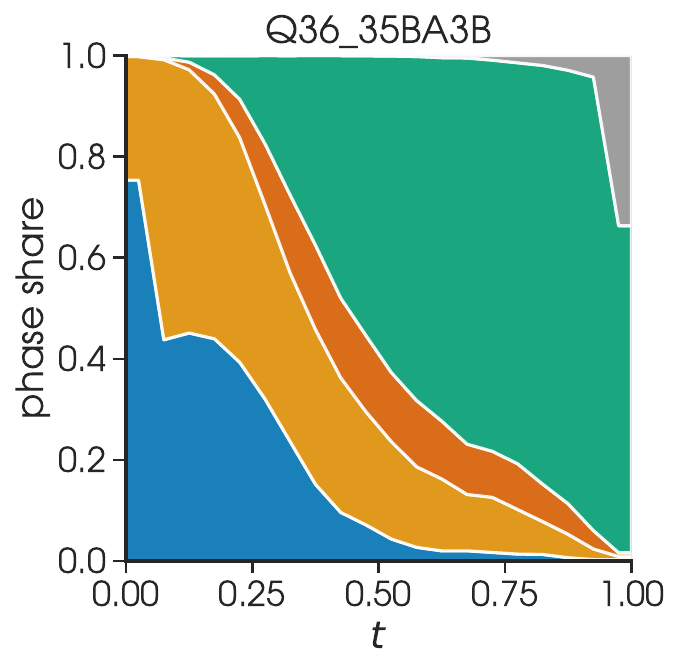} &
\includegraphics[width=0.31\linewidth]{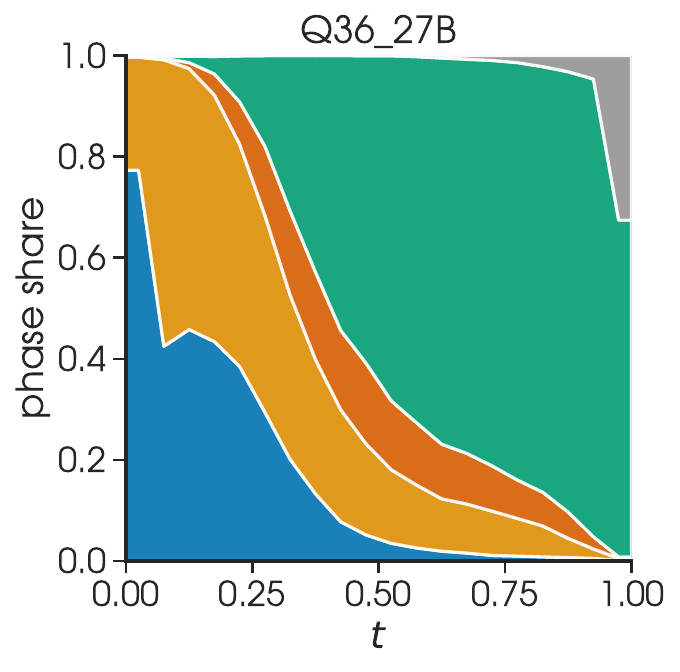} &
\includegraphics[width=0.31\linewidth]{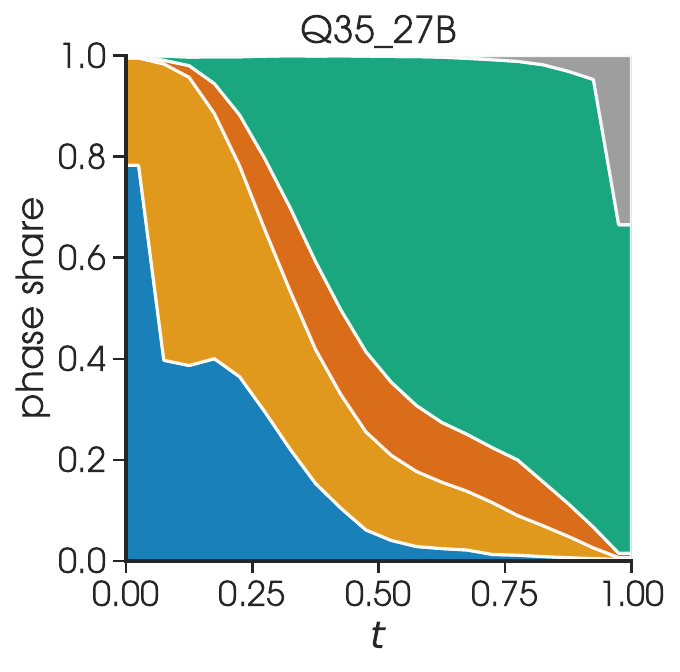} \\
\end{tabular}
\caption{Phase composition (part 2 of 2): Gemini, Grok and Qwen families. Continued from Figure~\ref{fig:metrics-sbv-phase}.}
\label{fig:metrics-sbv-phase-b}
\end{figure}

\paragraph{Per-model phase composition.}
Figures~\ref{fig:metrics-sbv-phase}--\ref{fig:metrics-sbv-phase-b} split each trajectory across the four R6 phases (Section~\ref{sec:results:trajectory-metrics}). GPT-5 trajectories collapse explore/localize into implement/verify, while Qwen3-coder and Gemini spread time more evenly.

\begin{figure}[h]
\centering
\includegraphics[width=0.85\linewidth]{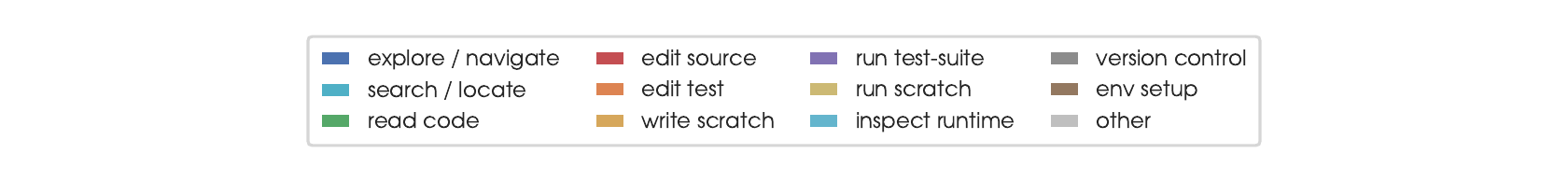}\\[2pt]
\setlength{\tabcolsep}{1pt}
\renewcommand{\arraystretch}{0.5}
\begin{tabular}{ccc}
\includegraphics[width=0.31\linewidth]{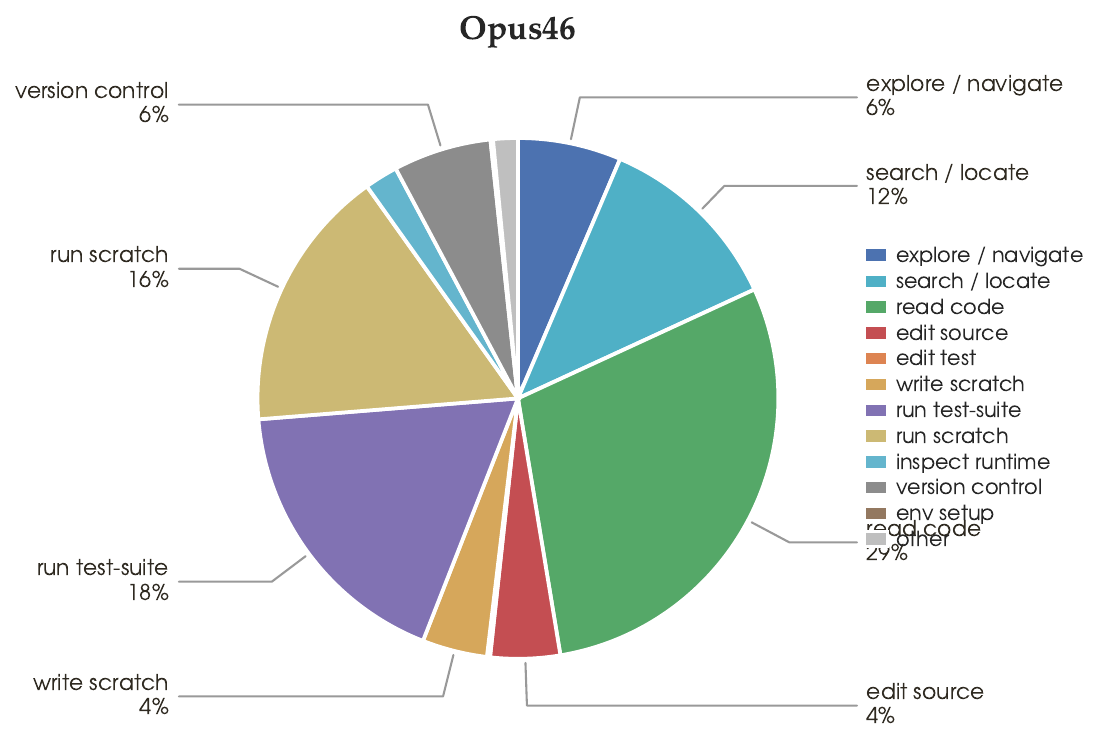} &
\includegraphics[width=0.31\linewidth]{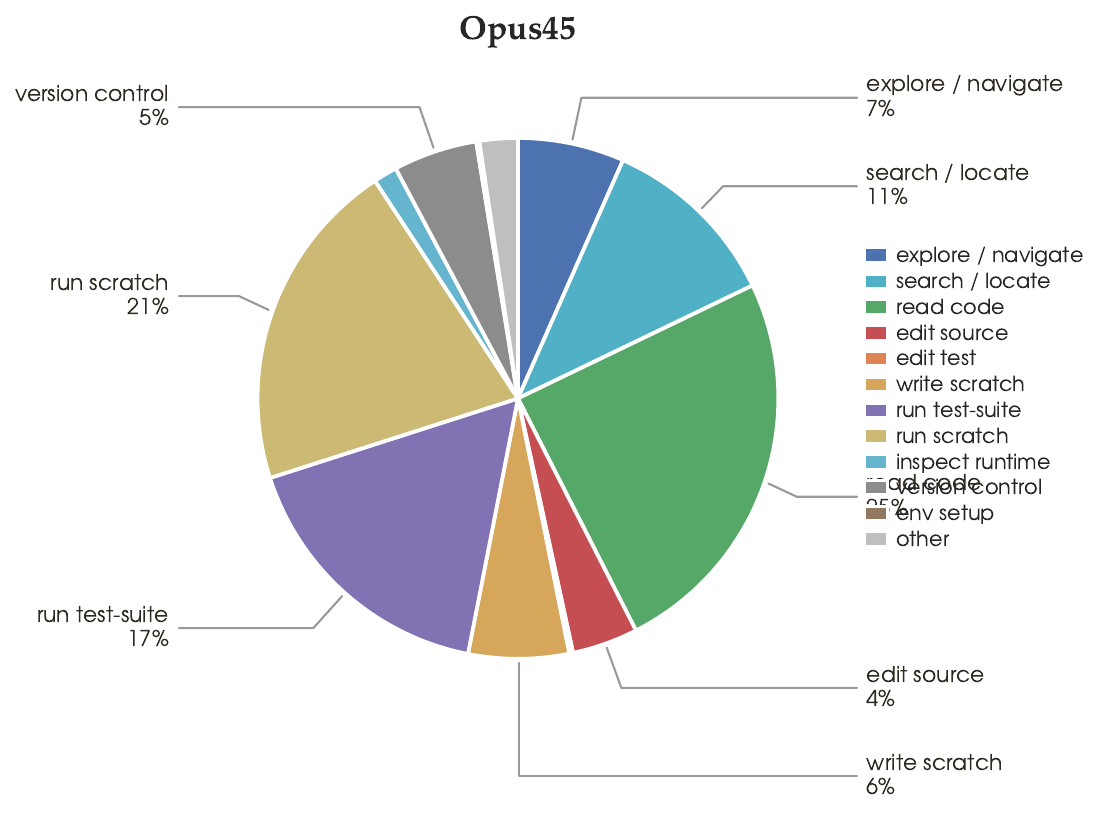} &
\includegraphics[width=0.31\linewidth]{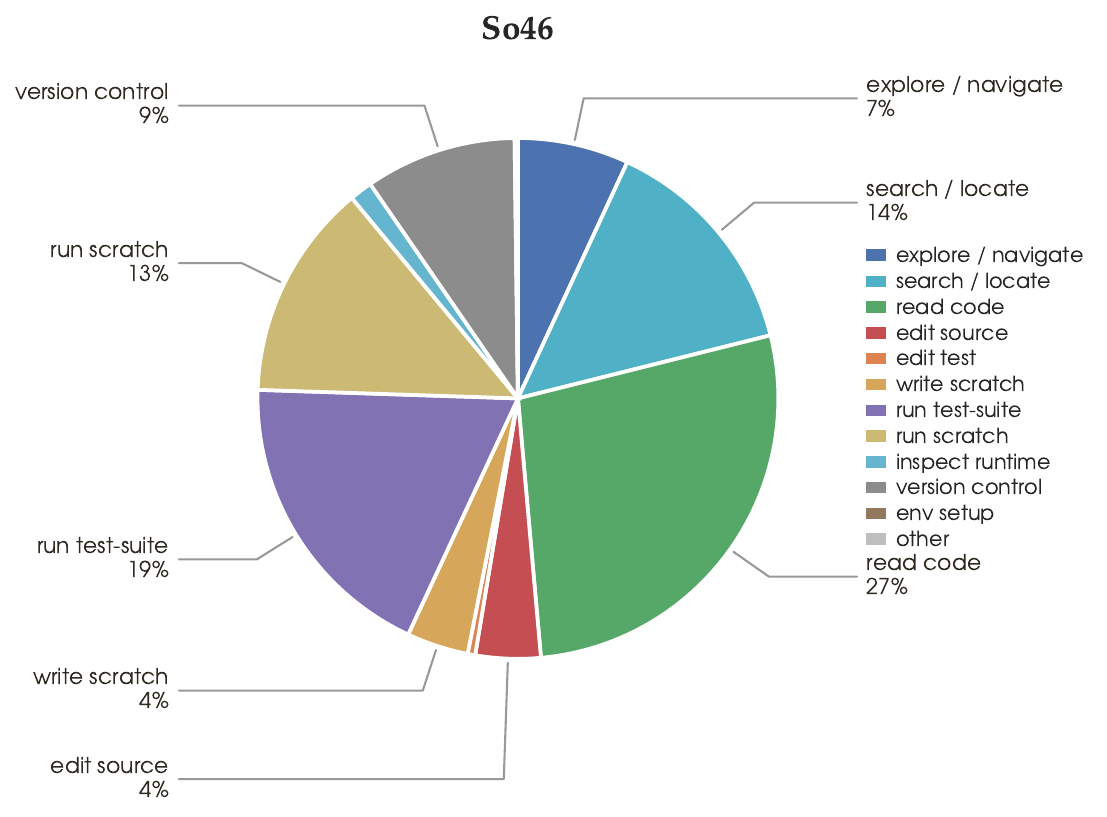} \\
\includegraphics[width=0.31\linewidth]{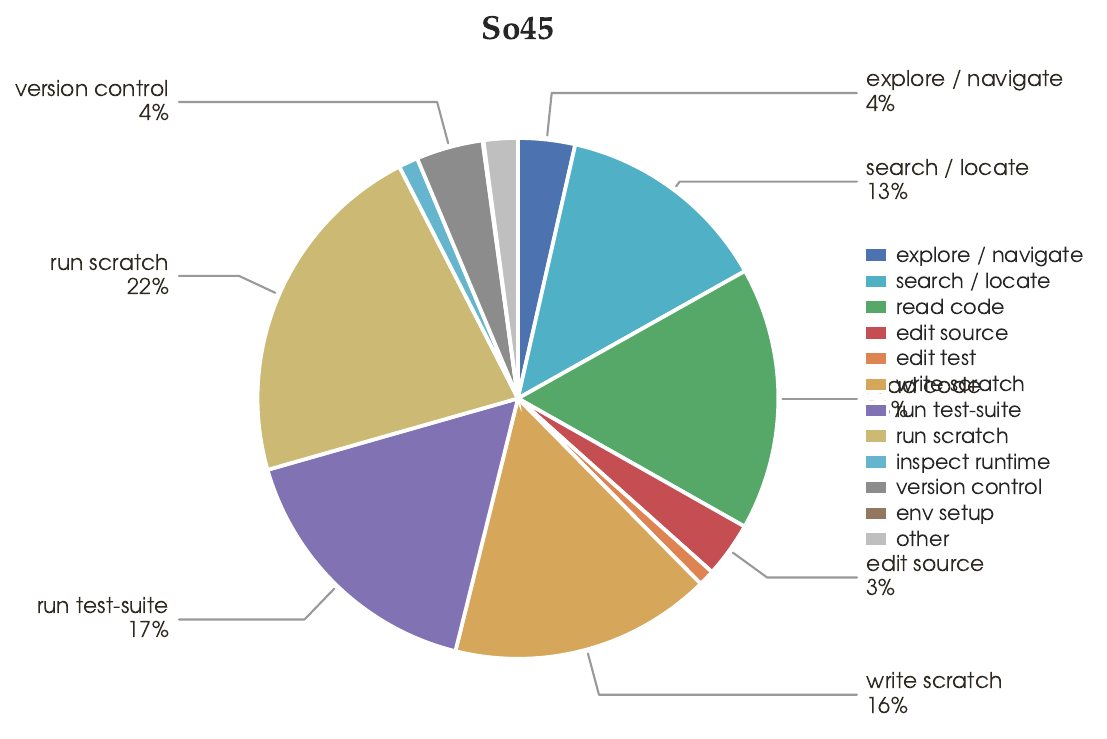} &
\includegraphics[width=0.31\linewidth]{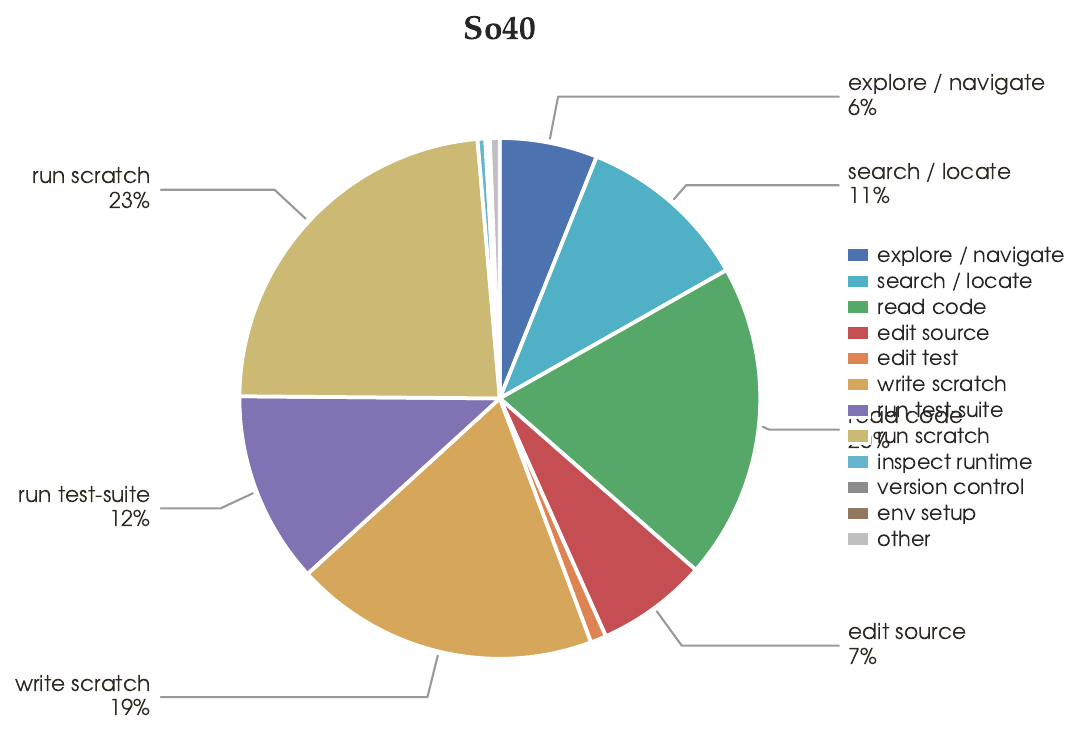} &
\includegraphics[width=0.31\linewidth]{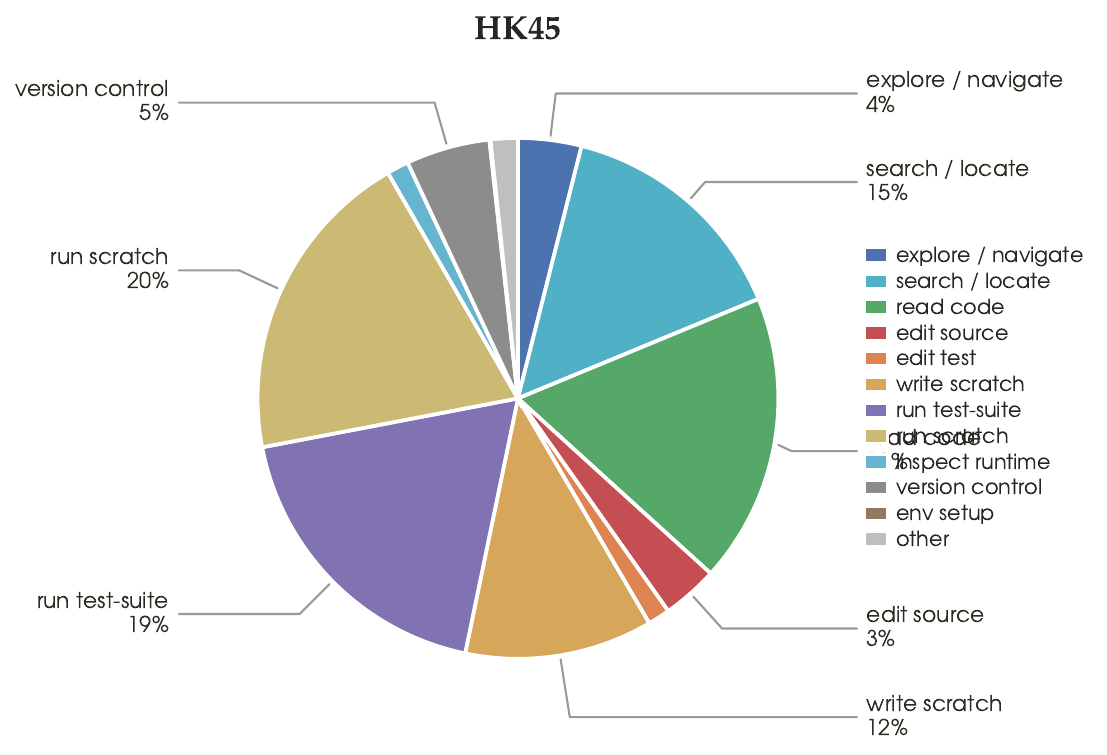} \\
\includegraphics[width=0.31\linewidth]{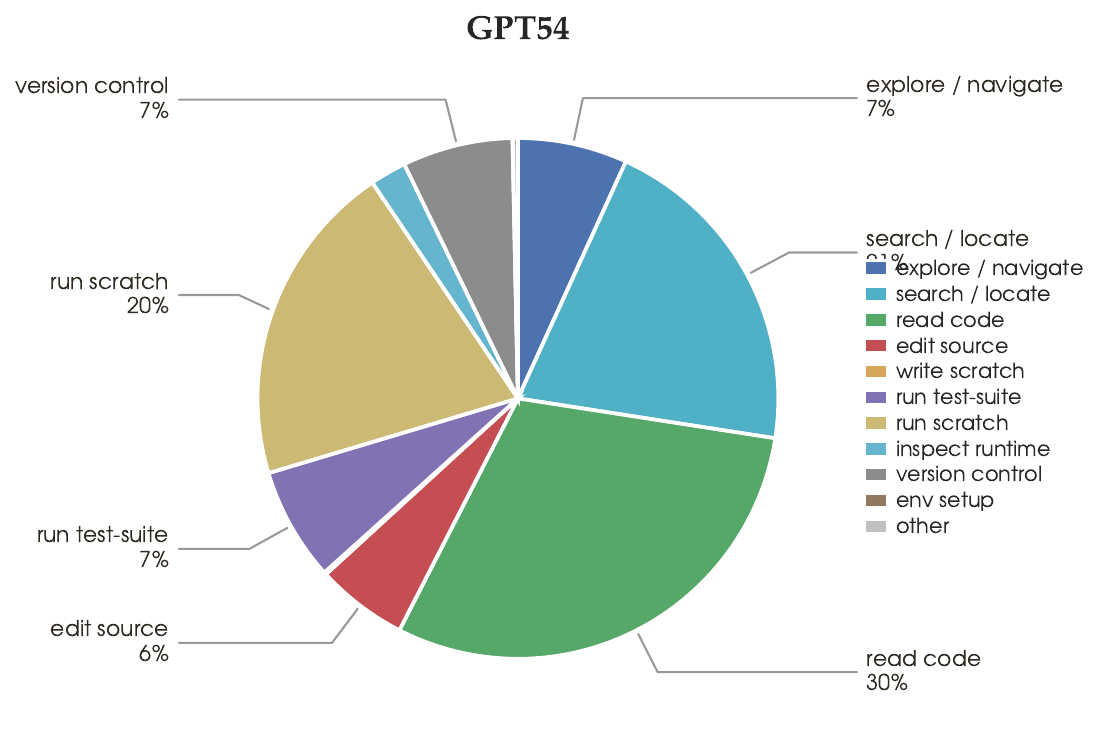} &
\includegraphics[width=0.31\linewidth]{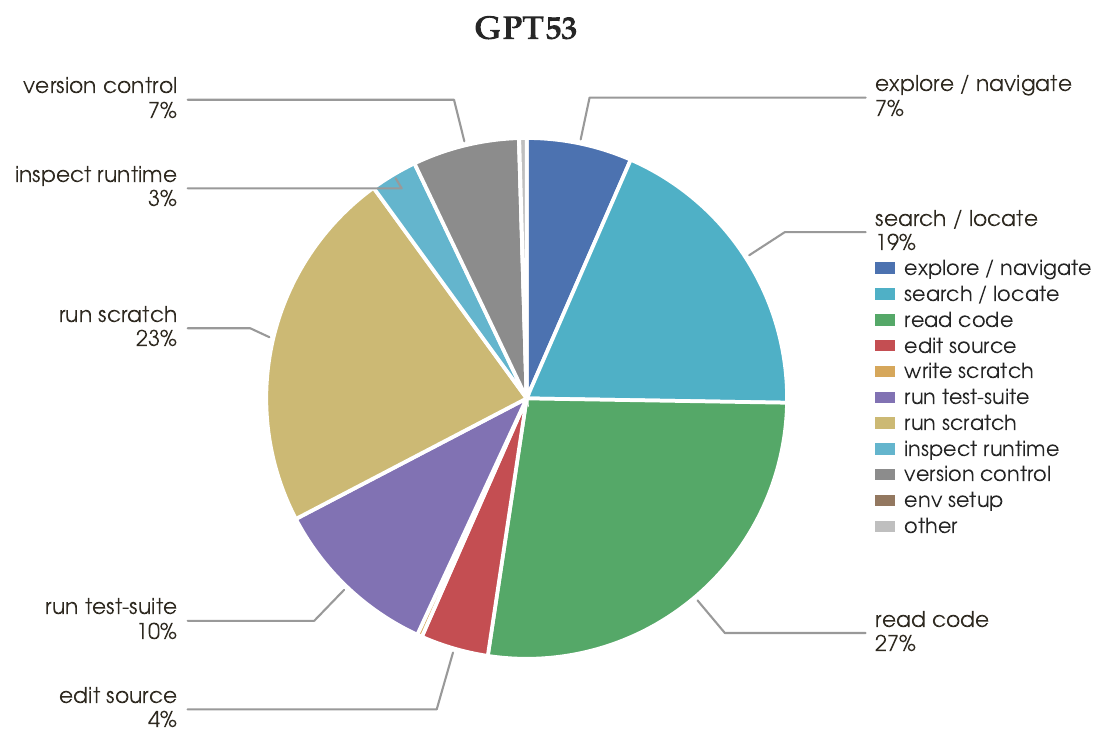} &
\includegraphics[width=0.31\linewidth]{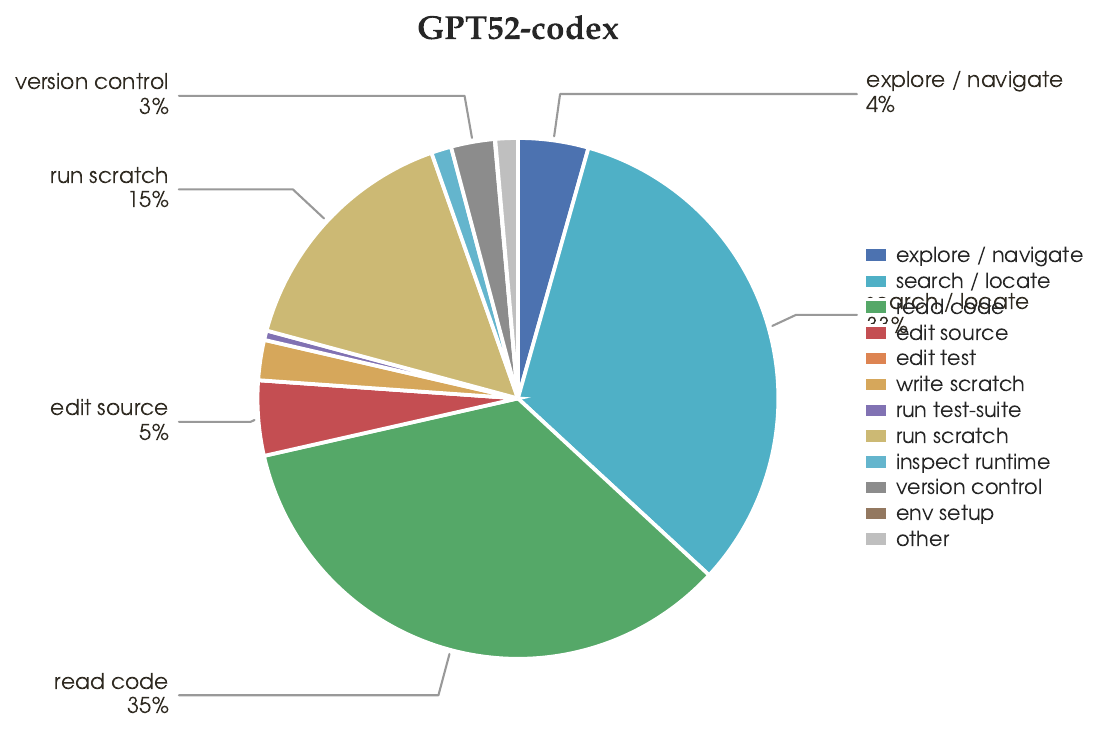} \\
\includegraphics[width=0.31\linewidth]{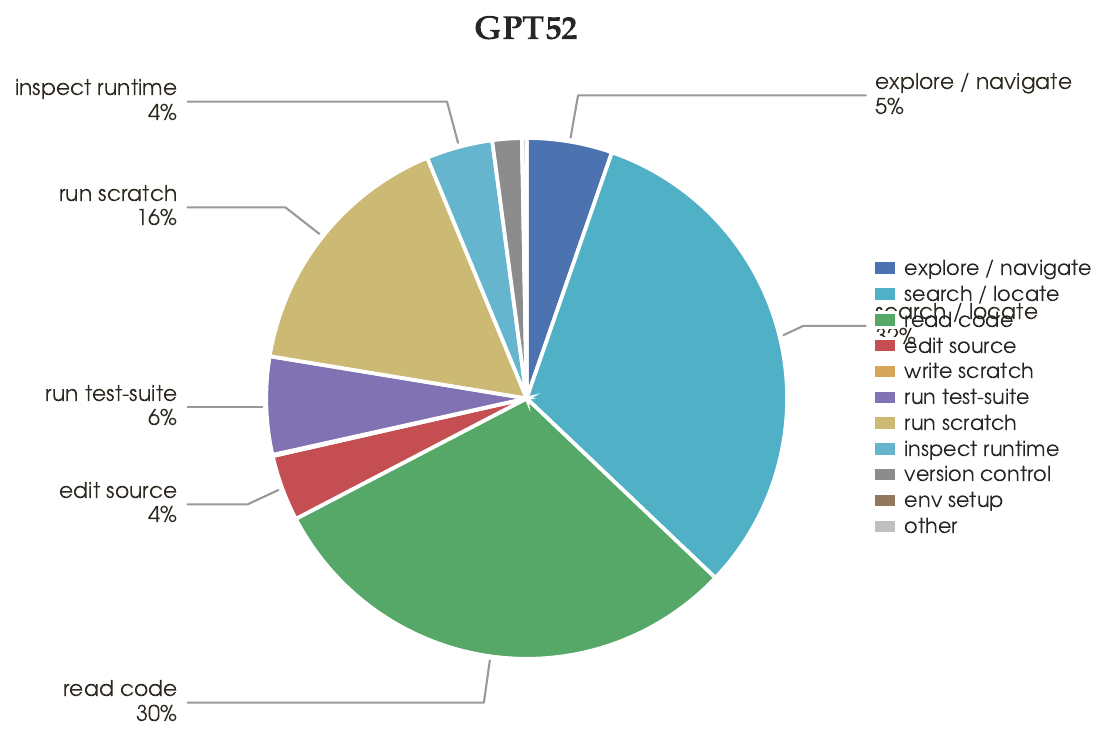} &
\includegraphics[width=0.31\linewidth]{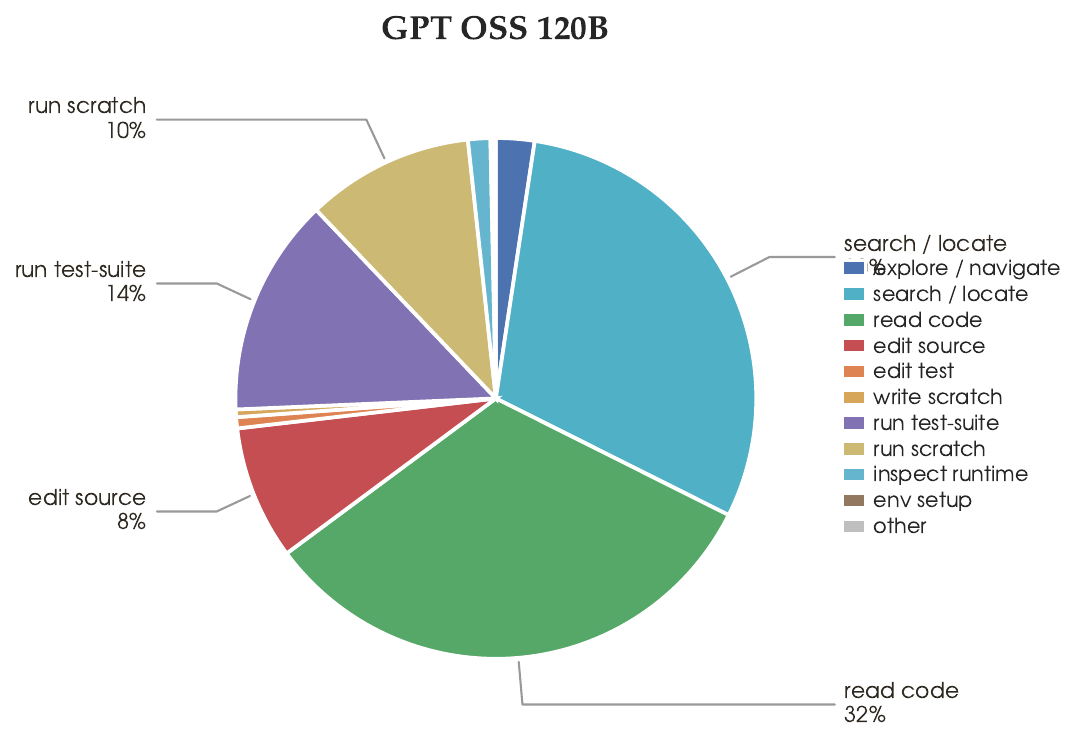} &
\includegraphics[width=0.31\linewidth]{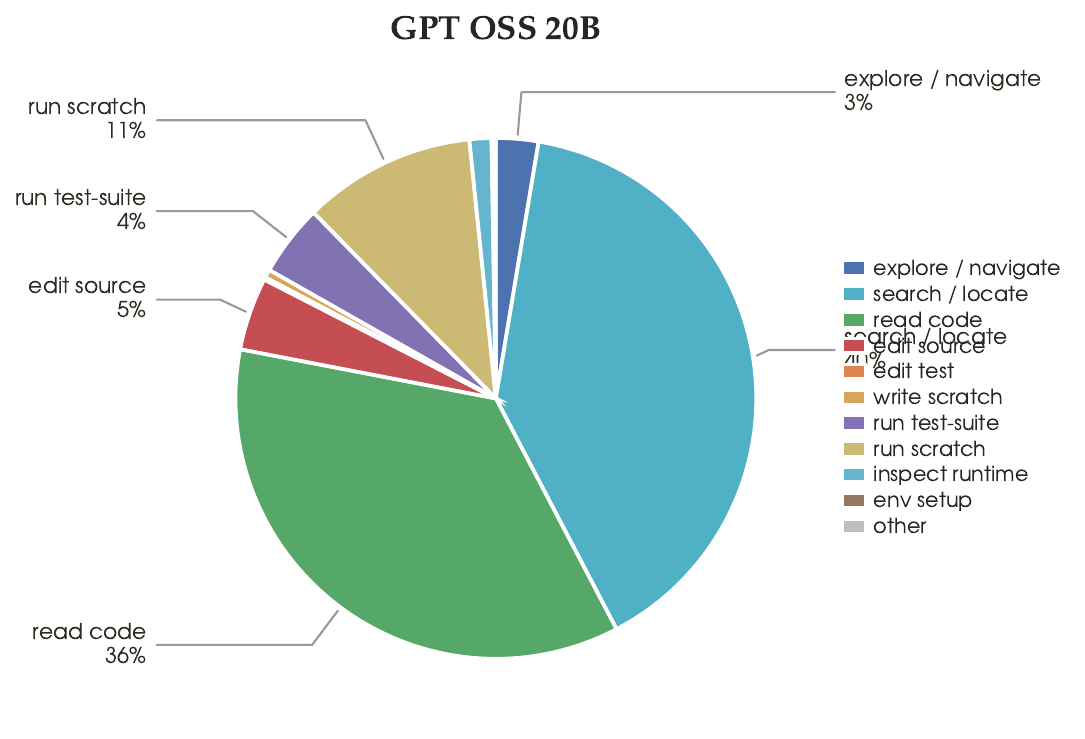} \\
\end{tabular}
\caption{Tool-call distribution per model on SWE-Bench-Verified (part 1 of 2): Anthropic Claude and OpenAI families. Each cell shows the share of calls in each R1 category (see Table\,\ref{tab:judge-buckets} for LLM judge rubrics) for one model.}
\label{fig:metrics-sbv-tooldist}
\end{figure}

\begin{figure}[p]
\centering
\includegraphics[width=0.85\linewidth]{swe_verif/tool_distribution_legend.pdf}\\[2pt]
\setlength{\tabcolsep}{1pt}
\renewcommand{\arraystretch}{0.5}
\begin{tabular}{ccc}
\includegraphics[width=0.31\linewidth]{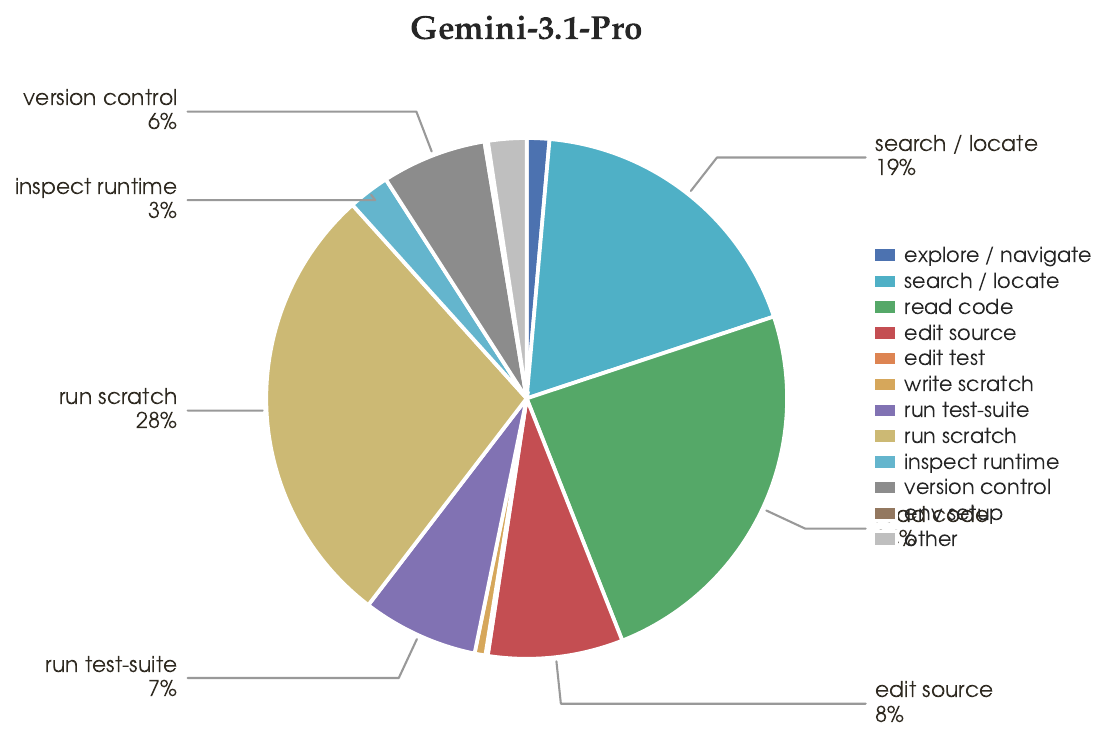} &
\includegraphics[width=0.31\linewidth]{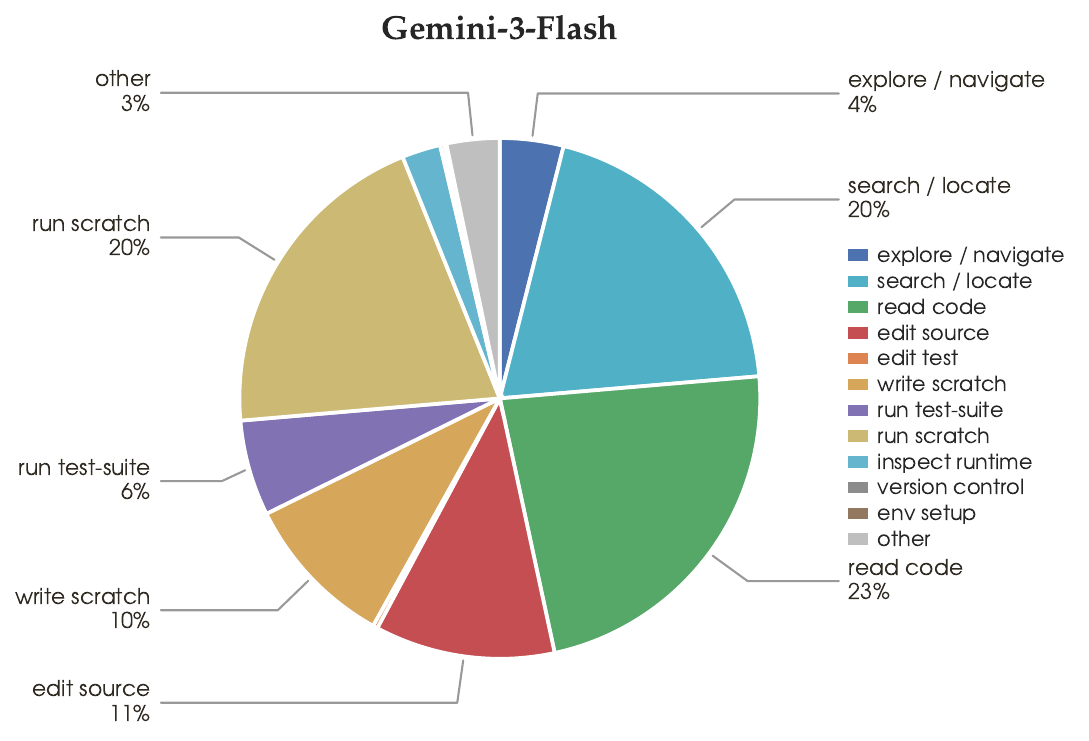} & \\
\includegraphics[width=0.31\linewidth]{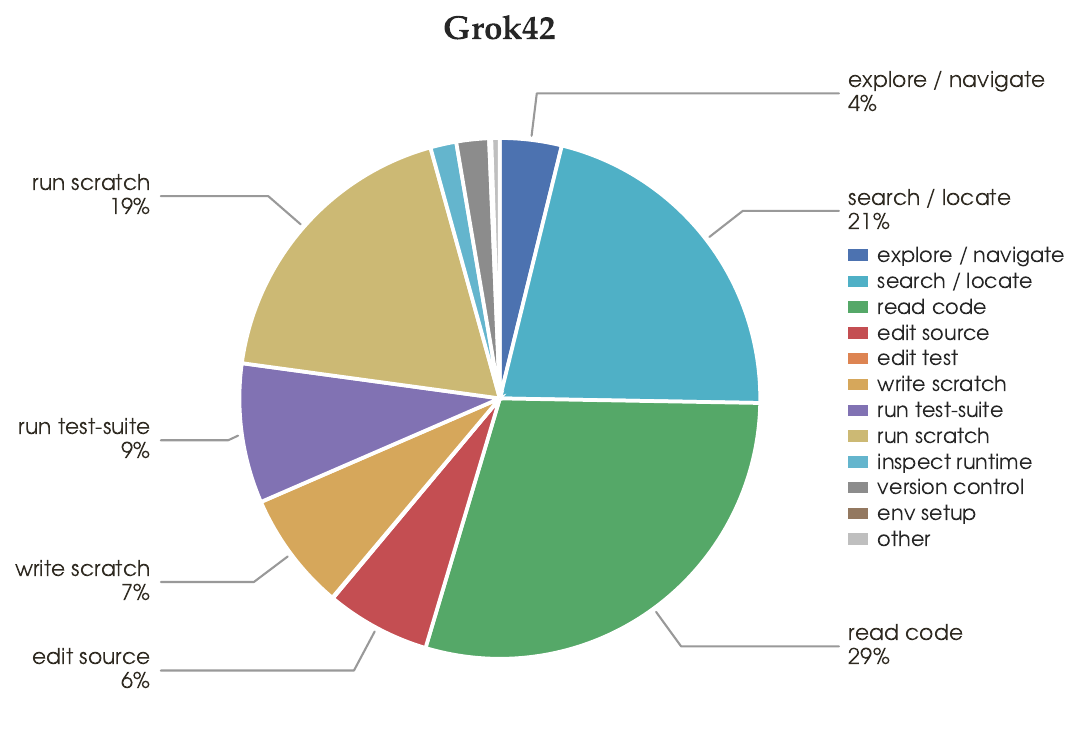} & & \\
\includegraphics[width=0.31\linewidth]{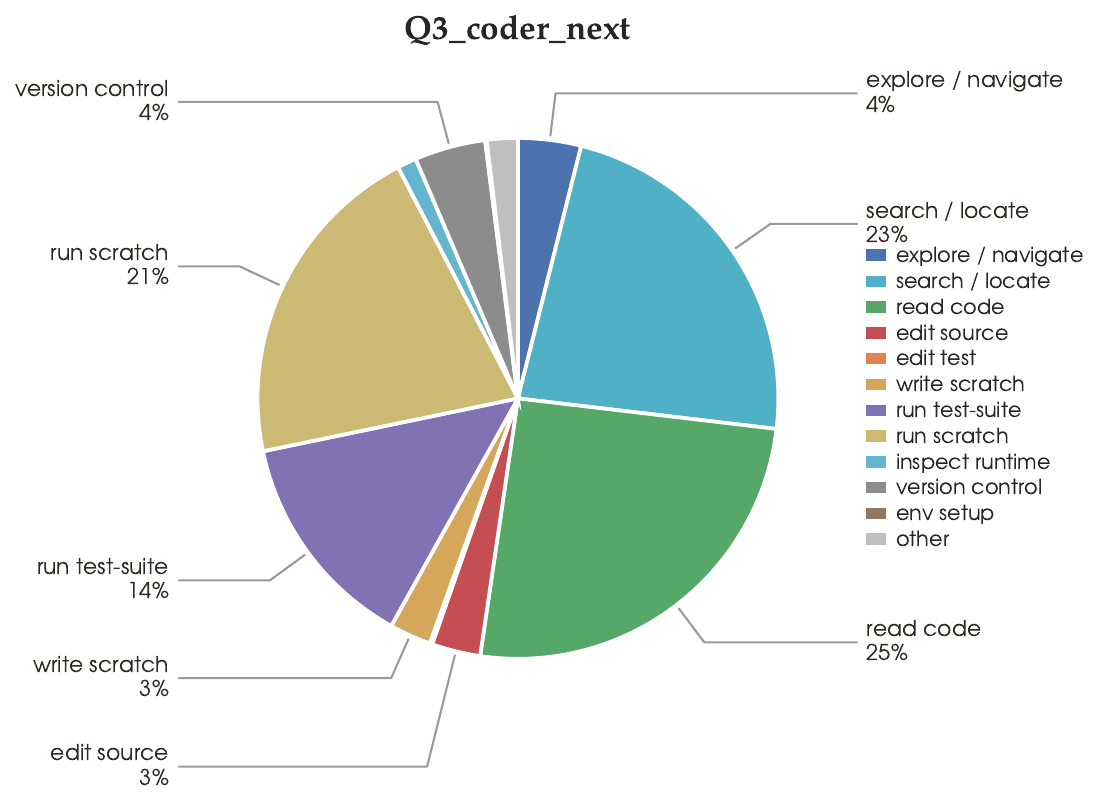} &
\includegraphics[width=0.31\linewidth]{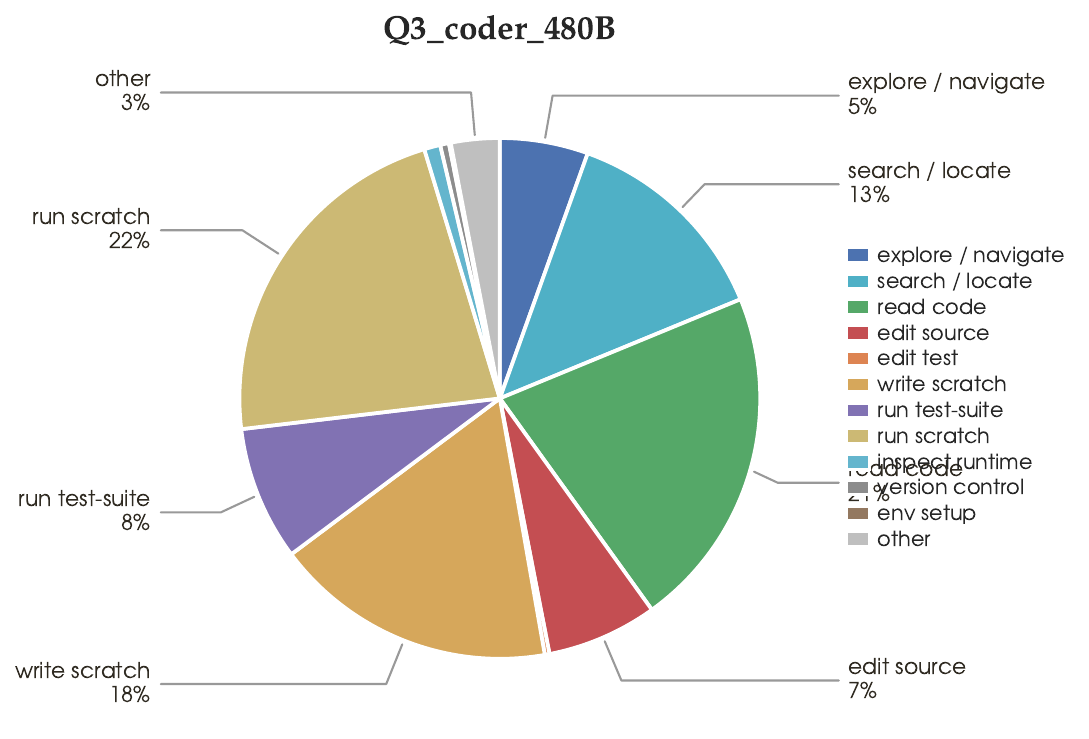} &
\includegraphics[width=0.31\linewidth]{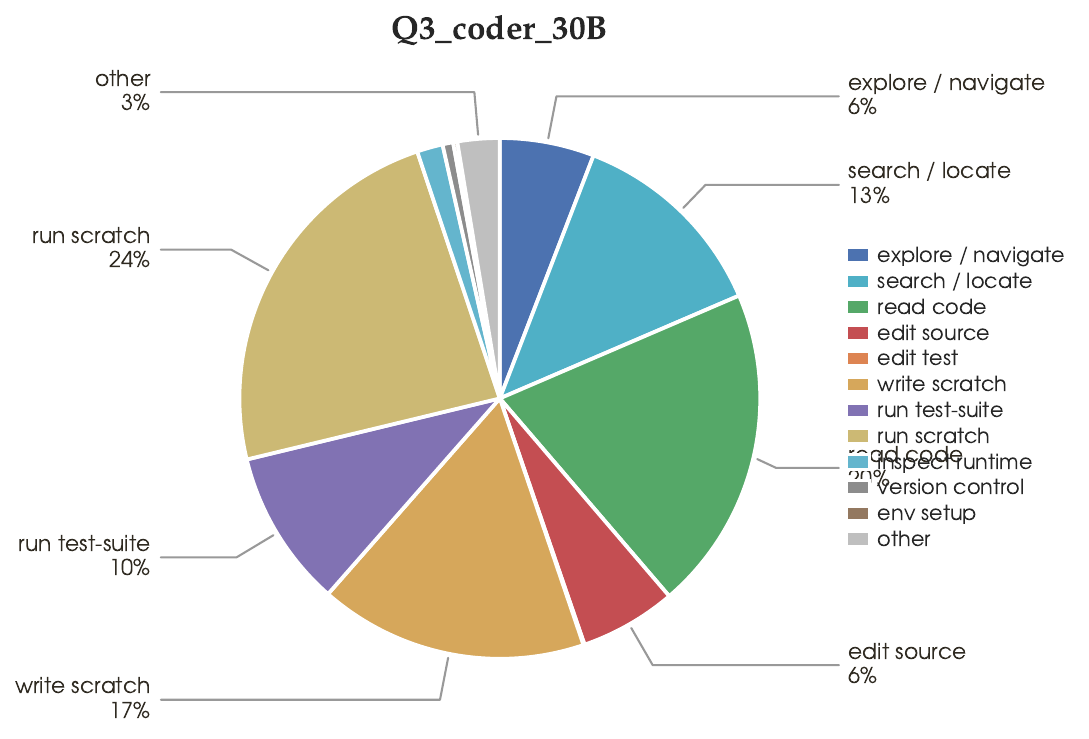} \\
\includegraphics[width=0.31\linewidth]{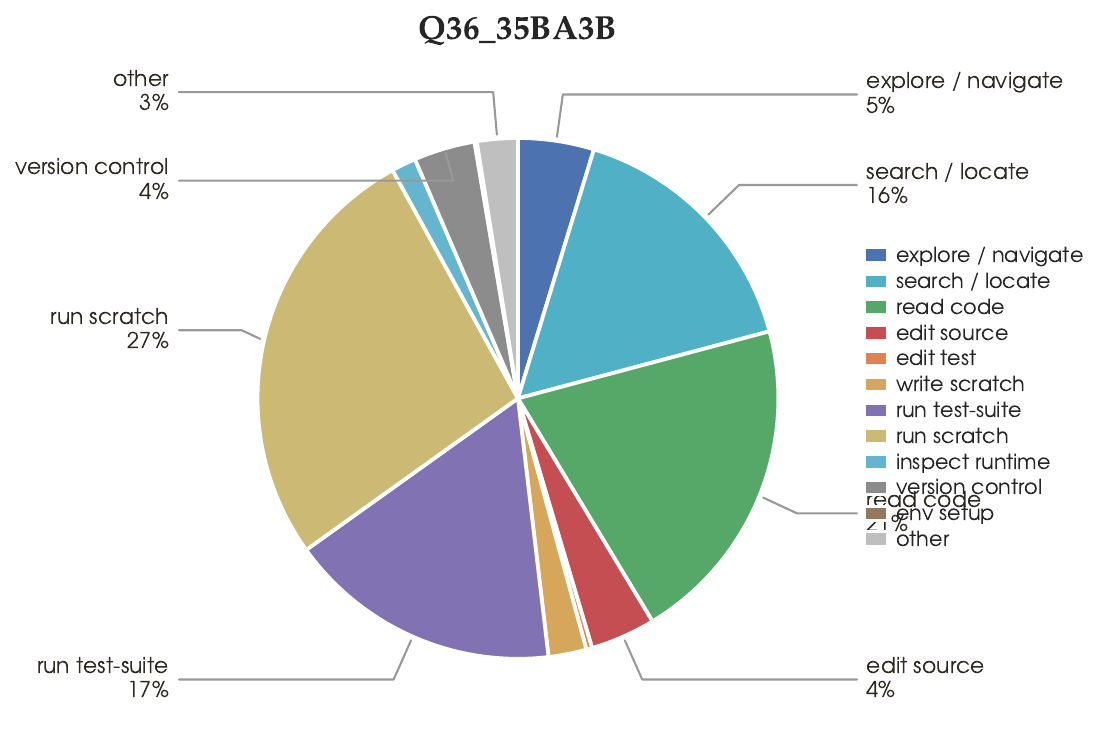} &
\includegraphics[width=0.31\linewidth]{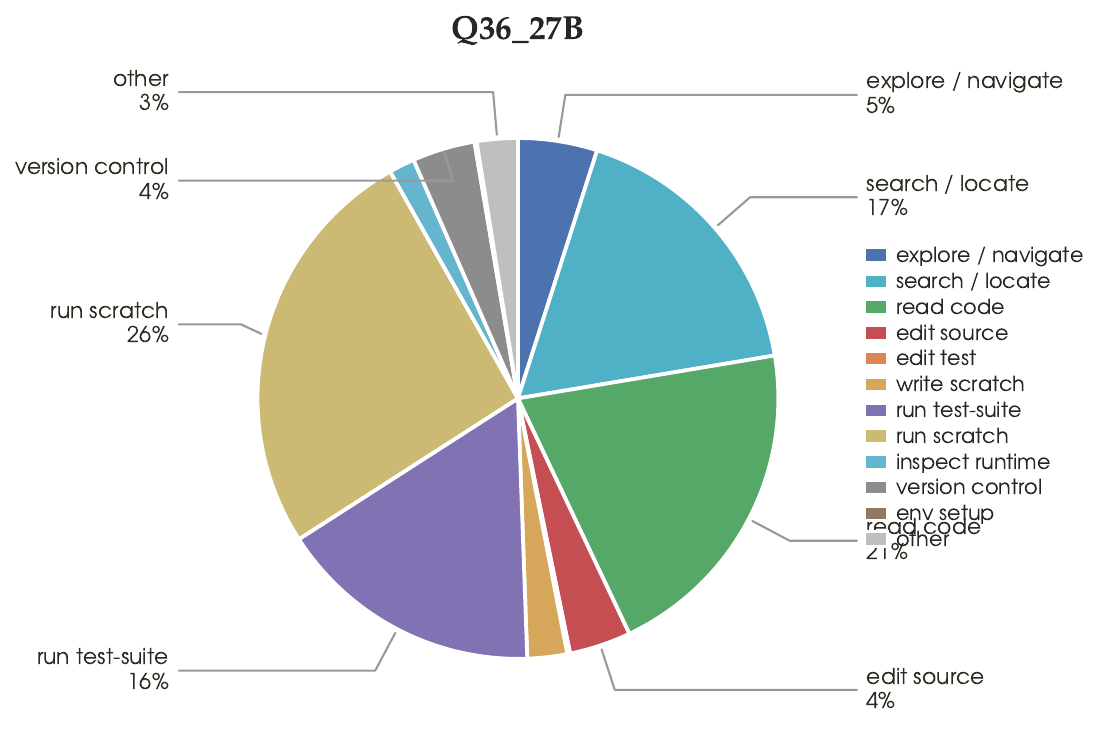} &
\includegraphics[width=0.31\linewidth]{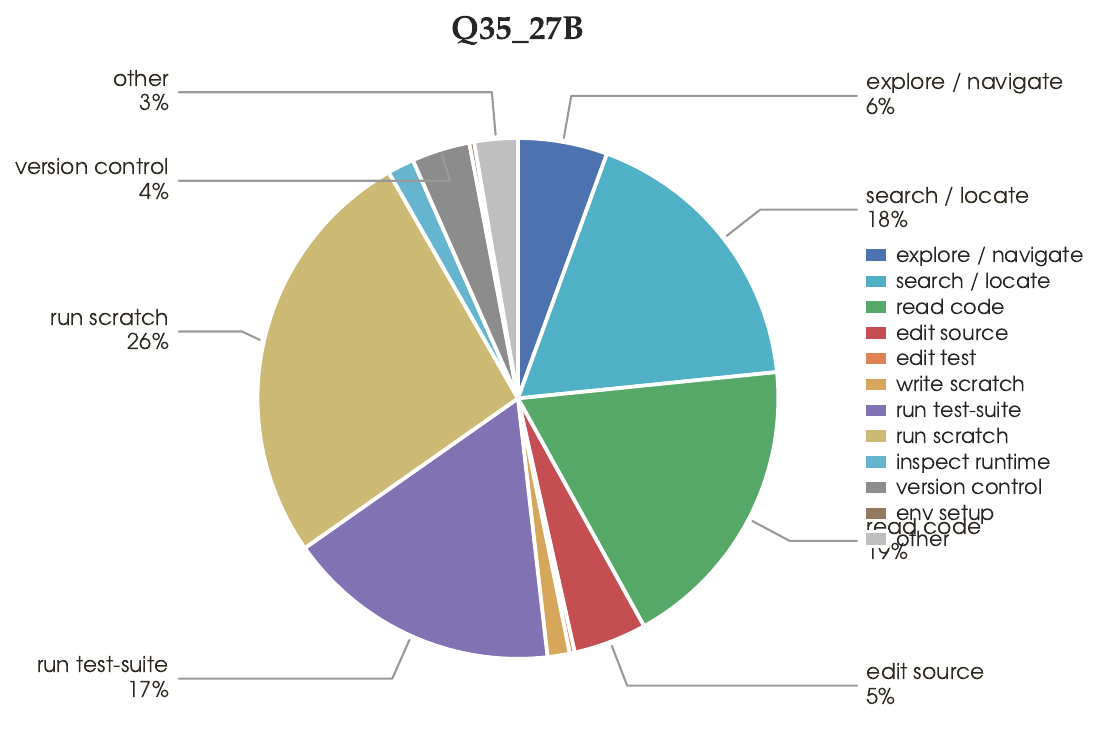} \\
\end{tabular}
\caption{Tool-call distribution per model on SWE-Bench-Verified (part 2 of 2): Gemini, Grok and Qwen families. Continued from Figure~\ref{fig:metrics-sbv-tooldist}.}
\label{fig:metrics-sbv-tooldist-b}
\end{figure}

\clearpage
\paragraph{Per-model tool-call distribution.}
Figures~\ref{fig:metrics-sbv-tooldist}--\ref{fig:metrics-sbv-tooldist-b} show how each model's tool calls split across the R1 categories (\texttt{edit\_source}, \texttt{read\_code}, \texttt{run\_suite}, \texttt{search\_locate}, \texttt{vcs}, etc.). Decisive models concentrate mass on a small number of categories (e.g. Gemini family is heavy on \texttt{edit\_source}), while exploratory models spread mass across many stages). Gemini carrying larger \texttt{edit\_source} is consistent with higher backtracking as we observed before.
The GPT and Gemini families show almost no \texttt{write\_scratch} share not because they skip
reproduction, but because they run them as a single bundled command. Specifically, they pipe the script into
\texttt{python} through an stdin heredoc, or writing and running it in one call, which the judge
scores as one \texttt{run\_scratch} action. Claude and Qwen-Coder instead author the scratch file in a
separate \texttt{write\_scratch} step before running it, so their reproduction surfaces in both buckets.

\FloatBarrier

\subsection{Metrics: SWE-Bench-Pro}
\label{app:metrics:sbp}

\begin{figure}[h]
    \centering
    \includegraphics[width=\linewidth]{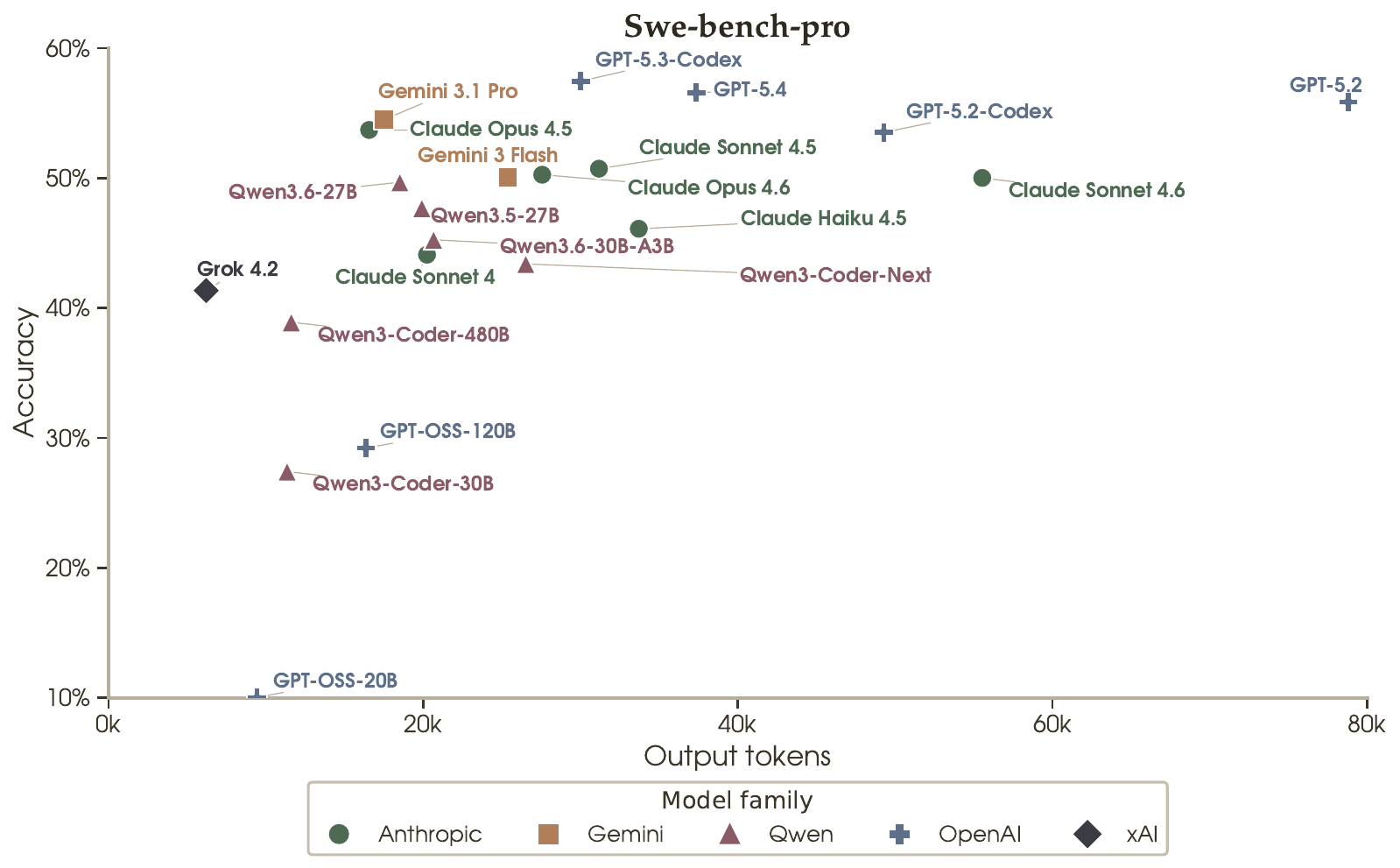}  
    \caption{SWE-Bench-Pro~\citep{deng2025swebenchpro}: total output tokens vs.\ pass@1, one point per model. Output tokens are for each problem instance (summed over all LLM calls in an agent run), averaged across 731 instances and 5 runs per instance. Pass@1 is the mean of the five per-run resolution rates. Colors group models by family. Top-left corresponds to high accuracy with the fewest output tokens spent reasoning and acting. All Pass@1 values reported here are from the Base public containers; see Appendix~\ref{app:swe-pro-git-leakage} for the Sanitized comparison that controls for git-history leakage.}
    \label{fig:metrics-sbp-pass-vs-tokens}
\end{figure}

Figure~\ref{fig:metrics-sbp-pass-vs-tokens} reports the same token-vs-pass@1 Pareto view for SWE-Bench-Pro that Figure~\ref{fig:metrics-sbv-pass-vs-tokens} reports for SWE-Bench-Verified.
The $x$-axis is total output tokens for a problem instance (summed over all LLM calls in an agent run), averaged across $731$
instances and 5 runs per instance; the $y$-axis is the mean pass@1 across those five runs. Models are colored by family.

\begin{figure}[p]
\centering
\setlength{\tabcolsep}{1pt}
\renewcommand{\arraystretch}{0.5}
\begin{tabular}{ccc}
\includegraphics[width=0.31\linewidth]{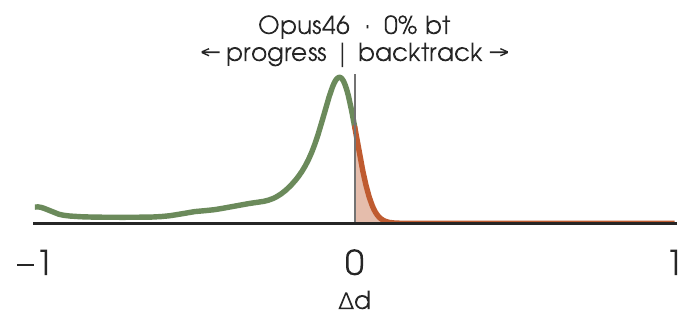} &
\includegraphics[width=0.31\linewidth]{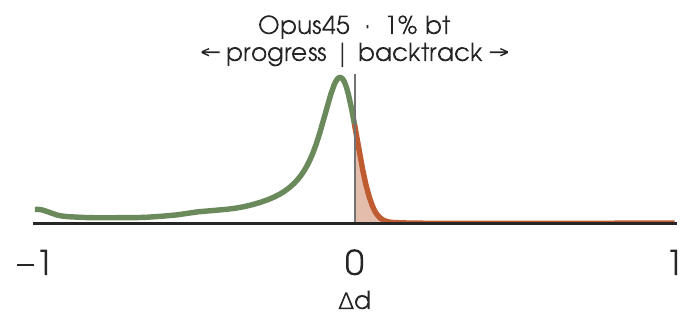} &
\includegraphics[width=0.31\linewidth]{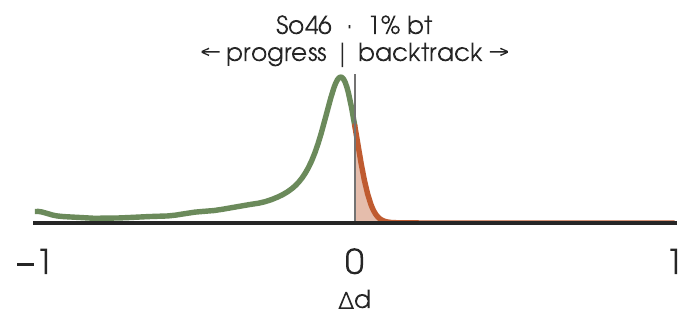} \\
\includegraphics[width=0.31\linewidth]{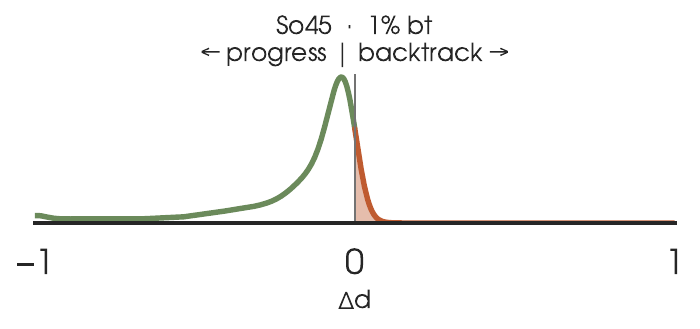} &
\includegraphics[width=0.31\linewidth]{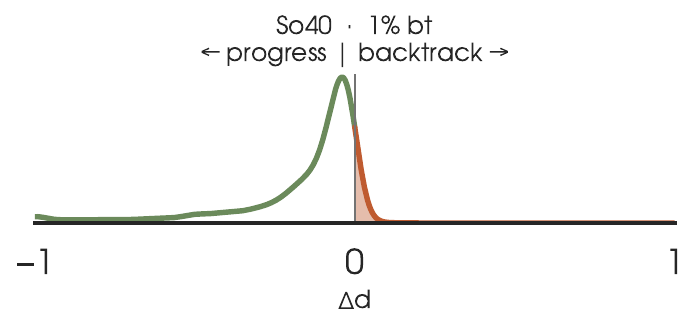} &
\includegraphics[width=0.31\linewidth]{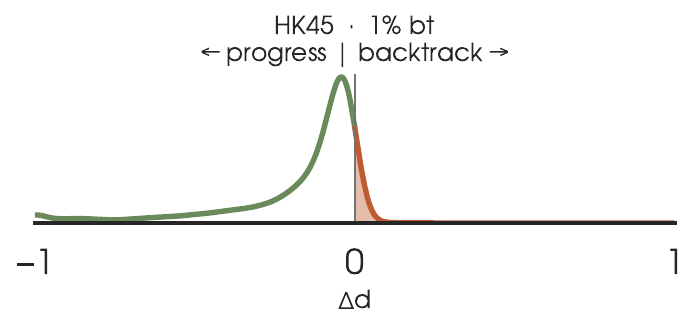} \\
\includegraphics[width=0.31\linewidth]{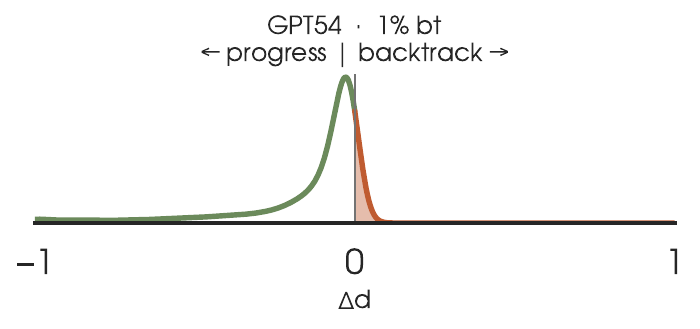} &
\includegraphics[width=0.31\linewidth]{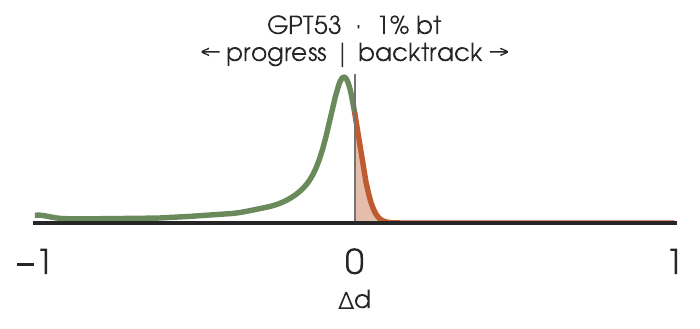} &
\includegraphics[width=0.31\linewidth]{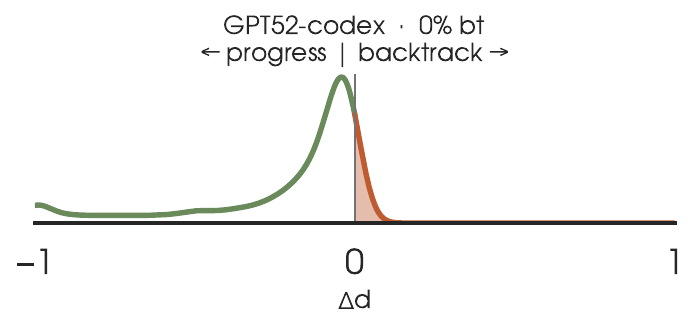} \\
\includegraphics[width=0.31\linewidth]{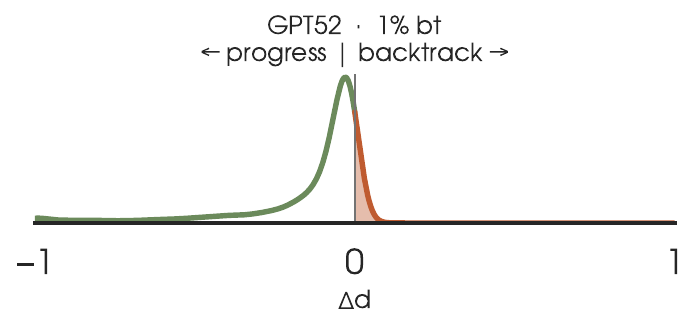} &
\includegraphics[width=0.31\linewidth]{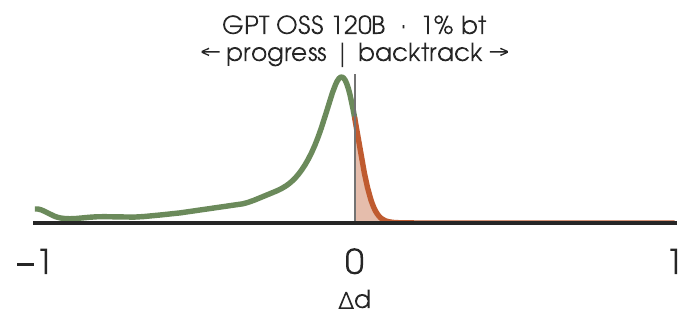} &
\includegraphics[width=0.31\linewidth]{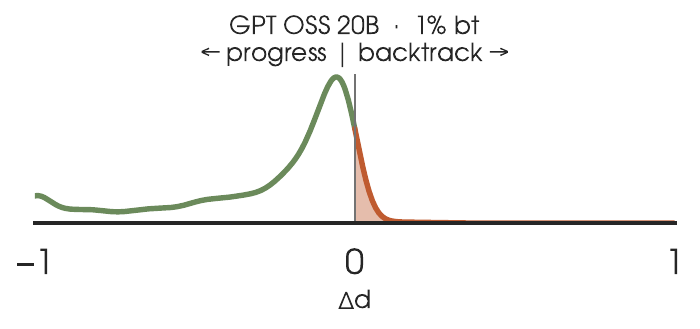} \\
\end{tabular}
\caption{\textbf{Local backtracking.} Histograms of $\Delta D$ on SWE-Bench-Pro (part 1 of 2): Anthropic Claude and OpenAI families. Each cell is one model; green = progress edits, red = local backtracking edits. Annotation shows the per-model local-backtrack share.}
\label{fig:metrics-sbp-backtracking}
\end{figure}

\begin{figure}[p]
\centering
\setlength{\tabcolsep}{1pt}
\renewcommand{\arraystretch}{0.5}
\begin{tabular}{ccc}
\includegraphics[width=0.31\linewidth]{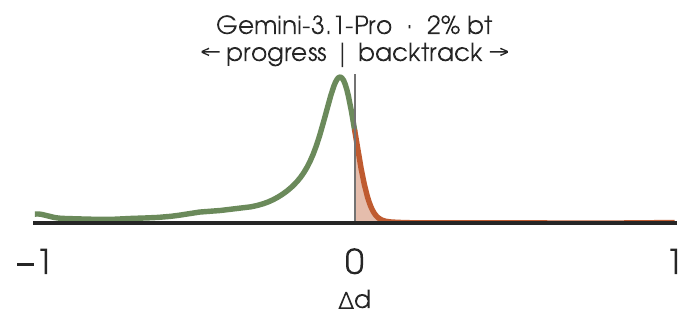} &
\includegraphics[width=0.31\linewidth]{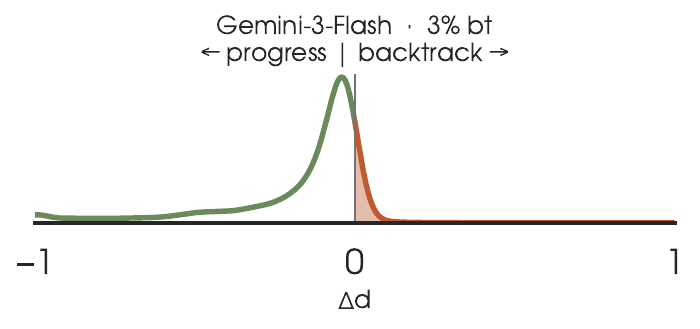} & \\
\includegraphics[width=0.31\linewidth]{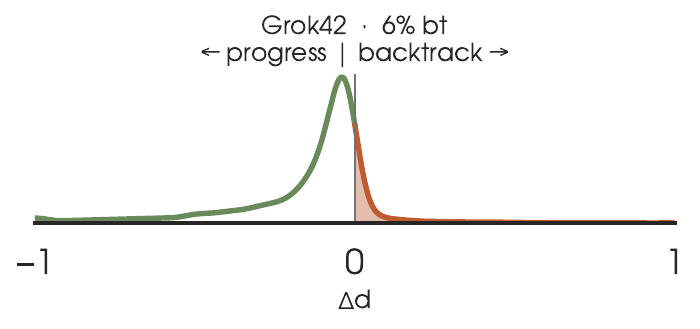} & & \\
\includegraphics[width=0.31\linewidth]{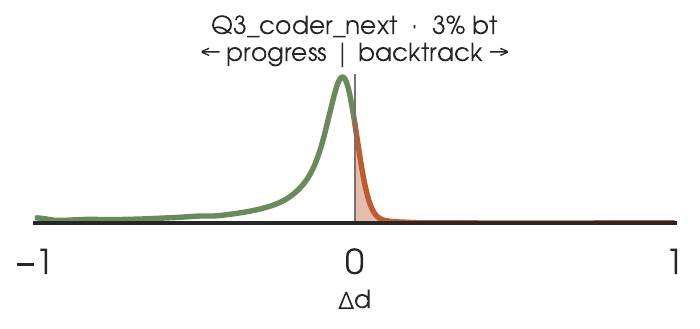} &
\includegraphics[width=0.31\linewidth]{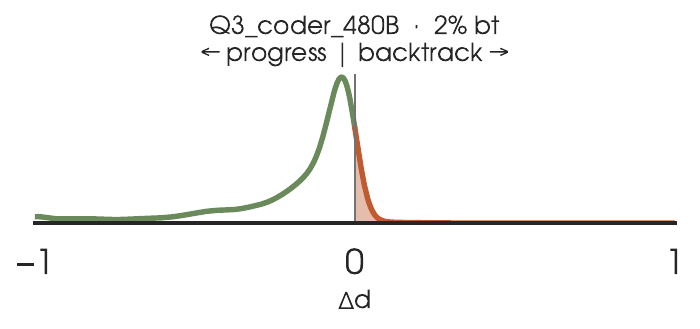} &
\includegraphics[width=0.31\linewidth]{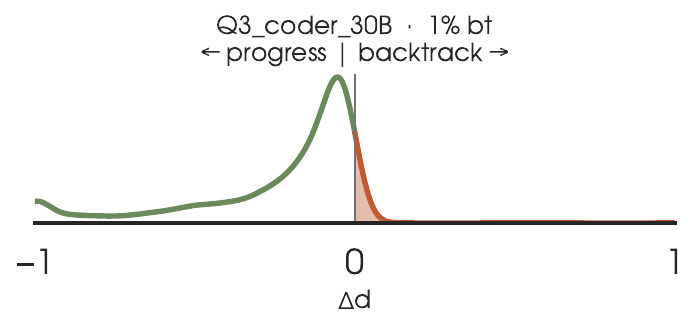} \\
\includegraphics[width=0.31\linewidth]{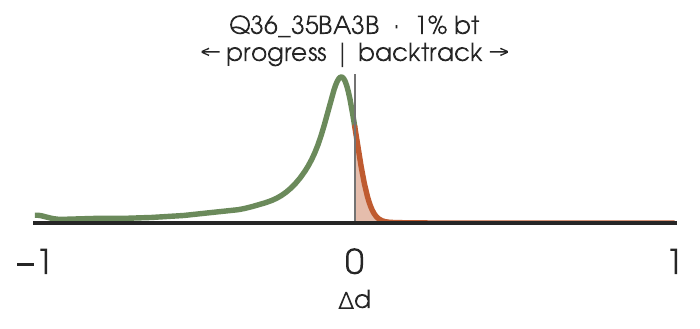} &
\includegraphics[width=0.31\linewidth]{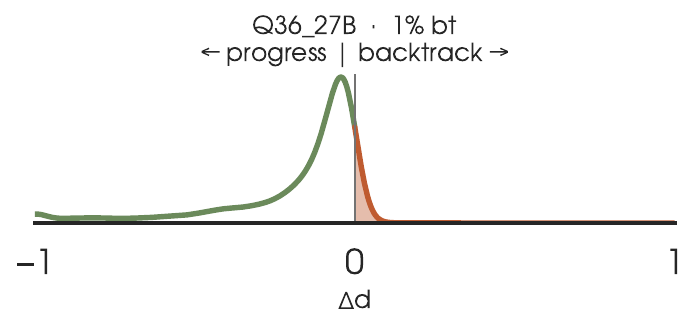} &
\includegraphics[width=0.31\linewidth]{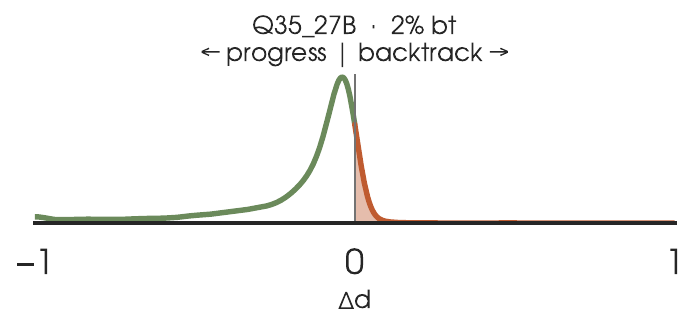} \\
\end{tabular}
\caption{\textbf{Local backtracking.} Histograms of $\Delta D$ on SWE-Bench-Pro (part 2 of 2): Gemini, Grok and Qwen families. Continued from Figure~\ref{fig:metrics-sbp-backtracking}.}
\label{fig:metrics-sbp-backtracking-b}
\end{figure}

Table~\ref{tab:metrics-sbp-distance-backtrack-summary} gives the same compact distance-and-local-backtracking summary on SBP that Table~\ref{tab:metrics-sbv-distance-backtrack-summary} reports on SBV. Two observations stand out compared to SWE-Bench-Verified: \textbf{backtracking reduces across the board on SWE-Bench-Pro} --- the entire field sits under $6\%$, from $0.3\%$ in GPT-5.2-codex to $5.9\%$ in Grok 4.2. The Gemini has a much lower backtracking rate here compared to SWE-Bench-Verified (3.1 Pro goes from $14.2 \to 1.7\%$; Flash from $16.9 \to 2.7\%$); Grok 4.2 has the highest backtracking rate. \textbf{Arrival-time is higher on average compared to SBV} ($t @ D{\le}0.1$ is $0.53$--$0.95$ on SBP vs.\ $0.37$--$0.95$ on SBV), consistent with SBP's larger and longer-horizon repository changes. Note that, since SBP tasks require a large set of changes to the source code, and are more involved  (spread across more files), compared to atomic changes in SBV, large backtracking steps are less likely as models usually make edits one file at a time. We can also see this distinction in backtracking density Figures\,\ref{fig:metrics-sbv-backtracking}-\ref{fig:metrics-sbv-backtracking-b} where SBV carries significant mass for $\Delta D\sim1$ which is visibly absent in SBP Figures\,\ref{fig:metrics-sbp-backtracking}-\ref{fig:metrics-sbp-backtracking-b}

\begin{table}[h]
\centering
\small
\caption{Distance, local backtracking and outcome summary over resolved SWE-Bench-Pro trajectories per model. Local backtrack \%, arrival-time $t$ @ $D\!\le\!0.1$ and area-under-curve (AUC) are computed from the faithful state-replay curves over $731$-instance (5 runs each) per model. Pass@1 is the SSA published value (Base containers; matches Appendix~\ref{app:passrates}).}
\label{tab:metrics-sbp-distance-backtrack-summary}
\begin{tabular}{@{}lrrrr@{}}
\toprule
Model & Local backtrack \% & $t$ @ $D\!\le\!0.1$ & AUC & pass@1 \% \\
\midrule
Opus 4.8           &  0.8 & 0.63 & 0.44 & 68.48 \\
Opus 4.6           &  0.4 & 0.74 & 0.55 & 57.45 \\
Opus 4.5           &  1.1 & 0.68 & 0.50 & 54.58 \\
Sonnet 4.6         &  0.6 & 0.84 & 0.68 & 53.67 \\
Sonnet 4.5         &  0.9 & 0.53 & 0.36 & 50.15 \\
Sonnet 4.0         &  1.0 & 0.68 & 0.44 & 44.26 \\
Haiku 4.5          &  1.0 & 0.63 & 0.42 & 46.37 \\
\midrule
GPT-5.5            &  1.6 & 0.74 & 0.46 & 59.42 \\
GPT-5.4            &  0.5 & 0.84 & 0.61 & 59.90 \\
GPT-5.3            &  0.6 & 0.84 & 0.57 & 58.57 \\
GPT-5.2            &  0.8 & 0.89 & 0.63 & 56.47 \\
GPT-5.2-codex      &  0.3 & 0.95 & 0.71 & 53.87 \\
GPT-OSS 120B       &  1.1 & 0.95 & 0.57 & 43.61 \\
GPT-OSS 20B        &  0.7 & 0.95 & 0.63 & 39.67 \\
\midrule
Gemini 3.1 Pro     &  1.7 & 0.89 & 0.51 & 54.00 \\
Gemini 3 Flash     &  2.7 & 0.79 & 0.46 & 49.43 \\
\midrule
Grok 4.2           &  5.9 & 0.95 & 0.70 & 51.57 \\
\midrule
Qwen3-Coder Next   &  2.7 & 0.74 & 0.47 & 57.75 \\
Qwen3-Coder 480B   &  1.8 & 0.68 & 0.45 & 38.90 \\
Qwen3-Coder 30B    &  0.7 & 0.68 & 0.48 & 27.41 \\
Qwen3.6 35B-A3B    &  1.3 & 0.63 & 0.45 & 46.70 \\
Qwen3.6 27B        &  0.9 & 0.68 & 0.48 & 51.40 \\
Qwen3.5 27B        &  2.0 & 0.68 & 0.46 & 47.44 \\
\bottomrule
\end{tabular}
\end{table}

\begin{figure}[p]
\centering
\includegraphics[width=0.75\linewidth]{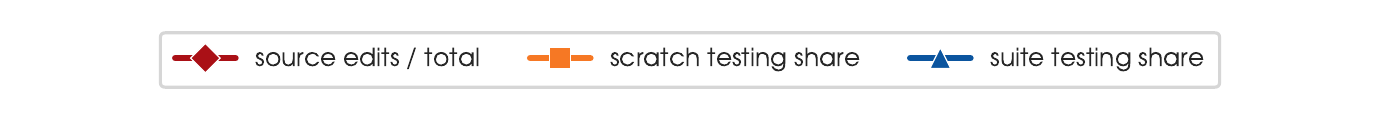}\\[2pt]
\setlength{\tabcolsep}{1pt}
\renewcommand{\arraystretch}{0.5}
\begin{tabular}{ccc}
\includegraphics[width=0.31\linewidth]{swe_pro/edit_read_ratio/opus46.pdf} &
\includegraphics[width=0.31\linewidth]{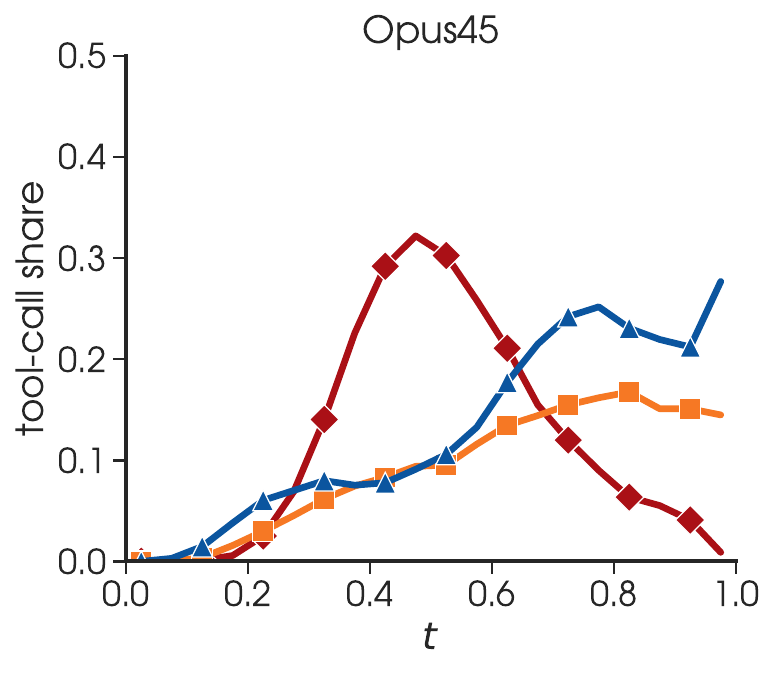} &
\includegraphics[width=0.31\linewidth]{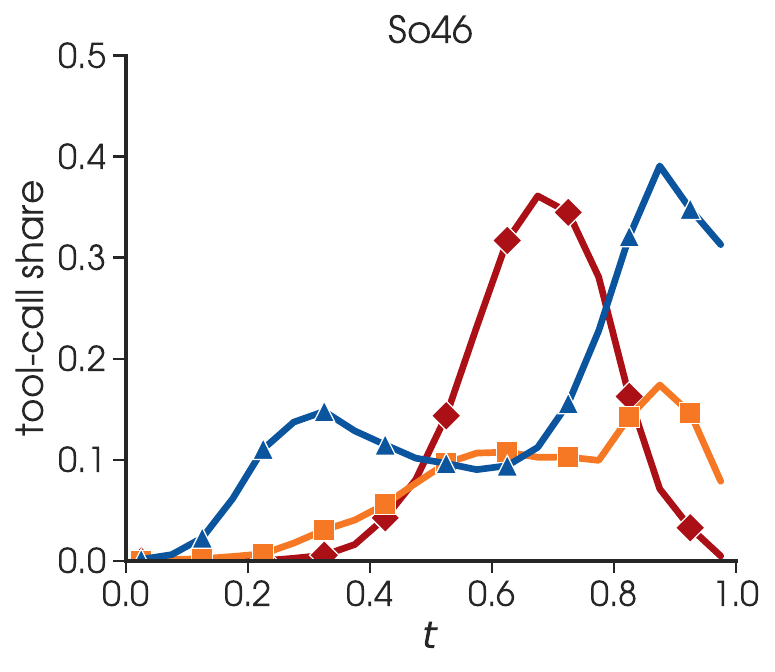} \\
\includegraphics[width=0.31\linewidth]{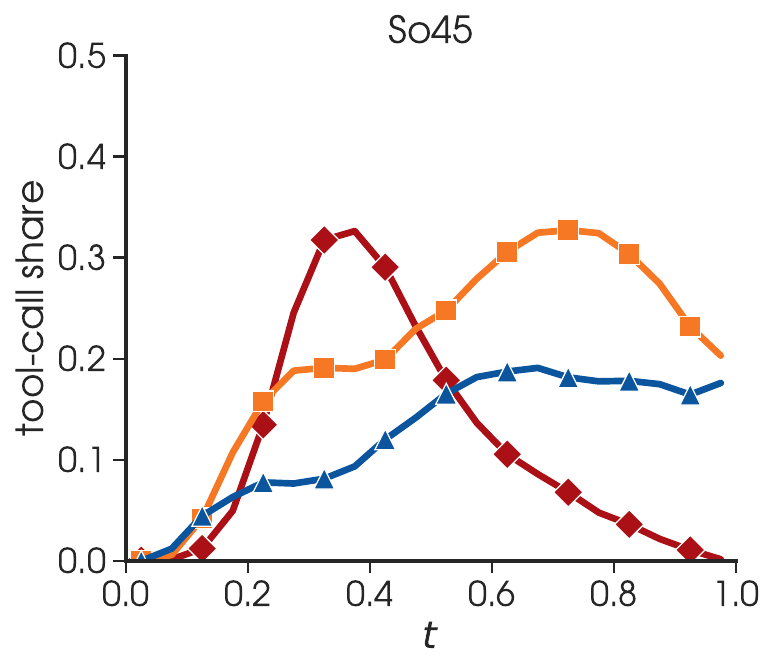} &
\includegraphics[width=0.31\linewidth]{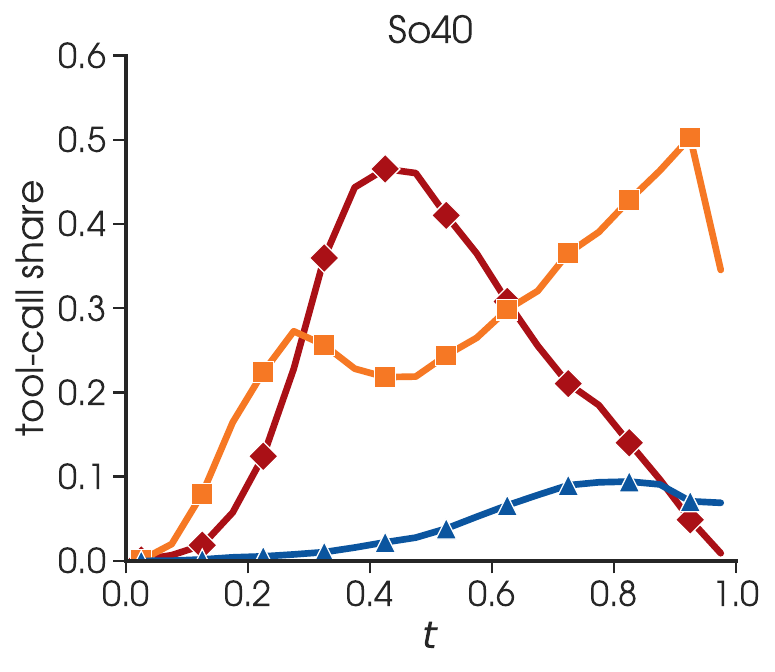} &
\includegraphics[width=0.31\linewidth]{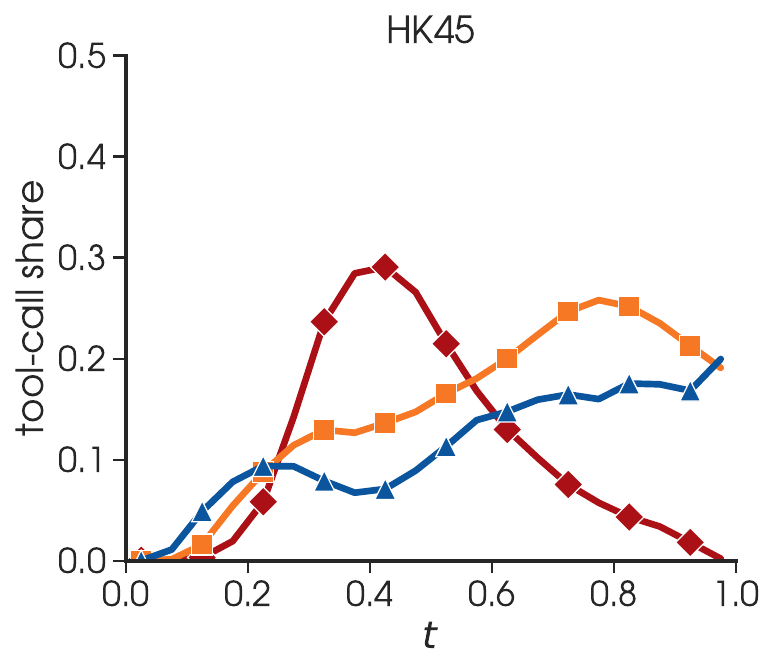} \\
\includegraphics[width=0.31\linewidth]{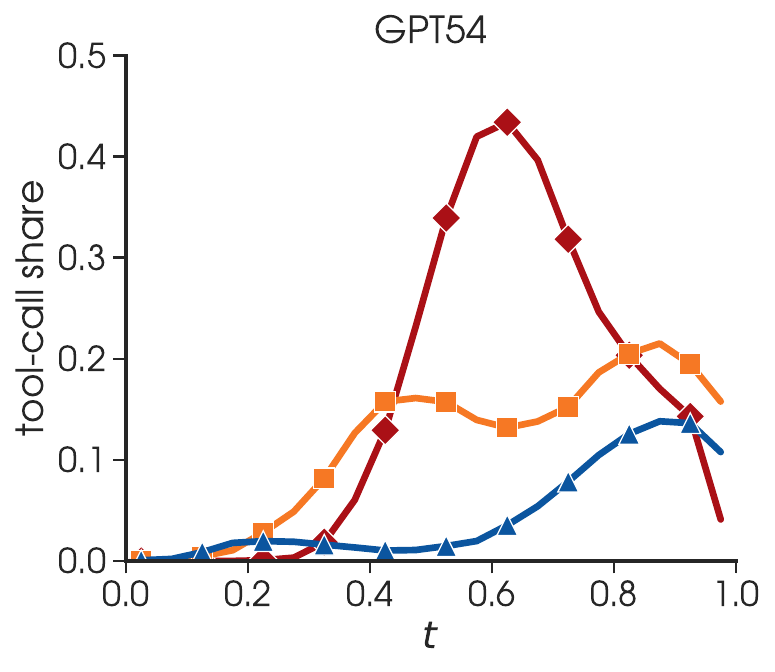} &
\includegraphics[width=0.31\linewidth]{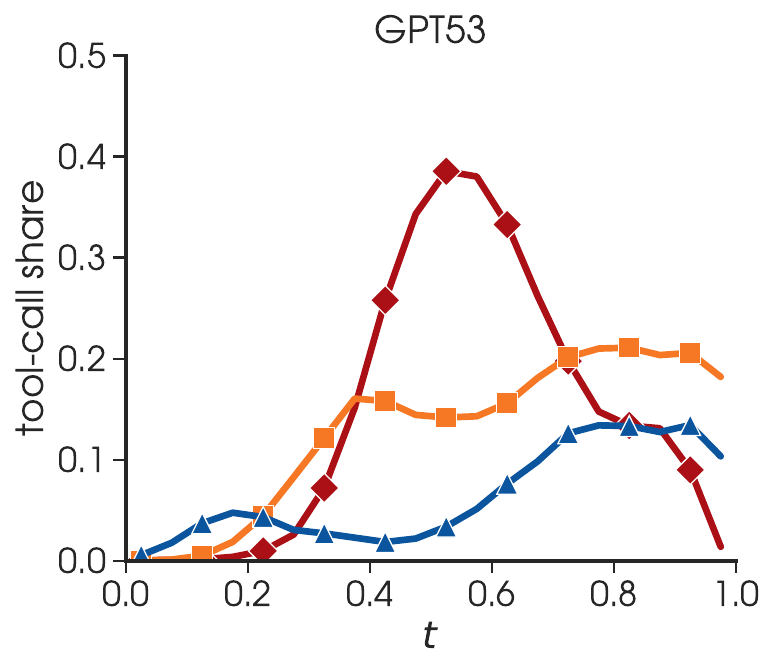} &
\includegraphics[width=0.31\linewidth]{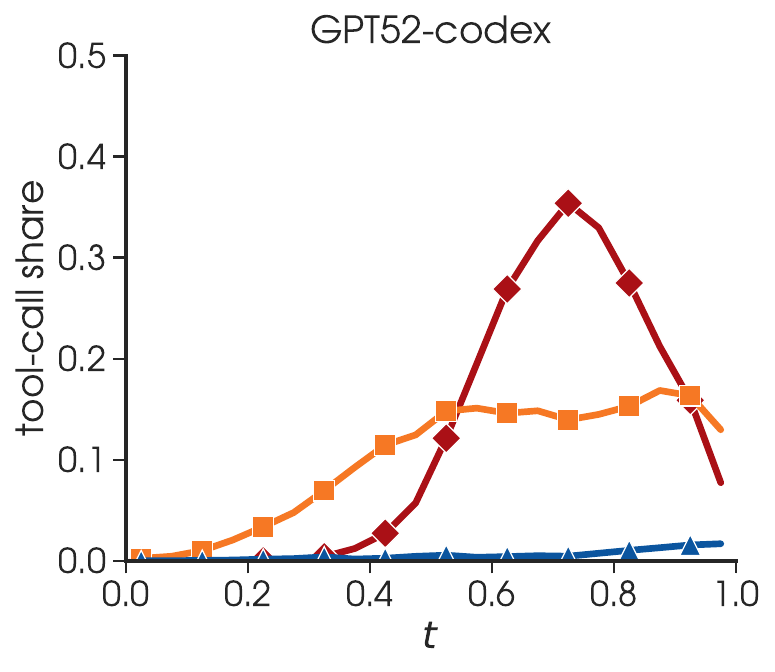} \\
\includegraphics[width=0.31\linewidth]{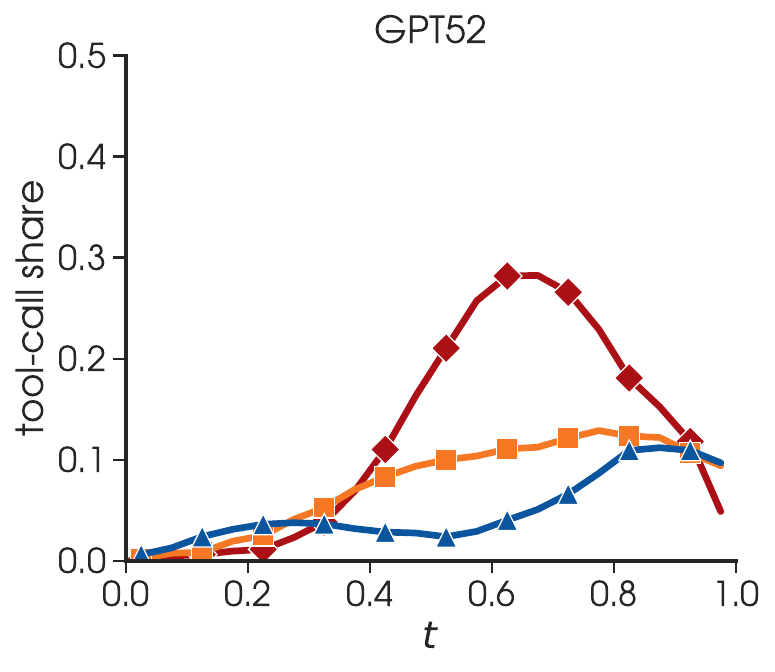} &
\includegraphics[width=0.31\linewidth]{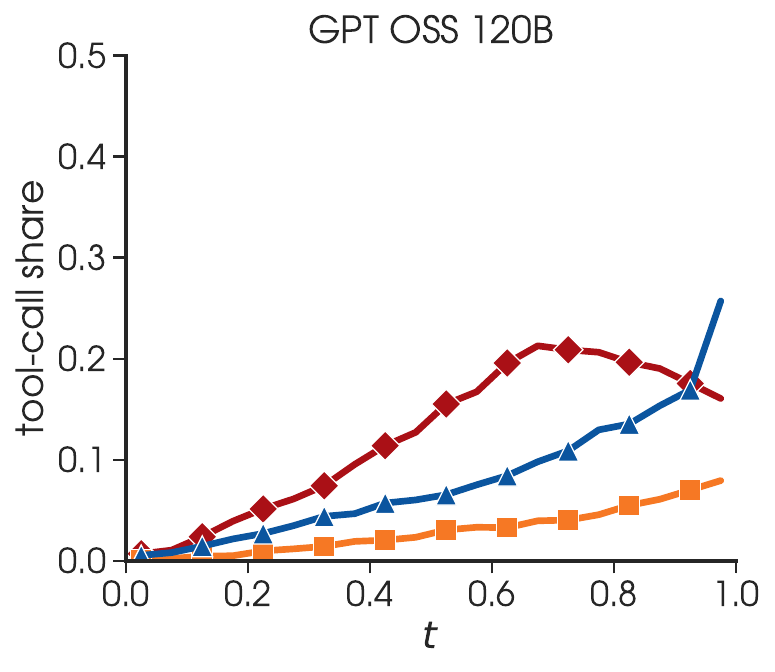} &
\includegraphics[width=0.31\linewidth]{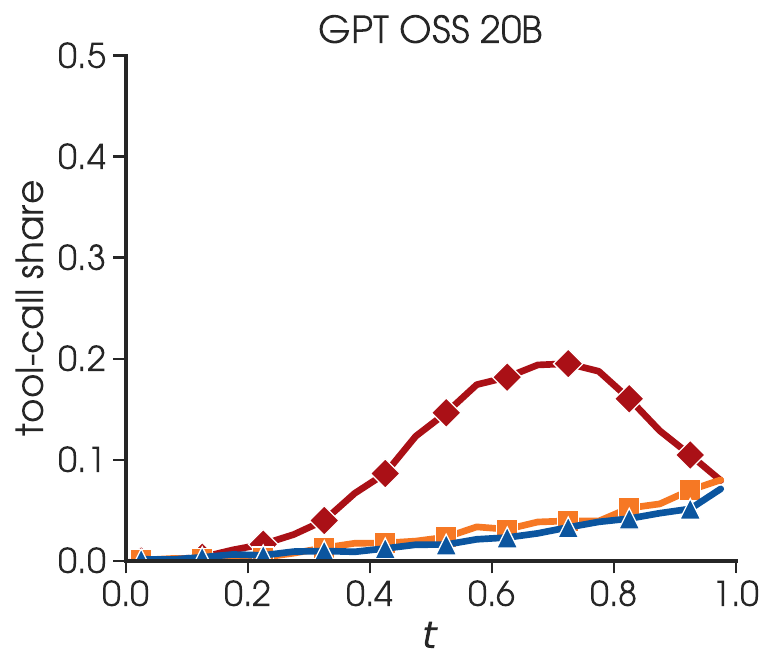} \\
\end{tabular}
\caption{Edit/test ratio per cycle on SWE-Bench-Pro (part 1 of 2): Anthropic Claude and OpenAI families. Each subplot shows the model's distribution of ratios.}
\label{fig:metrics-sbp-editread}
\end{figure}

\begin{figure}[p]
\centering
\includegraphics[width=0.75\linewidth]{swe_pro/edit_read_ratio_legend.pdf}\\[2pt]
\setlength{\tabcolsep}{1pt}
\renewcommand{\arraystretch}{0.5}
\begin{tabular}{ccc}
\includegraphics[width=0.31\linewidth]{swe_pro/edit_read_ratio/gemini_3.1_pro.pdf} &
\includegraphics[width=0.31\linewidth]{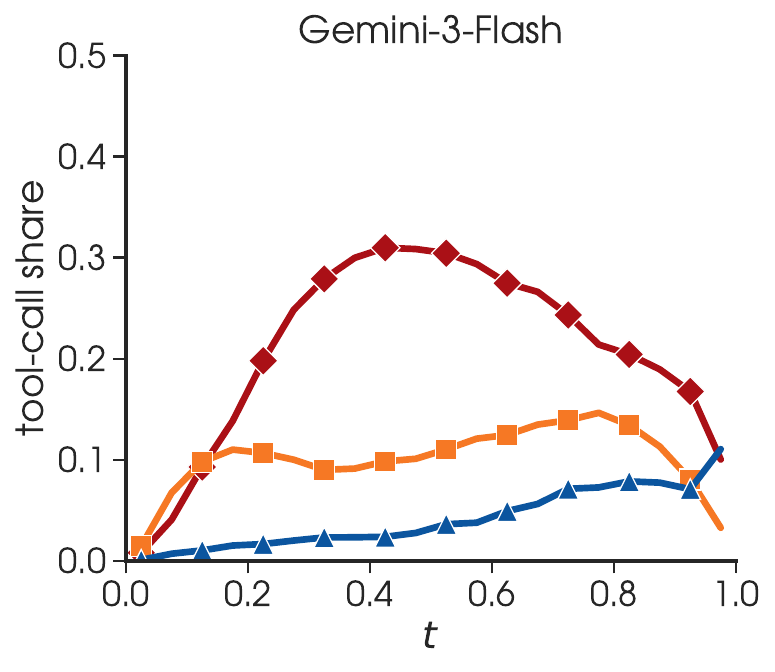} & \\
\includegraphics[width=0.31\linewidth]{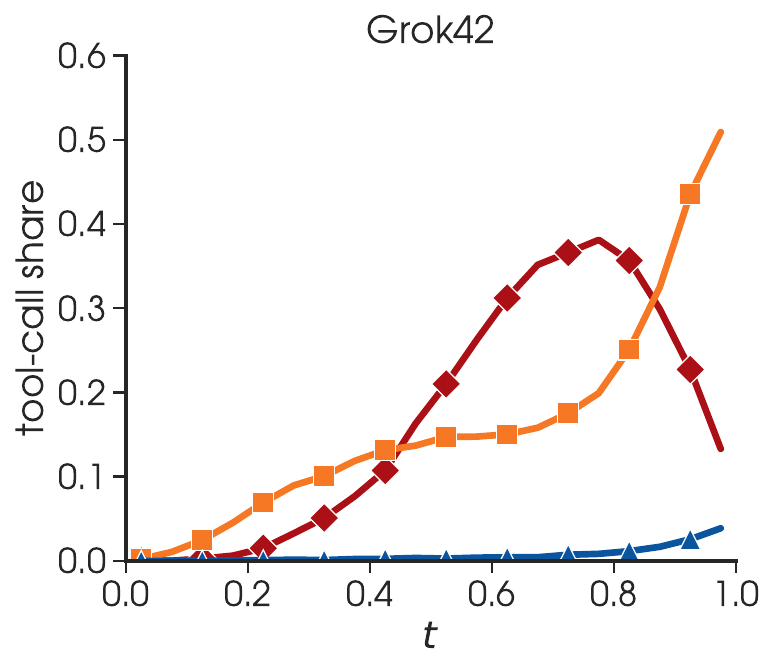} & & \\
\includegraphics[width=0.31\linewidth]{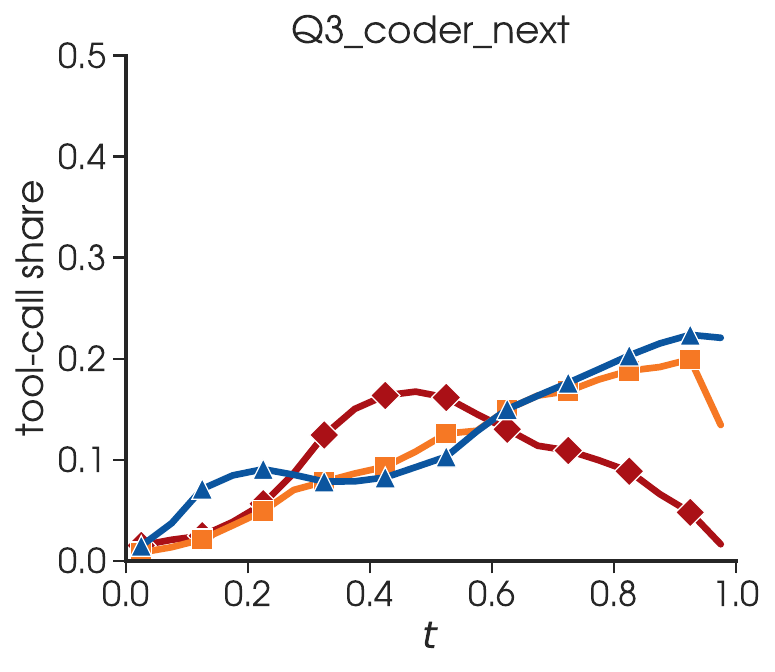} &
\includegraphics[width=0.31\linewidth]{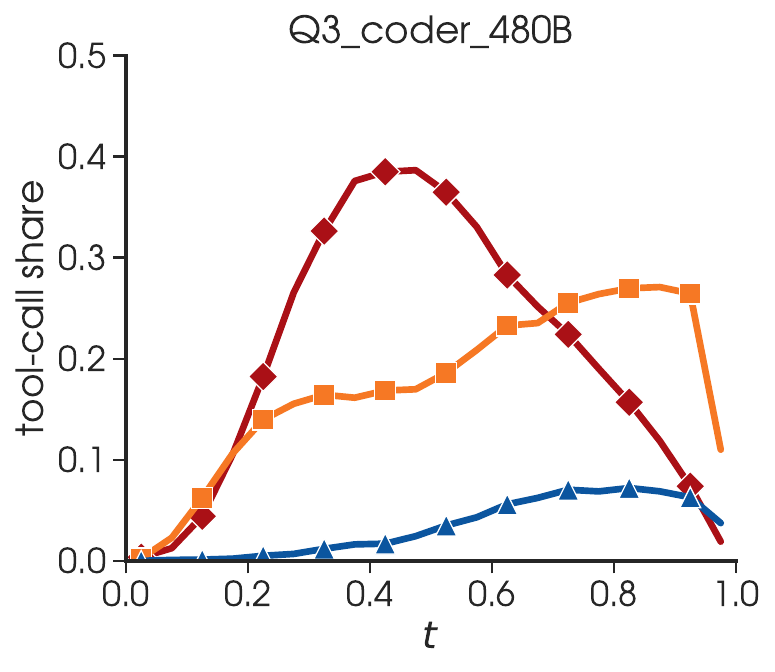} &
\includegraphics[width=0.31\linewidth]{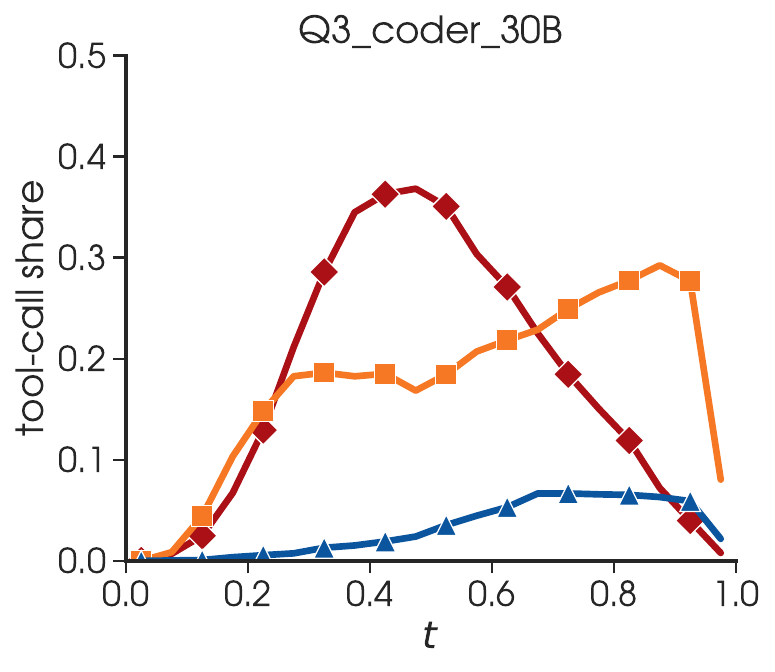} \\
\includegraphics[width=0.31\linewidth]{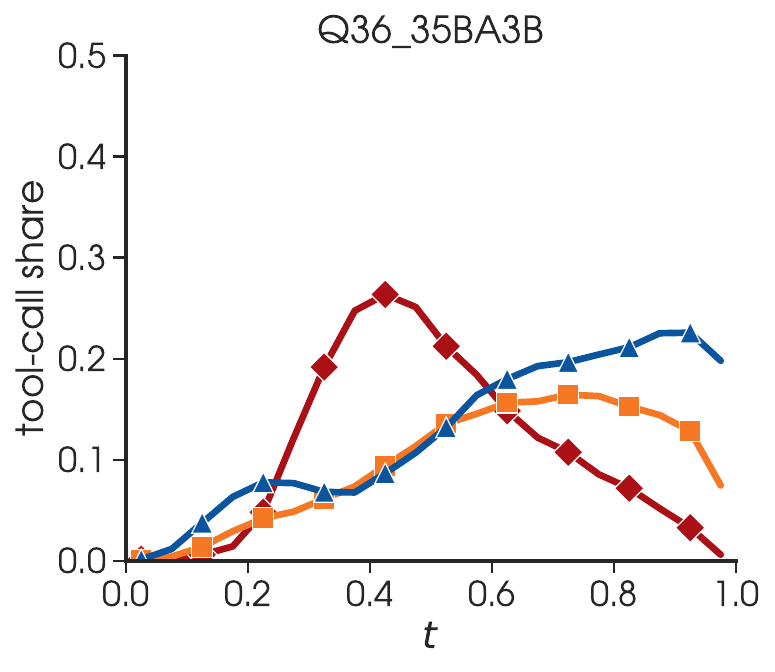} &
\includegraphics[width=0.31\linewidth]{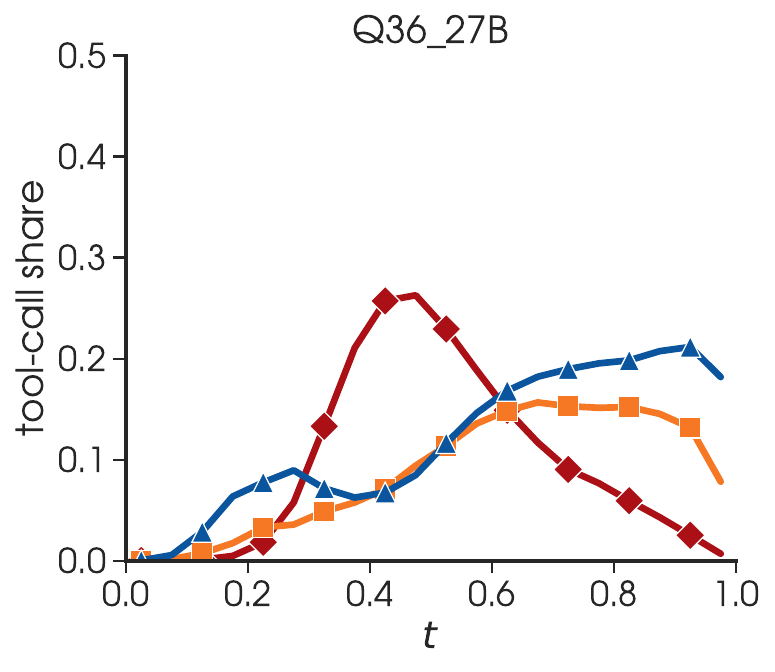} &
\includegraphics[width=0.31\linewidth]{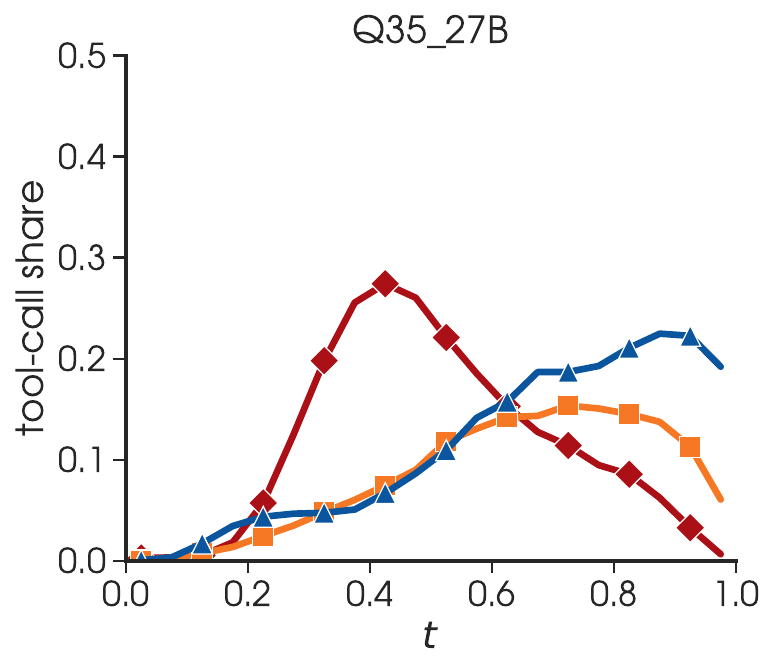} \\
\end{tabular}
\caption{Edit/test ratio per cycle on SWE-Bench-Pro (part 2 of 2): Gemini, Grok and Qwen families. Continued from Figure~\ref{fig:metrics-sbp-editread}.}
\label{fig:metrics-sbp-editread-b}
\end{figure}

\begin{figure}[p]
\centering
\setlength{\tabcolsep}{1pt}
\renewcommand{\arraystretch}{0.5}
\begin{tabular}{ccc}
\includegraphics[width=0.31\linewidth]{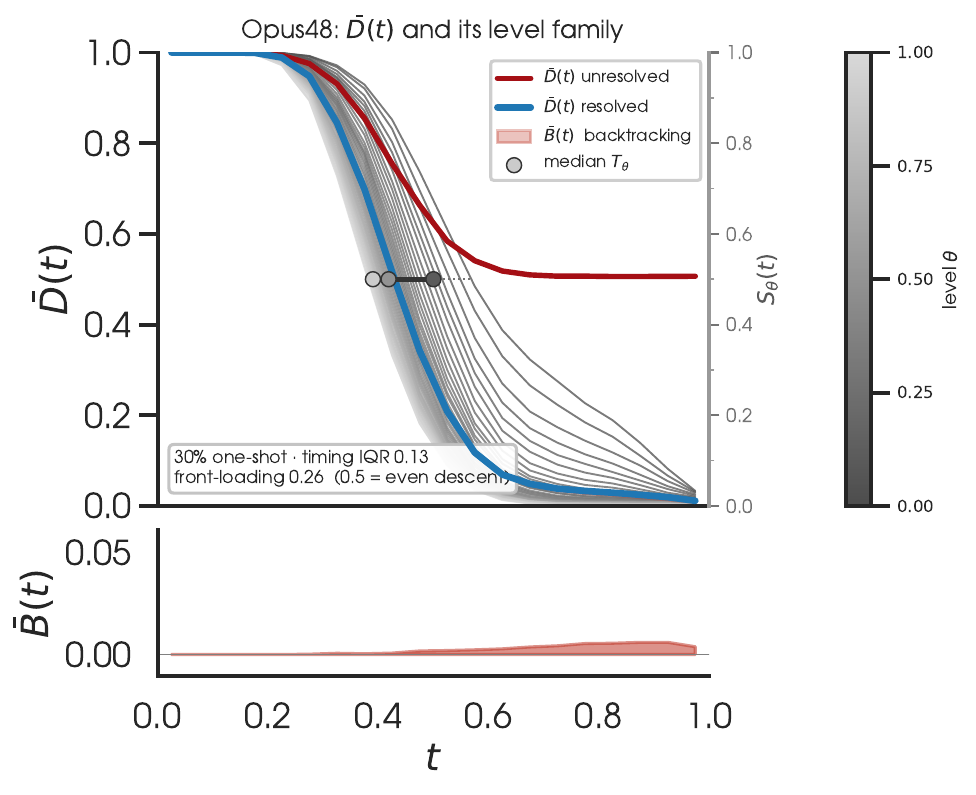} &
\includegraphics[width=0.31\linewidth]{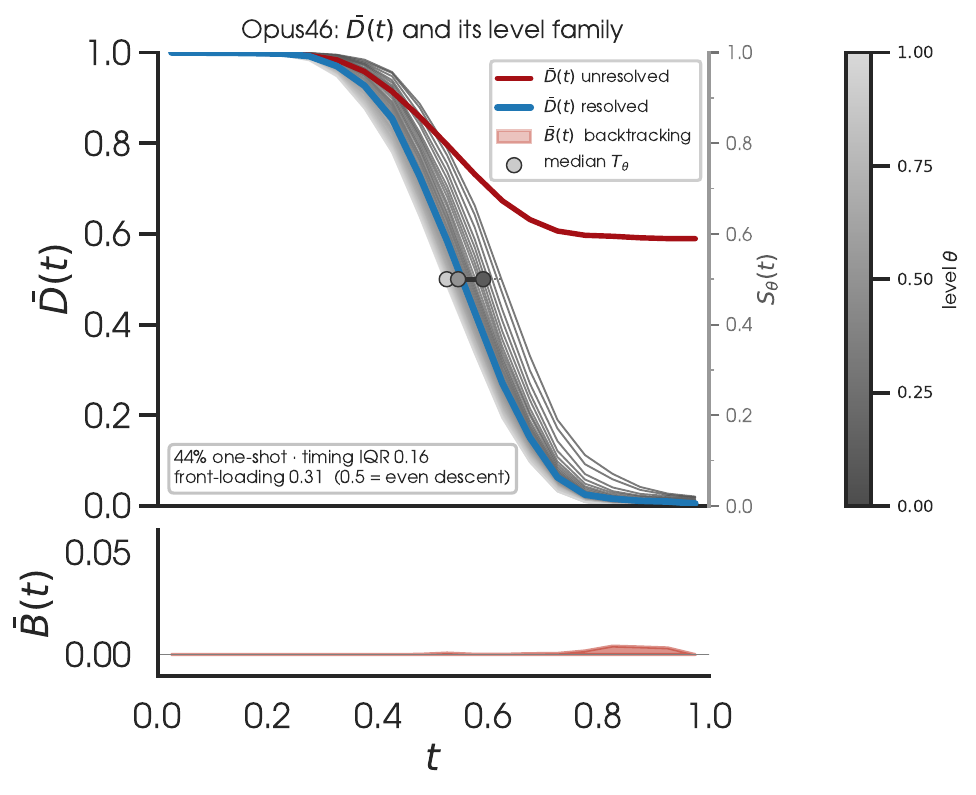} &
\includegraphics[width=0.31\linewidth]{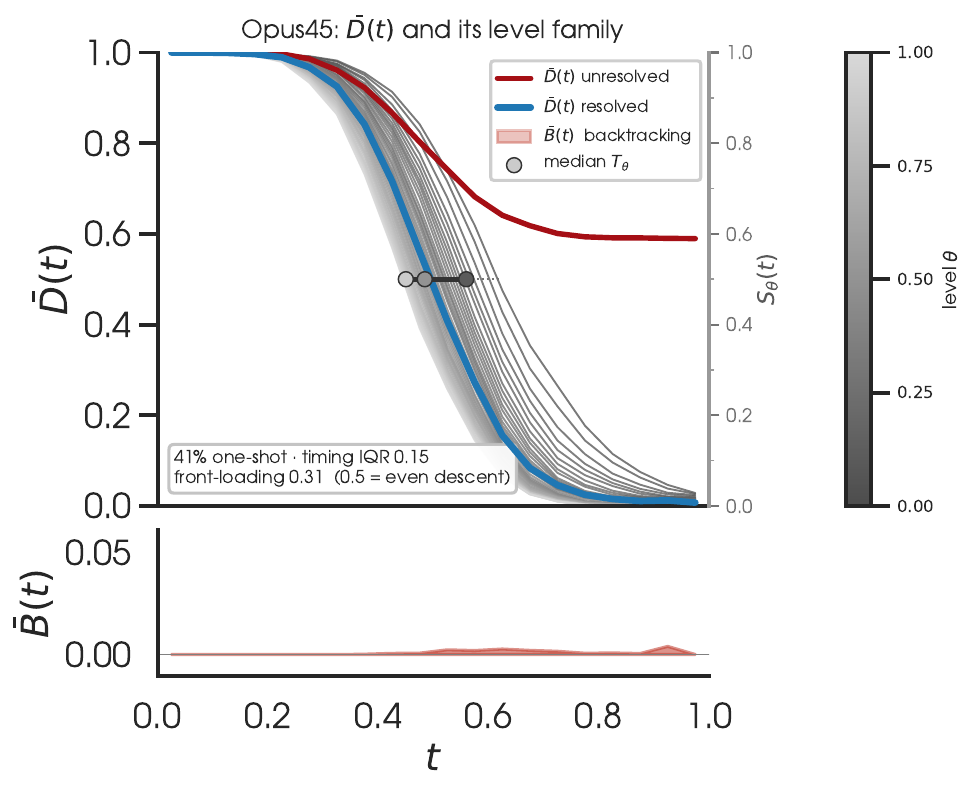} \\
\includegraphics[width=0.31\linewidth]{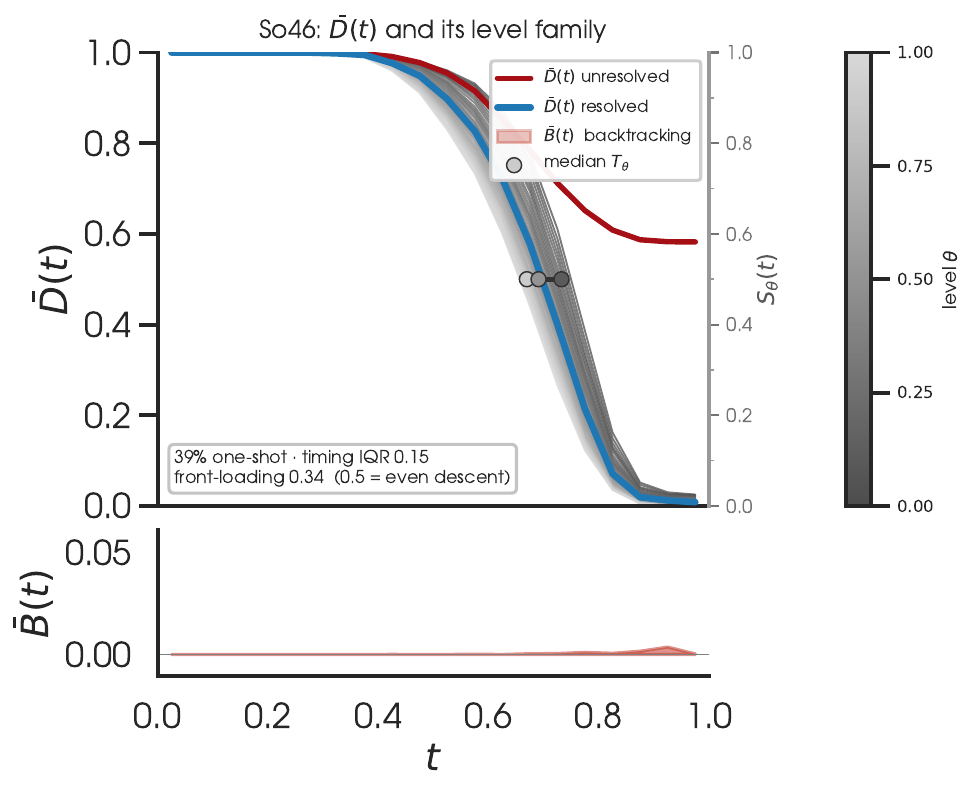} &
\includegraphics[width=0.31\linewidth]{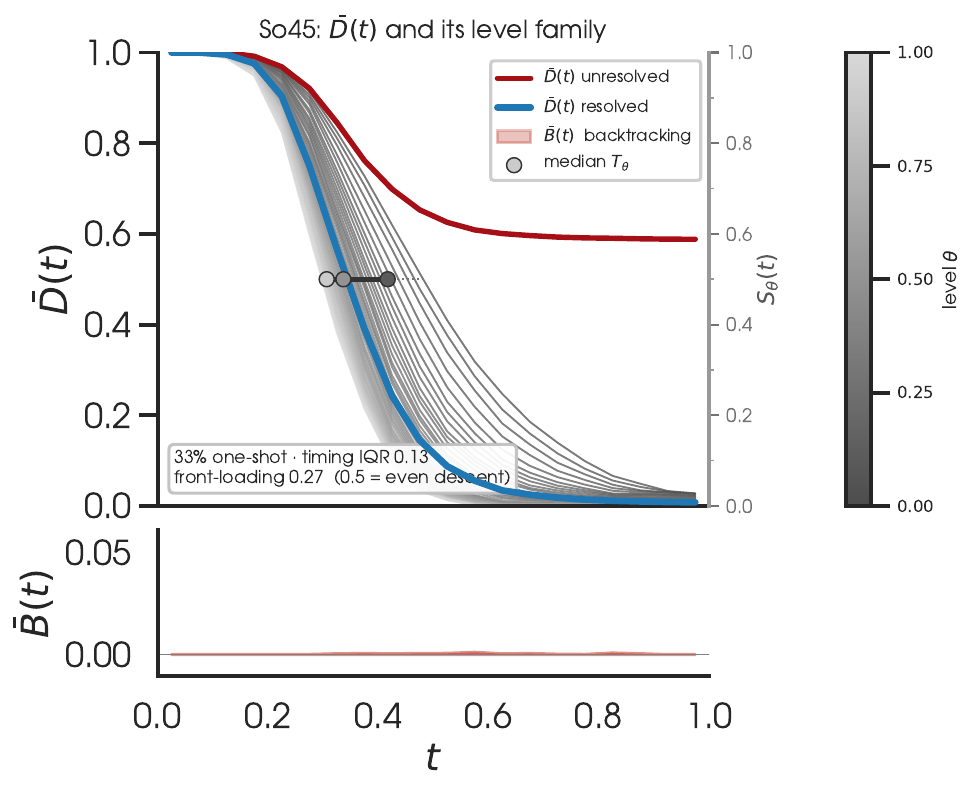} &
\includegraphics[width=0.31\linewidth]{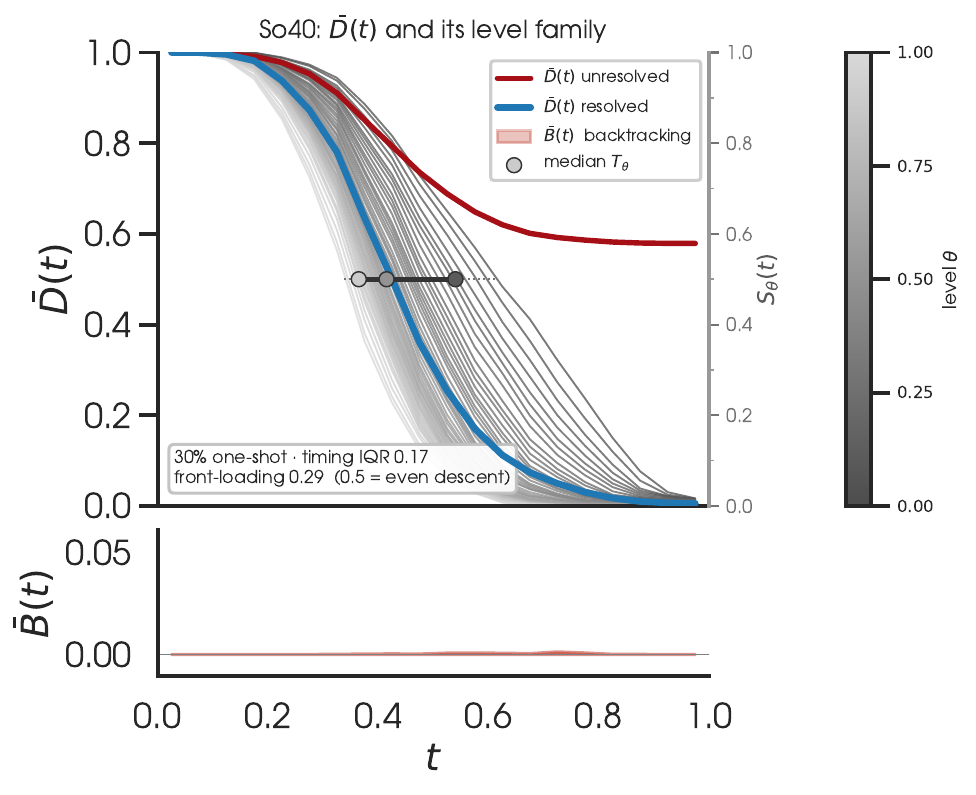} \\
\includegraphics[width=0.31\linewidth]{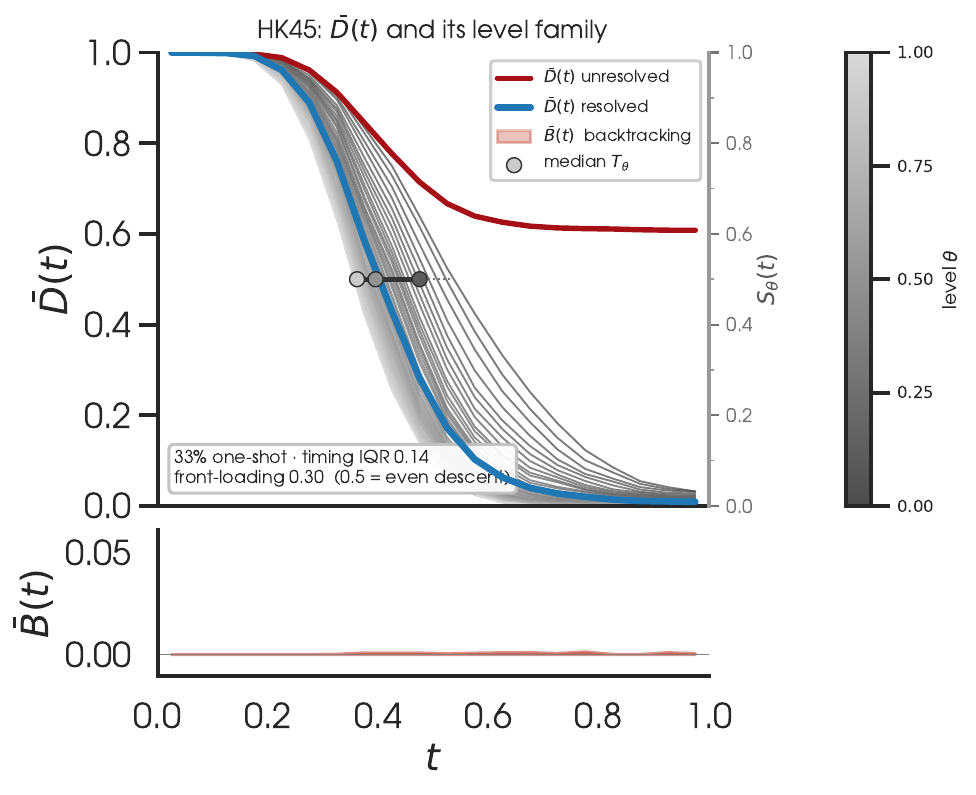} &
\includegraphics[width=0.31\linewidth]{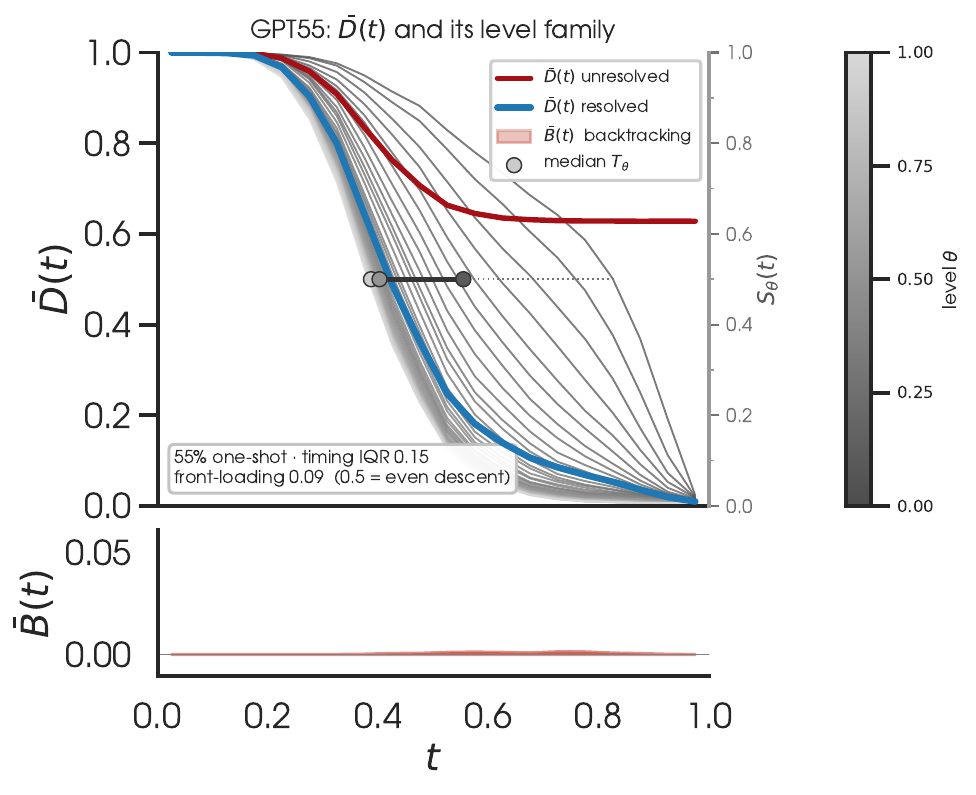} &
\includegraphics[width=0.31\linewidth]{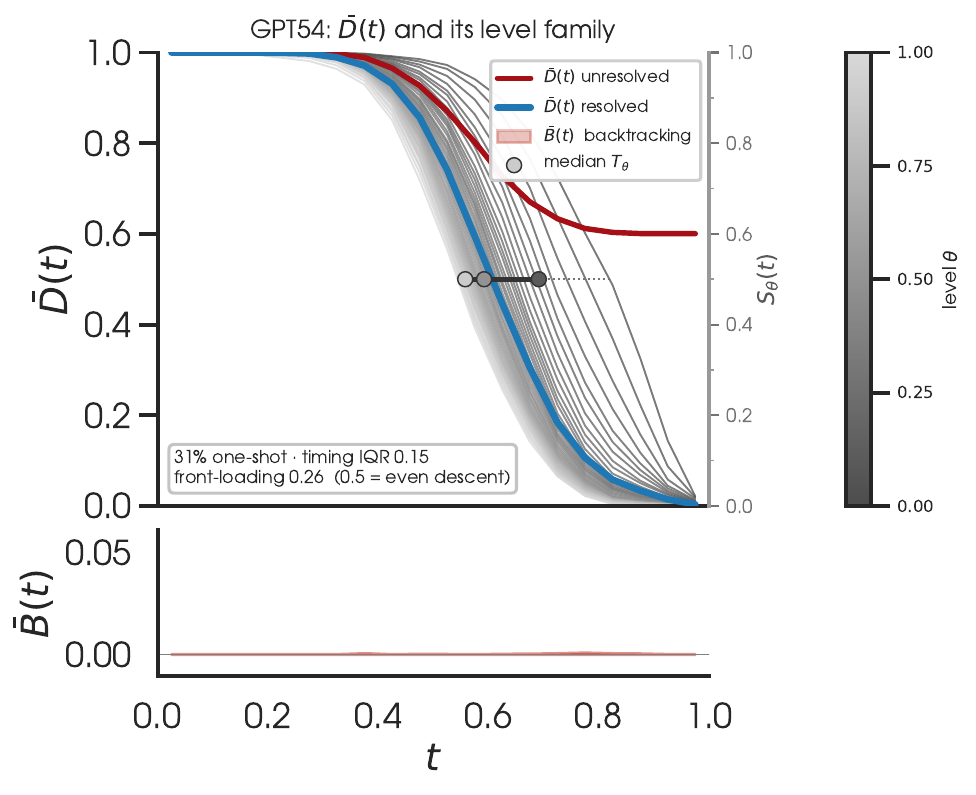} \\
\includegraphics[width=0.31\linewidth]{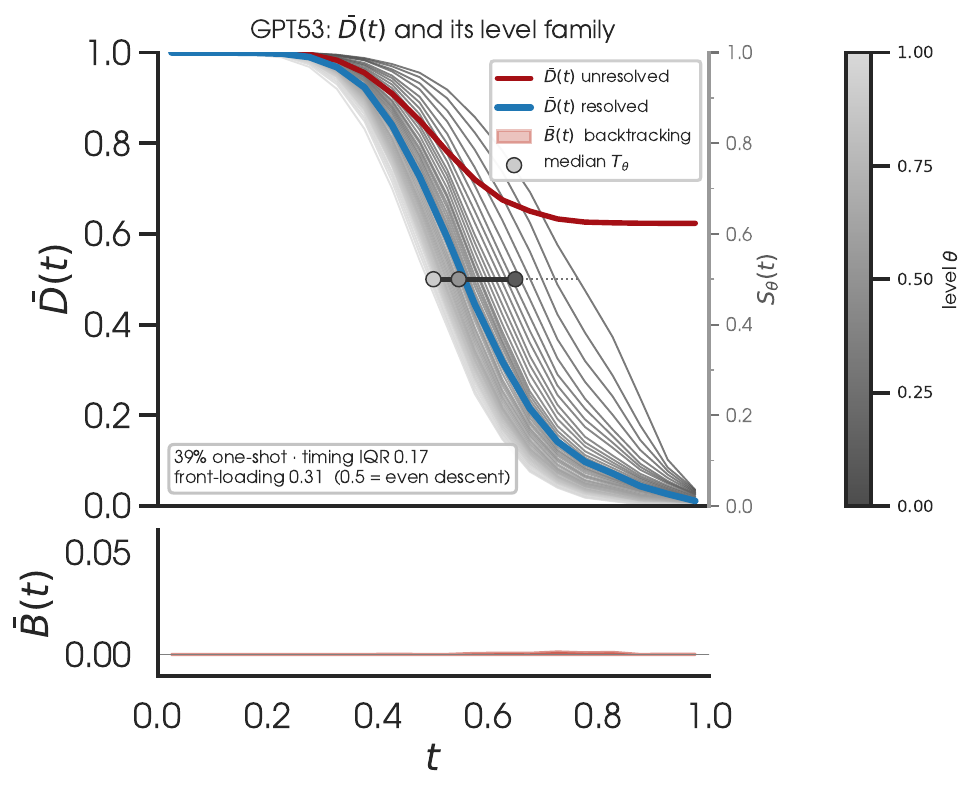} &
\includegraphics[width=0.31\linewidth]{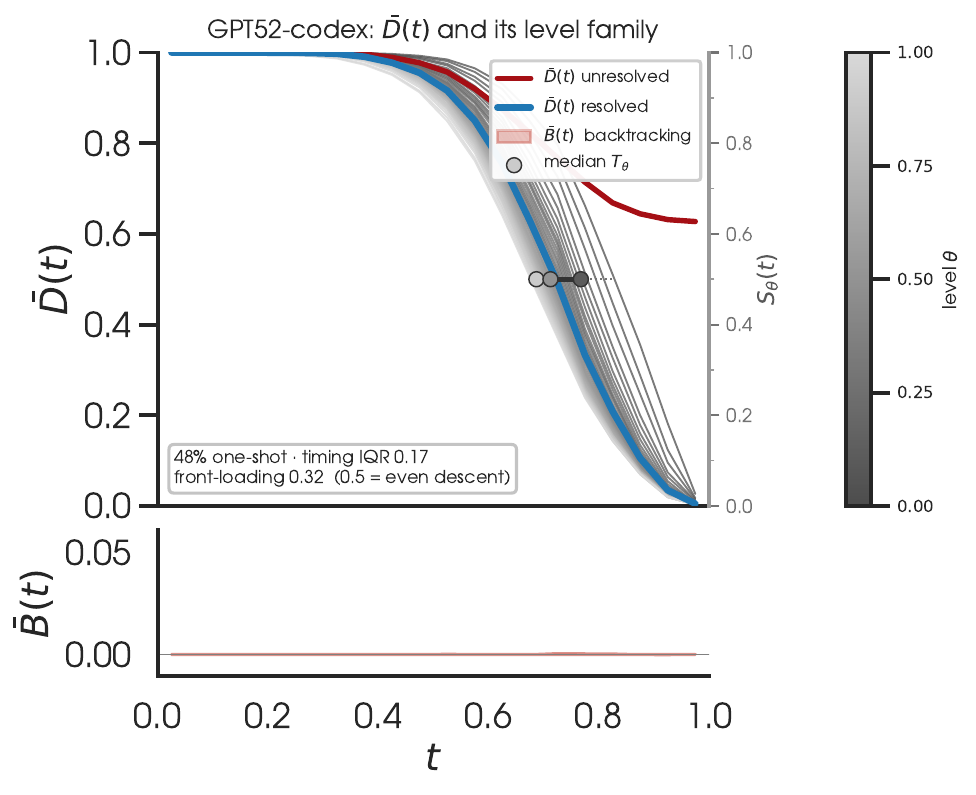} &
\includegraphics[width=0.31\linewidth]{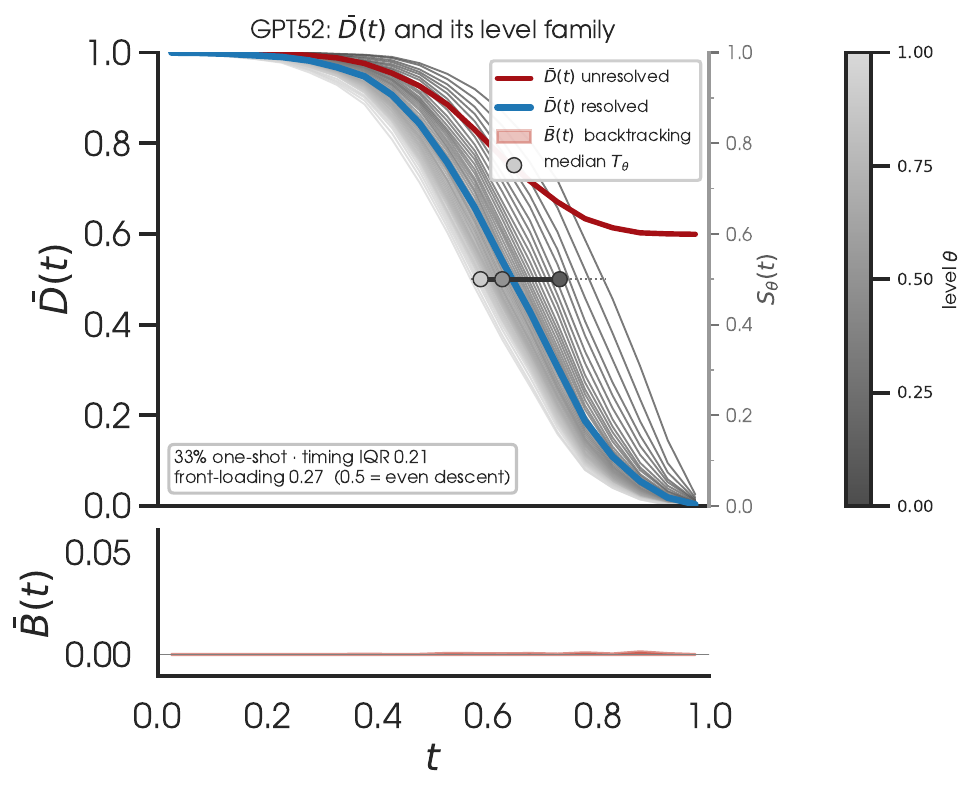} \\
\includegraphics[width=0.31\linewidth]{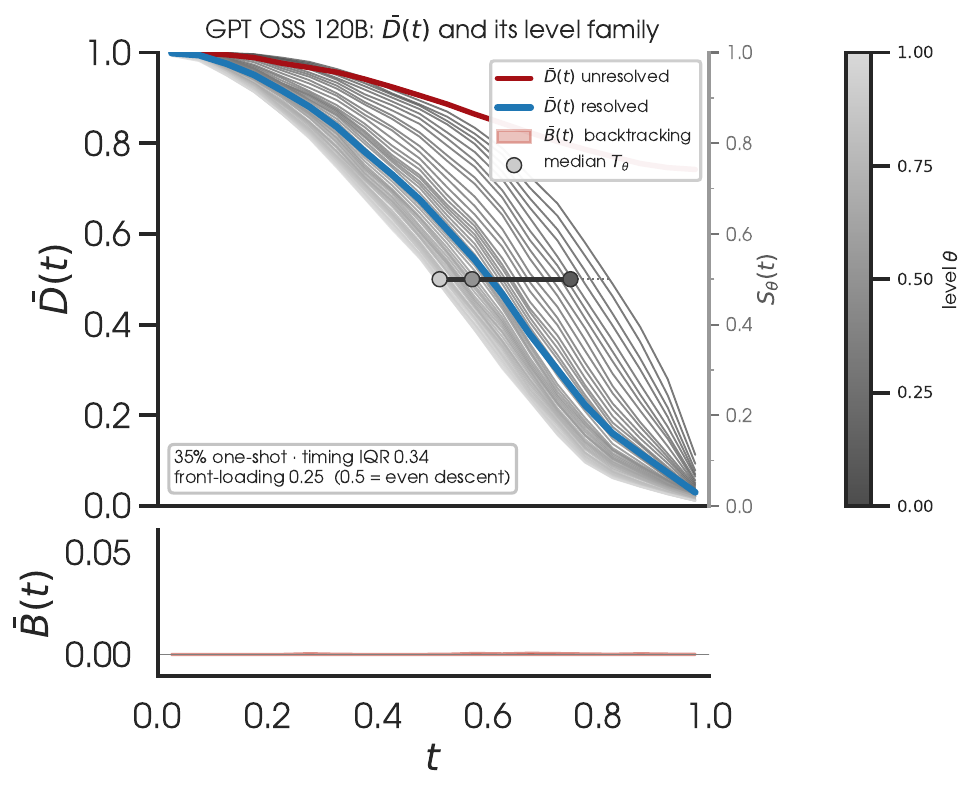} &
\includegraphics[width=0.31\linewidth]{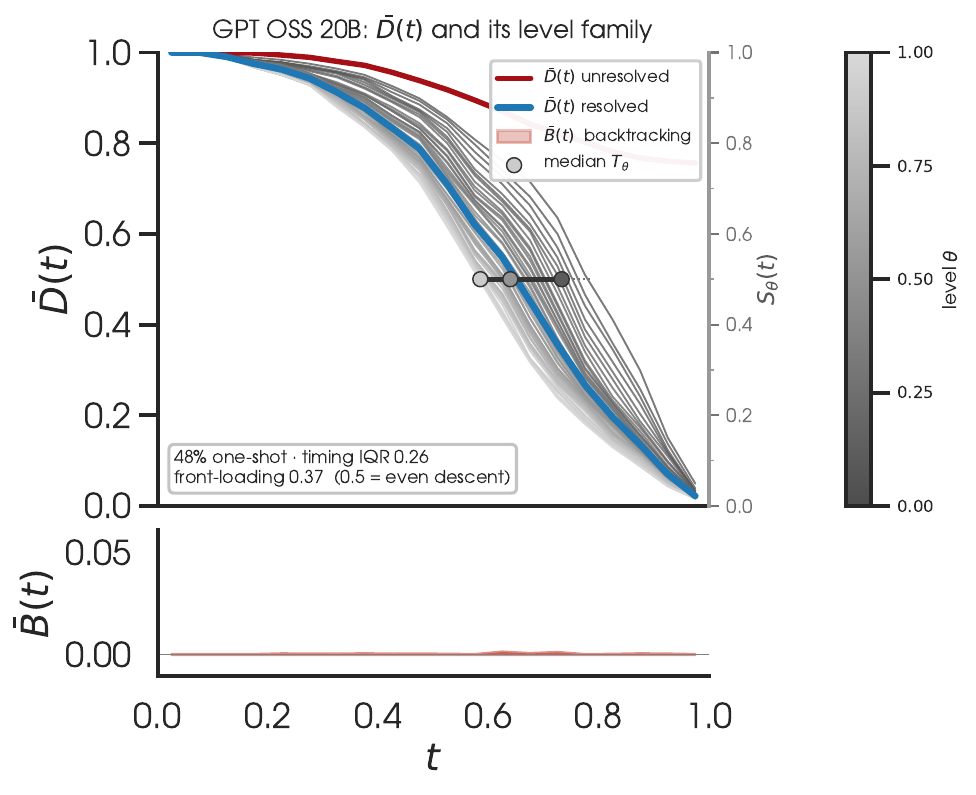} &
 \\
\end{tabular}
\caption{Level families on SWE-Bench-Pro (part 1 of 2): Claude and GPT families. Grey curves are the level sets $S_\theta(t)$ of the resolved trajectories (Section~\ref{sec:distance}), the blue curve is their mean $\bar D(t)$, the dark-red curve is the mean $\bar D(t)$ of the unresolved runs (which plateaus where they stop), the markers give the median crossing time $T_\theta$ (the median over runs of the first-entry time $T_r(\theta)$, i.e.\ where $S_\theta$ crosses $1/2$) at $\theta=0.5$ with its interquartile range, and the strip beneath each panel is the backtracking term $\bar B(t)$. The inset reports the one-shot share (runs that land the patch in a single edit); the timing IQR, the interquartile range of the runs' half-way times $T_r(0.5)$, i.e.\ how dispersed solve timing is across runs; and the front-loading index $(T_{0.5}-T_{0.9})/(T_{0.1}-T_{0.9})$, the position of the median half-way time between the median crossing times of levels $0.9$ and $0.1$ ($0.5$ = even descent; lower = most of the distance falls early).}
\label{fig:metrics-sbp-soldist}
\end{figure}

\begin{figure}[p]
\centering
\setlength{\tabcolsep}{1pt}
\renewcommand{\arraystretch}{0.5}
\begin{tabular}{ccc}
\includegraphics[width=0.31\linewidth]{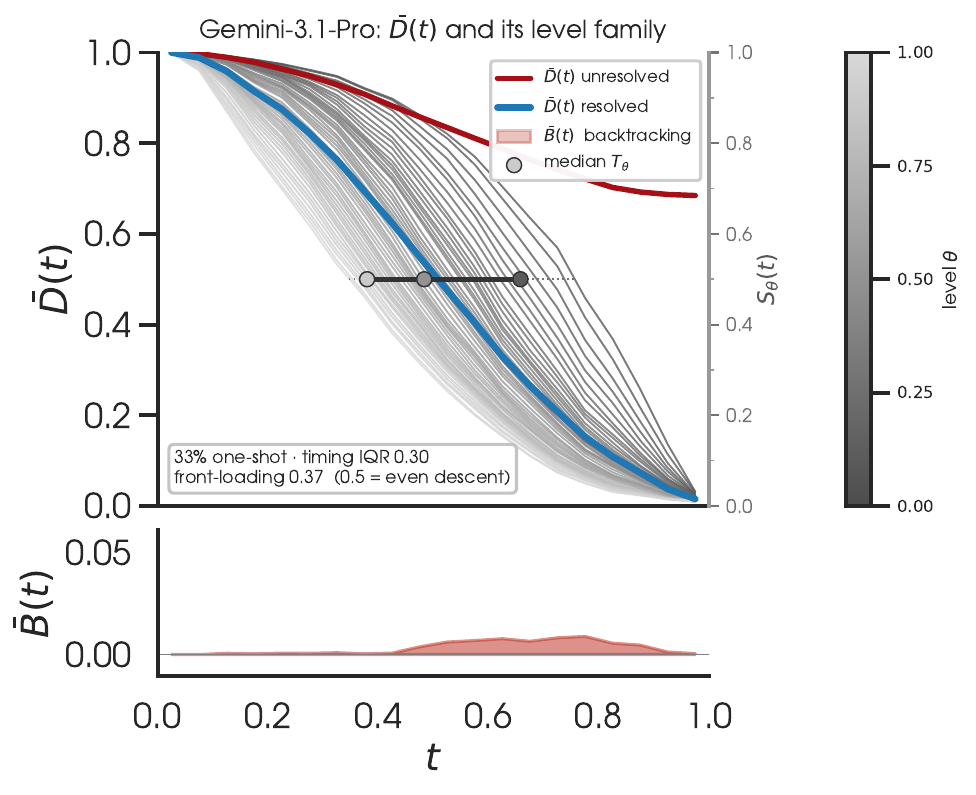} &
\includegraphics[width=0.31\linewidth]{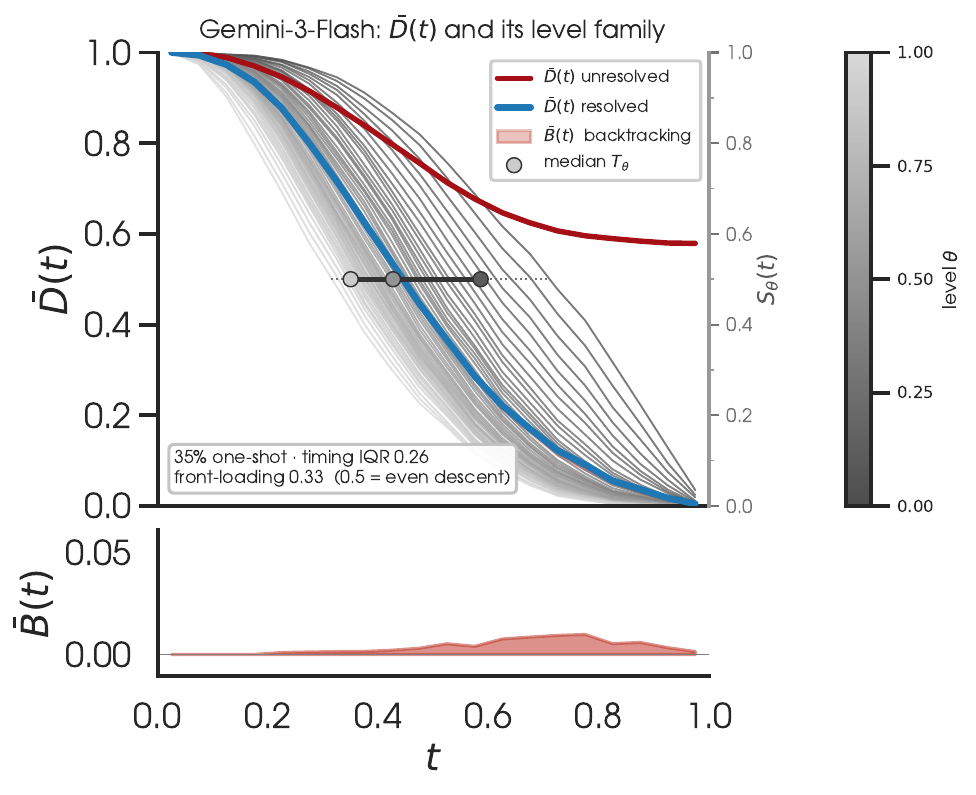} &
\includegraphics[width=0.31\linewidth]{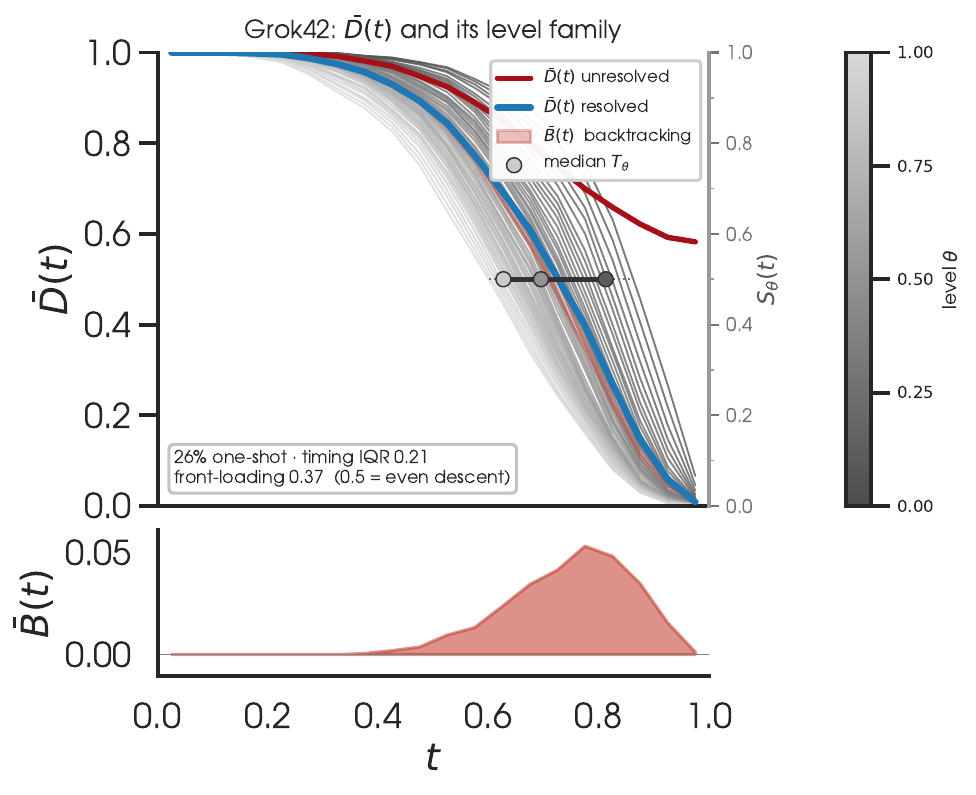} \\
\includegraphics[width=0.31\linewidth]{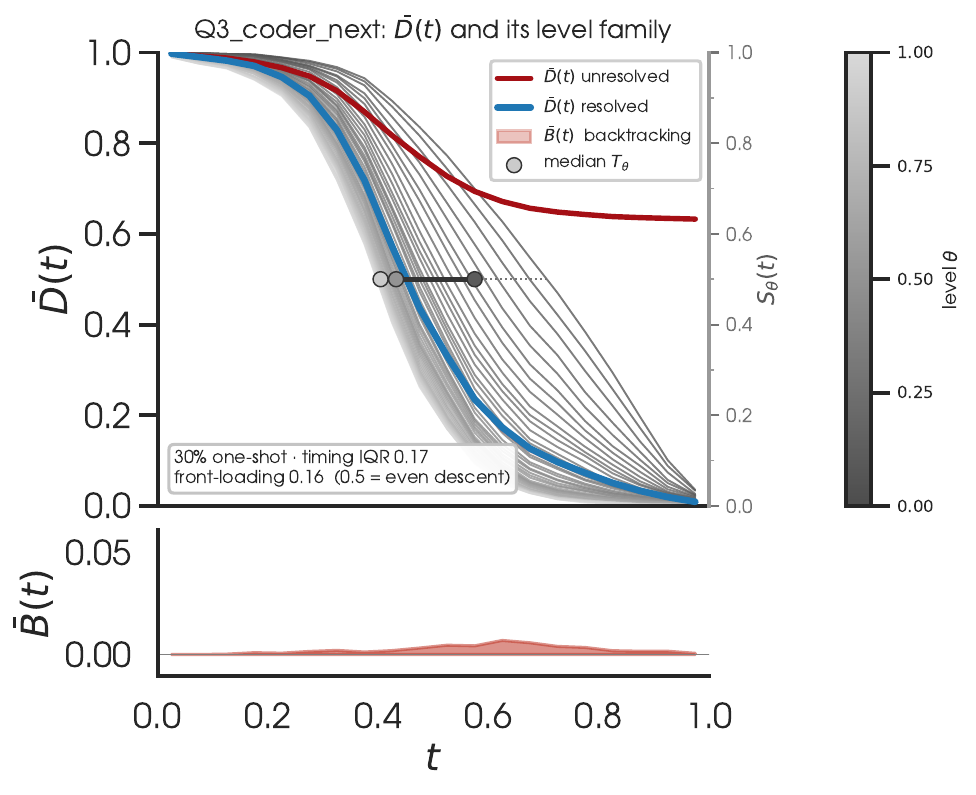} &
\includegraphics[width=0.31\linewidth]{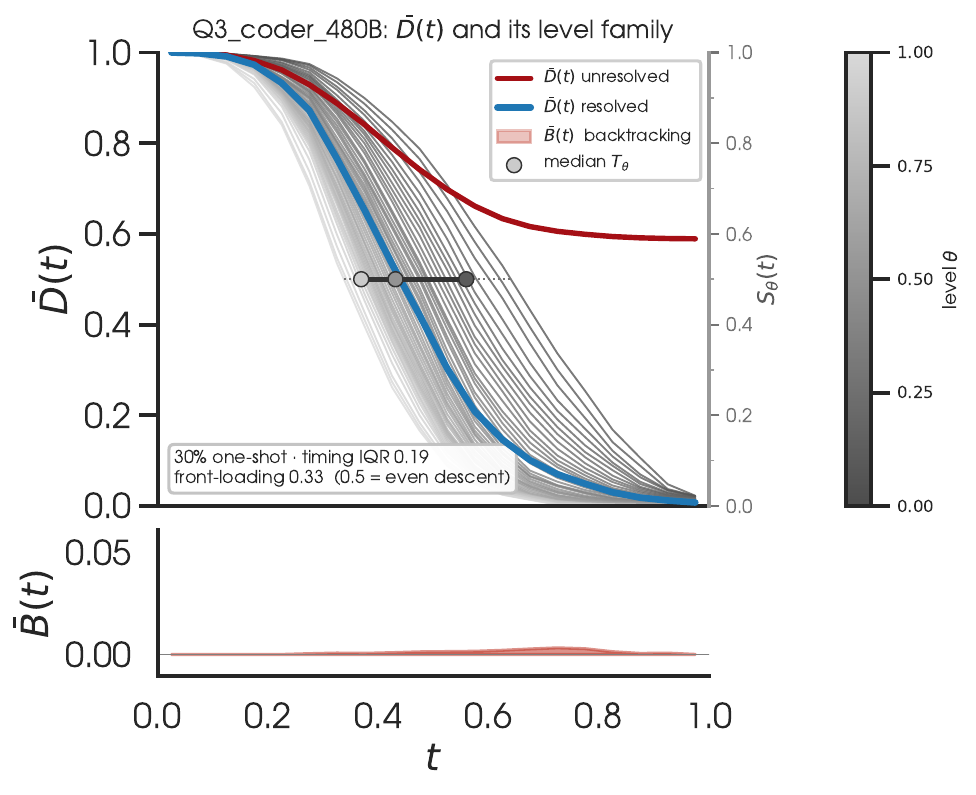} &
\includegraphics[width=0.31\linewidth]{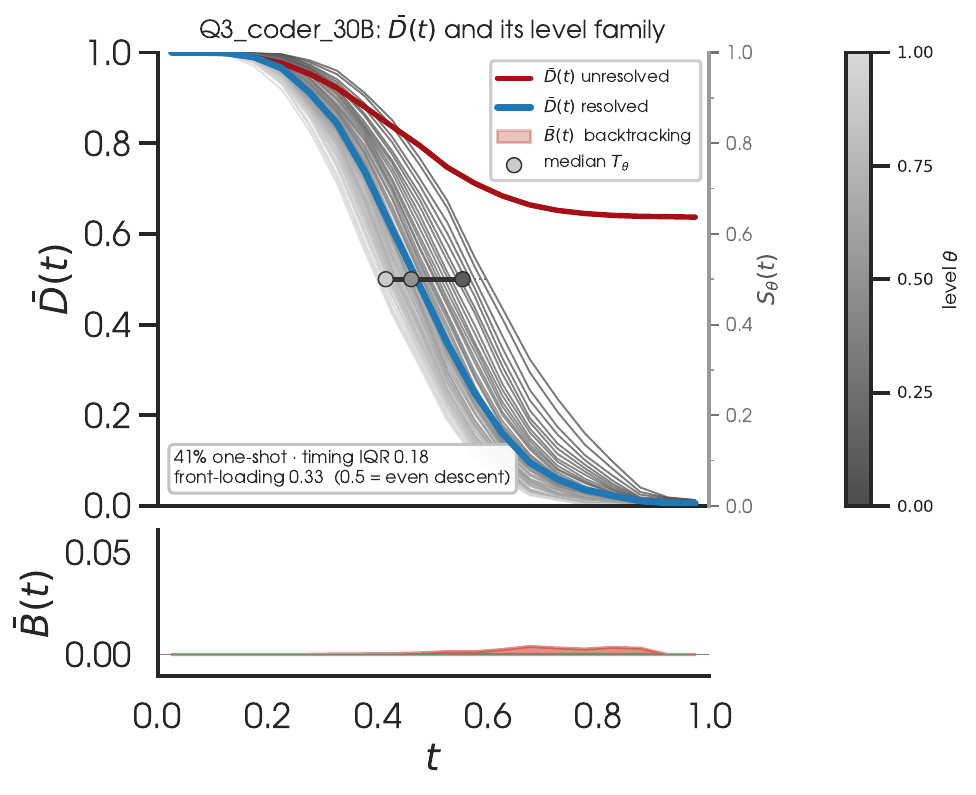} \\
\includegraphics[width=0.31\linewidth]{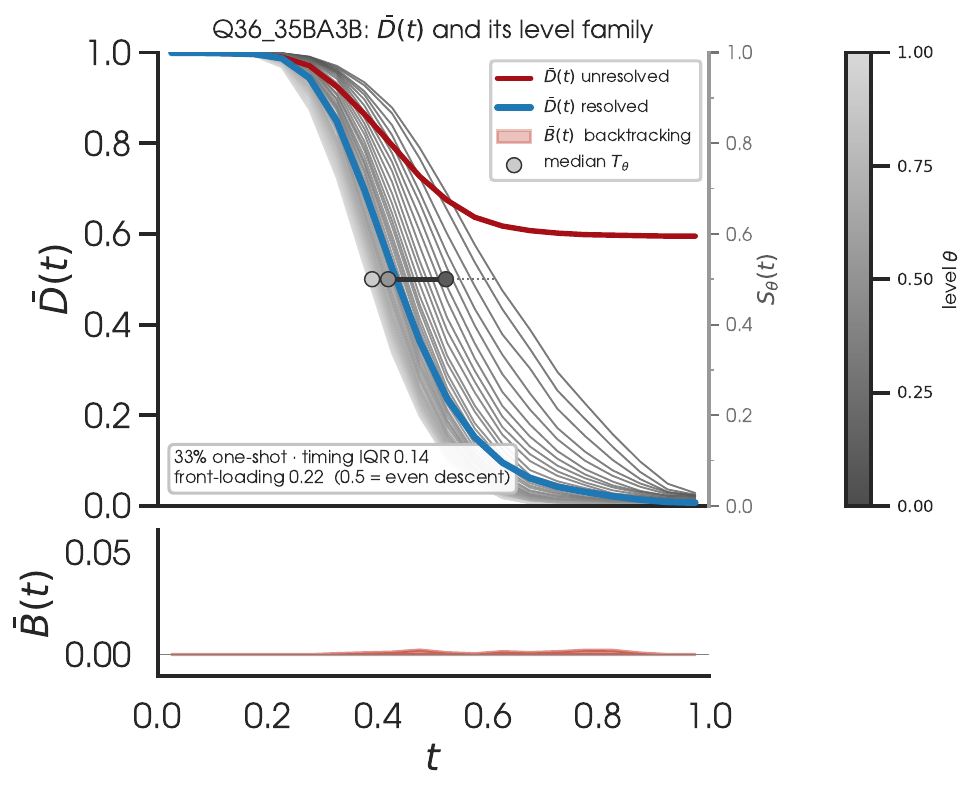} &
\includegraphics[width=0.31\linewidth]{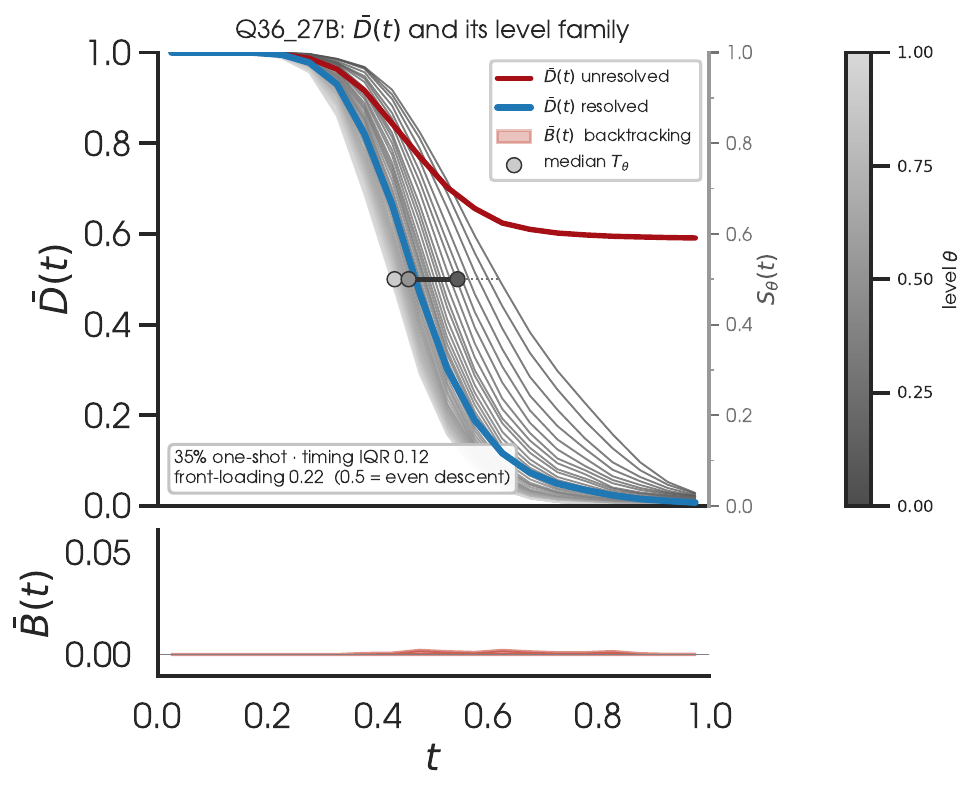} &
\includegraphics[width=0.31\linewidth]{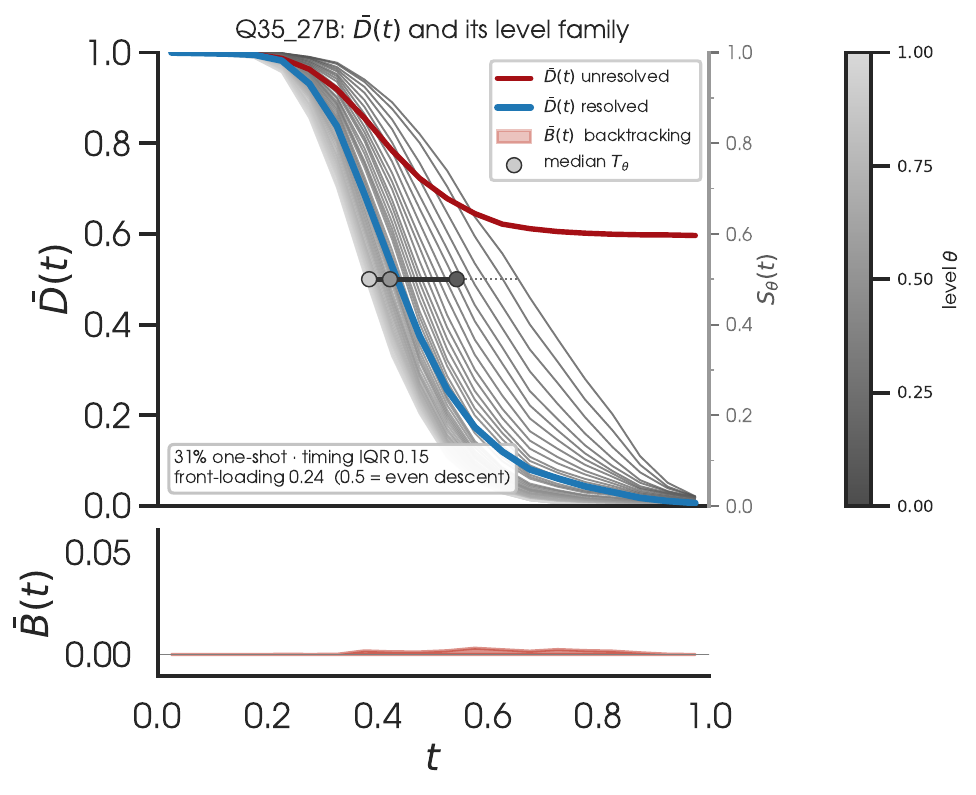} \\
\end{tabular}
\caption{Level families on SWE-Bench-Pro (part 2 of 2): Gemini, Grok and Qwen families. Continued from Figure~\ref{fig:metrics-sbp-soldist}.}
\label{fig:metrics-sbp-soldist-b}
\end{figure}

\begin{figure}[p]
\centering
\setlength{\tabcolsep}{1pt}
\renewcommand{\arraystretch}{0.5}
\begin{tabular}{ccc}
\includegraphics[width=0.31\linewidth]{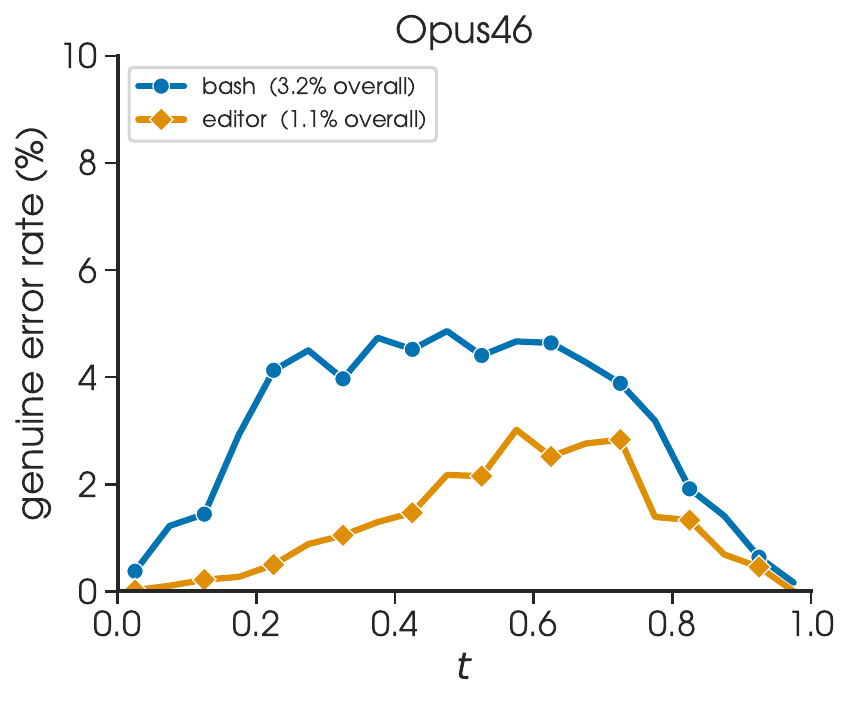} &
\includegraphics[width=0.31\linewidth]{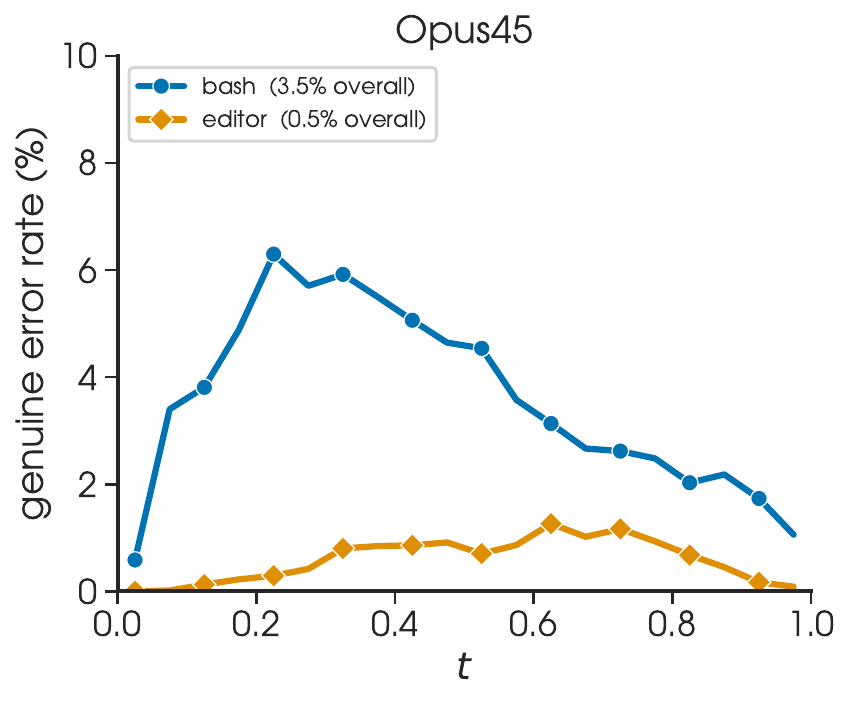} &
\includegraphics[width=0.31\linewidth]{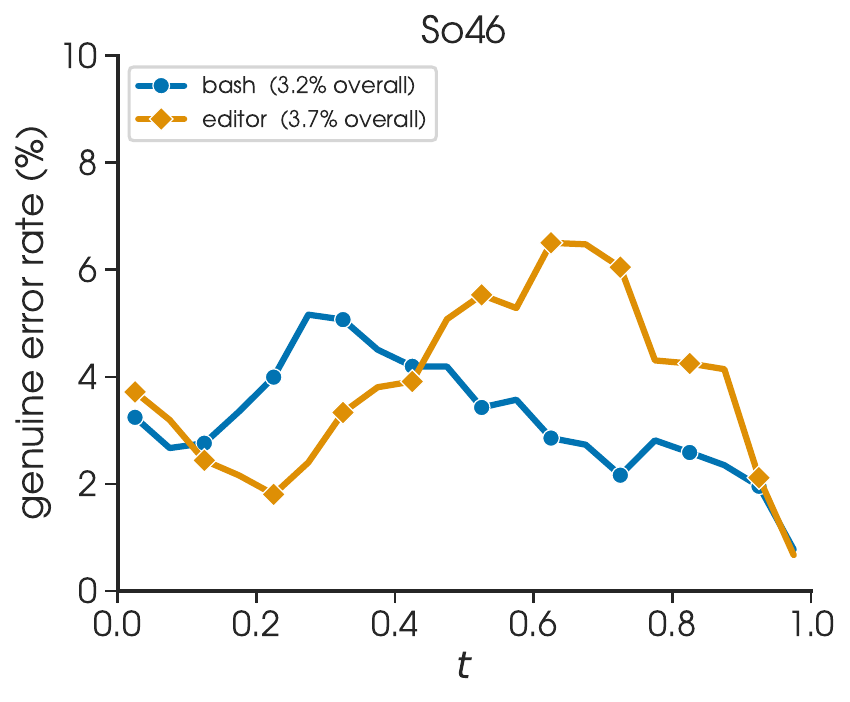} \\
\includegraphics[width=0.31\linewidth]{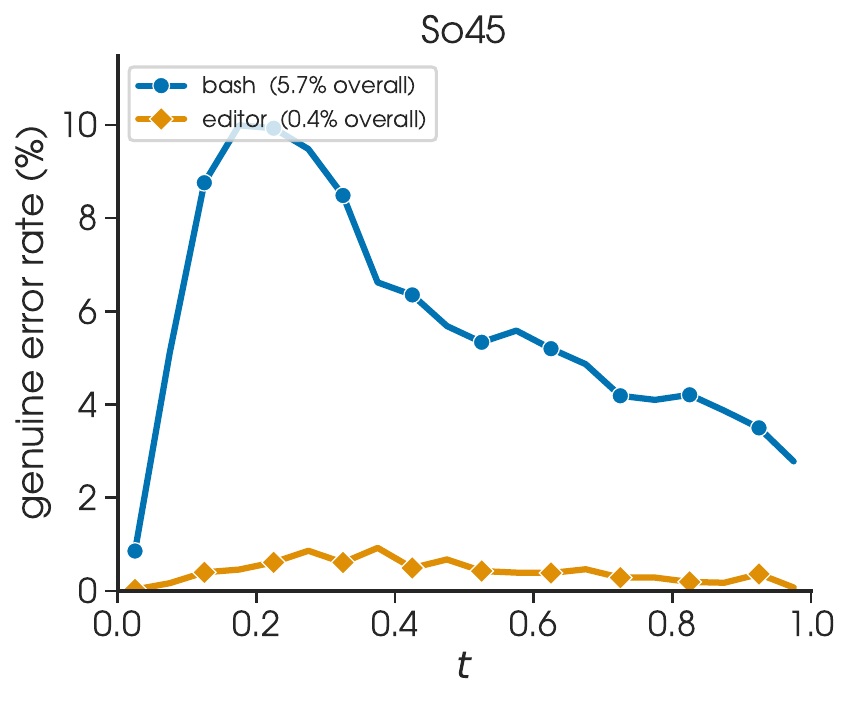} &
\includegraphics[width=0.31\linewidth]{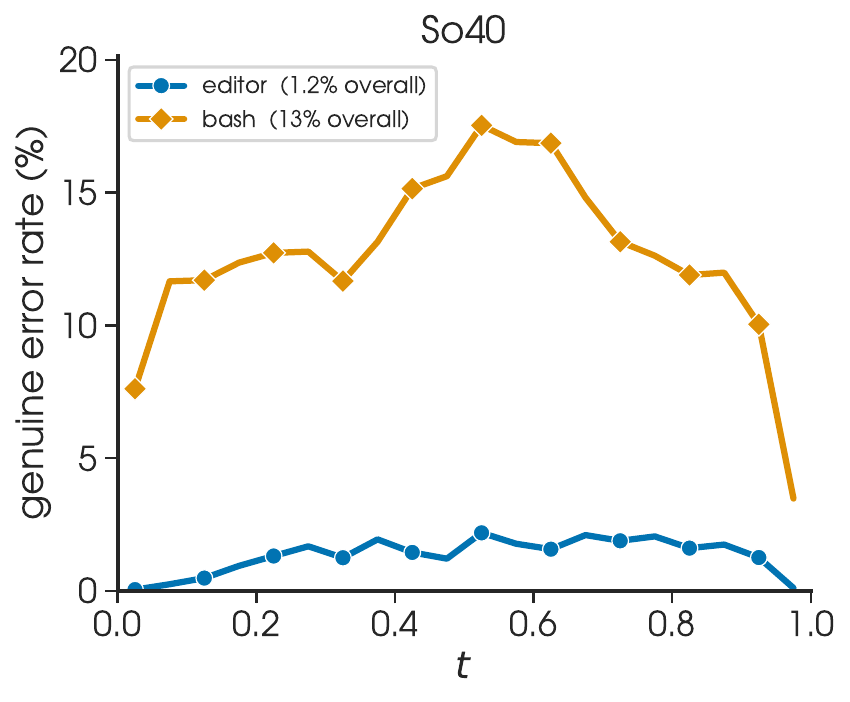} &
\includegraphics[width=0.31\linewidth]{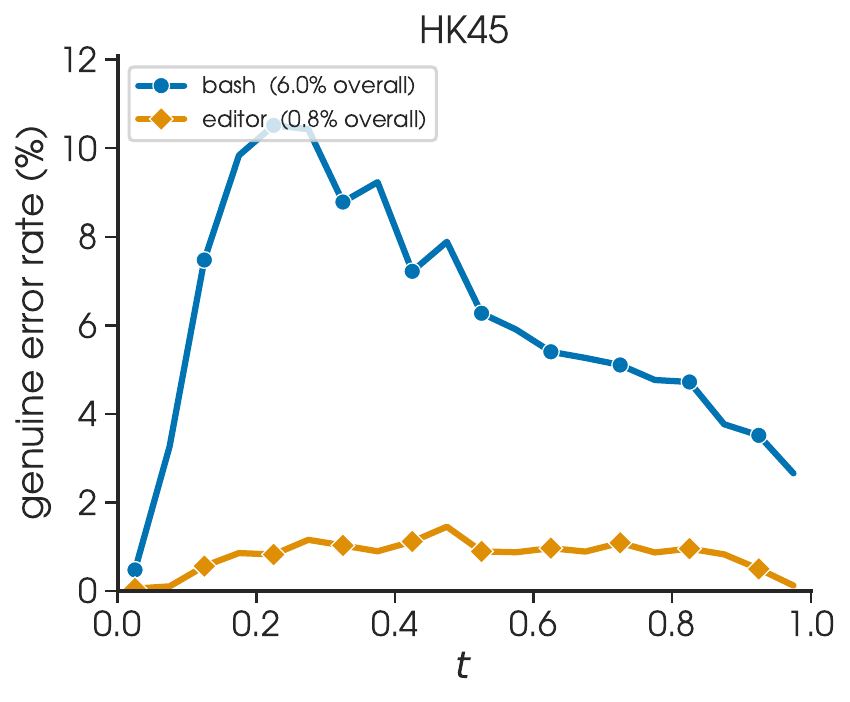} \\
\includegraphics[width=0.31\linewidth]{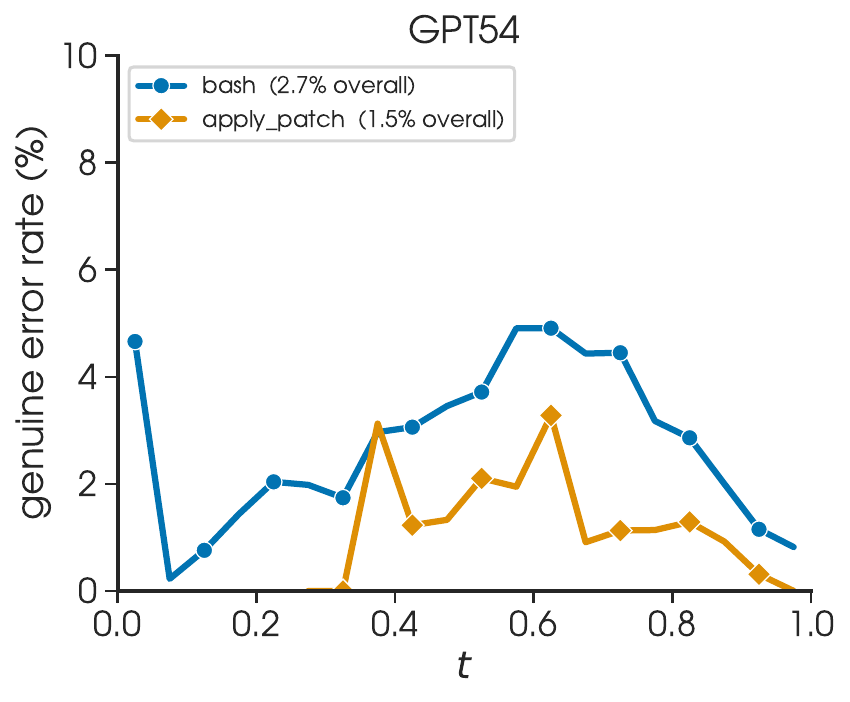} &
\includegraphics[width=0.31\linewidth]{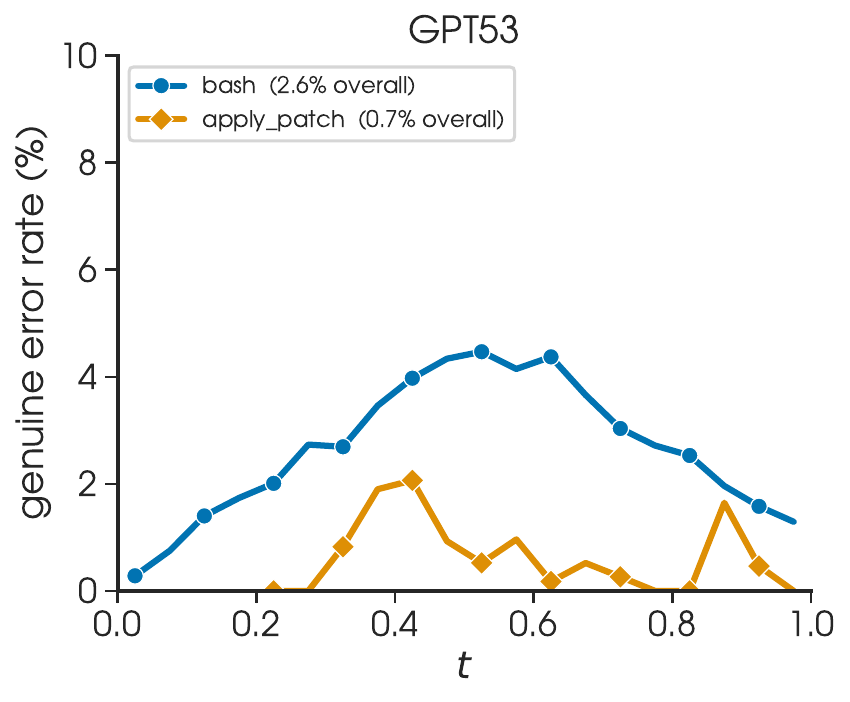} &
\includegraphics[width=0.31\linewidth]{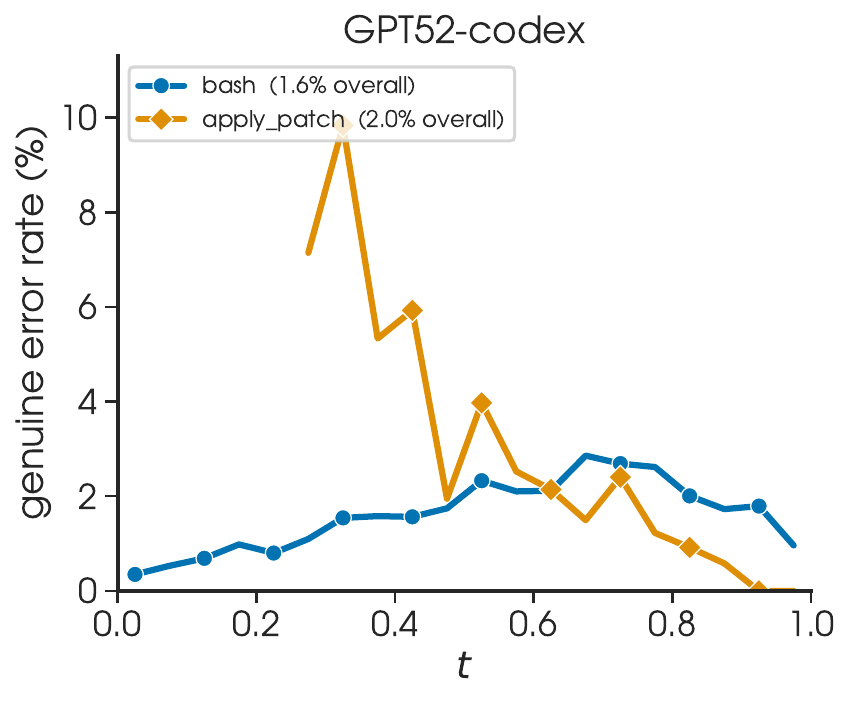} \\
\includegraphics[width=0.31\linewidth]{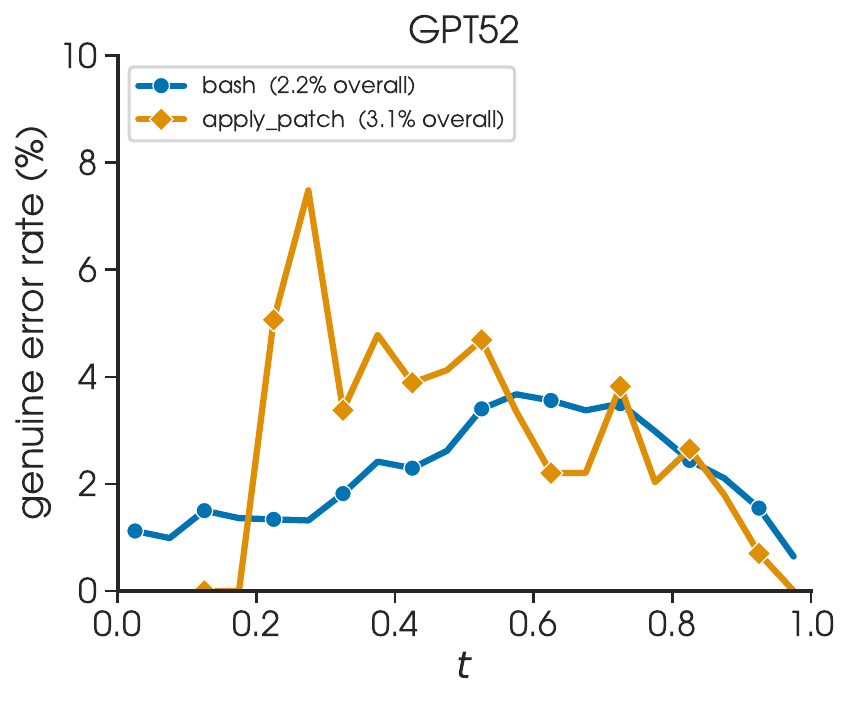} &
\includegraphics[width=0.31\linewidth]{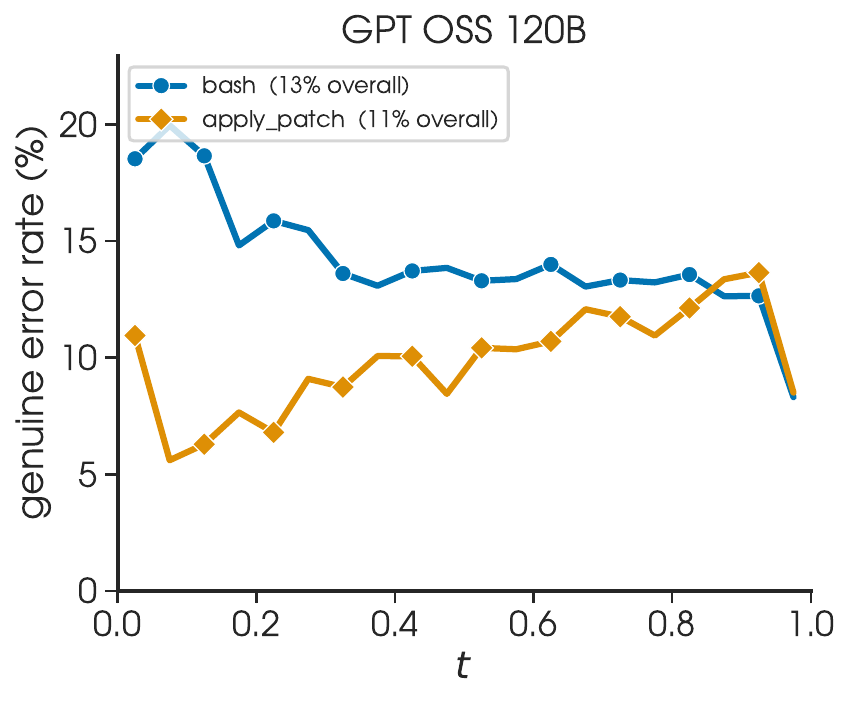} &
\includegraphics[width=0.31\linewidth]{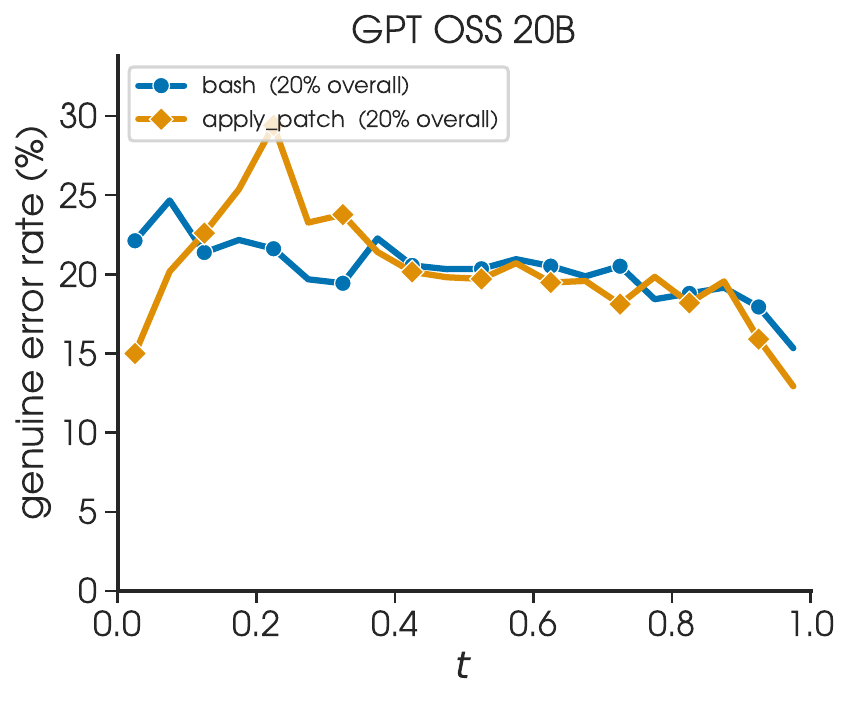} \\
\end{tabular}
\caption{Genuine tool-error rate per cycle on SWE-Bench-Pro (part 1 of 2): Anthropic Claude and OpenAI families. Benign exits (\texttt{grep} no-match, \texttt{pytest} test failures, etc.)\ are excluded.}
\label{fig:metrics-sbp-toolerr}
\end{figure}

\begin{figure}[p]
\centering
\setlength{\tabcolsep}{1pt}
\renewcommand{\arraystretch}{0.5}
\begin{tabular}{ccc}
\includegraphics[width=0.31\linewidth]{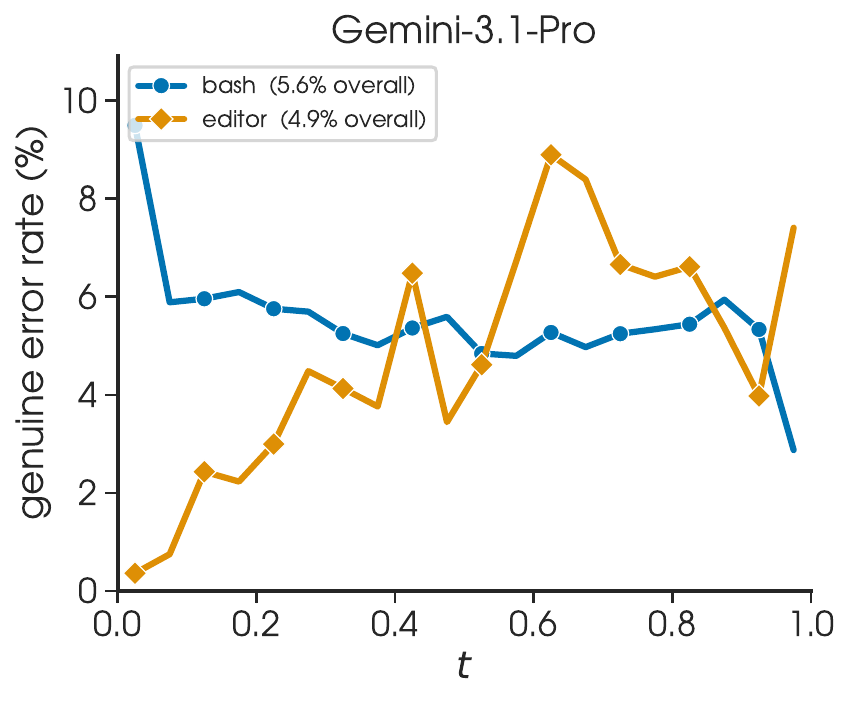} &
\includegraphics[width=0.31\linewidth]{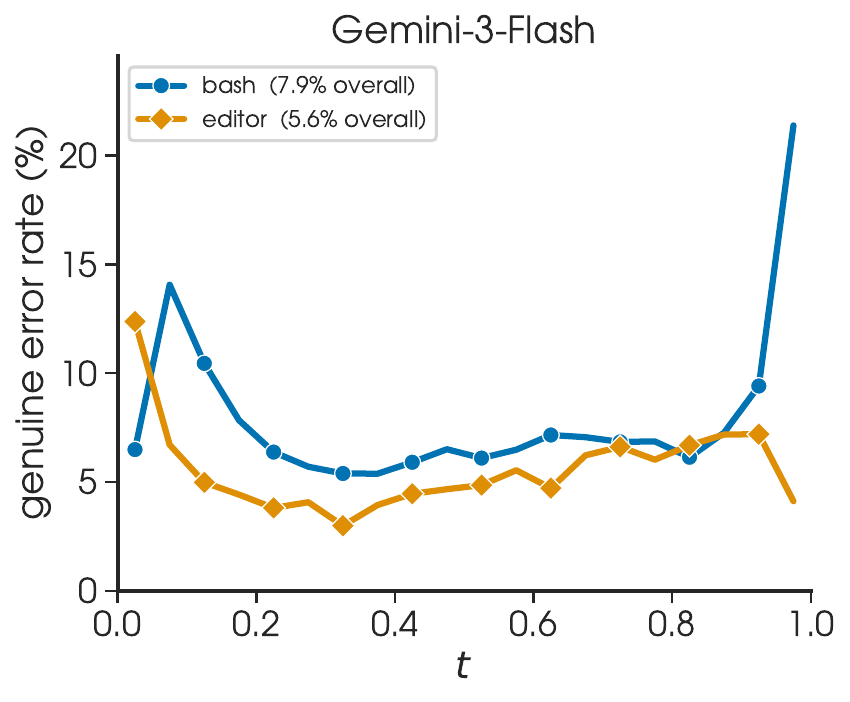} & \\
\includegraphics[width=0.31\linewidth]{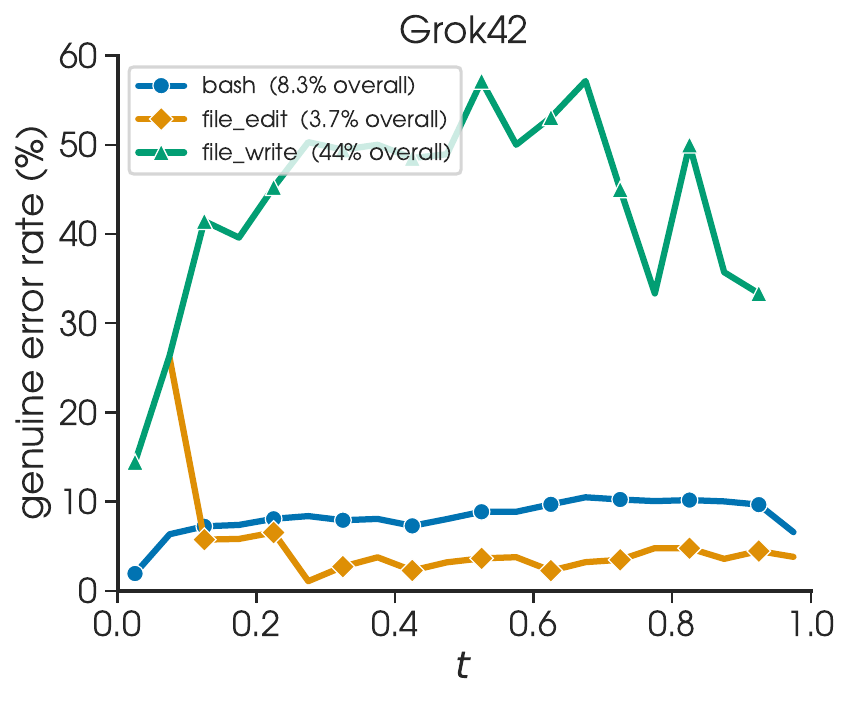} & & \\
\includegraphics[width=0.31\linewidth]{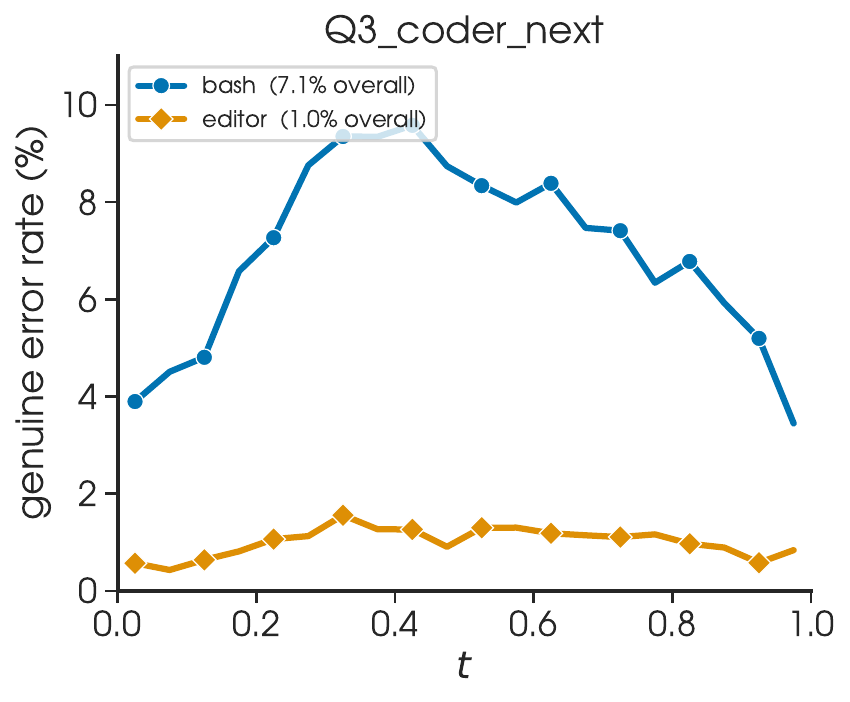} &
\includegraphics[width=0.31\linewidth]{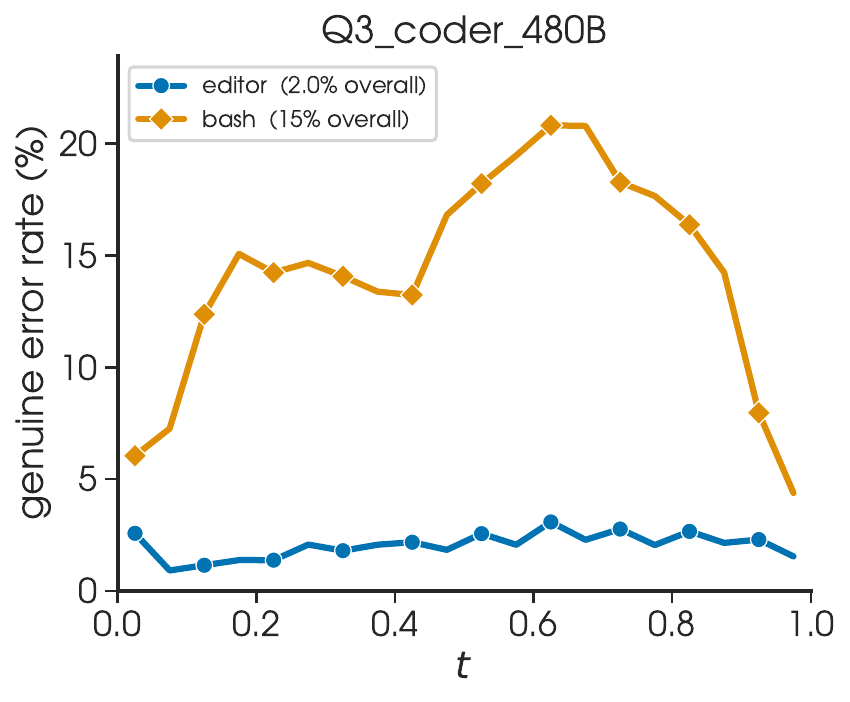} &
\includegraphics[width=0.31\linewidth]{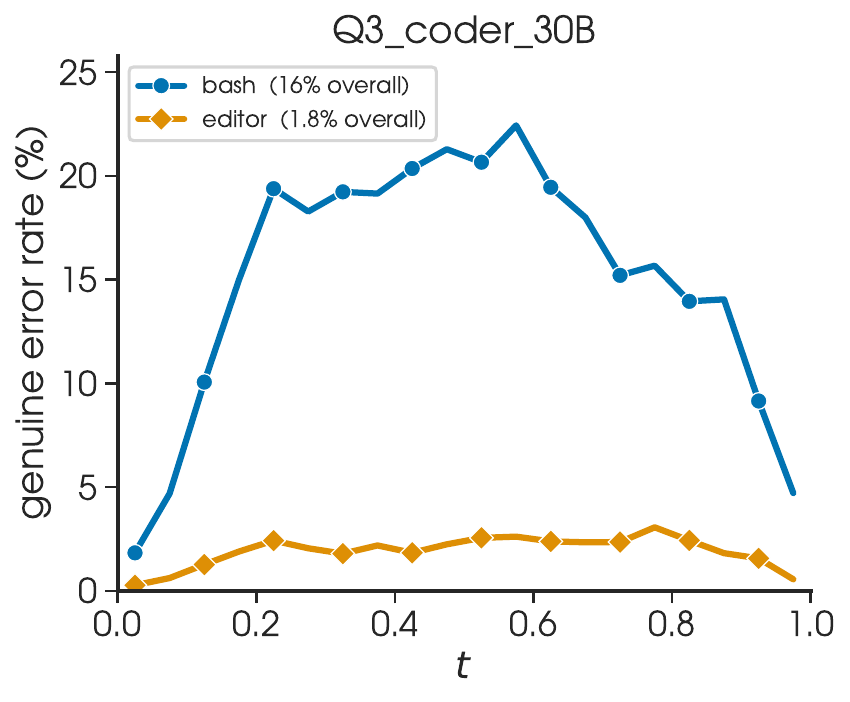} \\
\includegraphics[width=0.31\linewidth]{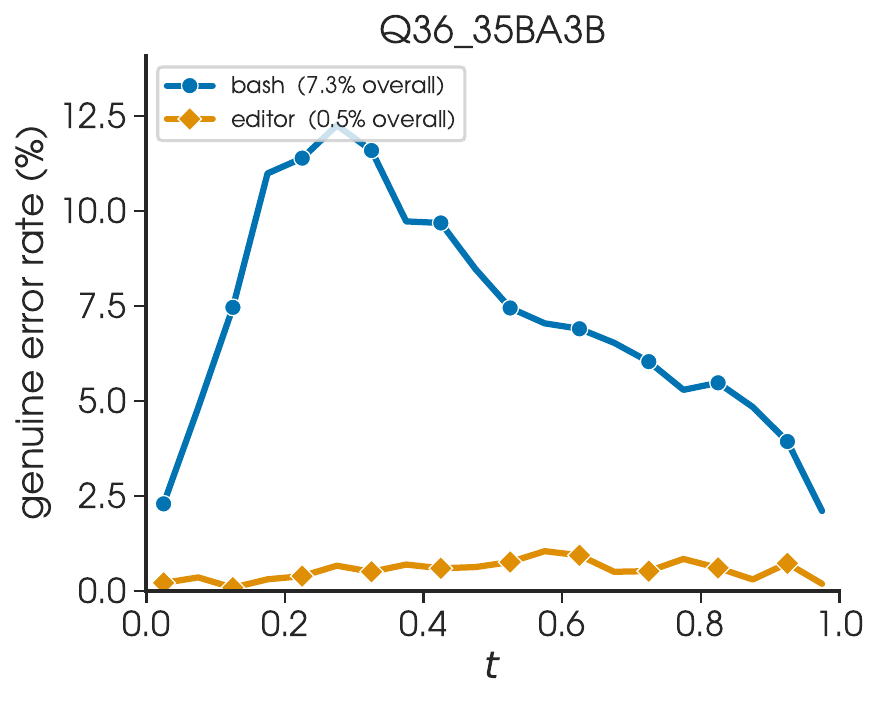} &
\includegraphics[width=0.31\linewidth]{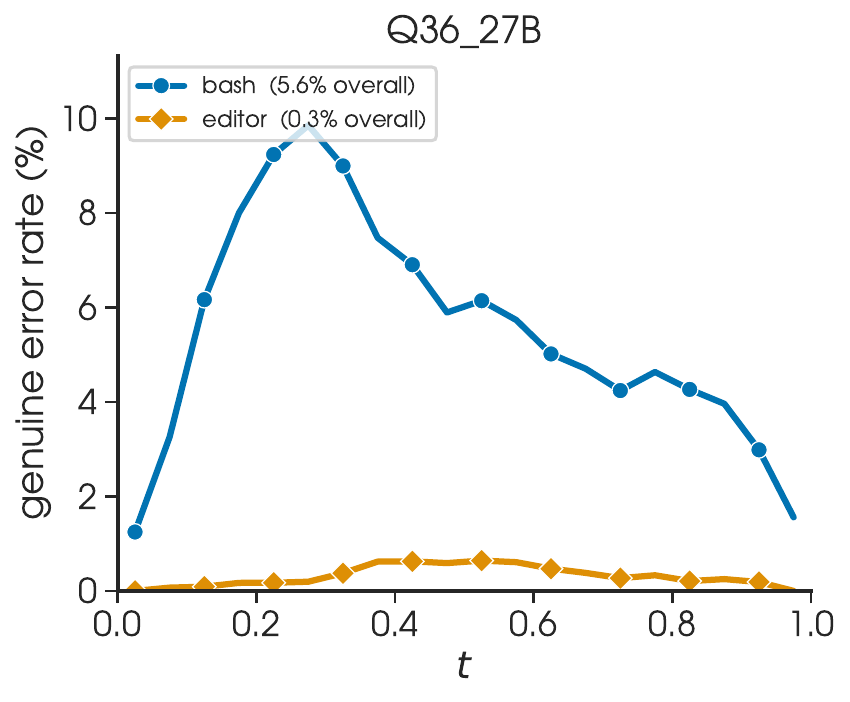} &
\includegraphics[width=0.31\linewidth]{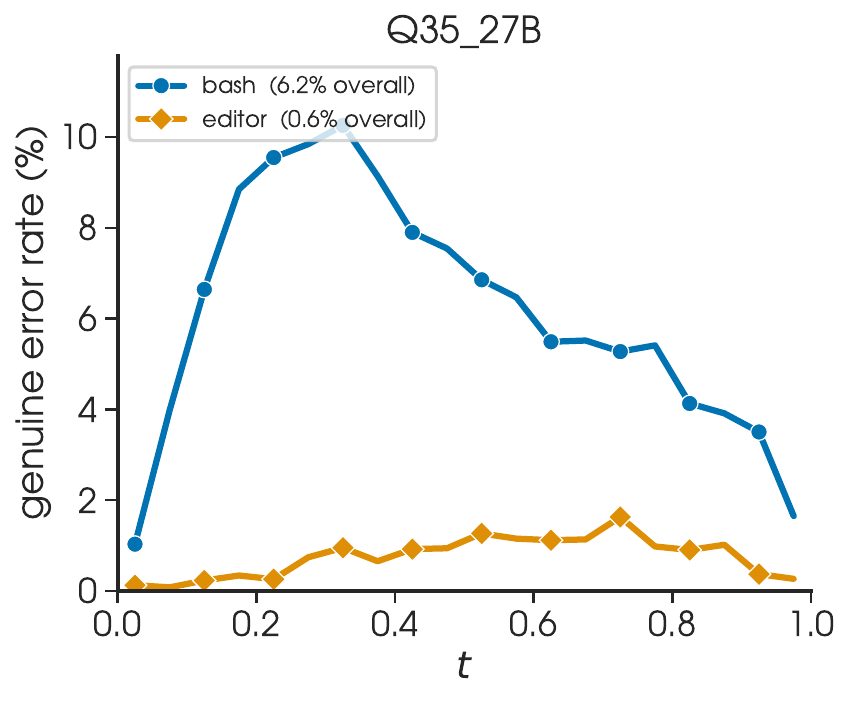} \\
\end{tabular}
\caption{Genuine tool-error rate per cycle on SWE-Bench-Pro (part 2 of 2): Gemini, Grok and Qwen families. Continued from Figure~\ref{fig:metrics-sbp-toolerr}.}
\label{fig:metrics-sbp-toolerr-b}
\end{figure}

\begin{figure}[p]
\centering
\includegraphics[width=0.65\linewidth]{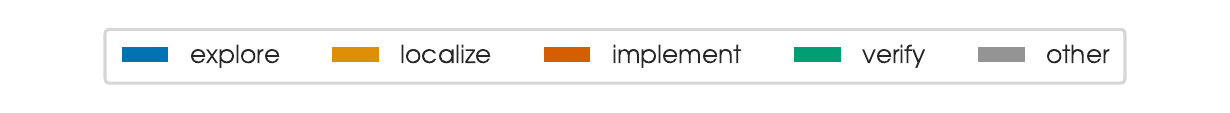}\\[2pt]
\setlength{\tabcolsep}{1pt}
\renewcommand{\arraystretch}{0.5}
\begin{tabular}{ccc}
\includegraphics[width=0.31\linewidth]{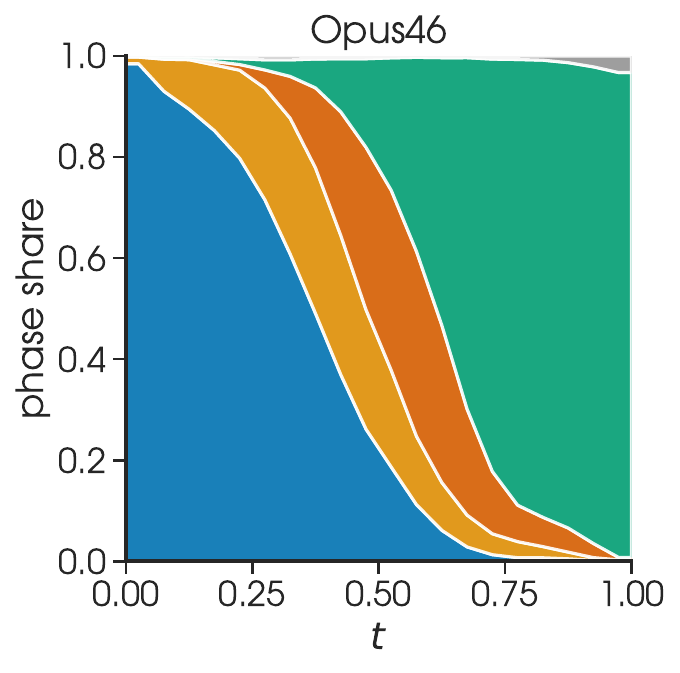} &
\includegraphics[width=0.31\linewidth]{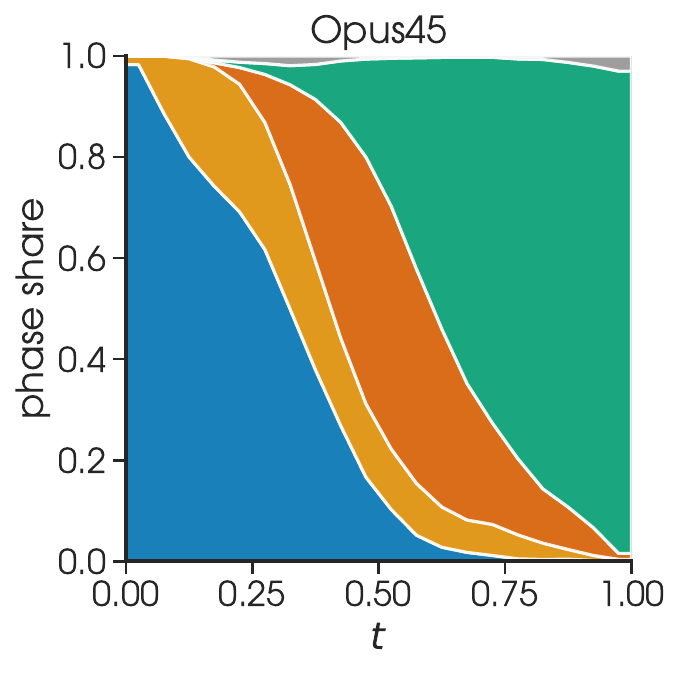} &
\includegraphics[width=0.31\linewidth]{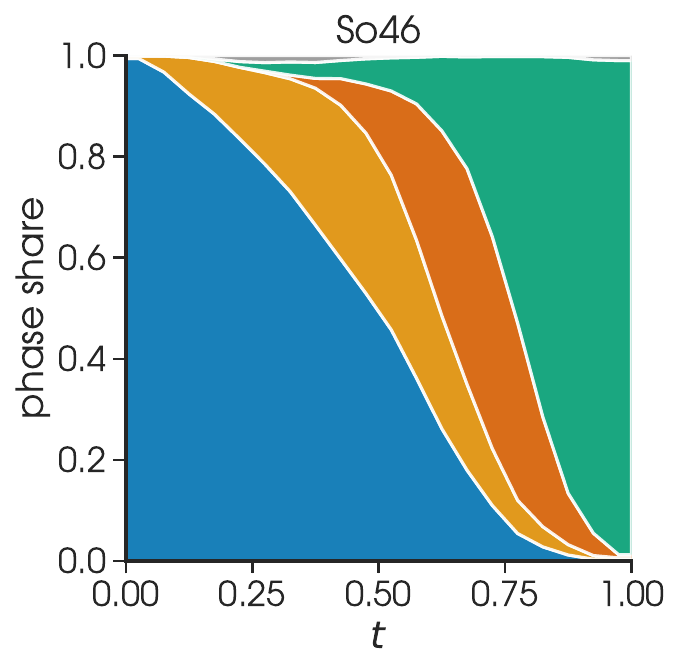} \\
\includegraphics[width=0.31\linewidth]{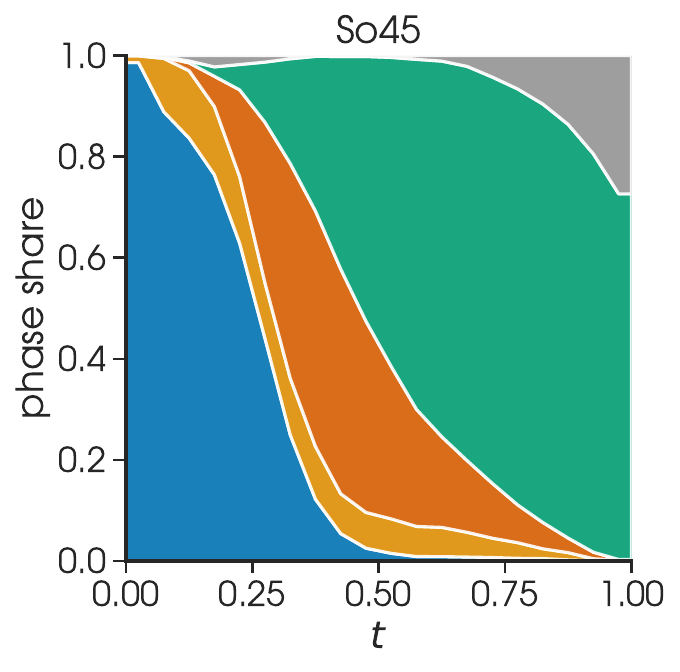} &
\includegraphics[width=0.31\linewidth]{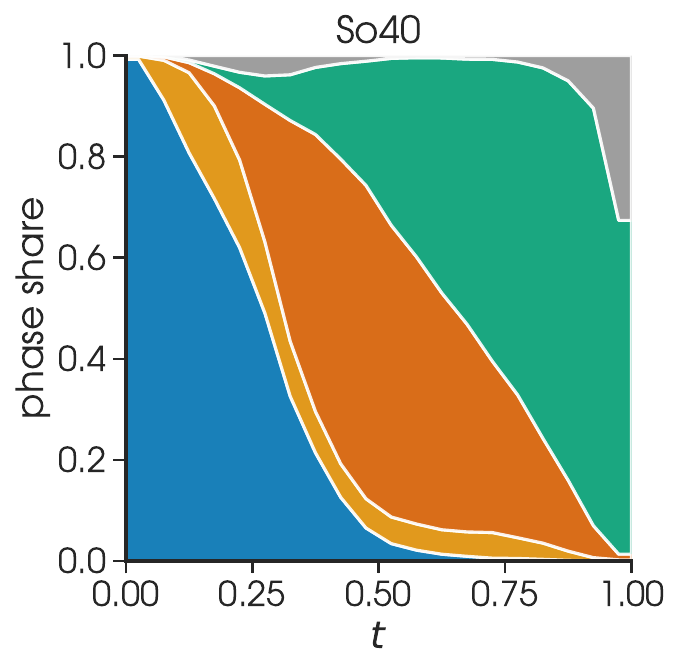} &
\includegraphics[width=0.31\linewidth]{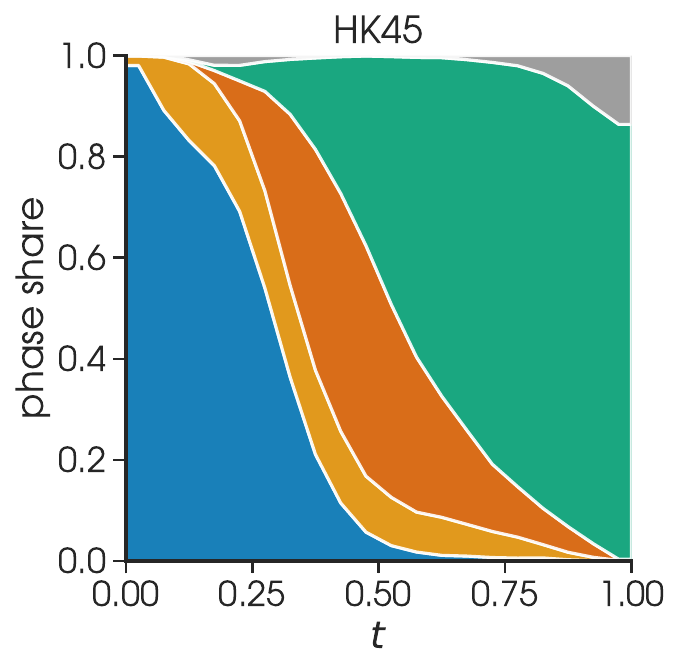} \\
\includegraphics[width=0.31\linewidth]{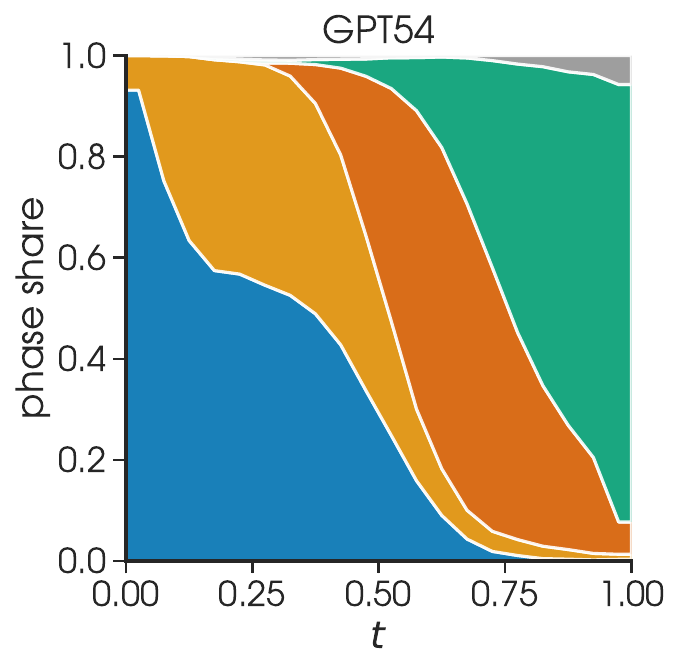} &
\includegraphics[width=0.31\linewidth]{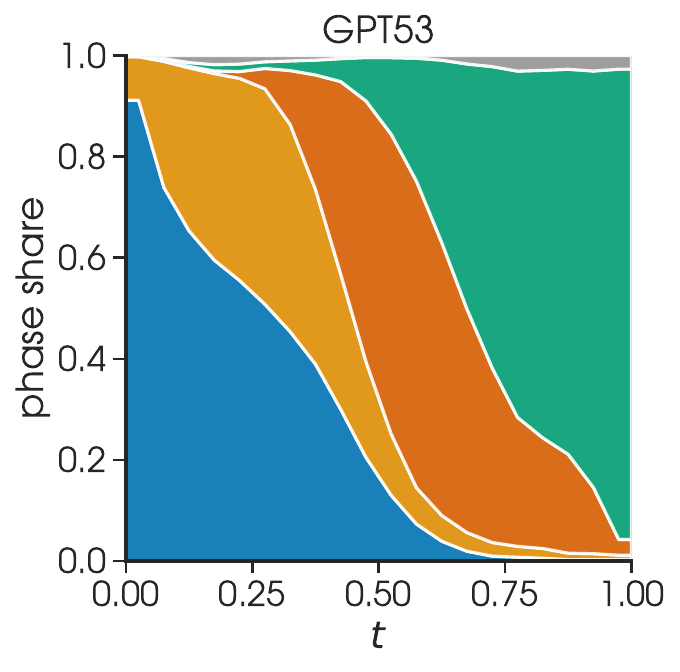} &
\includegraphics[width=0.31\linewidth]{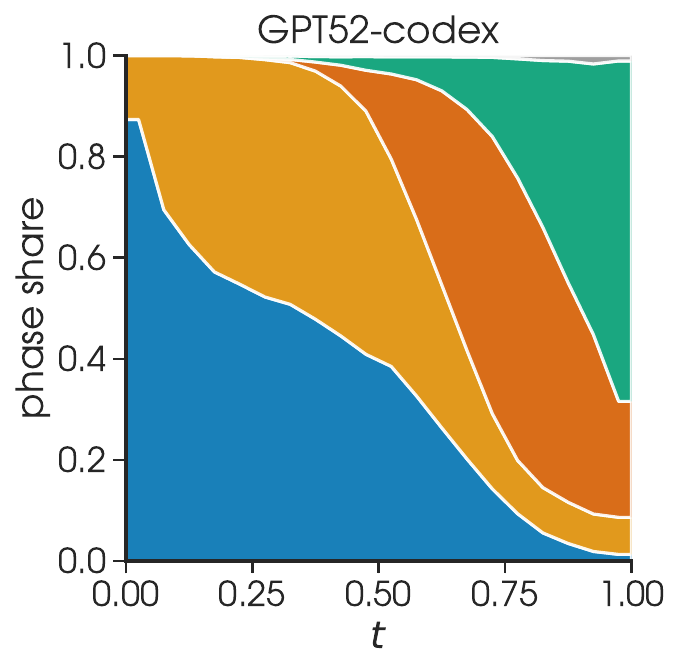} \\
\includegraphics[width=0.31\linewidth]{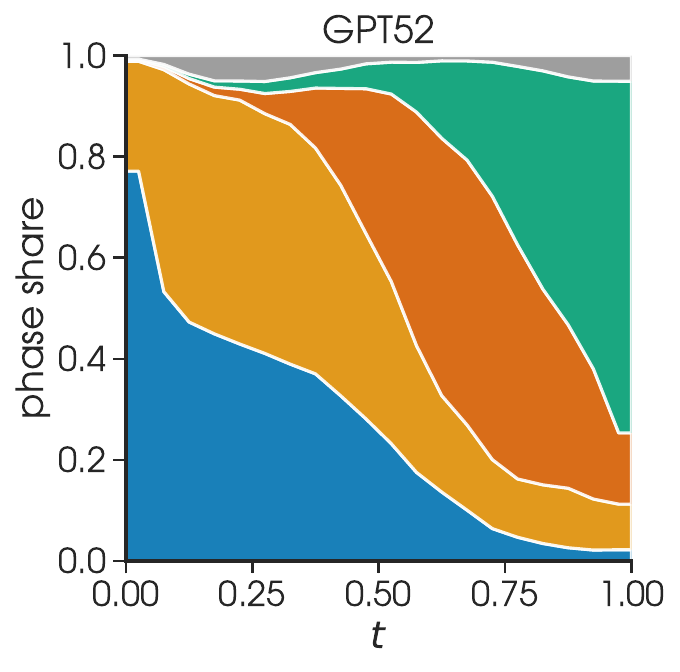} &
\includegraphics[width=0.31\linewidth]{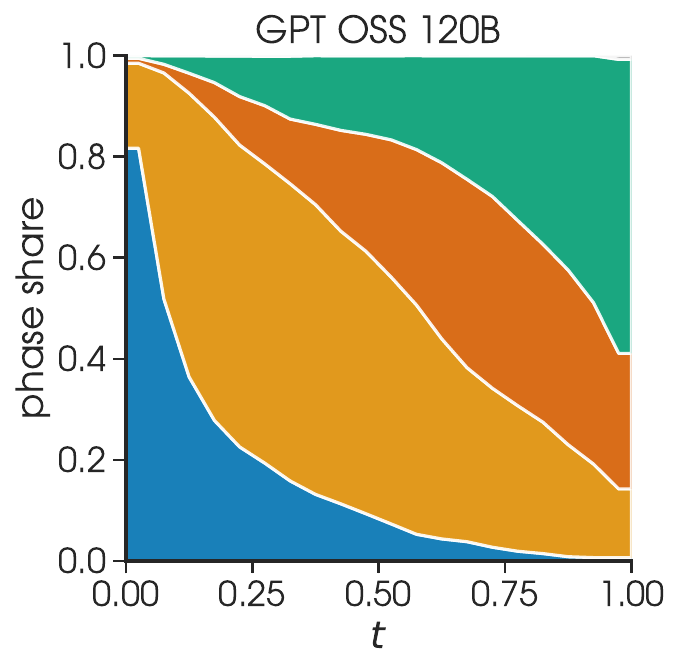} &
\includegraphics[width=0.31\linewidth]{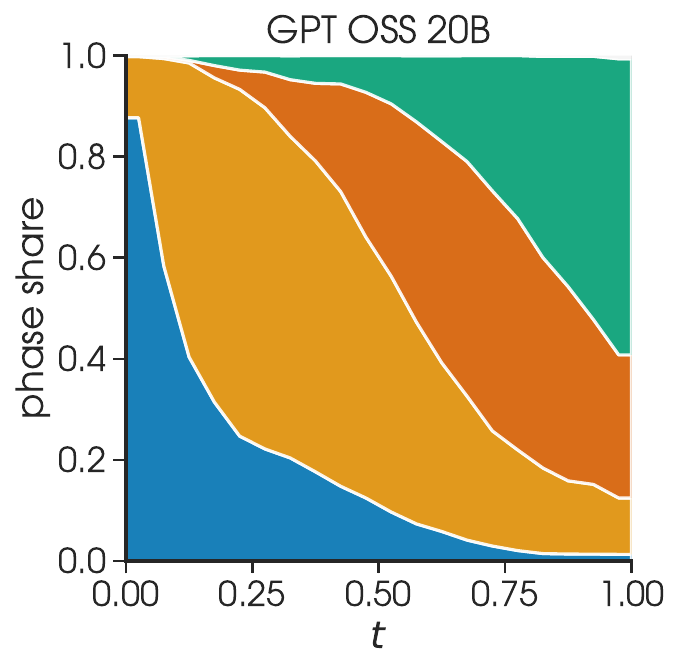} \\
\end{tabular}
\caption{\textbf{Phase composition surfaces model-specific schedules.} Per-cycle share of the four R6 (see LLM judge rubrics in Table\,\ref{tab:judge-buckets}) phases (\texttt{explore},
\texttt{localize}, \texttt{implement}, \texttt{verify}) vs.\ normalised cycle
position $t$ on SWE-Bench-Pro, for Claude and GPT families (part 1 of 2). Given the
same tasks, models schedule their cycles differently. The implementation peak shifts in position and width, and the 
verify phase carries different long-tail mass.}
\label{fig:metrics-sbp-phase}
\end{figure}

\begin{figure}[p]
\centering
\includegraphics[width=0.65\linewidth]{swe_pro/phase_legend.pdf}\\[2pt]
\setlength{\tabcolsep}{1pt}
\renewcommand{\arraystretch}{0.5}
\begin{tabular}{ccc}
\includegraphics[width=0.31\linewidth]{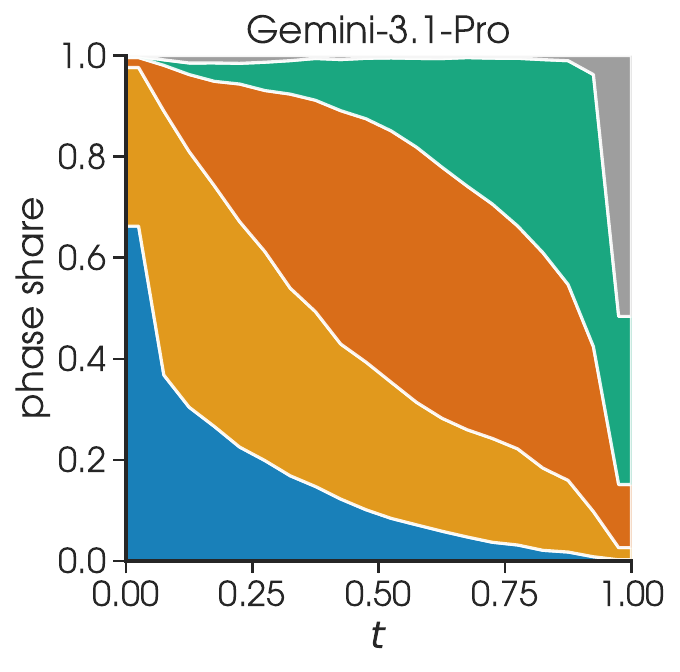} &
\includegraphics[width=0.31\linewidth]{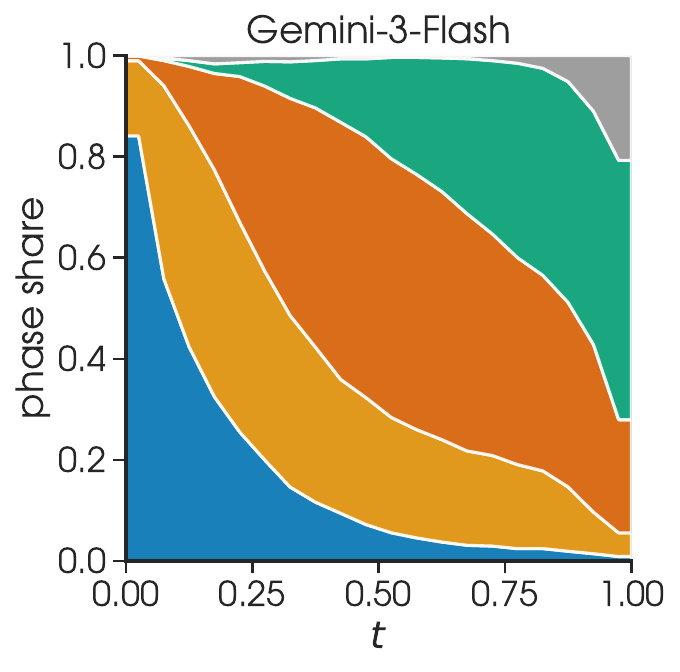} & \\
\includegraphics[width=0.31\linewidth]{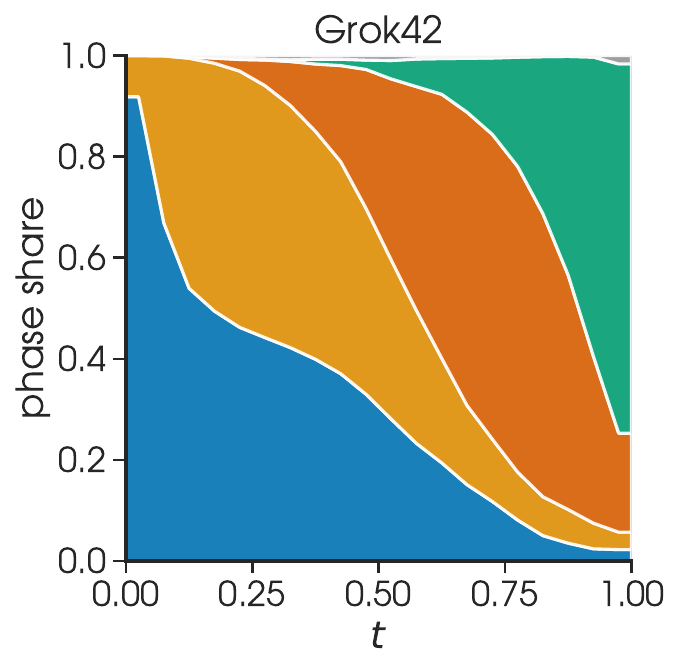} & & \\
\includegraphics[width=0.31\linewidth]{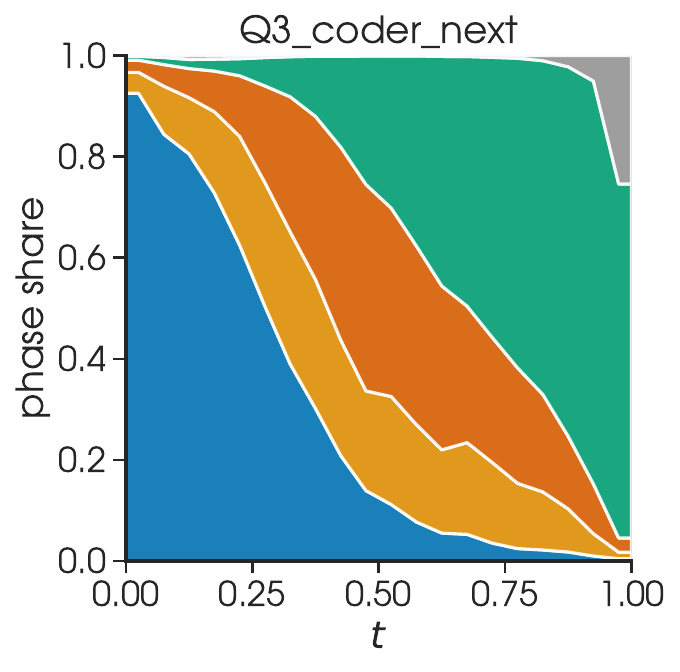} &
\includegraphics[width=0.31\linewidth]{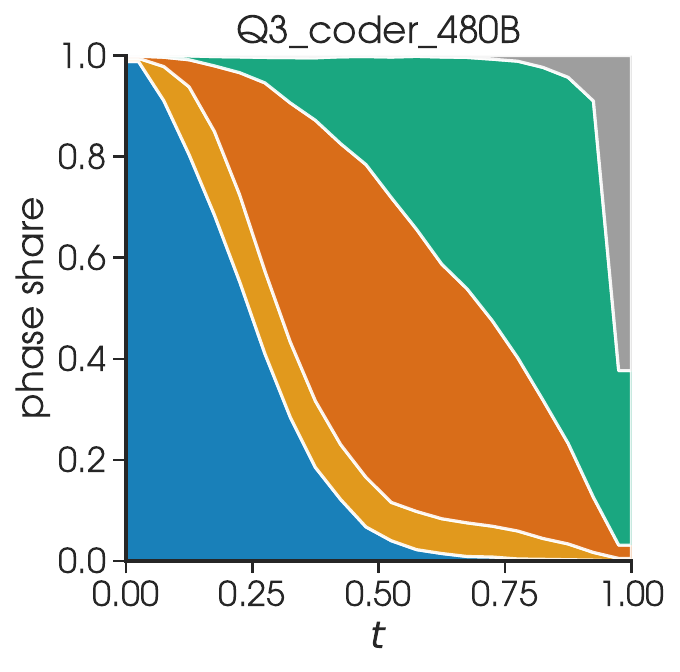} &
\includegraphics[width=0.31\linewidth]{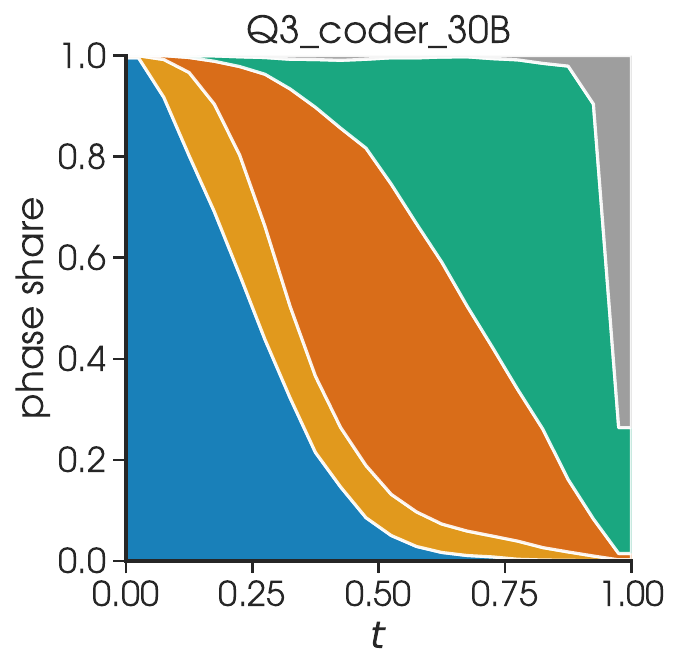} \\
\includegraphics[width=0.31\linewidth]{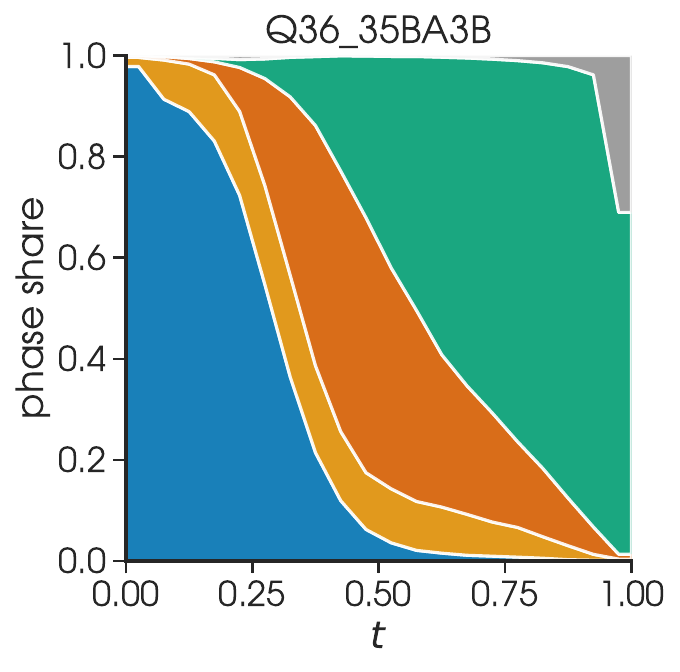} &
\includegraphics[width=0.31\linewidth]{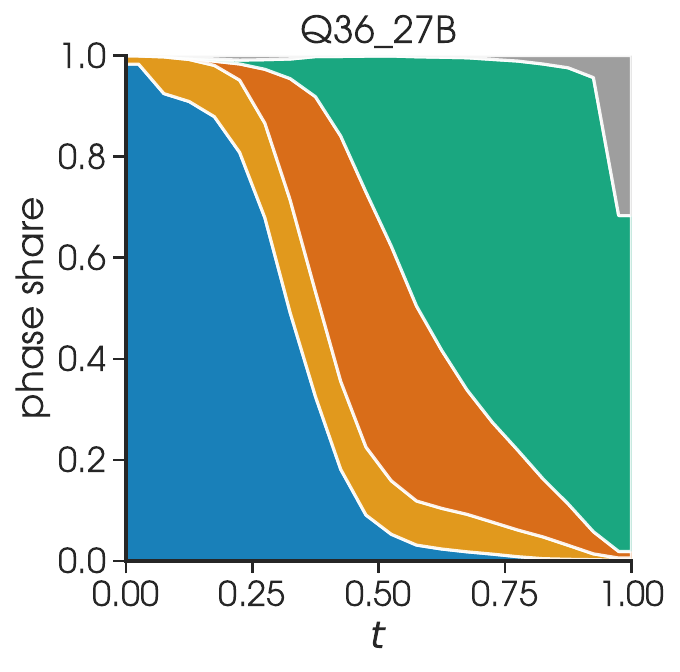} &
\includegraphics[width=0.31\linewidth]{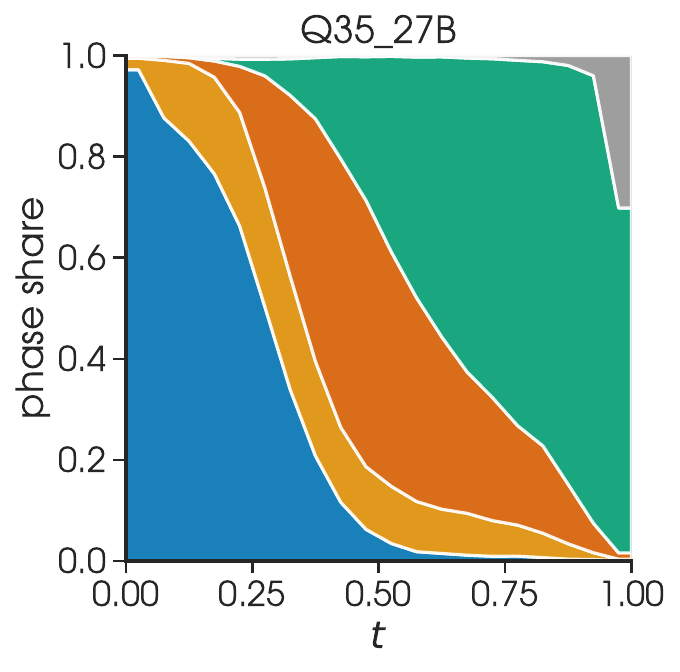} \\
\end{tabular}
\caption{Phase composition on SWE-Bench-Pro (part 2 of 2): Gemini, Grok and Qwen families. Continued from Figure~\ref{fig:metrics-sbp-phase}.}
\label{fig:metrics-sbp-phase-b}
\end{figure}

\begin{figure}[p]
\centering
\includegraphics[width=0.85\linewidth]{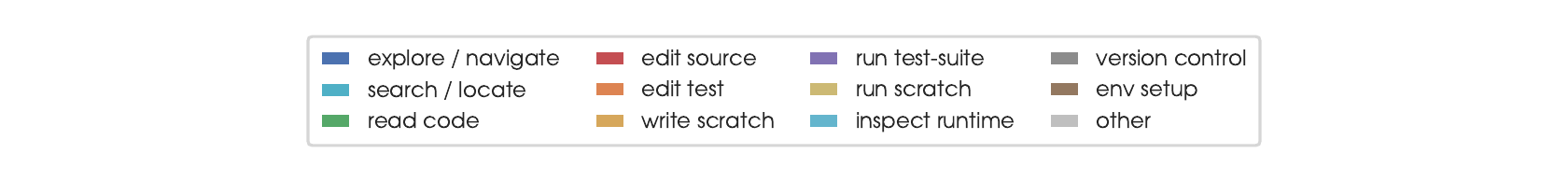}\\[2pt]
\setlength{\tabcolsep}{1pt}
\renewcommand{\arraystretch}{0.5}
\begin{tabular}{ccc}
\includegraphics[width=0.31\linewidth]{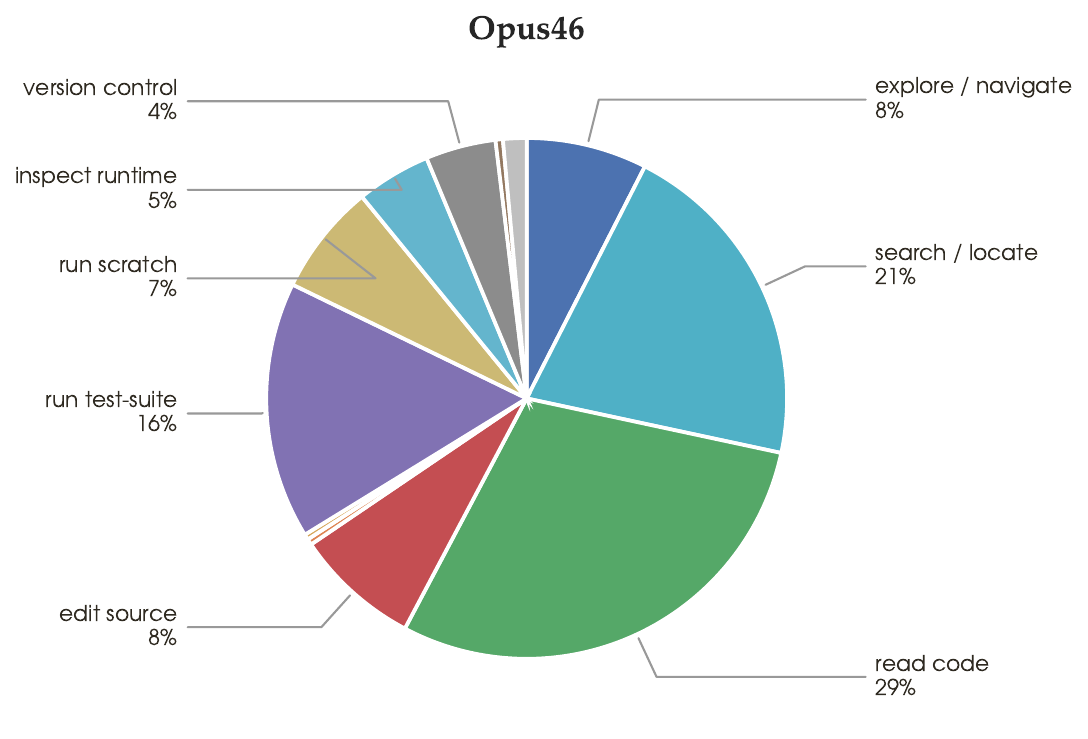} &
\includegraphics[width=0.31\linewidth]{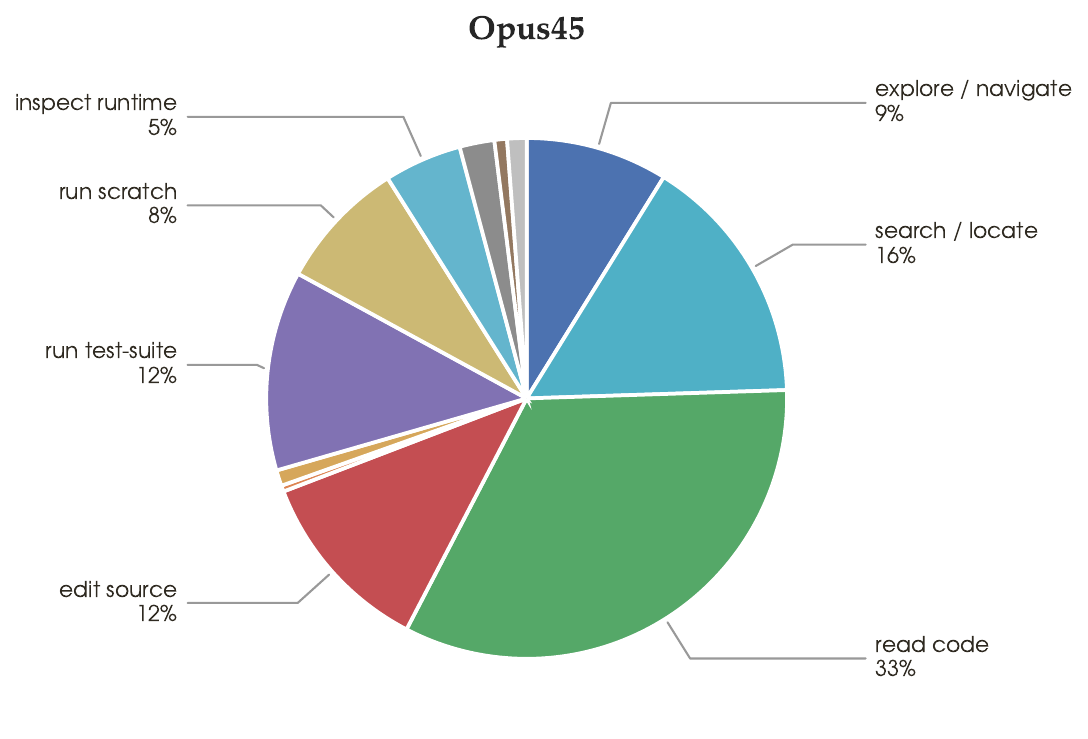} &
\includegraphics[width=0.31\linewidth]{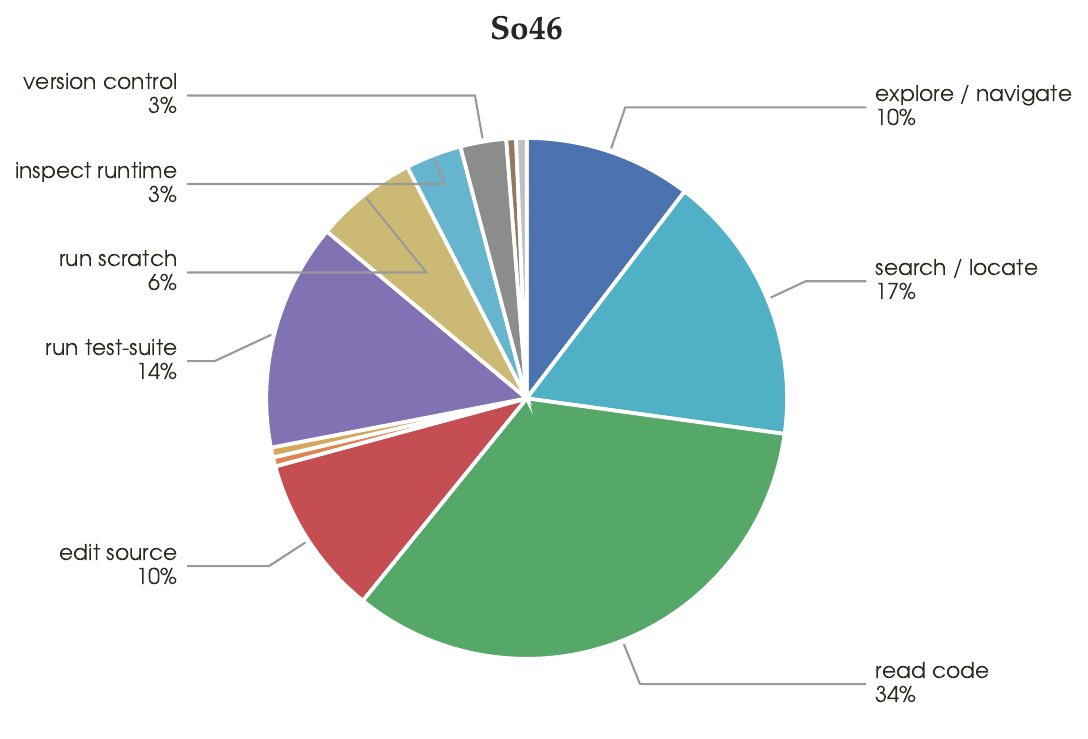} \\
\includegraphics[width=0.31\linewidth]{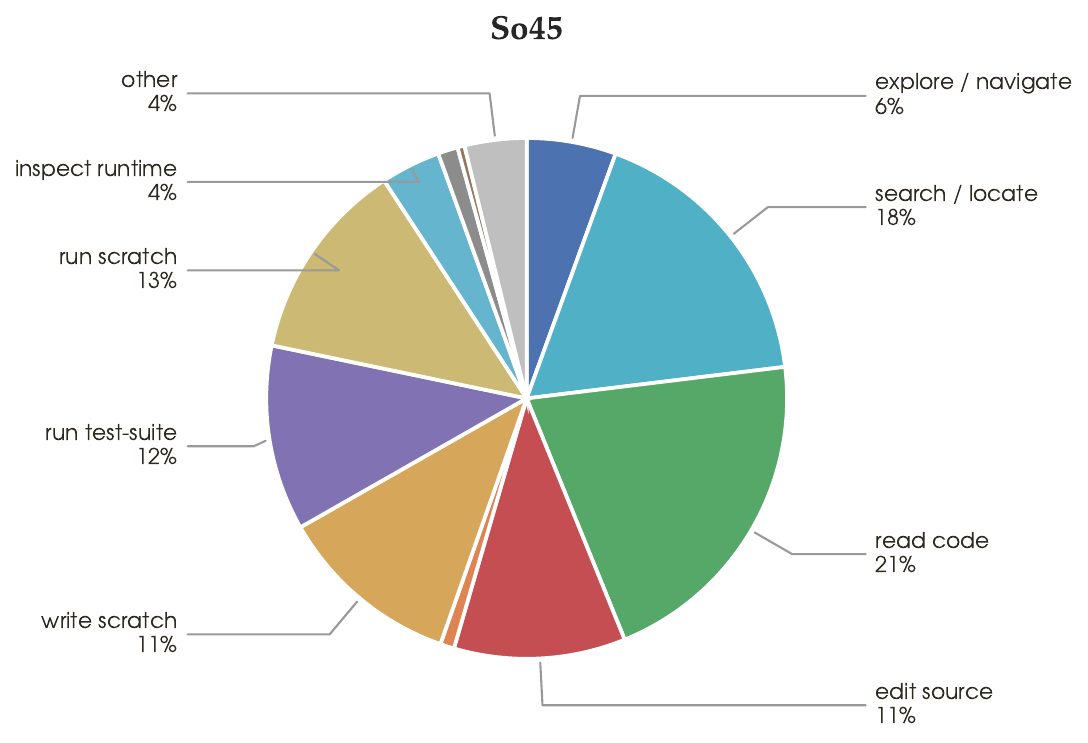} &
\includegraphics[width=0.31\linewidth]{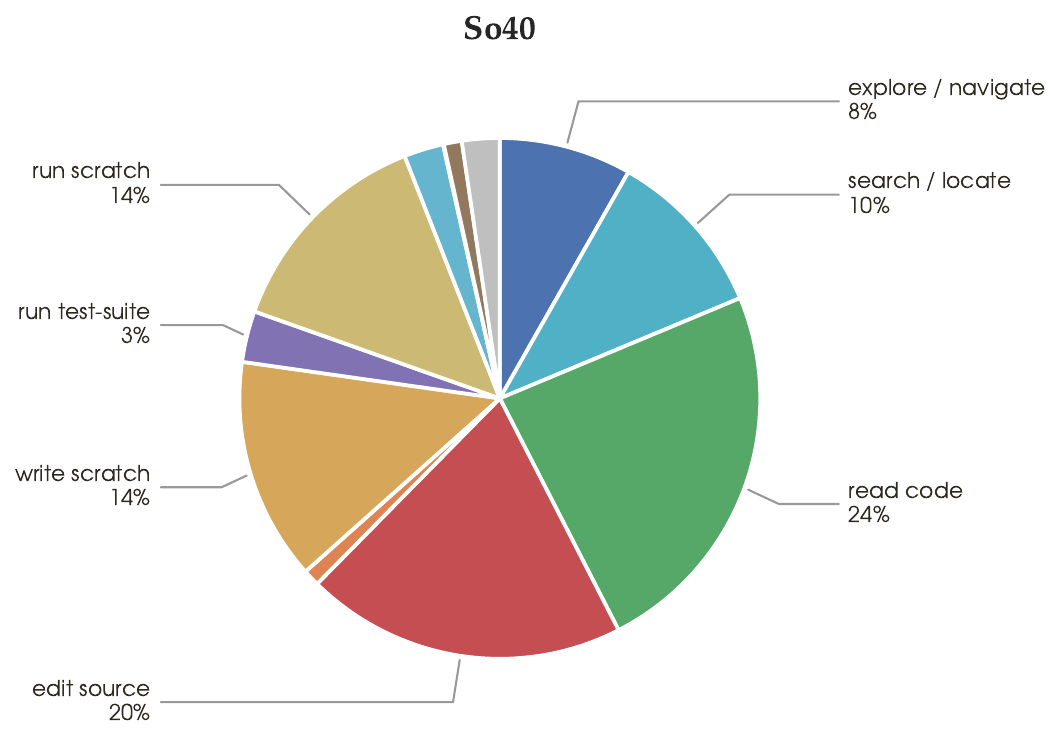} &
\includegraphics[width=0.31\linewidth]{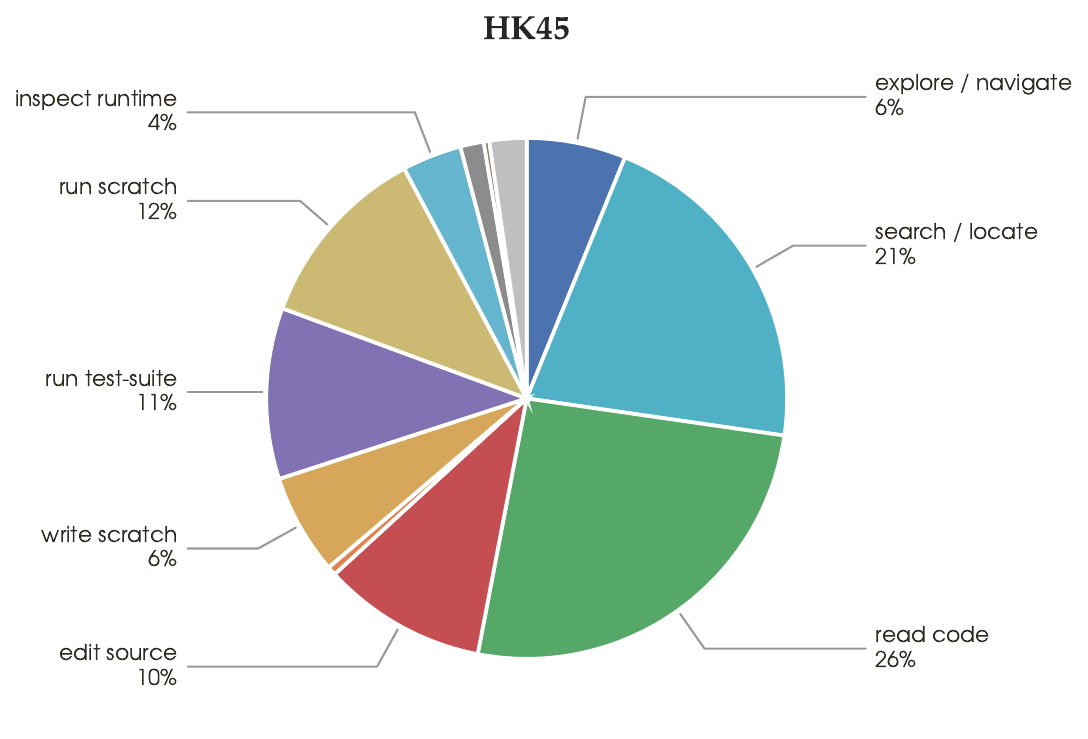} \\
\includegraphics[width=0.31\linewidth]{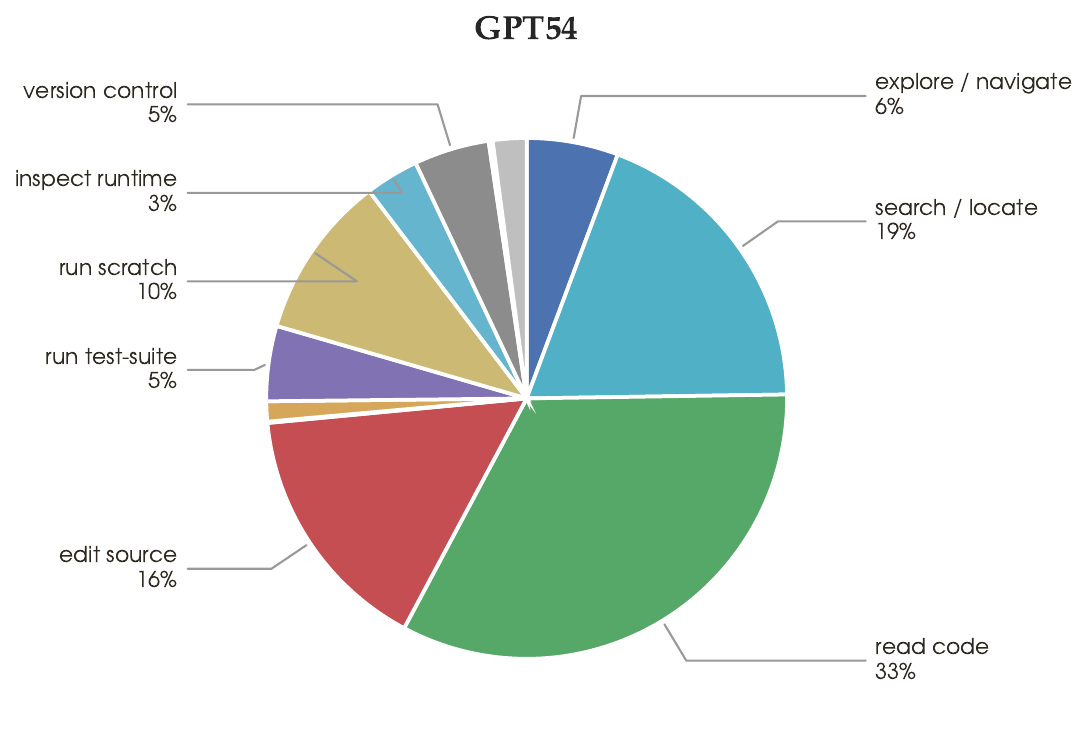} &
\includegraphics[width=0.31\linewidth]{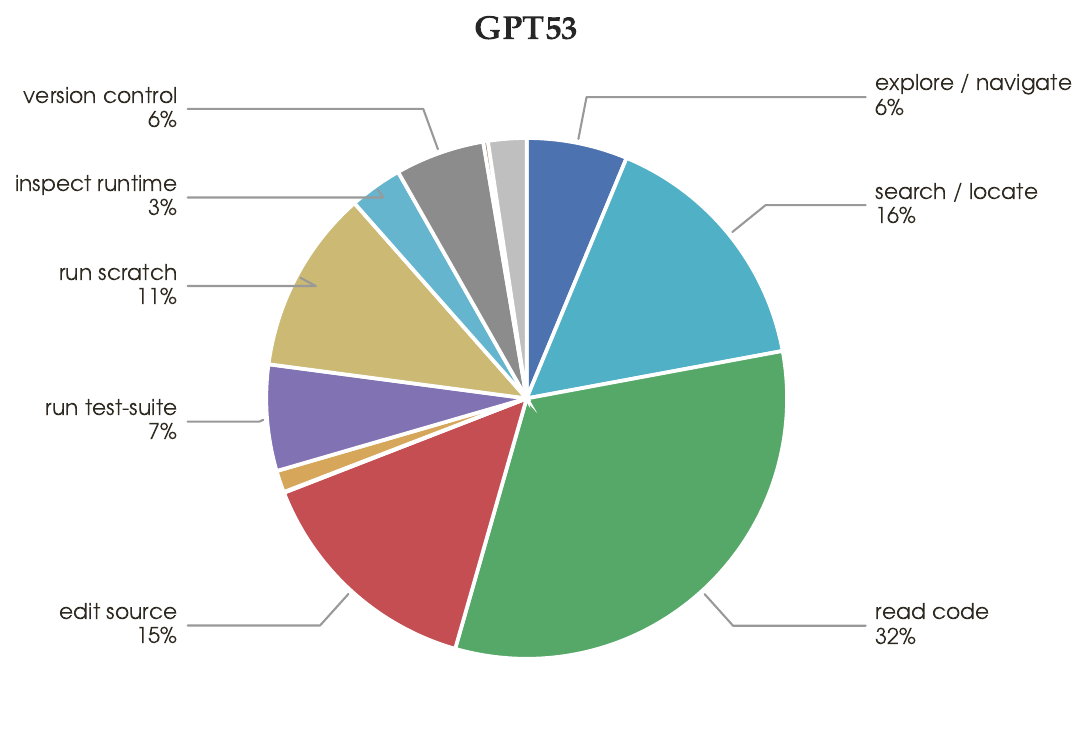} &
\includegraphics[width=0.31\linewidth]{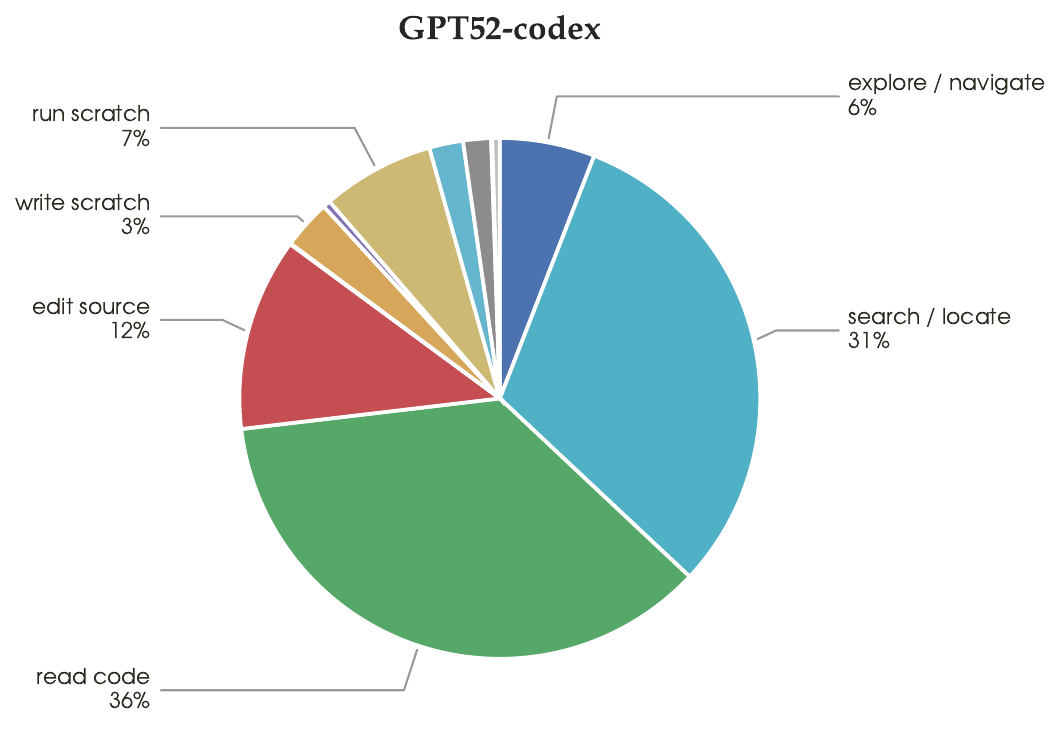} \\
\includegraphics[width=0.31\linewidth]{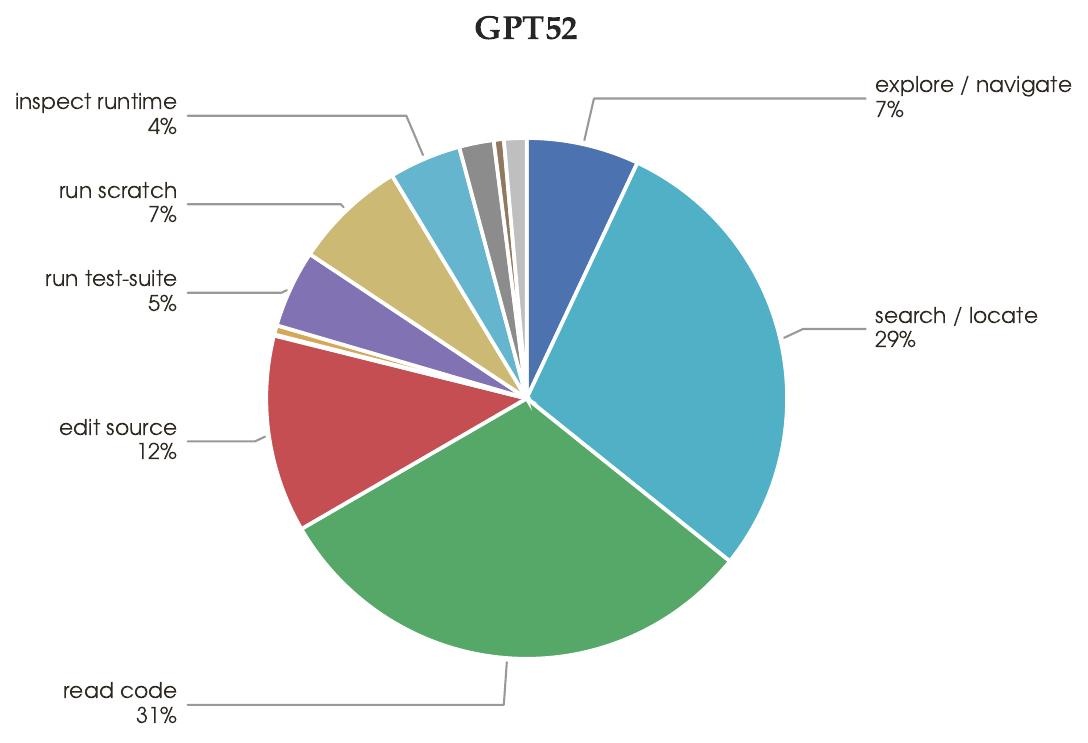} &
\includegraphics[width=0.31\linewidth]{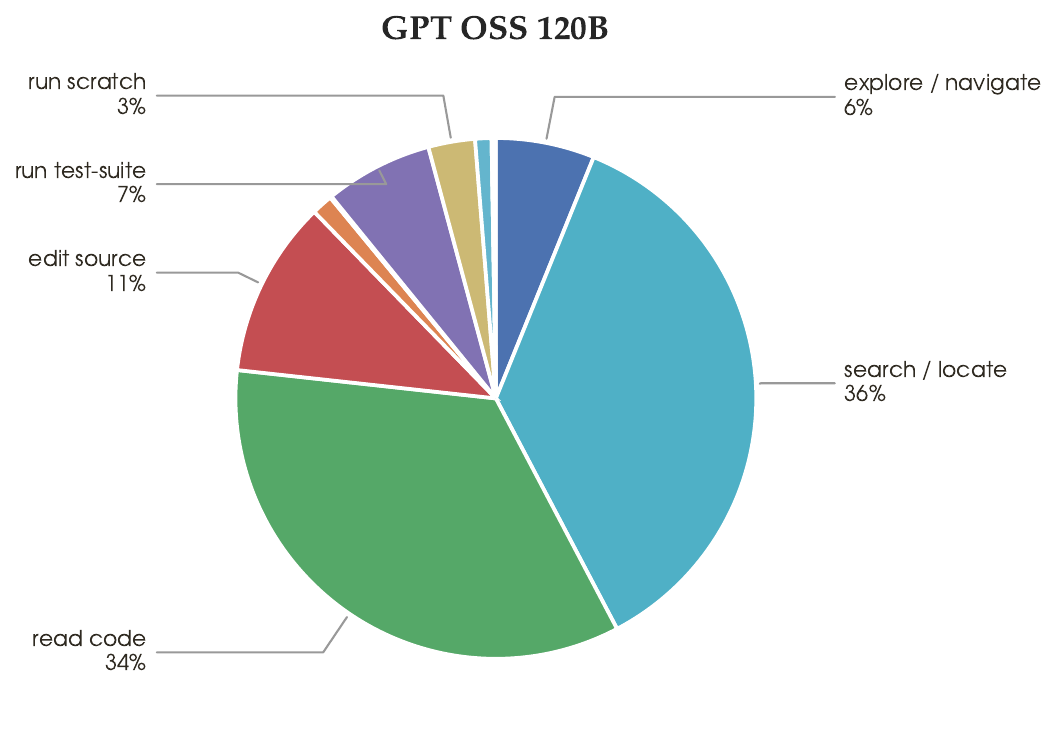} &
\includegraphics[width=0.31\linewidth]{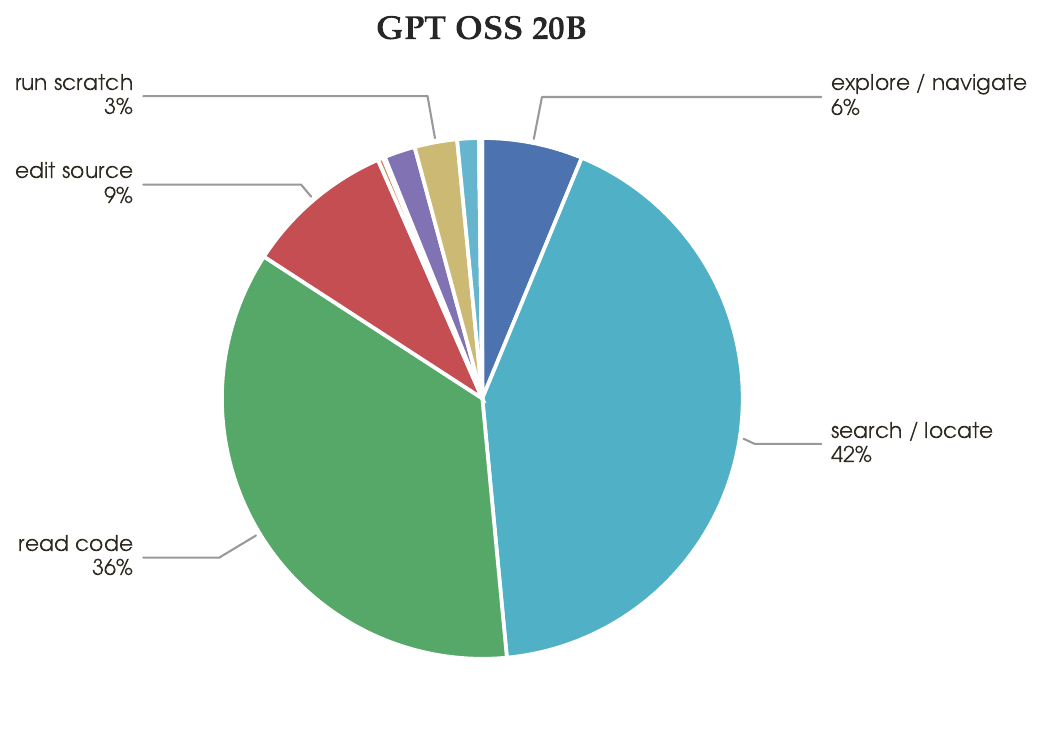} \\
\end{tabular}
\caption{Tool-call distribution per model on SWE-Bench-Pro (part 1 of 2): Anthropic Claude and OpenAI families. Each cell shows the share of calls in each R1 category (see Table\,\ref{tab:judge-buckets} for LLM judge rubrics) for one model.}
\label{fig:metrics-sbp-tooldist}
\end{figure}

\begin{figure}[p]
\centering
\includegraphics[width=0.85\linewidth]{swe_pro/tool_distribution_legend.pdf}\\[2pt]
\setlength{\tabcolsep}{1pt}
\renewcommand{\arraystretch}{0.5}
\begin{tabular}{ccc}
\includegraphics[width=0.31\linewidth]{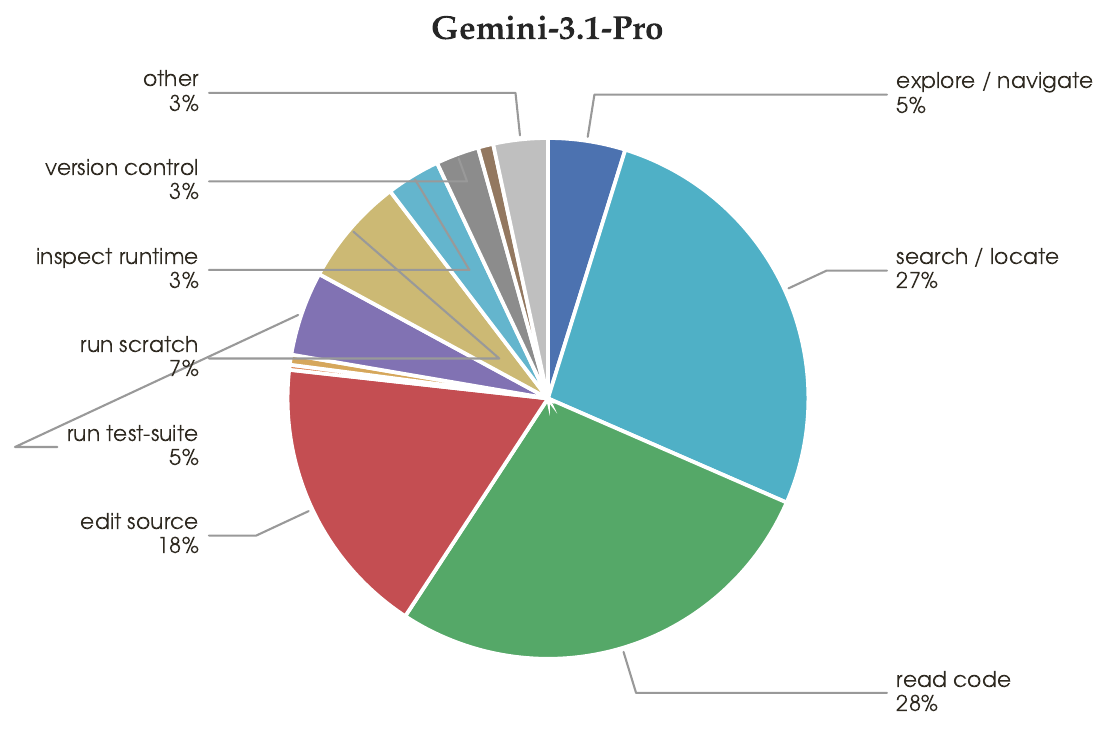} &
\includegraphics[width=0.31\linewidth]{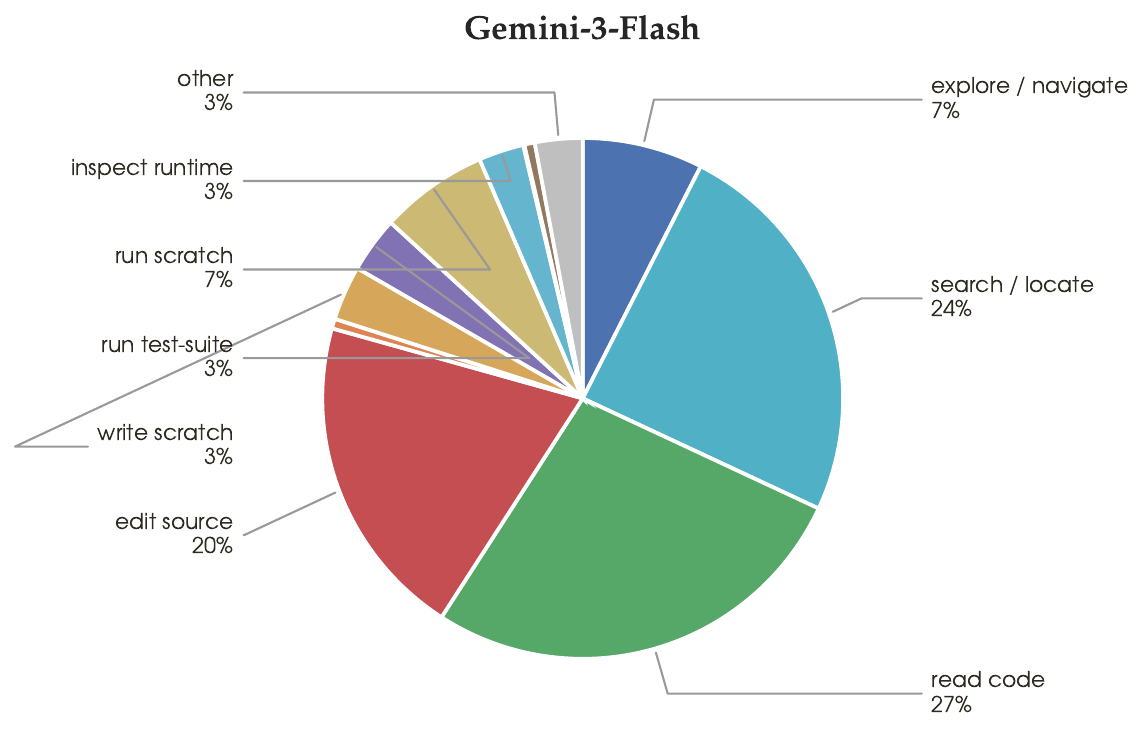} & \\
\includegraphics[width=0.31\linewidth]{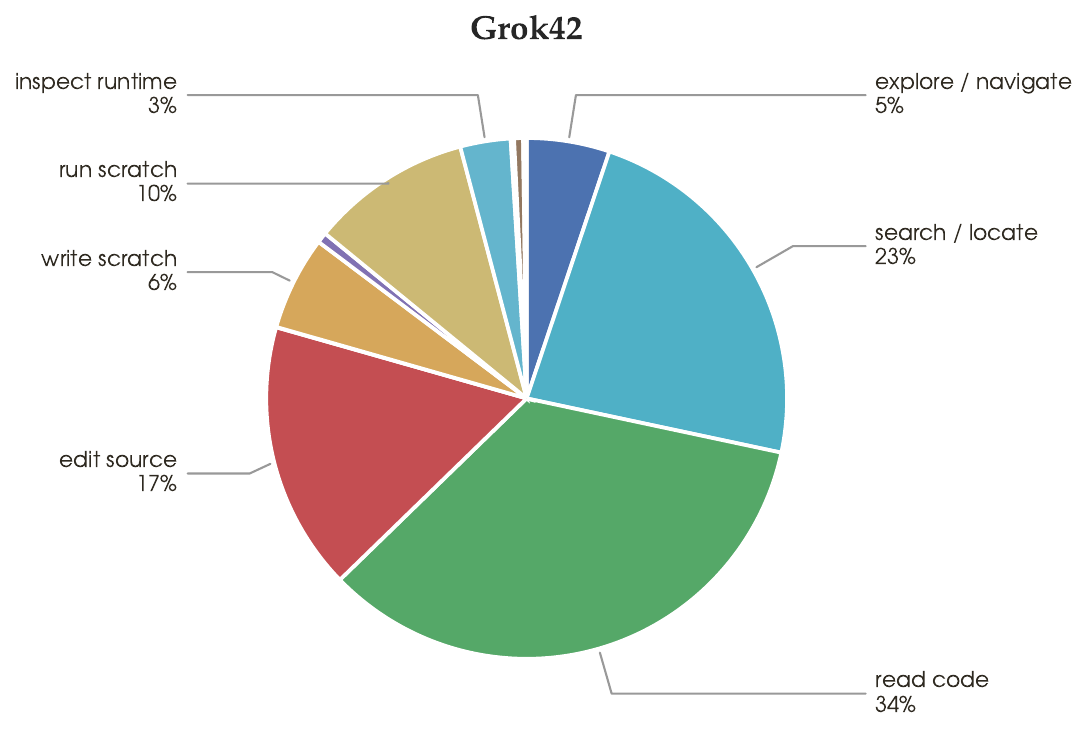} & & \\
\includegraphics[width=0.31\linewidth]{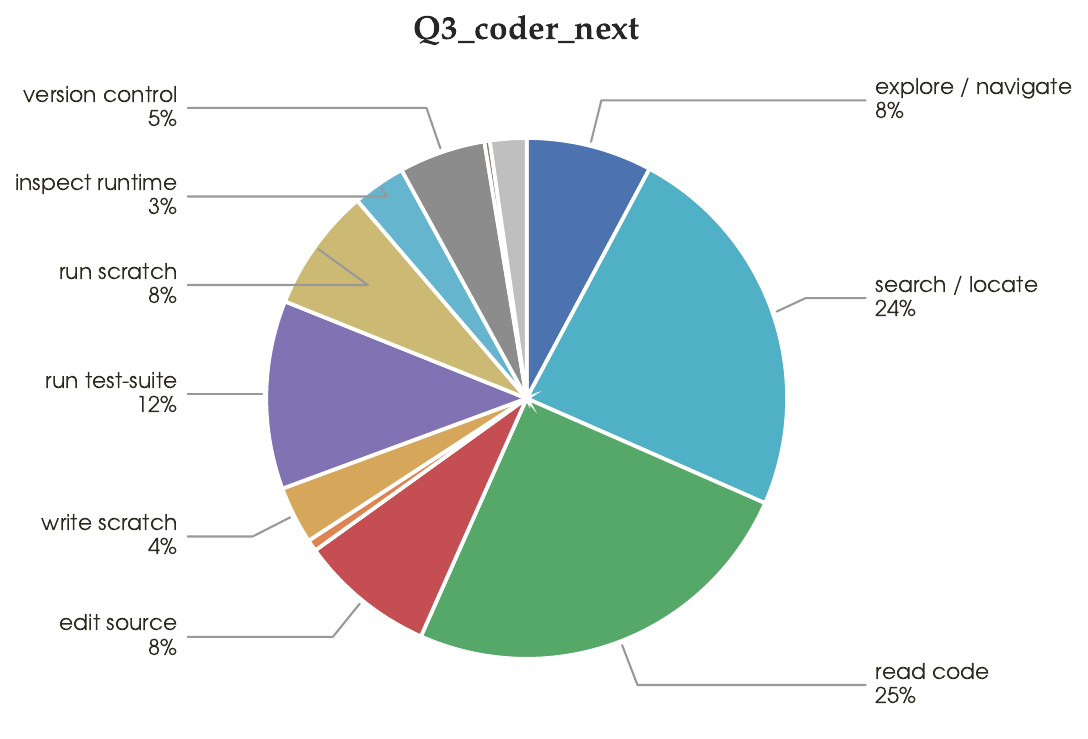} &
\includegraphics[width=0.31\linewidth]{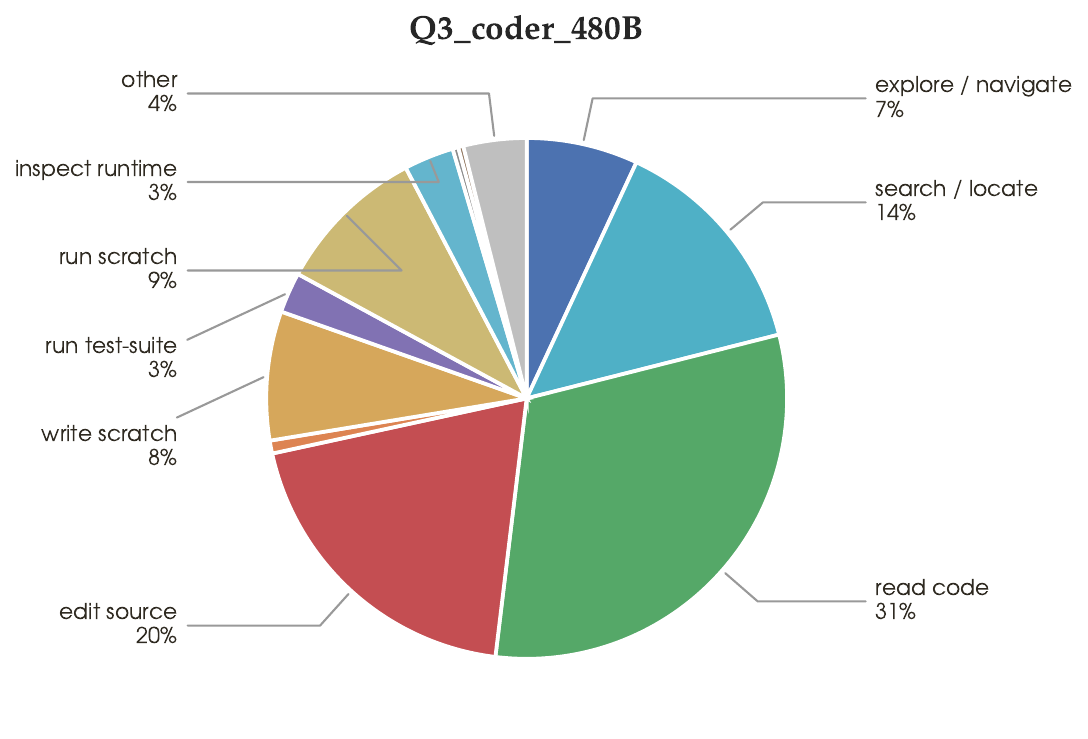} &
\includegraphics[width=0.31\linewidth]{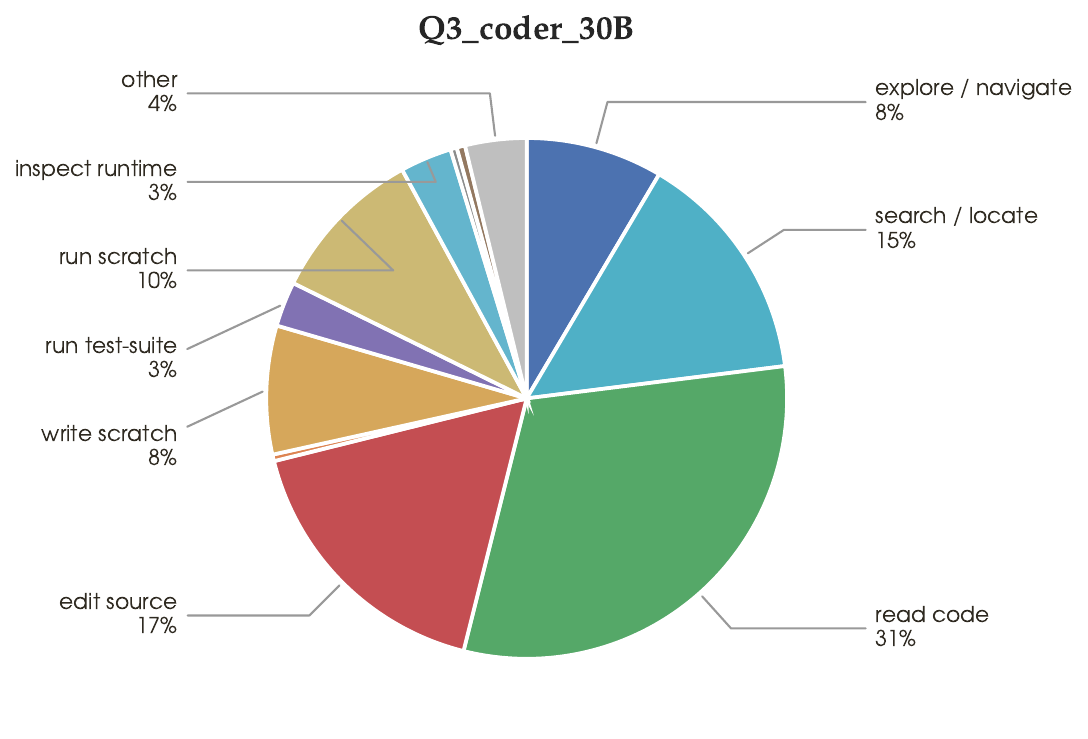} \\
\includegraphics[width=0.31\linewidth]{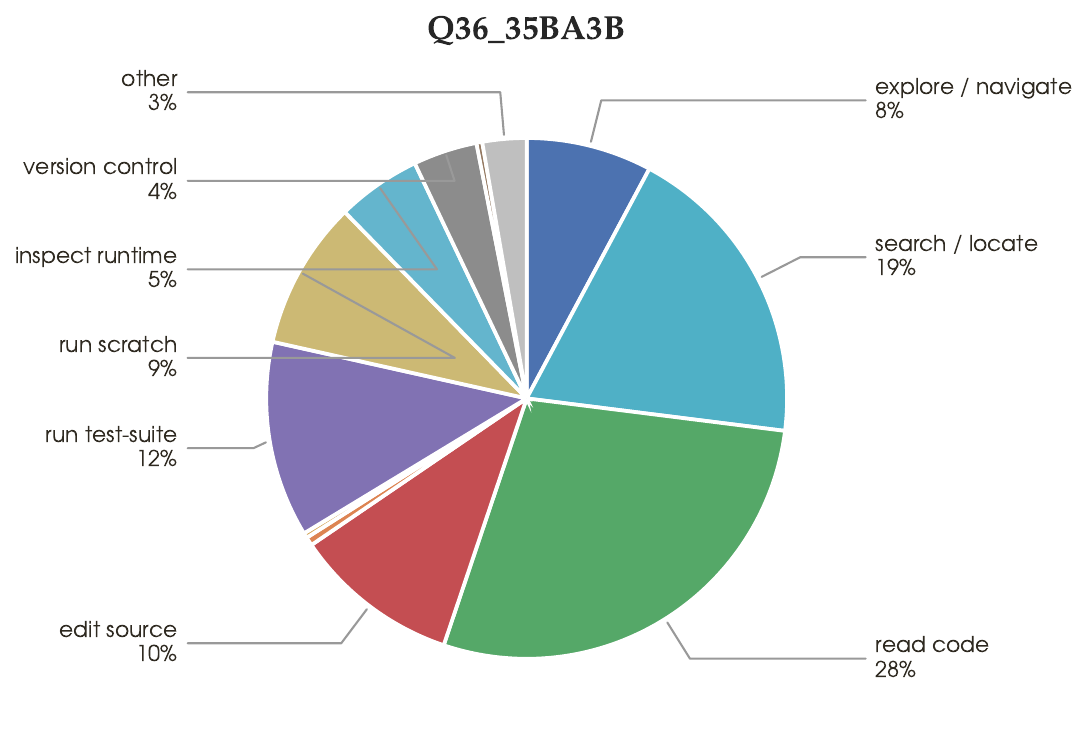} &
\includegraphics[width=0.31\linewidth]{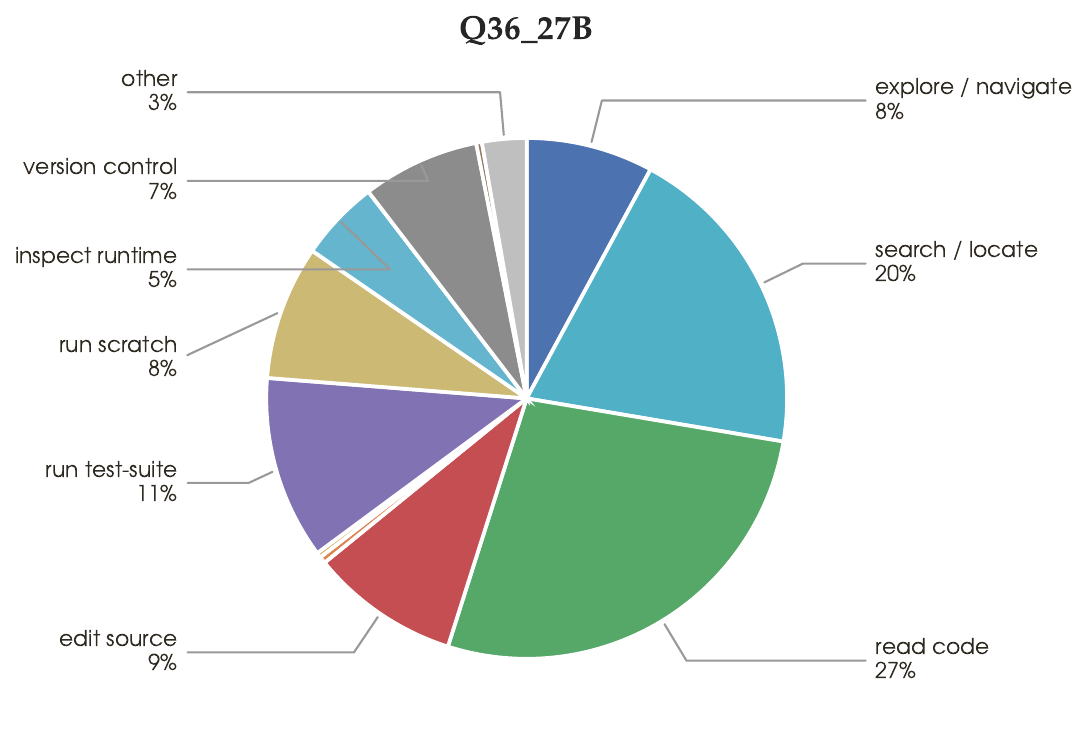} &
\includegraphics[width=0.31\linewidth]{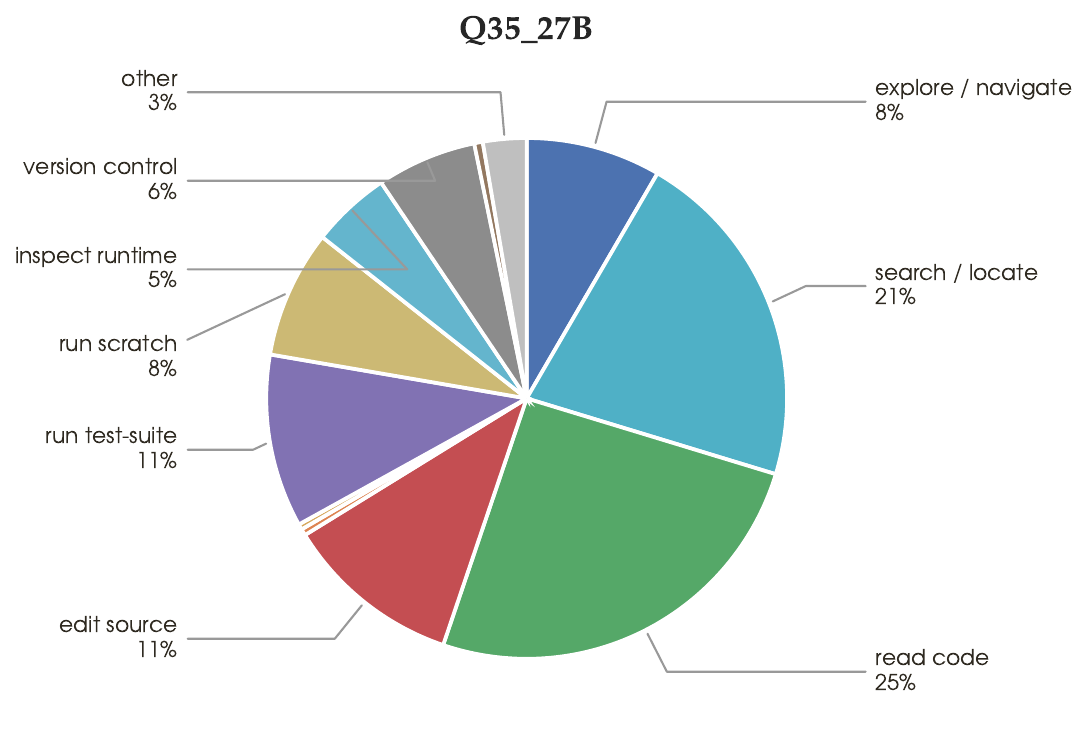} \\
\end{tabular}
\caption{Tool-call distribution per model on SWE-Bench-Pro (part 2 of 2): Gemini, Grok and Qwen families. Continued from Figure~\ref{fig:metrics-sbp-tooldist}.}
\label{fig:metrics-sbp-tooldist-b}
\end{figure}

Figures~\ref{fig:metrics-sbp-backtracking}--\ref{fig:metrics-sbp-tooldist-b} show the same six per-model behavioral metrics on SWE-Bench-Pro that Section~\ref{app:metrics:sbv} reports for SWE-Bench-Verified. Note that all trajectories were for the Base public containers and not the Sanitized SBP trajectories, so per-model breakdowns in this subsection inherit the Base minus Sanitized caveat of Appendix~\ref{app:swe-pro-git-leakage}.

\paragraph{Per-model backtracking distributions (SBP).}
Figure~\ref{fig:metrics-sbp-backtracking}--\ref{fig:metrics-sbp-backtracking-b} show the per-edit $\Delta D$ histogram for each model on SBP: green bars are progress edits ($\Delta D < 0$), red bars are backtracking edits ($\Delta D > 0$); the annotated percentage is the per-model backtrack-edit share.

\paragraph{Per-model edit/read ratio (SBP).}
Figure~\ref{fig:metrics-sbp-editread}--\ref{fig:metrics-sbp-editread-b} shows the per-cycle ratio of edit-oriented tool calls to total
ones across resolved + unresolved trajectories. A high ratio means the agent mostly edits and a low ratio means it mostly explores/test.

\paragraph{Per-model level families (SBP).}
Figures~\ref{fig:metrics-sbp-soldist}--\ref{fig:metrics-sbp-soldist-b} show the level family, mean $\bar D(t)$ and backtracking term $\bar B(t)$ of the resolved trajectories for each model on SBP, in the format of Figure~\ref{fig:metrics-sbv-soldist}.

\paragraph{Per-model genuine-error rate over cycles (SBP).}
Figure~\ref{fig:metrics-sbp-toolerr}--\ref{fig:metrics-sbp-toolerr-b} plot the rate of genuine tool errors per cycle (R8 $=$ \texttt{genuine}, see Section~\ref{app:metrics:rubric}), broken down by tool family.

\paragraph{Per-model phase composition (SBP).}
Figure~\ref{fig:metrics-sbp-phase}--\ref{fig:metrics-sbp-phase-b} show how each trajectory's time is split across the four R6 phases on SBP.

\paragraph{Per-model tool-call distribution (SBP).}
Figures~\ref{fig:metrics-sbp-tooldist}--\ref{fig:metrics-sbp-tooldist-b} show how each model's tool calls split across the R1 categories on SWE-Bench-Pro.

\FloatBarrier

\subsection{Metrics: Terminal-Bench-2}
\label{app:metrics:tb2}

The Terminal-Bench-2 (TB2) metrics in this section are computed by the same windowed judge pipeline as on SWE (Section~\ref{app:metrics:judge}), but with a TB2-specific rubric variant. TB2 tasks \emph{build or author} a deliverable under \texttt{/app} rather than patch a checked-out repo under \texttt{/testbed}, ship no project test suite, and are graded by an externally-run success command on the same container. The judge, therefore, reuses the \texttt{R1}, \texttt{R3}, \texttt{R5}, \texttt{R7}, and \texttt{R8} fields of Table~\ref{tab:judge-buckets} unchanged and only re-shapes the task-dependent entries. Table~\ref{tab:judge-buckets-tb2} lists the differences.

\begin{table}[h]
\centering
\small
\setlength{\tabcolsep}{4pt}
\caption{Terminal-Bench-2 judge rubric: the only fields that differ from the SWE rubric of Table~\ref{tab:judge-buckets}. \texttt{R1}, \texttt{R3}, \texttt{R5}, \texttt{R7}, \texttt{R8} are unchanged.}
\label{tab:judge-buckets-tb2}
\begin{tabular}{@{}>{\raggedright\arraybackslash}p{2.2cm}>{\raggedright\arraybackslash}p{5.0cm}>{\raggedright\arraybackslash}p{6.1cm}@{}}
\toprule
Field & SWE rubric (Table~\ref{tab:judge-buckets}) & TB2 rubric \\
\midrule
Workspace & \texttt{/testbed} (repo at \texttt{base\_commit}) & \texttt{/app} (task workspace; no golden patch) \\
\texttt{R2 target\_kind} & \texttt{project\_source} \textbar{} \texttt{project\_test} \textbar{} \texttt{scratch} \textbar{} \texttt{build\_config} \textbar{} \texttt{vcs\_internal} \textbar{} \texttt{none} & \texttt{task\_artifact} \textbar{} \texttt{upstream\_source} \textbar{} \texttt{scratch} \textbar{} \texttt{build\_config} \textbar{} \texttt{vcs\_internal} \textbar{} \texttt{none} (splits the edited artifact into the deliverable vs.\ 3rd-party source being built/patched) \\
\texttt{R4 test\_activity} & \texttt{suite} \textbar{} \texttt{scratch} \textbar{} \texttt{none} & \texttt{success\_check} \textbar{} \texttt{scratch} \textbar{} \texttt{none} (no project suite; \texttt{success\_check} is the task's declared success oracle) \\
\texttt{R6 phase} & \texttt{explore} \textbar{} \texttt{localize} \textbar{} \texttt{implement} \textbar{} \texttt{verify} \textbar{} \texttt{other} & \texttt{setup\_env} \textbar{} \texttt{build} \textbar{} \texttt{explore} \textbar{} \texttt{implement} \textbar{} \texttt{verify} \textbar{} \texttt{cleanup} \textbar{} \texttt{other} (drops \texttt{localize}; promotes \texttt{setup\_env}/\texttt{build} out of \texttt{other}; adds post-solve \texttt{cleanup}) \\
New-file default & A freshly created \texttt{/testbed} file is usually \texttt{scratch}; the SWE-Pro addendum carves out new package-tree source as \texttt{edit\_source}. & Inverted: a new \texttt{/app} file is the deliverable by default (\texttt{task\_artifact}/\texttt{edit\_source}) unless it is clearly probe-shaped or under \texttt{/tmp}. \\
\bottomrule
\end{tabular}
\end{table}

\begin{figure}[h]
    \centering
    \includegraphics[width=\linewidth]{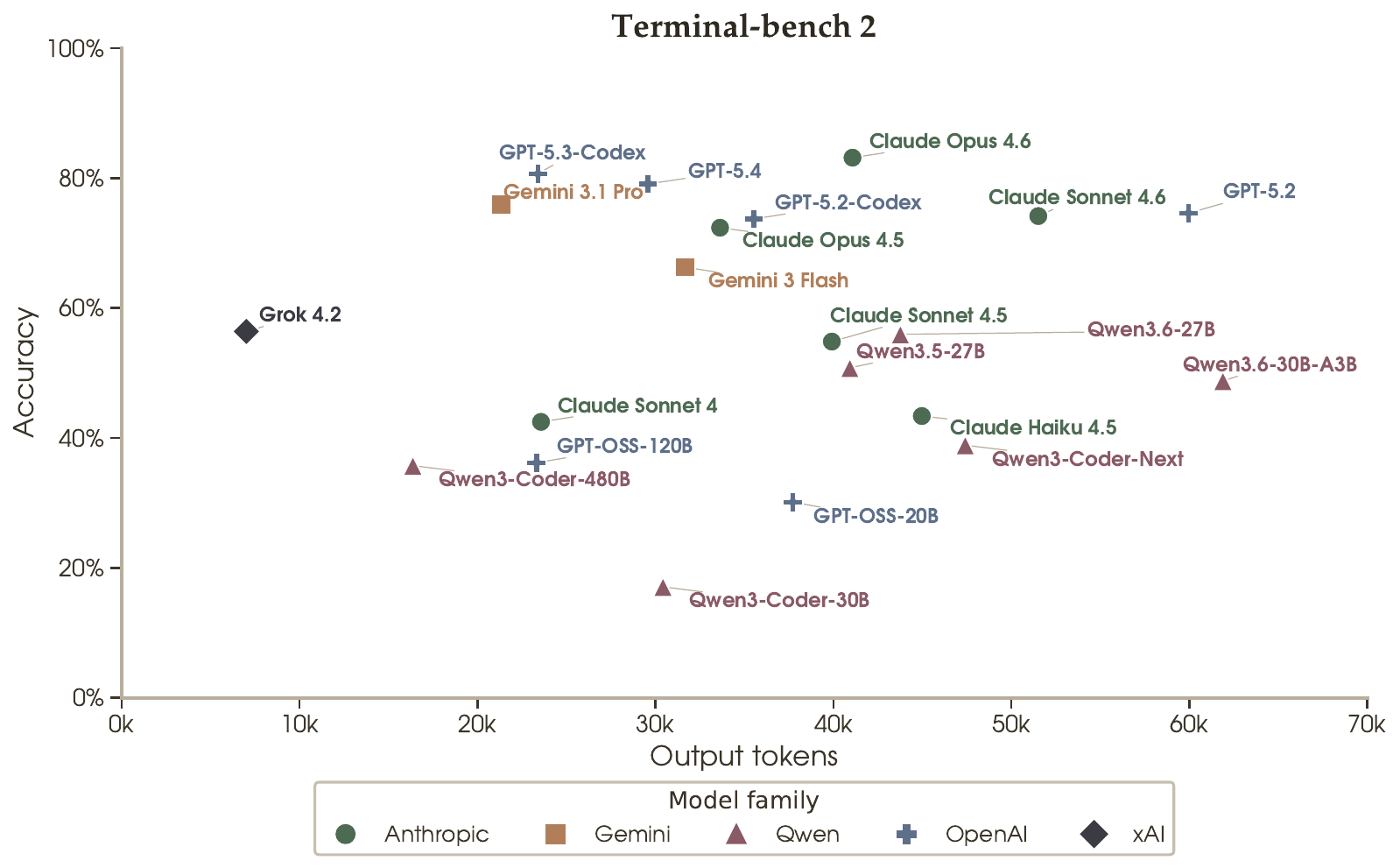}
    \caption{Terminal-Bench-2: total output tokens vs.\ pass@1, one point per model. Output tokens are for each problem instance (summed over all LLM calls in an agent run), averaged across 89 instances and 5 runs per instance . Pass@1 is the mean of the five per-run resolution rates. Colors group models by family. Top-left corresponds to high accuracy with the fewest output tokens spent reasoning and acting.}
    \label{fig:metrics-tb2-pass-vs-tokens}
\end{figure}

Figure~\ref{fig:metrics-tb2-pass-vs-tokens} reports the same token-vs-pass@1 Pareto view for Terminal-Bench-2 that Figures~\ref{fig:metrics-sbv-pass-vs-tokens} and~\ref{fig:metrics-sbp-pass-vs-tokens} report for SWE-Bench-Verified and SWE-Bench-Pro respectively. Open-source families plotted here use the Unconstrained setting (the only one we ran for them). Closed-source models use the Constrained setting to match Appendix~\ref{app:passrates}.

\begin{figure}[p]
\centering
\includegraphics[width=0.65\linewidth]{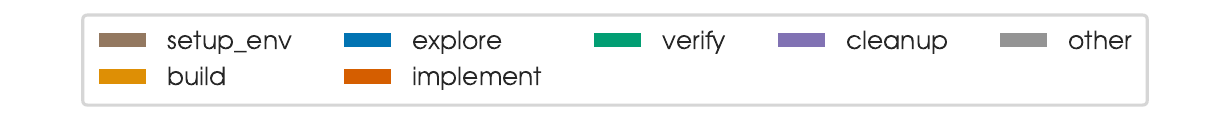}\\[2pt]
\setlength{\tabcolsep}{1pt}
\renewcommand{\arraystretch}{0.5}
\begin{tabular}{ccc}
\includegraphics[width=0.31\linewidth]{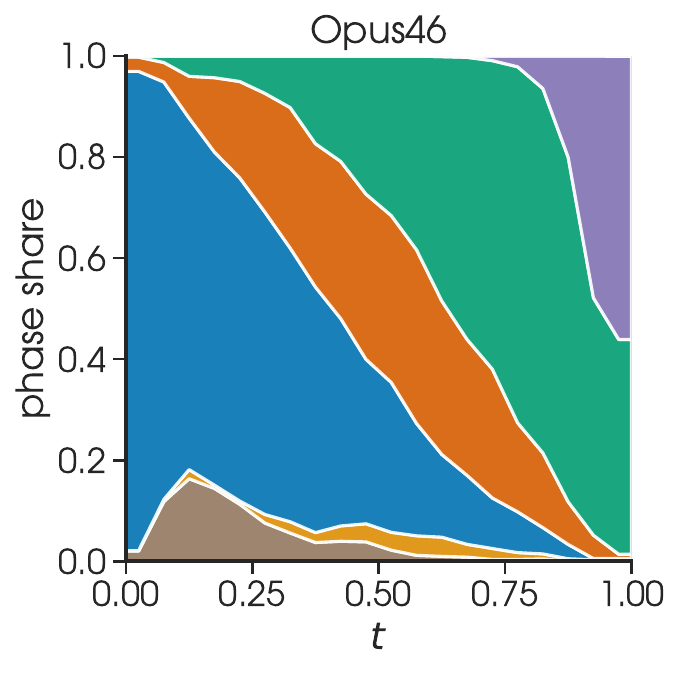} &
\includegraphics[width=0.31\linewidth]{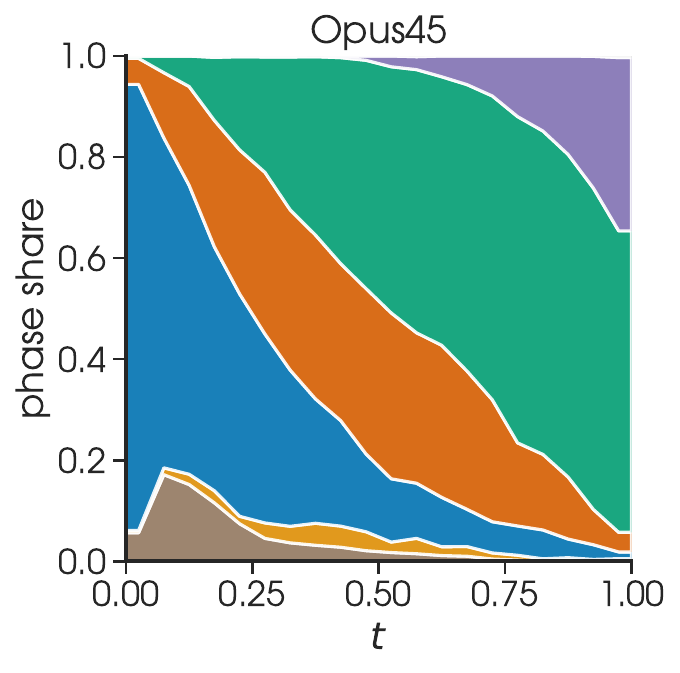} &
\includegraphics[width=0.31\linewidth]{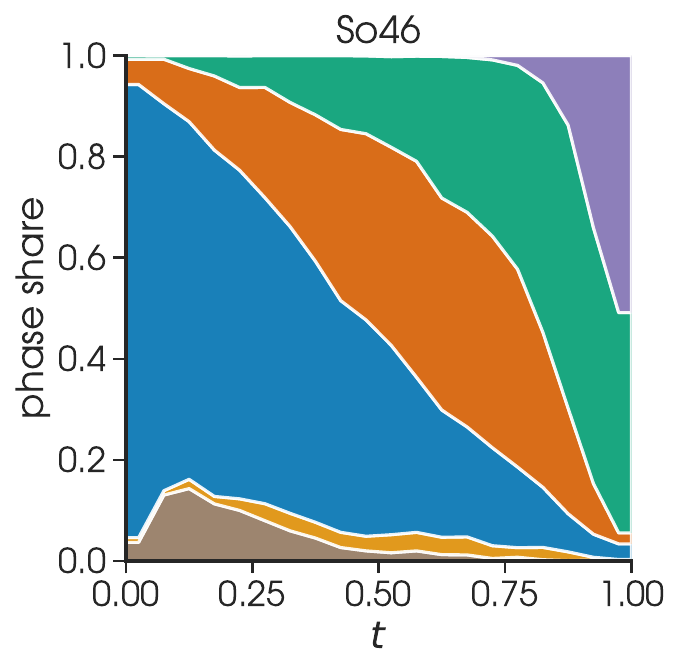} \\
\includegraphics[width=0.31\linewidth]{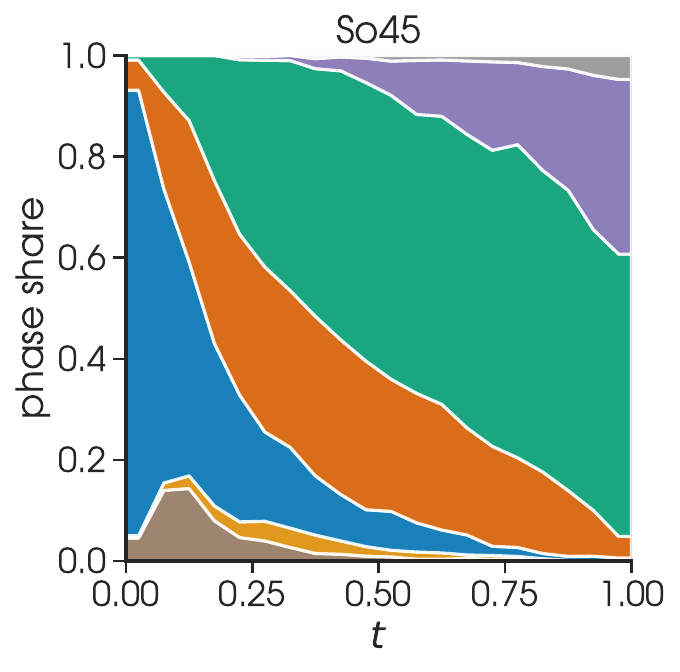} &
\includegraphics[width=0.31\linewidth]{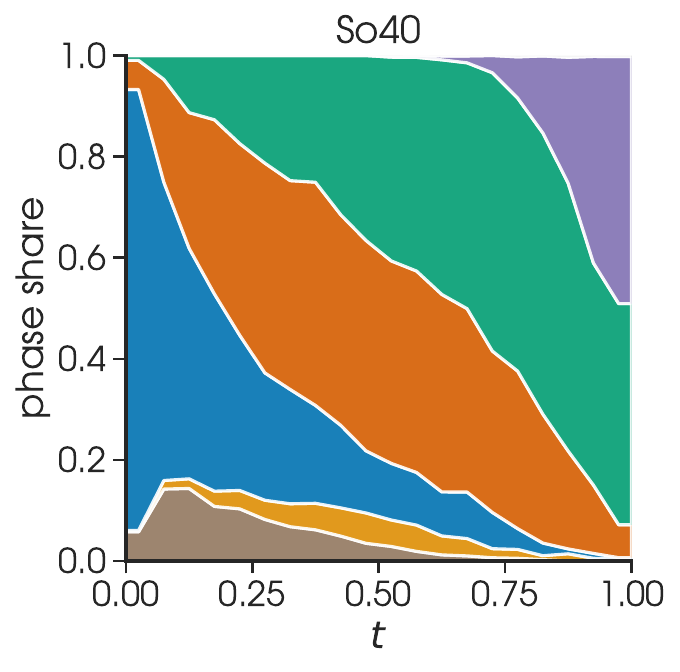} &
\includegraphics[width=0.31\linewidth]{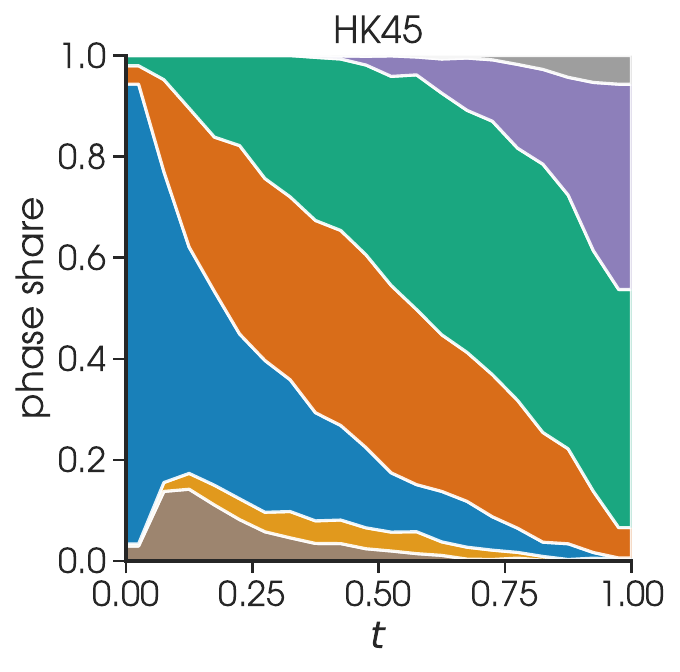} \\
\includegraphics[width=0.31\linewidth]{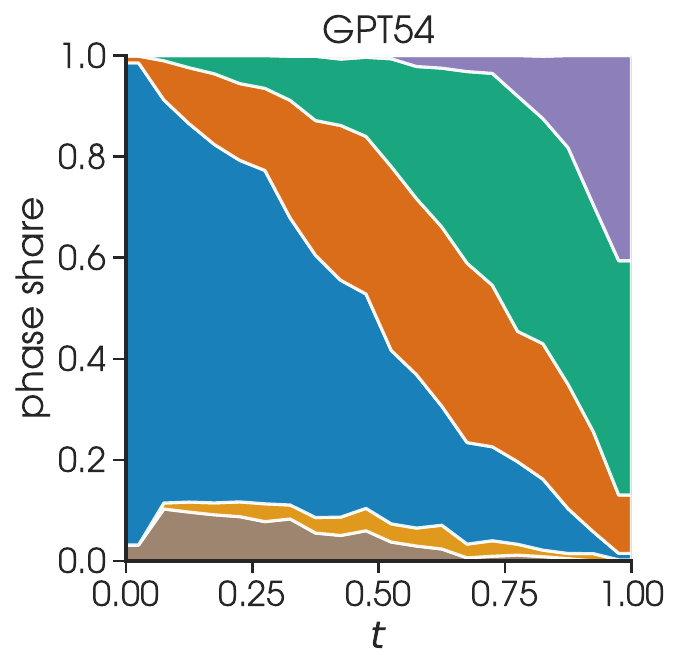} &
\includegraphics[width=0.31\linewidth]{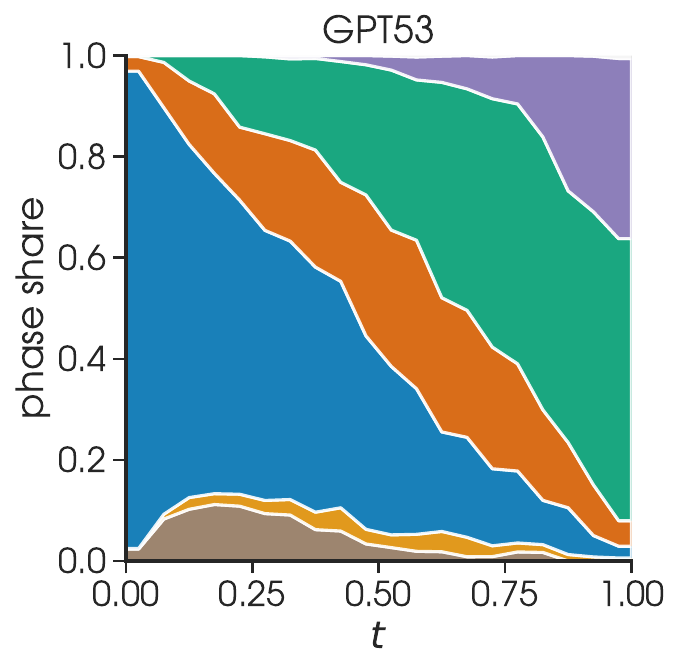} &
\includegraphics[width=0.31\linewidth]{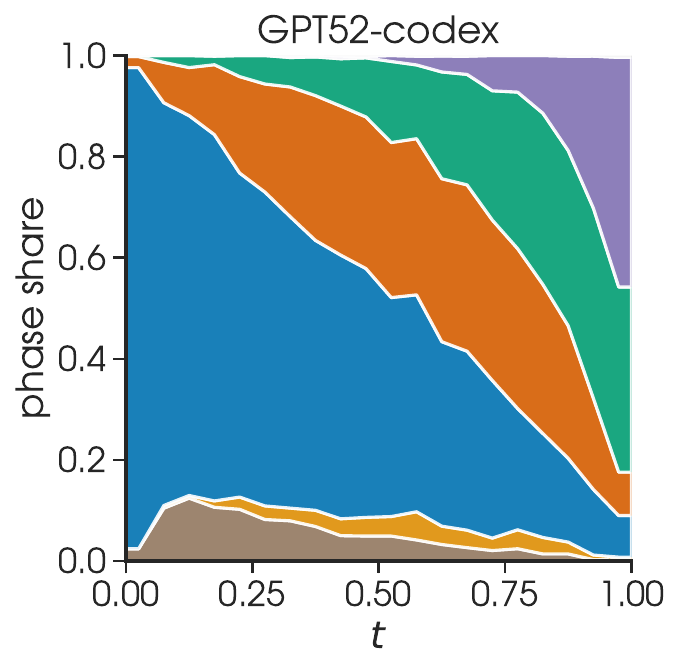} \\
\includegraphics[width=0.31\linewidth]{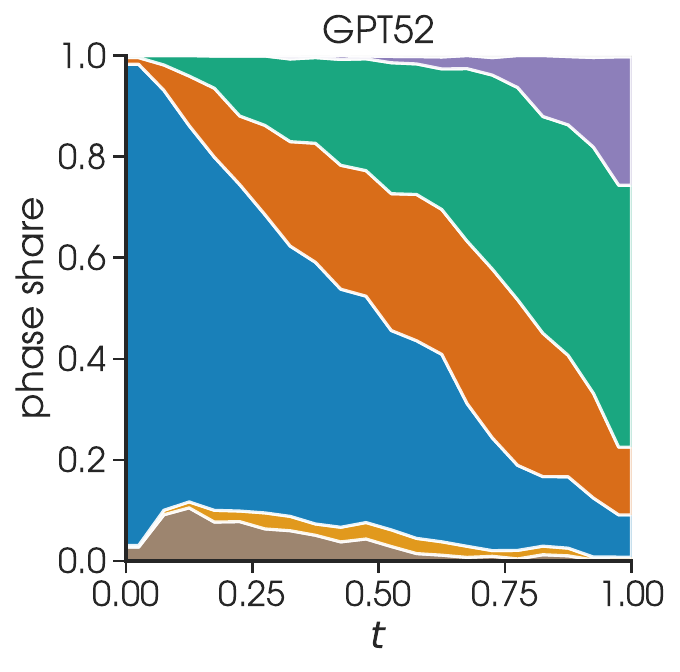} &
\includegraphics[width=0.31\linewidth]{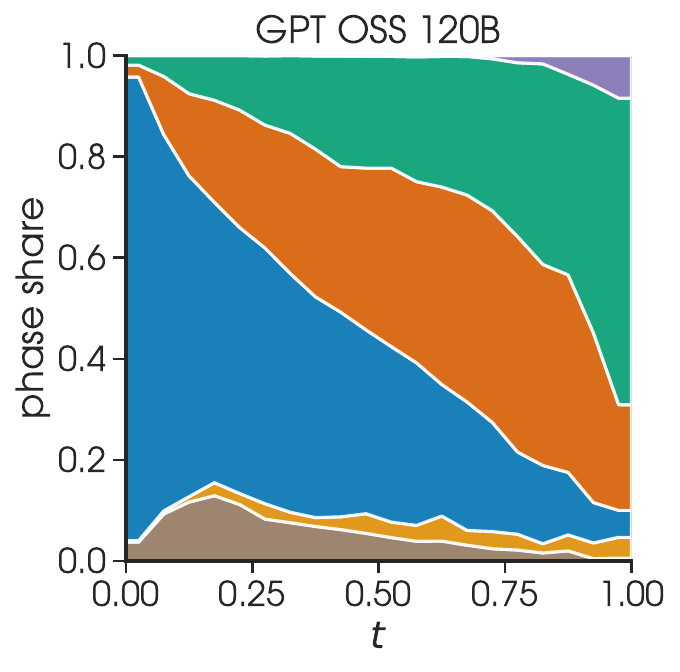} &
\includegraphics[width=0.31\linewidth]{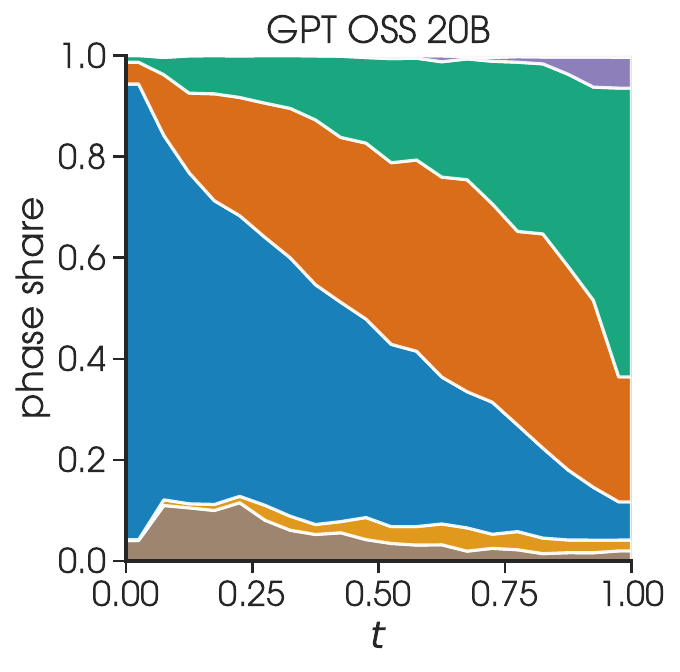} \\
\end{tabular}
\caption{\textbf{Phase composition surfaces model-specific schedules.} Per-cycle share of the four R6 (see LLM judge rubrics in Table\,\ref{tab:judge-buckets}) phases (\texttt{explore},
\texttt{localize}, \texttt{implement}, \texttt{verify}) vs.\ normalised cycle
position $t$ on Terminal-Bench-2, for Claude and GPT families (part 1 of 2). Given the
same tasks, models schedule their cycles differently. The implementation peak shifts in position and width, and the 
verify phase carries different long-tail mass.}
\label{fig:metrics-tb2-phase}
\end{figure}

\begin{figure}[p]
\centering
\includegraphics[width=0.65\linewidth]{tb2/phase_legend.pdf}\\[2pt]
\setlength{\tabcolsep}{1pt}
\renewcommand{\arraystretch}{0.5}
\begin{tabular}{ccc}
\includegraphics[width=0.31\linewidth]{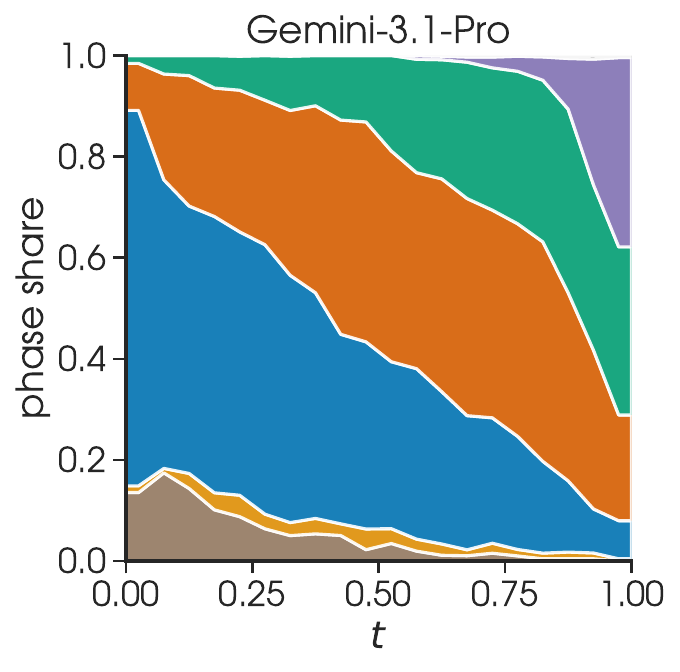} &
\includegraphics[width=0.31\linewidth]{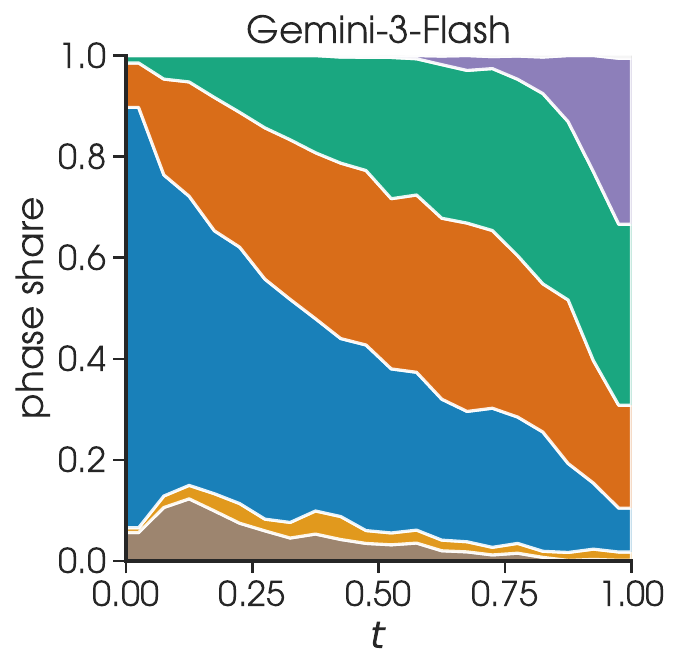} & \\
\includegraphics[width=0.31\linewidth]{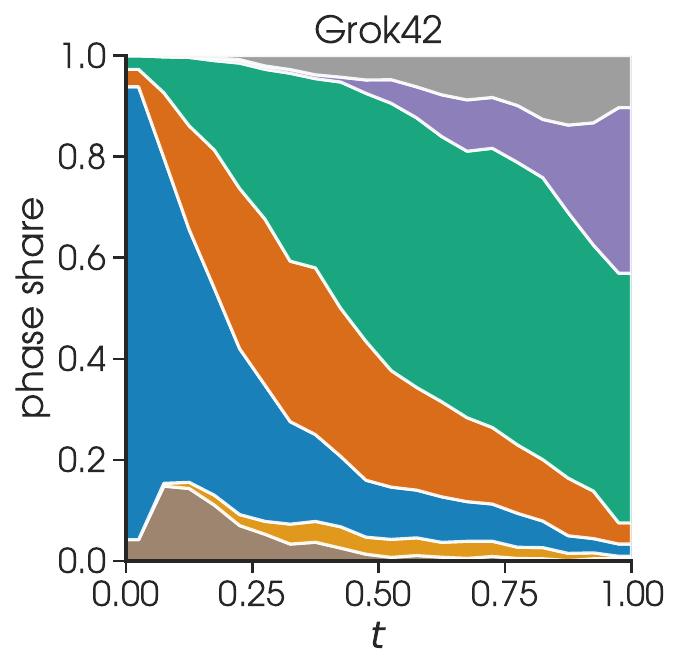} & & \\
\includegraphics[width=0.31\linewidth]{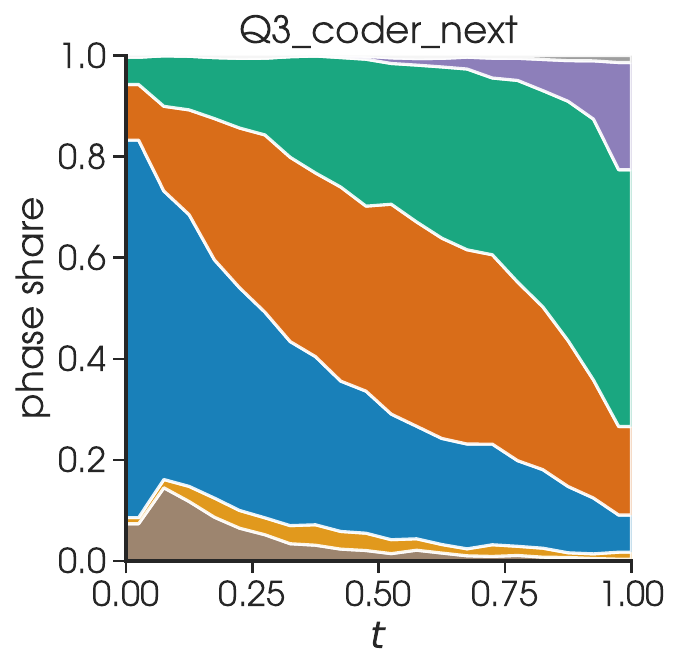} &
\includegraphics[width=0.31\linewidth]{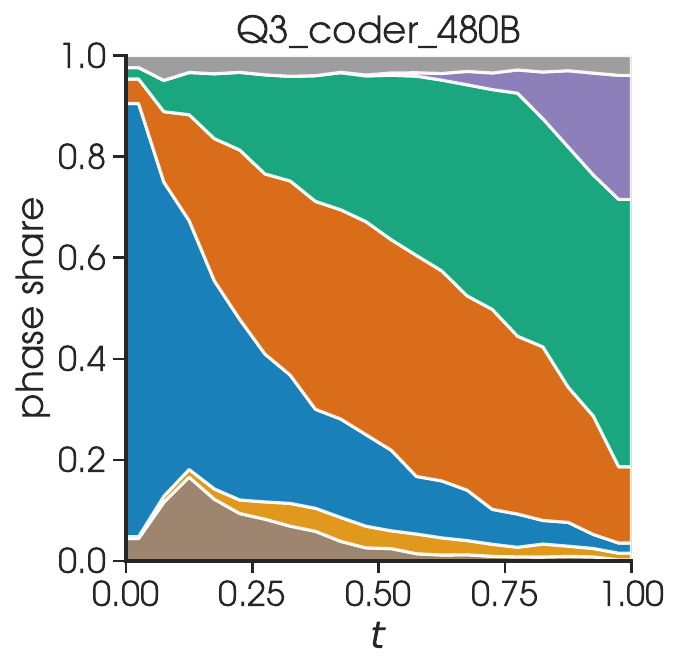} &
\includegraphics[width=0.31\linewidth]{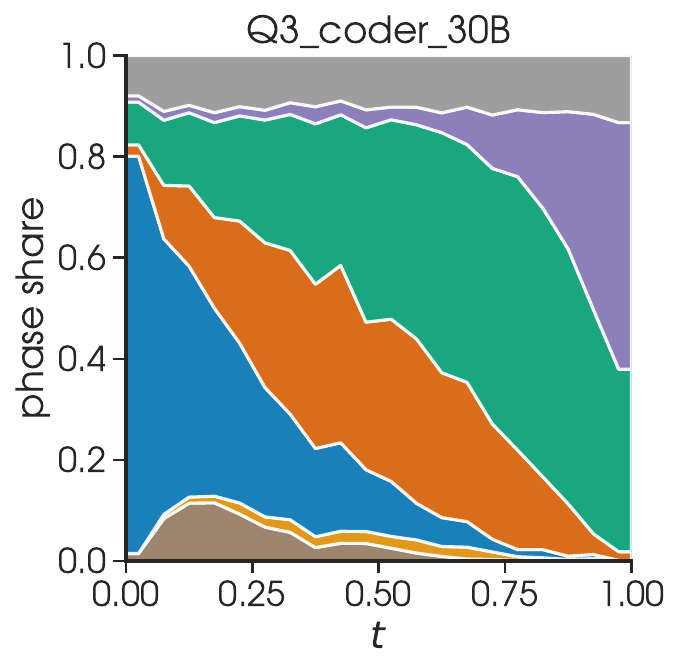} \\
\includegraphics[width=0.31\linewidth]{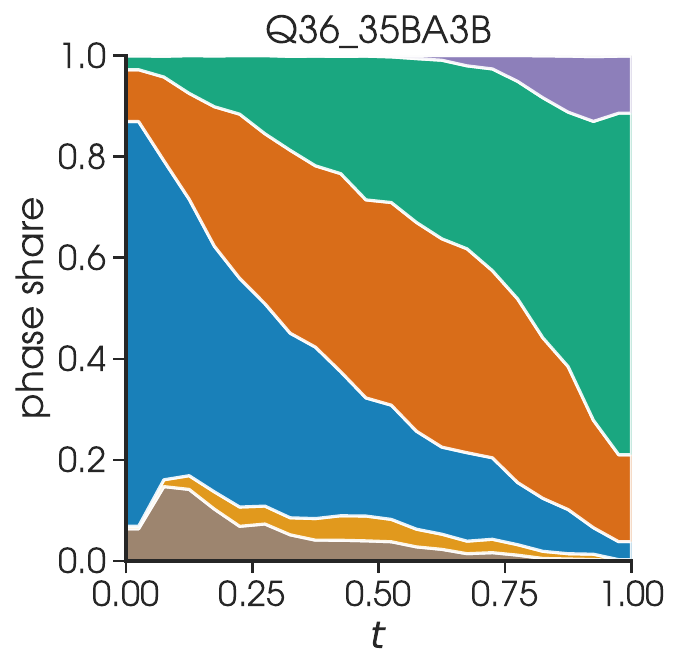} &
\includegraphics[width=0.31\linewidth]{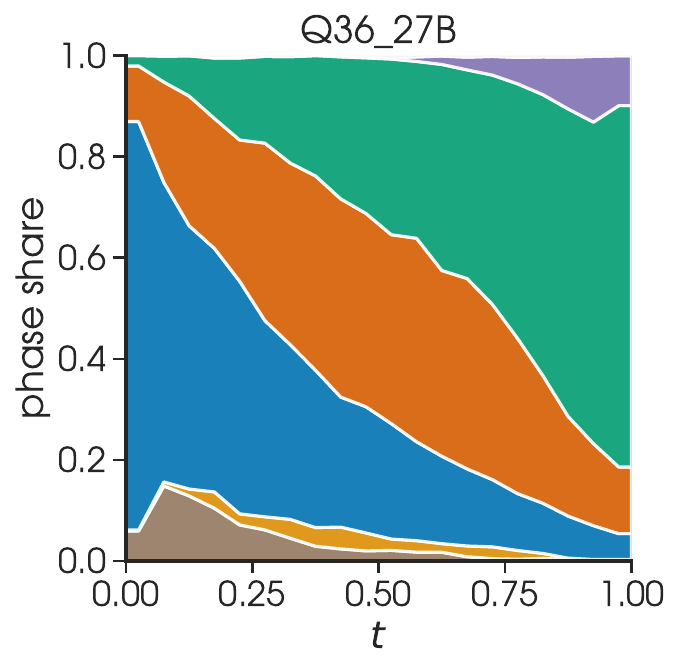} &
\includegraphics[width=0.31\linewidth]{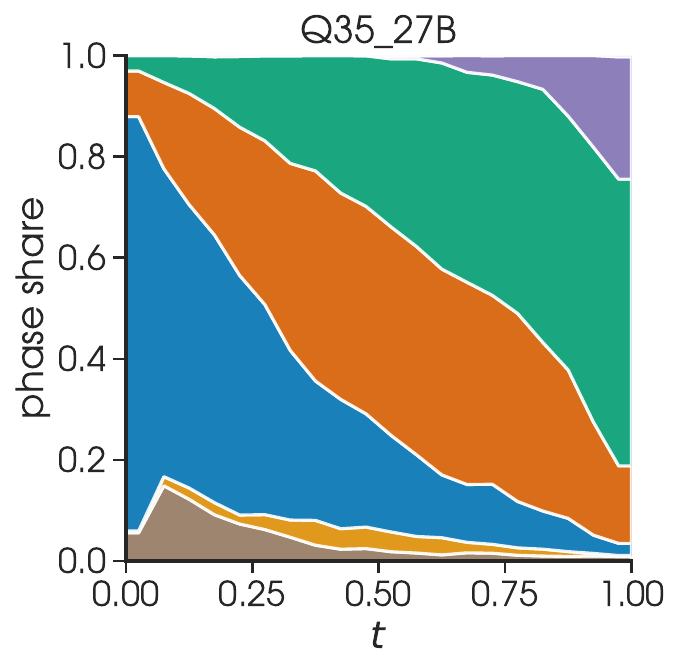} \\
\end{tabular}
\caption{Phase composition on Terminal-Bench-2 (part 2 of 2): Gemini, Grok and Qwen families. Continued from Figure~\ref{fig:metrics-tb2-phase}.}
\label{fig:metrics-tb2-phase-b}
\end{figure}

\begin{figure}[p]
\centering
\setlength{\tabcolsep}{1pt}
\renewcommand{\arraystretch}{0.5}
\begin{tabular}{ccc}
\includegraphics[width=0.31\linewidth]{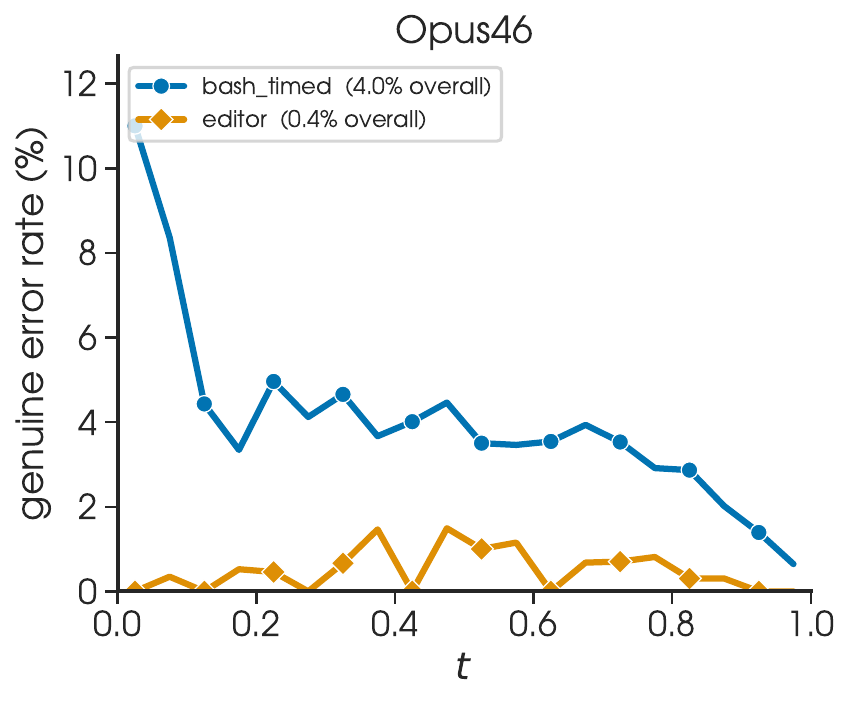} &
\includegraphics[width=0.31\linewidth]{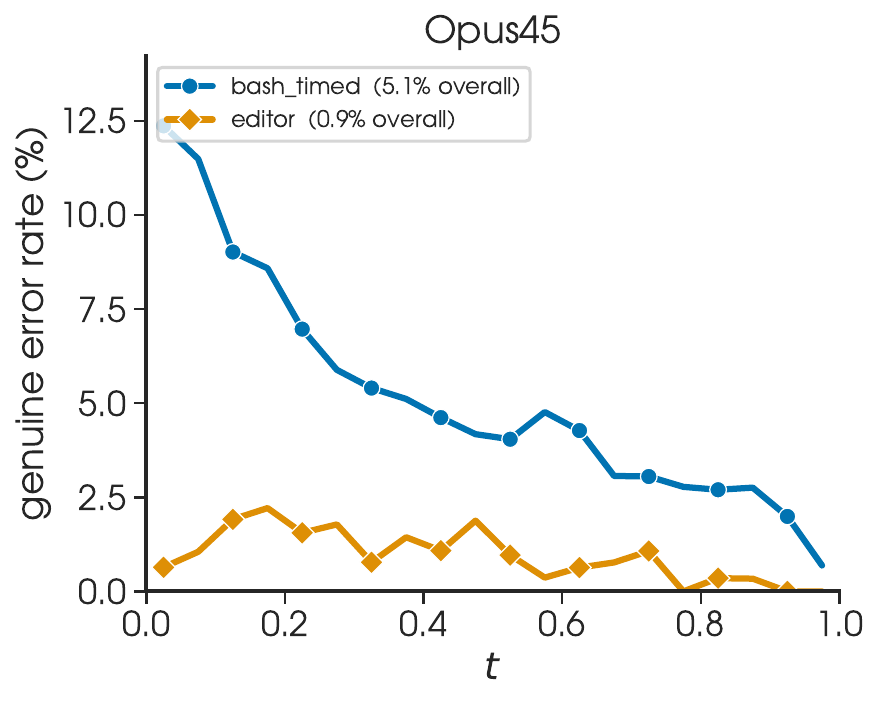} &
\includegraphics[width=0.31\linewidth]{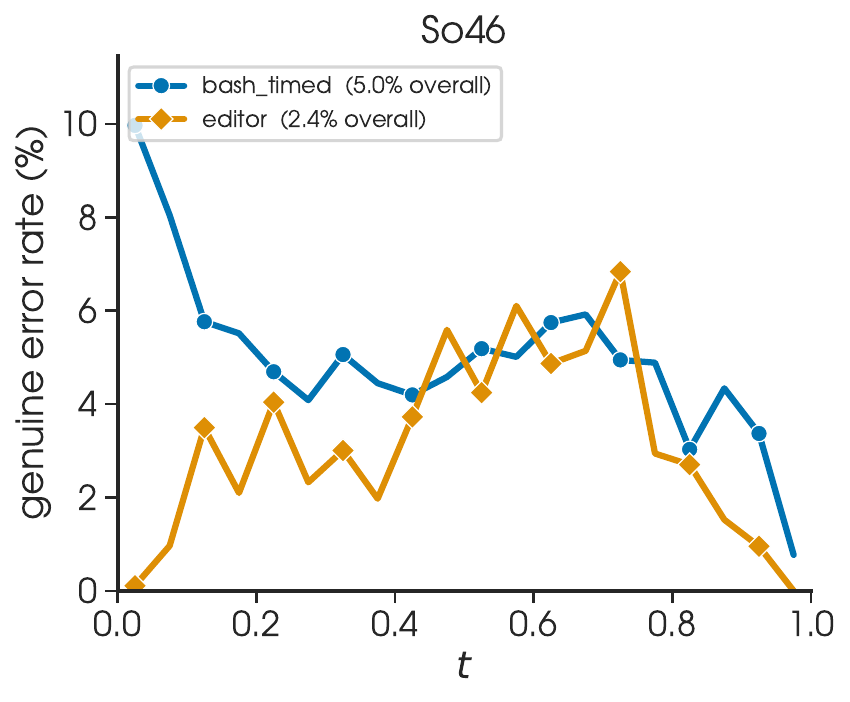} \\
\includegraphics[width=0.31\linewidth]{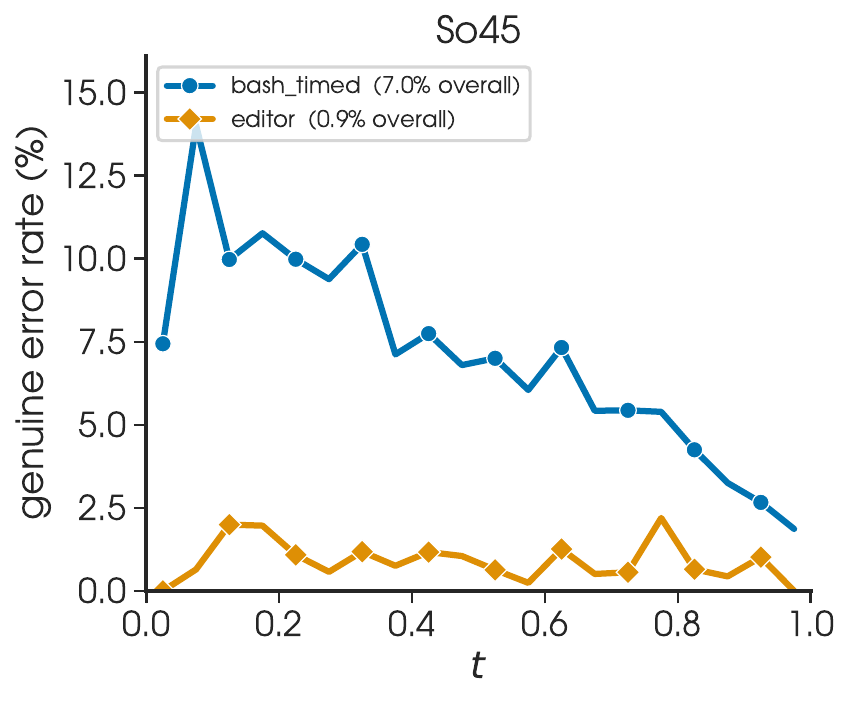} &
\includegraphics[width=0.31\linewidth]{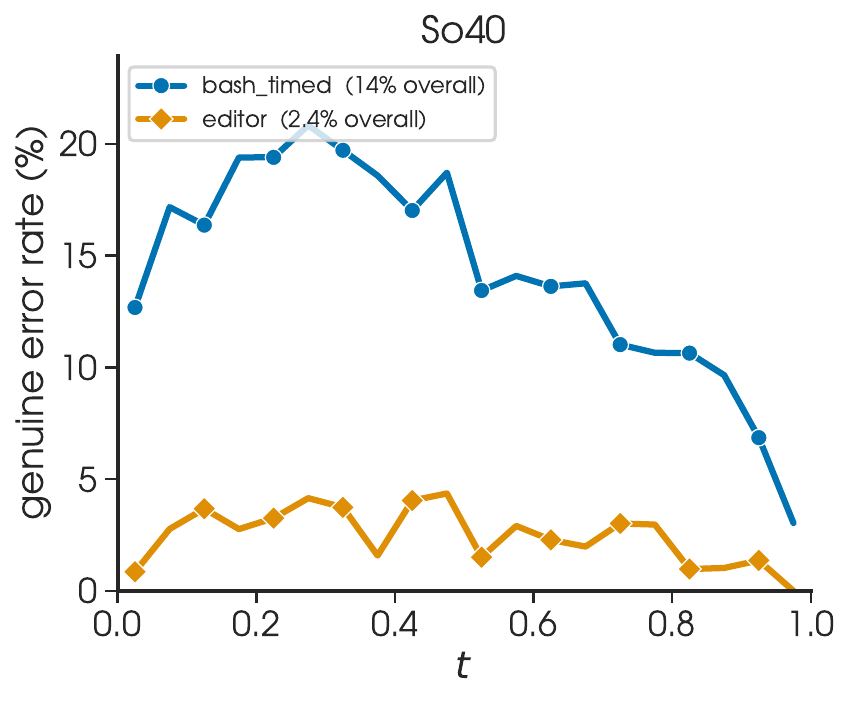} &
\includegraphics[width=0.31\linewidth]{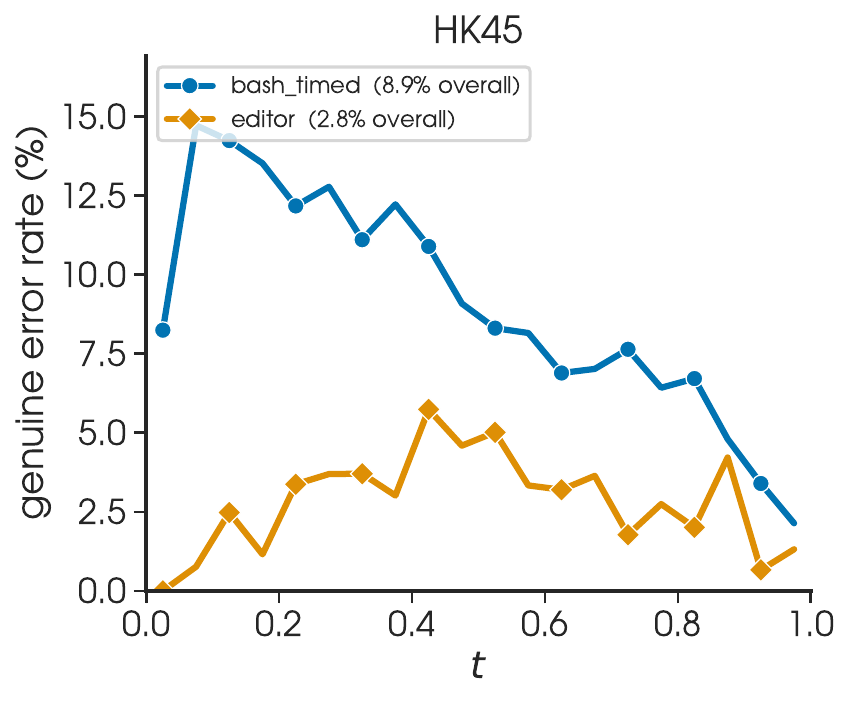} \\
\includegraphics[width=0.31\linewidth]{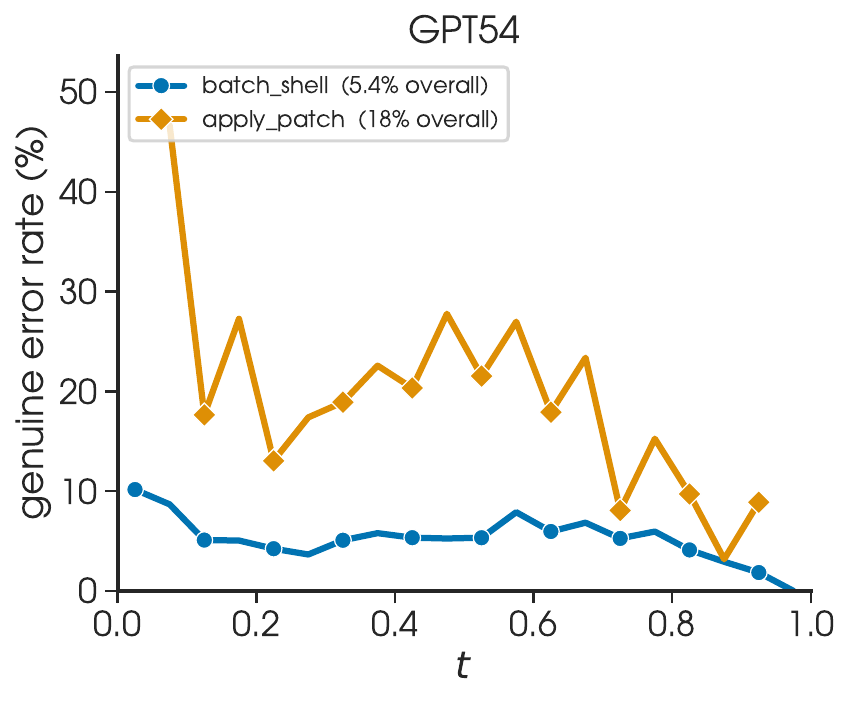} &
\includegraphics[width=0.31\linewidth]{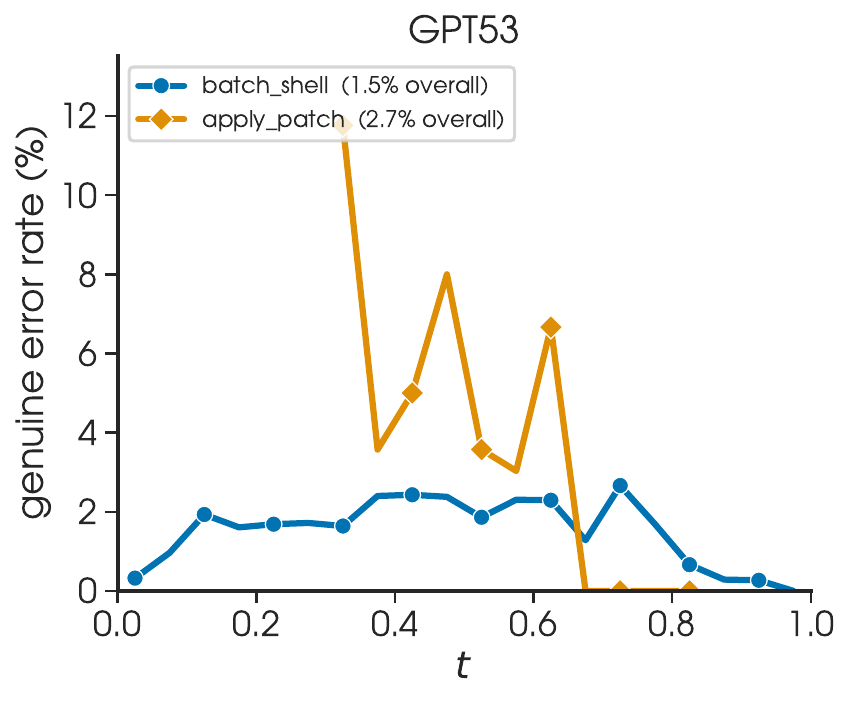} &
\includegraphics[width=0.31\linewidth]{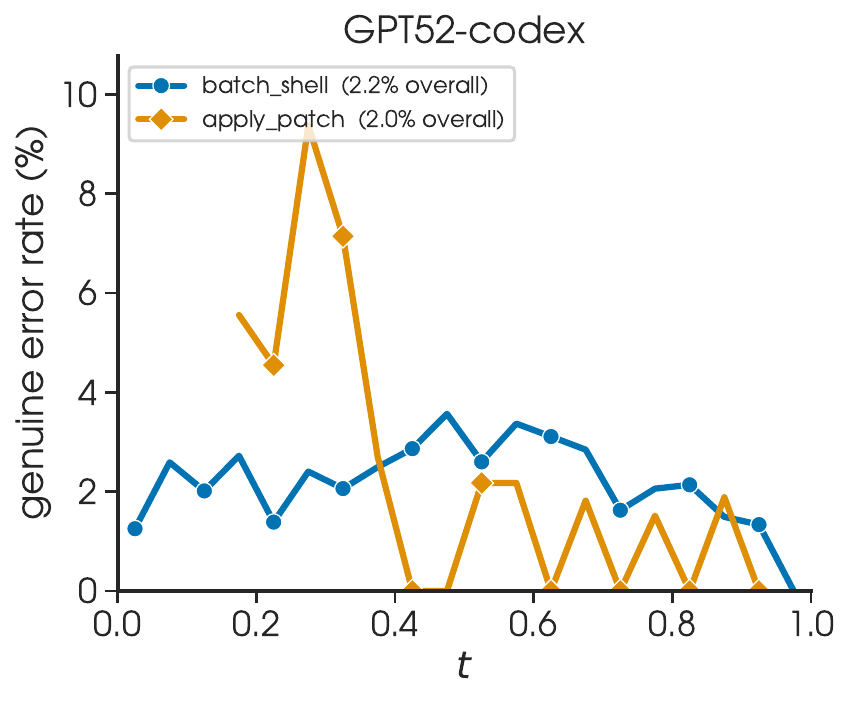} \\
\includegraphics[width=0.31\linewidth]{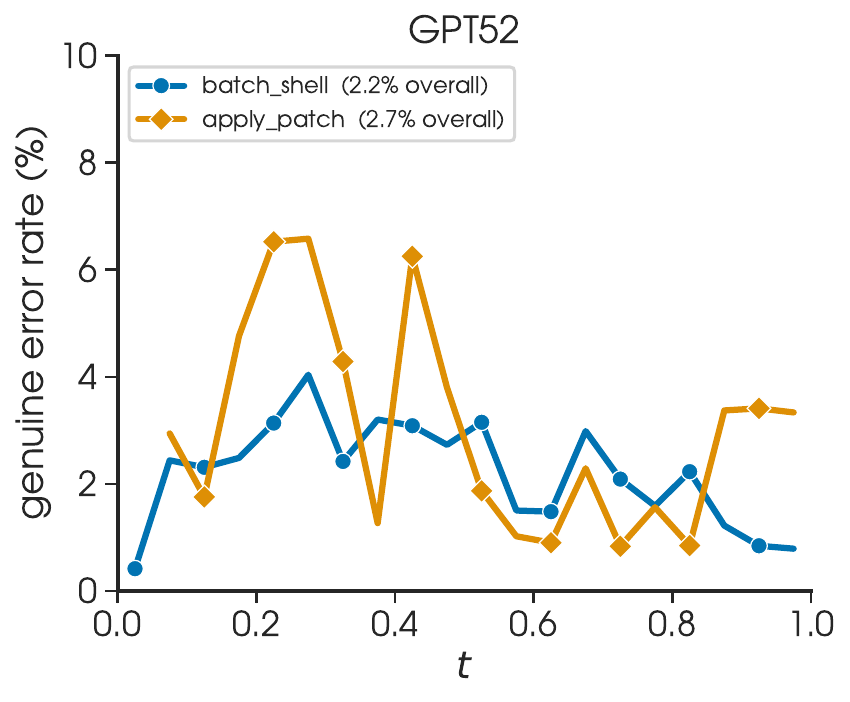} &
\includegraphics[width=0.31\linewidth]{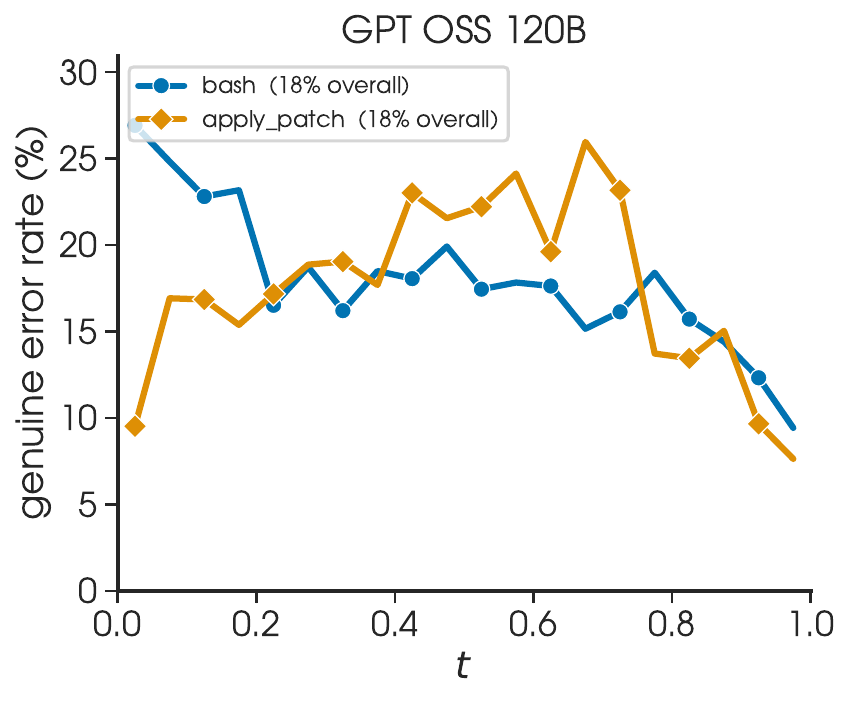} &
\includegraphics[width=0.31\linewidth]{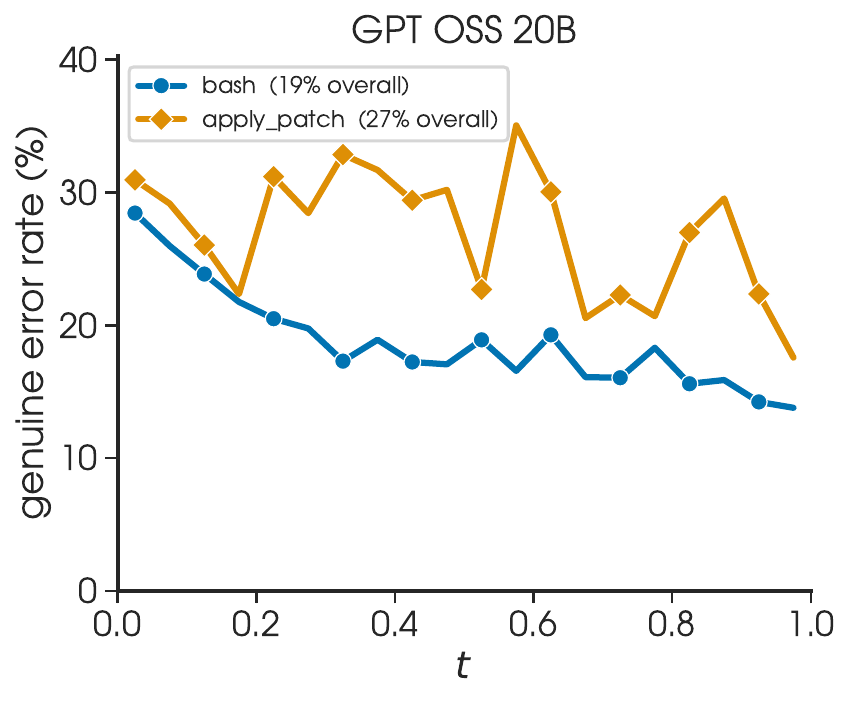} \\
\end{tabular}
\caption{Genuine tool-error rate per cycle on Terminal-Bench-2 (part 1 of 2): Anthropic Claude and OpenAI families. Benign exits excluded; tool families broken out within each sub-plot.}
\label{fig:metrics-tb2-toolerr}
\end{figure}

\begin{figure}[p]
\centering
\setlength{\tabcolsep}{1pt}
\renewcommand{\arraystretch}{0.5}
\begin{tabular}{ccc}
\includegraphics[width=0.31\linewidth]{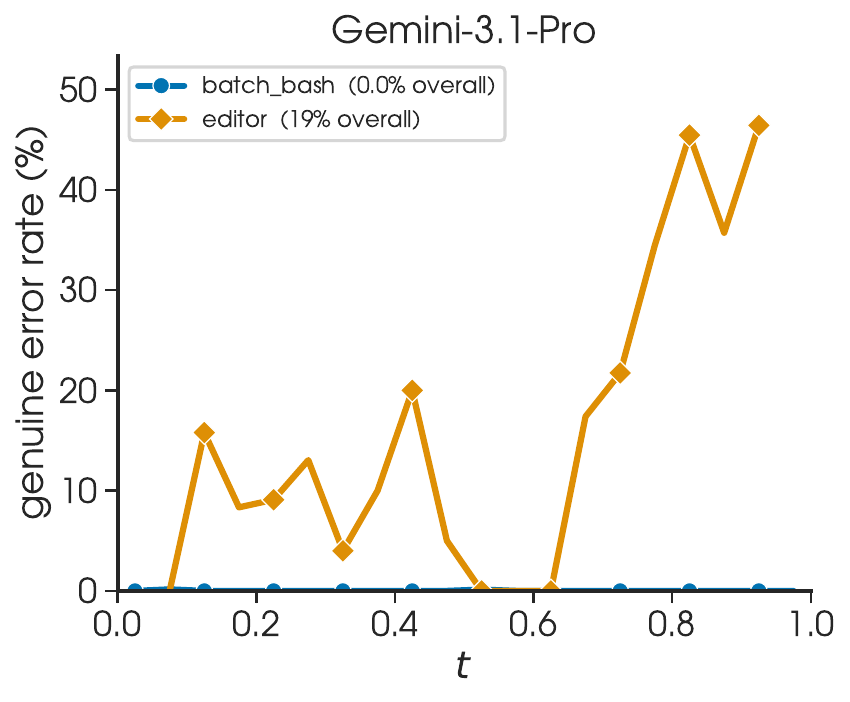} &
\includegraphics[width=0.31\linewidth]{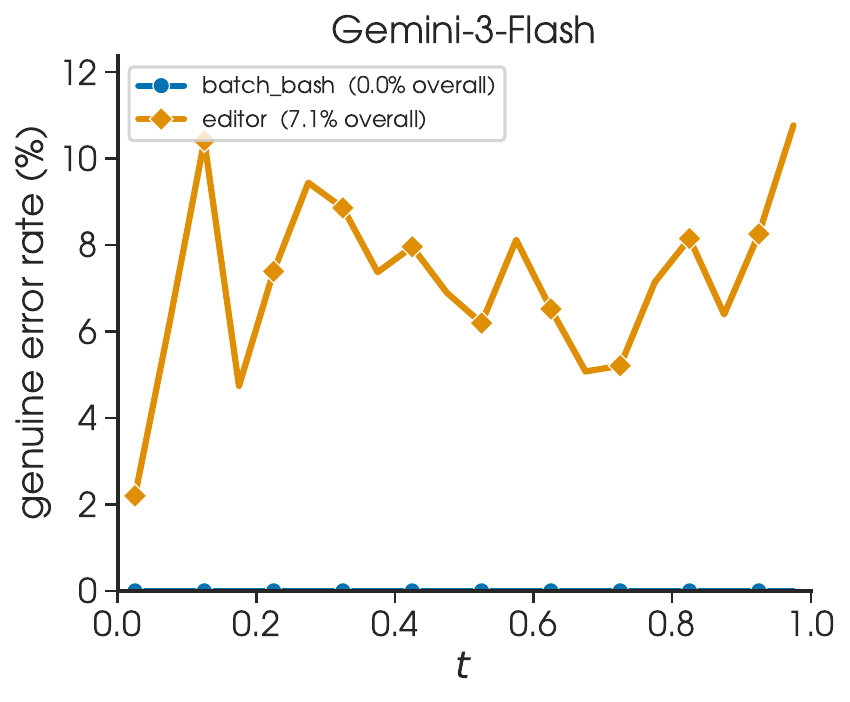} & \\
\includegraphics[width=0.31\linewidth]{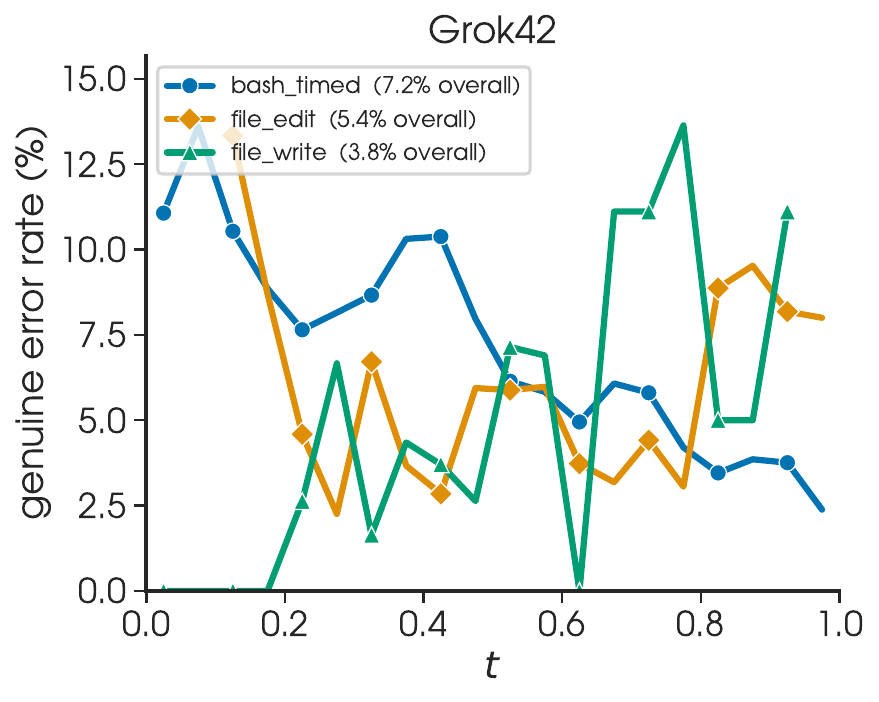} & & \\
\includegraphics[width=0.31\linewidth]{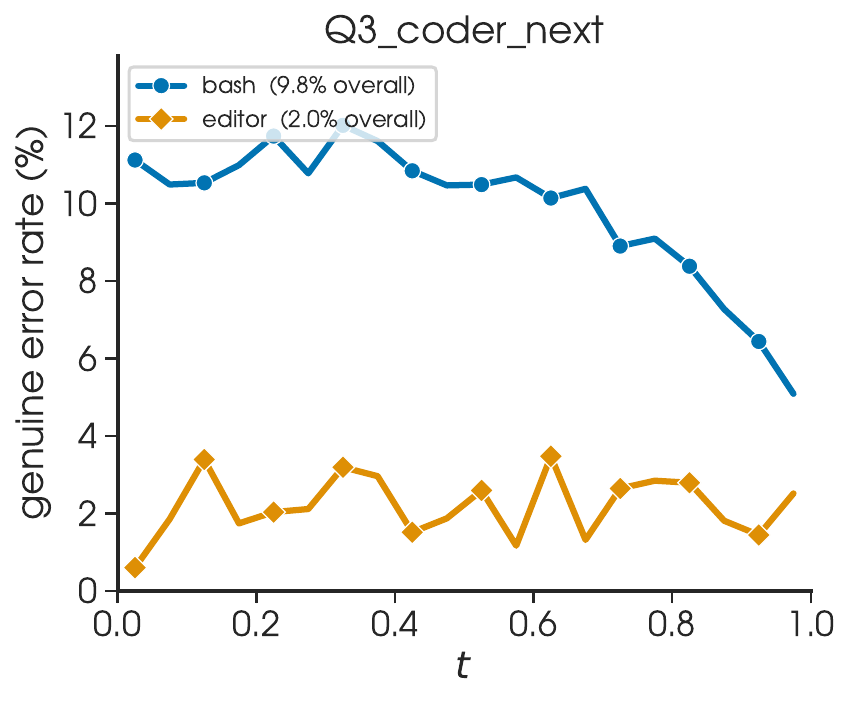} &
\includegraphics[width=0.31\linewidth]{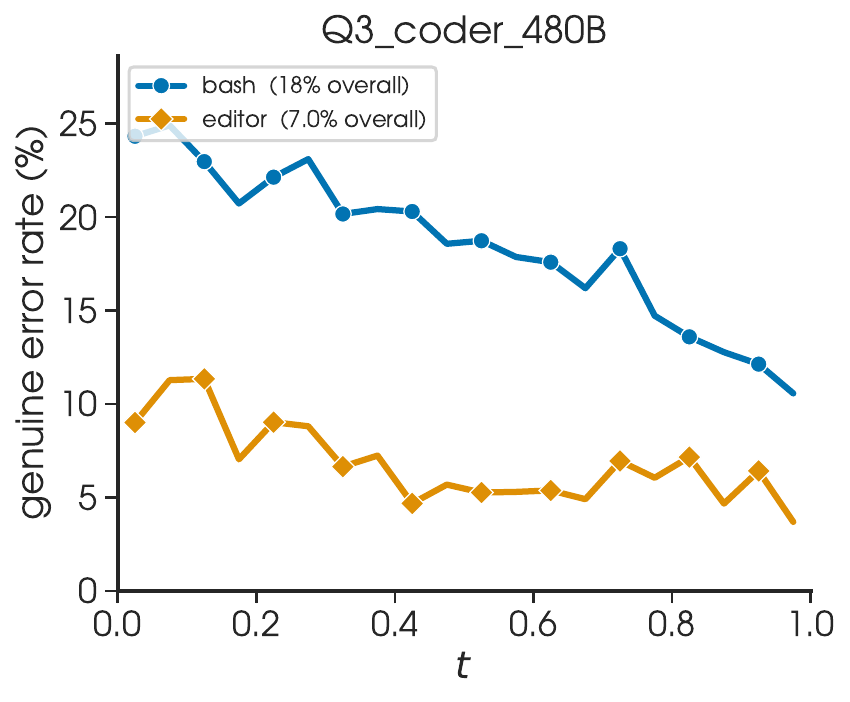} &
\includegraphics[width=0.31\linewidth]{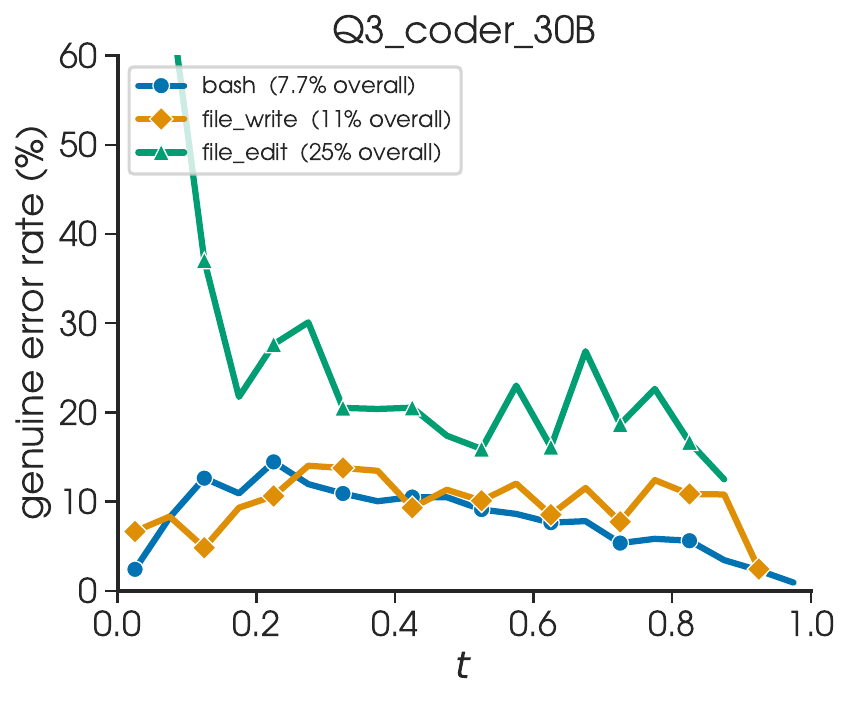} \\
\includegraphics[width=0.31\linewidth]{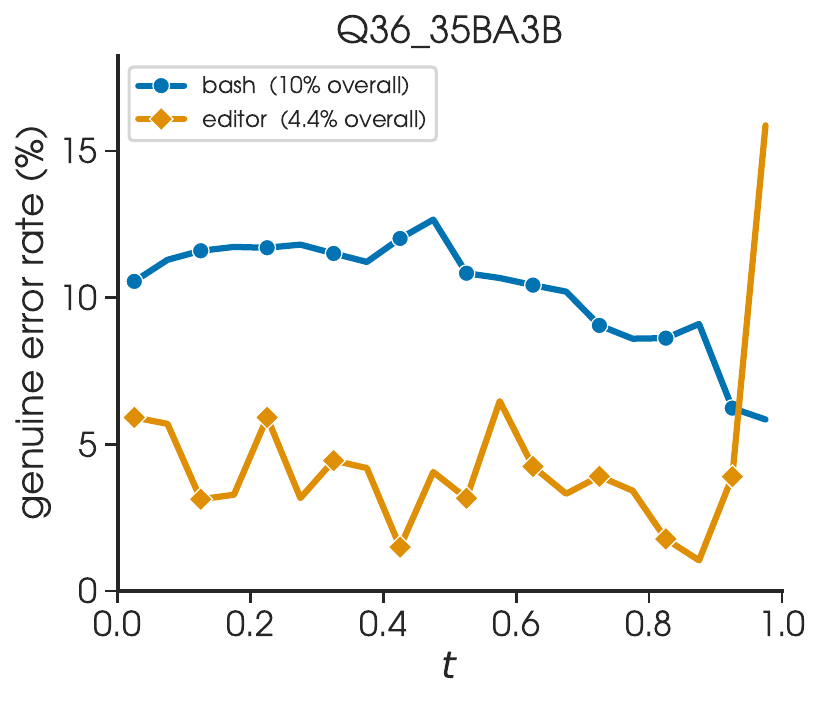} &
\includegraphics[width=0.31\linewidth]{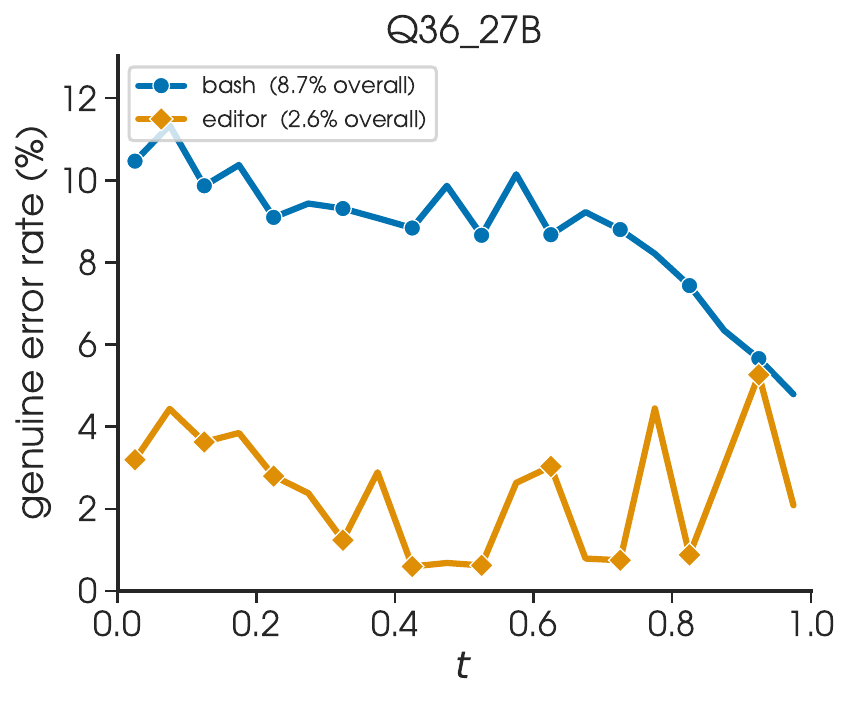} &
\includegraphics[width=0.31\linewidth]{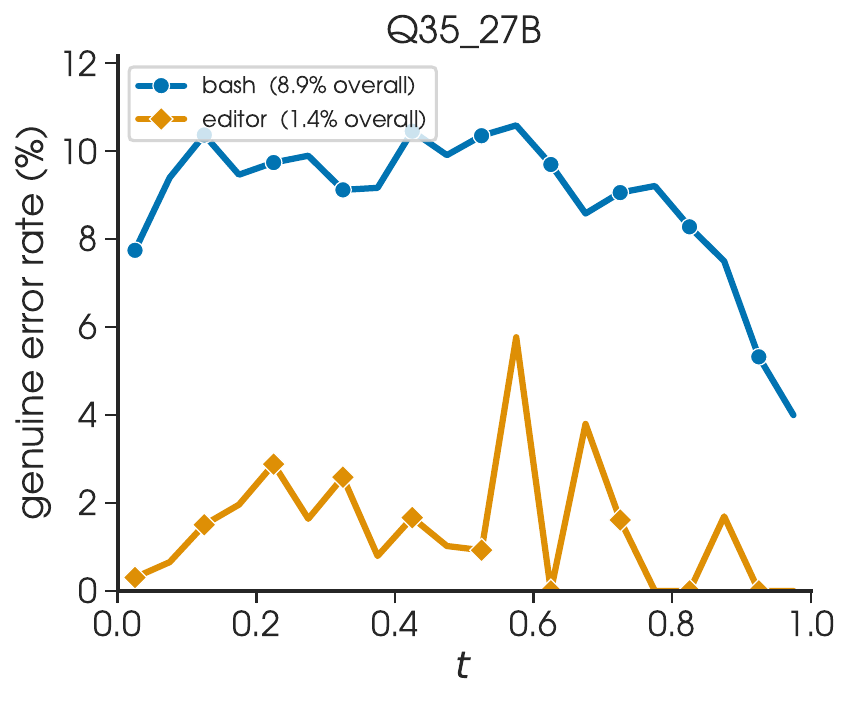} \\
\end{tabular}
\caption{Genuine tool-error rate per cycle on Terminal-Bench-2 (part 2 of 2): Gemini, Grok and Qwen families. Continued from Figure~\ref{fig:metrics-tb2-toolerr}.}
\label{fig:metrics-tb2-toolerr-b}
\end{figure}

\begin{figure}[p]
\centering
\includegraphics[width=0.75\linewidth]{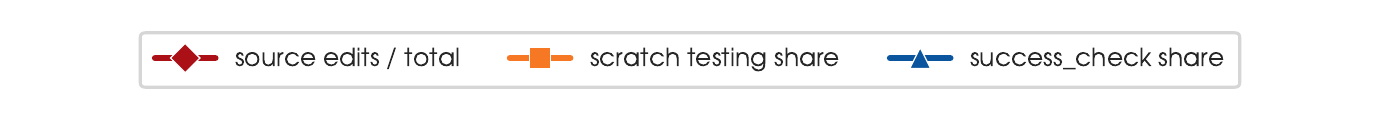}\\[2pt]
\setlength{\tabcolsep}{1pt}
\renewcommand{\arraystretch}{0.5}
\begin{tabular}{ccc}
\includegraphics[width=0.31\linewidth]{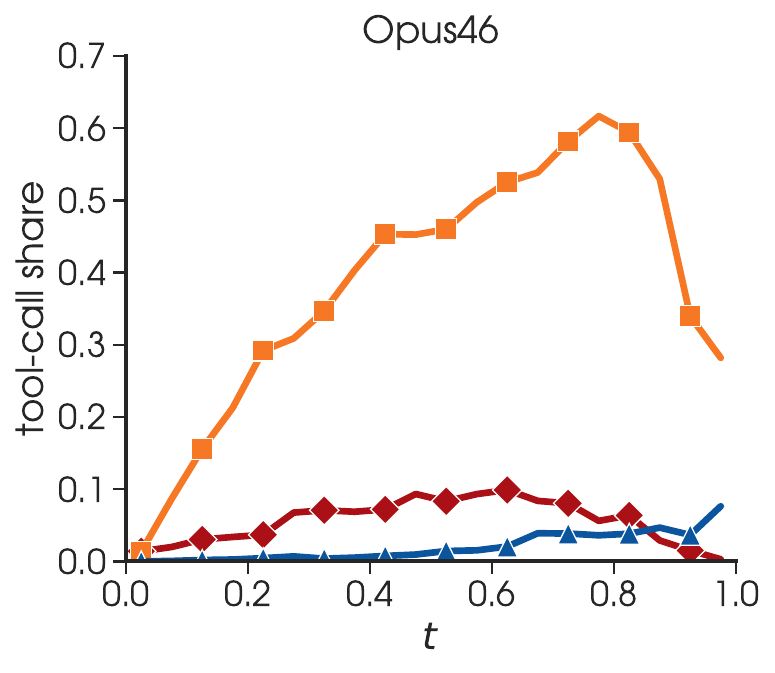} &
\includegraphics[width=0.31\linewidth]{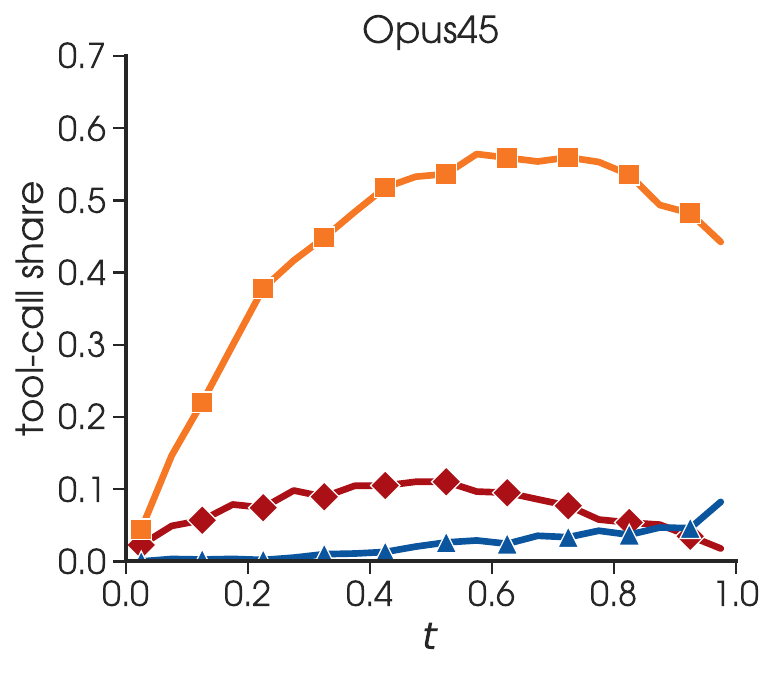} &
\includegraphics[width=0.31\linewidth]{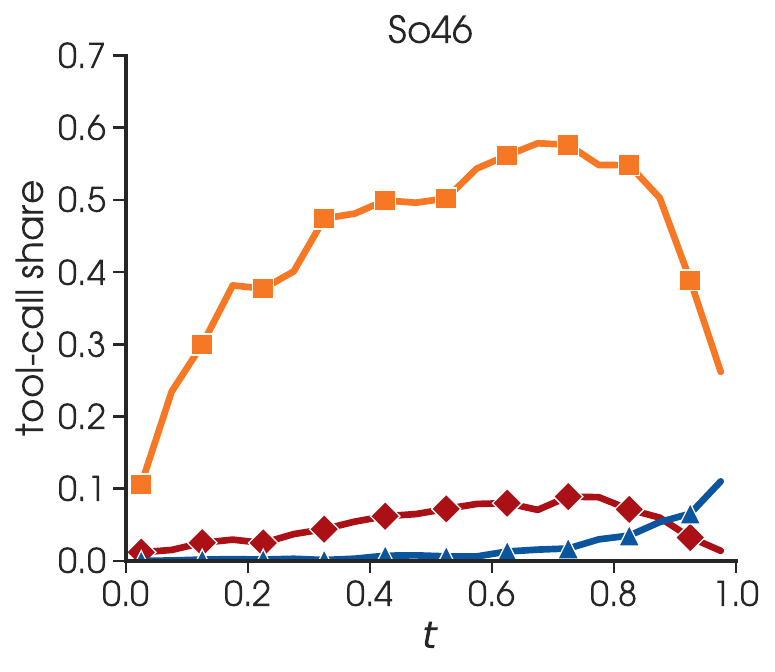} \\
\includegraphics[width=0.31\linewidth]{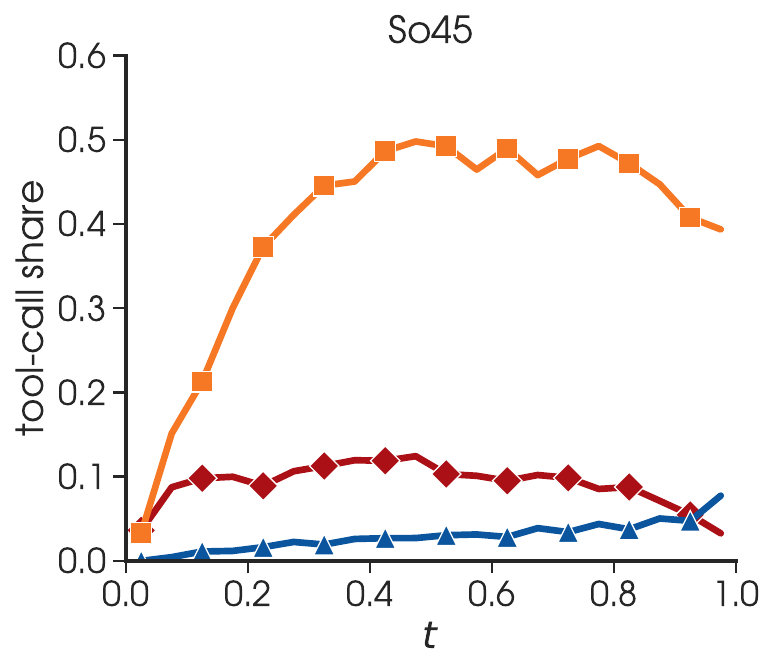} &
\includegraphics[width=0.31\linewidth]{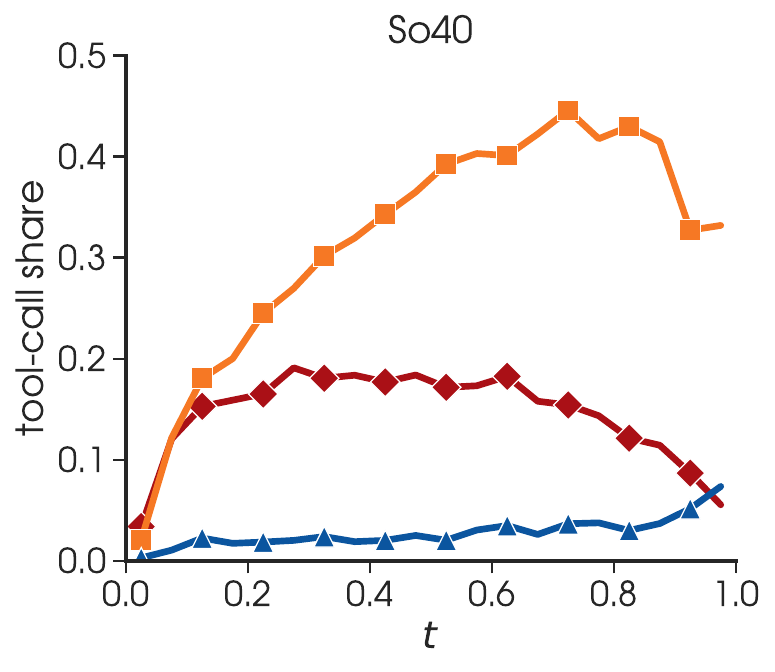} &
\includegraphics[width=0.31\linewidth]{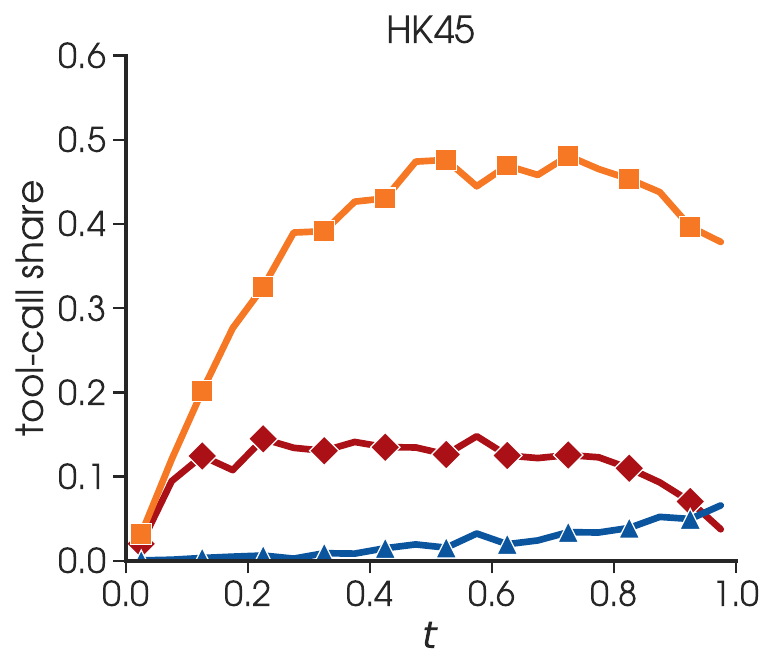} \\
\includegraphics[width=0.31\linewidth]{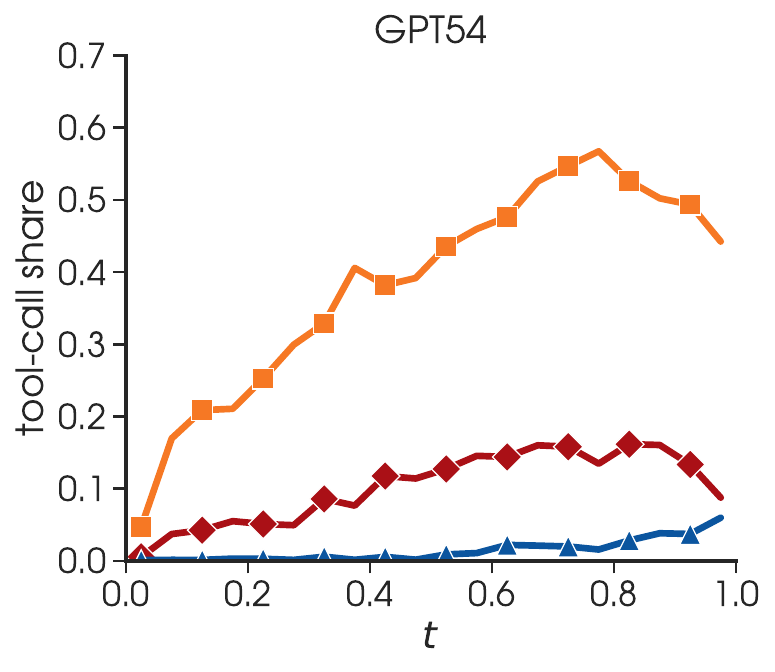} &
\includegraphics[width=0.31\linewidth]{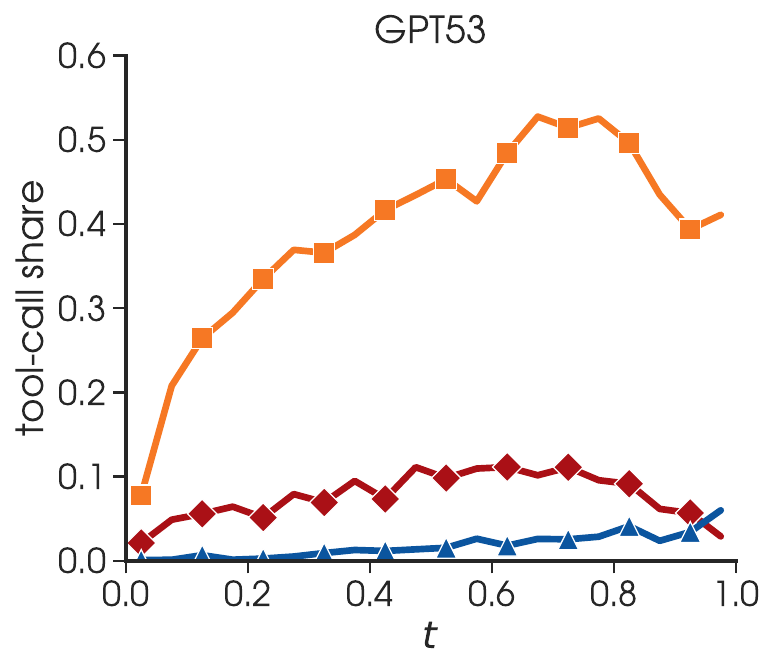} &
\includegraphics[width=0.31\linewidth]{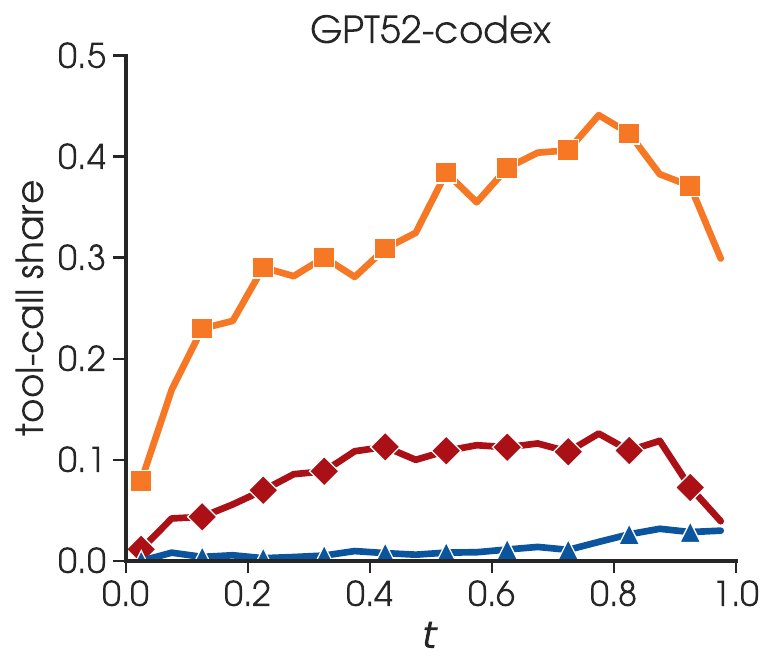} \\
\includegraphics[width=0.31\linewidth]{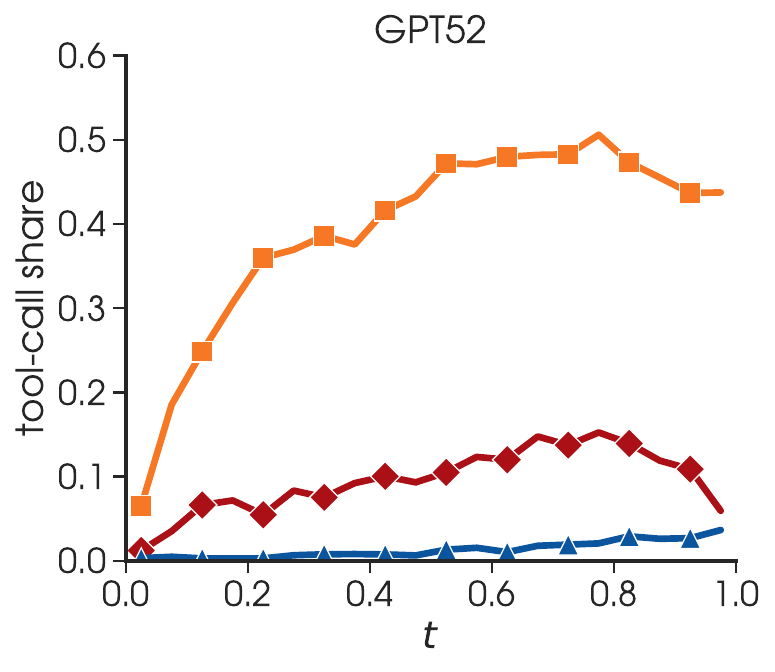} &
\includegraphics[width=0.31\linewidth]{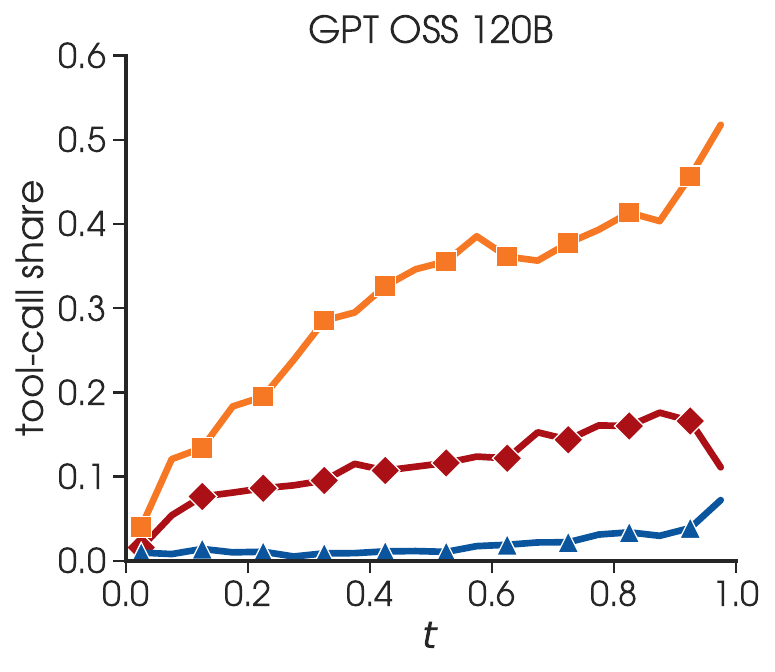} &
\includegraphics[width=0.31\linewidth]{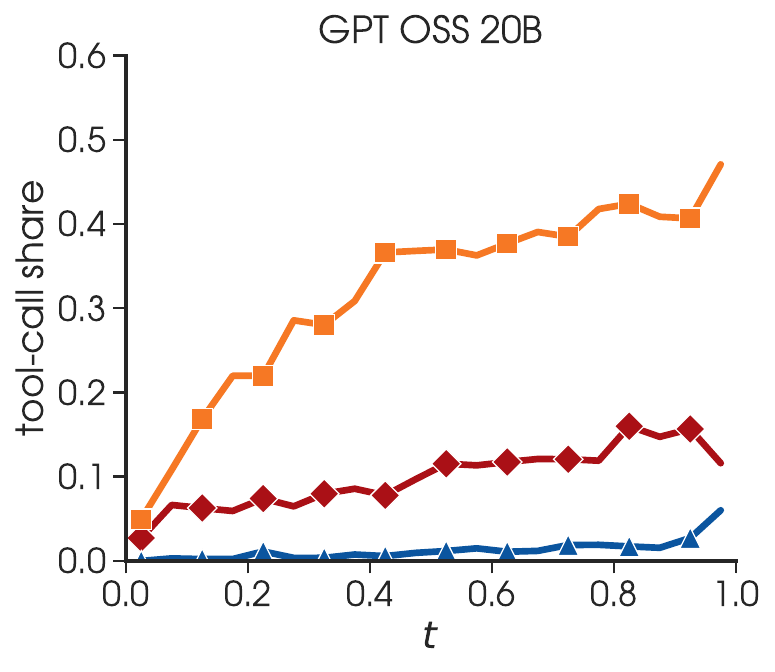} \\
\end{tabular}
\caption{Edit/test ratio per cycle on Terminal-Bench-2 (part 1 of 2): Anthropic Claude and OpenAI families. Each cell shows the model's distribution of ratios.}
\label{fig:metrics-tb2-editread}
\end{figure}

\begin{figure}[p]
\centering
\includegraphics[width=0.75\linewidth]{tb2/edit_read_ratio_legend.pdf}\\[2pt]
\setlength{\tabcolsep}{1pt}
\renewcommand{\arraystretch}{0.5}
\begin{tabular}{ccc}
\includegraphics[width=0.31\linewidth]{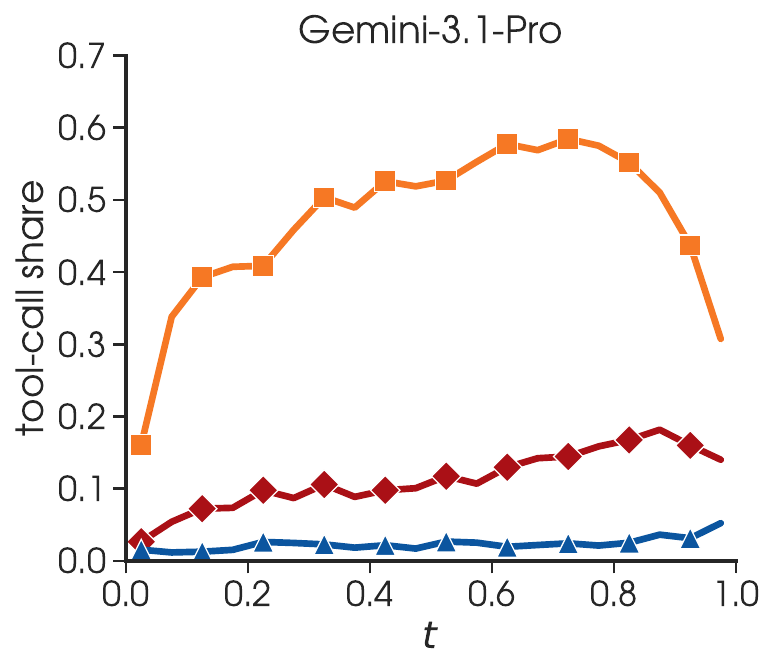} &
\includegraphics[width=0.31\linewidth]{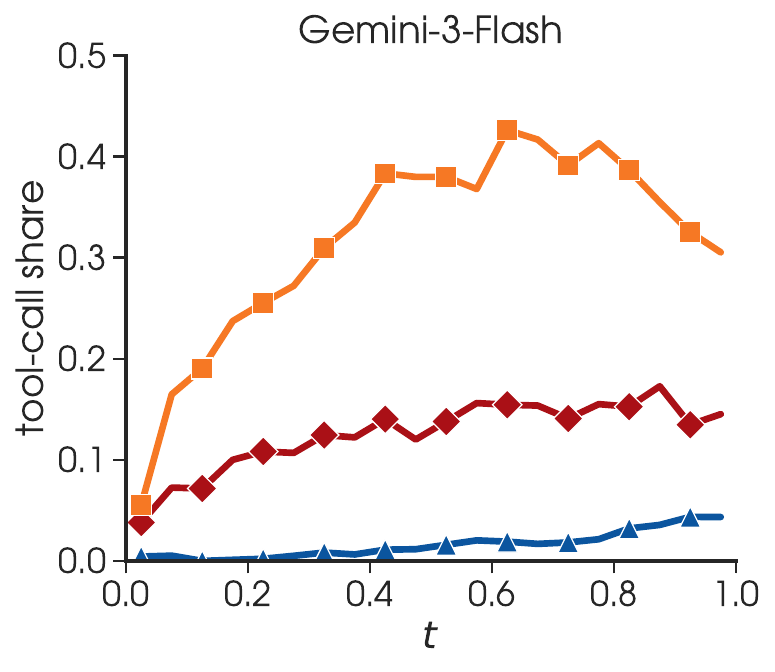} & \\
\includegraphics[width=0.31\linewidth]{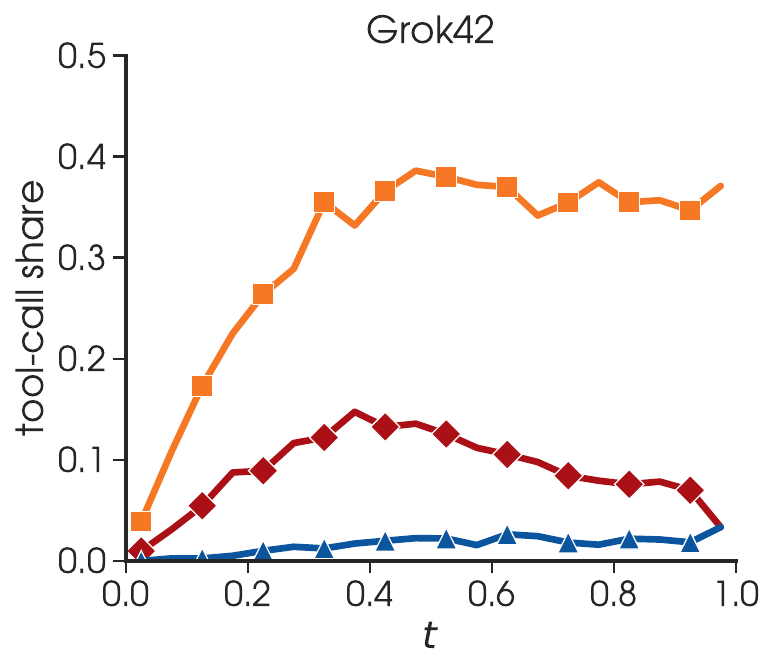} & & \\
\includegraphics[width=0.31\linewidth]{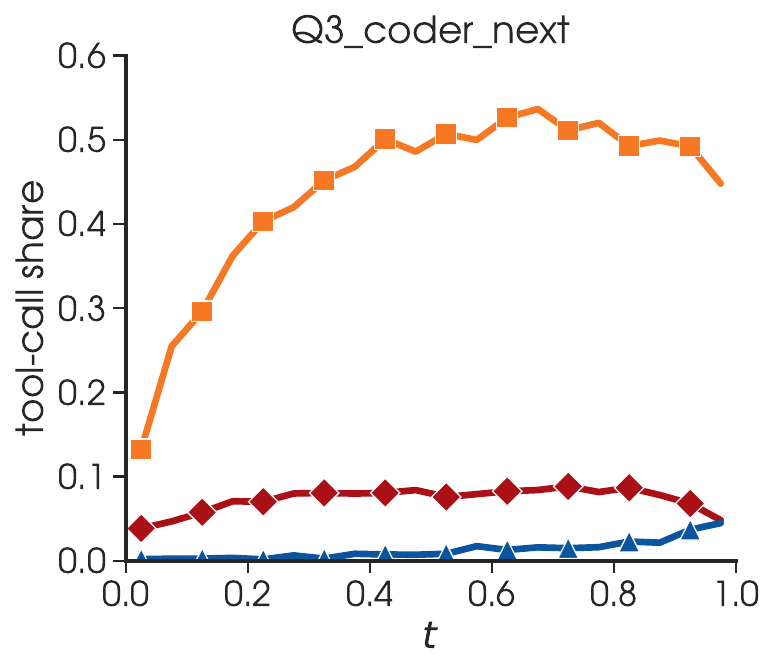} &
\includegraphics[width=0.31\linewidth]{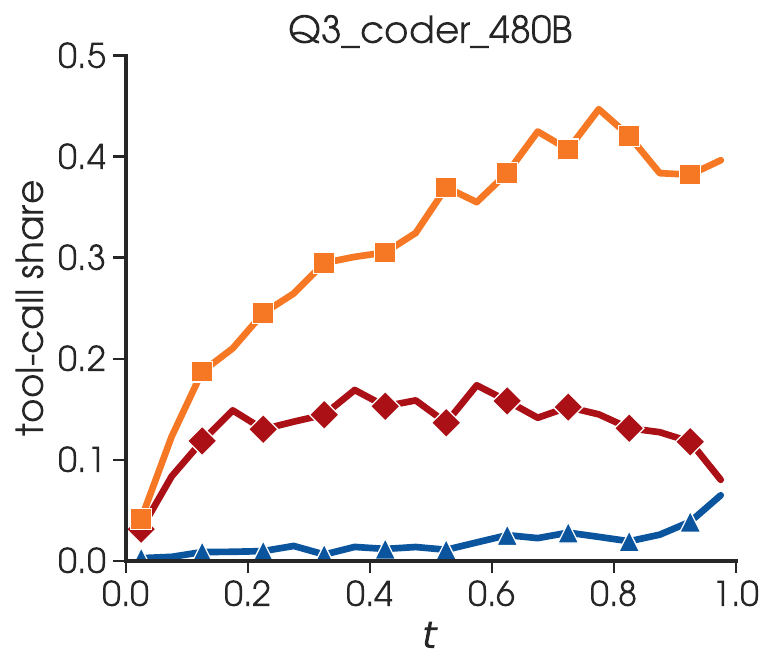} &
\includegraphics[width=0.31\linewidth]{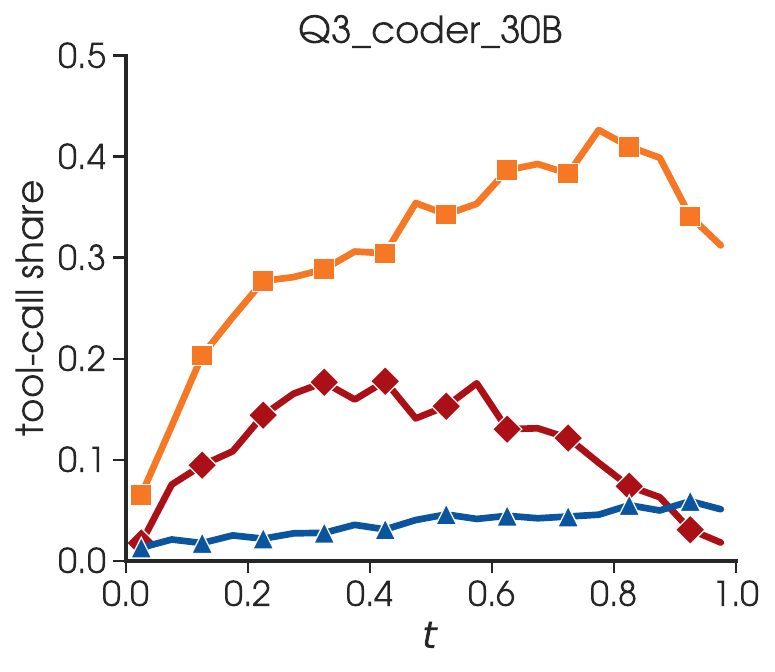} \\
\includegraphics[width=0.31\linewidth]{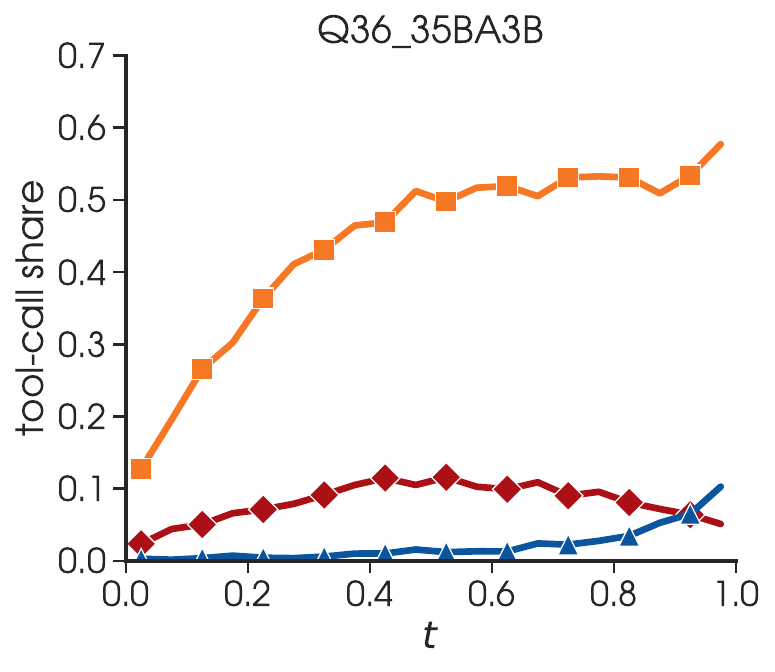} &
\includegraphics[width=0.31\linewidth]{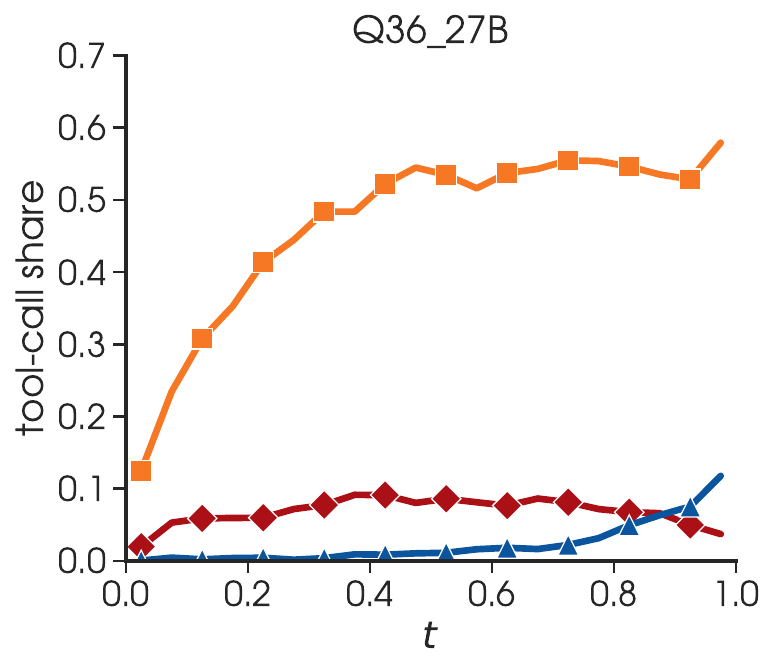} &
\includegraphics[width=0.31\linewidth]{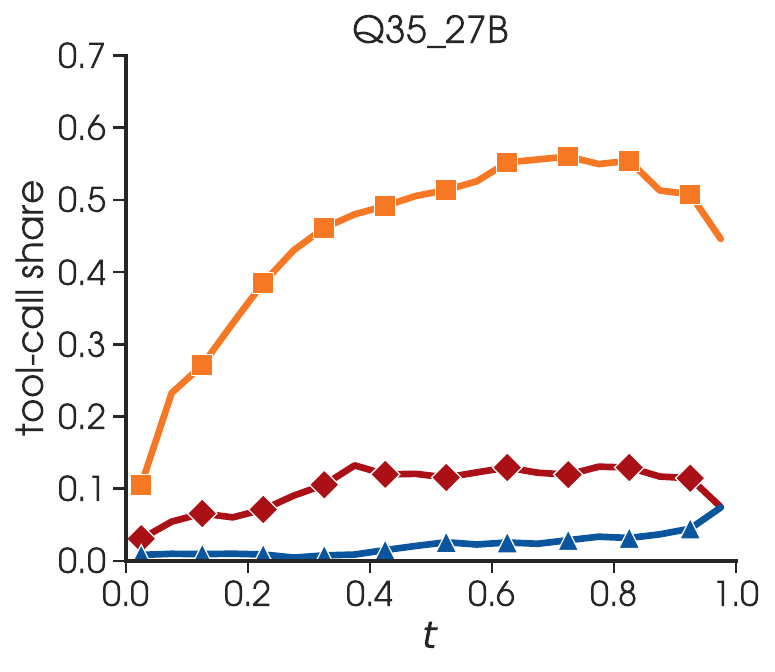} \\
\end{tabular}
\caption{Edit/test ratio per cycle on Terminal-Bench-2 (part 2 of 2): Gemini, Grok and Qwen families. Continued from Figure~\ref{fig:metrics-tb2-editread}.}
\label{fig:metrics-tb2-editread-b}
\end{figure}

\begin{figure}[p]
\centering
\includegraphics[width=0.85\linewidth]{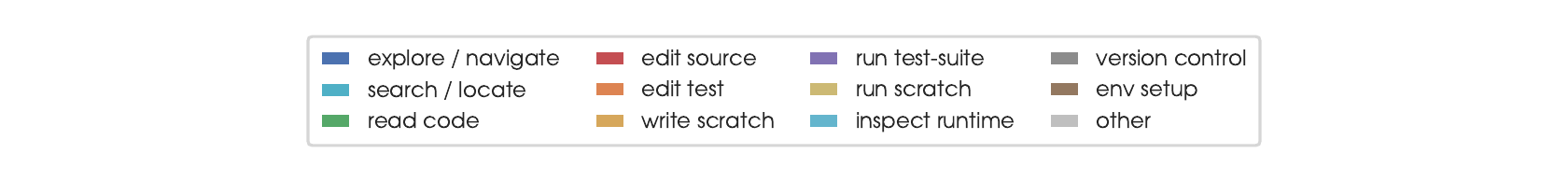}\\[2pt]
\setlength{\tabcolsep}{1pt}
\renewcommand{\arraystretch}{0.5}
\begin{tabular}{ccc}
\includegraphics[width=0.31\linewidth]{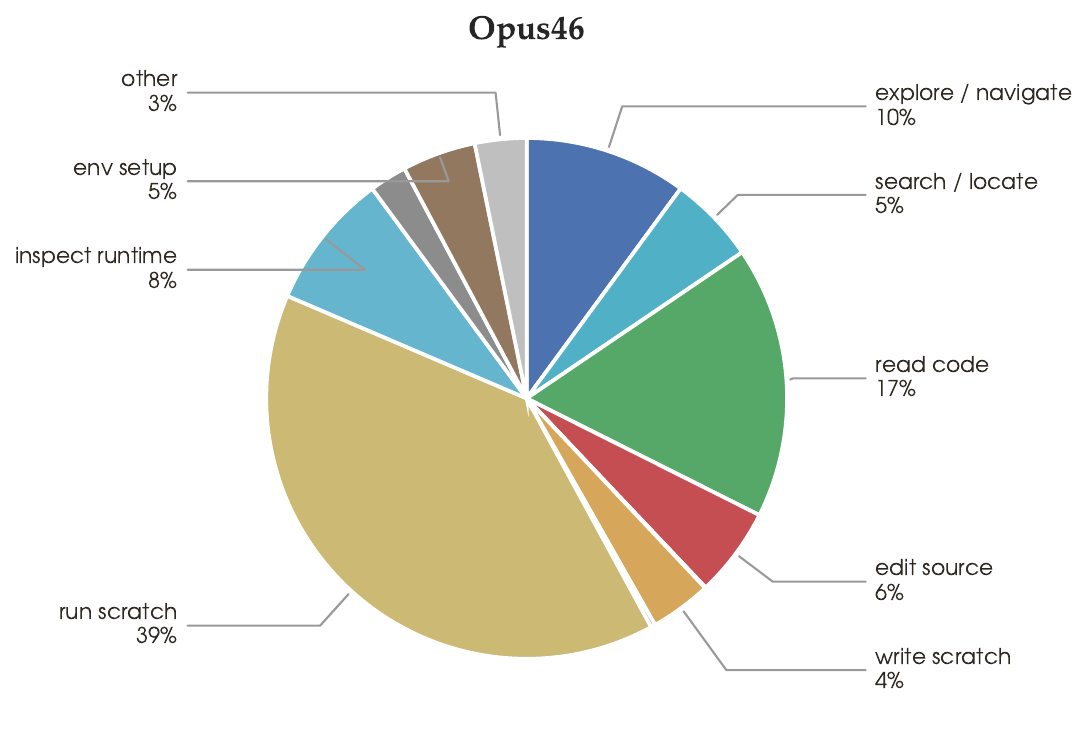} &
\includegraphics[width=0.31\linewidth]{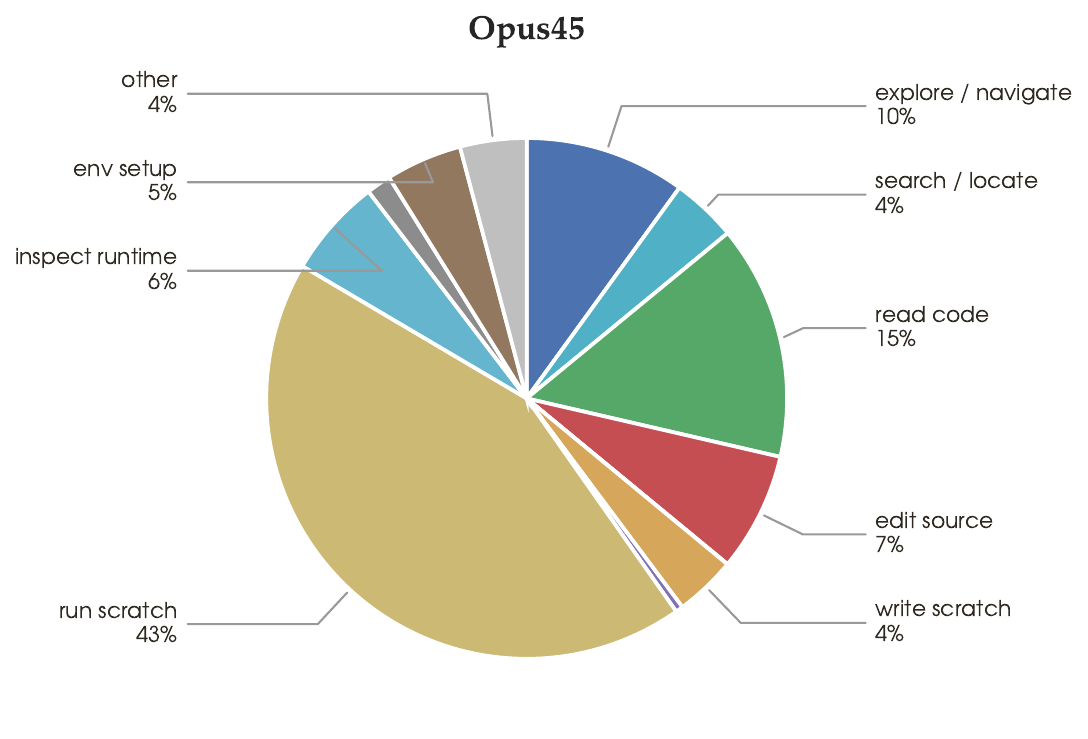} &
\includegraphics[width=0.31\linewidth]{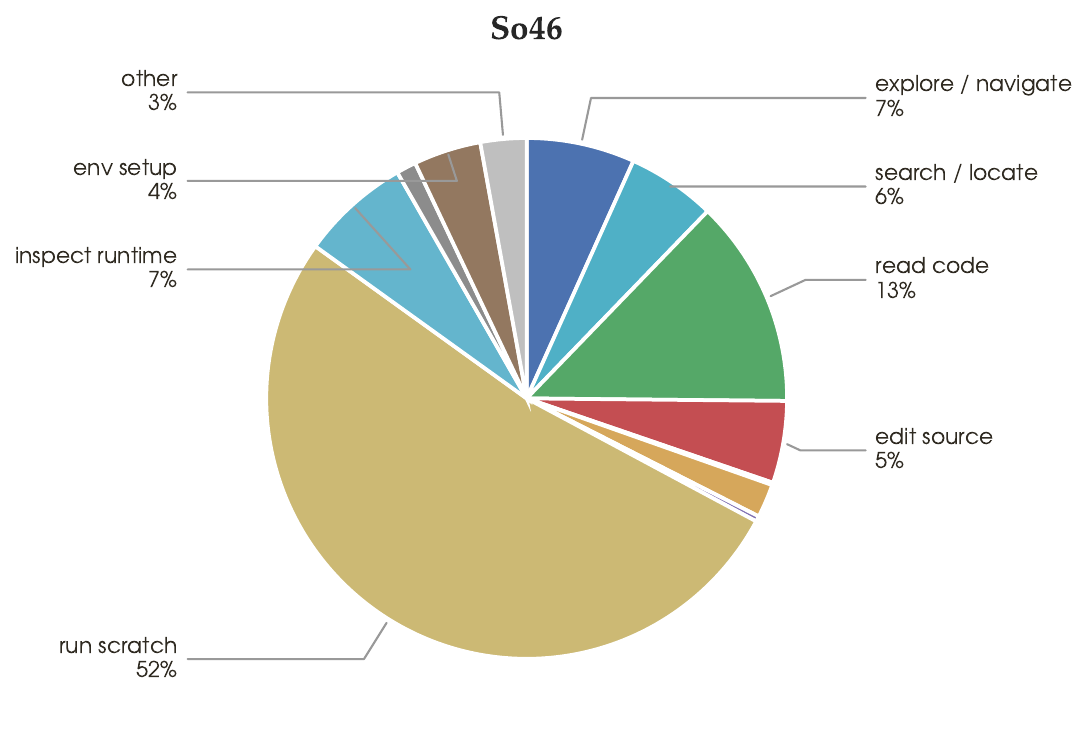} \\
\includegraphics[width=0.31\linewidth]{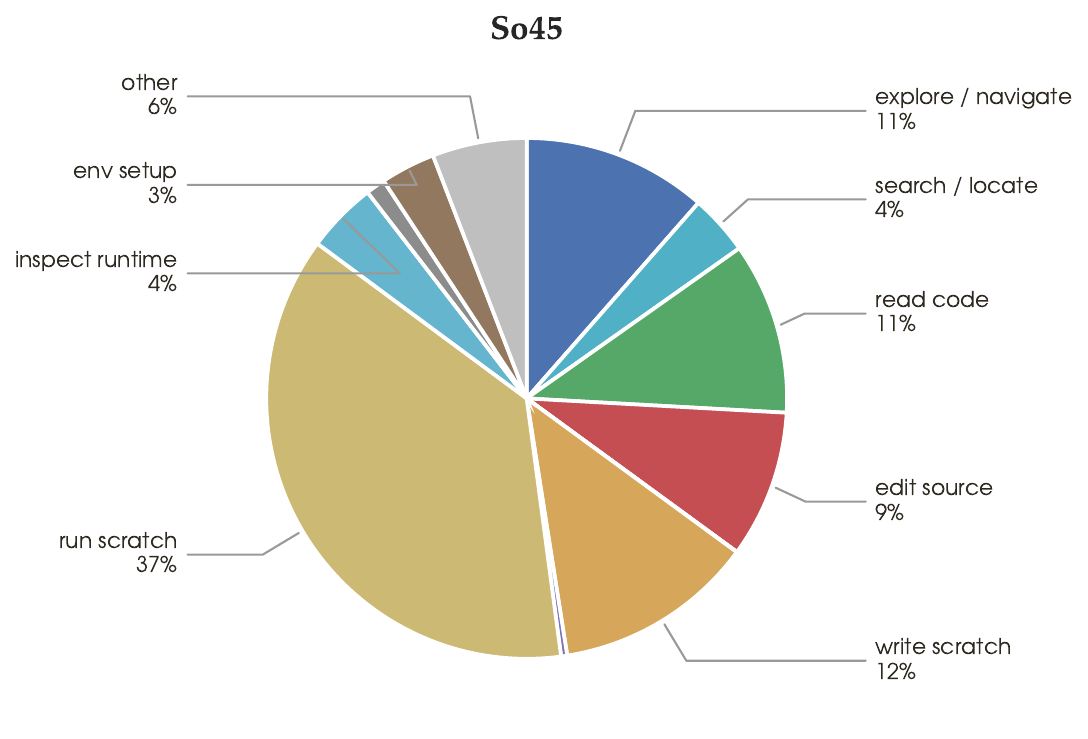} &
\includegraphics[width=0.31\linewidth]{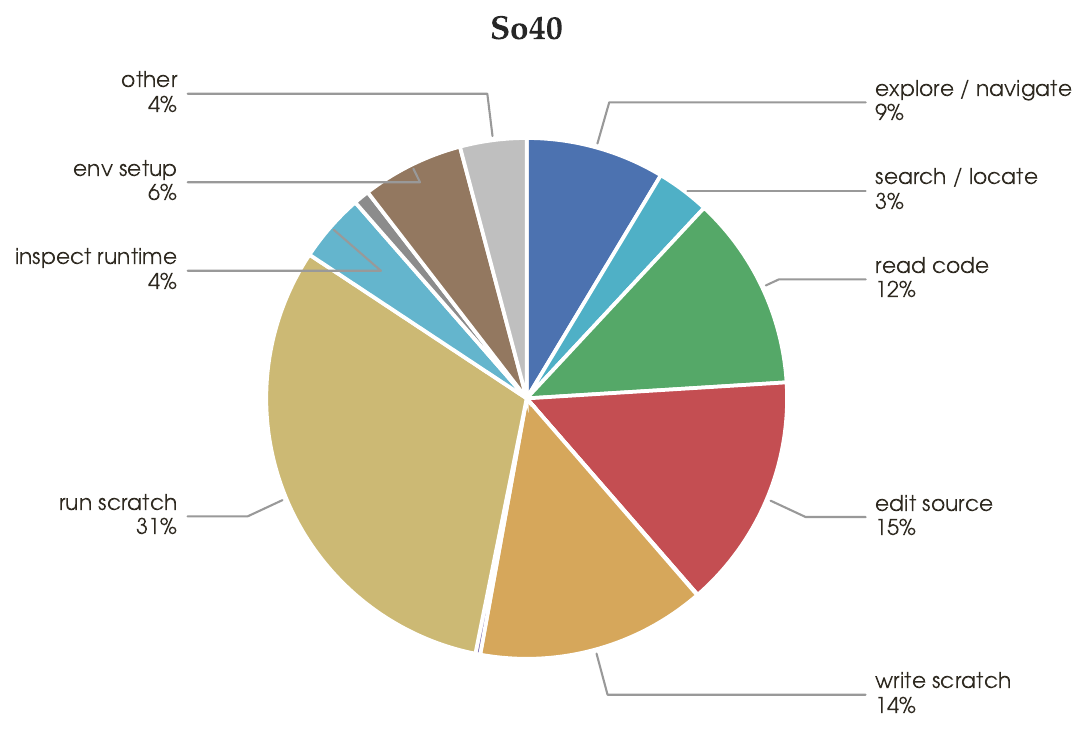} &
\includegraphics[width=0.31\linewidth]{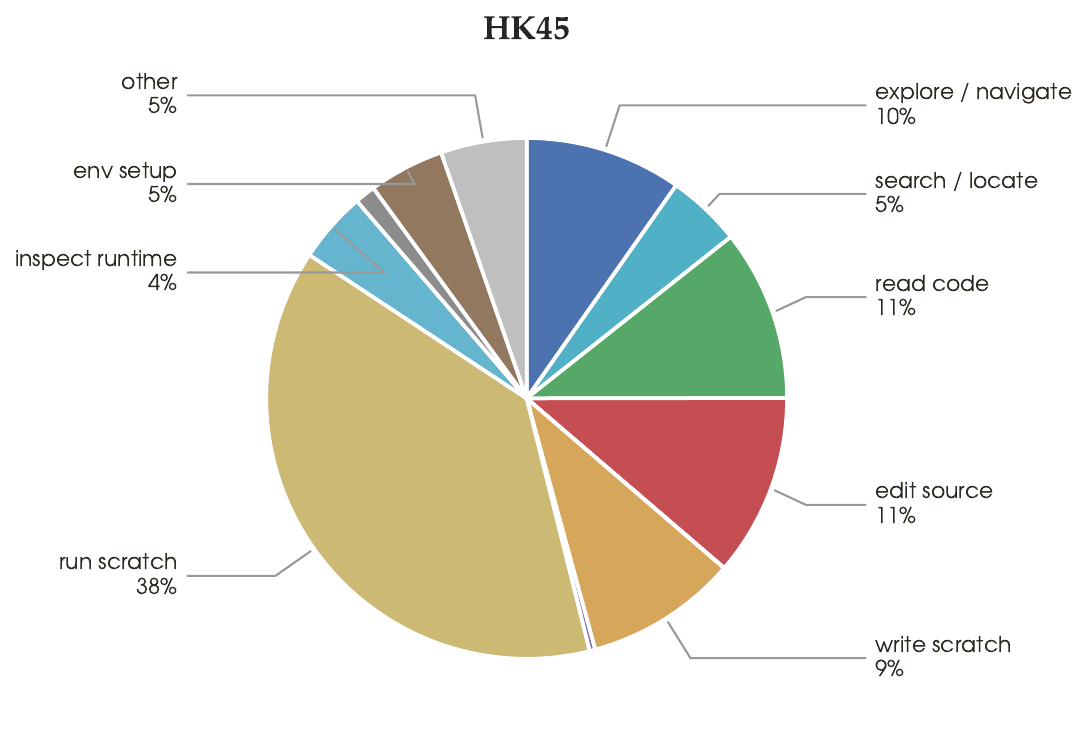} \\
\includegraphics[width=0.31\linewidth]{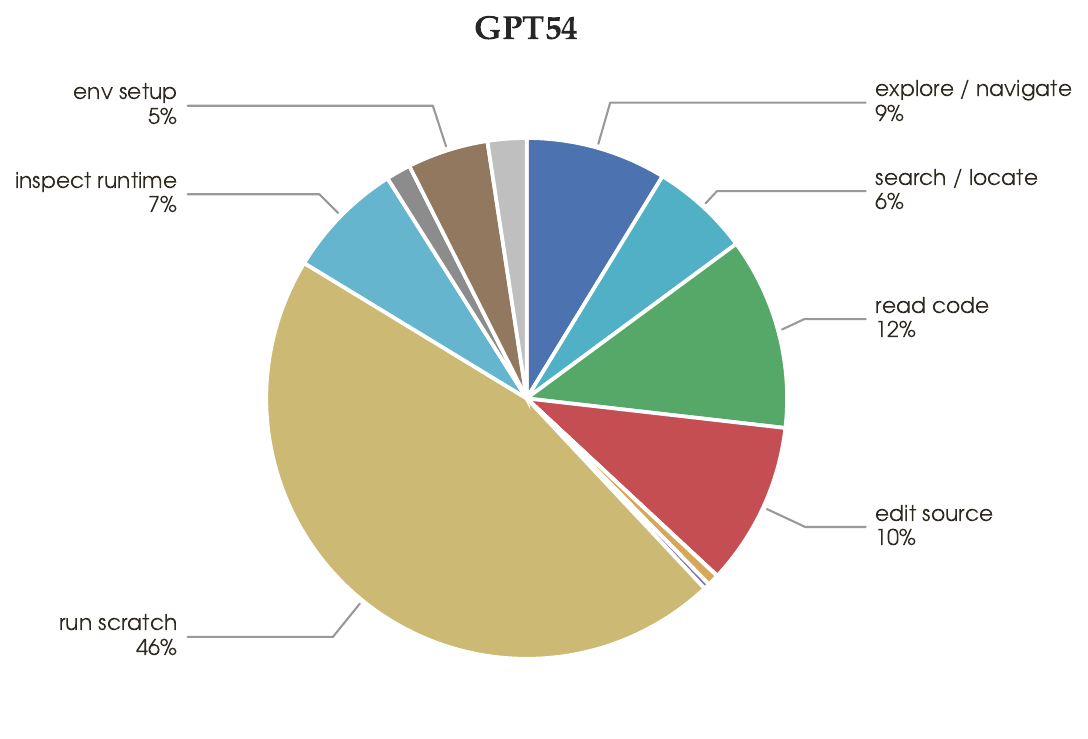} &
\includegraphics[width=0.31\linewidth]{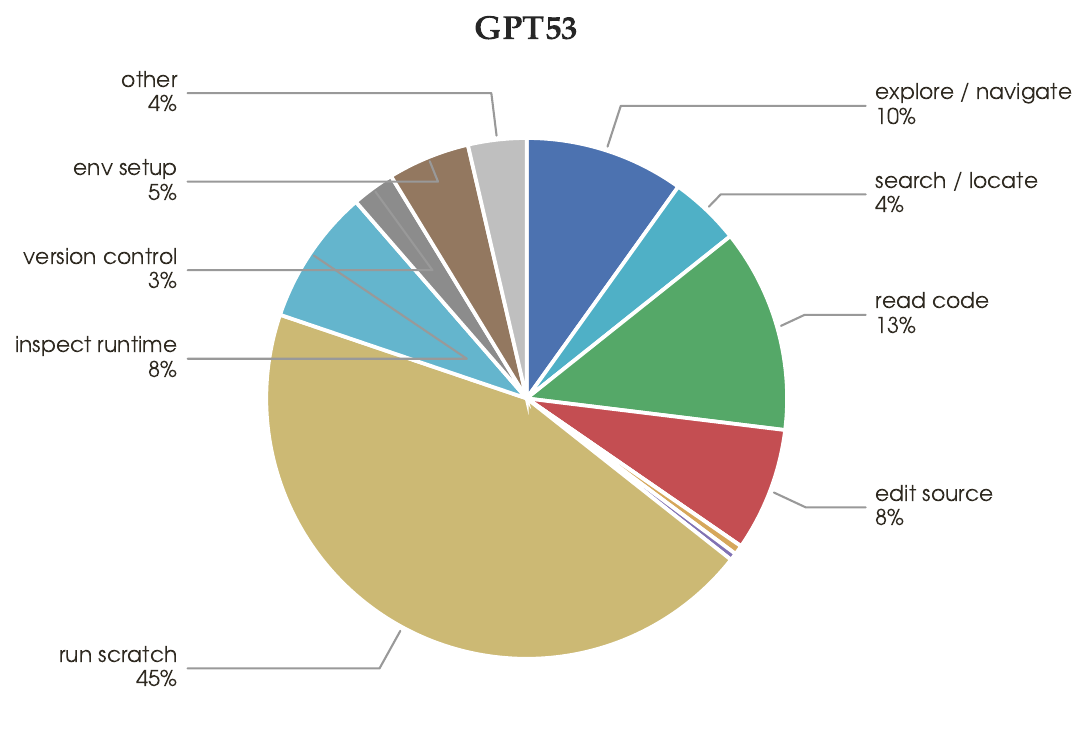} &
\includegraphics[width=0.31\linewidth]{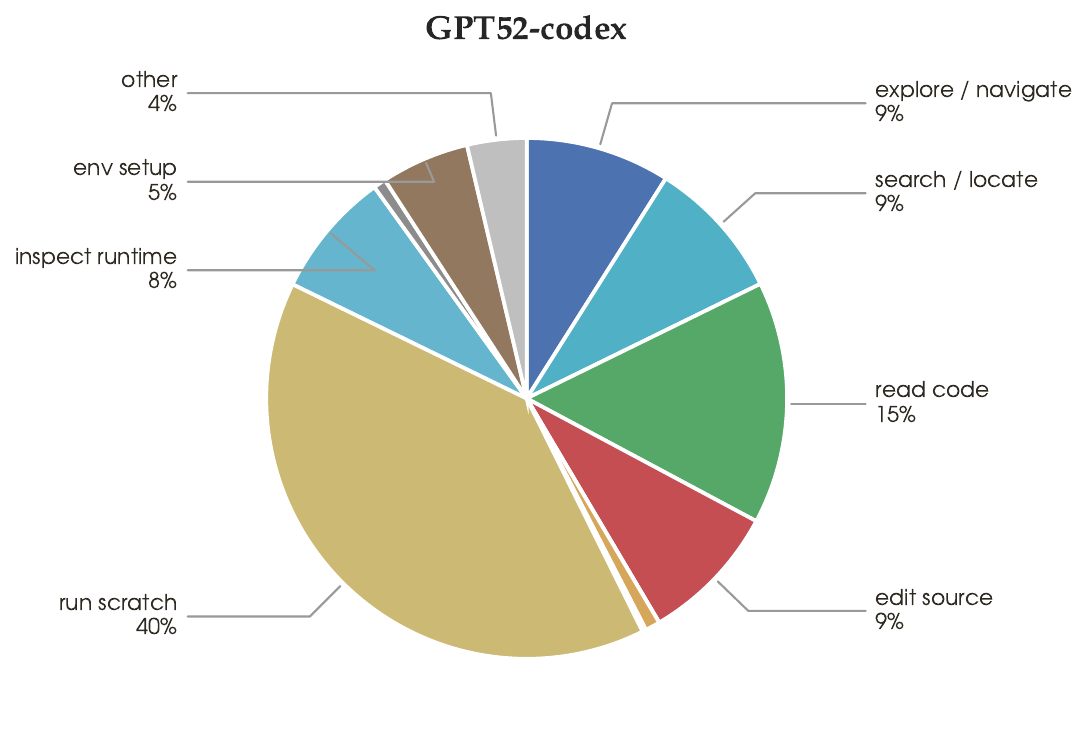} \\
\includegraphics[width=0.31\linewidth]{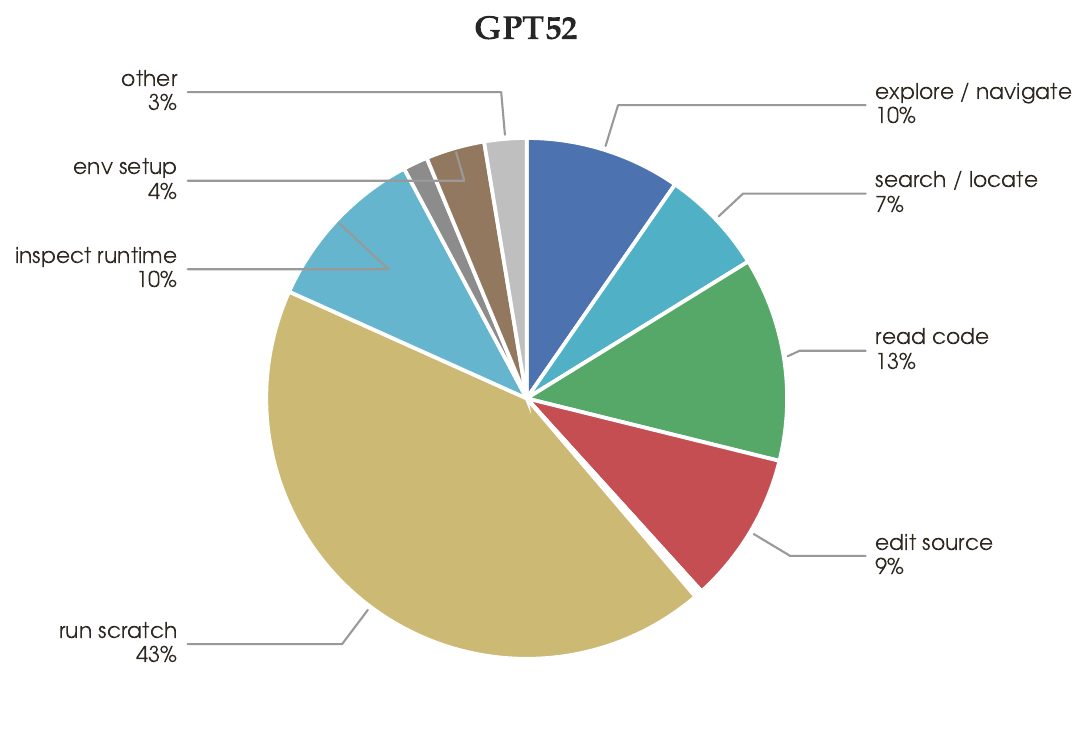} &
\includegraphics[width=0.31\linewidth]{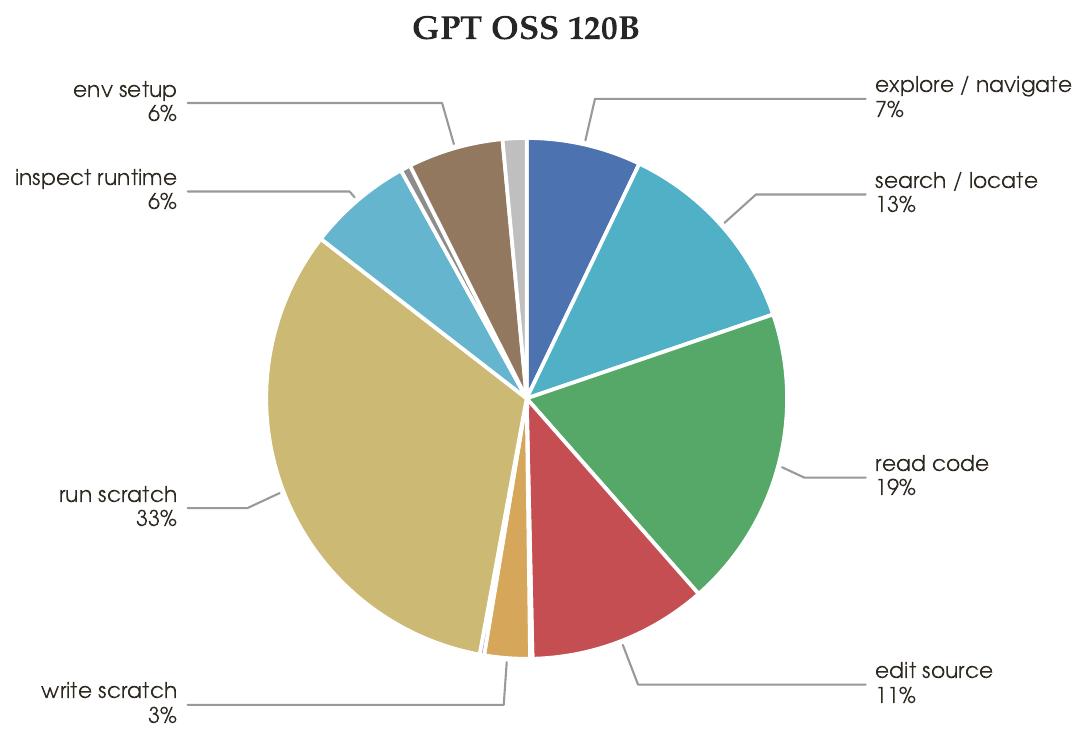} &
\includegraphics[width=0.31\linewidth]{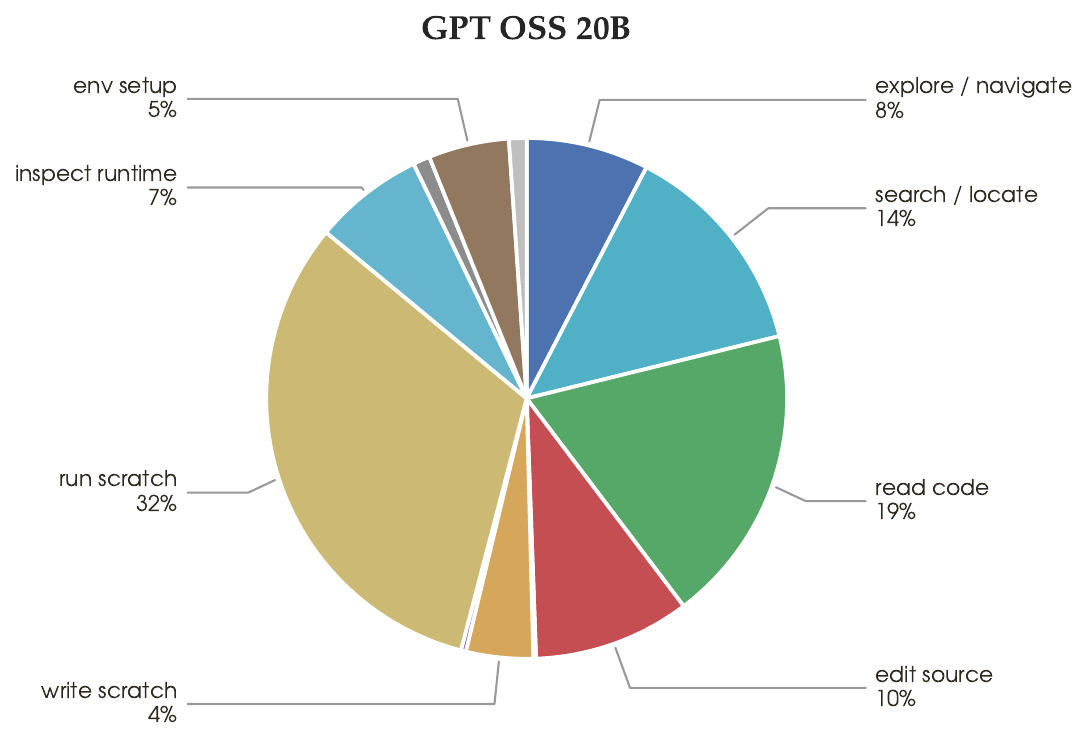} \\
\end{tabular}
\caption{Tool-call distribution per model on Terminal-Bench-2 (part 1 of 2): Anthropic Claude and OpenAI families. Each cell shows the share of calls in each R1 category (see Table\,\ref{tab:judge-buckets} for LLM judge rubrics) for one model.}
\label{fig:metrics-tb2-tooldist}
\end{figure}

\begin{figure}[p]
\centering
\includegraphics[width=0.85\linewidth]{tb2/tool_distribution_legend.pdf}\\[2pt]
\setlength{\tabcolsep}{1pt}
\renewcommand{\arraystretch}{0.5}
\begin{tabular}{ccc}
\includegraphics[width=0.31\linewidth]{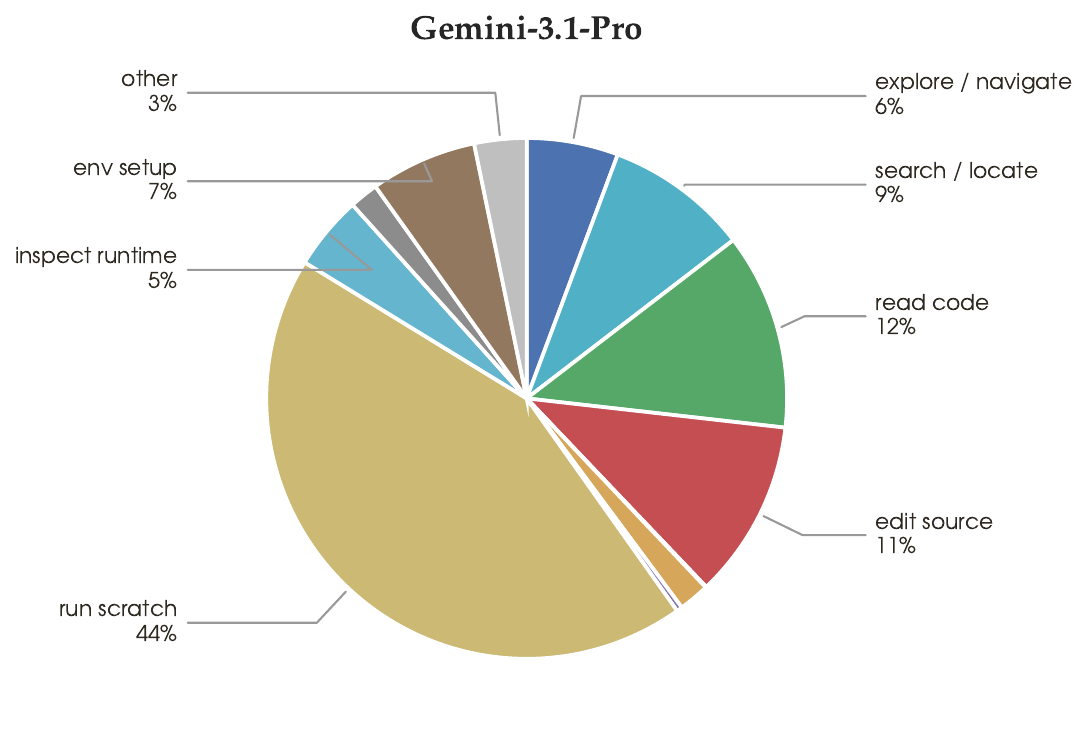} &
\includegraphics[width=0.31\linewidth]{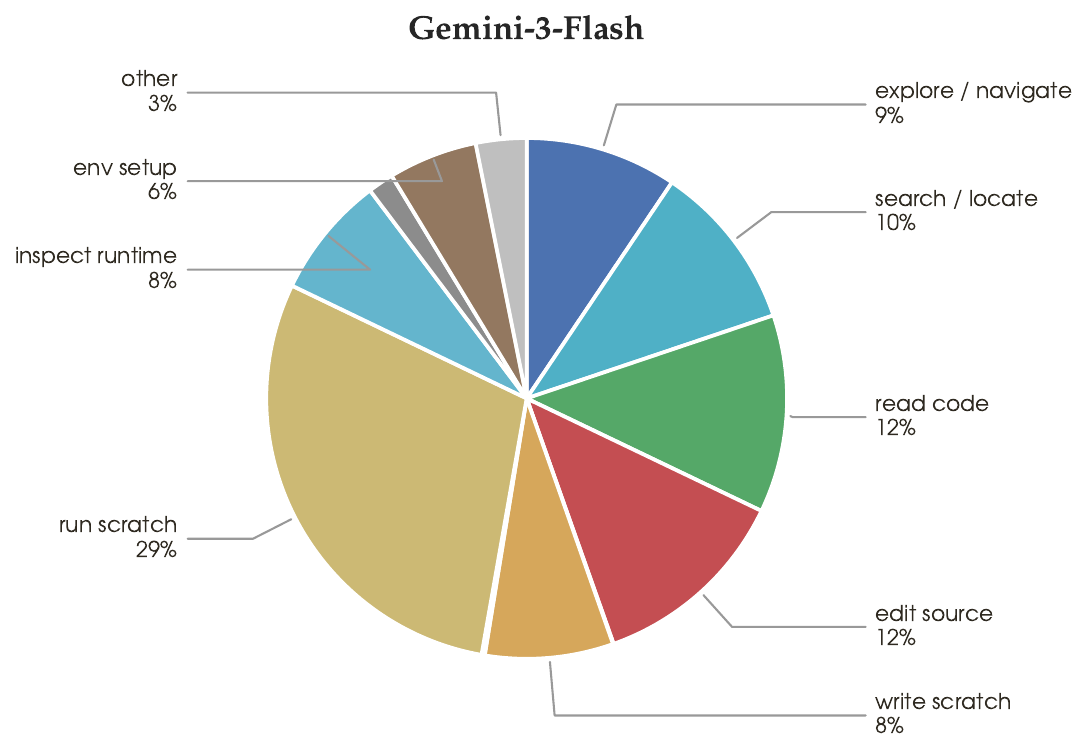} & \\
\includegraphics[width=0.31\linewidth]{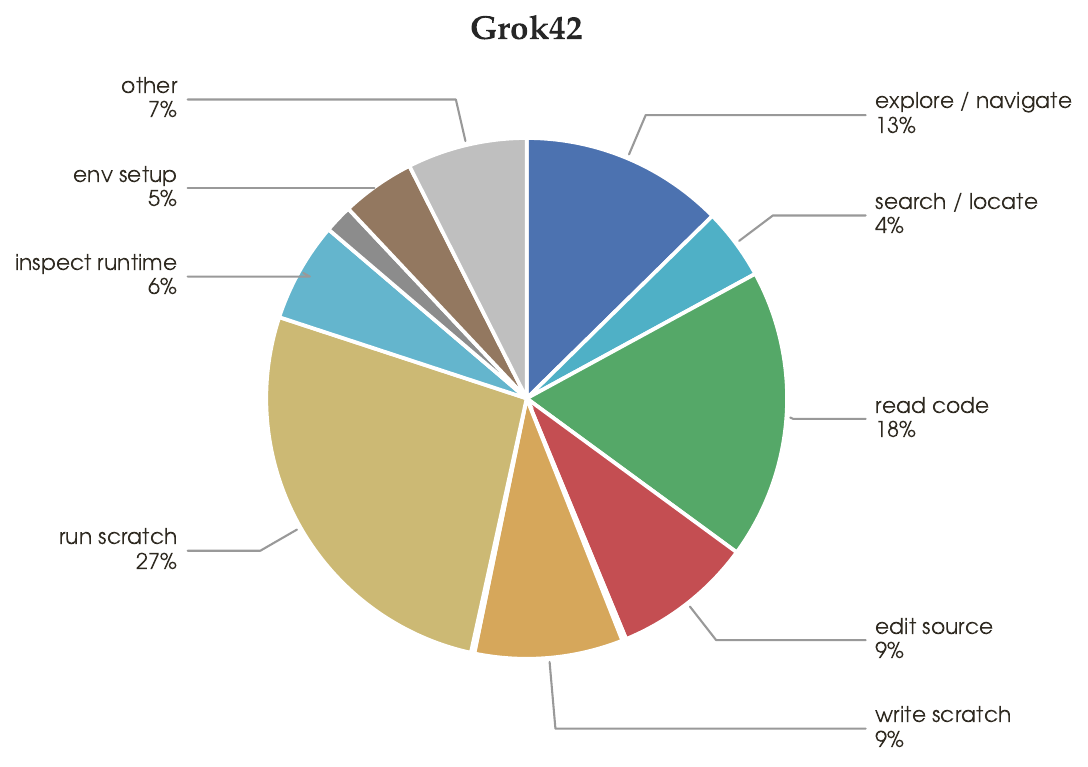} & & \\
\includegraphics[width=0.31\linewidth]{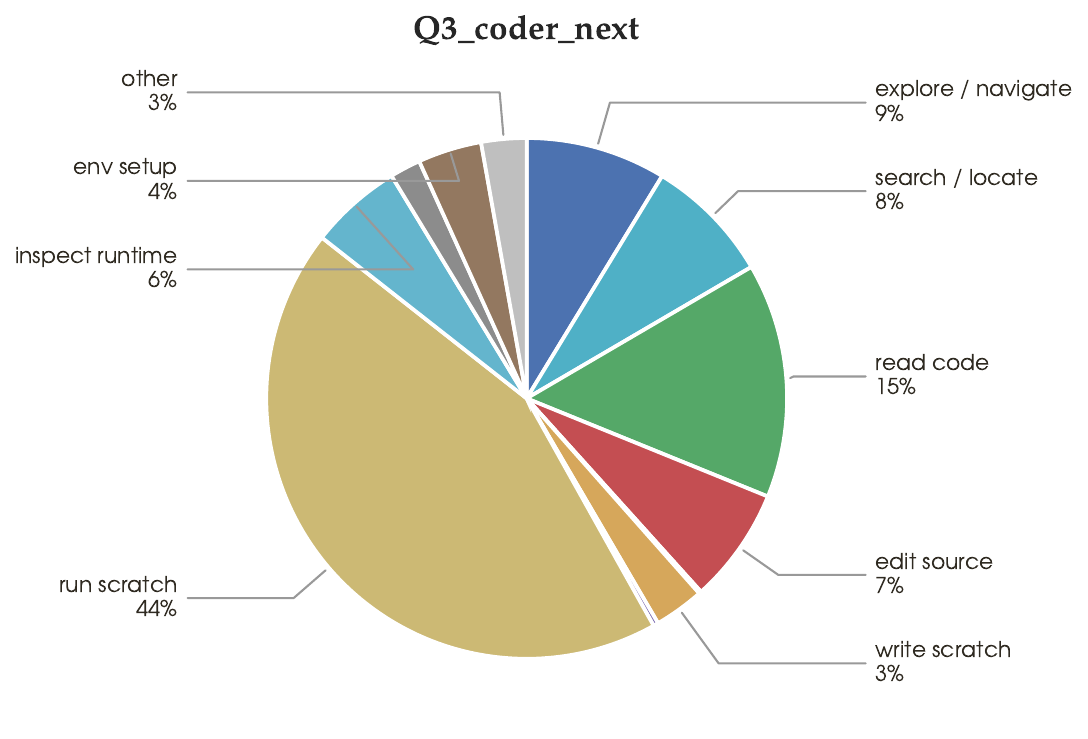} &
\includegraphics[width=0.31\linewidth]{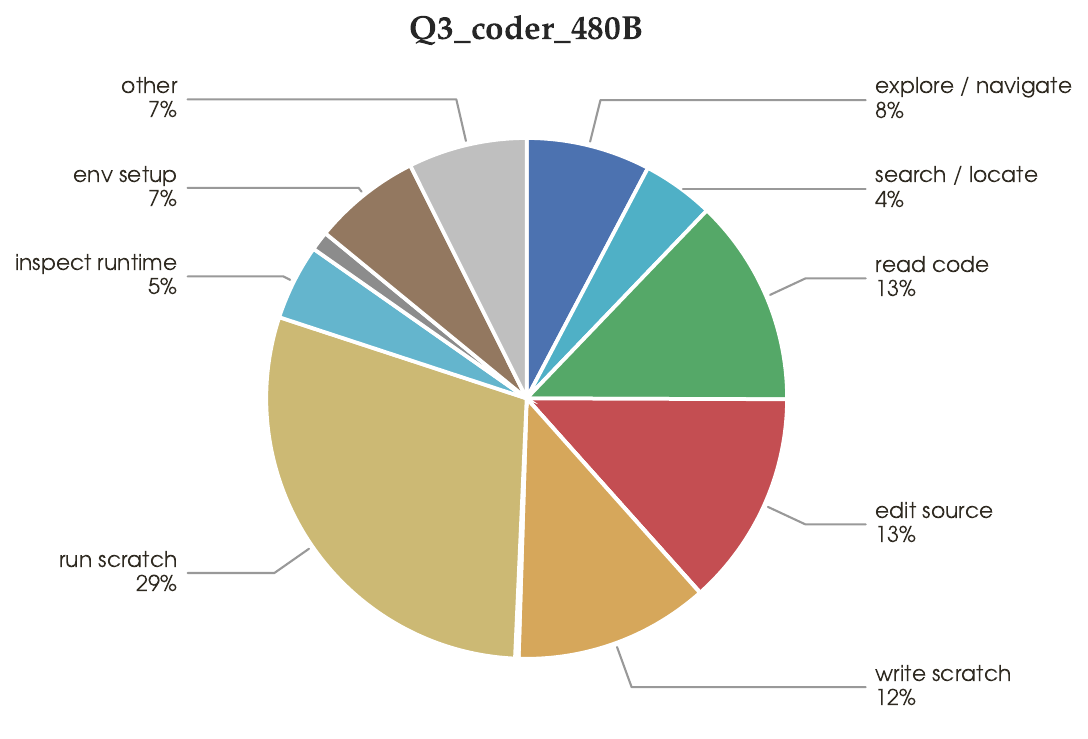} &
\includegraphics[width=0.31\linewidth]{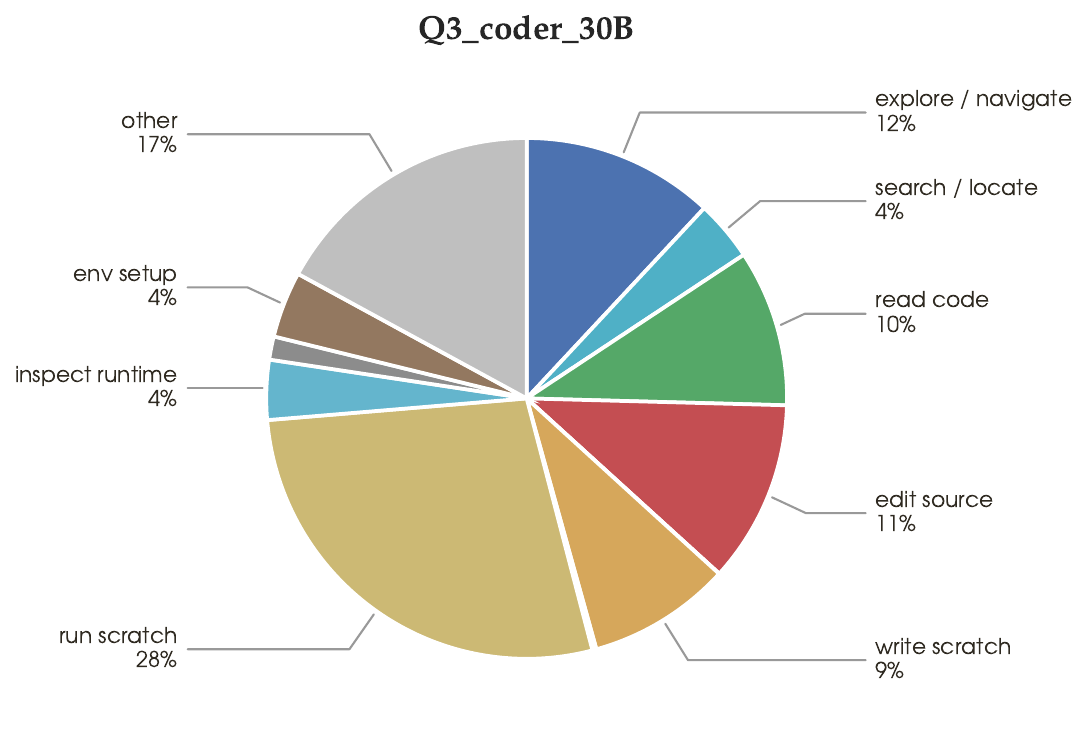} \\
\includegraphics[width=0.31\linewidth]{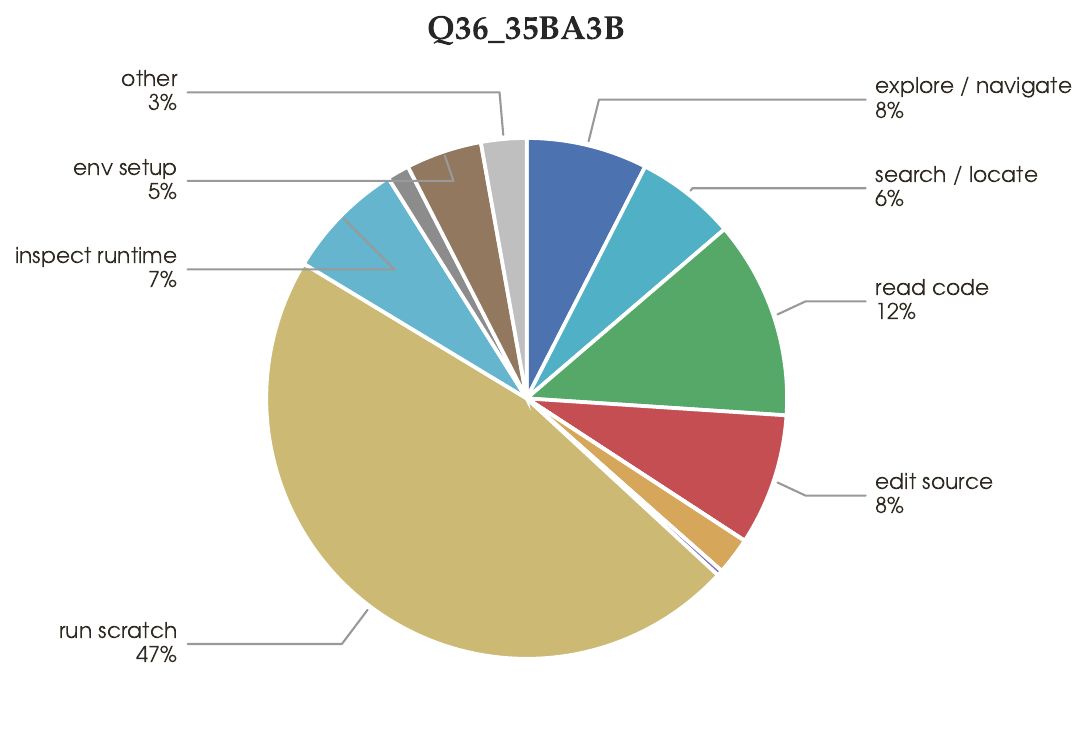} &
\includegraphics[width=0.31\linewidth]{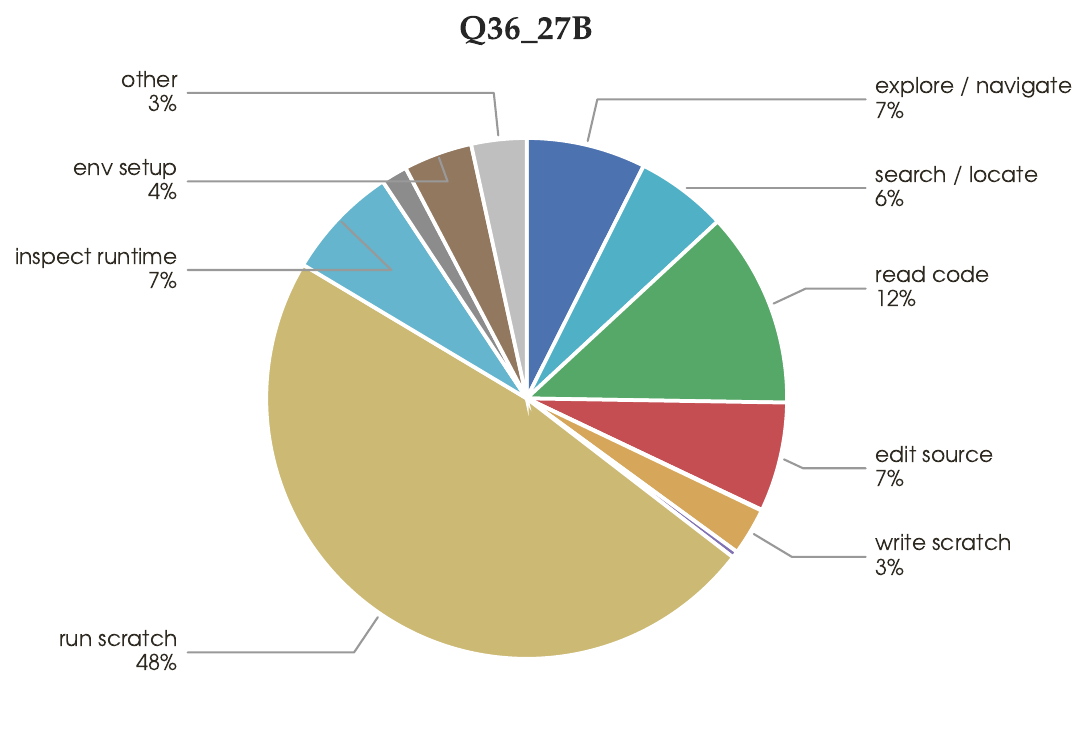} &
\includegraphics[width=0.31\linewidth]{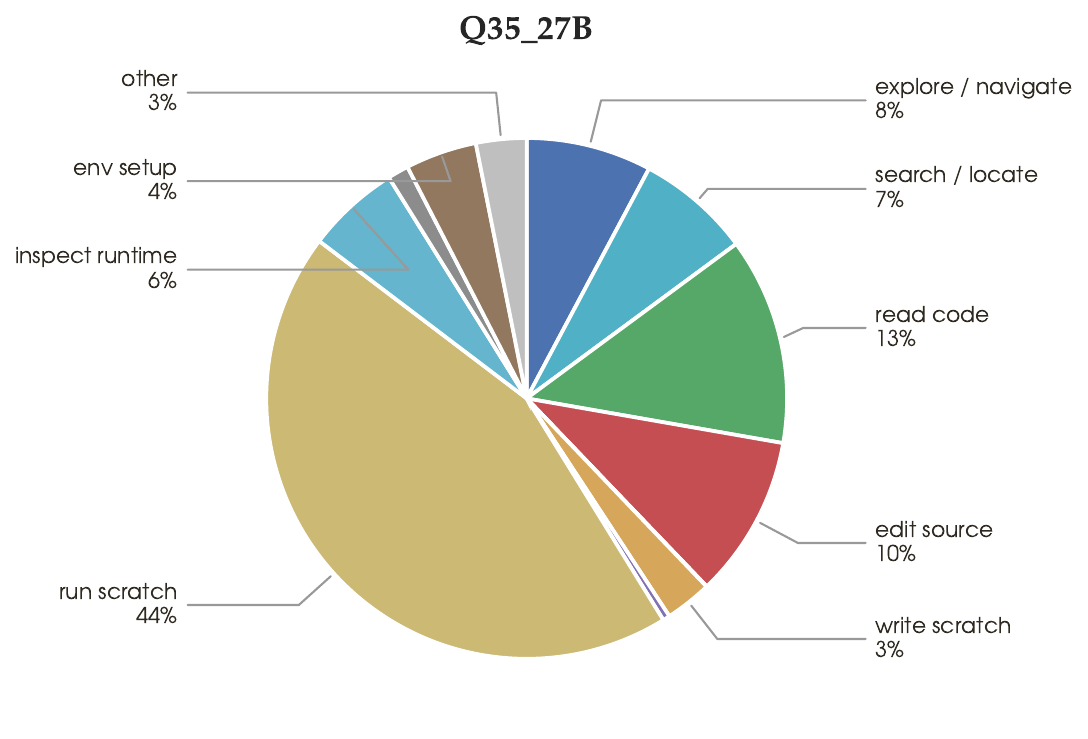} \\
\end{tabular}
\caption{Tool-call distribution per model on Terminal-Bench-2 (part 2 of 2): Gemini, Grok and Qwen families. Continued from Figure~\ref{fig:metrics-tb2-tooldist}.}
\label{fig:metrics-tb2-tooldist-b}
\end{figure}

Figures~\ref{fig:metrics-tb2-phase}--\ref{fig:metrics-tb2-tooldist-b} present per-model behavioral breakdowns from Section~\ref{app:metrics:sbv} on Terminal-Bench-2. The two trajectory-replay panels ($\Delta D$ histograms and solution-distance curves) are omitted because TB2 does not expose per-instance gold patches in a unified-diff form, so the empirical solution manifold $\tilde{S}_i$ used in Sections~\ref{sec:distance}--\ref{app:metrics:sbv} cannot be constructed. The tool-use, phase and error metrics shown here are computed with the TB2-specific rubric variant introduced at the top of this section (Table~\ref{tab:judge-buckets-tb2}).

\paragraph{Per-model phase composition (TB2).}
Figures~\ref{fig:metrics-tb2-phase}--\ref{fig:metrics-tb2-phase-b} show how each TB2 trajectory's time is split across the TB2 R6 phases (Table~\ref{tab:judge-buckets-tb2}) --- \texttt{setup\_env} (install/configure the environment), \texttt{build} (run build tooling against upstream/system code), \texttt{explore} (orient and read docs/source), \texttt{implement} (write or patch the deliverable), \texttt{verify} (run the container/self tests), and \texttt{cleanup} (post-solve teardown). TB2 tasks are heavier on \texttt{setup\_env} / \texttt{build} than SWE-Bench instances, which have no analogous phases.

\paragraph{Per-model genuine-error rate over cycles (TB2).}
Figures~\ref{fig:metrics-tb2-toolerr}--\ref{fig:metrics-tb2-toolerr-b} plot the rate of \emph{genuine} tool errors per cycle (R8 $=$ \texttt{genuine}, see Section~\ref{app:metrics:rubric}), broken out by tool family. Benign non-zero exits (\texttt{grep} no-match, expected non-zero exit codes, etc.)\ are excluded so spikes here reflect real harness friction, not task outcomes.

\paragraph{Per-model edit/read ratio (TB2).}
Figures~\ref{fig:metrics-tb2-editread}--\ref{fig:metrics-tb2-editread-b} shows the per-cycle ratio of edit-oriented tool calls to total
ones across resolved + unresolved trajectories. A high ratio means the agent mostly edits and a low ratio means it mostly explores/test.

\paragraph{Per-model tool-call distribution (TB2).}
Figures~\ref{fig:metrics-tb2-tooldist}--\ref{fig:metrics-tb2-tooldist-b} show how each model's tool calls split across the R1 categories on Terminal-Bench-2. The categories are identical to the SWE figures, with \texttt{env\_setup} and \texttt{inspect\_runtime} carrying notably more mass on TB2 as the containers often do not contain all the dependencies (e.g. \texttt{pytorch}, \texttt{R} etc) required to solve the task.

\clearpage
\subsection{Metrics: Deep-SWE}
\label{app:metrics:deep}

The figures in this section show the per-model behavioral metrics of Appendix~\ref{app:metrics:sbv} for the 17 models evaluated on Deep-SWE, in the same order and format, so that each metric can be read against its SWE-Bench-Verified counterpart. Models are ordered by family, with families in alphabetical order: Claude (Opus 5.5, Opus 5.0, Opus 4.8, Sonnet 5.0, Fable 5), Gemini (3.8 Flash), GPT (GPT-6 Astra, Sol and Luna; GPT-5.6 Sol, Terra and Luna; GPT-5.5), Grok (4.7), Kimi (K3), Muse Spark (1.3) and Qwen (Qwen3.8 27B). Deep-SWE tasks require larger repairs than SWE-Bench-Verified, and the differences are visible in every metric: the patch is built over many edits rather than one, the level families fan out over most of the trajectory, and the backtracking-edit share is well below the SWE-Bench-Verified range for every model.

Table~\ref{tab:metrics-deep-distance-backtrack-summary} gives the same compact distance-and-local-backtracking summary on Deep-SWE that Table~\ref{tab:metrics-sbv-distance-backtrack-summary} reports on SWE-Bench-Verified. Local backtracking sits an order of magnitude below the SWE-Bench-Verified range ($0.2$--$2.0\%$ against $0.9$--$16.9\%$) and arrival is late for every model ($t$ @ $D\!\le\!0.1$ between $0.74$ and $0.95$), since a Deep-SWE patch is assembled over many edits rather than landed in one; the spread in AUC ($0.39$--$0.70$) is nonetheless as wide as on SWE-Bench-Verified.

\begin{table}[h]
\centering
\small
\caption{Distance, local backtracking and outcome summary over resolved Deep-SWE trajectories per model. Local backtrack \%, arrival-time $t$ @ $D\!\le\!0.1$ and area-under-curve (AUC) are computed from the faithful state-replay curves over $113$ instances (5 runs each) per model; pass@1 is from Appendix~\ref{app:passrates}. Families in alphabetical order.}
\label{tab:metrics-deep-distance-backtrack-summary}
\begin{tabular}{@{}lrrrr@{}}
\toprule
Model & Local backtrack \% & $t$ @ $D\!\le\!0.1$ & AUC & pass@1 \% \\
\midrule
Opus 5.5           &  2.0 & 0.95 & 0.58 & 74.33 \\
Opus 5.0           &  1.6 & 0.89 & 0.47 & 72.56 \\
Opus 4.8           &  1.2 & 0.79 & 0.50 & 61.28 \\
Sonnet 5.0         &  1.7 & 0.79 & 0.39 & 57.25 \\
Fable 5            &  1.1 & 0.79 & 0.47 & 68.84 \\
\midrule
Gemini 3.8 Flash   &  0.6 & 0.79 & 0.56 & 73.10 \\
\midrule
GPT-6-Astra        &  0.2 & 0.79 & 0.52 & 77.70 \\
GPT-6-Sol          &  0.6 & 0.84 & 0.48 & 76.10 \\
GPT-6-Luna         &  1.2 & 0.74 & 0.45 & 64.24 \\
GPT-5.6-Sol        &  1.1 & 0.74 & 0.46 & 72.56 \\
GPT-5.6-Terra      &  1.2 & 0.74 & 0.45 & 69.20 \\
GPT-5.6-Luna       &  1.4 & 0.79 & 0.43 & 65.13 \\
GPT-5.5            &  0.5 & 0.74 & 0.47 & 74.51 \\
\midrule
Grok 4.7           &  1.2 & 0.95 & 0.70 & 72.39 \\
\midrule
Kimi-K3            &  0.9 & 0.79 & 0.45 & 77.00 \\
\midrule
Muse Spark 1.3     &  0.5 & 0.79 & 0.54 & 75.22 \\
\midrule
Qwen3.8 27B        &  1.3 & 0.84 & 0.44 & 44.77 \\
\bottomrule
\end{tabular}
\end{table}

\begin{figure}[p]
\centering
\setlength{\tabcolsep}{1pt}
\renewcommand{\arraystretch}{0.5}
\begin{tabular}{ccc}
\includegraphics[width=0.31\linewidth]{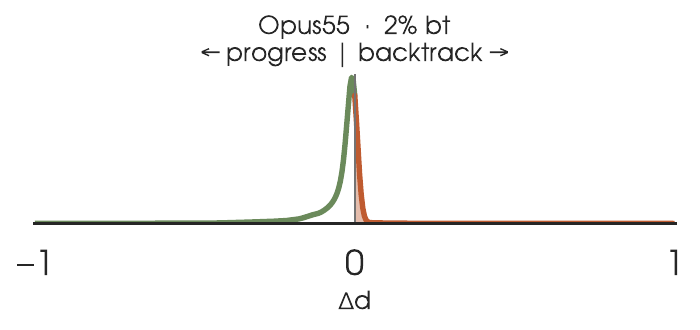} &
\includegraphics[width=0.31\linewidth]{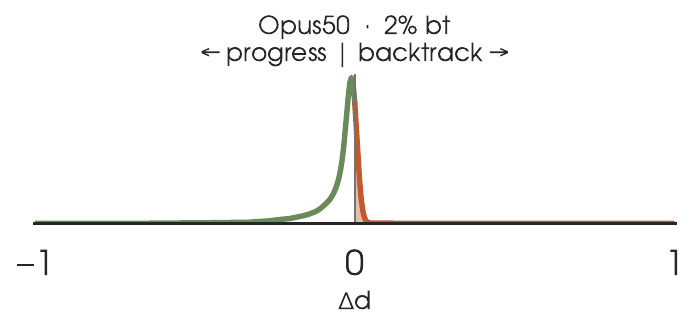} &
\includegraphics[width=0.31\linewidth]{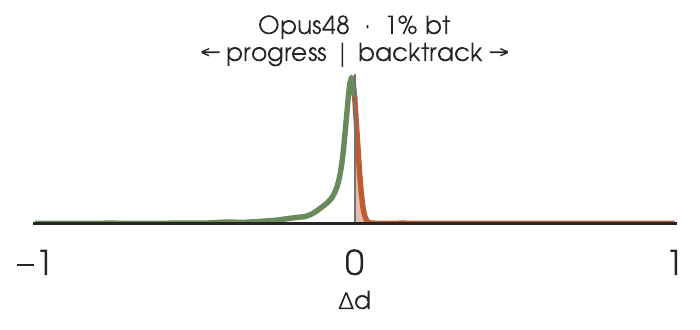} \\
\includegraphics[width=0.31\linewidth]{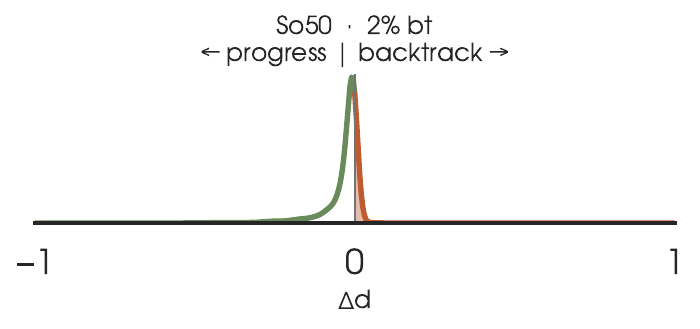} &
\includegraphics[width=0.31\linewidth]{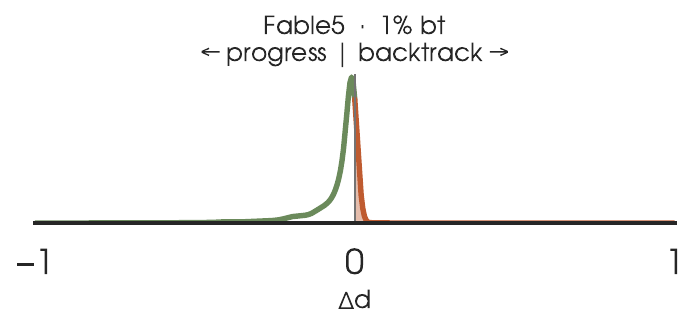} &
\includegraphics[width=0.31\linewidth]{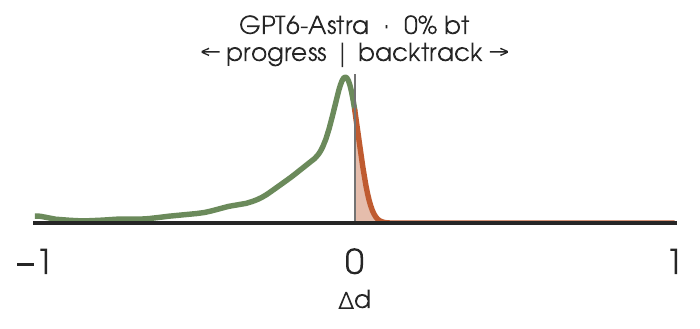} \\
\includegraphics[width=0.31\linewidth]{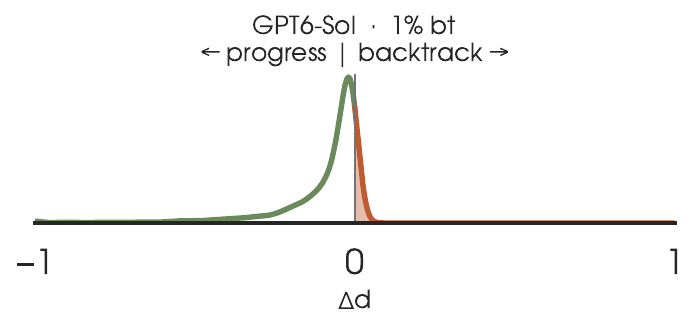} &
\includegraphics[width=0.31\linewidth]{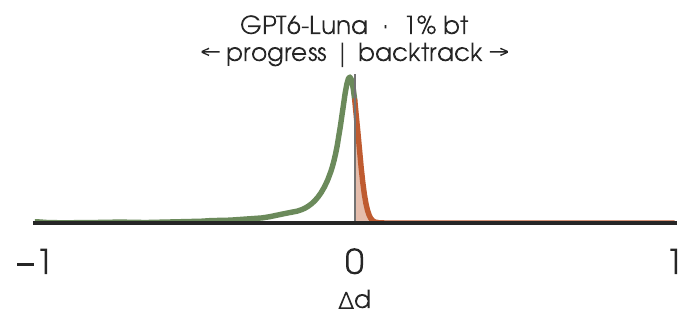} &
\includegraphics[width=0.31\linewidth]{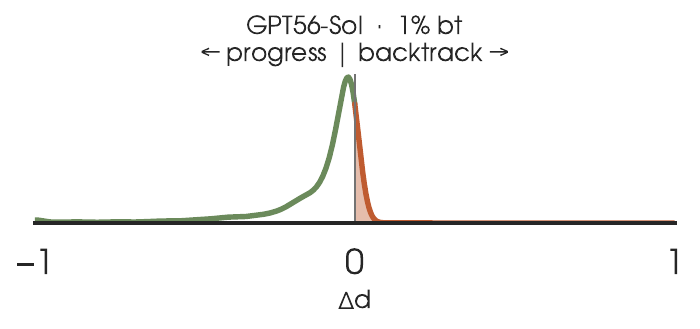} \\
\includegraphics[width=0.31\linewidth]{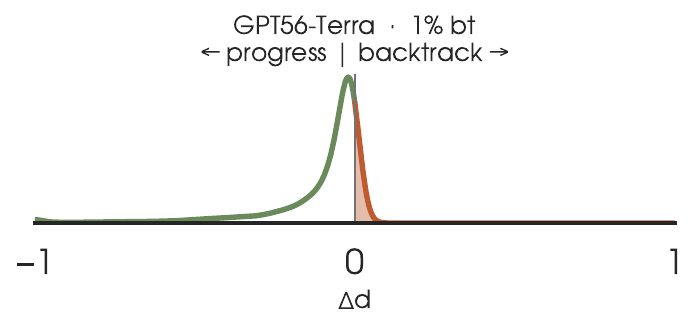} &
\includegraphics[width=0.31\linewidth]{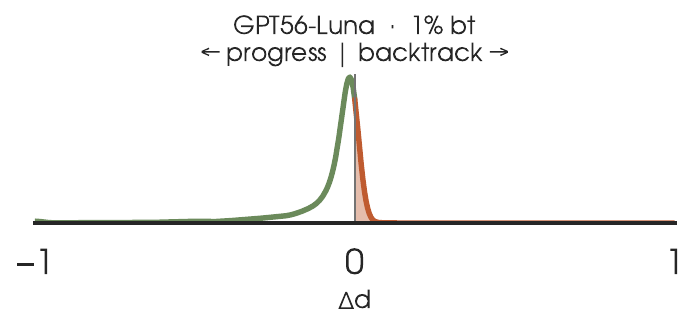} &
\includegraphics[width=0.31\linewidth]{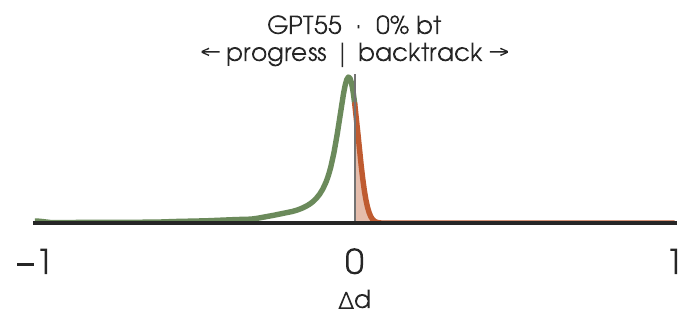} \\
\end{tabular}
\caption{\textbf{Local backtracking.} Histograms of $\Delta D$ on Deep-SWE (part 1 of 2): Claude and GPT families. Each cell is one model, green = progress edits, red = local backtracking edits. Annotation shows the per-model local-backtrack share.}
\label{fig:metrics-deep-backtracking}
\end{figure}

\begin{figure}[p]
\centering
\setlength{\tabcolsep}{1pt}
\renewcommand{\arraystretch}{0.5}
\begin{tabular}{ccc}
\includegraphics[width=0.31\linewidth]{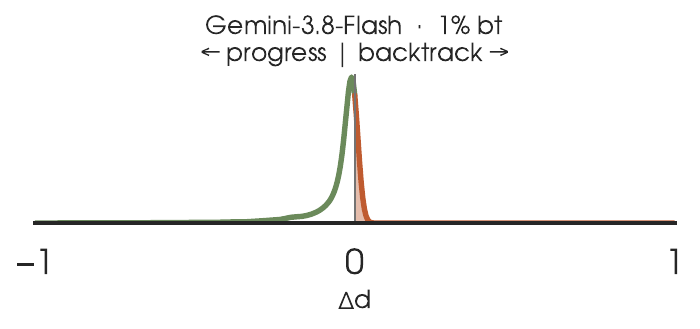} &
\includegraphics[width=0.31\linewidth]{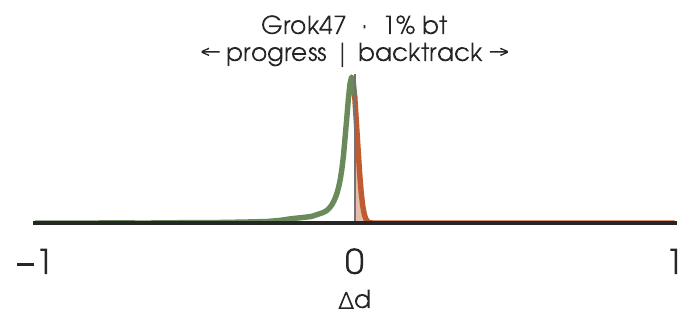} &
\includegraphics[width=0.31\linewidth]{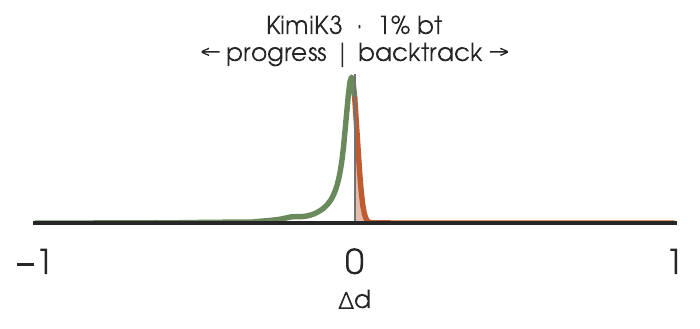} \\
\includegraphics[width=0.31\linewidth]{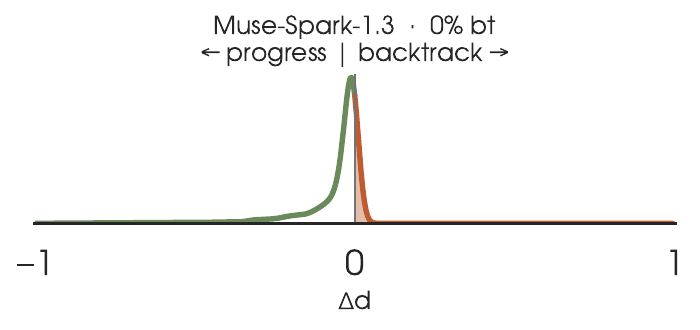} &
\includegraphics[width=0.31\linewidth]{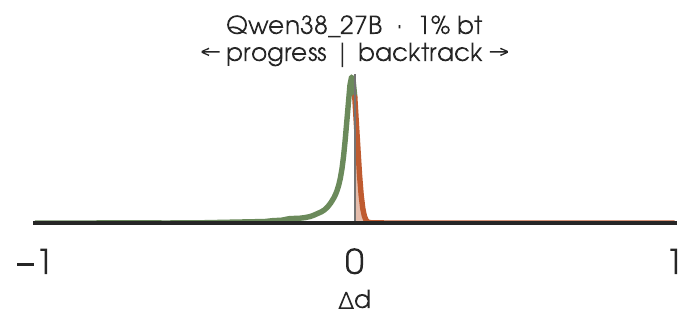} &
 \\
\end{tabular}
\caption{\textbf{Local backtracking.} Histograms of $\Delta D$ on Deep-SWE (part 2 of 2): Gemini, Grok, Kimi, Muse Spark and Qwen families. Continued from Figure~\ref{fig:metrics-deep-backtracking}.}
\label{fig:metrics-deep-backtracking-b}
\end{figure}

\begin{figure}[p]
\centering
\includegraphics[width=0.75\linewidth]{swe_verif/edit_read_ratio_legend.pdf}\\[2pt]
\setlength{\tabcolsep}{1pt}
\renewcommand{\arraystretch}{0.5}
\begin{tabular}{ccc}
\includegraphics[width=0.31\linewidth]{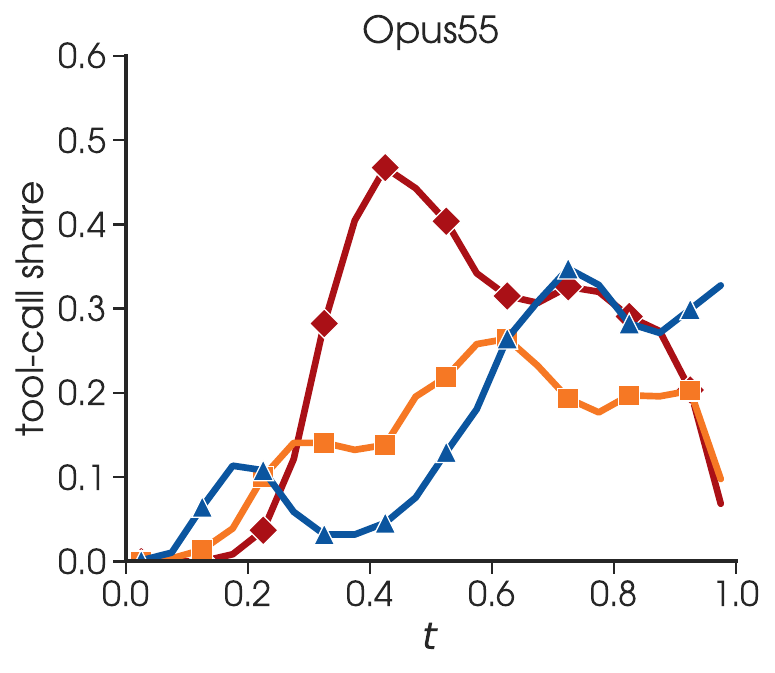} &
\includegraphics[width=0.31\linewidth]{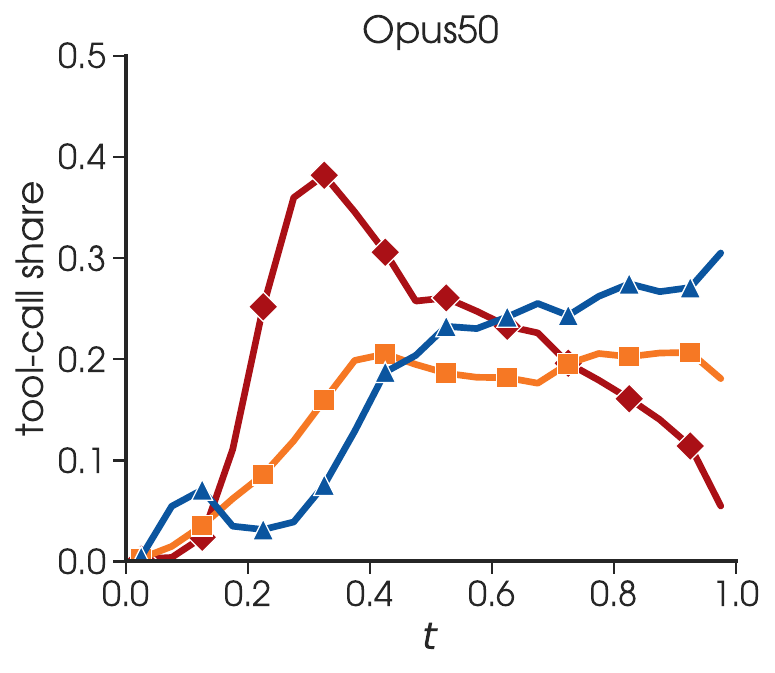} &
\includegraphics[width=0.31\linewidth]{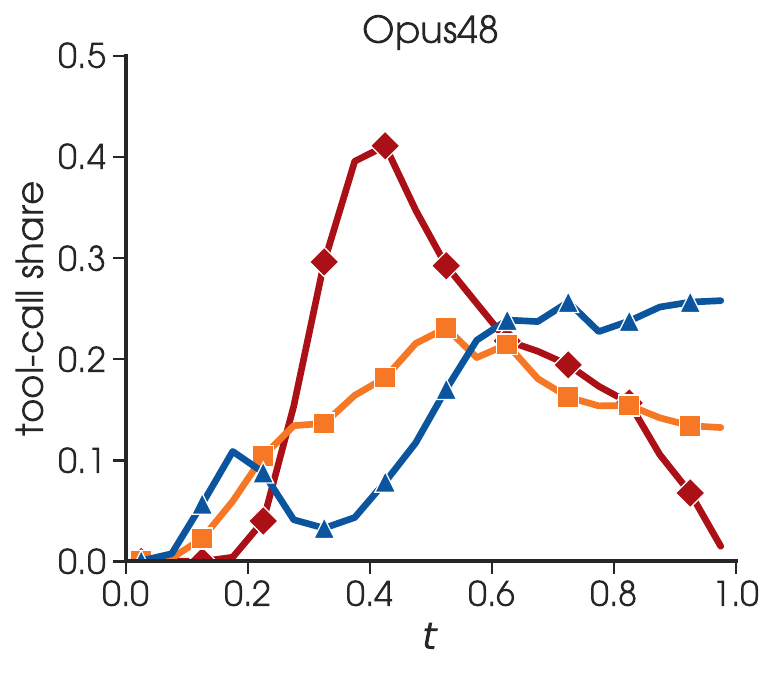} \\
\includegraphics[width=0.31\linewidth]{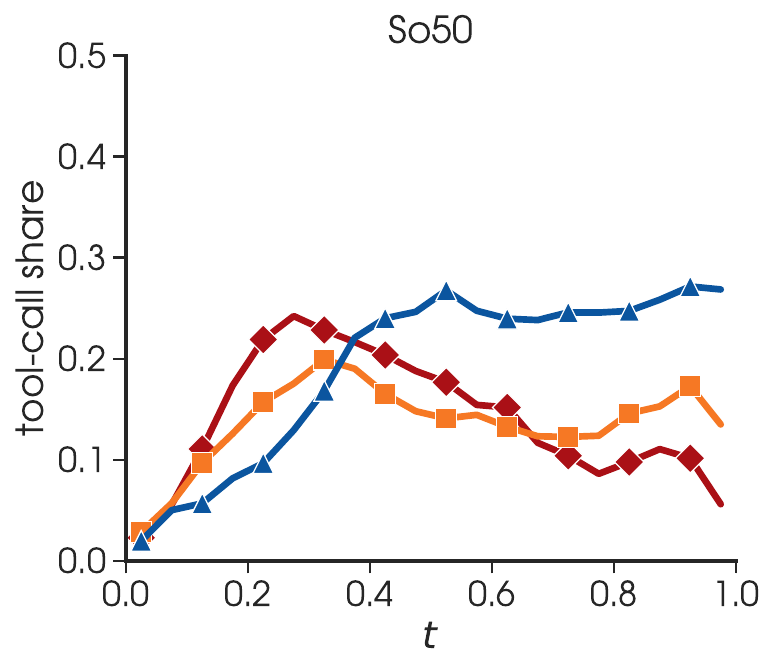} &
\includegraphics[width=0.31\linewidth]{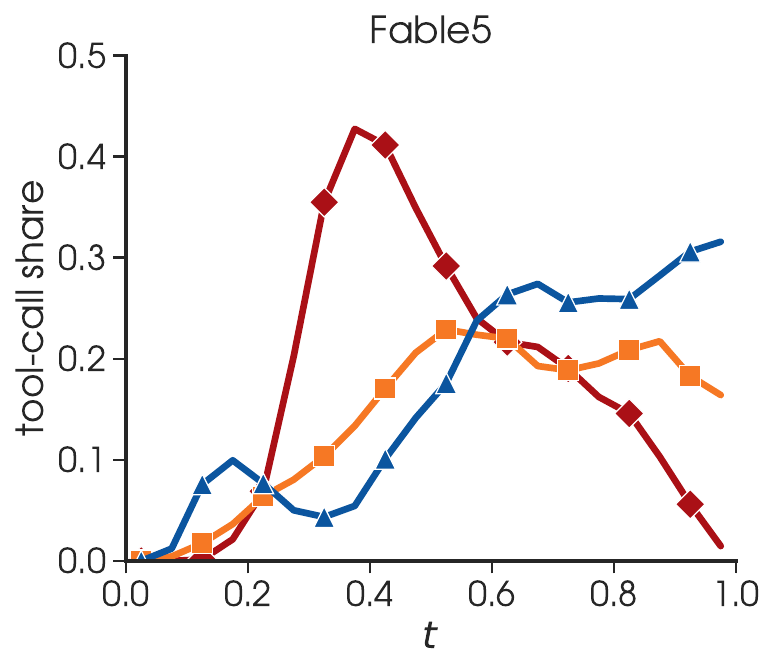} &
\includegraphics[width=0.31\linewidth]{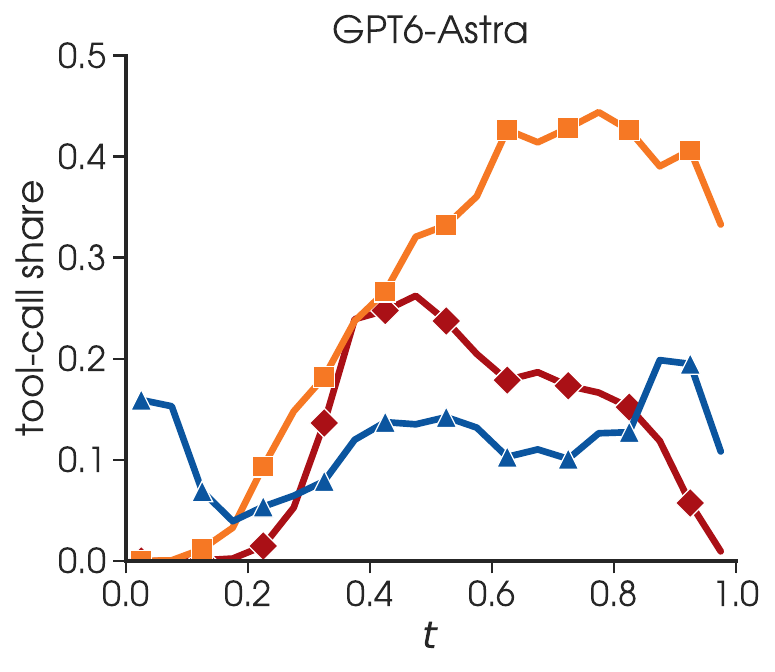} \\
\includegraphics[width=0.31\linewidth]{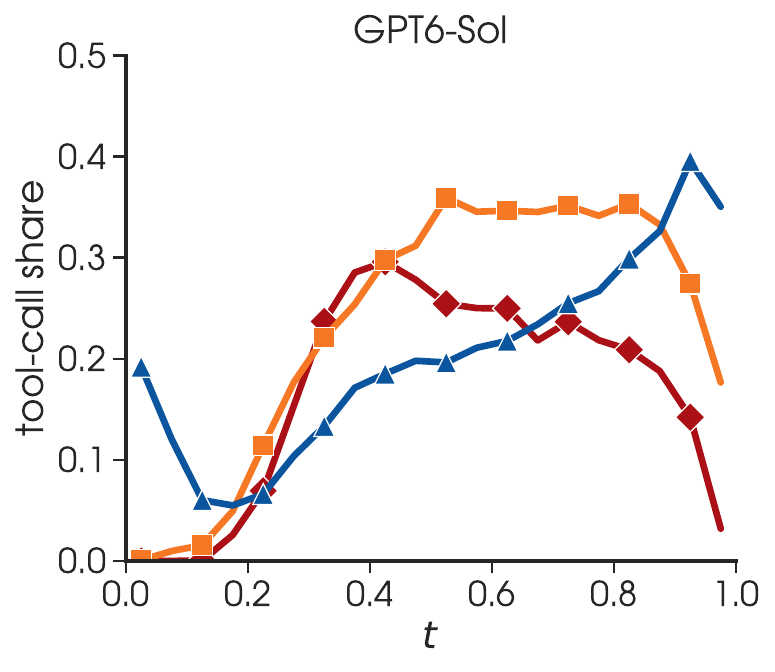} &
\includegraphics[width=0.31\linewidth]{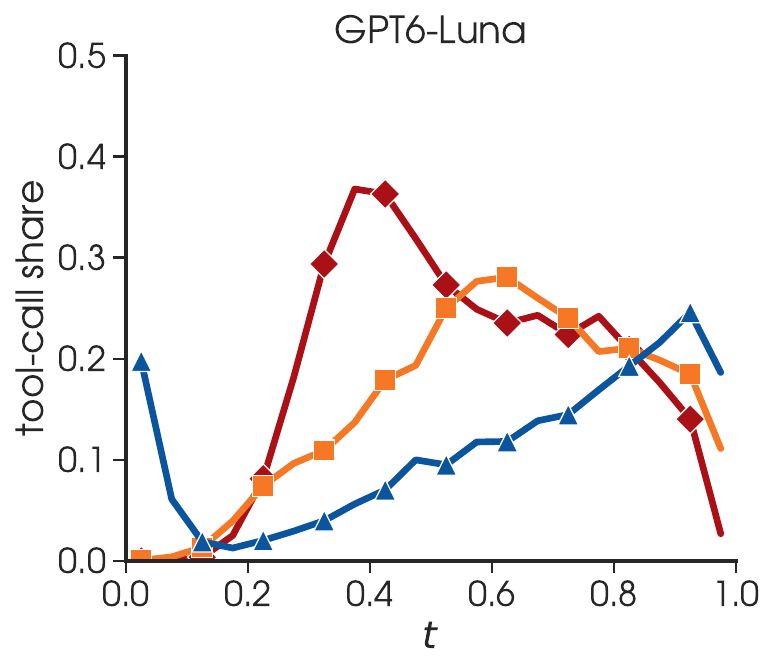} &
\includegraphics[width=0.31\linewidth]{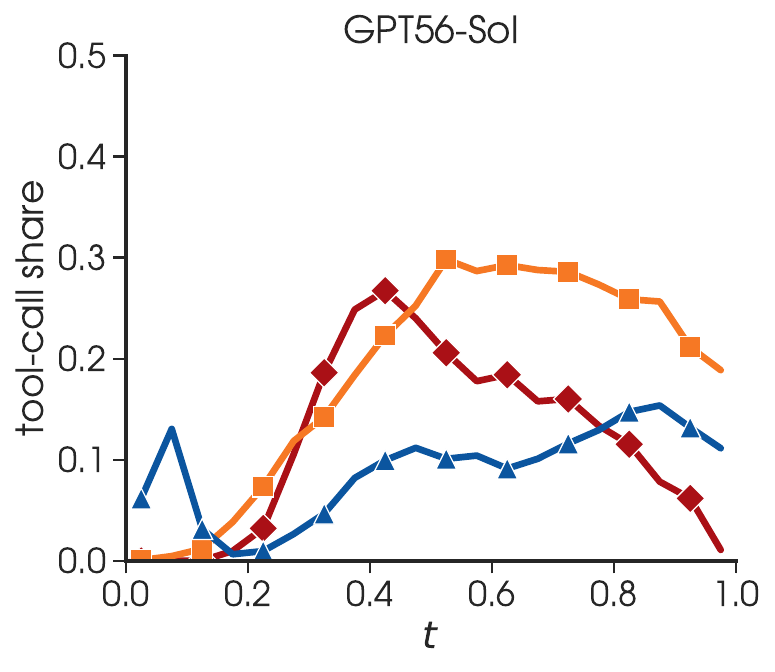} \\
\includegraphics[width=0.31\linewidth]{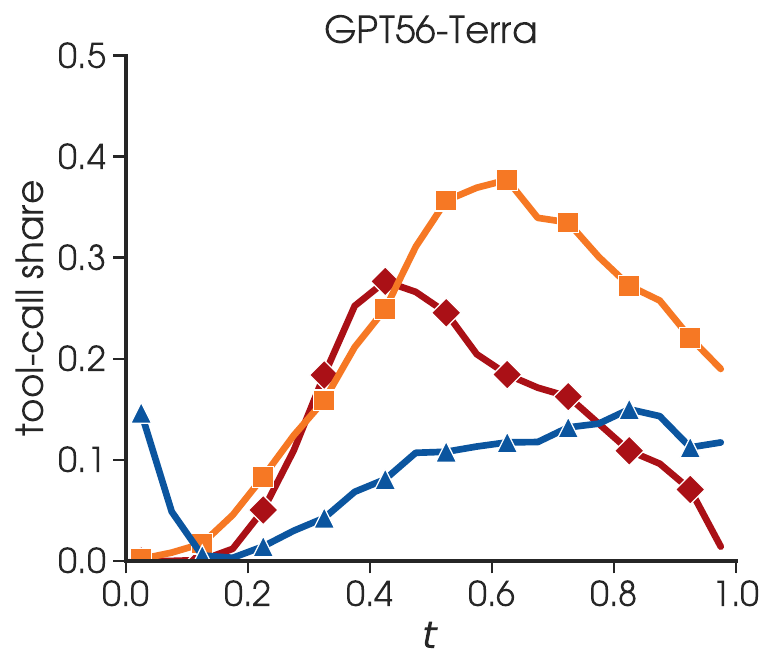} &
\includegraphics[width=0.31\linewidth]{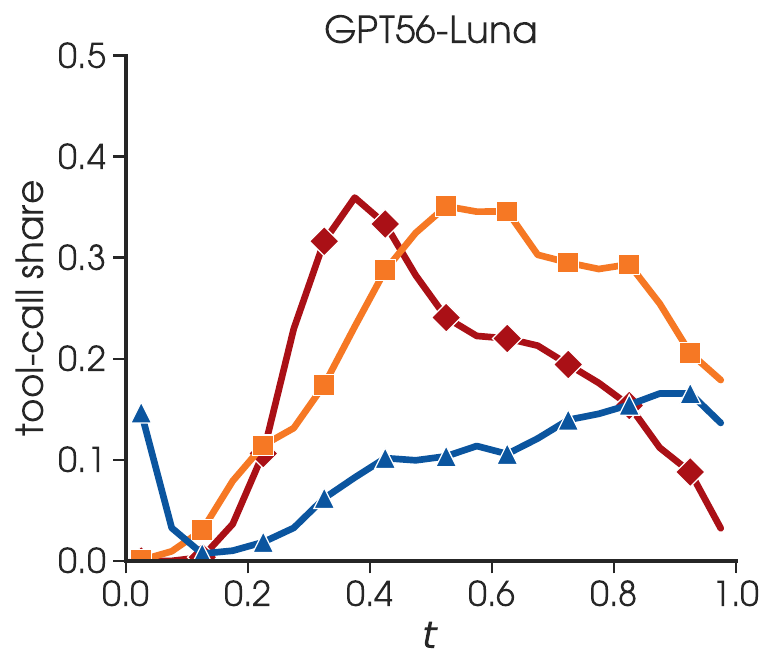} &
\includegraphics[width=0.31\linewidth]{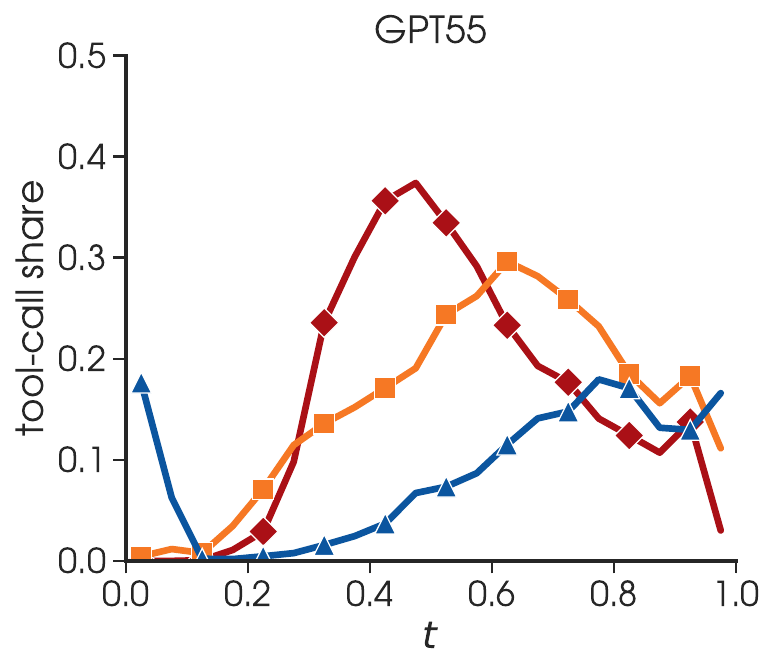} \\
\end{tabular}
\caption{Edit/test ratio per cycle on Deep-SWE (part 1 of 2): Claude and GPT families. Each cell shows the model's share of source edits, scratch tests and suite tests over normalized cycle position.}
\label{fig:metrics-deep-editread}
\end{figure}

\begin{figure}[p]
\centering
\includegraphics[width=0.75\linewidth]{swe_verif/edit_read_ratio_legend.pdf}\\[2pt]
\setlength{\tabcolsep}{1pt}
\renewcommand{\arraystretch}{0.5}
\begin{tabular}{ccc}
\includegraphics[width=0.31\linewidth]{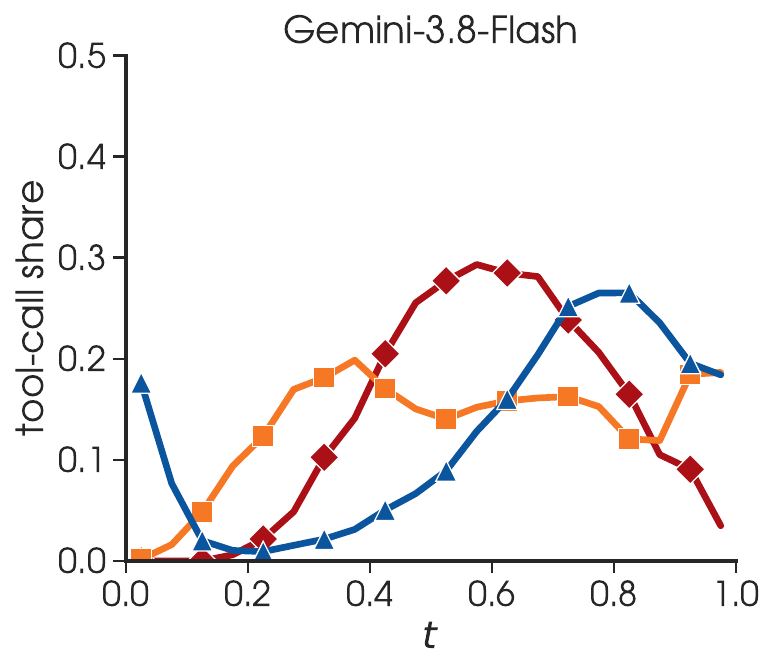} &
\includegraphics[width=0.31\linewidth]{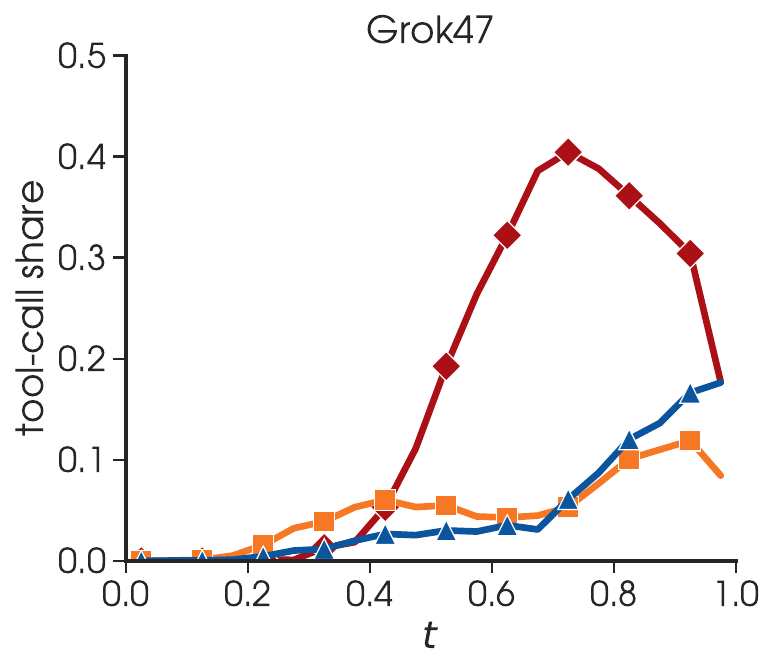} &
\includegraphics[width=0.31\linewidth]{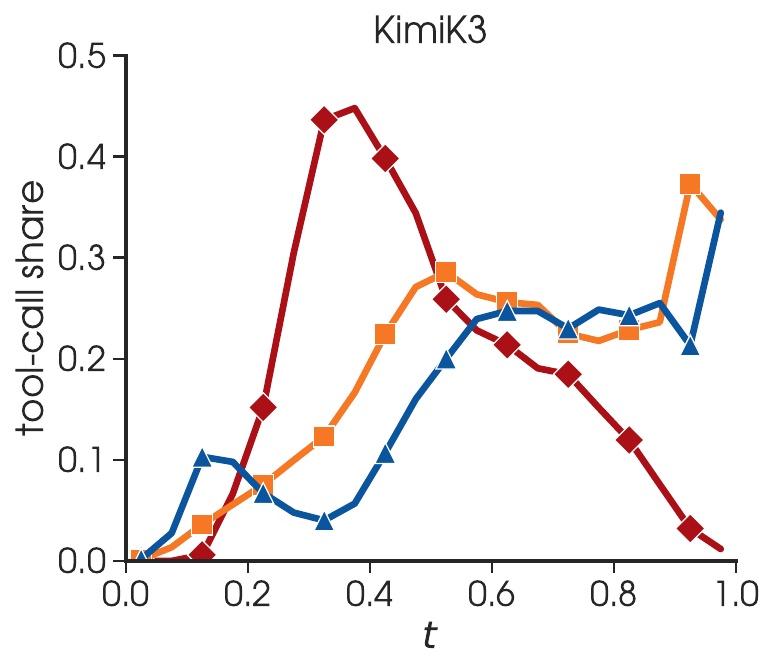} \\
\includegraphics[width=0.31\linewidth]{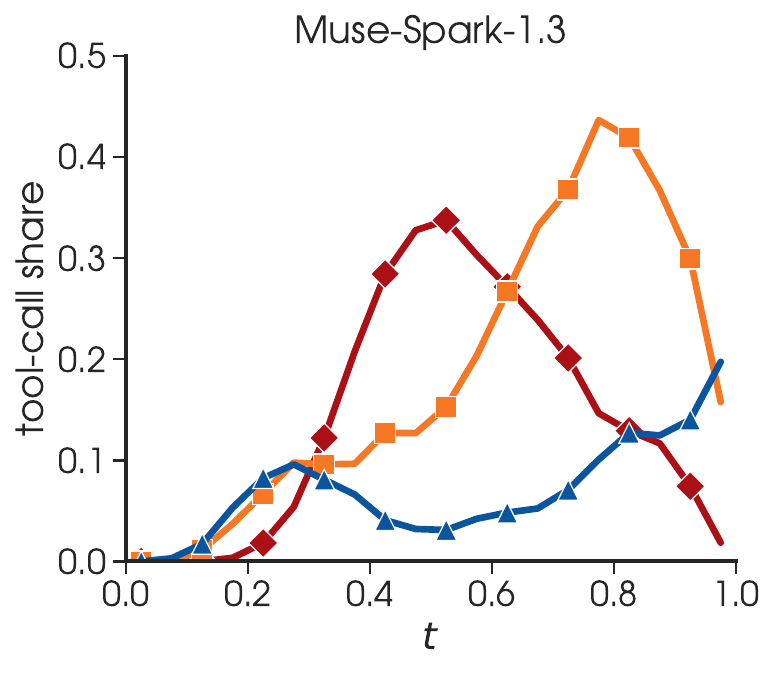} &
\includegraphics[width=0.31\linewidth]{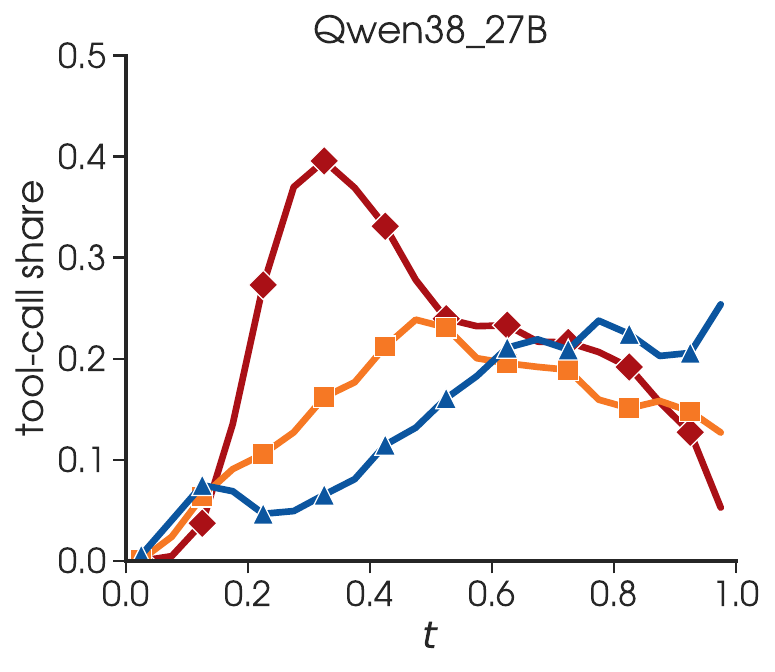} &
 \\
\end{tabular}
\caption{Edit/test ratio per cycle on Deep-SWE (part 2 of 2): Gemini, Grok, Kimi, Muse Spark and Qwen families. Continued from Figure~\ref{fig:metrics-deep-editread}.}
\label{fig:metrics-deep-editread-b}
\end{figure}

\begin{figure}[p]
\centering
\setlength{\tabcolsep}{1pt}
\renewcommand{\arraystretch}{0.5}
\begin{tabular}{ccc}
\includegraphics[width=0.31\linewidth]{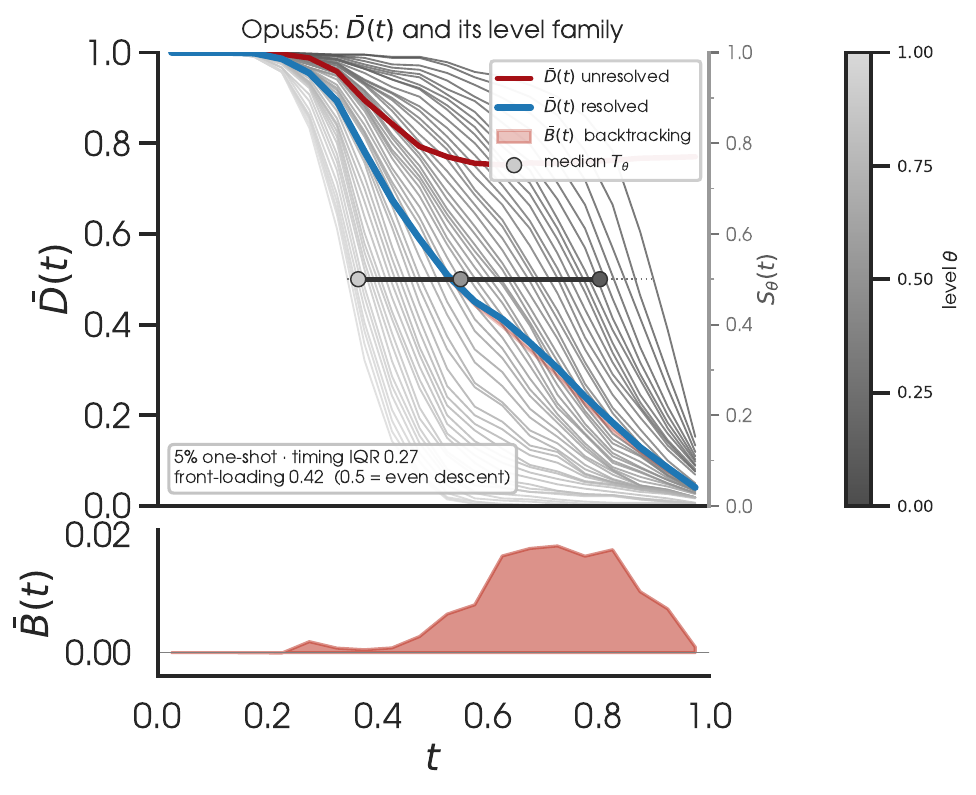} &
\includegraphics[width=0.31\linewidth]{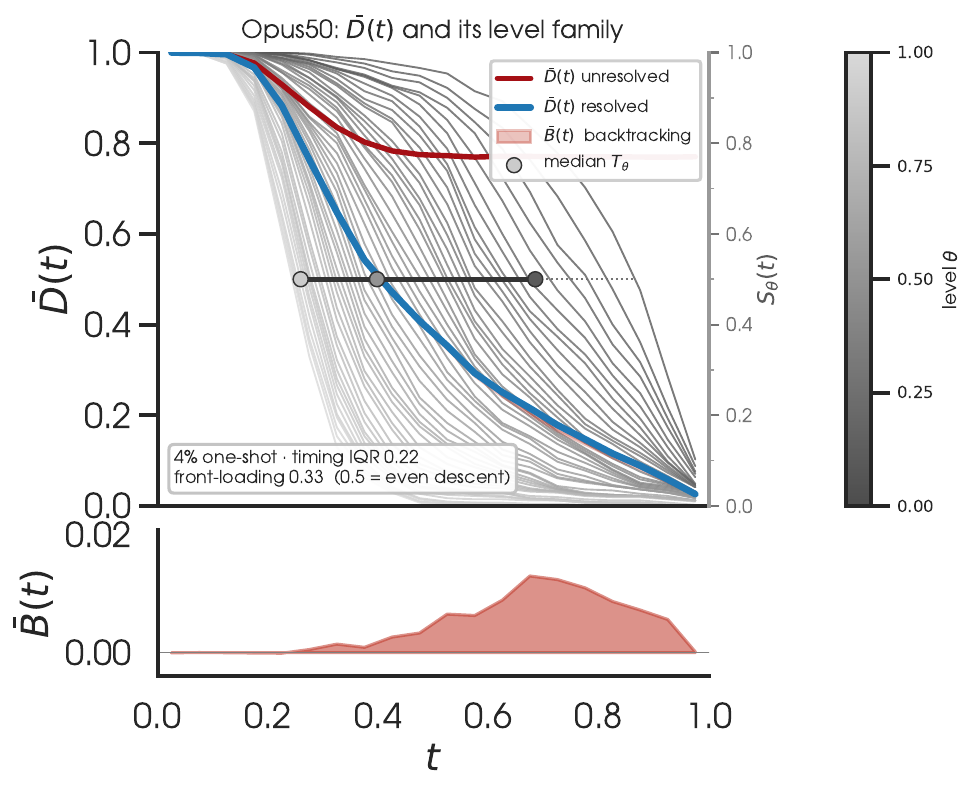} &
\includegraphics[width=0.31\linewidth]{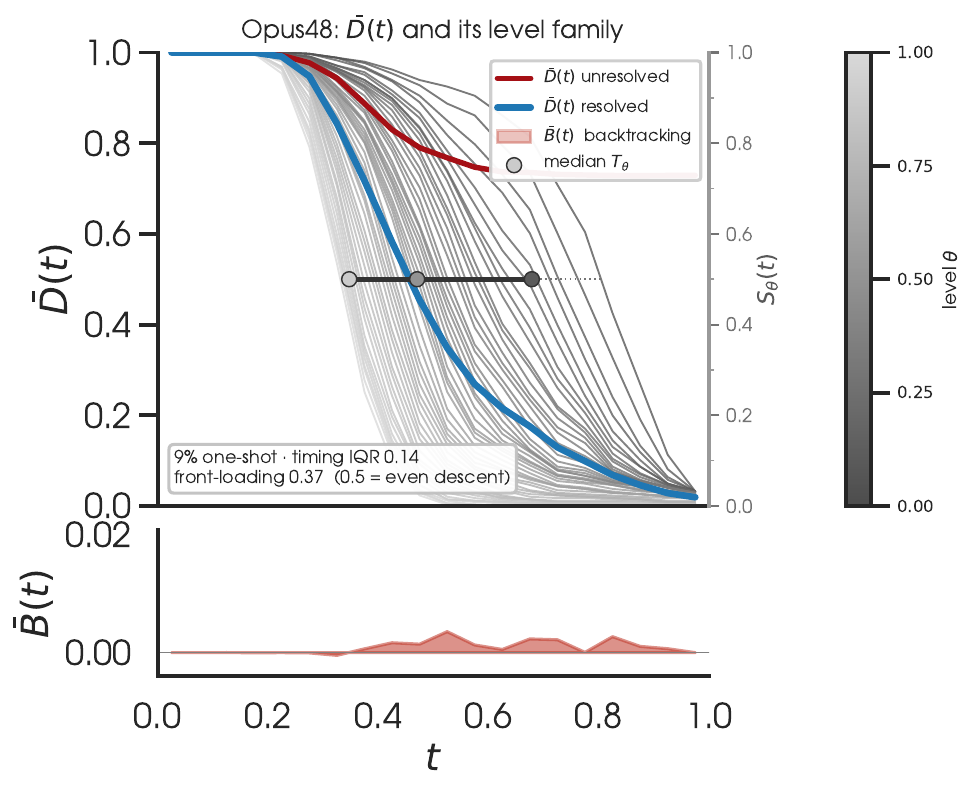} \\
\includegraphics[width=0.31\linewidth]{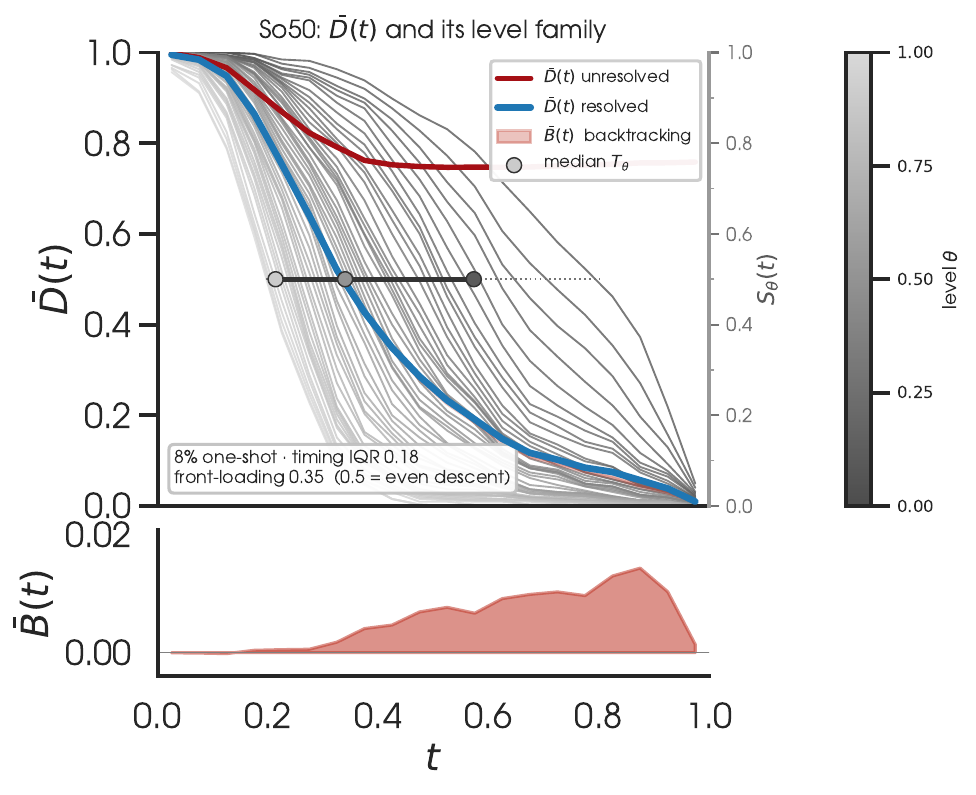} &
\includegraphics[width=0.31\linewidth]{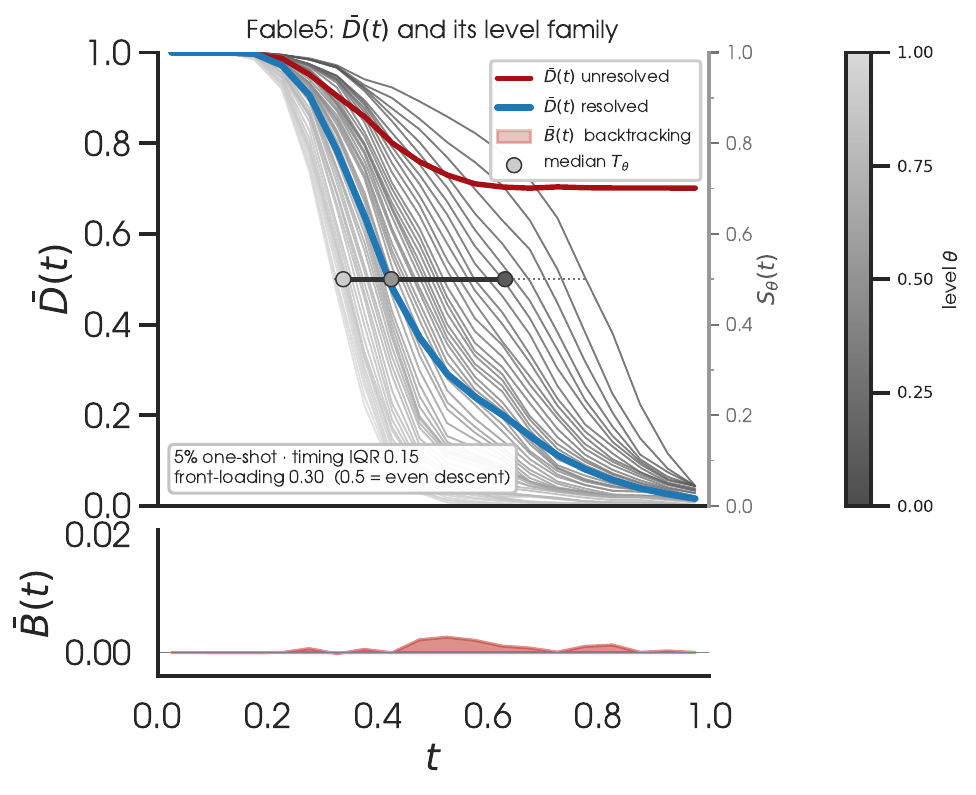} &
\includegraphics[width=0.31\linewidth]{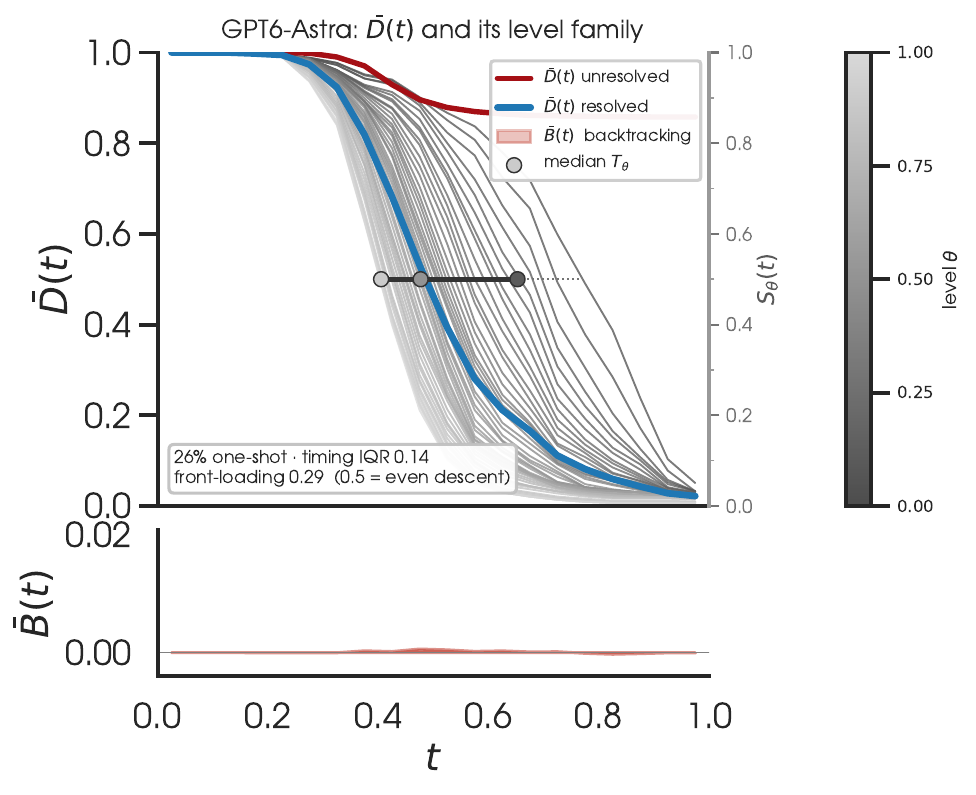} \\
\includegraphics[width=0.31\linewidth]{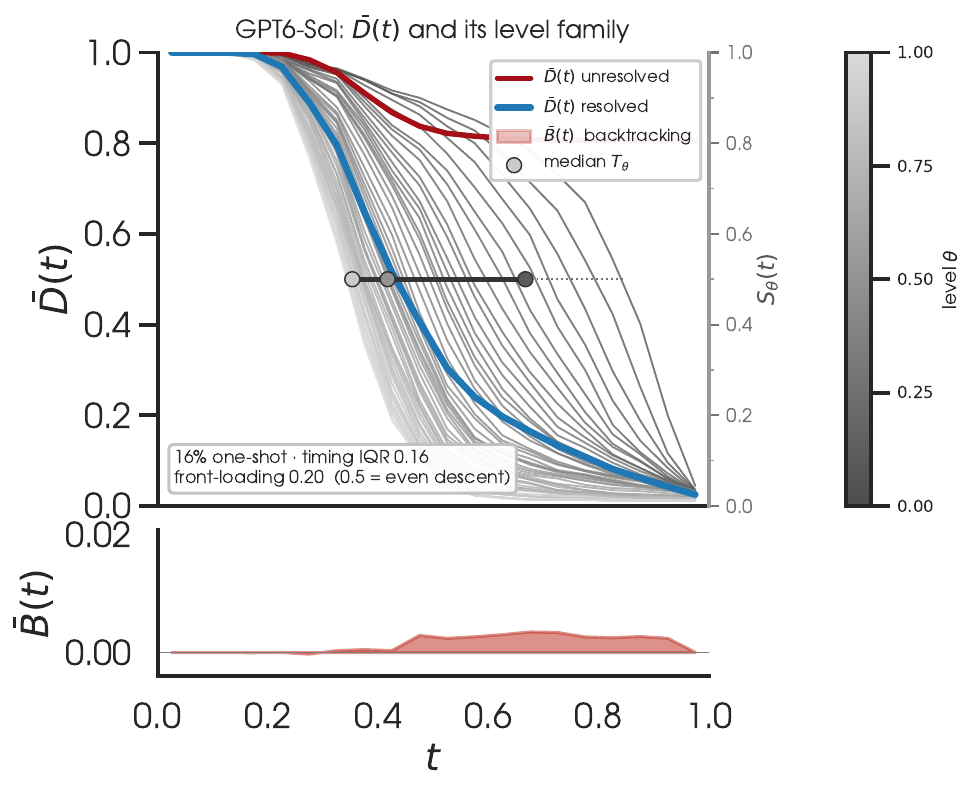} &
\includegraphics[width=0.31\linewidth]{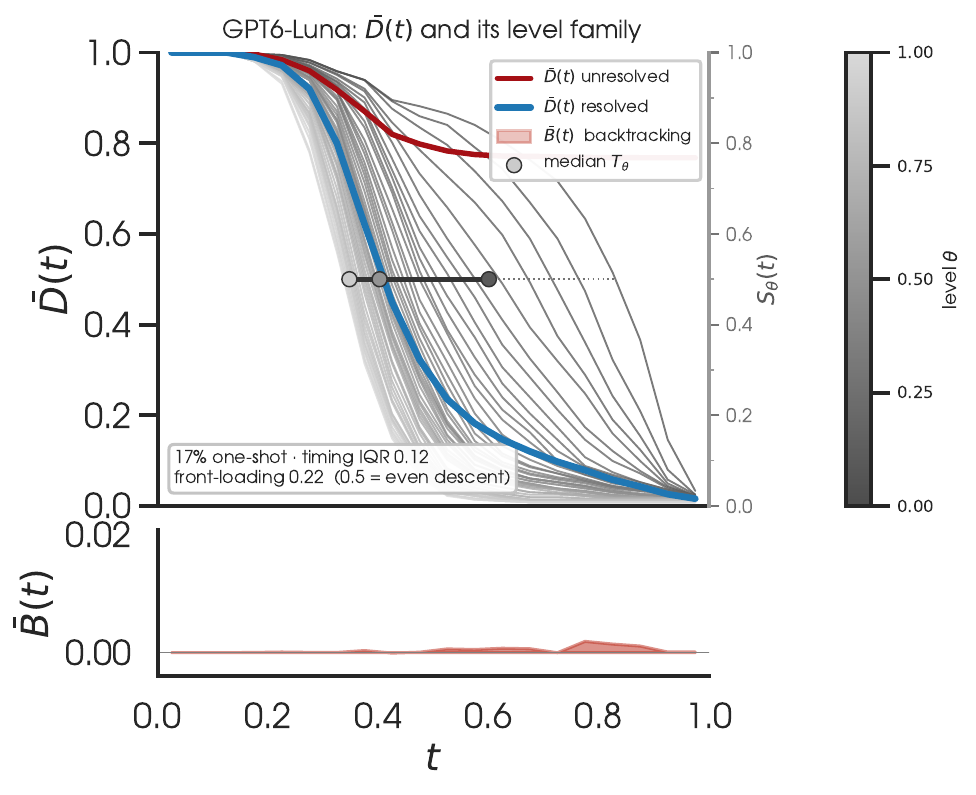} &
\includegraphics[width=0.31\linewidth]{swe_deep/solution_distance_fan/gpt56_sol.pdf} \\
\includegraphics[width=0.31\linewidth]{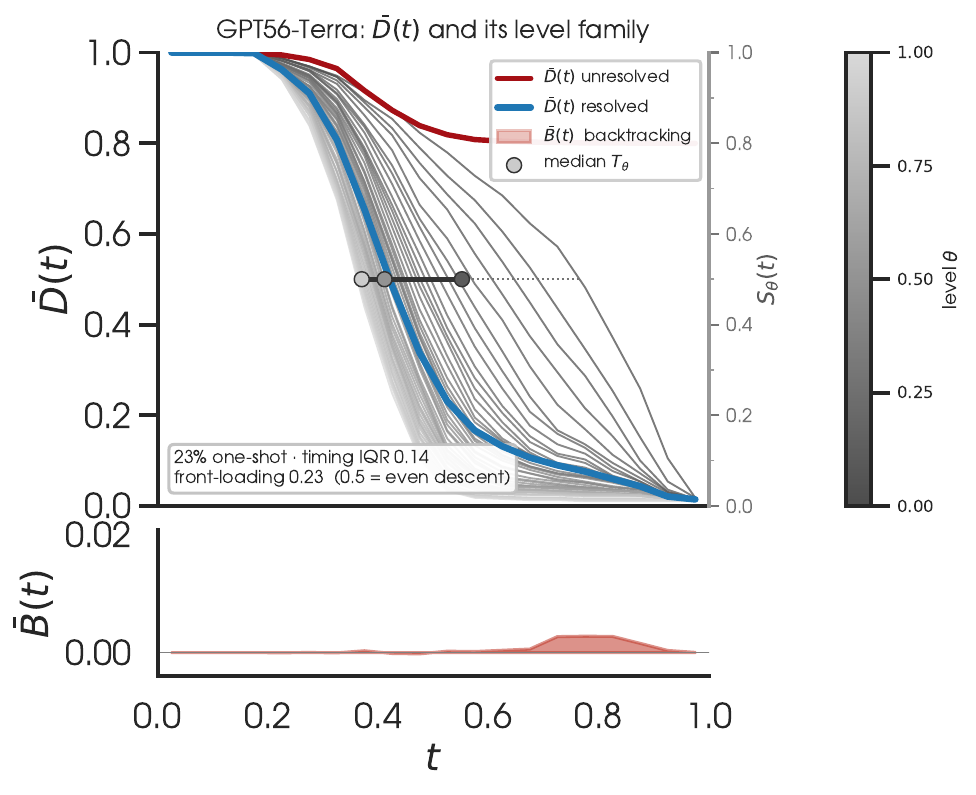} &
\includegraphics[width=0.31\linewidth]{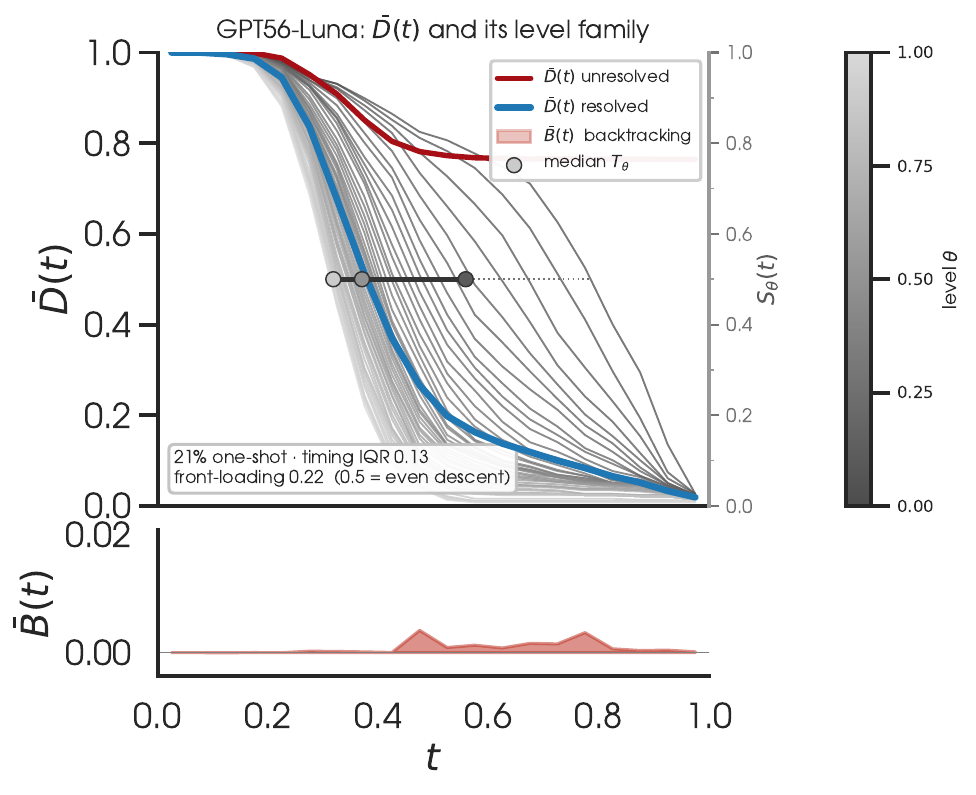} &
\includegraphics[width=0.31\linewidth]{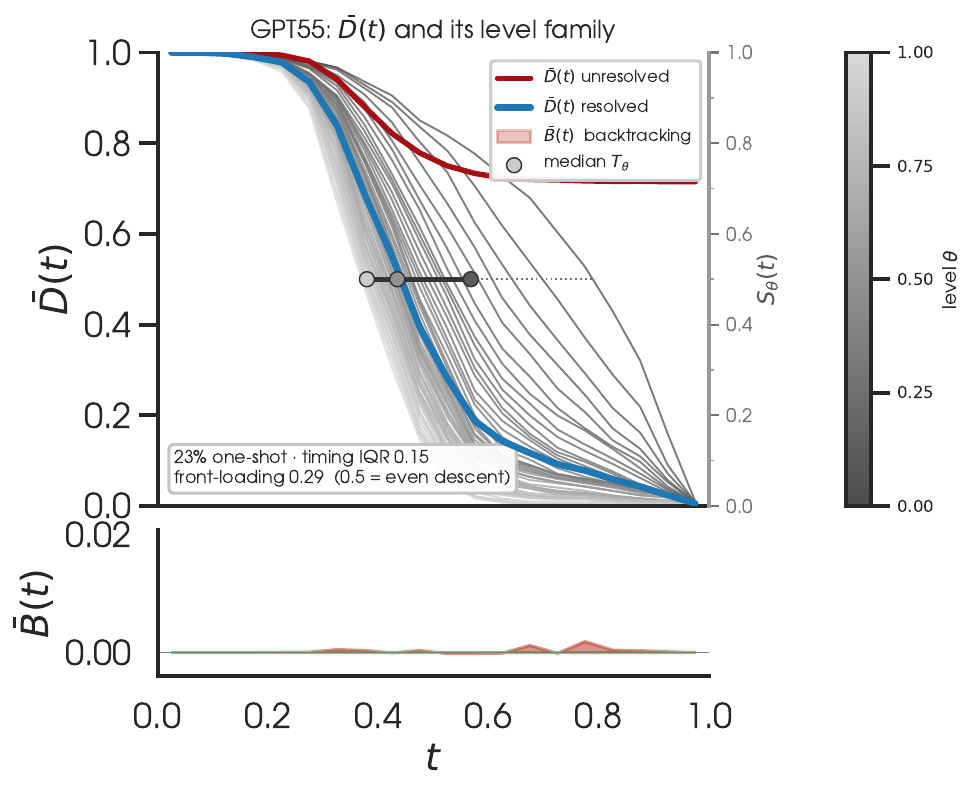} \\
\end{tabular}
\caption{Level families on Deep-SWE (part 1 of 2): Claude and GPT families. Grey curves are the level sets $S_\theta(t)$ of the resolved trajectories (Section~\ref{sec:distance}), the blue curve is their mean $\bar D(t)$, the dark-red curve is the mean $\bar D(t)$ of the unresolved runs (which plateaus where they stop), the markers give the median crossing time $T_\theta$ (the median over runs of the first-entry time $T_r(\theta)$, i.e.\ where $S_\theta$ crosses $1/2$) at $\theta=0.5$ with its interquartile range, and the strip beneath each panel is the backtracking term $\bar B(t)$. The inset reports the one-shot share (runs that land the patch in a single edit); the timing IQR, the interquartile range of the runs' half-way times $T_r(0.5)$, i.e.\ how dispersed solve timing is across runs; and the front-loading index $(T_{0.5}-T_{0.9})/(T_{0.1}-T_{0.9})$, the position of the median half-way time between the median crossing times of levels $0.9$ and $0.1$ ($0.5$ = even descent; lower = most of the distance falls early).}
\label{fig:metrics-deep-soldist}
\end{figure}

\begin{figure}[p]
\centering
\setlength{\tabcolsep}{1pt}
\renewcommand{\arraystretch}{0.5}
\begin{tabular}{ccc}
\includegraphics[width=0.31\linewidth]{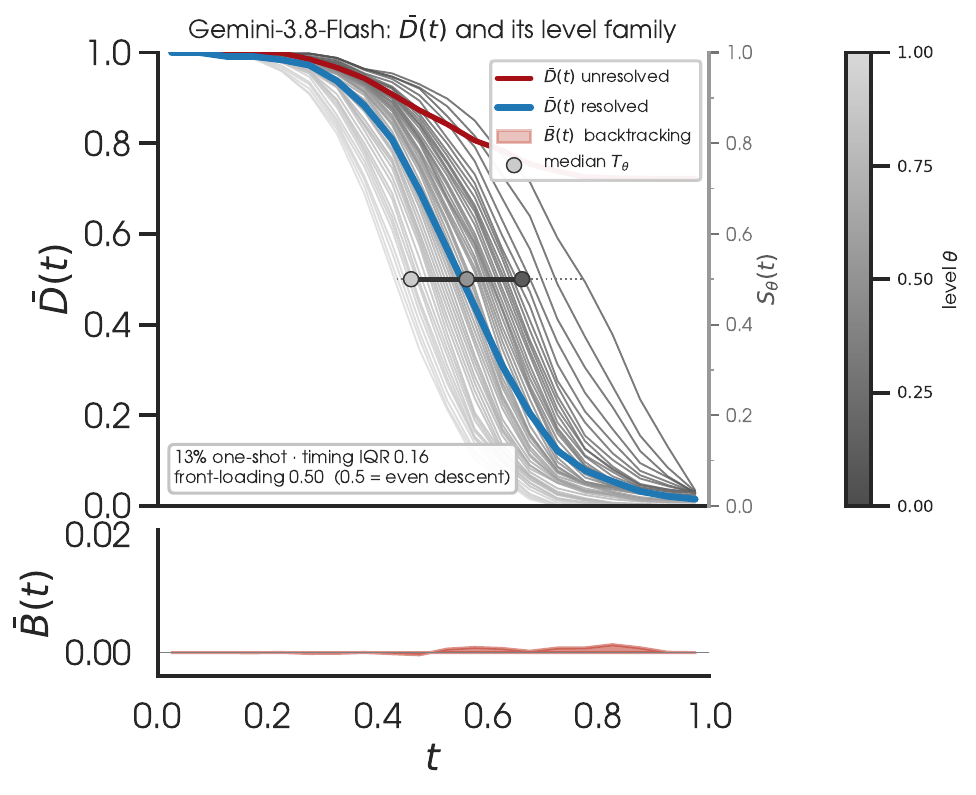} &
\includegraphics[width=0.31\linewidth]{swe_deep/solution_distance_fan/grok47.pdf} &
\includegraphics[width=0.31\linewidth]{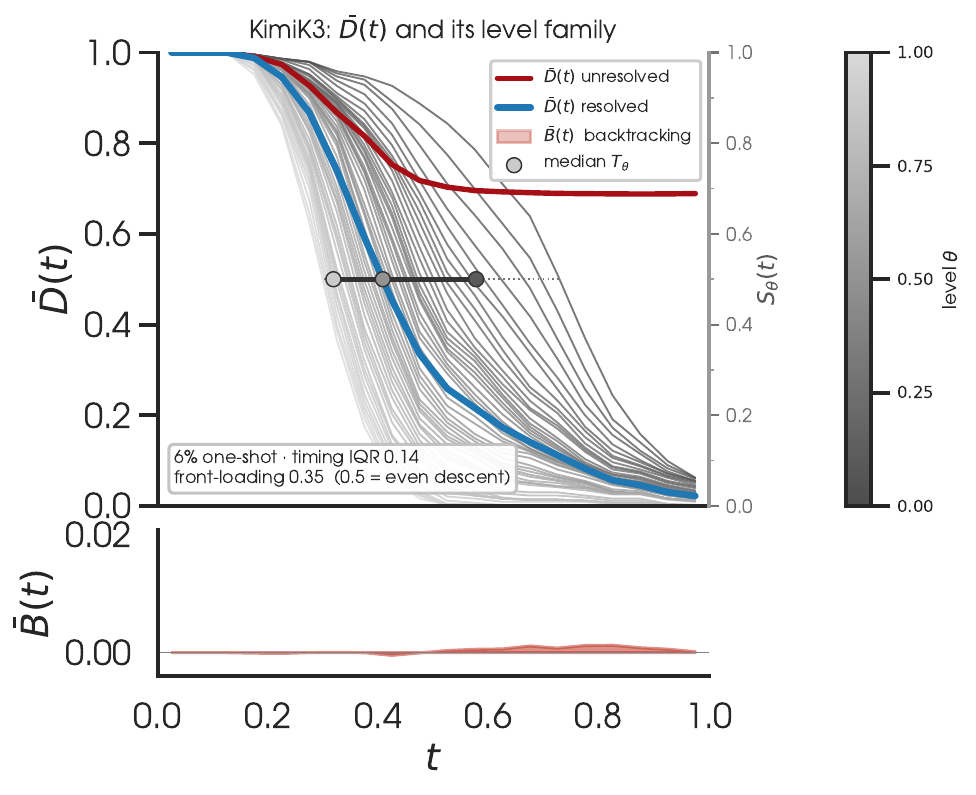} \\
\includegraphics[width=0.31\linewidth]{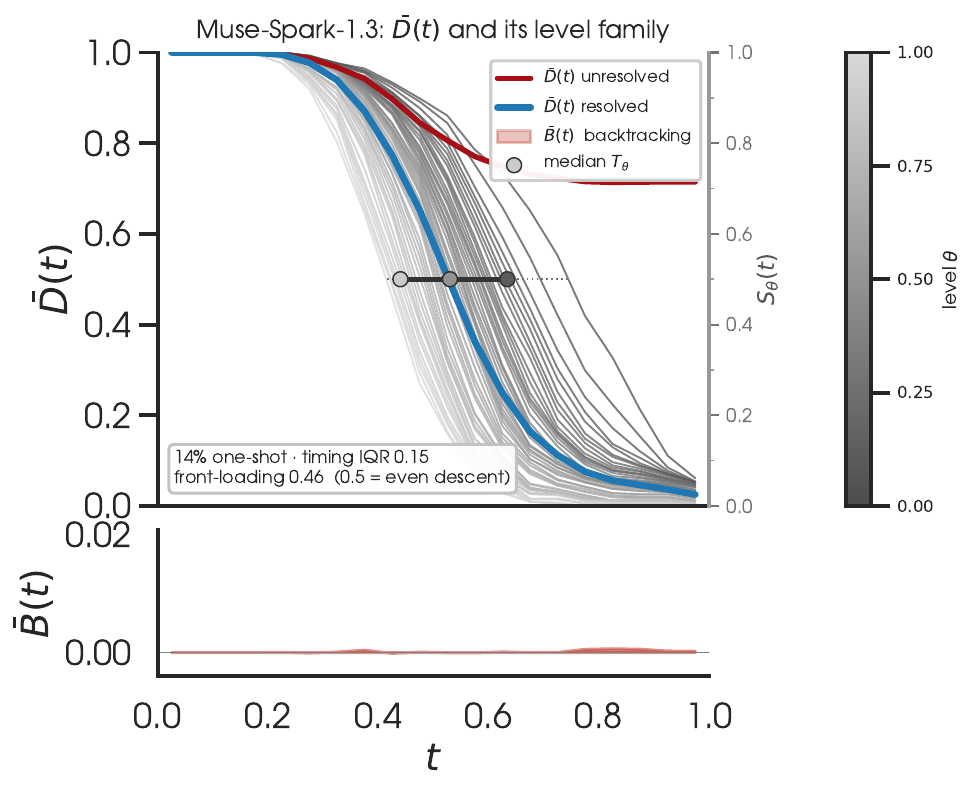} &
\includegraphics[width=0.31\linewidth]{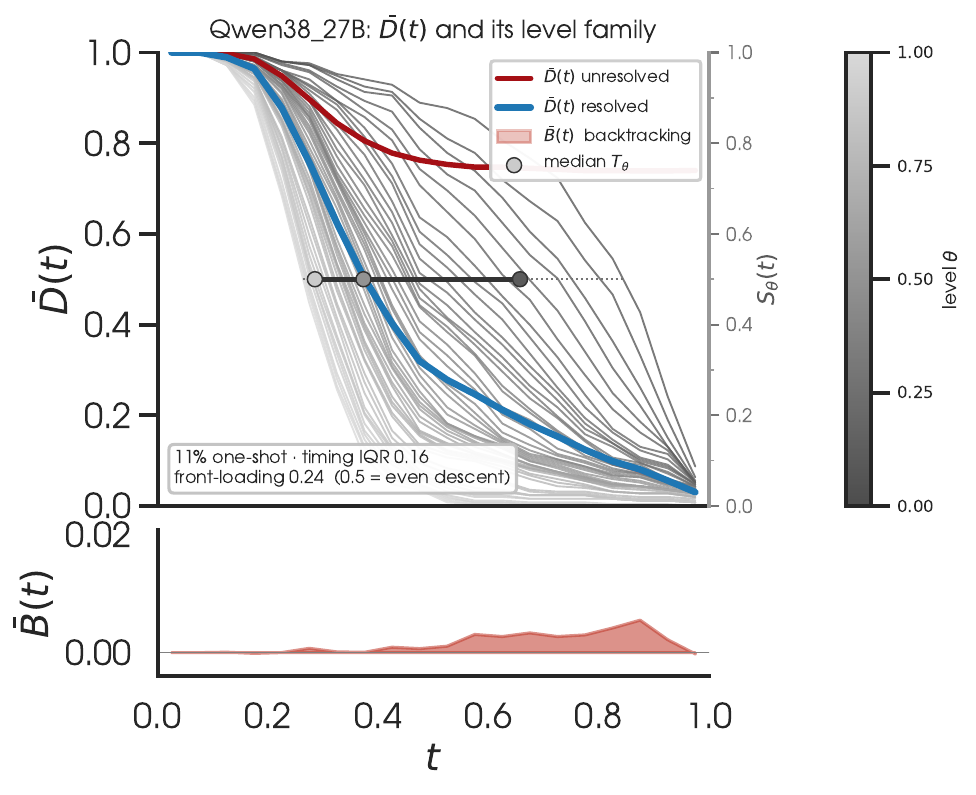} &
 \\
\end{tabular}
\caption{Level families on Deep-SWE (part 2 of 2): Gemini, Grok, Kimi, Muse Spark and Qwen families. Continued from Figure~\ref{fig:metrics-deep-soldist}.}
\label{fig:metrics-deep-soldist-b}
\end{figure}

\begin{figure}[p]
\centering
\setlength{\tabcolsep}{1pt}
\renewcommand{\arraystretch}{0.5}
\begin{tabular}{ccc}
\includegraphics[width=0.31\linewidth]{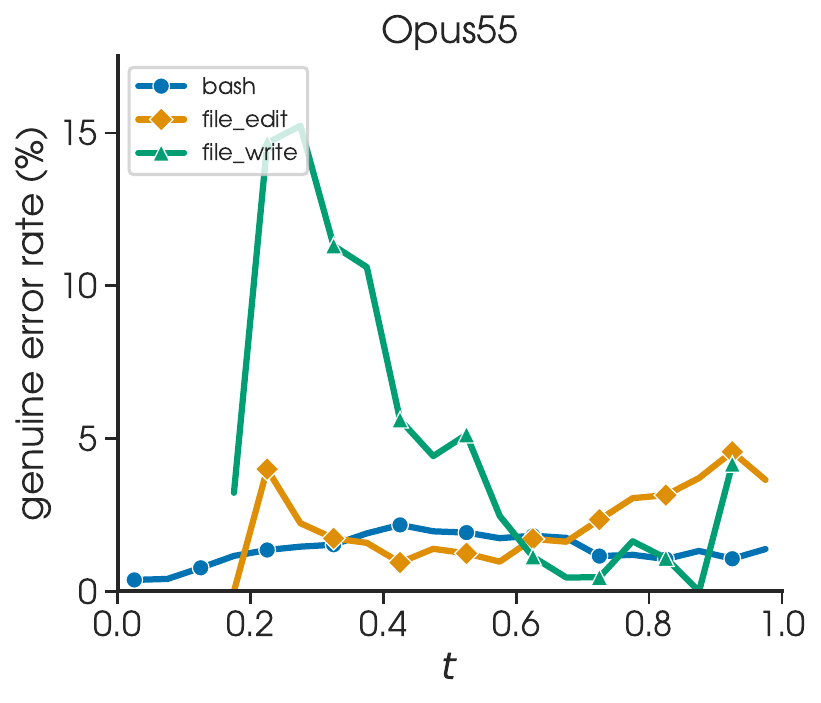} &
\includegraphics[width=0.31\linewidth]{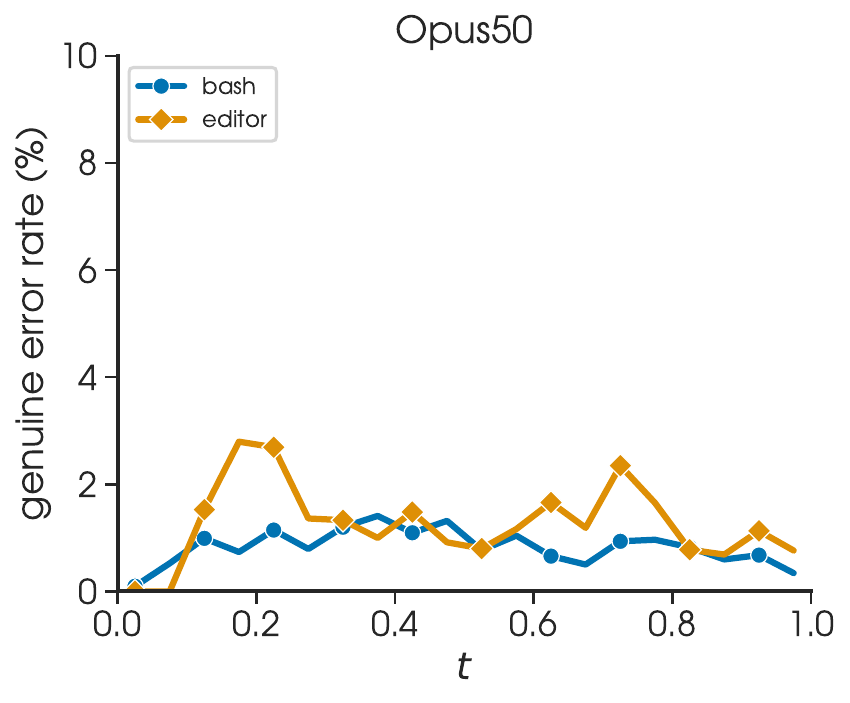} &
\includegraphics[width=0.31\linewidth]{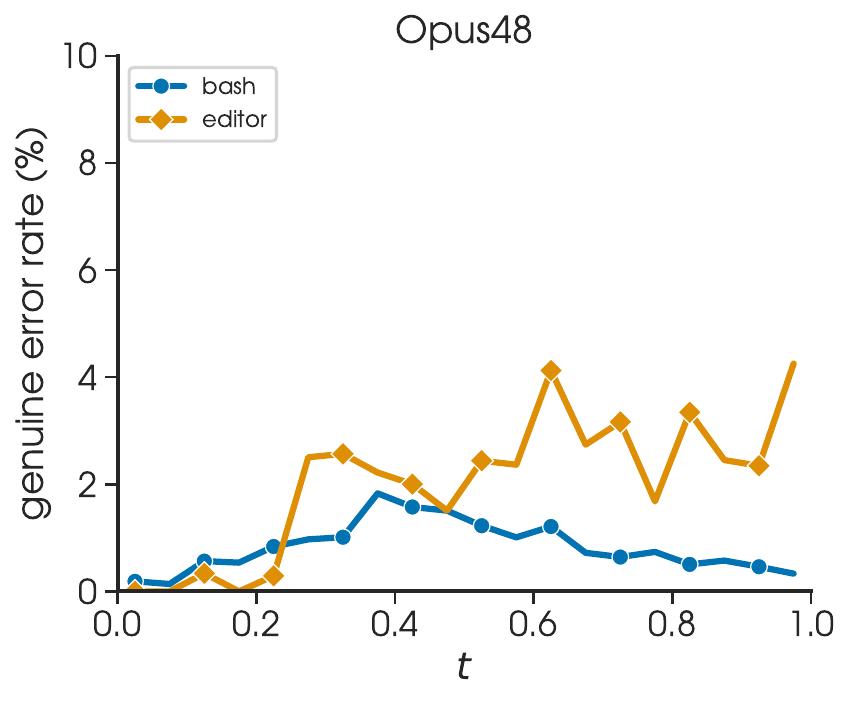} \\
\includegraphics[width=0.31\linewidth]{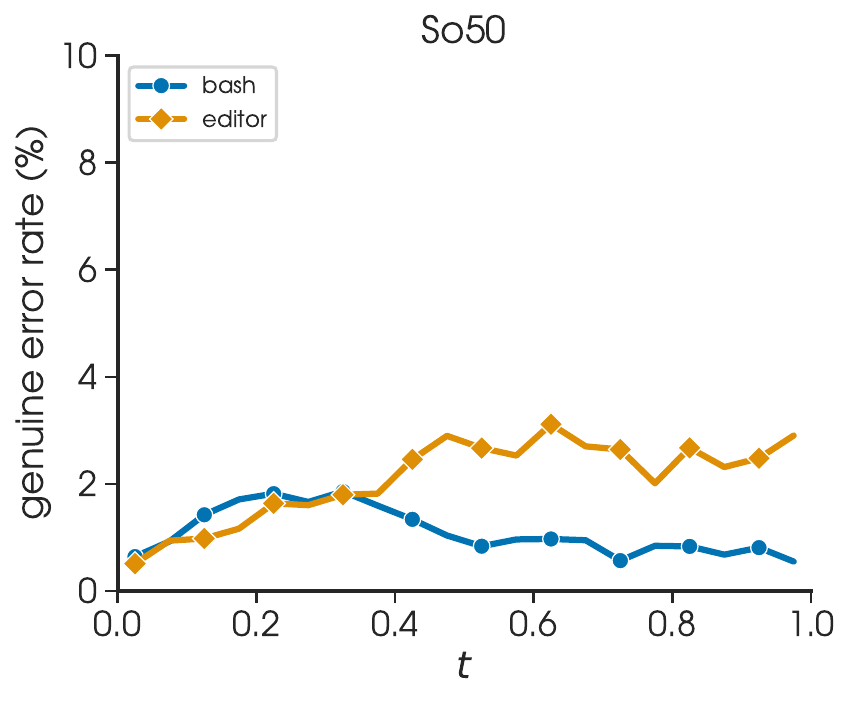} &
\includegraphics[width=0.31\linewidth]{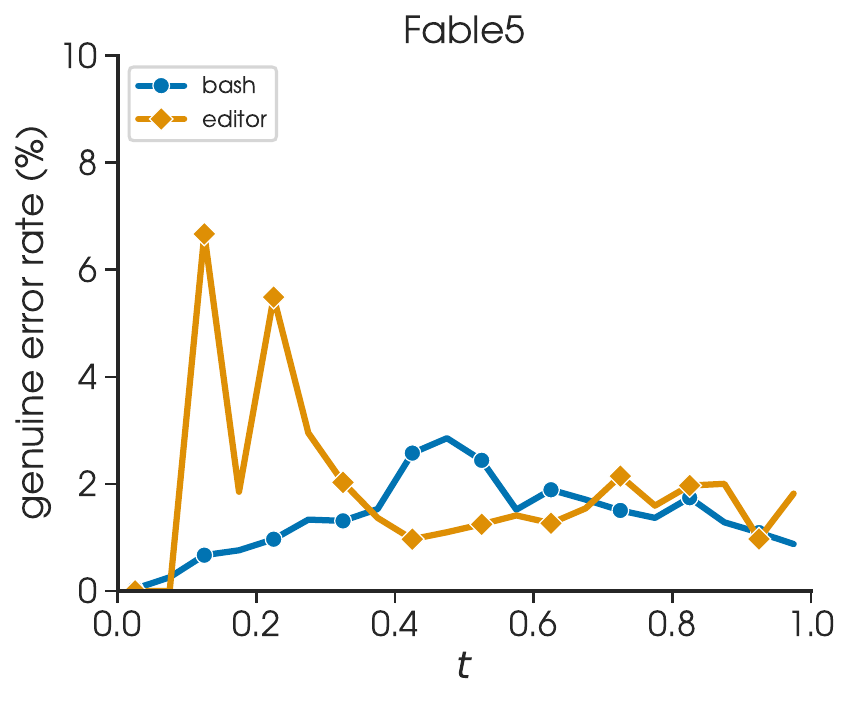} &
\includegraphics[width=0.31\linewidth]{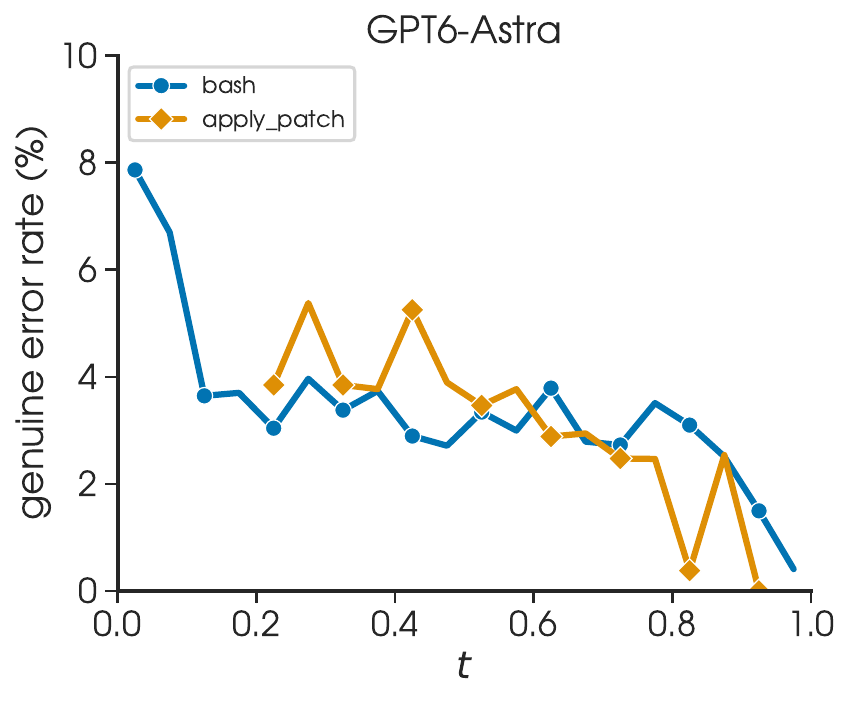} \\
\includegraphics[width=0.31\linewidth]{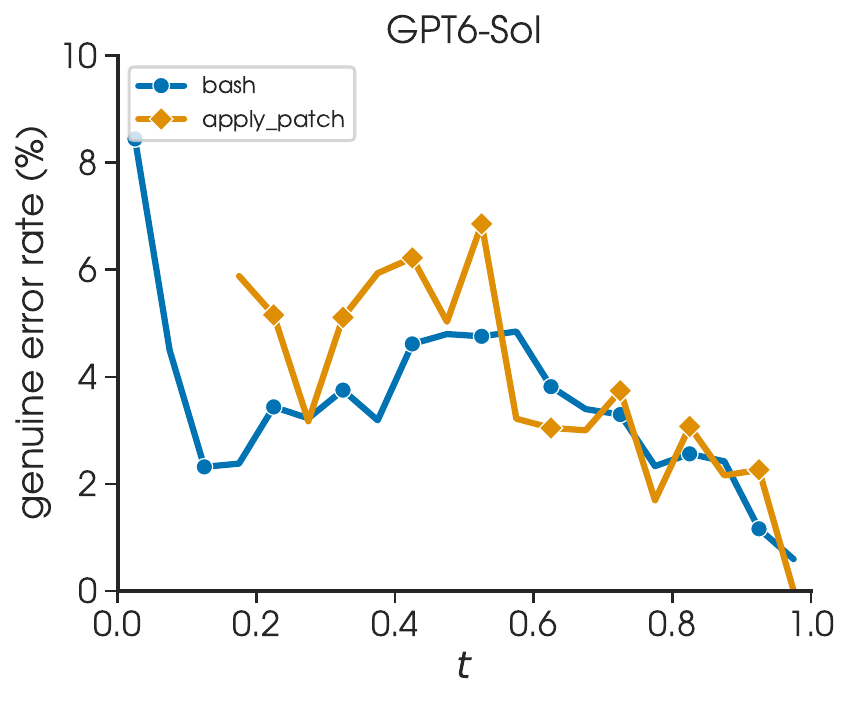} &
\includegraphics[width=0.31\linewidth]{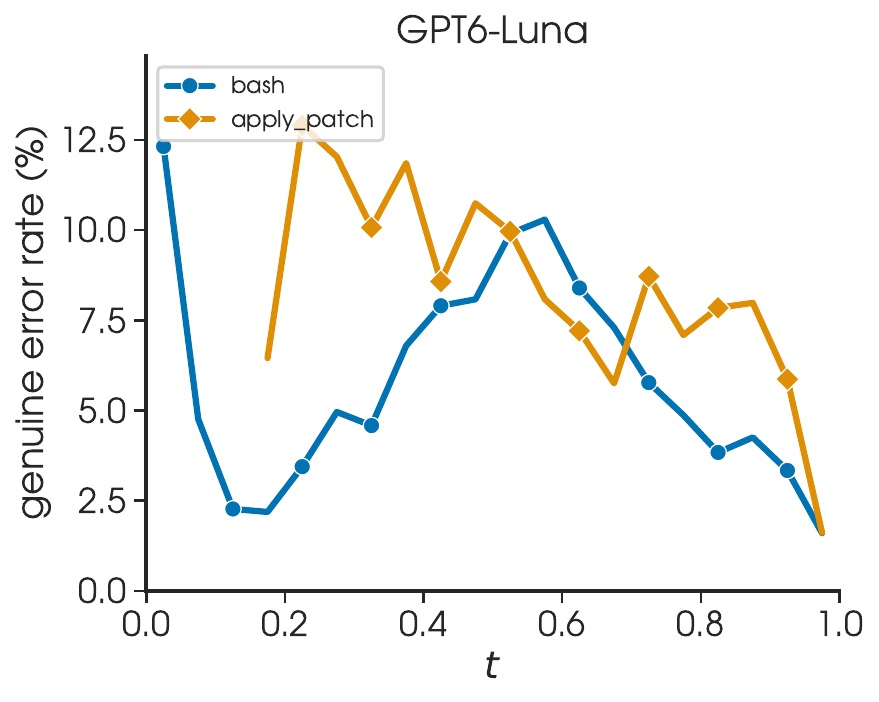} &
\includegraphics[width=0.31\linewidth]{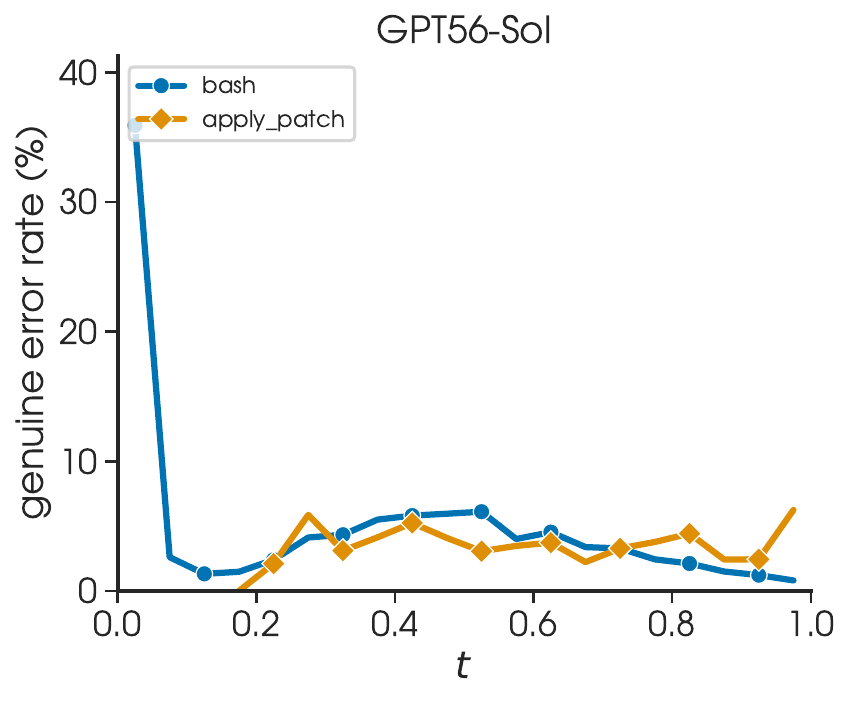} \\
\includegraphics[width=0.31\linewidth]{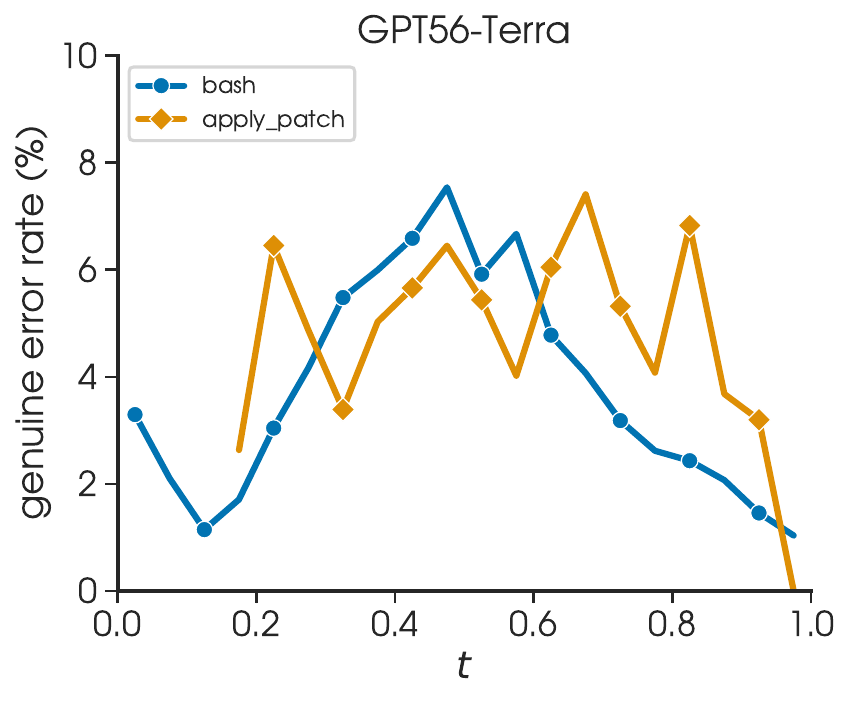} &
\includegraphics[width=0.31\linewidth]{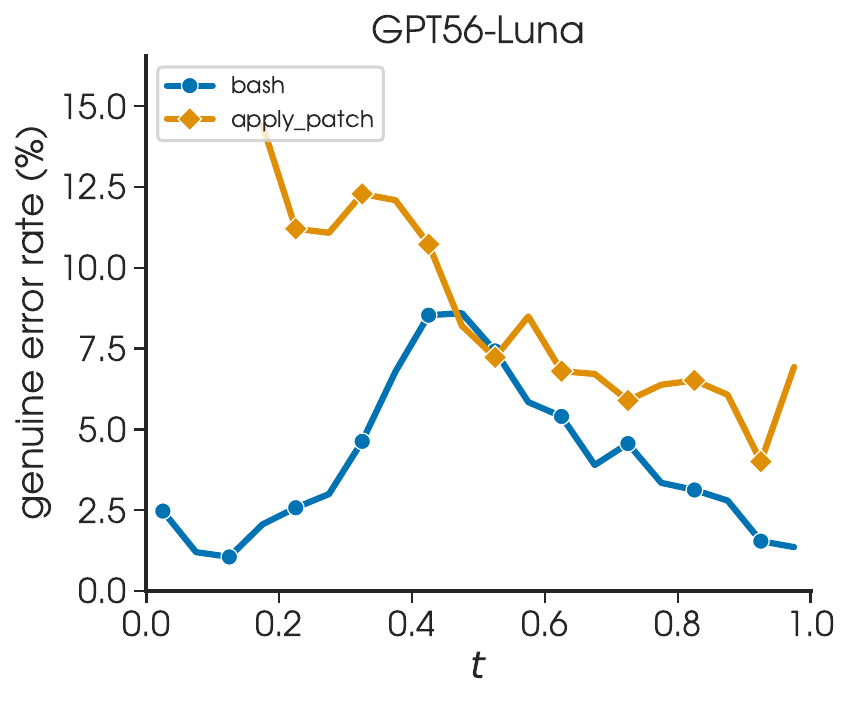} &
\includegraphics[width=0.31\linewidth]{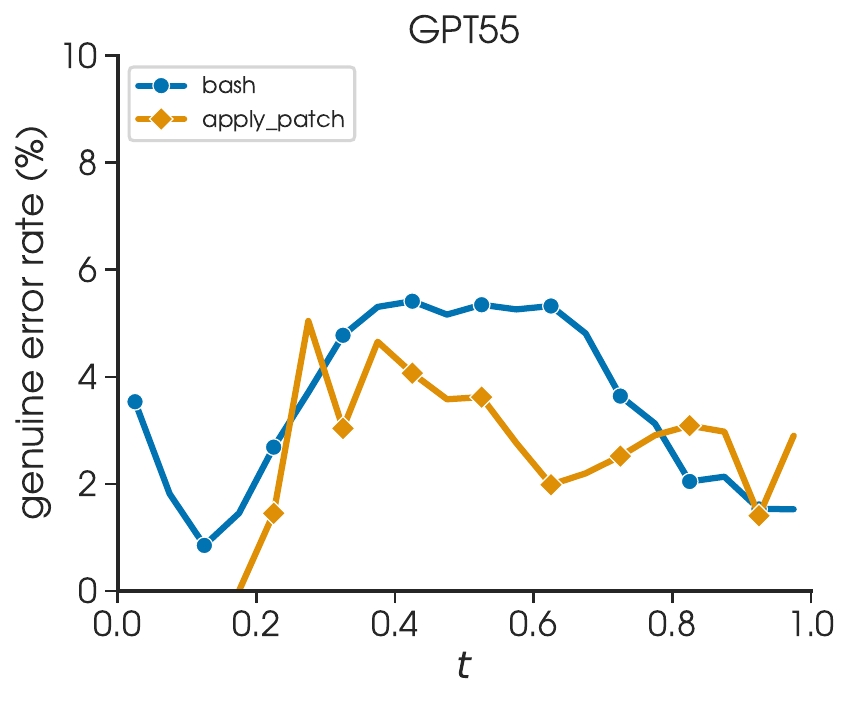} \\
\end{tabular}
\caption{Genuine tool-error rate per cycle on Deep-SWE (part 1 of 2): Claude and GPT families. Benign exits (\texttt{grep} no-match, \texttt{pytest} test failures, etc.)\ are excluded. Tool families broken out within each sub-plot.}
\label{fig:metrics-deep-toolerr}
\end{figure}

\begin{figure}[p]
\centering
\setlength{\tabcolsep}{1pt}
\renewcommand{\arraystretch}{0.5}
\begin{tabular}{ccc}
\includegraphics[width=0.31\linewidth]{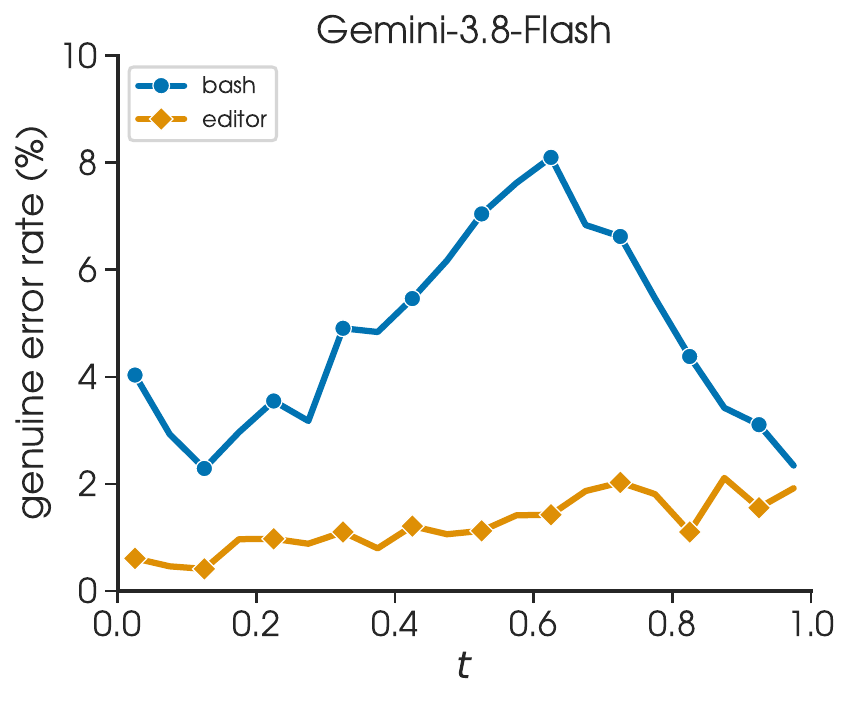} &
\includegraphics[width=0.31\linewidth]{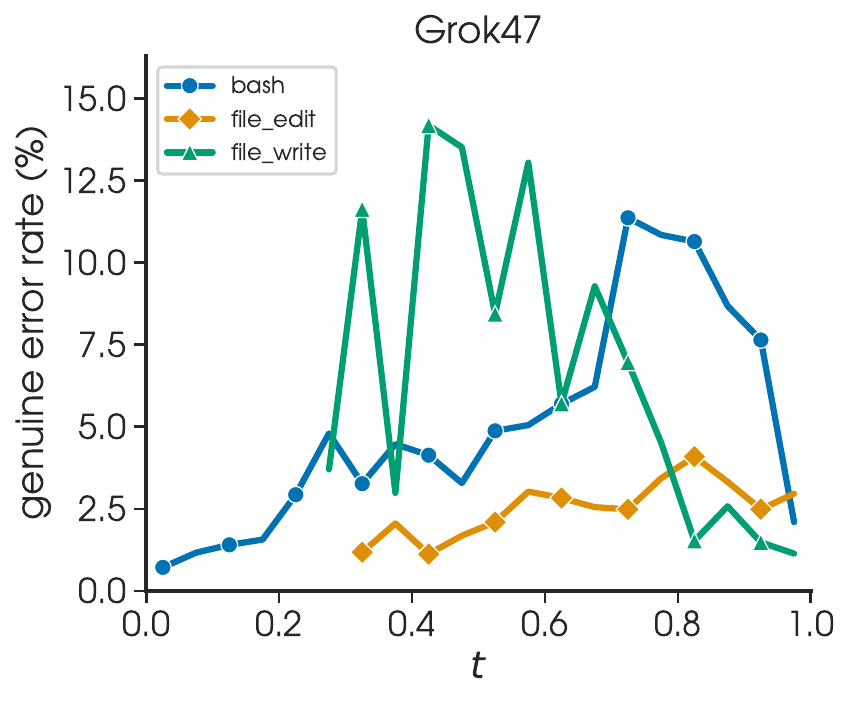} &
\includegraphics[width=0.31\linewidth]{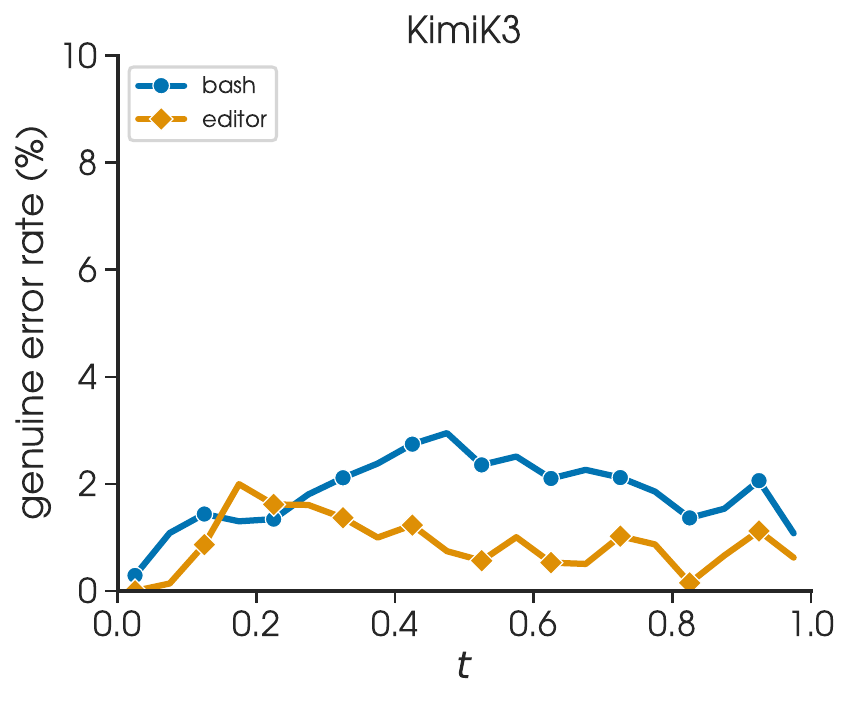} \\
\includegraphics[width=0.31\linewidth]{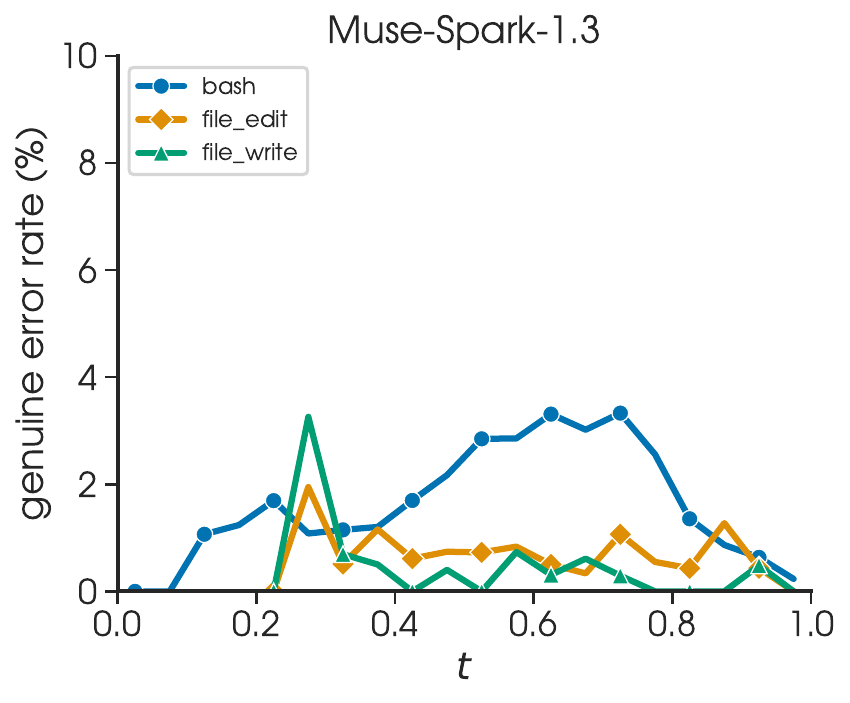} &
\includegraphics[width=0.31\linewidth]{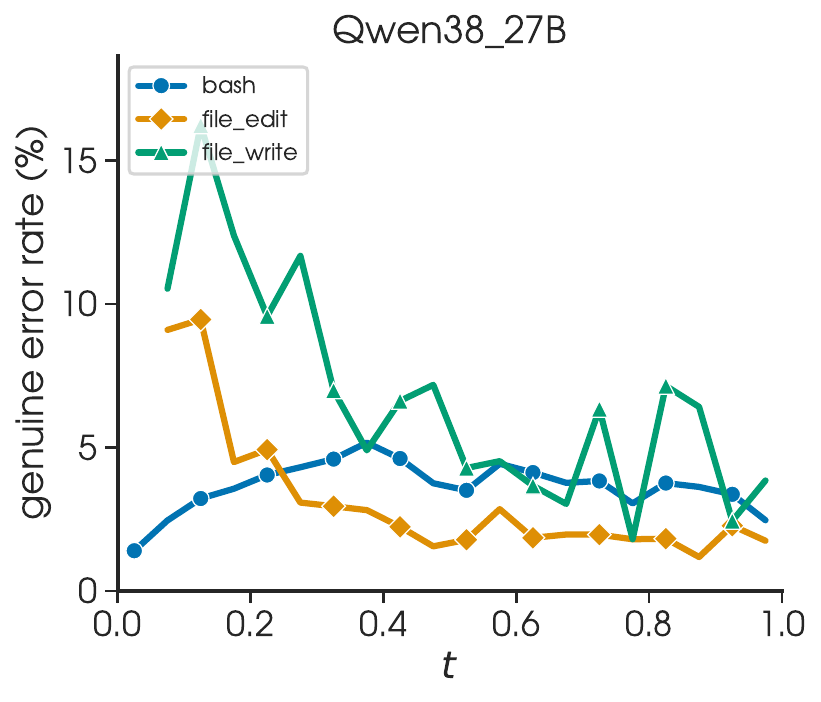} &
 \\
\end{tabular}
\caption{Genuine tool-error rate per cycle on Deep-SWE (part 2 of 2): Gemini, Grok, Kimi, Muse Spark and Qwen families. Continued from Figure~\ref{fig:metrics-deep-toolerr}.}
\label{fig:metrics-deep-toolerr-b}
\end{figure}

\begin{figure}[p]
\centering
\includegraphics[width=0.65\linewidth]{swe_verif/phase_legend.pdf}\\[2pt]
\setlength{\tabcolsep}{1pt}
\renewcommand{\arraystretch}{0.5}
\begin{tabular}{ccc}
\includegraphics[width=0.31\linewidth]{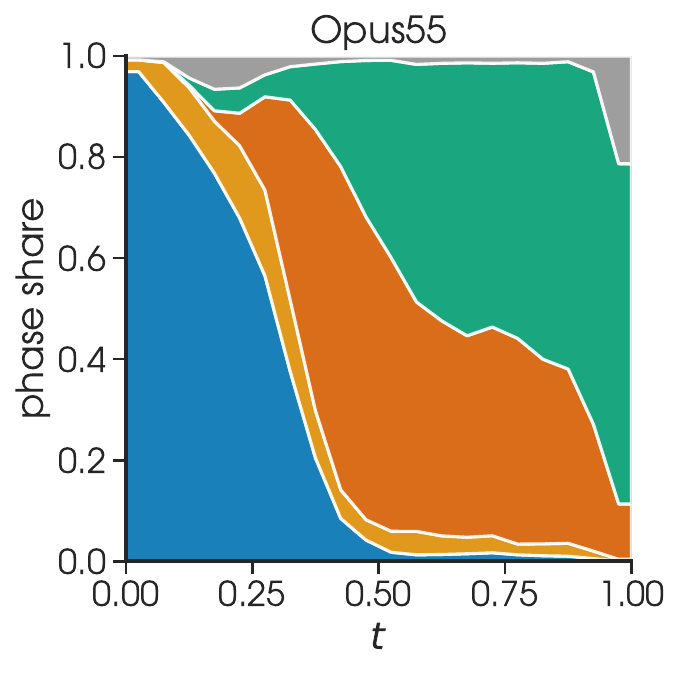} &
\includegraphics[width=0.31\linewidth]{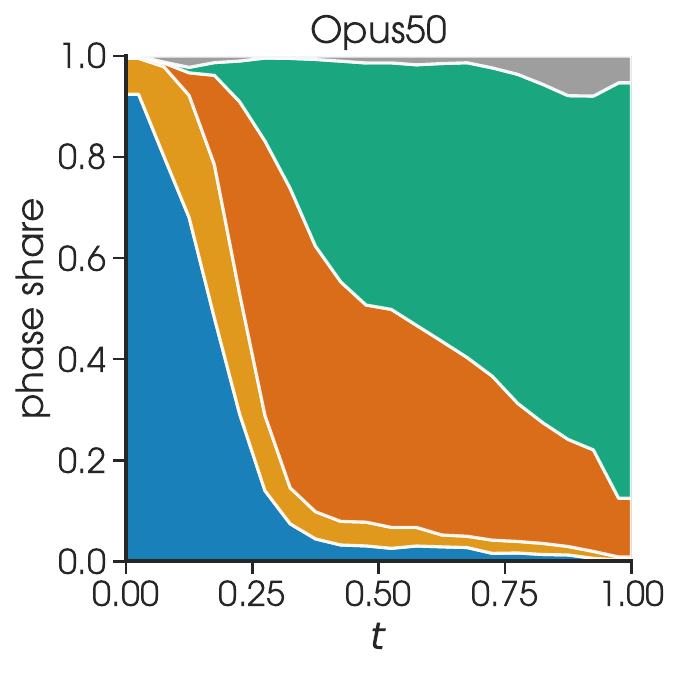} &
\includegraphics[width=0.31\linewidth]{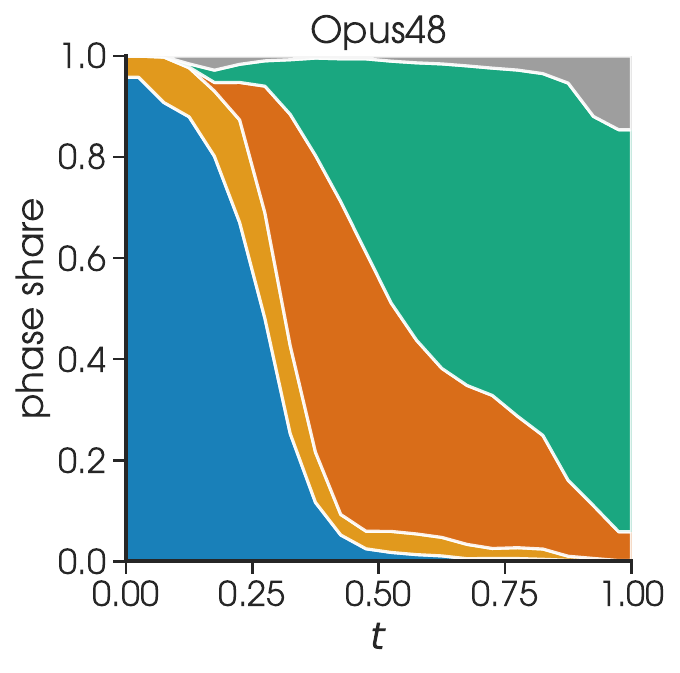} \\
\includegraphics[width=0.31\linewidth]{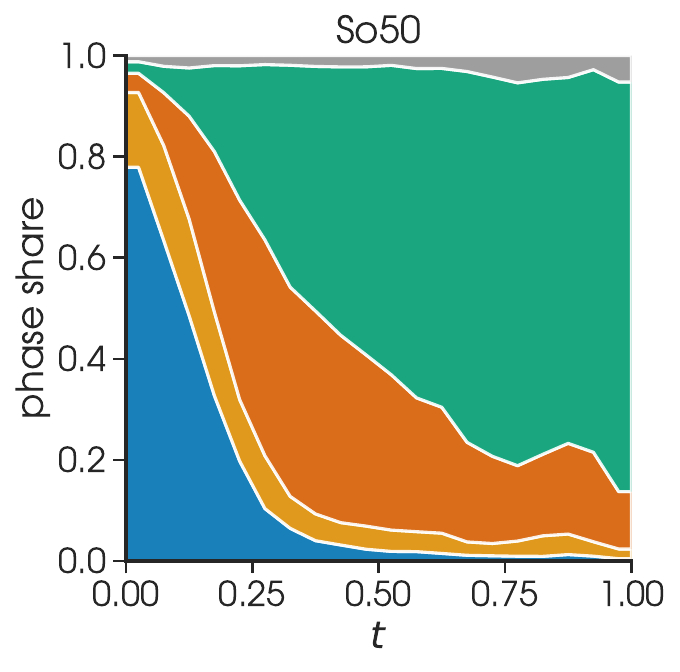} &
\includegraphics[width=0.31\linewidth]{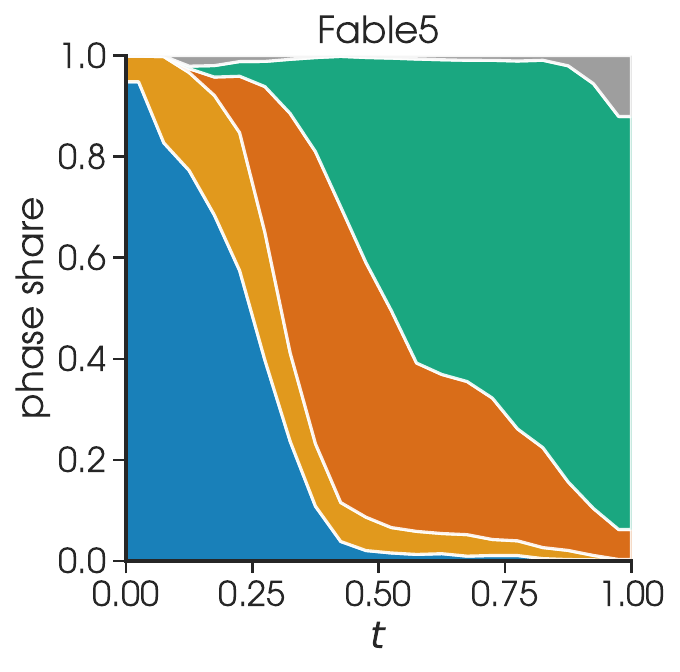} &
\includegraphics[width=0.31\linewidth]{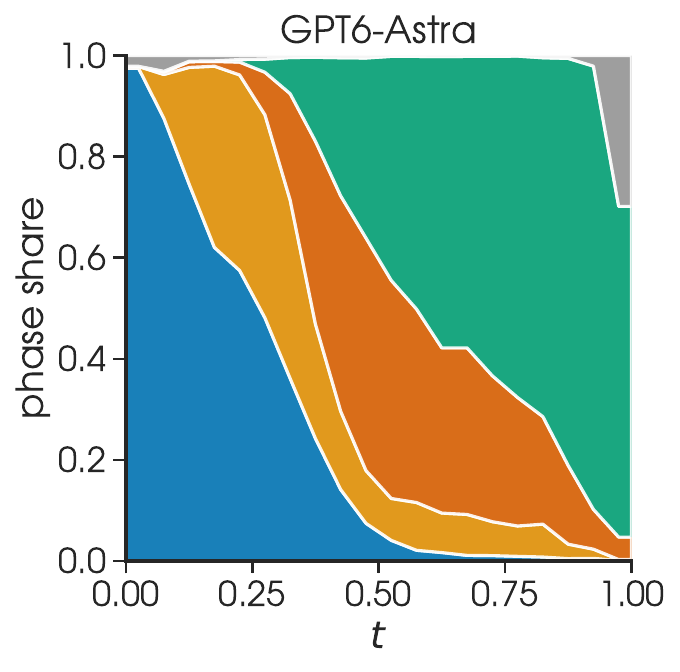} \\
\includegraphics[width=0.31\linewidth]{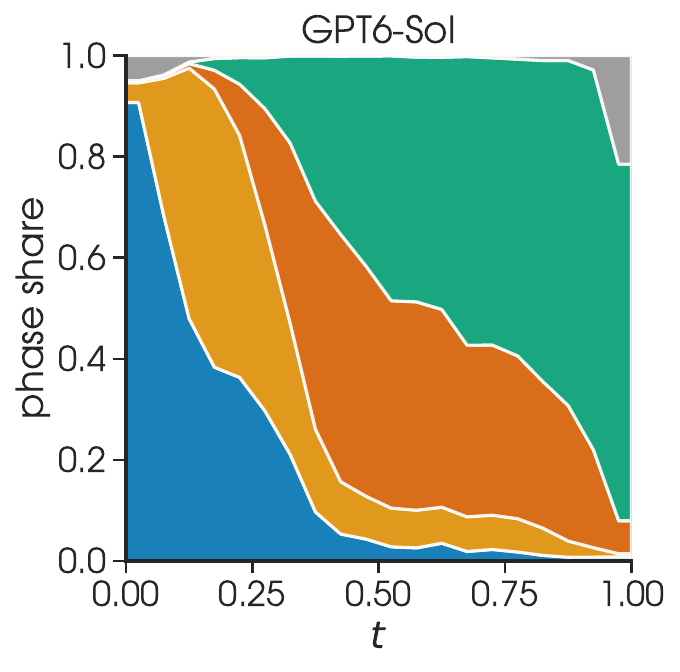} &
\includegraphics[width=0.31\linewidth]{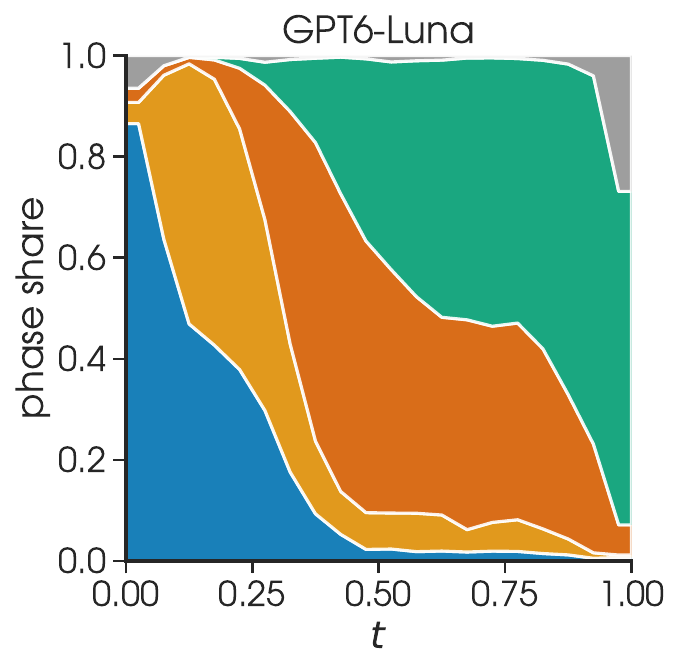} &
\includegraphics[width=0.31\linewidth]{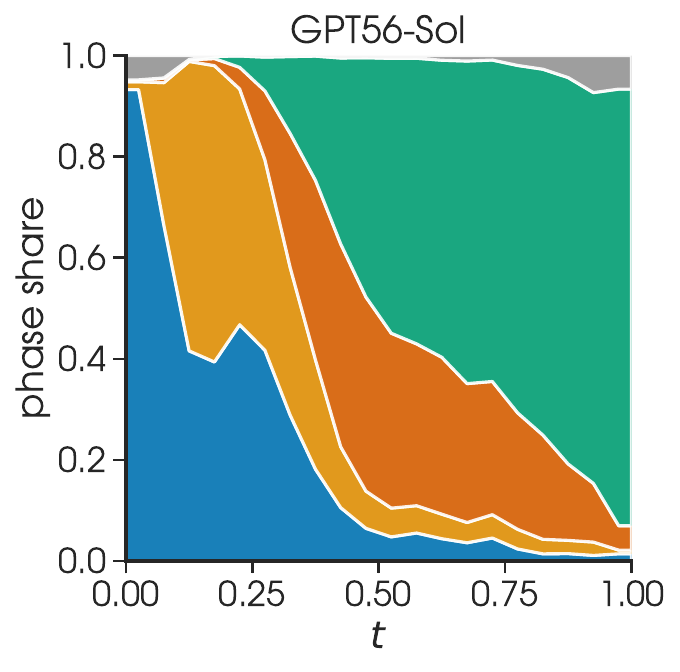} \\
\includegraphics[width=0.31\linewidth]{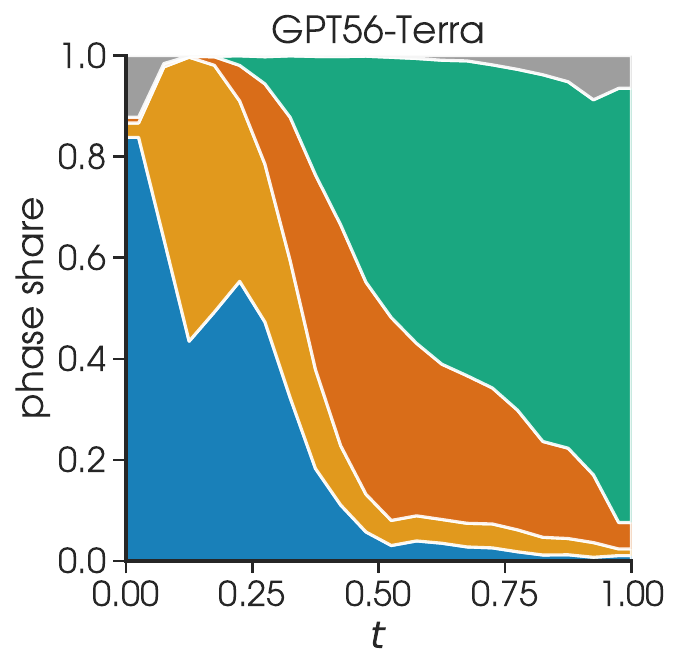} &
\includegraphics[width=0.31\linewidth]{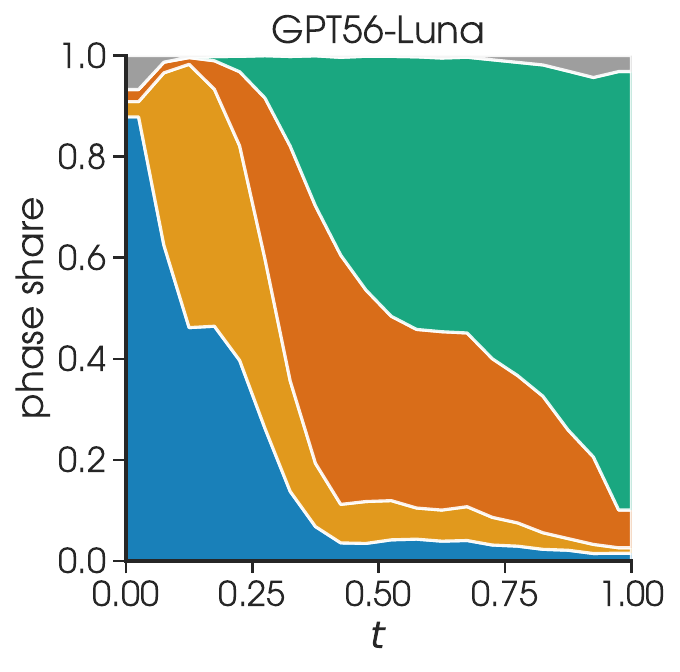} &
\includegraphics[width=0.31\linewidth]{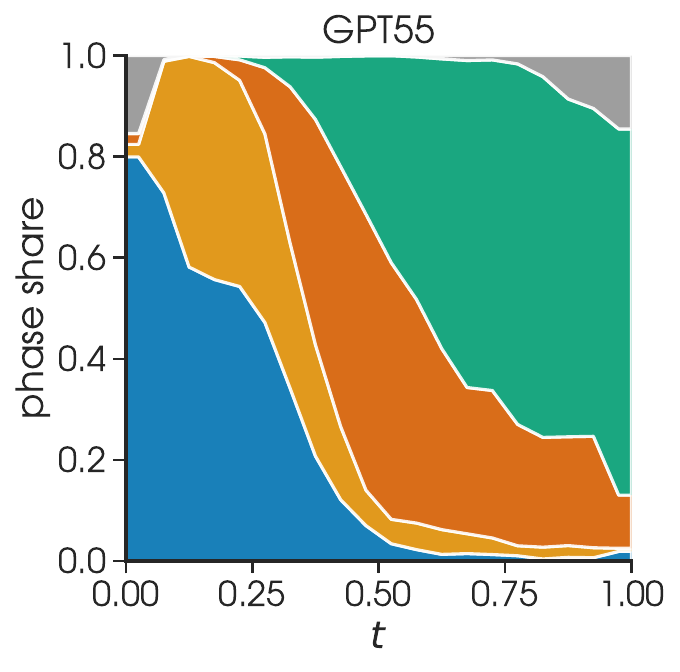} \\
\end{tabular}
\caption{\textbf{Phase composition surfaces model-specific schedules.} Per-cycle share of the four R6 (see LLM judge rubrics in Table\,\ref{tab:judge-buckets}) phases (\texttt{explore},
\texttt{localize}, \texttt{implement}, \texttt{verify}) vs.\ normalised cycle
position $t$ on Deep-SWE, for Claude and GPT families (part 1 of 2). Given the
same tasks, models schedule their cycles differently. The implementation peak shifts in position and width, and the 
verify phase carries different long-tail mass.}
\label{fig:metrics-deep-phase}
\end{figure}

\begin{figure}[p]
\centering
\includegraphics[width=0.65\linewidth]{swe_verif/phase_legend.pdf}\\[2pt]
\setlength{\tabcolsep}{1pt}
\renewcommand{\arraystretch}{0.5}
\begin{tabular}{ccc}
\includegraphics[width=0.31\linewidth]{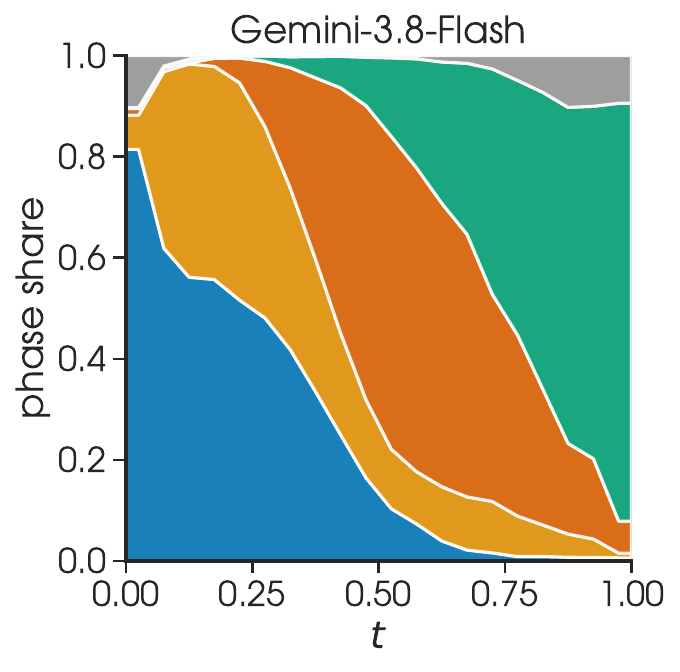} &
\includegraphics[width=0.31\linewidth]{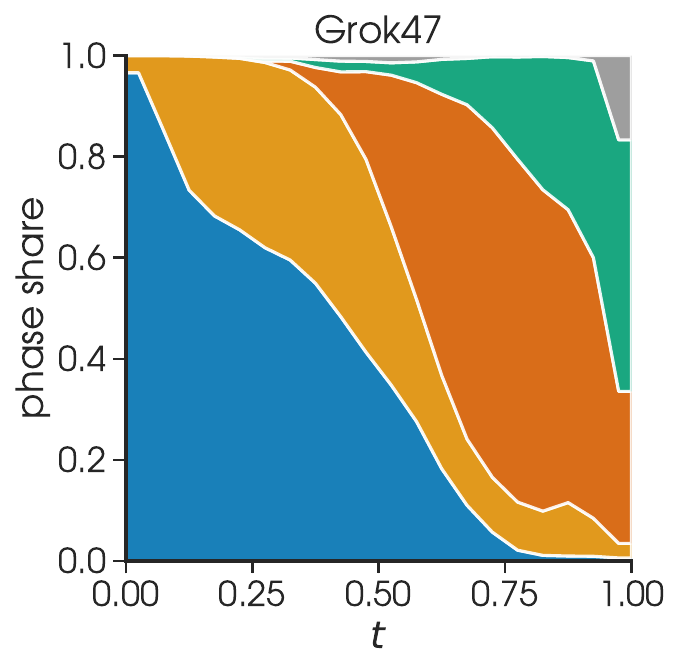} &
\includegraphics[width=0.31\linewidth]{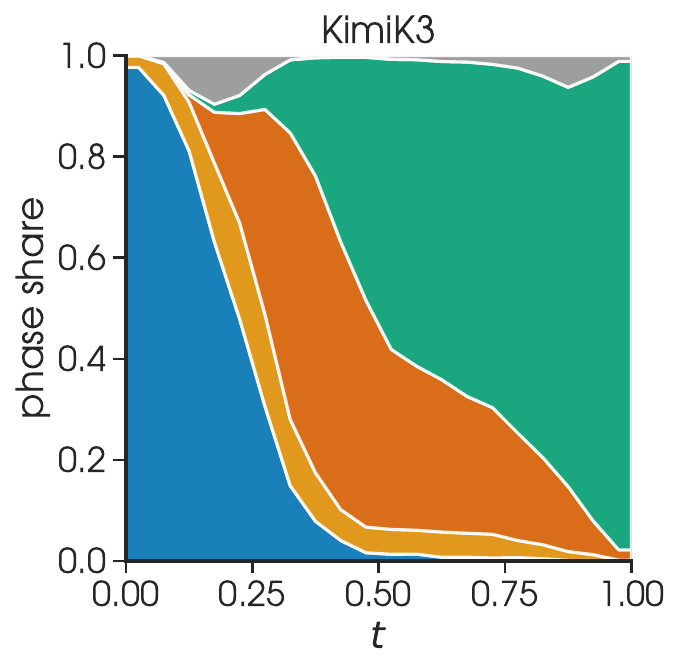} \\
\includegraphics[width=0.31\linewidth]{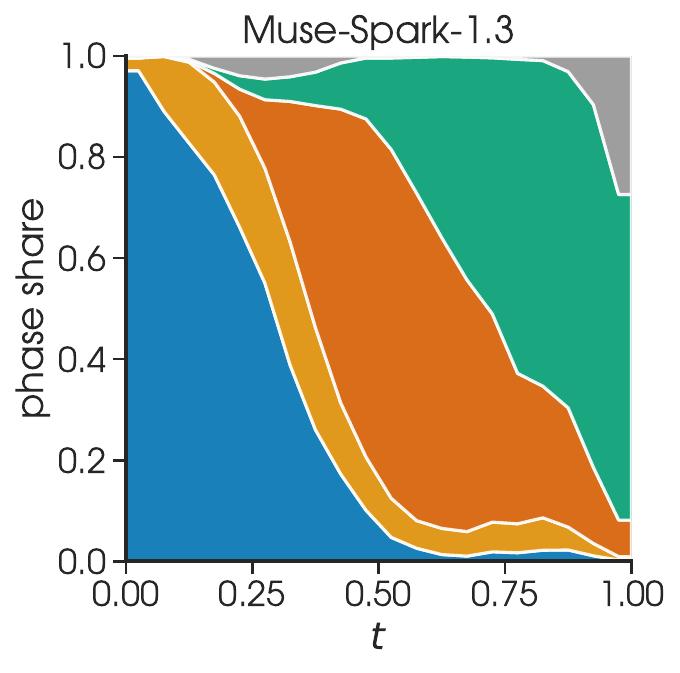} &
\includegraphics[width=0.31\linewidth]{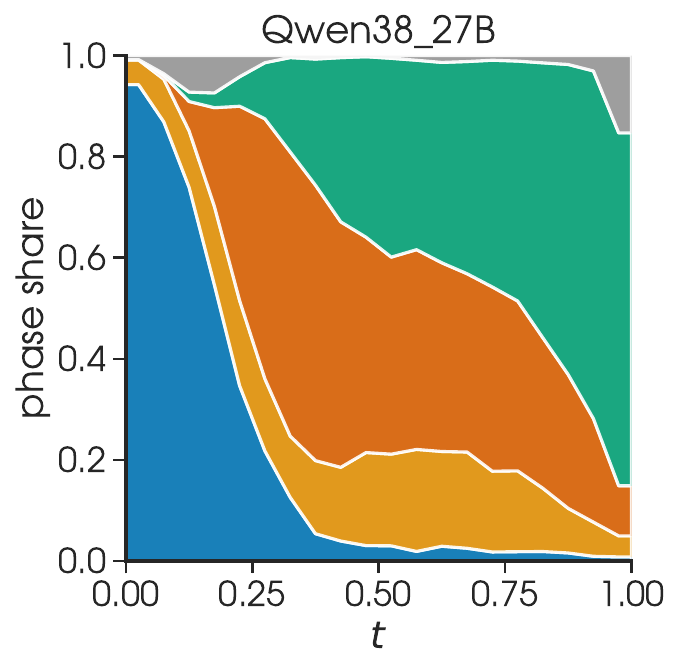} &
 \\
\end{tabular}
\caption{Phase composition on Deep-SWE (part 2 of 2): Gemini, Grok, Kimi, Muse Spark and Qwen families. Continued from Figure~\ref{fig:metrics-deep-phase}.}
\label{fig:metrics-deep-phase-b}
\end{figure}

\paragraph{Per-model figures (Deep-SWE).}
Each figure pair below follows the format of its SWE-Bench-Verified counterpart. Figures~\ref{fig:metrics-deep-backtracking}--\ref{fig:metrics-deep-backtracking-b} show the per-edit $\Delta D$ histogram for each model, as in Figure~\ref{fig:metrics-sbv-backtracking}: green bars are progress edits ($\Delta D<0$), red bars are backtracking edits ($\Delta D>0$), and the annotation is the backtrack-edit share. Figures~\ref{fig:metrics-deep-editread}--\ref{fig:metrics-deep-editread-b} show the per-cycle share of source-editing, scratch-testing and suite-testing tool calls across resolved and unresolved trajectories, as in Figure~\ref{fig:metrics-sbv-editread}. Figures~\ref{fig:metrics-deep-soldist}--\ref{fig:metrics-deep-soldist-b} show the level family, mean $\bar D(t)$ and backtracking term $\bar B(t)$ of the resolved trajectories for each model, as in Figure~\ref{fig:metrics-sbv-soldist}. Figures~\ref{fig:metrics-deep-toolerr}--\ref{fig:metrics-deep-toolerr-b} plot the rate of \emph{genuine} tool errors per cycle (R8 = \texttt{genuine}, see Section~\ref{app:metrics:rubric}), broken out by tool family (\texttt{bash} vs.\ \texttt{editor}), as in Figure~\ref{fig:metrics-sbv-toolerr}; benign non-zero exits are excluded. Figures~\ref{fig:metrics-deep-phase}--\ref{fig:metrics-deep-phase-b} show how each trajectory's time is split across the four R6 phases (\texttt{explore}, \texttt{localize}, \texttt{implement}, \texttt{verify}; see Table\,\ref{tab:judge-buckets}), as in Figure~\ref{fig:metrics-sbv-phase}.

\clearpage
\section{Per-Model Pass Rates by Family}
\label{app:passrates}

This appendix gives the full numbers summarized by Tables~\ref{tab:results-reproduction} and~\ref{tab:results-reproduction-deep}: for every model and every benchmark it was run on, SSA's five-run pass@1 with its $95\%$ confidence interval, pass@2 where reported, and the figure the provider publishes together with its source. Tables are organized by model family, in alphabetical order, so that each can be read against its own provider's published figures; they are not intended for cross-family comparison. Terminal-Bench-2 is reported under both the Constrained setting (the benchmark's memory and time budgets) and the Unconstrained setting (budgets removed, an upper bound), with the published figure placed under the setting the provider reports. Opus~5.5, Opus~5.0, Opus~4.8, Sonnet~5.0, Fable~5 and GPT-5.5 were added after the main sweep: they were run on Deep-SWE and, where a row appears below, on SWE-Bench-Verified or SWE-Bench-Pro, but not on Terminal-Bench-2, so they do not appear in Table~\ref{tab:results-reproduction}.

\begin{table}[h]
\centering
\caption{Per-model pass rates, Claude family, grouped by benchmark. SSA pass@1 is the mean of five per-run resolution rates with a $95\%$ confidence interval over all per-instance trials; pass@2 is the fraction of instances resolved by at least one of two attempts (reported for the SWE benchmarks only). ``Published'' is the provider's model-card figure; for Deep-SWE it is the DataCurve v1.1 artifact (\texttt{mini-swe-agent}, best reasoning-effort setting, pass@1 over four attempts) or, where marked $^{\ast}$, the vendor's own system or model card. ``---'' no figure published; ``n/a'' not reported. Mark as in Table~\ref{tab:results-reproduction}. SWE-Bench-Pro figures are from the Base public containers.}
\label{tab:app-pass-claude}
\footnotesize
\setlength{\tabcolsep}{6pt}
\renewcommand{\arraystretch}{1.0}
\begin{tabular}{@{}lcccc@{}}
\toprule
Model & SSA pass@1 (\%) & pass@2 (\%) & Published (\%) & Mark \\
\midrule
\famrow{5}{SWE-Bench-Verified} \\
\modelname{Opus 5.0} & $96.60 \pm 1.85$ & 97.28 & 96.0 & \rep \\
\modelname{Opus 4.8} & $88.80 \pm 1.23$ & 91.66 & 88.6 & \rep \\
\modelname{Opus 4.6} & $80.80 \pm 1.54$ & 82.54 & 80.8 & \rep \\
\modelname{Opus 4.5} & $80.48 \pm 1.55$ & 83.30 & 80.9 & \rep \\
\modelname{Sonnet 5.0} & $84.04 \pm 1.43$ & 86.96 & 85.0 & \rep \\
\modelname{Sonnet 4.6} & $79.00 \pm 1.59$ & 82.06 & 79.6 & \rep \\
\modelname{Sonnet 4.5} & $76.40 \pm 1.66$ & 79.94 & 77.2 & \rep \\
\modelname{Sonnet 4.0} & $72.00 \pm 1.76$ & 76.82 & 72.7 & \rep \\
\modelname{Haiku 4.5} & $72.60 \pm 1.74$ & 77.30 & 73.3 & \rep \\
\addlinespace[2pt]
\famrow{5}{SWE-Bench-Pro} \\
\modelname{Opus 5.0} & $79.53 \pm 1.30$ & 82.67 & 79.2 & \rep \\
\modelname{Opus 4.8} & $68.48 \pm 1.50$ & 73.02 & --- & \nopub \\
\modelname{Opus 4.6} & $57.45 \pm 1.60$ & 63.10 & 53.4 & \exc \\
\modelname{Opus 4.5} & $54.58 \pm 1.61$ & 58.72 & 52.0 & \exc \\
\modelname{Sonnet 4.6} & $53.67 \pm 1.61$ & 59.83 & --- & \nopub \\
\modelname{Sonnet 4.5} & $50.15 \pm 1.62$ & 56.22 & --- & \nopub \\
\modelname{Sonnet 4.0} & $44.26 \pm 1.61$ & 51.09 & --- & \nopub \\
\modelname{Haiku 4.5} & $46.37 \pm 1.61$ & 53.14 & --- & \nopub \\
\modelname{Fable 5} & $79.12 \pm 1.31$ & 82.85 & 80.0 & \rep \\
\addlinespace[2pt]
\famrow{5}{Terminal-Bench-2 (Constrained)} \\
\modelname{Opus 4.6} & $73.93 \pm 4.08$ & n/a & 65.4 & \exc \\
\modelname{Opus 4.5} & $65.39 \pm 4.40$ & n/a & 59.8 & \exc \\
\modelname{Sonnet 4.6} & $64.71 \pm 4.44$ & n/a & 59.1 & \exc \\
\modelname{Sonnet 4.5} & $49.21 \pm 4.63$ & n/a & 51.0 & \rep \\
\modelname{Sonnet 4.0} & $41.80 \pm 4.56$ & n/a & --- & \nopub \\
\modelname{Haiku 4.5} & $41.57 \pm 4.56$ & n/a & 41.0 & \rep \\
\addlinespace[2pt]
\famrow{5}{Terminal-Bench-2 (Unconstrained)} \\
\modelname{Opus 4.6} & $83.14 \pm 3.48$ & n/a & --- & \nopub \\
\modelname{Opus 4.5} & $72.35 \pm 4.15$ & n/a & --- & \nopub \\
\modelname{Sonnet 4.6} & $74.15 \pm 4.07$ & n/a & --- & \nopub \\
\modelname{Sonnet 4.5} & $54.83 \pm 4.62$ & n/a & --- & \nopub \\
\modelname{Sonnet 4.0} & $42.47 \pm 4.59$ & n/a & --- & \nopub \\
\modelname{Haiku 4.5} & $43.37 \pm 4.60$ & n/a & --- & \nopub \\
\addlinespace[2pt]
\famrow{5}{Deep-SWE} \\
\modelname{Opus 5.5} & $74.33 \pm 3.60$ & 80.70 & 74.2$^{\ast}$ & \rep \\
\modelname{Opus 5.0} & $72.56 \pm 3.68$ & 81.77 & 68.8$^{\ast}$ & \exc \\
\modelname{Opus 4.8} & $61.28 \pm 4.02$ & 75.66 & 59.0 & \rep \\
\modelname{Sonnet 5.0} & $57.25 \pm 4.07$ & 73.18 & 53.9 & \rep \\
\modelname{Fable 5} & $68.84 \pm 3.82$ & 77.78 & 69.9 & \rep \\
\bottomrule
\end{tabular}
\end{table}

\begin{table}[h]
\centering
\caption{Per-model pass rates, Gemini family, grouped by benchmark. SSA pass@1 is the mean of five per-run resolution rates with a $95\%$ confidence interval over all per-instance trials; pass@2 is the fraction of instances resolved by at least one of two attempts (reported for the SWE benchmarks only). ``Published'' is the provider's model-card figure; for Deep-SWE it is the DataCurve v1.1 artifact (\texttt{mini-swe-agent}, best reasoning-effort setting, pass@1 over four attempts) or, where marked $^{\ast}$, the vendor's own system or model card. ``---'' no figure published; ``n/a'' not reported. Mark as in Table~\ref{tab:results-reproduction}. SWE-Bench-Pro figures are from the Base public containers.}
\label{tab:app-pass-gemini}
\footnotesize
\setlength{\tabcolsep}{6pt}
\renewcommand{\arraystretch}{1.0}
\begin{tabular}{@{}lcccc@{}}
\toprule
Model & SSA pass@1 (\%) & pass@2 (\%) & Published (\%) & Mark \\
\midrule
\famrow{5}{SWE-Bench-Verified} \\
\modelname{Gemini 3.1 Pro} & $80.80 \pm 1.54$ & 84.14 & 80.6 & \rep \\
\modelname{Gemini 3 Flash} & $78.32 \pm 2.08$ & 81.76 & 78.0 & \rep \\
\addlinespace[2pt]
\famrow{5}{SWE-Bench-Pro} \\
\modelname{Gemini 3.1 Pro} & $54.00 \pm 1.61$ & 58.72 & 54.2 & \rep \\
\modelname{Gemini 3 Flash} & $49.43 \pm 1.62$ & 55.40 & --- & \nopub \\
\addlinespace[2pt]
\famrow{5}{Terminal-Bench-2 (Constrained)} \\
\modelname{Gemini 3.1 Pro} & $72.13 \pm 4.17$ & n/a & 68.5 & \rep \\
\modelname{Gemini 3 Flash} & $63.59 \pm 4.47$ & n/a & 47.6 & \exc \\
\addlinespace[2pt]
\famrow{5}{Terminal-Bench-2 (Unconstrained)} \\
\modelname{Gemini 3.1 Pro} & $76.00 \pm 4.00$ & n/a & --- & \nopub \\
\modelname{Gemini 3 Flash} & $66.29 \pm 4.39$ & n/a & --- & \nopub \\
\addlinespace[2pt]
\famrow{5}{Deep-SWE} \\
\modelname{Gemini 3.8 Flash} & $73.10 \pm 3.65$ & 81.76 & 73.7$^{\ast}$ & \rep \\
\bottomrule
\end{tabular}
\end{table}

\begin{table}[h]
\centering
\caption{Per-model pass rates, GPT family, grouped by benchmark. SSA pass@1 is the mean of five per-run resolution rates with a $95\%$ confidence interval over all per-instance trials; pass@2 is the fraction of instances resolved by at least one of two attempts (reported for the SWE benchmarks only). ``Published'' is the provider's model-card figure; for Deep-SWE it is the DataCurve v1.1 artifact (\texttt{mini-swe-agent}, best reasoning-effort setting, pass@1 over four attempts) or, where marked $^{\ast}$, the vendor's own system or model card. ``---'' no figure published; ``n/a'' not reported. Mark as in Table~\ref{tab:results-reproduction}. SWE-Bench-Pro figures are from the Base public containers.}
\label{tab:app-pass-openai}
\footnotesize
\setlength{\tabcolsep}{6pt}
\renewcommand{\arraystretch}{1.0}
\begin{tabular}{@{}lcccc@{}}
\toprule
Model & SSA pass@1 (\%) & pass@2 (\%) & Published (\%) & Mark \\
\midrule
\famrow{5}{SWE-Bench-Verified} \\
\modelname{GPT-5.4} & $80.16 \pm 1.56$ & 83.40 & --- & \nopub \\
\modelname{GPT-5.3} & $79.52 \pm 1.58$ & 82.20 & --- & \nopub \\
\modelname{GPT-5.2} & $79.28 \pm 1.58$ & 81.72 & 80.0 & \rep \\
\modelname{GPT-5.2-codex} & $76.80 \pm 1.65$ & 79.98 & --- & \nopub \\
\modelname{GPT-OSS 120B} & $64.64 \pm 1.87$ & 71.90 & 62.0 & \exc \\
\modelname{GPT-OSS 20B} & $59.84 \pm 1.92$ & 68.54 & 60.0 & \rep \\
\addlinespace[2pt]
\famrow{5}{SWE-Bench-Pro} \\
\modelname{GPT-5.5} & $59.42 \pm 1.60$ & 61.79 & 58.6 & \rep \\
\modelname{GPT-5.4} & $59.90 \pm 1.58$ & 64.77 & 57.7 & \exc \\
\modelname{GPT-5.3} & $58.57 \pm 1.59$ & 62.59 & 56.8 & \exc \\
\modelname{GPT-5.2} & $56.47 \pm 1.60$ & 60.00 & 56.0 & \rep \\
\modelname{GPT-5.2-codex} & $53.87 \pm 1.61$ & 62.59 & 56.4 & \abv \\
\modelname{GPT-OSS 120B} & $43.61 \pm 1.60$ & 51.10 & --- & \nopub \\
\modelname{GPT-OSS 20B} & $39.67 \pm 1.58$ & 48.22 & --- & \nopub \\
\addlinespace[2pt]
\famrow{5}{Terminal-Bench-2 (Constrained)} \\
\modelname{GPT-5.4} & $72.58 \pm 4.02$ & n/a & 75.1 & \rep \\
\modelname{GPT-5.3} & $75.05 \pm 4.02$ & n/a & 77.3 & \rep \\
\modelname{GPT-5.2} & $62.69 \pm 4.49$ & n/a & 62.2 & \rep \\
\modelname{GPT-5.2-codex} & $68.76 \pm 4.63$ & n/a & 64.7 & \rep \\
\addlinespace[2pt]
\famrow{5}{Terminal-Bench-2 (Unconstrained)} \\
\modelname{GPT-5.4} & $79.10 \pm 3.78$ & n/a & --- & \nopub \\
\modelname{GPT-5.3} & $80.67 \pm 3.67$ & n/a & --- & \nopub \\
\modelname{GPT-5.2} & $74.60 \pm 4.04$ & n/a & --- & \nopub \\
\modelname{GPT-5.2-codex} & $73.70 \pm 4.07$ & n/a & --- & \nopub \\
\modelname{GPT-OSS 120B} & $36.17 \pm 4.46$ & n/a & --- & \nopub \\
\modelname{GPT-OSS 20B} & $30.11 \pm 4.26$ & n/a & --- & \nopub \\
\addlinespace[2pt]
\famrow{5}{Deep-SWE} \\
\modelname{GPT-6-Astra} & $77.70 \pm 3.43$ & 81.85 & 74.1$^{\ast}$ & \exc \\
\modelname{GPT-6-Sol} & $76.10 \pm 3.51$ & 81.95 & 68.8$^{\ast}$ & \exc \\
\modelname{GPT-6-Luna} & $64.24 \pm 3.95$ & 76.01 & 66.6$^{\ast}$ & \rep \\
\modelname{GPT-5.6-Sol} & $72.56 \pm 3.68$ & 81.32 & 72.7$^{\ast}$ & \rep \\
\modelname{GPT-5.6-Terra} & $69.20 \pm 3.80$ & 77.52 & 69.6 & \rep \\
\modelname{GPT-5.6-Luna} & $65.13 \pm 3.93$ & 77.25 & 67.2 & \rep \\
\modelname{GPT-5.5} & $74.51 \pm 3.51$ & 85.04 & 67.0 & \exc \\
\bottomrule
\end{tabular}
\end{table}

\begin{table}[h]
\centering
\caption{Per-model pass rates, Grok family, grouped by benchmark. SSA pass@1 is the mean of five per-run resolution rates with a $95\%$ confidence interval over all per-instance trials; pass@2 is the fraction of instances resolved by at least one of two attempts (reported for the SWE benchmarks only). ``Published'' is the provider's model-card figure; for Deep-SWE it is the vendor's own system card, marked $^{\ast}$. ``---'' no figure published; ``n/a'' not reported. Mark as in Table~\ref{tab:results-reproduction}. SWE-Bench-Pro figures are from the Base public containers.}
\label{tab:app-pass-xai}
\footnotesize
\setlength{\tabcolsep}{6pt}
\renewcommand{\arraystretch}{1.0}
\begin{tabular}{@{}lcccc@{}}
\toprule
Model & SSA pass@1 (\%) & pass@2 (\%) & Published (\%) & Mark \\
\midrule
\famrow{5}{SWE-Bench-Verified} \\
\modelname{Grok 4.2} & $77.56 \pm 1.63$ & 80.76 & --- & \nopub \\
\addlinespace[2pt]
\famrow{5}{SWE-Bench-Pro} \\
\modelname{Grok 4.2} & $51.57 \pm 1.62$ & 57.31 & --- & \nopub \\
\addlinespace[2pt]
\famrow{5}{Terminal-Bench-2 (Constrained)} \\
\modelname{Grok 4.2} & $52.35 \pm 4.64$ & n/a & --- & \nopub \\
\addlinespace[2pt]
\famrow{5}{Terminal-Bench-2 (Unconstrained)} \\
\modelname{Grok 4.2} & $56.40 \pm 4.61$ & n/a & --- & \nopub \\
\addlinespace[2pt]
\famrow{5}{Deep-SWE} \\
\modelname{Grok 4.7} & $72.39 \pm 3.68$ & 83.18 & 71.0$^{\ast}$ & \rep \\
\bottomrule
\end{tabular}
\end{table}

\begin{table}[h]
\centering
\caption{Per-model pass rates, Kimi family, grouped by benchmark. SSA pass@1 is the mean of five per-run resolution rates with a $95\%$ confidence interval over all per-instance trials; pass@2 is the fraction of instances resolved by at least one of two attempts (reported for the SWE benchmarks only). ``Published'' is the provider's model-card figure; for Deep-SWE it is the DataCurve v1.1 artifact (\texttt{mini-swe-agent}, best reasoning-effort setting, pass@1 over four attempts) or, where marked $^{\ast}$, the vendor's own system or model card. ``---'' no figure published; ``n/a'' not reported. Mark as in Table~\ref{tab:results-reproduction}. SWE-Bench-Pro figures are from the Base public containers.}
\label{tab:app-pass-kimi}
\footnotesize
\setlength{\tabcolsep}{6pt}
\renewcommand{\arraystretch}{1.0}
\begin{tabular}{@{}lcccc@{}}
\toprule
Model & SSA pass@1 (\%) & pass@2 (\%) & Published (\%) & Mark \\
\midrule
\famrow{5}{Deep-SWE} \\
\modelname{Kimi-K3} & $77.00 \pm 3.47$ & 86.46 & 67.5$^{\ast}$ & \exc \\
\bottomrule
\end{tabular}
\end{table}

\begin{table}[h]
\centering
\caption{Per-model pass rates, Muse Spark family, grouped by benchmark. SSA pass@1 is the mean of five per-run resolution rates with a $95\%$ confidence interval over all per-instance trials; pass@2 is the fraction of instances resolved by at least one of two attempts (reported for the SWE benchmarks only). ``Published'' is the provider's model-card figure; for Deep-SWE it is the DataCurve v1.1 artifact (\texttt{mini-swe-agent}, best reasoning-effort setting, pass@1 over four attempts) or, where marked $^{\ast}$, the vendor's own system or model card. ``---'' no figure published; ``n/a'' not reported. Mark as in Table~\ref{tab:results-reproduction}. SWE-Bench-Pro figures are from the Base public containers.}
\label{tab:app-pass-musespark}
\footnotesize
\setlength{\tabcolsep}{6pt}
\renewcommand{\arraystretch}{1.0}
\begin{tabular}{@{}lcccc@{}}
\toprule
Model & SSA pass@1 (\%) & pass@2 (\%) & Published (\%) & Mark \\
\midrule
\famrow{5}{Deep-SWE} \\
\modelname{Muse Spark 1.3} & $75.22 \pm 3.56$ & 86.10 & 75.4$^{\ast}$ & \rep \\
\bottomrule
\end{tabular}
\end{table}

\begin{table}[h]
\centering
\caption{Per-model pass rates, Qwen family, grouped by benchmark. SSA pass@1 is the mean of five per-run resolution rates with a $95\%$ confidence interval over all per-instance trials; pass@2 is the fraction of instances resolved by at least one of two attempts (reported for the SWE benchmarks only). ``Published'' is the provider's model-card figure; for Deep-SWE it is the vendor's own model card, marked $^{\ast}$. ``---'' no figure published; ``n/a'' not reported. Mark as in Table~\ref{tab:results-reproduction}. SWE-Bench-Pro figures are from the Base public containers.}
\label{tab:app-pass-qwen}
\footnotesize
\setlength{\tabcolsep}{6pt}
\renewcommand{\arraystretch}{1.0}
\begin{tabular}{@{}lcccc@{}}
\toprule
Model & SSA pass@1 (\%) & pass@2 (\%) & Published (\%) & Mark \\
\midrule
\famrow{5}{SWE-Bench-Verified} \\
\modelname{Qwen3-Coder Next} & $72.44 \pm 1.75$ & 78.20 & 71.3 & \rep \\
\modelname{Qwen3-Coder 480B} & $67.56 \pm 1.83$ & 73.66 & 69.6 & \abv \\
\modelname{Qwen3-Coder 30B} & $51.36 \pm 1.95$ & 59.46 & 51.6 & \rep \\
\modelname{Qwen3.6 35B-A3B} & $72.92 \pm 1.74$ & 77.76 & 73.4 & \rep \\
\modelname{Qwen3.6 27B} & $76.24 \pm 1.67$ & 80.52 & 77.2 & \rep \\
\modelname{Qwen3.5 27B} & $74.36 \pm 1.71$ & 79.50 & 72.4 & \exc \\
\addlinespace[2pt]
\famrow{5}{SWE-Bench-Pro} \\
\modelname{Qwen3-Coder Next} & $57.75 \pm 1.60$ & 69.82 & 44.3 & \exc \\
\modelname{Qwen3-Coder 480B} & $38.90 \pm 1.58$ & 46.42 & --- & \nopub \\
\modelname{Qwen3-Coder 30B} & $27.41 \pm 1.44$ & 33.96 & --- & \nopub \\
\modelname{Qwen3.6 35B-A3B} & $46.70 \pm 1.61$ & 56.47 & 49.5 & \abv \\
\modelname{Qwen3.6 27B} & $51.40 \pm 1.62$ & 58.49 & 53.5 & \abv \\
\modelname{Qwen3.5 27B} & $47.44 \pm 1.61$ & 66.06 & 51.2 & \abv \\
\addlinespace[2pt]
\famrow{5}{Terminal-Bench-2 (Unconstrained)} \\
\modelname{Qwen3-Coder Next} & $38.87 \pm 4.53$ & n/a & 36.2 & \rep \\
\modelname{Qwen3-Coder 480B} & $35.73 \pm 4.45$ & n/a & --- & \nopub \\
\modelname{Qwen3-Coder 30B} & $17.07 \pm 3.50$ & n/a & --- & \nopub \\
\modelname{Qwen3.6 35B-A3B} & $48.76 \pm 4.64$ & n/a & 51.5 & \rep \\
\modelname{Qwen3.6 27B} & $56.00 \pm 4.61$ & n/a & 59.3 & \rep \\
\modelname{Qwen3.5 27B} & $50.78 \pm 4.65$ & n/a & 41.6 & \exc \\
\addlinespace[2pt]
\famrow{5}{Deep-SWE} \\
\modelname{Qwen3.8 27B} & $44.77 \pm 4.10$ & 61.68 & 42.2$^{\ast}$ & \rep \\
\bottomrule
\end{tabular}
\end{table}

\clearpage
\section{Terminal-Bench-2: Infrastructure Sensitivity}
\label{app:tb2-infra}

Terminal-Bench-2 imposes both compute limits (task memory, storage, CPU) and time limits (agent and verifier timeouts). While these limits standardize the use of compute to make benchmark scores comparable, they also have the unintended side-effect of conflating agent quality with infrastructure stability. Dependency installation, backend latency, and retries draw on the same budget as problem
solving~\citep{merrill2026terminalbench,kwon2023vllm}. From looking at the logs, we noticed two system factors that stood out. First, the inference
backend's throughput (tokens-per-minute and requests-per-minute) must cover all
concurrent projects for the full duration of evaluation. High invoker-latency variance,
timeouts, and retries eat into the total time budget allocated for problem-solving and reduce the resolution rate. 
Second, packing multiple projects onto one node
(e.g.\ Harbor with \texttt{n\_concurrent}~$>1$) splits available network bandwidth across
them, slowing the dependency downloads required by nearly every task and leaving
less time before the agent is interrupted.

\textbf{Command batching interacts with each family's reasoning budget differently.} This split is a
direct consequence of the reasoning/tool-use relation. Since most tool calls are
shell commands, an obvious mitigation of high interaction
overhead due to sequential command execution is a batch interface that executes
several commands per turn. The effect is mixed. Batching forces the model to maintain a coherent
terminal state across steps, which raises reasoning complexity. For Claude, batching shifts
effort from round-trips into in-context reasoning; for Gemini and Grok, 
batch execution helps because it does not trigger additional
reasoning. Under the constrained setting, batching is therefore not a
uniform win, and it mirrors the failure mode of
Section~\ref{sec:methods}: the right operating point on the
reasoning-versus-interaction trade-off is family-specific, not global.

\textbf{Relaxing the time and memory limits raises absolute pass@1 by
$5$--$10$ points.} We interpret the
Unconstrained setting (Appendix~\ref{app:passrates}) as an upper-bound of model performance on Terminal-Bench-2.
Across models, removing the memory and agent-time limits lifts pass@1 by about
$5$--$10$ absolute points, with most of the gains coming from a handful of projects that
repeatedly time out under the constraints but solve reliably without them ---
\texttt{make-doom-for-mips}, \texttt{torch-pipeline-parallelism},
\texttt{gpt2-codegolf}, \texttt{caffe-cifar-10}, and \texttt{train-fasttext}.

\clearpage
\section{Evaluation Integrity}
\label{app:eval-integrity}

\textbf{The public SWE-Bench-Pro containers leak future git history.} Distinct from
training-data contamination~\citep{magar2022contamination,sainz2023contamination,jain2024livecodebench},
the public images leave post-\texttt{base\_commit} objects reachable even
though \texttt{HEAD} is checked out at the instance's \texttt{base\_commit}. An agent can recover the intended fix through ordinary commands such as
\texttt{git log -p} or \texttt{git show <future-sha>}. We compare the public
\emph{Base} containers against \emph{Sanitized} ones whose working tree is
identical but whose future commits and dev-branches are pruned before the agent starts. 
The containers expose the fix to any agent that inspects git history in the course of ordinary 
exploration, so removing them lowers measured pass@1 most for the models whose exploration 
includes git history: Opus~4.6 by
$6.87$ points ($57.45\%\!\to\!50.58\%$, $4.67\%$ of instances including a \texttt{git show} of the gold-patch commit), GPT-5.4 by $3.32$ points
($5.33\%$ of instances with \texttt{git show}), and Sonnet~4.6
by $3.66$ points. Other models are flat or move within trajectory noise, confirming the leakage channel exists
but is exercised unevenly. In an ablation where the model is explicitly instructed to  ``inspect git history'' in its system prompt, leakage effect is significantly larger: it raises Qwen3.6~27B by $+17.13\%$ (compared to sanitized)
and Qwen3-Coder Next to a $+21.12\%$ (compared to sanitized). We, therefore, treat Sanitized as the cleaner estimate of model
ability and recommend that SWE-Bench-Pro comparisons either use Sanitized
scores or explicitly report leakage exposure. The sanitization
protocol and full per-model Base-Sanitized table are in
Appendix~\ref{app:swe-pro-git-leakage}.

\textbf{A few residual evaluation artifacts cap the achievable ceiling.} On
SWE-Bench-Verified, instances such as
\texttt{astropy-8872} and \texttt{astropy-8707} fail even when the agent's patch
is correct, because of setup inconsistencies that require fixes to the
evaluation environment; some \texttt{psf\_requests} instances fail
intermittently on nonresponsive external test URLs and need manual patching for
reliable scoring~\citep{jimenez2024swebench}. On SWE-Bench-Pro, run on Amazon ECS\footnote{https://aws.amazon.com/ecs/},
$3$ of $731$ instances consistently fail under AWS-specific environment
assumptions, and an approximate $0.41\%$ ceiling loss is suffered uniformly across all
SSA runs.

\section{SWE-Bench-Pro Git Leakage}
\label{app:swe-pro-git-leakage}

The public SWE-Bench-Pro~\citep{deng2025swebenchpro} containers are checked out at
the instance's \texttt{base\_commit}, but they can still retain future git
objects, branches, tags, or reflogs. This creates a runtime leakage channel, i.e., 
an agent can inspect repository history during evaluation and place
post-\texttt{base\_commit} information into its context. This differs from
training-data contamination~\citep{magar2022contamination,sainz2023contamination,jain2024livecodebench} as
the model need not memorize a solution if the benchmark environment itself can
reveal it through ordinary commands such as \texttt{git log -p},
\texttt{git branch -a}, or \texttt{git show <future-sha>}.

\subsection{Sanitized evaluation}
\label{app:swe-pro-git-leakage-protocol}

For a SWE-Pro instance with cutoff point \texttt{base\_commit}, a temporally
isolated environment should expose only the checked-out source tree, tests,
dependencies, and repository history up to that commit. We therefore compare two
regimes. \textbf{Base} uses the official public container as released:
\texttt{HEAD} is at \texttt{base\_commit}, but future git state may remain
reachable. \textbf{Sanitized} keeps the working tree identical to Base while
removing future history before the agent starts: remotes and refs are removed and a
single local \texttt{main} branch is recreated at the original \texttt{HEAD}.

The main impact statistic is \textbf{Base--Sanitized Pass@1}. If future information does not matter to the agent, removing it should not systematically reduce Pass@1 beyond
ordinary trajectory variance. If the agent benefited from future information, Base accuracy should degrade under Sanitized evaluation.

\subsection{Direct leakage metric}
\label{app:swe-pro-git-leakage-proxy}

Let $\mathcal{I}_{\mathrm{SBP}}$ be the 731 public SWE-Pro instances, and let
$s_i$ be the gold fix commit for instance $i$. For each model, we count a
\emph{confirmed direct leak} when a resolved Base trajectory contains
\texttt{git show} applied to $s_i$ in the agent context before termination. Let $L_i$ be the leakage indicator that is exactly $1$ when the context contains
\texttt{git show <gold-patch-sha>} for that instance AND the run solves the Base
task. The reported direct leakage
percentage is
\[
  \mathrm{DirectLeakage}
  = \mathbb{E}[L_i].
\]
The empirical expectation is over all public SWE-Bench-Pro instances (731), so the number is
directly comparable to Pass@1.

This metric is intentionally a lower bound. It captures only exact-SHA exposure.
It misses future information obtained through \texttt{git log -p},
\texttt{git diff} against branches or tags, \texttt{git blame}, merge commits,
cherry-picks, release commits, nearby commits that contain the same edit,
follow-up commits that reveal the intended invariant, and commit messages that
localize the fix without exposing the exact gold SHA. Thus, direct leakage is
confirmed evidence of exposure, not an estimate of all possible leakage. The
Base minus Sanitized Pass@1 gap is the broader aggregate measure.

\begin{figure}[t]
    \centering
    \includegraphics[width=\linewidth]{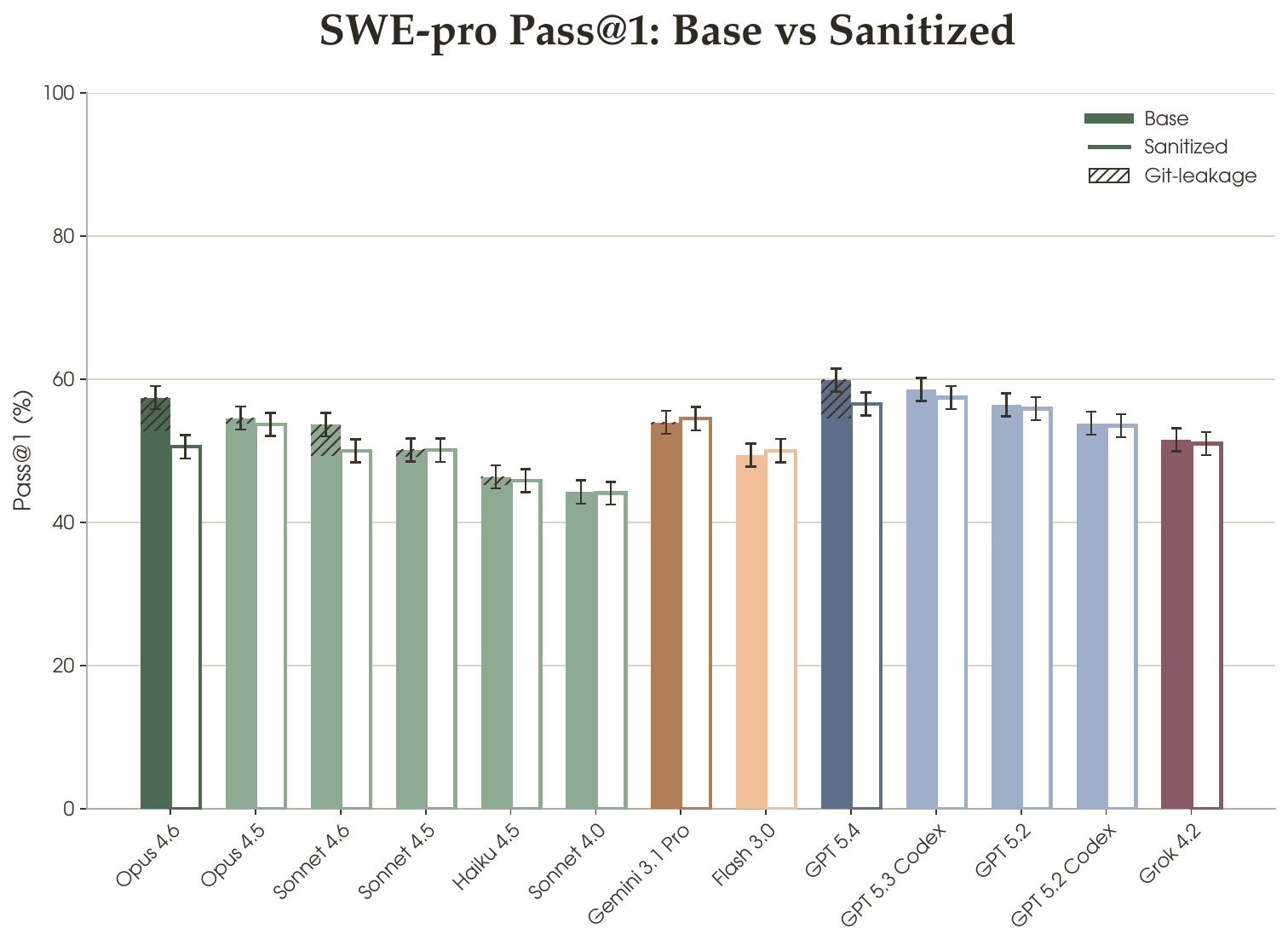}
    \caption{SWE-Pro Pass@1 under Base and Sanitized containers. The shaded
    overlay is the confirmed direct leakage fraction: resolved Base instances
    whose agent context contains \protect\texttt{git show <gold-patch-sha>},
    normalized by all 731 SWE-Bench-Pro instances.}
    \label{fig:app-swe-pro-base-vs-sanitized}
\end{figure}

\subsection{Empirical impact}
\label{app:swe-pro-git-leakage-impact}

Figure~\ref{fig:app-swe-pro-base-vs-sanitized} shows that sanitization can
materially change measured SWE-Bench-Pro performance. Opus 4.6 drops from
57.45\% Pass@1 in Base to 50.58\% under Sanitized evaluation, a 6.87\%
decrease, with 4.67\% of its Base instances include a \texttt{git show} of the gold-patch commit. GPT 5.4 drops from 59.90\% to
56.58\% and has the largest direct leakage fraction, 5.33\%. Sonnet 4.6 also
shows a clear signal: 53.67\% Base, 50.01\% Sanitized, and 4.32\% confirmed
direct leakage.

\begin{table}[t]
\centering
\small
\caption{SWE-Pro Base and Sanitized Pass@1, with a fraction of instances in which a Base run accessed the gold-patch commit. Leakage is counted when a resolved Base trajectory contains
\protect\texttt{git show <gold-patch-sha>} in the agent context; percentages use all
731 SWE-Pro instances as the denominator.}
\label{tab:app-swe-pro-git-leakage}
\begin{tabular}{lrrrr}
\toprule
Model & Base \% & Sanitized \% & Gap \% & Direct leakage \% \\
\midrule
Opus 4.6 & 57.45 & 50.58 & 6.87 & 4.67 \\
Opus 4.5 & 54.58 & 53.70 & 0.88 & 0.65 \\
Sonnet 4.6 & 53.67 & 50.01 & 3.66 & 4.32 \\
Sonnet 4.5 & 50.15 & 50.10 & 0.05 & 0.90 \\
Haiku 4.5 & 46.37 & 45.82 & 0.55 & 1.09 \\
Sonnet 4.0 & 44.26 & 44.10 & 0.16 & 0.00 \\
Gemini 3.1 Pro & 54.00 & 54.50 & -0.50 & 0.21 \\
Flash 3.0 & 49.43 & 50.04 & -0.61 & 0.00 \\
GPT 5.4 & 59.90 & 56.58 & 3.32 & 5.33 \\
GPT 5.3 Codex & 58.57 & 57.45 & 1.12 & 0.00 \\
GPT 5.2 & 56.47 & 55.90 & 0.57 & 0.00 \\
GPT 5.2 Codex & 53.87 & 53.51 & 0.36 & 0.00 \\
Grok 4.2 & 51.57 & 51.02 & 0.55 & 0.00 \\
\bottomrule
\end{tabular}
\end{table}

Table~\ref{tab:app-swe-pro-git-leakage} gives the numerical view. The gap and
the direct leakage fraction are related but not interchangeable. The gap can be
larger than direct leakage because the metric misses indirect and nearby-commit
leakage. It can also be smaller because an exact \texttt{git show} exposure may
not be necessary for the final solve, or the agent may not realize to overwrite its current solution with the exposed one. Near-zero gaps or small Sanitized improvements, such as Gemini 3.1
Pro and Flash 3.0, are best read as ordinary trajectory variance. The central
finding is that Base containers expose a real future-information channel, and its
measured effect depends on each model's tool-use behavior.

\subsection{Open-source models and explicit git instructions}
\label{app:swe-pro-git-leakage-opensource}

The results above use each model's default instructions (prompts and tool spec). To stress-test the
channel, we also evaluate four Qwen models with \textbf{Git Instructions}, a
prompt variant that explicitly asks the agent to inspect git history while
debugging. This regime estimates how much Pass@1 can increase when the leakage
path is made salient rather than discovered naturally.

The Git Instructions experiment appends the following block to the agent's system
prompt, directing it to consult version-control history before editing:

\begin{tcolorbox}[promptbox, title=Git Instructions (appended to system prompt)]
\begin{lstlisting}[style=prompt]
Additional instructions:
- Before editing ANY file <file>, you MUST first run `git log -p -- <file>` AND `git blame <file>`. Reading history in your head does not count.
- For each symbol named in the <pr_description>, you MUST run `git log -S'<symbol>' --all` to find when/why it was introduced or changed.
- If a git command returns no useful history, state that explicitly and proceed -- do not silently skip the call.
\end{lstlisting}
\end{tcolorbox}

\begin{figure}[t]
    \centering
    \includegraphics[width=0.95\linewidth]{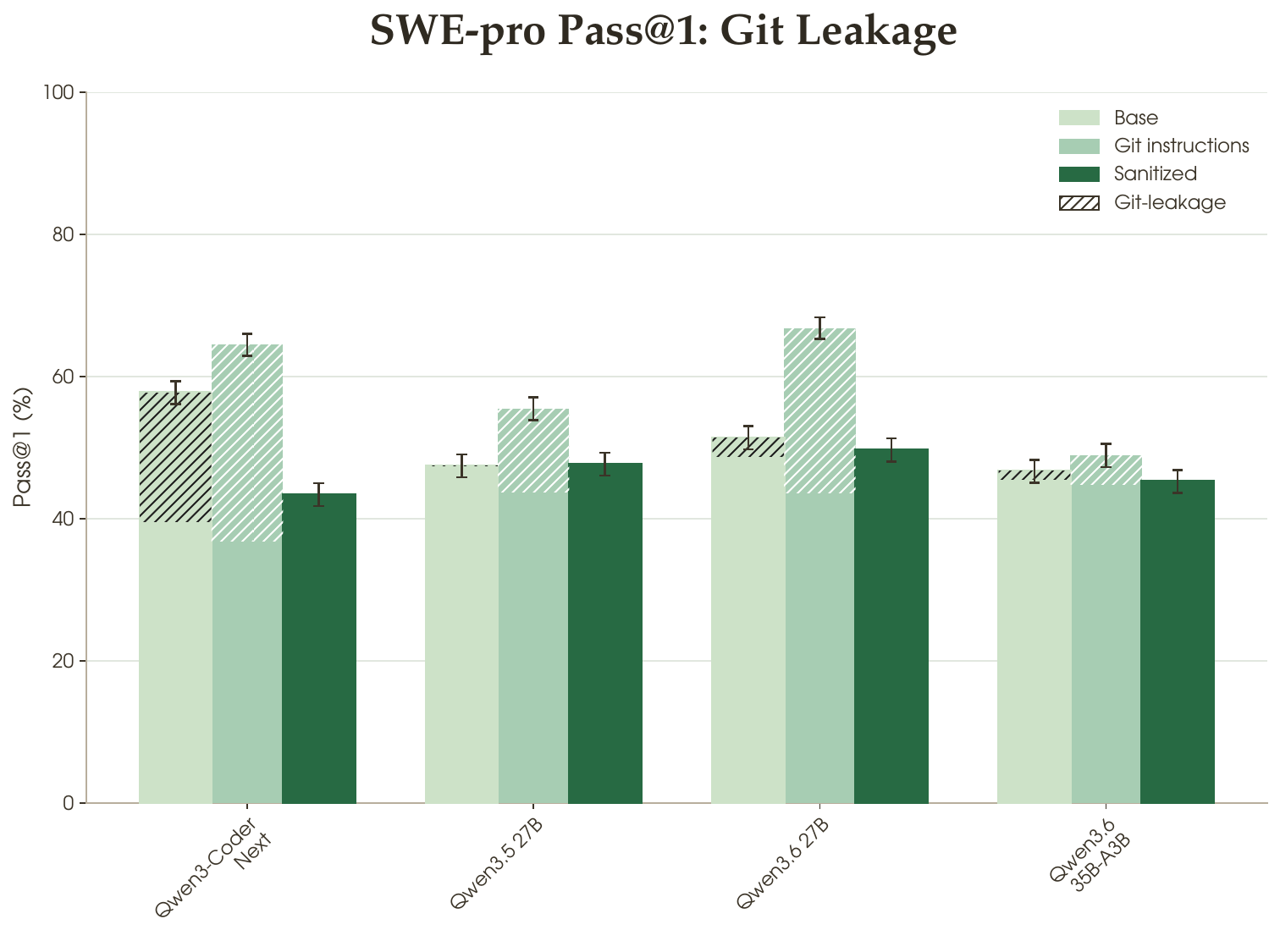}
    \caption{SWE-Bench-Pro Pass@1 for Qwen models under Base, Git Instructions, and
    Sanitized regimes. Explicit git instructions raise Pass@1 for all four
    models; Sanitized scores remove future git history while keeping the working
    tree fixed.}
    \label{fig:app-swe-pro-opensource-leakage}
\end{figure}

\begin{table}[t]
\centering
\small
\setlength{\tabcolsep}{3pt}
\caption{Qwen models on SWE-Bench-Pro under Base, Git Instructions, and Sanitized
regimes. Pass@1 cells are percentages with bootstrap 95\% confidence intervals
where available; leakage columns are percentage-point estimates attributable to
git leakage.}
\label{tab:app-swe-pro-opensource-leakage}
\begin{tabular}{@{}lrrrrr@{}}
\toprule
Model & Base (\%) & Base leak. (\%) & Git Instr. (\%) & Git leak. (\%) & Sanitized (\%) \\
\midrule
Qwen3-Coder Next & $57.75 \pm 1.60$ & 18.11 & $64.51 \pm 1.55$ & 27.60 & $43.39 \pm 1.60$ \\
Qwen3.5 27B      & $47.44 \pm 1.61$ &  0.01 & $55.51 \pm 1.61$ & 11.79 & $47.68 \pm 1.61$ \\
Qwen3.6 27B      & $51.40 \pm 1.62$ &  2.60 & $66.81 \pm 1.52$ & 23.14 & $49.68 \pm 1.62$ \\
Qwen3.6 35B-A3B  & $46.70 \pm 1.61$ &  1.17 & $48.91 \pm 1.62$ &  4.02 & $45.22 \pm 1.61$ \\
\bottomrule
\end{tabular}
\end{table}

Figure~\ref{fig:app-swe-pro-opensource-leakage} and
Table~\ref{tab:app-swe-pro-opensource-leakage} show the expected pattern.
Explicit git instructions raise Pass@1 for Qwen3.6~27B by 17.13\% and for
Qwen3.5~27B by 7.83\%. Qwen3-Coder Next also gains 21.12\% under Git
Instructions, carries the largest Git-Instructions leakage estimate (27.60\%),
and drops to 43.39\% after sanitization. Across all four models,
Sanitized scores are below Git Instructions scores, confirming that future git
information can account for a substantial part of measured SWE-Bench-Pro performance.

For reporting, Sanitized is the cleaner estimate of SWE-Bench-Pro repair ability
because it preserves the checked-out task state while removing future repository
states that should be unobservable at the benchmark cutoff point. Base scores are
still useful for documenting the public containers as released, but model
comparisons should either use Sanitized results or explicitly report git-leakage
exposure.

\section{Configurations}
\label{sec:appendix-configs}

All runs use an AWS PCS\footnote{https://aws.amazon.com/pcs/} cluster of \texttt{c7.48xlarge} instances with a
maximum concurrency of $10$. Claude models are served through Amazon Bedrock\footnote{https://aws.amazon.com/bedrock/} at
production capacity; OpenAI, Gemini, and Grok models are accessed through their
commercial APIs. Open-weights models are self-served via vLLM \citep{kwon2023vllm} (see exact serving commands in Appendix~\ref{app:vllm}). We use
vendor-recommended high-reasoning settings wherever a model card documents them
(e.g.\ \texttt{adaptive} thinking at \texttt{max} effort for the Claude~4.x family,
reasoning effort \texttt{xhigh} for the GPT-5 family, thinking level \texttt{high} for
Gemini); the complete per-model inference settings, provider clients, prompt
tags, bound tools, and reasoning controls for every benchmark
configuration are summarized below, and shipped with the open-sourced SSA package for reproducibility.

We enforce strict evaluation hygiene. Internet access is disabled for
SWE-Bench-Verified and SWE-Bench-Pro and enabled for Terminal-Bench-2, as its
tasks require it. For both SWE benchmarks we use the standard Docker
environments, whose working trees are checked out at the instance's
\texttt{base\_commit} and are \emph{intended} to expose only history up to that
revision~\citep{jimenez2024swebench,deng2025swebenchpro,magar2022contamination,sainz2023contamination,jain2024livecodebench}.
Because the public SWE-Bench-Pro images can nonetheless retain reachable future
git objects, we separately audit and sanitize that temporal-leakage channel in
Appendices~\ref{app:eval-integrity} and~\ref{app:swe-pro-git-leakage}.

The repository ships 80 benchmark configurations spanning the four
benchmarks reported in Section~\ref{sec:results}: SWE-Bench-Verified
(SBV; 21 configs), SWE-Bench-Pro (SBP; 21 configs), Terminal-Bench-2
(TB2; 21 configs) and Deep-SWE (17 configs). Each config is a Hydra-mergeable~\citep{yadan2019hydra} YAML that fixes the
model identifier, the invoker (provider client), the prompt tag (resolved
against \texttt{SimpleResolverAgent.yaml}, see Section~\ref{app:prompts}),
the bound tools, and the reasoning controls.

The summaries below report the raw values stored in
\texttt{src/ssa/configs/}. In summary, For Anthropic models, \texttt{thinking=enabled
(Nk)} gives the extended-thinking token budget, \texttt{thinking=adaptive}
is the adaptive-thinking mode, \texttt{interleaved} indicates the
\texttt{interleaved-thinking-2025-05-14} beta header, and \texttt{effort} is
\texttt{output\_config.effort} (the \texttt{effort-2025-11-24} beta).
\texttt{level=HIGH} is Gemini's \texttt{thinking\_config.thinking\_level};
\texttt{effort=xhigh} is OpenAI's \texttt{reasoning.effort}. For the
open-weights (vLLM) models, \texttt{temperature}, \texttt{top\_p}, \texttt{top\_k},
\texttt{max\_tokens} and \texttt{rep\_penalty} are the sampling parameters. \texttt{defaults} marks a config that sets no sampling or reasoning overrides. Configs not listed inherit
defaults from \texttt{default.yaml} (\texttt{max\_iterations = 500};
\texttt{retry\_on\_no\_change.max\_attempts = 1}; \texttt{timeout = 200s} for
SWE-Bench and Deep-SWE, \texttt{120s} for Terminal-Bench-2. Internet access disabled for
SWE-Bench and Deep-SWE, enabled for Terminal-Bench-2 except where noted.

\clearpage
\subsection{SWE-Bench-Verified}
\label{app:configs:sbv}{\normalsize\setlength{\tabcolsep}{4pt}
\begin{longtable}{@{}>{\raggedright\arraybackslash}p{3.9cm}>{\raggedright\arraybackslash}p{1.3cm}>{\raggedright\arraybackslash}p{4.2cm}>{\raggedright\arraybackslash}p{3.6cm}@{}}
\caption{SBV configurations (21).}\label{tab:configs-sbv-full}\\
\toprule
Model & Invoker & Inference config & Prompt tag \\
\midrule
\endfirsthead
\multicolumn{4}{c}{\tablename\ \thetable{} -- continued}\\
\toprule
Model & Invoker & Inference config & Prompt tag \\
\midrule
\endhead
\midrule\multicolumn{4}{r}{continued on next page}\\
\endfoot
\bottomrule
\endlastfoot
\texttt{claude-haiku-4-5-\allowbreak{}20251001-v1:0} & bedrock & thinking=enabled (128k), interleaved & \texttt{Anthropic\_v2} \\
\texttt{claude-opus-4-5-\allowbreak{}20251101-v1:0} & bedrock & effort=high & \texttt{Anthropic\_v2} \\
\texttt{claude-opus-4-6-v1} & bedrock & thinking=adaptive, effort=max & \texttt{Anthropic\_v2} \\
\texttt{claude-sonnet-4-\allowbreak{}20250514-v1:0} & bedrock & thinking=enabled (200k), interleaved & \texttt{swe\_generic\_v1} \\
\texttt{claude-sonnet-4-5-\allowbreak{}20250929-v1:0} & bedrock & thinking=enabled (200k), interleaved & \texttt{Anthropic\_v2} \\
\texttt{claude-sonnet-4-6} & bedrock & thinking=adaptive, effort=high & \texttt{Anthropic\_v2} \\
\texttt{gemini-3-flash-preview} & gemini & temp=1.0, level=HIGH & \texttt{swe\_generic\_v2} \\
\texttt{gemini-3.\allowbreak{}1-pro-preview} & gemini & temp=1.0, level=HIGH & \texttt{swe\_generic\_v2} \\
\texttt{gpt-5.\allowbreak{}2} & openai & effort=xhigh & \texttt{swe\_generic\_v2\_gpt} \\
\texttt{gpt-5.\allowbreak{}2-codex} & openai & effort=xhigh & \texttt{swe\_generic\_v2\_gpt} \\
\texttt{gpt-5.\allowbreak{}3-codex} & openai & effort=xhigh & \texttt{swe\_generic\_v2\_gpt} \\
\texttt{gpt-5.\allowbreak{}4} & openai & effort=xhigh & \texttt{swe\_generic\_v2\_gpt} \\
\texttt{gpt-oss-120b} & openai & temp=1.0, top\_p=1.0, effort=high & \texttt{swe\_generic\_v2\_gpt} \\
\texttt{gpt-oss-20b} & openai & temp=1.0, top\_p=1.0, max\_tokens=16,384, effort=high & \texttt{swe\_generic\_v2\_gpt} \\
\texttt{grok-4.\allowbreak{}20-reasoning} & xai & defaults & \texttt{swe\_generic\_v3} \\
\texttt{qwen3-coder-30b-\allowbreak{}a3b-instruct} & openai & temp=0.7, top\_p=0.8, top\_k=20, rep\_penalty=1.05, max\_tokens=65,536 & \texttt{swe\_generic\_think\_v2} \\
\texttt{qwen3-coder-480b-\allowbreak{}a35b-instruct} & openai & temp=0.7, top\_p=0.8, top\_k=20, rep\_penalty=1.05, max\_tokens=65,536 & \texttt{swe\_generic\_think\_v2} \\
\texttt{qwen3-coder-next} & openai & defaults & \texttt{swe\_generic\_think\_v1} \\
\texttt{qwen3.\allowbreak{}5-27b} & openai & temp=1.0, top\_p=0.95, top\_k=20, thinking=disabled & \texttt{swe\_generic\_think\_v1} \\
\texttt{qwen3.\allowbreak{}6-27b} & openai & temp=1.0, top\_p=0.95, top\_k=20, thinking=enabled & \texttt{swe\_generic\_think\_v1} \\
\texttt{qwen3.\allowbreak{}6-35b-a3b} & openai & temp=1.0, top\_p=0.95, top\_k=20, thinking=enabled & \texttt{swe\_generic\_think\_v1} \\
\end{longtable}
}

\clearpage
\subsection{SWE-Bench-Pro}
\label{app:configs:sbp}{\normalsize\setlength{\tabcolsep}{4pt}
\begin{longtable}{@{}>{\raggedright\arraybackslash}p{3.9cm}>{\raggedright\arraybackslash}p{1.3cm}>{\raggedright\arraybackslash}p{4.2cm}>{\raggedright\arraybackslash}p{3.6cm}@{}}
\caption{SBP configurations (21).}\label{tab:configs-sbp-full}\\
\toprule
Model & Invoker & Inference config & Prompt tag \\
\midrule
\endfirsthead
\multicolumn{4}{c}{\tablename\ \thetable{} -- continued}\\
\toprule
Model & Invoker & Inference config & Prompt tag \\
\midrule
\endhead
\midrule\multicolumn{4}{r}{continued on next page}\\
\endfoot
\bottomrule
\endlastfoot
\texttt{claude-haiku-4-5-\allowbreak{}20251001-v1:0} & bedrock & thinking=enabled (128k), interleaved & \texttt{swe\_generic\_v2} \\
\texttt{claude-opus-4-5-\allowbreak{}20251101-v1:0} & bedrock & effort=high & \texttt{swe\_generic\_v2} \\
\texttt{claude-opus-4-6-v1} & bedrock & thinking=adaptive, effort=max & \texttt{swe\_generic\_v2} \\
\texttt{claude-sonnet-4-\allowbreak{}20250514-v1:0} & bedrock & thinking=enabled (200k), interleaved & \texttt{swe\_generic\_v2} \\
\texttt{claude-sonnet-4-5-\allowbreak{}20250929-v1:0} & bedrock & thinking=enabled (200k), interleaved & \texttt{swe\_generic\_v2} \\
\texttt{claude-sonnet-4-6} & bedrock & thinking=adaptive, effort=max & \texttt{swe\_generic\_v2} \\
\texttt{gemini-3-flash-preview} & gemini & temp=1.0, level=HIGH & \texttt{swe\_generic\_v2} \\
\texttt{gemini-3.\allowbreak{}1-pro-preview} & gemini & temp=1.0, level=HIGH & \texttt{swe\_generic\_v2} \\
\texttt{gpt-5.\allowbreak{}2} & openai & effort=xhigh & \texttt{swe\_generic\_v2\_gpt} \\
\texttt{gpt-5.\allowbreak{}2-codex} & openai & effort=xhigh & \texttt{swe\_generic\_v2\_gpt} \\
\texttt{gpt-5.\allowbreak{}4} & openai & effort=xhigh & \texttt{swe\_generic\_v2\_gpt} \\
\texttt{gpt-oss-120b} & openai & max\_tokens=16,384, effort=high & \texttt{swe\_generic\_v2\_gpt} \\
\texttt{gpt-oss-20b} & openai & max\_tokens=16,384, effort=high & \texttt{swe\_generic\_v2\_gpt} \\
\texttt{grok-4.\allowbreak{}20-0309-reasoning} & xai & defaults & \texttt{swe\_generic\_v2} \\
\texttt{qwen3-coder-30b-\allowbreak{}a3b-instruct} & openai & temp=0.7, top\_p=0.8, top\_k=20, rep\_penalty=1.05, max\_tokens=65,536 & \texttt{swe\_generic\_v2} \\
\texttt{qwen3-coder-480b-\allowbreak{}a35b-instruct} & openai & temp=0.7, top\_p=0.8, top\_k=20, rep\_penalty=1.05, max\_tokens=65,536 & \texttt{swe\_generic\_v2} \\
\texttt{qwen3-coder-next} & openai & temp=1.0, top\_p=0.95, top\_k=40 & \texttt{swe\_generic\_v2} \\
\texttt{qwen3.\allowbreak{}5-27b} & openai & temp=1.0, top\_p=0.95, thinking=enabled & \texttt{swe\_generic\_v2} \\
\texttt{qwen3.\allowbreak{}6-27b} & openai & temp=1.0, top\_p=0.95, thinking=enabled & \texttt{swe\_generic\_v2} \\
\texttt{qwen3.\allowbreak{}6-35b-a3b} & openai & temp=1.0, top\_p=0.95, thinking=enabled & \texttt{swe\_generic\_v2} \\
\end{longtable}
}

\clearpage
\subsection{Terminal-Bench-2}
\label{app:configs:tb2}{\normalsize\setlength{\tabcolsep}{4pt}
\begin{longtable}{@{}>{\raggedright\arraybackslash}p{3.9cm}>{\raggedright\arraybackslash}p{1.3cm}>{\raggedright\arraybackslash}p{4.2cm}>{\raggedright\arraybackslash}p{3.6cm}@{}}
\caption{TB2 configurations (21).}\label{tab:configs-tb2-full}\\
\toprule
Model & Invoker & Inference config & Prompt tag \\
\midrule
\endfirsthead
\multicolumn{4}{c}{\tablename\ \thetable{} -- continued}\\
\toprule
Model & Invoker & Inference config & Prompt tag \\
\midrule
\endhead
\midrule\multicolumn{4}{r}{continued on next page}\\
\endfoot
\bottomrule
\endlastfoot
\texttt{claude-haiku-4-5-\allowbreak{}20251001-v1:0} & bedrock & thinking=enabled (128k), interleaved & \texttt{tb2} \\
\texttt{claude-opus-4-5-\allowbreak{}20251101-v1:0} & bedrock & thinking=enabled (128k), interleaved, effort=high & \texttt{tb2} \\
\texttt{claude-opus-4-6-v1} & bedrock & thinking=adaptive, effort=max & \texttt{tb2} \\
\texttt{claude-sonnet-4-\allowbreak{}20250514-v1:0} & bedrock & thinking=enabled (200k), interleaved & \texttt{tb2} \\
\texttt{claude-sonnet-4-5-\allowbreak{}20250929-v1:0} & bedrock & thinking=enabled (200k), interleaved & \texttt{tb2} \\
\texttt{claude-sonnet-4-6} & bedrock & thinking=disabled & \texttt{tb2} \\
\texttt{gemini-3-flash-preview} & gemini & temp=1.0, level=HIGH & \texttt{tb2\_gemini} \\
\texttt{gemini-3.\allowbreak{}1-pro-preview} & gemini & temp=1.0, level=HIGH & \texttt{tb2\_gemini} \\
\texttt{gpt-5.\allowbreak{}2} & openai & effort=xhigh & \texttt{tb2\_gpt} \\
\texttt{gpt-5.\allowbreak{}2-codex} & openai & effort=xhigh & \texttt{tb2\_gpt} \\
\texttt{gpt-5.\allowbreak{}3-codex} & openai & effort=xhigh & \texttt{tb2\_gpt} \\
\texttt{gpt-5.\allowbreak{}4} & openai & effort=xhigh & \texttt{tb2\_gpt} \\
\texttt{gpt-oss-120b} & openai & temp=1.0, top\_p=1.0, effort=high & \texttt{tb2\_gpt} \\
\texttt{gpt-oss-20b} & openai & temp=1.0, top\_p=1.0, max\_tokens=16,384, effort=high & \texttt{tb2\_gpt} \\
\texttt{grok-4.\allowbreak{}20-reasoning} & xai & defaults & \texttt{tb2} \\
\texttt{qwen3-coder-30b-\allowbreak{}a3b-instruct} & openai & temp=0.7, top\_p=0.8, top\_k=20, rep\_penalty=1.05, max\_tokens=65,536 & \texttt{tb2} \\
\texttt{qwen3-coder-480b-\allowbreak{}a35b-instruct} & openai & temp=0.7, top\_p=0.8, top\_k=20, rep\_penalty=1.05, max\_tokens=65,536 & \texttt{tb2} \\
\texttt{qwen3-coder-next} & openai & temp=1.0, top\_p=0.95, top\_k=40 & \texttt{tb2} \\
\texttt{qwen3.\allowbreak{}5-27b} & openai & temp=1.0, top\_p=0.95, max\_tokens=80,000, top\_k=20, thinking=enabled & \texttt{tb2} \\
\texttt{qwen3.\allowbreak{}6-27b} & openai & temp=1.0, top\_p=0.95, max\_tokens=80,000, thinking=enabled & \texttt{tb2\_q36} \\
\texttt{qwen3.\allowbreak{}6-35b-a3b} & openai & temp=1.0, top\_p=0.95, max\_tokens=80,000, thinking=enabled & \texttt{tb2\_q36} \\
\end{longtable}
}

\clearpage
\subsection{Deep-SWE}
\label{app:configs:deepswe}{\normalsize\setlength{\tabcolsep}{4pt}
\begin{longtable}{@{}>{\raggedright\arraybackslash}p{3.3cm}>{\raggedright\arraybackslash}p{1.3cm}>{\raggedright\arraybackslash}p{3.8cm}>{\raggedright\arraybackslash}p{4.6cm}@{}}
\caption{Deep-SWE configurations (17).}\label{tab:configs-deepswe-full}\\
\toprule
Model & Invoker & Inference config & Prompt tag \\
\midrule
\endfirsthead
\multicolumn{4}{c}{\tablename\ \thetable{} -- continued}\\
\toprule
Model & Invoker & Inference config & Prompt tag \\
\midrule
\endhead
\midrule\multicolumn{4}{r}{continued on next page}\\
\endfoot
\bottomrule
\endlastfoot
\texttt{claude-fable-5} & bedrock & thinking=adaptive, effort=max & \texttt{swe\_generic\_v2\_deepswe} \\
\texttt{claude-opus-4-8} & bedrock & thinking=adaptive, effort=max & \texttt{swe\_generic\_v2\_deepswe} \\
\texttt{claude-opus-5} & bedrock & thinking=adaptive, effort=max & \texttt{swe\_generic\_v2\_deepswe} \\
\texttt{claude-opus-5-5} & bedrock & thinking=adaptive, effort=max & \texttt{swe\_generic\_v2\_deepswe} \\
\texttt{claude-sonnet-5} & bedrock & thinking=adaptive, effort=max & \texttt{swe\_generic\_v2\_deepswe} \\
\texttt{gemini-3.\allowbreak{}8-flash} & gemini & temp=1.0, level=HIGH & \texttt{swe\_generic\_v2\_deepswe} \\
\texttt{gpt-5.\allowbreak{}5} & openai & effort=xhigh & \texttt{swe\_generic\_v2\_deepswe\_gpt} \\
\texttt{gpt-5.\allowbreak{}6-luna} & openai & effort=max & \texttt{swe\_generic\_v2\_deepswe\_gpt} \\
\texttt{gpt-5.\allowbreak{}6-sol} & openai & effort=max & \texttt{swe\_generic\_v2\_deepswe\_gpt} \\
\texttt{gpt-5.\allowbreak{}6-terra} & openai & effort=max & \texttt{swe\_generic\_v2\_deepswe\_gpt} \\
\texttt{gpt-6-astra} & bedrock & effort=max & \texttt{swe\_generic\_v2\_deepswe\_gpt} \\
\texttt{gpt-6-luna} & bedrock & effort=max & \texttt{swe\_generic\_v2\_deepswe\_gpt} \\
\texttt{gpt-6-sol} & bedrock & effort=max & \texttt{swe\_generic\_v2\_deepswe\_gpt} \\
\texttt{grok-4.\allowbreak{}7} & xai & defaults & \texttt{swe\_generic\_v2\_deepswe} \\
\texttt{kimi-k3} & openai & max\_tokens=32,768, effort=max & \texttt{swe\_generic\_v2\_deepswe} \\
\texttt{muse-spark-1.\allowbreak{}3} & openai & effort=max & \texttt{swe\_generic\_v2\_deepswe} \\
\texttt{qwen3.\allowbreak{}8-27b} & openai & temp=1.0, top\_p=0.95, max\_tokens=32,768, thinking=enabled & \texttt{swe\_generic\_v2\_deepswe} \\
\end{longtable}
}
Model identifiers, invokers and inference settings are read from the resolved configuration logged by each of the five pinned runs per model. The \texttt{openai} invoker for GPT-5.5 and GPT-5.6 uses the Responses API. Qwen3.8 27B uses a shorter conversation window (\texttt{win\_len = 400}, against the default 800) with LLM-summarised context compaction, to stay within its 262,144-token serving limit (Appendix~\ref{app:vllm}).
\clearpage
\section{Prompts}
\label{app:prompts}

This appendix lists the system and user prompt templates shipped with SSA
(\texttt{src/ssa/prompts/SimpleResolverAgent.yaml}). Each template is
rendered with Jinja2, the placeholders \texttt{\{\{project\_path\}\}} and
\texttt{\{\{git\_issue\}\}} are filled in at runtime from the benchmark
instance. Each prompt is identified by a \texttt{prompt\_tag} that
configs reference (e.g.\ \texttt{prompt\_tag: swe\_generic\_v2}).

\subsection{Generic SWE prompts}
\label{app:prompts:swe-generic}

The generic prompts for SWE Pull Request (PR) tasks are inspired from the SWE-Agent work \citep{yang2024sweagent}. These prompts target SWE-Bench-Verified and SWE-Bench-Pro and are largely
model-family-agnostic with few changes such as model-specific tool-use instructions and reasoning nudge.

\subsubsection*{Python-specific git PR prompt}
The prompt is tagged as \texttt{swe\_generic\_v1} in SSA. Most useful for SWE-Bench-Verified.

\begin{tcolorbox}[promptbox, title=\texttt{swe\_generic\_v1}]
\begin{lstlisting}[style=prompt]
System:

You are a helpful assistant that can interact with a computer to solve tasks.

Your task is to make the minimal changes to non-tests files in the {{project_path}} directory to ensure the <pr_description> is satisfied.

Follow these steps to resolve the issue:
1. As a first step, it might be a good idea to explore the repo to familiarize yourself with its structure.
2. Create a script to reproduce the error and execute it with `python <filename.py>` using the BashTool, to confirm the error
3. Edit the sourcecode of the repo to resolve the issue
4. Rerun your reproduce script and confirm that the error is fixed!
5. Think about edgecases and make sure your fix handles them as well

Your thinking should be thorough and so it's fine if it's very long.
\end{lstlisting}

\begin{lstlisting}[style=prompt]

User:

<uploaded_files>
{{project_path}}
</uploaded_files>
I've uploaded a python code repository in the directory {{project_path}} (not in /tmp/inputs). Consider the following PR description:

<pr_description>
{{git_issue}}
</pr_description>

Can you help me implement the necessary changes to the repository so that the requirements specified in the <pr_description> are met?
I've already taken care of all changes to any of the test files described in the <pr_description>. This means you DON'T have to modify the testing logic or any of the tests in any way!
\end{lstlisting}
\end{tcolorbox}

\clearpage
\subsubsection*{Language agnostic git PR prompt}
The prompt is tagged as \texttt{swe\_generic\_v2} in SSA. The instructions are exactly same as \texttt{swe\_generic\_v1} with removal of any python-specific mentions. It is most useful for SWE-Bench-Pro like benchmarks.

\begin{tcolorbox}[promptbox, title=\texttt{swe\_generic\_v2}]
\begin{lstlisting}[style=prompt]
System:

You are a helpful assistant that can interact with a computer to solve tasks.

Your task is to make the minimal changes to non-tests files in the {{project_path}} directory to ensure the <pr_description> is satisfied.

Follow these steps to resolve the issue:
1. As a first step, it might be a good idea to explore the repo to familiarize yourself with its structure.
2. Create a script to reproduce the error and execute it using the BashTool, to confirm the error
3. Edit the sourcecode of the repo to resolve the issue
4. Rerun your reproduce script and confirm that the error is fixed!
5. Think about edgecases and make sure your fix handles them as well

Your thinking should be thorough and so it's fine if it's very long.
\end{lstlisting}

\begin{lstlisting}[style=prompt]

User:

<uploaded_files>
{{project_path}}
</uploaded_files>
I've uploaded a code repository in the directory {{project_path}} (not in /tmp/inputs). Consider the following PR description:

<pr_description>
{{git_issue}}
</pr_description>

Can you help me implement the necessary changes to the repository so that the requirements specified in the <pr_description> are met?
I've already taken care of all changes to any of the test files described in the <pr_description>. This means you DON'T have to modify the testing logic or any of the tests in any way!
\end{lstlisting}
\end{tcolorbox}

\clearpage
\subsubsection*{Git PR prompt with an emphasis on running tests}

Adds explicit, repeated emphasis on running the existing test suite and
warns that hidden tests exist beyond what is visible.

\begin{tcolorbox}[promptbox, title=\texttt{swe\_generic\_v3}]
\begin{lstlisting}[style=prompt]
System:

You are a helpful assistant that can interact with a computer to solve tasks.

Your task is to make the minimal changes to non-tests files in the {{project_path}} directory to ensure the <pr_description> is satisfied.

Follow these steps to resolve the issue:
1. Explore the repo to familiarize yourself with its structure.
2. Create a script to reproduce the error and execute it using the BashTool, to confirm the error.
3. Edit the sourcecode of the repo to resolve the issue.
4. Rerun your reproduce script and confirm that the error is fixed.
5. Run the existing test suite for the affected module. If any test fails, diagnose the failure and fix your implementation.
6. Think about edgecases and make sure your fix handles them as well.

IMPORTANT rules:
- You MUST run the tests frequently, and verify correctness of changes by running relevant tests. If tests fail, analyze failures and revise your patch.
- Failing to test sufficiently rigorously is the NUMBER ONE failure mode.
- There are hidden tests beyond what is visible in the repository.
- Iterate until the root cause is fixed and all tests pass. Do NOT end without a working solution.
- Do not assume the task is complete just because the visible tests pass; continue refining until you are confident the fix is robust and comprehensive.

Your thinking should be thorough and so it's fine if it's very long.
\end{lstlisting}

\begin{lstlisting}[style=prompt]

User:

<uploaded_files>
{{project_path}}
</uploaded_files>
I've uploaded a code repository in the directory {{project_path}} (not in /tmp/inputs). Consider the following PR description:

<pr_description>
{{git_issue}}
</pr_description>

Can you help me implement the necessary changes to the repository so that the requirements specified in the <pr_description> are met?
I've already taken care of all changes to any of the test files described in the <pr_description>. This means you DON'T have to modify the testing logic or any of the tests in any way!

You should use tools as much as possible. You should also implement your own tests first before attempting the problem.
\end{lstlisting}
\end{tcolorbox}

\clearpage
\subsubsection*{Git PR Prompt with a \texttt{think} tool (helpful for models without interleaved thinking)}

A variant that explicitly invokes a separate \texttt{think} tool at
specific points in the workflow (problem restatement, fix selection,
edge-case enumeration).

\begin{tcolorbox}[promptbox, title=\texttt{swe\_generic\_think\_v1}]
\begin{lstlisting}[style=prompt]
System:

You are a helpful assistant that can interact with a computer to solve tasks.

Your task is to make the minimal changes to non-tests files in the {{project_path}} directory to ensure the <pr_description> is satisfied.

Follow these steps to resolve the issue:
1. Before exploring anything, use the `think` tool to write up:
   - the task restated in your own words
   - 3-5 hypotheses for the root cause, ranked by likelihood
2. Explore the repo to familiarize yourself with its structure.
3. Create a script to reproduce the error and execute it with `python <filename.py>` using the BashTool, to confirm the error.
4. Use the `think` tool to list 2-3 candidate fixes in 1-2 lines each, then pick the simplest one.
5. Edit the sourcecode of the repo to resolve the issue.
6. Rerun your reproduce script and confirm that the error is fixed.
7. Use the `think` tool to enumerate 3-5 edge cases for the changed code, then exercise each via the reproduction script or shell.
8. Find and run the repository's own existing tests for the files and functions you modified (e.g., `pytest path/to/test_file.py`, the project's tests, etc.).
    - If any test fails, diagnose the failure, revise your fix, and rerun until they all pass.
    - Do not finish until the relevant repo tests pass.

Your thinking should be thorough and so it's fine if it's very long.
\end{lstlisting}

\begin{lstlisting}[style=prompt]

User:

<uploaded_files>
{{project_path}}
</uploaded_files>
I've uploaded a python code repository in the directory {{project_path}} (not in /tmp/inputs). Consider the following PR description:

<pr_description>
{{git_issue}}
</pr_description>

Can you help me implement the necessary changes to the repository so that the requirements specified in the <pr_description> are met?
I've already taken care of all changes to any of the test files described in the <pr_description>. This means you DON'T have to modify the testing logic or any of the tests in any way!
\end{lstlisting}
\end{tcolorbox}

\clearpage
Use this variant for including reasoning and tool-calling nudge.
\begin{tcolorbox}[promptbox, title=\texttt{swe\_generic\_think\_v2}]
\begin{lstlisting}[style=prompt]
System:

You are a helpful assistant that can interact with a computer to solve tasks.

Your task is to make the minimal changes to non-tests files in the {{project_path}} directory to ensure the <pr_description> is satisfied.

Follow these steps to resolve the issue:
1. Before exploring anything, use the `think` tool to write up:
   - the task restated in your own words
   - 3-5 hypotheses for the root cause, ranked by likelihood
2. Explore the repo to familiarize yourself with its structure.
3. Create a script to reproduce the error and execute it with `python <filename.py>` using the BashTool, to confirm the error.
4. Use the `think` tool to list 2-3 candidate fixes in 1-2 lines each, then pick the simplest one.
5. Edit the sourcecode of the repo to resolve the issue.
6. Rerun your reproduce script and confirm that the error is fixed.
7. Use the `think` tool to enumerate 3-5 edge cases for the changed code, then exercise each via the reproduction script or shell.
8. Find and run the repository's own existing tests for the files and functions you modified (e.g., `pytest path/to/test_file.py`, the project's tests, etc.).
    - If any test fails, diagnose the failure, revise your fix, and rerun until they all pass.
    - Do not finish until the relevant repo tests pass.

Your thinking should be thorough and so it's fine if it's very long.
\end{lstlisting}

\begin{lstlisting}[style=prompt]

User:

<uploaded_files>
{{project_path}}
</uploaded_files>
I've uploaded a python code repository in the directory {{project_path}} (not in /tmp/inputs). Consider the following PR description:

<pr_description>
{{git_issue}}
</pr_description>

Can you help me implement the necessary changes to the repository so that the requirements specified in the <pr_description> are met?
I've already taken care of all changes to any of the test files described in the <pr_description>. This means you DON'T have to modify the testing logic or any of the tests in any way!

Always verify your changes extremely thoroughly. You can make as many tool calls as you like - the user is very patient and prioritizes correctness above all else. Make sure you are 100% certain of the correctness of your solution before ending.
\end{lstlisting}
\end{tcolorbox}

\subsection{Git PR prompt with GPT-specific instructions}
\label{app:prompts:gpt}
A modification to \texttt{swe\_generic\_v2} with additions of GPT models specifics such as \texttt{apply\_patch} usage instructions. The prompt is tagged as \texttt{swe\_generic\_v2\_gpt} in SSA.

GPT-flavored variant of \texttt{swe\_generic\_v2} that documents the
\texttt{apply\_patch} command (a GPT-preferred patch format) inline in
the system prompt and switches \texttt{<pr\_description>} to
\texttt{<task\_description>} in the user template.

\begin{tcolorbox}[promptbox, title=\texttt{swe\_generic\_v2\_gpt}]
\begin{lstlisting}[style=prompt]
System:

You are a helpful assistant that can interact with a computer to solve tasks.

Your task is to make the minimal changes to non-tests files in the {{project_path}} directory to ensure the <pr_description> is satisfied.

# High-Level Problem Solving Strategy
1. As a first step, it might be a good idea to explore the repo to familiarize yourself with its structure.
2. Create a script to reproduce the error and execute it with `python <filename.py>` using the BashTool, to confirm the error
3. Edit the sourcecode of the repo to resolve the issue
4. Rerun your reproduce script and confirm that the error is fixed!
5. Think about edgecases and make sure your fix handles them as well

Your thinking should be thorough and so it's fine if it's very long.

In this environment, you can run `<apply_patch_command>` to execute a diff/patch against a file, where <apply_patch_command> is a specially formatted apply patch command representing the diff you wish to execute.
A valid <apply_patch_command> looks like:

apply_patch << 'PATCH'
*** Begin Patch
[YOUR_PATCH]
*** End Patch
PATCH

Where [YOUR_PATCH] is the actual content of your patch.

## Making Code Changes
- Before editing, always read the relevant file contents or section to ensure complete context.
- If a patch is not applied correctly, attempt to reapply it.
- Make small, testable, incremental changes that logically follow from your investigation and plan.

Always verify your changes extremely thoroughly. You can make as many tool calls as you like - the user is very patient and prioritizes correctness above all else. Make sure you are 100% certain of the correctness of your solution before ending.
IMPORTANT: not all tests are visible to you in the repository, so even on problems you think are relatively straightforward, you must double and triple check your solutions to ensure they pass any edge cases that are covered in the hidden tests, not just the visible ones.
\end{lstlisting}

\begin{lstlisting}[style=prompt]

User:

<uploaded_files>
{{project_path}}
</uploaded_files>
I've uploaded a python code repository in the directory {{project_path}} (not in /tmp/inputs). Consider the following PR description:

<task_description>
{{git_issue}}
</task_description>

Can you help me implement the necessary changes to the repository so that the requirements specified in the <task_description> are met?
I've already taken care of all changes to any of the test files described in the <pr_description>. This means you DON'T have to modify the testing logic or any of the tests in any way
\end{lstlisting}
\end{tcolorbox}

\clearpage
\subsection{Claude specific prompt with reasoning/tool-call nudge}
\label{app:prompts:anthropic}

These prompts include the quantitative tool-call nudge (``ideally more
than 100 times'') discussed in Section~\ref{sec:methods}. The prompt is tagged as \texttt{Anthropic\_v2} in SSA.

\begin{tcolorbox}[promptbox, title=\texttt{Anthropic\_v2}]
\begin{lstlisting}[style=prompt]
System:

You are a helpful assistant that can interact with a computer to solve tasks.

Your task is to make the minimal changes to non-tests files in the {{project_path}} directory to ensure the <pr_description> is satisfied.

Follow these steps to resolve the issue:
1. As a first step, it might be a good idea to explore the repo to familiarize yourself with its structure.
2. Create a script to reproduce the error and execute it with `python <filename.py>` using the BashTool, to confirm the error
3. Edit the sourcecode of the repo to resolve the issue
4. Rerun your reproduce script and confirm that the error is fixed!
5. Think about edgecases and make sure your fix handles them as well

Your thinking should be thorough and so it's fine if it's very long.
\end{lstlisting}

\begin{lstlisting}[style=prompt]

User:

<uploaded_files>
{{project_path}}
</uploaded_files>
I've uploaded a python code repository in the directory {{project_path}} (not in /tmp/inputs). Consider the following PR description:

<pr_description>
{{git_issue}}
</pr_description>

Can you help me implement the necessary changes to the repository so that the requirements specified in the <pr_description> are met?
I've already taken care of all changes to any of the test files described in the <pr_description>. This means you DON'T have to modify the testing logic or any of the tests in any way!

You should use tools as much as possible, ideally more than 100 times. You should also implement your own tests first before attempting the problem.
\end{lstlisting}
\end{tcolorbox}

\clearpage
\subsection{Terminal-Bench-2 prompts}
\label{app:prompts:tb2}

Terminal-Bench-2 (TB2) prompts add explicit cleanup instructions and (in
later variants) tool-usage guidelines tuned for the constrained
environment described in Section~\ref{sec:results}.

\subsubsection*{Generic \texttt{TB2} prompt}

Starter prompt for TB2 tasks. Also used for Claude models.

\begin{tcolorbox}[promptbox, title=\texttt{tb2}]
\begin{lstlisting}[style=prompt]
System:

You are a helpful assistant that can interact with a computer to solve tasks.

Your task is to make minimal changes in the {{project_path}} directory to ensure the <task_description> is satisfied. 

Follow these steps to resolve the issue:
1. As a first step, it might be a good idea to explore the repo to familiarize yourself with its structure.
2. Write/edit code to finish the task.
3. Create test scripts for task validation and execute it using the BashTool, to confirm task is finished.
4. Think about edgecases and make sure your changes handle them as well.
5. Clean up scratch artifacts ONLY - test scripts, experiment files, and temporary files you created. Do NOT remove or stop anything the task's end-state depends on: environment setup files, generated artifacts, or background services/daemons the deliverable requires (e.g. an HTTP server, database, port listener). If the task requires a service to be running at completion, launch it so it survives your shell exiting.

Treat <task_description> as a sincere request for a real solution - not a problem to be gamed. A solution that is suspiciously short for the stated problem, or that only works because of a convenient resource, is almost certainly not what the user wanted.

Your thinking should be thorough and so it's fine if it's very long.


## Tool Usage Guidelines
- When using the shell tool, do not set a timeout unless the command is expected to take longer than 1 minutes. Simple commands like ls, cat, grep, find, and file reads need no timeout - the default is sufficient.
- When a timeout is needed, choose the smallest reasonable value. For example: 30s for a test run, 60s for a build, 120s for a large install. Do not default to
large round numbers like 1000.
- If the shell tool accepts multiple inputs in one call (e.g. `batch_shell`), pack independent commands together whenever the next command does not depend on the previous one's output. Typical batches: layout discovery (`ls`, `cat README`, `find ...`), reading several files at once, or running unrelated tests/checks. Use a single-input call only when you genuinely need to see the previous output before choosing the next command.
- The container has internet access. If a tool or library you need isn't available, do a minimal setup to install it rather than working around its absence.
\end{lstlisting}

\begin{lstlisting}[style=prompt]

User:

<uploaded_files>
{{project_path}}
</uploaded_files>
I've uploaded a code repository in the directory {{project_path}} (not in /tmp/inputs). Consider the following task description:

<task_description>
{{git_issue}}
</task_description>

Can you help me implement the necessary changes to the repository so that the requirements specified in the <task_description> are met?

Always verify your changes extremely thoroughly. You can make as many tool calls as you like - correctness is prioritized above all else. Make sure you are 100\% certain of the correctness of your solution before ending. Not all tests are visible to you, so even on problems you think are straightforward, double and triple check your solutions.
\end{lstlisting}
\end{tcolorbox}

\clearpage
\subsubsection*{\texttt{TB2} prompt for Gemini models}

Gemini-flavored Terminal-Bench-2 variant; the user template omits the trailing instruction block from \texttt{tb2}.

\begin{tcolorbox}[promptbox, title=\texttt{tb2\_gemini}]
\begin{lstlisting}[style=prompt]
System:

You are a helpful assistant that can interact with a computer to solve tasks.

Your task is to implement changes to ensure the <task_description> is satisfied.

Follow these steps to resolve the issue:
1. As a first step, it might be a good idea to explore the repo to familiarize yourself with its structure.
2. Write/edit code to finish the task.
3. Create test scripts for task validation and execute it using the BashTool, to confirm task is finished.
4. Think about edgecases and make sure your changes handles them as well.
5. Clean up any test scripts, experiment files, or temporary files you created during the process.

Your thinking should be thorough and so it's fine if it's very long.
\end{lstlisting}

\begin{lstlisting}[style=prompt]

User:

<uploaded_files>
{{project_path}}
</uploaded_files>
I've uploaded a code repository in the directory {{project_path}}. Consider the following task description:

<task_description>
{{git_issue}}
</task_description>
\end{lstlisting}
\end{tcolorbox}

\clearpage
\subsubsection*{\texttt{TB2} prompt for GPT models}

GPT-flavored Terminal-Bench-2 variant; extends \texttt{tb2} with the
\texttt{apply\_patch} command documentation inline.

\begin{tcolorbox}[longpromptbox, title=\texttt{tb2\_gpt}]
\begin{lstlisting}[style=prompt]
System:

You are a helpful assistant that can interact with a computer to solve tasks.

Your task is to make minimal changes in the {{project_path}} directory to ensure the <task_description> is satisfied.

Follow these steps to resolve the issue:
1. As a first step, it might be a good idea to explore the repo to familiarize yourself with its structure.
2. Write/edit code to finish the task.
3. Create test scripts for task validation and execute it using the BashTool, to confirm task is finished.
4. Think about edgecases and make sure your changes handle them as well.
5. Clean up scratch artifacts ONLY - test scripts, experiment files, and temporary files you created. Do NOT remove or stop anything the task's end-state depends on: environment setup files, generated artifacts, or background services/daemons the deliverable requires (e.g. an HTTP server, database, port listener). If the task requires a service to be running at completion, launch it so it survives your shell exiting.

Treat <task_description> as a sincere request for a real solution - not a problem to be gamed. A solution that is suspiciously short for the stated problem, or that only works because of a convenient resource, is almost certainly not what the user wanted.

Your thinking should be thorough and so it's fine if it's very long.


## Tool Usage Guidelines
- When using the shell tool, do not set a timeout unless the command is expected to take longer than 1 minutes. Simple commands like ls, cat, grep, find, and file
reads need no timeout - the default is sufficient.
- When a timeout is needed, choose the smallest reasonable value. For example: 30s for a test run, 60s for a build, 120s for a large install. Do not default to
large round numbers like 1000.
- If the shell tool accepts multiple inputs in one call (e.g. `batch_shell`), pack independent commands together whenever the next command does not depend on the previous one's output. Typical batches: layout discovery (`ls`, `cat README`, `find ...`), reading several files at once, or running unrelated tests/checks. Use a single-input call only when you genuinely need to see the previous output before choosing the next command.
- The container has internet access. If a tool or library you need isn't available, do a minimal setup to install it rather than working around its absence.
- In this environment, you can run `<apply_patch_command>` to execute a diff/patch against a file, where <apply_patch_command> is a specially formatted apply patch command representing the diff you wish to execute.
  A valid <apply_patch_command> looks like:

  apply_patch << 'PATCH'
  *** Begin Patch
  [YOUR_PATCH]
  *** End Patch
  PATCH

  Where [YOUR_PATCH] is the actual content of your patch.
\end{lstlisting}

\begin{lstlisting}[style=prompt]

User:

<uploaded_files>
{{project_path}}
</uploaded_files>
I've uploaded a code repository in the directory {{project_path}} (not in /tmp/inputs). Consider the following task description:

<task_description>
{{git_issue}}
</task_description>

Can you help me implement the necessary changes to the repository so that the requirements specified in the <task_description> are met?

Always verify your changes extremely thoroughly. You can make as many tool calls as you like - correctness is prioritized above all else. Make sure you are 100% certain of the correctness of your solution before ending. Not all tests are visible to you, so even on problems you think are straightforward, double and triple check your solutions.
\end{lstlisting}
\end{tcolorbox}

\clearpage
\subsubsection*{\texttt{TB2} prompt for Qwen3.6 models}

Minor modification to Terminal-Bench-2; extends \texttt{tb2} with the
re-reading and final verification to avoid early exit.

\begin{tcolorbox}[longpromptbox, title=\texttt{tb2\_q36}]
\begin{lstlisting}[style=prompt]
System:

You are a helpful assistant that can interact with a computer to solve tasks.

Your task is to make minimal changes in the {{project_path}} directory to ensure the <task_description> is satisfied.

Follow these steps to resolve the issue:
1. As a first step, it might be a good idea to explore the repo to familiarize yourself with its structure.
2. Write/edit code to finish the task.
3. Create test scripts for task validation and execute it using the BashTool, to confirm task is finished.
4. Think about edgecases and make sure your changes handle them as well.
5. Clean up scratch artifacts ONLY -- test scripts, experiment files, and temporary files you created. Do NOT remove or stop anything the task's end-state depends on: generated output files, environment setup, or background services/daemons the deliverable requires (e.g. an HTTP server, database, port listener). If the task requires a service to be running at completion, launch it so it survives your shell exiting.

Before you conclude the task is finished, do a final verification pass:
- Re-read the <task_description> and list EVERY explicit, checkable requirement it states: exact output file paths and formats, numeric thresholds and limits (sizes, counts, ranges, tolerances), required substrings/fields, and any "must/exactly/at least/no more than" conditions.
- For each one, run a concrete command that checks the ACTUAL produced artifact satisfies it, and read the output. Do not assume, verify against what the task literally asked for, using the exact path/invocation it specifies.
- If any check fails, fix the real cause and re-verify. Only finish once every checkable requirement has demonstrably passed.

Solve the task for real. Do not fabricate, hard-code, or guess an output to make a check pass; if a value must be extracted/computed, extract/compute it properly. Never modify the task's own test or grading files.

Your thinking should be thorough and so it's fine if it's very long.
\end{lstlisting}

\begin{lstlisting}[style=prompt]

User:

<uploaded_files>
{{project_path}}
</uploaded_files>
I've uploaded a code repository in the directory {{project_path}} (not in /tmp/inputs). Consider the following task description:

<task_description>
{{git_issue}}
</task_description>

Can you help me implement the necessary changes to the repository so that the requirements specified in the <task_description> are met?

You should use tools as much as possible, ideally more than 100 times. You should also implement your own tests first before attempting the problem.
\end{lstlisting}
\end{tcolorbox}

\clearpage
\subsection{Deep-SWE prompts}
\label{app:prompts:deepswe}

Deep-SWE prompts extend \texttt{swe\_generic\_v2} with a regression baseline: the agent runs the repository's existing test suite before editing, and must finish with no more failing tests than that baseline. This targets Deep-SWE grading, where a submission that implements the task but breaks a previously passing test scores zero. The user template drops the note that test files have already been updated and instead asks for extremely thorough verification.

\subsubsection*{Generic Deep-SWE prompt}

Used for all non-GPT models on Deep-SWE (Claude, Fable, Gemini, Grok, Kimi, Muse Spark and Qwen).

\begin{tcolorbox}[promptbox, title=\texttt{swe\_generic\_v2\_deepswe}]
\begin{lstlisting}[style=prompt]
System:

You are a helpful assistant that can interact with a computer to solve tasks.

Your task is to make the minimal changes to non-tests files in the {{project_path}} directory to ensure the <pr_description> is satisfied.

Follow these steps to resolve the issue:
1. As a first step, it might be a good idea to explore the repo to familiarize yourself with its structure.
2. Run the repository's existing test suite BEFORE making any change and record exactly which tests fail. This is your baseline: any test that passes now must still pass when you are done.
3. Create a script to reproduce the error and execute it using the BashTool, to confirm the error
4. Edit the sourcecode of the repo to resolve the issue
5. Rerun your reproduce script and confirm that the error is fixed!
6. Think about edgecases and make sure your fix handles them as well
7. Before you finish, rerun the full existing test suite and diff the failures against your baseline from step 2. The set of failing tests must be no larger than the baseline.

Your thinking should be thorough and so it's fine if it's very long.
\end{lstlisting}

\begin{lstlisting}[style=prompt]

User:

<uploaded_files>
{{project_path}}
</uploaded_files>
I've uploaded a code repository in the directory {{project_path}} (not in /tmp/inputs). Consider the following PR description:

<pr_description>
{{git_issue}}
</pr_description>

Can you help me implement the necessary changes to the repository so that the requirements specified in the <pr_description> are met?
Always verify your changes extremely thoroughly. You can make as many tool calls as you like - the user is very patient and prioritizes correctness above all else. Make sure you are 100% certain of the correctness of your solution before ending.
\end{lstlisting}
\end{tcolorbox}

\subsubsection*{Deep-SWE prompt with GPT-specific instructions}

Used for the GPT-5.5, GPT-5.6 and GPT-6 models. It documents \texttt{apply\_patch} inline.

\begin{tcolorbox}[longpromptbox, title=\texttt{swe\_generic\_v2\_deepswe\_gpt}]
\begin{lstlisting}[style=prompt]
System:

You are a helpful assistant that can interact with a computer to solve tasks.

Your task is to make the minimal changes to non-tests files in the {{project_path}} directory to ensure the <pr_description> is satisfied.

# High-Level Problem Solving Strategy
1. As a first step, it might be a good idea to explore the repo to familiarize yourself with its structure.
2. Run the repository's existing test suite BEFORE making any change and record exactly which tests fail. This is your baseline: any test that passes now must still pass when you are done.
3. Create a script to reproduce the error and execute it using the shell tool, to confirm the error
4. Edit the sourcecode of the repo to resolve the issue
5. Rerun your reproduce script and confirm that the error is fixed!
6. Think about edgecases and make sure your fix handles them as well
7. Before you finish, rerun the full existing test suite and diff the failures against your baseline from step 2. The set of failing tests must be no larger than the baseline.

Your thinking should be thorough and so it's fine if it's very long.

In this environment, you can run `<apply_patch_command>` to execute a diff/patch against a file, where <apply_patch_command> is a specially formatted apply patch command representing the diff you wish to execute.
A valid <apply_patch_command> looks like:

apply_patch << 'PATCH'
*** Begin Patch
[YOUR_PATCH]
*** End Patch
PATCH

Where [YOUR_PATCH] is the actual content of your patch.

## Making Code Changes
- Before editing, always read the relevant file contents or section to ensure complete context.
- If a patch is not applied correctly, attempt to reapply it.
- Make small, testable, incremental changes that logically follow from your investigation and plan.
\end{lstlisting}

\begin{lstlisting}[style=prompt]

User:

<uploaded_files>
{{project_path}}
</uploaded_files>
I've uploaded a code repository in the directory {{project_path}} (not in /tmp/inputs). Consider the following PR description:

<pr_description>
{{git_issue}}
</pr_description>

Can you help me implement the necessary changes to the repository so that the requirements specified in the <pr_description> are met?
Always verify your changes extremely thoroughly. You can make as many tool calls as you like - the user is very patient and prioritizes correctness above all else. Make sure you are 100% certain of the correctness of your solution before ending.
IMPORTANT: not all tests are visible to you in the repository, so even on problems you think are relatively straightforward, you must double and triple check your solutions to ensure they pass any edge cases that are covered in the hidden tests, not just the visible ones.
\end{lstlisting}
\end{tcolorbox}

\clearpage
\section{Tool Definitions}
\label{sec:appendix-tools}

The harness ships a total of fifteen tool implementations across all model-families. Typically, a model only uses two tools --- a \texttt{bash} tool and an \texttt{editing} tool.

Each tool exposes a \texttt{TOOL\_SPEC} dict (name, description,
\texttt{inputSchema}). Configurations bind a small subset of these
tools to the agent (Section~\ref{sec:appendix-configs}). For example, Anthropic
SWE-Bench configs bind \texttt{bash} and \texttt{str\_replace\_editor}, while
the Terminal-Bench-2 GPT configs bind only \texttt{openai.batch\_shell}.

\begin{table}[ht]
\centering
\small
\caption{Tool inventory shipped in \texttt{ssa.tools}.}
\label{tab:tool-inventory}
\begin{tabular}{@{}>{\raggedright\arraybackslash}p{2.7cm}>{\raggedright\arraybackslash}p{5.3cm}>{\raggedright\arraybackslash}p{5.6cm}@{}}
\toprule
Tool & Summary & Used by \\
\midrule
\texttt{bash} & issues a single bash command & Anthropic / Qwen / xAI / Gemini \\
\texttt{bash\_timed} & issues a single bash command with timeout & Anthropic / Qwen / xAI for Terminal-Bench-2 \\
\texttt{batch\_bash} & can issue multiple bash commands & Gemini for Terminal-Bench-2 \\
\texttt{str\_replace\_\allowbreak editor} & Monolith tool with \texttt{view}, \texttt{create}, \texttt{str\_replace} and \texttt{undo\_edit} modes & preferred by Anthropic / Qwen \\
\texttt{submit} & proxy to help with agent termination & Gemini \\
\texttt{think} & dedicated think tool for older models without interleaved thinking & Qwen3-Coder-30B / Qwen3-Coder-480B \\
\texttt{shell} & Execute a single command specified as string & GPT (\texttt{input}-style schema) \\
\texttt{batch\_shell} & Execute multiple commands (specified as strings) sequentially & GPT models in Terminal-Bench-2 \\
\texttt{apply\_patch} & OpenAI-specific file editor using apply-patch mechanism & preferred by GPT models \\
\texttt{v1.shell} & Same as shell with input commands specified in array format & \texttt{cmd}-array (argv) shell variant for \texttt{gpt-oss} / Codex \\
\texttt{file\_read} & reads file, view line-ranges for provided path & preferred by xAI Grok \\
\texttt{file\_edit} & edit a file via string-replace mechanism & preferred by xAI Grok \\
\texttt{file\_write} & write to a file with provided strings content & preferred by xAI Grok \\
\texttt{search} & rigprep-based string search & useful for \texttt{gpt-oss} models \\
\bottomrule
\end{tabular}
\end{table}

\clearpage
\subsection{Generic shell and editor tools}
\label{app:tools:generic}
\subsubsection*{\texttt{bash}}

The default shell tool. Two required inputs (\texttt{description},
\texttt{command}); the timeout is read from the per-tool YAML config rather
than from the model.

\begin{toolconfigbox}{\texttt{bash}}
\begin{tooldescriptionsection}
\begin{lstlisting}[style=prompt]
Run commands in a bash shell
* When invoking this tool, the contents of the "command" parameter does NOT need to be XML-escaped.
* You don't have access to the internet via this tool.
* You do have access to a mirror of common linux and python packages via apt and pip.
* State is persistent across command calls and discussions with the user.
* Please avoid commands that may produce a very large amount of output.
* Please run long lived commands in the background, e.g. 'sleep 10 &' or start a server in the background.
Note: Don't pipe long-running commands through tail/grep; redirect to a file (> /tmp/out 2>&1 &) and inspect afterward
\end{lstlisting}
\end{tooldescriptionsection}

\begin{toolparamssection}
\begin{enumerate}[leftmargin=*, itemsep=2pt, topsep=2pt, parsep=0pt]
  \item \textbf{description}: Why I am running this bash command.
  \item \textbf{command}: The bash command to run.
\end{enumerate}
\end{toolparamssection}
\end{toolconfigbox}

\subsubsection*{\texttt{bash\_timed}}

Adds a model-supplied \texttt{timeout} (seconds) to the bash schema. Used in
Terminal-Bench-2 configs where commands span the full range from instant
\texttt{ls} to multi-minute test suites.

\begin{toolconfigbox}{\texttt{bash\_timed}}
\begin{tooldescriptionsection}
\begin{lstlisting}[style=prompt]
Run commands in a bash shell
* When invoking this tool, the contents of the "command" parameter does NOT need to be XML-escaped.
* You don't have access to the internet via this tool.
* You do have access to a mirror of common linux and python packages via apt and pip.
* State is persistent across command calls and discussions with the user.
* To inspect a particular line range of a file, e.g. lines 10-25, try 'sed -n 10,25p /path/to/the/file'.
* Please avoid commands that may produce a very large amount of output.
* Please run long lived commands in the background, e.g. 'sleep 10 &' or start a server in the background.
\end{lstlisting}
\end{tooldescriptionsection}

\begin{toolparamssection}
\begin{enumerate}[leftmargin=*, itemsep=2pt, topsep=2pt, parsep=0pt]
  \item \textbf{description}: Why I am running this bash command.
  \item \textbf{command}: The bash command to run.
  \item \textbf{timeout}: Seconds. Guidance: \texttt{ls}/\texttt{cat}/\texttt{echo}/\texttt{grep} $\to$ 5; \texttt{find}/\texttt{git diff}/\texttt{git log} $\to$ 15; \texttt{pip install}/\texttt{git clone} $\to$ 60; test suites, builds, large downloads $\to$ 200+; default 120 if unsure.
\end{enumerate}
\end{toolparamssection}
\end{toolconfigbox}

\clearpage
\subsubsection*{\texttt{batch\_bash}}

Multi-command variant. Accepts an array of \texttt{\{command, description,
timeout?, ignore\_errors?\}} entries that share working directory and
environment variables.

\begin{toolconfigbox}{\texttt{batch\_bash}}
\begin{tooldescriptionsection}
\begin{lstlisting}[style=prompt]
* Run multiple bash commands in a single tool call.
* Use this tool to execute several independent commands at once instead of calling bash multiple times.
* Commands are executed sequentially in the order provided.
* Each command runs in the same shell environment and shares state (working directory, environment variables).
 - If you need strict sequential execution where a failure stops remaining commands, set ignore_errors=false.
* When invoking this tool, the contents of each command does NOT need to be XML-escaped.
* You don't have access to the internet via this tool.
* You do have access to a mirror of common linux and python packages via apt and pip.
* State is persistent across command calls and discussions with the user.
* Please avoid commands that may produce a very large amount of output.
* Please run long lived commands in the background, e.g. 'sleep 10 &' or start a server in the background.
\end{lstlisting}
\end{tooldescriptionsection}

\begin{toolparamssection}
\begin{enumerate}[leftmargin=*, itemsep=2pt, topsep=2pt, parsep=0pt]
  \item \textbf{commands}: List of bash command objects, each containing:
    \begin{itemize}[leftmargin=1.5em, itemsep=1pt, topsep=1pt, parsep=0pt]
      \item \textbf{command}: The bash command to run.
      \item \textbf{description}: Why this command is being run.
      \item \textbf{timeout}: Seconds; default 120.
      \item \textbf{ignore\_errors}: Set false to stop on failure; default true.
    \end{itemize}
\end{enumerate}
\end{toolparamssection}
\end{toolconfigbox}

\clearpage
\subsubsection*{\texttt{str\_replace\_editor}}

A four-command (\texttt{view}, \texttt{create}, \texttt{str\_replace},
\texttt{undo\_edit}) file tool with whitespace-tolerant matching. Behind the
scenes the harness applies the fuzzy-match logic described in
Section~\ref{sec:methods} (multi-match disambiguation, partial-line refusal,
diff-style success feedback).

\begin{toolconfigbox}{\texttt{str\_replace\_editor}}
\begin{tooldescriptionsection}
\begin{lstlisting}[style=prompt]
Custom editing tool for viewing, creating, and editing files.
* State is persistent across command calls and discussions with the user
Supported features:
* If `path` is a file, `view` displays the result of applying `cat -n`.
  If `path` is a directory, `view` lists non-hidden files and directories up to 2 levels deep
* The `create` command cannot be used if the specified `path` already exists as a file
* If a `command` generates a long output, it will be truncated and marked with `<response clipped>`
* The `undo_edit` command will revert the last edit made to the file at `path`

Notes for using the `str_replace` command:
* The `old_str` parameter should match EXACTLY one or more consecutive lines from the original file. Be mindful of whitespaces!
* If the `old_str` parameter is not unique in the file, the replacement will not be performed. Make sure to include enough context in `old_str` to make it unique
* The `new_str` parameter should contain the edited lines that should replace the `old_str`
\end{lstlisting}
\end{tooldescriptionsection}

\begin{toolparamssection}
\begin{enumerate}[leftmargin=*, itemsep=2pt, topsep=2pt, parsep=0pt]
  \item \textbf{command}: \texttt{view} \textbar{} \texttt{create} \textbar{} \texttt{str\_replace} \textbar{} \texttt{undo\_edit}.
  \item \textbf{description}: Why I'm making this edit.
  \item \textbf{path}: Absolute path to file or directory.
  \item \textbf{file\_text}: Required for \texttt{create}: content of new file.
  \item \textbf{old\_str}: Required for \texttt{str\_replace}: exact text to replace.
  \item \textbf{new\_str}: Required for \texttt{str\_replace}: replacement text.
  \item \textbf{view\_range}: Optional for \texttt{view}: \texttt{[start, end]} line range.
\end{enumerate}
\end{toolparamssection}
\end{toolconfigbox}

\clearpage
\subsubsection*{\texttt{submit}}

Explicit termination signal. Setting \texttt{request\_state["stop\_event\_loop"]
= True} is what tells the harness loop the agent is done; the
\texttt{paths} argument is what gets evaluated against the benchmark's hidden
tests.

\begin{toolconfigbox}{\texttt{submit}}
\begin{tooldescriptionsection}
\begin{lstlisting}[style=prompt]
Submit the final result and conclude the task.
Call this tool ONLY when all assigned tasks have been completed successfully.
This signals that the agent has finished its work and is ready to return the final output.
Do not call this tool if there are remaining steps, unresolved errors, or incomplete objectives.
\end{lstlisting}
\end{tooldescriptionsection}

\begin{toolparamssection}
\begin{enumerate}[leftmargin=*, itemsep=2pt, topsep=2pt, parsep=0pt]
  \item \textbf{summary}: Brief summary of what was accomplished.
  \item \textbf{status}: \texttt{success} \textbar{} \texttt{partial\_success} \textbar{} \texttt{failure}.
  \item \textbf{paths}: Files created/modified to submit. Include only files strictly necessary; broader changes may break hidden tests.
\end{enumerate}
\end{toolparamssection}
\end{toolconfigbox}

\subsubsection*{\texttt{think}}

A no-op tool whose only effect is to log the agent's thought. Used as a
deliberation step --- the model is asked to enumerate hypotheses or candidate
fixes before committing to a multi-step action.

\begin{toolconfigbox}{\texttt{think}}
\begin{tooldescriptionsection}
\begin{lstlisting}[style=prompt]
Structure your reasoning before committing to a multi-step action. This tool does not read files, run code, or change the repository - it forces a deliberation step so you act on a considered plan rather than the first idea.

Call it in these situations:
1. After localizing a bug, before writing the patch - state the root cause in one sentence, then list 2-4 candidate fixes. For each, note whether it addresses the root cause or just the observed symptom, and pick the minimal change that fixes the cause.
2. After a failed test or reproduction - list 2-4 hypotheses for the failure ranked by likelihood, and identify the cheapest next action that would discriminate between them.
3. Before a refactor or multi-file change - outline the approach, the files touched, and one concrete risk (e.g., a caller you might break).

Skip for trivial actions: one-line fixes, obvious next commands, simple file lookups, or steps where the next action is already determined. Do not use it to narrate what you just did.

Format: 3-5 concise bullets. No prose paragraphs. No restating the problem.
\end{lstlisting}
\end{tooldescriptionsection}

\begin{toolparamssection}
\begin{enumerate}[leftmargin=*, itemsep=2pt, topsep=2pt, parsep=0pt]
  \item \textbf{thought}: Your structured reasoning (3--5 bullets, no prose paragraphs).
\end{enumerate}
\end{toolparamssection}
\end{toolconfigbox}

\clearpage
\subsection{OpenAI-specific tools}
\label{app:tools:openai}
\subsubsection*{\texttt{shell}}

Input field is named \texttt{input} (not \texttt{command}) and exposes a model-controlled \texttt{timeout}. Used for the GPT model-family.

\begin{toolconfigbox}{\texttt{shell}}
\begin{tooldescriptionsection}
\begin{lstlisting}[style=prompt]
Run commands in a bash shell
* When invoking this tool, the contents of the "command" parameter does NOT need to be XML-escaped.
* You don't have access to the internet via this tool.
* State is persistent across command calls and discussions with the user.
* To inspect a particular line range of a file, e.g. lines 10-25, try 'sed -n 10,25p /path/to/the/file'.
* Please avoid commands that may produce a very large amount of output.
* Please run long lived commands in the background, e.g. 'sleep 10 &' or start a server in the background.
\end{lstlisting}
\end{tooldescriptionsection}

\begin{toolparamssection}
\begin{enumerate}[leftmargin=*, itemsep=2pt, topsep=2pt, parsep=0pt]
  \item \textbf{description}: Why I am running this bash command.
  \item \textbf{input}: The shell command to run.
  \item \textbf{timeout}: Seconds; set conservatively to exceed expected runtime: 5--10s instant; 30--60s installs; 120--180s test suites/builds; 300--600s extreme.
\end{enumerate}
\end{toolparamssection}
\end{toolconfigbox}

\subsubsection*{\texttt{batch\_shell}}

Batched variant of \texttt{shell} for Terminal-Bench-2 configs. The
description explicitly tells the model when packing inputs is preferred
(layout discovery, parallel test runs, multi-file reads) and when a single
input is correct (the next command depends on the previous output).

\begin{toolconfigbox}{\texttt{batch\_shell}}
\begin{tooldescriptionsection}
\begin{lstlisting}[style=prompt]
Run one or more bash commands in a single tool call.

When to pack multiple inputs in one call (PREFERRED whenever possible):
* Independent exploration steps whose next command does not depend on the previous output, e.g. layout discovery: `ls -la`, `cat README.md`, `find . -name pyproject.toml`, `git log --oneline -20`.
* Running several unrelated test files or check commands together, e.g. `pytest tests/test_a.py`, `pytest tests/test_b.py`, `ruff check src/`.
* Reading multiple files or inspecting multiple symbols at once, e.g. `sed -n '1,80p' a.py`, `sed -n '1,80p' b.py`, `grep -n foo src/`.

When to use a SINGLE input:
* The next command genuinely depends on the previous output (e.g. you need to see a file before deciding what to edit).
* You are running a single long command or a script.
\end{lstlisting}
\end{tooldescriptionsection}

\begin{toolparamssection}
\begin{enumerate}[leftmargin=*, itemsep=2pt, topsep=2pt, parsep=0pt]
  \item \textbf{inputs}: List of shell command objects, each containing:
    \begin{itemize}[leftmargin=1.5em, itemsep=1pt, topsep=1pt, parsep=0pt]
      \item \textbf{input}: The shell command to run.
      \item \textbf{description}: Why this command is being run.
      \item \textbf{timeout}: Seconds; default 10, max 600.
      \item \textbf{ignore\_errors}: Set false to stop on failure; default true.
    \end{itemize}
\end{enumerate}
\end{toolparamssection}
\end{toolconfigbox}

\clearpage
\subsubsection*{\texttt{apply\_patch}}

A standalone diff-applying tool that wraps OpenAI's \texttt{*** Begin Patch /
*** End Patch} format. Available as a tool call for configs that prefer it
over heredoc patching inside \texttt{shell}.

\begin{toolconfigbox}{\texttt{apply\_patch}}
\begin{tooldescriptionsection}
\begin{lstlisting}[style=prompt]
Apply a patch to one or more files using the apply_patch format. The `patch` value MUST start with `*** Begin Patch` and end with `*** End Patch`. Example `patch` value:
*** Begin Patch
*** Update File: path/to/file.py
@@
- old line
+ new line
*** End Patch
\end{lstlisting}
\end{tooldescriptionsection}

\begin{toolparamssection}
\begin{enumerate}[leftmargin=*, itemsep=2pt, topsep=2pt, parsep=0pt]
  \item \textbf{patch}: \texttt{apply\_patch} payload bounded by \texttt{*** Begin Patch} / \texttt{*** End Patch}.
  \item \textbf{description}: Why I'm making this edit.
\end{enumerate}
\end{toolparamssection}
\end{toolconfigbox}

\subsubsection*{\texttt{v1.shell}}

A second OpenAI shell variant that takes a \texttt{cmd} argv array rather
than a single \texttt{input} string, matching the native shape that
\texttt{gpt-oss-120b} and Codex models emit. The harness collapses
\texttt{["bash", "-lc", "<body>"]} back into the body string before
execution.

\begin{toolconfigbox}{\texttt{v1.shell}}
\begin{tooldescriptionsection}
\begin{lstlisting}[style=prompt]
Run commands in a bash shell.
* The "cmd" field is an argv array. The conventional invocation is ["bash", "-lc", "<your command>"].
* When invoking this tool, the contents of the command do NOT need to be XML-escaped.
* You don't have access to the internet via this tool.
* State is persistent across command calls.
* To inspect a particular line range of a file, e.g. lines 10-25, try 'sed -n 10,25p /path/to/the/file'.
* Please avoid commands that may produce a very large amount of output.
* Please run long lived commands in the background, e.g. 'sleep 10 &' or start a server in the background.
\end{lstlisting}
\end{tooldescriptionsection}

\begin{toolparamssection}
\begin{enumerate}[leftmargin=*, itemsep=2pt, topsep=2pt, parsep=0pt]
  \item \textbf{cmd}: Argv array; typically \texttt{["bash", "-lc", "<cmd>"]}.
\end{enumerate}
\end{toolparamssection}
\end{toolconfigbox}

\clearpage
\subsection{xAI-specific tools}
\label{app:tools:xai}
\subsubsection*{\texttt{file\_read}}

A read-only file viewer split out of \texttt{str\_replace\_editor}. Grok-4.20
emits a variety of line-range parameter names (\texttt{lineStart},
\texttt{startLine}, \texttt{from\_line}, etc.); the implementation contains a
\texttt{\_infer\_line\_range} fallback that scans \texttt{tool\_input} for any
key matching a line-endpoint pattern --- an example of harness-side
tolerance for model preferrence.

\begin{toolconfigbox}{\texttt{file\_read}}
\begin{tooldescriptionsection}
\begin{lstlisting}[style=prompt]
Tool for viewing files and directories.
* If `path` is a file, `view` displays the result of applying `cat -n`.
  If `path` is a directory, `view` lists non-hidden files and directories up to 2 levels deep.
* Use `view_range` to display specific line ranges of a file, e.g. [11, 20] shows lines 11-20.
* If output is too long, it will be truncated and marked with `<response clipped>`.
\end{lstlisting}
\end{tooldescriptionsection}

\begin{toolparamssection}
\begin{enumerate}[leftmargin=*, itemsep=2pt, topsep=2pt, parsep=0pt]
  \item \textbf{path}: Absolute path to file or directory.
  \item \textbf{view\_range}: \texttt{[start, end]} line range; use \texttt{[start, -1]} for ``from start to EOF.''
\end{enumerate}
\end{toolparamssection}
\end{toolconfigbox}

\subsubsection*{\texttt{file\_edit}}

The xAI counterpart of \texttt{str\_replace\_editor}'s editing path. Two
commands (\texttt{str\_replace}, \texttt{undo\_edit}); creation is split out
into \texttt{xai.file\_write} so the model has to commit to one mode of
modification per call.

\begin{toolconfigbox}{\texttt{file\_edit}}
\begin{tooldescriptionsection}
\begin{lstlisting}[style=prompt]
Tool for editing existing files.
* The `str_replace` command replaces an exact string in a file with a new string.
* The `undo_edit` command reverts the last edit made to the file.

Notes for using the `str_replace` command:
* The `old_str` parameter should match EXACTLY one or more consecutive lines from the original file. Be mindful of whitespaces!
* If the `old_str` parameter is not unique in the file, the replacement will not be performed. Make sure to include enough context in `old_str` to make it unique.
* The `new_str` parameter should contain the edited lines that should replace the `old_str`.
* To create new files, use the `file_write` tool instead.
\end{lstlisting}
\end{tooldescriptionsection}

\begin{toolparamssection}
\begin{enumerate}[leftmargin=*, itemsep=2pt, topsep=2pt, parsep=0pt]
  \item \textbf{command}: \texttt{str\_replace} \textbar{} \texttt{undo\_edit}.
  \item \textbf{description}: Why I'm making this edit.
  \item \textbf{path}: Absolute path to file.
  \item \textbf{old\_str}: Required for \texttt{str\_replace}: exact text to replace.
  \item \textbf{new\_str}: Required for \texttt{str\_replace}: replacement text.
\end{enumerate}
\end{toolparamssection}
\end{toolconfigbox}

\clearpage
\subsubsection*{\texttt{file\_write}}

File-creation tool. Fails if the target already exists.

\begin{toolconfigbox}{\texttt{file\_write}}
\begin{tooldescriptionsection}
\begin{lstlisting}[style=prompt]
Tool for creating new files.
* Creates a new file at the given path with the provided content.
* Fails if the file already exists - use `file_edit` to modify existing files.
* The parent directory must already exist.
\end{lstlisting}
\end{tooldescriptionsection}

\begin{toolparamssection}
\begin{enumerate}[leftmargin=*, itemsep=2pt, topsep=2pt, parsep=0pt]
  \item \textbf{path}: Absolute path for the new file.
  \item \textbf{file\_text}: Full content of the file to be created.
  \item \textbf{description}: Why I'm creating this file.
\end{enumerate}
\end{toolparamssection}
\end{toolconfigbox}

\subsubsection*{\texttt{search}}

Ripgrep-backed search tool. The description includes a hard rule
(``ALWAYS use this tool for search tasks. NEVER invoke grep or rg as a Bash
command'') because Grok otherwise tends to issue \texttt{grep} via the bash
tool and loses the structured output.

\begin{toolconfigbox}{\texttt{search}}
\begin{tooldescriptionsection}
\begin{lstlisting}[style=prompt]
A powerful search tool built on ripgrep.
* ALWAYS use this tool for search tasks. NEVER invoke grep or rg as a Bash command.
* `query` supports full regex syntax (ripgrep flavor), e.g. "log.*Error", "function\s+\w+".
* `path` restricts the search to a specific file or directory; defaults to the working directory.
* Returns matching lines with file path and line number, capped at `max_results` entries.
\end{lstlisting}
\end{tooldescriptionsection}

\begin{toolparamssection}
\begin{enumerate}[leftmargin=*, itemsep=2pt, topsep=2pt, parsep=0pt]
  \item \textbf{query}: Regex pattern (ripgrep syntax).
  \item \textbf{path}: File or directory to search in; defaults to cwd.
  \item \textbf{max\_results}: Max matching lines to return.
\end{enumerate}
\end{toolparamssection}
\end{toolconfigbox}

\section{vLLM Serving Configurations}
\label{app:vllm}

The open-weights configurations in Section~\ref{sec:appendix-configs} (Qwen-3.5/3.6/3.8 variants~\citep{yang2025qwen3}, gpt-oss-120b/20b~\citep{openai2025gpt5}) are hosted on a private vLLM~\citep{kwon2023vllm} endpoint stood up on an internal SLURM cluster with Amazon EC2 \texttt{p4.24xlarge} instances, with a LiteLLM~\citep{litellm} proxy. The \texttt{vllm serve} invocations used to bring those endpoints up are listed below.

\subsection{Per-model serve commands}
\label{app:vllm:serve}
\begin{tcolorbox}[vllmbox, title={\texttt{qwen3.5-27b}}]
\begin{lstlisting}[style=prompt]
vllm serve Qwen/Qwen3.5-27B \
  --host 0.0.0.0 --port 8000 \
  --tensor-parallel-size 8 \
  --distributed-executor-backend ray \
  --max-model-len 200000 \
  --gpu-memory-utilization 0.92 \
  --enable-prefix-caching \
  --enable-prompt-tokens-details \
  --reasoning-parser qwen3 \
  --enable-auto-tool-choice \
  --tool-call-parser qwen3_xml \
  --served-model-name qwen3.5-27b \
  --language-model-only
\end{lstlisting}
\end{tcolorbox}

\begin{tcolorbox}[vllmbox, title={\texttt{qwen3.6-35b-a3b} (SWE tasks)}]
\begin{lstlisting}[style=prompt]
vllm serve Qwen/Qwen3.6-35B-A3B \
  --host 0.0.0.0 --port 8100 \
  --tensor-parallel-size 8 \
  --distributed-executor-backend ray \
  --max-model-len 200000 \
  --gpu-memory-utilization 0.92 \
  --enable-prefix-caching \
  --enable-prompt-tokens-details \
  --reasoning-parser qwen3 \
  --enable-auto-tool-choice \
  --tool-call-parser qwen3_xml \
  --served-model-name qwen3.6-35b-a3b \
  --language-model-only
\end{lstlisting}
\end{tcolorbox}

\begin{tcolorbox}[vllmbox, title={\texttt{qwen3.6-35b-a3b} (Terminal-Bench-2)}]
\begin{lstlisting}[style=prompt]
vllm serve Qwen/Qwen3.6-35B-A3B \
  --host 0.0.0.0 --port 8100 \
  --tensor-parallel-size 8 \
  --distributed-executor-backend ray \
  --max-model-len 262144 \
  --gpu-memory-utilization 0.92 \
  --enable-prefix-caching \
  --enable-prompt-tokens-details \
  --reasoning-parser qwen3 \
  --enable-auto-tool-choice \
  --tool-call-parser qwen3_xml \
  --served-model-name qwen3.6-35b-a3b \
  --language-model-only
\end{lstlisting}
\end{tcolorbox}

\begin{tcolorbox}[vllmbox, title={\texttt{qwen3.6-27b} (SWE tasks)}]
\begin{lstlisting}[style=prompt]
vllm serve Qwen/Qwen3.6-27B \
  --host 0.0.0.0 --port 8200 \
  --tensor-parallel-size 8 \
  --distributed-executor-backend ray \
  --max-model-len 200000 \
  --gpu-memory-utilization 0.92 \
  --enable-prefix-caching \
  --enable-prompt-tokens-details \
  --reasoning-parser qwen3 \
  --enable-auto-tool-choice \
  --tool-call-parser qwen3_xml \
  --served-model-name qwen3.6-27b \
  --language-model-only
\end{lstlisting}
\end{tcolorbox}

\begin{tcolorbox}[vllmbox, title={\texttt{qwen3.6-27b} (Terminal-Bench-2)}]
\begin{lstlisting}[style=prompt]
vllm serve Qwen/Qwen3.6-27B \
  --host 0.0.0.0 --port 8200 \
  --tensor-parallel-size 8 \
  --distributed-executor-backend ray \
  --max-model-len 262144 \
  --gpu-memory-utilization 0.92 \
  --enable-prefix-caching \
  --enable-prompt-tokens-details \
  --reasoning-parser qwen3 \
  --enable-auto-tool-choice \
  --tool-call-parser qwen3_xml \
  --served-model-name qwen3.6-27b \
  --language-model-only
\end{lstlisting}
\end{tcolorbox}

\begin{tcolorbox}[vllmbox, title={\texttt{qwen3.8-27b} (Deep-SWE)}]
\begin{lstlisting}[style=prompt]
vllm serve Qwen/Qwen3.8-27B \
  --host 0.0.0.0 --port 8000 \
  --tensor-parallel-size 8 \
  --max-model-len 262144 \
  --gpu-memory-utilization 0.90 \
  --enable-prefix-caching \
  --enable-prompt-tokens-details \
  --reasoning-parser qwen3 \
  --enable-auto-tool-choice \
  --tool-call-parser qwen3_xml \
  --served-model-name qwen3.8-27b \
  --language-model-only
\end{lstlisting}
\end{tcolorbox}

Served with vLLM 0.28.0 using its default multiprocess executor (no Ray), one TP=8 replica per node.

\begin{tcolorbox}[vllmbox, title={\texttt{gpt-oss-120b}}]
\begin{lstlisting}[style=prompt]
vllm serve openai/gpt-oss-120b \
  --host 0.0.0.0 --port 8300 \
  --tensor-parallel-size 8 \
  --max-model-len 131072 \
  --gpu-memory-utilization 0.90 \
  --enable-prefix-caching \
  --enable-prompt-tokens-details \
  --reasoning-parser openai_gptoss \
  --enable-auto-tool-choice \
  --tool-call-parser openai \
  --served-model-name gpt-oss-120b
\end{lstlisting}
\end{tcolorbox}

MXFP4 datatype is auto-detected from the model config.

\begin{tcolorbox}[vllmbox, title={\texttt{gpt-oss-20b}}]
\begin{lstlisting}[style=prompt]
vllm serve openai/gpt-oss-20b \
  --host 0.0.0.0 --port 8400 \
  --tensor-parallel-size 2 \
  --max-model-len 131072 \
  --gpu-memory-utilization 0.90 \
  --enable-prefix-caching \
  --enable-prompt-tokens-details \
  --reasoning-parser openai_gptoss \
  --enable-auto-tool-choice \
  --tool-call-parser openai \
  --served-model-name gpt-oss-20b
\end{lstlisting}
\end{tcolorbox}

\subsection{A note on vLLM flags}
\label{app:vllm:choices}
\begin{itemize}[leftmargin=*, itemsep=2pt, topsep=2pt]
  \item \textbf{Compile mode is on for every model.} \texttt{ENFORCE\_EAGER=0} in every sbatch means no \texttt{--enforce-eager} and no dynamo disable: $\sim$2--3$\times$ decode speedup over eager.
  \item \textbf{No \texttt{--quantization} flag anywhere.} BF16 models load BF16 weights; gpt-oss auto-detects \texttt{mxfp4} from its checkpoint's \texttt{config.json}.
\end{itemize}

\clearpage
\section{Harmony-format parsing failures in the harness}
\label{app:gptoss}

This section describes the initial gap  we observed for the open-weights
gpt-oss evaluations (\texttt{gpt-oss-120b} and \texttt{gpt-oss-20b}) in the  officially reported pass@1. For \texttt{gpt-oss-120b}, on SWE-Bench-Verified, we observed 42\% pass@1 while it is reported at 62\% in the release card. The root-cause was a failure mode observed in the open-weights
\texttt{gpt-oss} evaluations when sampling was done at high temperatures, such as $1.0$. The affected outputs had a mix of non-existent tool-calls and empty tool-use payloads. After fixing the issue, the pass@1 of SSA increased to 64.64\% as reported in Appendix~\ref{app:passrates}.

This is an interface failure with disproportionate trajectory-level consequences.
In an agent loop, one corrupted tool turn is fed back as an environmental
observation. The model then reasons from a false account of its own action and spends subsequent turns repairing a call that was never malformed. The remainder of this appendix documents the symptom and the cache diagnostic that
exposed the affected batch (Section~\ref{app:gptoss:symptom}), the stateless
re-tokenization boundary that makes the corruption persistent once it reaches the
transcript (Section~\ref{app:gptoss:boundary}), token-level evidence
(Section~\ref{app:gptoss:observed}), and the retry-based mitigation
(Section~\ref{app:gptoss:patches}).

\subsection{The symptom: empty and malformed tool calls}
\label{app:gptoss:symptom}

We identified the issue after noticing an abnormal cache hit-rate of zero for \texttt{gpt-oss} experiments. In a healthy multi-turn agent trajectory the cache hit rate should rise towards 1.0 as each request resubmits a long prefix that the previous request has already processed and cached. Intermittently, we observed that the input cache was dropped in the middle of the trajectory (cache hit-rate of zero) followed by a cache-write.
It indicated that prefix matching before cache-read failed and hence a mismatch existed between agent-sent input context vs what existed in the server. Since, agent context and server cached context differs only at the last assistant message, these events provided a first-level diagnostic of tampered model generated output sequence.

In the logs, we found that \texttt{gpt-oss} turns intermittently decode into a tool-call whose argument body is empty, or into a
tool-call whose name contains literal Harmony control-token text. Both cases are problematic because they are presented to the agent as an ordinary tool outcomes.
The model is then shown a validation error for an action it did not actually emit in that form, and its next turn is pulled away from the original task.

The failure location is the Harmony-to-OpenAI (chat-completion format) response boundary. \texttt{gpt-oss} encodes
assistant output in the \emph{Harmony} format~\citep{openai2025harmony}, which
multiplexes internal reasoning, tool-use communication, and final answers onto one
token stream. The \texttt{analysis} channel carries the model's reasoning, the
\texttt{commentary} channel carries tool calls and tool-use planning, and the
\texttt{final} channel carries the user-visible response. Reserved control tokens
delimit channels and message bodies (Table~\ref{tab:harmony-tokens}). A serving
stack such as vLLM~\citep{kwon2023vllm} must demultiplex this stream and re-emit a
standard OpenAI-compatible response (for example, a chat-completion api). Reasoning stripped or separated, ordinary text for final answers, and a structured \texttt{\{name, arguments\}} object for
each parsed tool call. The observed corruption enters at this demultiplexing
stage, where an ill-formed Harmony boundary can still be returned to the client as
a nominally valid tool call.

\begin{table}[h]
\centering
\footnotesize
\caption{Harmony control tokens that delimit channels and messages. The two
assistant stop tokens, \texttt{<{|}return{|}>} and \texttt{<{|}call{|}>}, both halt
generation but signal different terminal states; the distinction is used by the
retry check in Section~\ref{app:gptoss:patches}.}\label{tab:harmony-tokens}
\begin{tabular}{@{}l l@{}}
\toprule
Token & Role in the stream \\
\midrule
\texttt{<{|}start{|}>}     & Begin a new message; followed by the role (and optional \texttt{to=functions.<name>} recipient) \\
\texttt{<{|}channel{|}>}   & Followed by the channel name (\texttt{analysis} / \texttt{commentary} / \texttt{final}) \\
\texttt{<{|}constrain{|}>} & Begin a constrained span (e.g.\ declaring the tool-argument body is \texttt{json}) \\
\texttt{<{|}message{|}>}   & Begin the message body (free-form text or the tool-argument JSON) \\
\texttt{<{|}end{|}>}       & End of a non-terminal message (more messages follow in the turn) \\
\texttt{<{|}return{|}>}    & End of the assistant turn with a final answer (normal end-of-stream) \\
\texttt{<{|}call{|}>}      & End of an assistant turn that emitted a tool call \\
\bottomrule
\end{tabular}
\end{table}

\subsection{The serving boundary: stateless re-tokenization}
\label{app:gptoss:boundary}

The main issue is the misplacement of Harmony control tokens in Table\,\ref{tab:harmony-tokens} in the parsed JSON output. Due to high sampling temperature of 1.0, the model has a higher chance of generating delimiter tokens (such as \texttt{<{|}channel{|}>}, \texttt{<{|}call{|}>}) at improper locations. This ill-generated token-sequence passes through Chat Completions API which converts it to structured JSON. See Figure\,\ref{fig:harmony-confuse-name-tokens}, where one incorrect generated token confused the system to parse the string into a tool-name that didn't exist.

Once a corrupted turn is admitted into the transcript, the serving path has no
mechanism for restoring the original token sequence. The Chat Completions API is
stateless at the token-level. At every turn, the harness resubmits the conversation as a structured JSON and the server reconstructs the prompt rather than reusing the token-ids generated in the previous turn. In the \texttt{gpt-oss} path, vLLM rebuilds Harmony \texttt{Message} objects from fields such as \texttt{content}, \texttt{reasoning}, and each \texttt{tool\_call}'s \texttt{name} and
\texttt{arguments}. It then re-encodes those reconstructed messages for the next
prefill. Each assistant turn is therefore decoded, parsed into JSON, returned to
the client, and re-tokenized from JSON when the conversation continues.

This decode--re-encode operation is not an identity map, which is the token
fidelity issue discussed by~\citet{gallouedec2026tito}. Tokenization is not
generally invertible as re-encoding decoded text need not recover the original token-ids. Ordinary text usually survives this round trip, which is why prompt caching
remains effective, but the reconstructed assistant turn is not byte-identical to
the model's original output. Prior \texttt{analysis} content is intentionally
dropped during reconstruction, and tool-call bodies are re-rendered as ordinary
\texttt{json} rather than as the original
\texttt{<{|}constrain{|}>json} span. These deterministic differences affect only
the final assistant segment of the prompt, while the longer shared prefix still
re-tokenizes identically and can be served from cache.

Note, a token-in/token-out (TITO) interface would avoid this failure by feeding the
original generated token-ids back into the model, rather than reconstructing them
from text. The JSON-based Chat Completions path cannot do that. It must rebuild
the prompt from structured fields, and therefore cannot distinguish a reserved
Harmony control token from a literal string that happens to spell the same text.
For example, once \texttt{<{|}channel{|}>} has leaked into decoded text, it
re-encodes as ordinary text tokens, not as the single reserved control-token id.
Consequently, a malformed Harmony turn that reaches the transcript remains
malformed on subsequent requests.

\subsection{Token-level evidence}
\label{app:gptoss:observed}

In healthy turns, Harmony control tokens separate the
reasoning message, the tool-call header, and the argument body. In failing turns,
those boundaries collapse: an \texttt{<{|}end{|}>} that should close the analysis
message, or a fresh
\texttt{<{|}start{|}>}\dots\texttt{<{|}channel{|}>}\dots\texttt{<{|}constrain{|}>}
header that should open the tool call, is missing or fused with adjacent text. As
a result, channel names, recipients, and message bodies are no longer separated in
the form expected by the parser. The figures below show the relevant decoded
\texttt{gpt-oss-120b} streams, together with the client-visible error and, the clean retry that follows.

When no clean argument span can be recovered, this boundary collapse yields the
empty-argument tool call described in Section~\ref{app:gptoss:symptom}. When a
control token is instead incorporated into the recipient, the same family of
failures surfaces as a \textit{corrupt tool name}. One observed server response named the
tool \texttt{shell<{|}channel{|}>commentary}, and SSA rejected it because no registered
tool has that name.

\begin{tcolorbox}[vllmbox, title={A control token folded into the tool name, rejected at the harness}]
\begin{lstlisting}[style=prompt]
Intended tool_name=shell
Decoded tool_name=<shell<|channel|>commentary>
SSA replies -> `invalid tool name'
\end{lstlisting}
\end{tcolorbox}

\paragraph{A corrupt tool name changes the trajectory.}
In Figure~\ref{fig:harmony-confuse-name}, the model attempts to read a specific
function, but a stray \texttt{<{|}channel{|}>} becomes part of the tool recipient.
The harness rejects the resulting name with \texttt{invalid tool name pattern}. The
model cannot observe the injected token, interprets the failure as its own usage
error, retries unsuccessfully, and then abandons the targeted read in favor of a
search. Thus the corrupted parse does not merely consume a turn, it removes an
intended observation from the trajectory. Figure~\ref{fig:harmony-confuse-name-tokens}
shows the corresponding raw streams.

\begin{figure}[ht]
\centering
\begin{minipage}{\linewidth}
\begin{turnbox}{TURN $k$ \textnormal{\textmd{--- a stray token corrupts the tool name}}}
\begin{intentsection}
The model wants to read a specific function: \emph{``Inspect \texttt{sympy/core/add.py} for
its \texttt{\_eval\_power}.''}
\end{intentsection}
\begin{executionsection}
It issues \texttt{file\_read} tool call, but a stray \texttt{<{|}channel{|}>} folds into the tool-name header, so the name parses as
\texttt{file\_read<{|}channel{|}>commentary}. 

The harness answers
\textcolor{execplum!85!black}{\texttt{tool\_name=<file\_read<{|}channel{|}>commentary> | invalid tool name pattern}}.
\end{executionsection}
\end{turnbox}

\begin{turnbox}{TURN $k{+}1$ \textnormal{\textmd{--- attributes the failure to its own usage}}}
\begin{intentsection}
Reading the rejection as its own mistake, model thinks: \emph{``Actually the tool name is
\texttt{file\_read}. Use that.''}
\end{intentsection}
\begin{executionsection}
It re-issues the same \texttt{file\_read}; and folds in the stray token, so the
same \textcolor{execplum!85!black}{\texttt{invalid tool name pattern}} returns. The model never gets to see the problematic token in the tool error message, so the retry does not help.
\end{executionsection}
\end{turnbox}

\begin{turnbox}{TURN $k{+}2$ \textnormal{\textmd{--- the targeted read is abandoned}}}
\begin{intentsection}
Unable to open the file, the model drops the inspection and switches to a different \texttt{search} action.
\end{intentsection}
\begin{executionsection}
\texttt{search(\ldots query="\_eval\_power"\ldots)} runs, but the targeted view of
\texttt{add.py} lines $340$--$380$ is \emph{dropped here}: the model proceeds on what it can
infer for now, having lost the read it set out to make.~\textcolor{execplum!85!black}{$\times$}
\end{executionsection}
\end{turnbox}
\end{minipage}
\caption{\textbf{A corrupted tool-name causes the model to abandon an intended read}
(\texttt{gpt-oss-120b}, \texttt{sympy-13974} (SWE-Bench-Verified). The model sets out to read \texttt{add.py} around
\texttt{\_eval\_power}. A stray \texttt{<{|}channel{|}>} is folded into the
\texttt{file\_read} recipient, producing
\texttt{file\_read<{|}channel{|}>commentary}. The harness rejects the call with
\texttt{invalid tool name pattern}. Because the injected token is invisible to the
model, it concludes that it used the tool incorrectly. After the retry fails, the
model pivots to \texttt{search} (Turn $k{+}2$), so the intended inspection of those
\texttt{add.py} lines is skipped at that point. Complete streams are shown in
Figure~\ref{fig:harmony-confuse-name-tokens}.}
\label{fig:harmony-confuse-name}
\end{figure}

\begin{figure}[ht]
\centering
\providecommand{\ct}[1]{\colorbox{black!8}{\texttt{\scriptsize #1}}}%
\providecommand{\ctblue}[1]{\colorbox{blue!55}{\texttt{\scriptsize\textcolor{white}{#1}}}}%
\setlength{\fboxsep}{1.3pt}%
\begin{tcolorbox}[vllmbox, title={Token-by-token streams}]
{\footnotesize\textbf{Turn $k$} --- the corrupted \texttt{file\_read}: a stray
\texttt{<{|}channel{|}>} is inserted after the recipient.}\par\smallskip
{\scriptsize\sloppy\raggedright\spaceskip=3pt plus 2pt minus 1pt
\ct{<{|}channel{|}>} analysis \ct{<{|}message{|}>} Inspect sympy\allowbreak/core\allowbreak/add.py for its \_eval\_power.
\ct{<{|}end{|}>}\ct{<{|}start{|}>} assistant \ct{<{|}channel{|}>} commentary to=functions.\allowbreak file\_read
\ctblue{<{|}channel{|}>} commentary code \ct{<{|}message{|}>}
\texttt{\{"path": "/testbed\allowbreak/sympy\allowbreak/core\allowbreak/add.py", "view\_range": [340, 380]\}}
\ct{<{|}call{|}>}\par}
\smallskip
{\footnotesize $\Rightarrow$ the recipient and the
\colorbox{blue!55}{\textcolor{white}{\texttt{\scriptsize <{|}channel{|}>}}} after it parse
together as the tool name \texttt{file\_read<{|}channel{|}>commentary}, so the harness returns
\textcolor{execplum!85!black}{\texttt{invalid tool name pattern}}.}
\smallskip\hrule\smallskip
{\footnotesize\textbf{Turn $k{+}1$} --- it attributes the failure to its own usage and retries, but the recipient now sits
on the \texttt{analysis} channel (a tool call must be on \texttt{commentary}):}\par\smallskip
{\scriptsize\sloppy\raggedright\spaceskip=3pt plus 2pt minus 1pt
\ct{<{|}channel{|}>} analysis \ct{<{|}message{|}>} Actually the tool name is file\_read. Use that.
\ct{<{|}end{|}>}\ct{<{|}start{|}>} assistant \ct{<{|}channel{|}>} analysis to=functions.\allowbreak file\_read
\ct{<{|}constrain{|}>} json \ct{<{|}message{|}>}
\texttt{\{"path": "/testbed\allowbreak/sympy\allowbreak/core\allowbreak/add.py", "view\_range": [340, 380]\}}
\ct{<{|}call{|}>}\par}
\smallskip
{\footnotesize $\Rightarrow$ on the wrong channel the call is not dispatched, so the
\texttt{file\_read} still does not land.}
\smallskip\hrule\smallskip
{\footnotesize\textbf{Turn $k{+}2$} --- the targeted read is abandoned and the model pivots to a \texttt{search}
(no stray token here):}\par\smallskip
{\scriptsize\sloppy\raggedright\spaceskip=3pt plus 2pt minus 1pt
\ct{<{|}channel{|}>} analysis \ct{<{|}message{|}>} Thus there is independent handling for Pow for base with power.
\ct{<{|}end{|}>}\ct{<{|}start{|}>} assistant \ct{<{|}channel{|}>} analysis to=functions.\allowbreak search code \ct{<{|}message{|}>}
\texttt{\{"path": "/testbed\allowbreak/sympy", "query": "if hasattr(base, \textbackslash"\allowbreak\_eval\_power\textbackslash")", "max\_results": 20\}}
\ct{<{|}call{|}>}\par}
\smallskip
{\footnotesize $\Rightarrow$ \texttt{search} runs; the inspection of \texttt{add.py} lines
$340$--$380$ is dropped here --- the corrupted parse redirected the trajectory.}
\end{tcolorbox}
\caption{\textbf{Token-level view of the corrupt-name example in
Figure~\ref{fig:harmony-confuse-name}.} Each reserved Harmony control token is a
grey chip. In Turn $k$ the single defect is the
\colorbox{blue!55}{\textcolor{white}{\texttt{\scriptsize <{|}channel{|}>}}} the server inserts
after \texttt{to=functions.file\_read}, where \texttt{code<{|}message{|}>} should follow. The
parser reads the recipient and that stray token together as the tool name
\texttt{file\_read<{|}channel{|}>commentary}, so the call is rejected. The model cannot see the
token, so it attributes the failure to its own usage, its retry (Turn $k{+}1$) puts the recipient on the
\texttt{analysis} channel and is silently not dispatched, so it pivots to a \texttt{search}
(Turn $k{+}2$). The read of \texttt{add.py} it set out to make is dropped at this point.}
\label{fig:harmony-confuse-name-tokens}
\end{figure}

\FloatBarrier

\subsection{The fix: surface a single retryable error}
\label{app:gptoss:patches}

The mitigation is to make a corrupted sample retryable before it becomes part of
the model-visible transcript. The failures arises
from rare token sequences at temperature $1.0$, and resampling the same prompt
typically yields a well-formed turn. The OpenAI-compatible server is therefore
hardened so that all observed corruption modes converge to a single explicit
signal, \texttt{HarmonyParseError} over HTTP~503, which SSA treats as retryable with backoff.

\paragraph{Evaluation after the fix}
Re-serving the identical agent configuration through a corrected vLLM build drops
these parse errors to \emph{zero}, localizing the failure to Harmony handling in the
model server rather than to the model weights or SSA's tool registry. The retry hook described above recovers all Harmony parsing errors increasing `pass@1' of \texttt{gpt-oss-120b} from $42\%$ to $64.64\%$ (reported model card number is $62\%$).

\end{document}